\documentclass[11pt,reqno,twoside]{book}

\usepackage{cmap} 

\usepackage[T1]{fontenc}
\usepackage[utf8]{inputenc}
\usepackage{graphicx}
\usepackage{placeins}

\usepackage{verbatim}

\usepackage[utf8]{inputenc}
\usepackage{imakeidx}
\makeindex[columns=2, title=Alphabetical Index,
           options= -s example\_style.ist]
\usepackage[nottoc]{tocbibind}

\usepackage{setspace}

\let\counterwithin\relax  
\usepackage{lmodern} 
\usepackage[scale=0.88]{tgheros} 

\usepackage{bm} 

\usepackage{bbold}

\usepackage{amsmath,amsbsy,amsgen,amscd,amsthm,amsfonts,amssymb,thmtools} 

\usepackage[centering,top=1.5in,bottom=1.2in,left=1.4in,right=1.4in]{geometry}

\usepackage{titling}
\usepackage{musicography}
\usepackage[sf,bf,compact]{titlesec}

\usepackage{booktabs,longtable,tabu} 
\usepackage[font=small,margin=12pt,labelfont={sf,bf},labelsep={space}]{caption}

\usepackage{enumitem}
\setitemize{itemsep=0pt} 
\setenumerate{itemsep=0pt}
\setlist[1]{labelindent=\parindent,
 }

\usepackage{tikz}

\renewcommand{\nabla}{D}
\newcommand{\Kt}{C_{\theta,j+1}}
\newcommand{\qf}{\mathfrak{q}}

\newcommand{\fgf}{g}

\newcommand{\vd}{v^{\dagger}}

\newcommand{\dw}{w^{\dagger}}
\newcommand{\yd}{y^{\dagger}}
\newcommand{\xid}{\xi^{\dagger}}
\newcommand{\etad}{\eta^{\dagger}}

\newcommand{\Yd}{Y^{\dagger}}
\newcommand{\Vd}{V^{\dagger}}
\newcommand{\Vs}{V^{\star}}
\newcommand{\Vm}{\Vmap}

\newcommand{\Gammas}{\mathsf{\Gamma}}
\newcommand{\Sigmas}{\mathsf{\Sigma}}

\newcommand{\Acs}{\mathsf{A}}
\newcommand{\ks}{\mathsf{k}}
\newcommand{\CRPS}{\mathsf{CRPS}}
\newcommand{\ES}{\mathsf{ES}}
\newcommand{\QS}{\mathsf{QS}}
\newcommand{\LS}{\mathsf{LS}}
\newcommand{\KS}{\mathsf{KS}}
\newcommand{\Id}{\mathrm{Id}}
\newcommand{\mS}{\mathsf{S}}
\newcommand{\hs}{\mathsf{h}}
\newcommand{\hC}{\hat{C}}
\newcommand{\hv}{\widehat{v}}
\newcommand{\hy}{\widehat{y}}

\newcommand{\normal}{\mathcal{N}}

\newcommand{\piap}{\pi^{\mathrm{ralg}}}
\newcommand{\pia}{\pi^{\mathrm{alg}}}
\newcommand{\piw}{\pi^{\mathrm{walg}}}
\newcommand{\piaw}{\widehat{\pi}^{\mathrm{alg}}}
\newcommand{\pie}{\pi^{\mathrm{EnKF}}}
\newcommand{\pieh}{\widehat{\pi}^{\mathrm{EnKF}}}

\newcommand{\pip}{\pi^{\mathrm{BPF}}}
\newcommand{\pop}{\pi^{\mathrm{OPF}}}

\newcommand{\dr}{\mathsf{d_{rand}}}

\newcommand{\hw}{\widehat{w}}
\newcommand{\hz}{\widehat{z}}

\definecolor{dark-gray}{gray}{0.3}
\definecolor{dkgray}{rgb}{.4,.4,.4}
\definecolor{dkblue}{rgb}{0,0,.5}
\definecolor{dkgreen}{rgb}{0,0.5,.0}
\definecolor{rust}{rgb}{0.5,0.1,0.1}

\usepackage{url}
\usepackage[colorlinks=true]{hyperref}
\hypersetup{linkcolor=dkblue}    
\hypersetup{citecolor=rust}      
\hypersetup{urlcolor=rust}     

\usepackage{doi}

\usepackage[final]{microtype} 

\newtheoremstyle{myThm} 
    {\topsep}                    
    {\topsep}                    
    {\itshape}                   
    {}                           
    {\sffamily\bfseries}                   
    {.}                          
    {.5em}                       
    {}  

\newtheoremstyle{myRem} 
    {\topsep}                    
    {\topsep}                    
    {}                   
    {}                           
    {\sffamily}                   
    {.}                          
    {.5em}                       
    {}  

\newtheoremstyle{myDef} 
    {\topsep}                    
    {\topsep}                    
    {}                   
    {}                           
    {\sffamily\bfseries}                   
    {.}                          
    {.5em}                       
    {}  

\theoremstyle{myThm}
\newtheorem{theorem}{Theorem}[chapter]
\newtheorem{lemma}[theorem]{Lemma}
\newtheorem{proposition}[theorem]{Proposition}
\newtheorem{corollary}[theorem]{Corollary}

\newtheorem{assumption}[theorem]{Assumption}
\newtheorem{dataassumption}[theorem]{Data Assumption}

\theoremstyle{myRem}
\newtheorem{remarkx}[theorem]{Remark}
 \newenvironment{remark}
  {\pushQED{\qed}\remarkx}
  {\popQED\endremarkx}

\theoremstyle{myDef}
\newtheorem{definitionx}[theorem]{Definition}
\newenvironment{definition}
  {\pushQED{\qed}\definitionx}
  {\popQED\enddefinitionx}

 \newtheorem{examplex}[theorem]{Example}
 \newenvironment{example}
  {\pushQED{\qed}\examplex}
  {\popQED\endexamplex}

\usepackage{fancyhdr}
\usepackage{nopageno} 
\let\originalleft\left
\let\originalright\right
\renewcommand{\left}{\mathopen{}\mathclose\bgroup\originalleft}
\renewcommand{\right}{\aftergroup\egroup\originalright}

\usepackage{mathtools}
\mathtoolsset{centercolon}  

\renewcommand{\phi}{\varphi}
\renewcommand{\div}{\mathrm{div}\,}

\newcommand{\thetas}{\theta^{\star}}
\newcommand{\betas}{\beta^{\star}}
\newcommand{\alphas}{\alpha^{\star}}
\newcommand{\psis}{\psi^{\star}}
\renewcommand{\ss}{s^{\star}}
\newcommand{\ps}{p^{\star}}
\newcommand{\qs}{q^{\star}}

\newcommand{\ms}{m^{\star}}
\newcommand{\Sigmast}{\Sigma^{\star}}
\newcommand{\gs}{g^{\star}}
\newcommand{\fs}{f^{\star}}

\newcommand{\pist}{\pi^{\star}}

\newcommand{\zerovct}{\vct{0}} 

\providecommand{\mathbbm}{\mathbb} 

\newcommand{\R}{\mathbbm{R}}

\newcommand{\N}{\mathbbm{N}}
\newcommand{\Z}{\mathbbm{Z}}
\newcommand{\la}{\langle}
\newcommand{\ra}{\rangle}

\definecolor{mygreen}{rgb}{0.1,0.75,0.2}

\newcommand{\dz}{d_z}
\newcommand{\ttheta}{\vartheta}
\newcommand{\mn}{\Upsilon^{N}}
\newcommand{\rn}{\rho^{N}}
\newcommand{\un}{u^{(n)}}
\newcommand{\UN}{U}

\newcommand{\xr}{u^{(r)}}

\newcommand{\psid}{\psi^{\dagger}}

\newcommand{\co}{\mathsf{c}}
\newcommand{\dm}{\mathsf{d}}

\renewcommand{\hm}{H_{\Upsilon}}
\newcommand{\hmn}{H_{\Upsilon^{\mbox {\tiny{{\rm N}}}}}}
\newcommand{\pr}{\rho}

\newcommand{\post}{\pi}
\newcommand{\intp}{\mathfrak{p}}
\newcommand{\intpw}{\mathfrak{p}^{\mathrm{walg}}}
\newcommand{\postg}{\pi_g}
\newcommand{\postr}{\rho}
\newcommand{\postv}{\varrho}
\newcommand{\Fr}{F_{\postr}}
\newcommand{\Fv}{F_{\postv}}

\newcommand{\noise}{\nu}
\newcommand{\loss}{\mathsf{L}}

\newcommand{\losss}{\ell}
\newcommand{\like}{\mathsf{l}}
\newcommand{\ELBO}{\mathsf{ELBO}}
\newcommand{\reg}{\mathsf{R}}

\newcommand{\V}{V}

\newcommand{\qopt}{q_{\text{\sc{opt}}}}

\newcommand{\refT}{\varrho}
\newcommand{\mkappa}{\mathbb{k}}

\newcommand{\pMC}{\post^N_{\mbox {\tiny{\rm MC}}}}

\newcommand{\Cpn}{C}

\newcommand{\Prob}{\operatorname{\mathbbm{P}}}
\newcommand{\Expect}{\operatorname{\mathbb{E}}}
\newcommand{\Sam}{N}
\newcommand{\sam}{n}

\newcommand{\du}{d}
\newcommand{\dy}{k}
\newcommand{\Ru}{\mathbbm{R}^\du}
\newcommand{\Ry}{\mathbbm{R}^\dy}
\newcommand{\Ruu}{\mathbbm{R}^{\du \times \du}}
\newcommand{\Ryu}{\mathbbm{R}^{\dy \times \du}}

\newcommand{\Gp}{G_{\mathrm{p}}}
\newcommand{\Gs}{G_{\mathrm{c}}}

\newcommand{\vct}[1]{{#1}}
\newcommand{\mtx}[1]{{#1}}

\newcommand{\umap}{u_{\mbox {\tiny{\rm MAP}}}}
\newcommand{\upm}{u_{\mbox {\tiny{\rm PM}}}}
\newcommand{\Vmap}{V_{\mbox {\tiny{\rm MAP}}}}
\newcommand{\vmap}{v_{\mbox {\tiny{\rm MAP}}}}

\newcommand{\vzmap}{v_{0,\mbox {\tiny{\rm MAP}}}}
\newcommand{\Zmap}{Z_{\mbox {\tiny{\rm MAP}}}}

\newcommand{\pmh}{p_{\mbox {\tiny{\rm MH}}}}

\newcommand{\Tr}{\mathrm{Tr}}

\newcommand{\cA}{\mathcal{A}}
\newcommand{\cB}{\mathcal{B}}

\newcommand{\grad}{\nabla}

\newcommand{\lambdar}{\lambda_{\mathsf{reg}}}

\newcommand{\cN}{\mathcal{N}}

\newcommand{\bbN}{\mathbb{N}}
\newcommand{\bbR}{\mathbb{R}}
\newcommand{\bbZ}{\mathbb{Z}}
\newcommand{\cG}{\mathcal{G}}
\newcommand{\bbP}{\mathbb{P}}
\newcommand{\cP}{\mathcal{P}}
\newcommand{\cQ}{\mathcal{Q}}

\newcommand{\cQt}{\mathcal{Q}_{\Theta}}
\newcommand{\cC}{\mathcal{C}}
\newcommand{\cTt}{\mathcal{T}_{\Theta}}
\newcommand{\cRt}{\mathcal{R}_{\Theta}}

\newcommand{\bbE}{\mathbb{E}}

\newcommand{\ind}{\perp\!\!\!\!\perp}

\newcommand{\dkl}{\mathsf{D}_{\mbox {\tiny{\rm KL}}}}

\newcommand{\dtv}{\mathsf{D}_{\mbox {\tiny{\rm TV}}}}
\newcommand{\dchi}{\mathsf{D}_{\mbox {\tiny{$ \chi^2$}}}}
\newcommand{\dhell}{\mathsf{D}_{\mbox {\tiny{\rm H}}}}
\newcommand{\dmmd}{\mathsf{D}_{\mbox {\tiny{\rm MMD}}}}
\newcommand{\den}{\mathsf{D}_{\mbox {\tiny{\rm E}}}}
\newcommand{\Nc}{\mathcal{N}}
\newcommand{\D}{\mathsf{D}}
\newcommand{\dd}{\mathsf{d}}
\newcommand{\ha}{h_{\alpha}}
\newcommand{\La}{L_{\alpha,\varrho}}
\newcommand{\qv}{q_{\varrho}}
\newcommand{\qr}{q_{\rho}}

\usepackage[]{algorithm}

\usepackage{graphicx, subfig}
\usepackage{algorithmic}
\usepackage{authblk}
\usepackage[square,numbers]{natbib}
\usepackage{chngcntr}
\usepackage{mathrsfs}
\counterwithin{table}{chapter}
\counterwithin{algorithm}{chapter}

\title{{\huge\emph{Machine Learning for Inverse Problems and Data Assimilation}}\\
} 

\author{\vspace{.55in}  Eviatar Bach $^{\musNatural{}}$, Ricardo Baptista $^{\barwedge{}}$, Daniel Sanz-Alonso $^{\musFlat{}},$ Andrew Stuart $^{\musSharp{}}$\\
	\vspace{.6in} 
 $^{\musNatural{}}$ University of Reading\vspace{.05in}
  \\  $^{\barwedge{}}$ University of Toronto\vspace{.05in}
 \\ $^{\musFlat{}}$  University of Chicago\vspace{.05in}
\\ $^{\musSharp{}}$ California Institute of Technology}

\vspace{.25in} 

\date{}

\makeatletter\@addtoreset{section}{part}\makeatother%
\numberwithin{equation}{chapter}

\renewcommand{\triangleq}{:=}
\newcommand{\upperRomannumeral}[1]{\uppercase\expandafter{\romannumeral#1}}

\renewcommand{\hat}{\widehat}
\renewcommand{\bar}{\overline}
\newcommand{\Kk}{{\mathcal{K}}}

\newcommand{\bepsilon}{b}
\newcommand{\bgamma}{\sigma}

\newcommand{\An}{{\mathsf{A}}}
\newcommand{\Pred}{{\mathsf{P}}}
\newcommand{\Gn}{{\mathsf{G}}}
\newcommand{\Un}{{\mathsf{U}}}

\newcommand{\J}{{\mathsf{J}}}
\newcommand{\F}{{\mathsf{F}}}

\newcommand{\apost}{\post^{{\mathsf{app}}}}
\newcommand{\hapost}{\hat{\post}^{{\mathsf{app}}}}

\newcommand{\joint}{\gamma}
\newcommand{\marg}{\kappa}

\DeclareMathOperator*{\argmax}{arg\,max}
\DeclareMathOperator*{\argmin}{arg\,min}
\DeclareMathOperator*{\arginf}{arg\,inf}

\begin{document}

\maketitle 

\newpage

\section*{\Large{\sffamily{Preface}}}

\subsection*{Aim and Overview}

The aim of this book is to demonstrate the potential for
ideas in machine learning\index{machine learning} to impact on the fields of
inverse problems\index{inverse problems} and data assimilation\index{data assimilation}. The perspective
is one that is primarily aimed at researchers from
inverse problems and/or data assimilation who wish to
see a mathematical presentation of machine learning as it
pertains to their fields. As a by-product, we include a 
succinct mathematical treatment of various \emph{fundamental 
underpinning topics} in machine learning, and adjacent areas of (computational)
mathematics. The fundamental underpinning material is summarized in
Part III. Part I of the book is concerned with
inverse problems, employing material from Part III;
Part II of the book is concerned with
data assimilation, employing material from Parts I and III. 
There are different ways to learn from this book, and consequently
there are competing ways in which the material can naturally be organized,
from a pedagogical perspective. In particular, as presented here, the material
allows application-oriented readers to proceed directly to the use of machine learning methodology, in either (or both) of inverse problems (Part I) and data
assimilation (Part II), dipping into the fundamental material in Part III  only as it 
arises and is used. Theory-oriented readers may instead prefer to
work through the fundamentals in Part III first and, having absorbed this material,
study the applications of machine learning from Parts I and II. Collectively
we have taught from this book in both ways and both have their merits. 

\subsection*{Overarching Concepts and Underpinning Fundamental Ideas}
Parts I and II, concerning inverse problems and data assimilation respectively, are organized
to parallel one another. Both start with the introductory Chapters \ref{ch1} and
\ref{lecture7}, framing the fields of inverse problems and data assimilation respectively.
We adopt a Bayesian perspective, deriving optimization approaches from it.
And in both chapters we include a section on what is one of the earliest, and most prevalent,
uses of machine learning in the fields of inverse problems and data assimilation, namely 
surrogate\index{surrogate} modeling.
There are then four overarching concepts that define the parallel organization of material 
in Parts I and II. In Chapters \ref{ch:VI} and \ref{ch:LG} we discuss the use of ideas from 
\emph{variational Bayes}\index{variational Bayes} to approximate probability measures from within tractable, and ideally expressive for the task at hand, subclasses of probability measures. Chapters \ref{ch:priorIP} and \ref{ch:DAML} study the \emph{machine learning
of prior models}\index{prior}, from data. In Chapters \ref{ch:transport} and \ref{ch:LG2} we focus on \emph{transport}\index{transport} as a unifying concept
for the learning of algorithms in inverse problems and data assimilation respectively.
Chapters \ref{ch:data-dependence} and \ref{ch:LG10} discuss \index{amortization}\emph{amortization}, demonstrating how the machine learning
ideas developed previously can be generalized
to learn dependence on the observations that define the inverse problems or data assimilation task; such
algorithms can then be reused on multiple instances of the observed data that define inverse
problems or data assimilation tasks.

The key underpinning fundamental ideas used in this book
are contained in Part III. In Chapter \ref{ch:distances} we study metrics\index{metric}, 
divergences\index{divergence} and scoring rules\index{scoring rule}. The first two provide ways of quantifying closeness of two probability measures;
scoring rules assign a notion of closeness to a probability measure and to a point and,
through averaging, can also be used to construct divergences. These notions of closeness
are used in two primary ways: to measure robustness of probability measures to perturbations in problem specification; and to define objective functions used in machine learning. Chapter \ref{ch:SL} is devoted to supervised learning\index{supervised learning}, focusing on the use of neural networks, random features and Gaussian processes to approximate functions from data. Chapter \ref{ch:UL} is devoted to the topics of unsupervised learning\index{unsupervised learning} and 
generative modeling\index{generative modeling}; in particular we introduce the use of 
\index{transport}transport as an important way of framing, and actioning, many tasks in unsupervised learning and generative modeling.   In Chapter \ref{ch:ts_forecasting} we discuss time-series forecasting\index{time-series forecasting}. Much of this material is
fundamental to new emerging methods for purely data-driven\index{data-driven} data assimilation. 
The ideas in this chapter are important for understanding the current landscape of data assimilation research, but not our focus when studying machine learning for data assimilation in Part II; this is because purely data-driven methods are rapidly evolving and not yet as established as methods that combine model-\index{model-driven} and data-driven approaches.
Chapter \ref{ch:optimization} describes optimization and sampling techniques that are broadly useful for defining algorithms in the fields of inverse problems and data assimilation. Optimization also plays a fundamental role in machine learning methods for inverse problems and data assimilation, since it is used for training machine learning models.

\subsection*{Existing Review Articles}

This is a rapidly evolving research area and there
are already several
articles that review aspects of the material we will cover. See, for example, \cite{arridge2019solving} for uses of ML in inverse problems,
and \cite{bocquet_surrogate_2023,cheng_machine_2023,wang_artificial_2026,bach_interface_2026,arcucci_convergence_2026} and Chapter 10 of \cite{chen_stochastic_2023} for uses of ML in data assimilation.
For background on neural networks and deep learning 
see \cite{goodfellow2016deep}; for background on Gaussian processes see
\cite{williams2006gaussian}. For an overview of metrics and divergences, and their
inter-relations, see \cite{gibbs2002choosing}; for a recent overview of scoring rules see \cite{waghmare2025proper}.

\subsection*{Prerequisites}

Even though the subjects of inverse problems and data assimilation are succinctly reviewed in this book, there remains an assumption of previous knowledge of these topics. The book \cite{sanz-alonso_inverse_2023} (also available in closely related form on arXiv) provides a good background on these topics, and we have tried to maintain similar notation. This book also assumes familiarity with linear algebra, probability, statistics, multivariable calculus, and matrix calculus; we establish notation in these areas in the following subsection. Useful graduate-level references on the topics of linear algebra, probability and statistics include~\cite{trefethen2022numerical, grimmett2020probability, wasserman2013all}. We also refer the reader to books on inverse problems~\cite{engl1996regularization, kaipio2006statistical, vogel2002computational} and data assimilation~\cite{asch2016data, reich2015probabilistic, law2015data}, that provide additional expositions on these topics from the mathematical and computational perspectives. And lastly
we refer to the texts 
\cite{abarbanel2013predicting,kalnay2003atmospheric,majda2012filtering,tarantola2005inverse}
which cover inverse problems and data assimilation from more applications-oriented perspectives.

\subsection*{Notation}

Throughout the book we adopt the following notational conventions:

\begin{itemize}

\item {\bf Sets} 
$\bbN$ denotes the set of positive integers $\{1,2,3, \ldots \}$
and
$\bbZ^+ := \bbN \cup \{0\}=\{0,1,2,3, \ldots \}$ denotes the set of non-negative integers. 
For $m,n \in \bbZ^+$ with $m \le n$ we denote by $[[m,n]]$ the set of all integers in the
closed interval $[m,n].$
$\bbR^d$ denotes the set of $d$-dimensional real vectors, and $\bbR^+$ denotes the non-negative reals.
The set $B(u,\delta) \subset \R^d$ denotes
the open ball of radius $\delta$ at $u$ in $\R^d$,
in the Euclidean norm. The symbol $\varnothing$ denotes the empty set. We also recall the Hausdorff distance\index{distance!Hausdorff}
between two sets $A,B \subset \R^d:$
$$h(A,B)=\max\bigl\{\sup_{a \in A} \inf_{b \in B}|a-b|, \, \sup_{b \in B} \inf_{a \in A}|a-b|\bigr\}.$$

\item {\bf Vector Spaces} 
Given $u \in \R^d$ we define  $u_{-i} \in \R^{d-1}$ by
$$u_{-i}:= (u_1, \ldots, u_{i-1}, u_{i+1}, \ldots, u_d) \in \R^{d-1};$$
we make the natural modification if $i \in \{1,d\}.$
The symbol $I_\du \in \bbR^{d \times d}$ 
denotes the identity matrix on $\Ru$, and
$\Id$ denotes the identity mapping. We denote the Euclidean norm on $\R^d$ by $|\cdot|$, noting
that it is induced by the inner-product $\langle a ,  b \rangle = a^\top b.$  We also use $|\cdot|$ 
to denote the induced norm on matrices and $|\cdot|_F$ to denote the Frobenius norm.
We let $\mathbb{S}^{r-1}:=\{ u \in \R^r:\, |u|=1\}$ denote the unit sphere in $r$-dimensional Euclidean space.

We say that symmetric matrix $A$
is \index{positive definite}positive definite (resp. positive semidefinite) if $\langle u, Au \rangle$ is positive (resp. non-negative)
for all $u \ne 0$, sometimes denoting this by  $A>0$ (resp. $A \ge 0).$ We let $\bbR^{d \times d}_{\textrm{sym}}$ denote the
subset of symmetric matrices in $\bbR^{d \times d}$ and $\bbR^{d \times d}_{\textrm{sym},>}$ the subset of positive definite symmetric matrices
and $\bbR^{d \times d}$ and $\bbR^{d \times d}_{\textrm{sym},\ge}$ the subset of positive semidefinite symmetric matrices. We use $\prec$ (resp. $\preceq$) to denote ordering
between matrices in the cone of positive definite (resp. semidefinite) symmetric matrices.
Likewise we use $\succ$ and $\succeq.$
We often wish to use covariance-weighted inner-product and norm and, to this end,
for covariance matrix $A>0,$ we define $|v|_A^2  = v^\top A^{-1} v $.
This norm $|\cdot|_A$ is induced by the weighted Euclidean inner-product 
$\langle \cdot \;,  \; \cdot \rangle_{A}:=\langle \cdot \;, A^{-1}\cdot \rangle.$ We use $A^+$ to denote the Moore--Penrose pseudoinverse of $A$.

The outer product $\otimes$ between two vectors $a,b \in \R^d$ is defined by the following identity,
assumed to hold for all vectors $c \in \R^d:$
$(a \otimes b)c=\langle b,c \rangle a.$ We also use $\otimes$ to denote the Kronecker product between matrices.
We use $\det$ and $\Tr$ to denote the determinant 
and trace functions on matrices. The Kronecker delta\index{Kronecker delta} is $\delta_{ij}=1$ for $i=j$ and
$0$ otherwise.

\item {\bf Probability} Unless otherwise stated, throughout this book random variables
\index{random variable} take values in Euclidean space, here denoted $\R^d.$
Any such random variable has probability\index{probability!distribution} distribution\index{distribution!probability} defined by a probability measure\index{probability!measure}, say $\mu$, on $\R^d.$ 
We write $u \sim \mu$ for random variable distributed as\index{distributed as} $\mu.$

We will mainly consider probability measures
characterized by a \emph{probability density function},\index{probability!density function} say $\rho$, with respect to Lebesgue measure;
we view $\rho$ as an element of $L^1(\R^d;\R)$ which is non-negative Lebesgue-a.e. in $\R^d$ and satisfies
$$\int_{\bbR^d} \rho(u)\,du=1.$$
Because we mainly work with probability density functions  we will sometimes blur the distinction between probability measures\index{probability!measure} and probability density functions.\index{probability!density function} In particular, when a random variable $u$ has probability density function\index{probability!density function} $\rho$ 
we will write $u \sim \rho$ rather than $u \sim \mu.$  We write i.i.d.\index{i.i.d.} for independent and identically distributed. We let $\cP(\R^d)$ denote the space of all probability measures\index{probability!measure}
(or probability density functions\index{probability!density function}) over $\R^d;$ we simply write $\cP$ if the Euclidean space $\R^d$ is clear.

An \emph{unnormalized density}\index{unnormalized density} $\tilde{\rho}$ is an element of $L^1(\R^d;\R)$ which is non-negative
Lebesgue-a.e. in $\R^d$ and has positive norm in $L^1.$ Then $\rho$ defined by
\begin{align*}
    \rho(u)&=Z^{-1}\widetilde{\rho},\\
    Z&= \int_{\bbR^d}\widetilde{\rho}(v)dv
\end{align*}
is a probability density function.

Occasionally we will need to
employ \index{Dirac measure}Dirac measures; these are not characterized by a probability 
density function\index{probability!density function} in  $L^1(\R^d;\R).$ However we will
use the notational convention that the \index{Dirac measure} Dirac measure at point $v$
has ``density'' $\delta(\cdot-v)$, also denoted by $\delta_v(\cdot).$ A particular use of Dirac
measures arises in the construction of an \emph{empirical measure}\index{probability!measure!empirical}, approximating another measure $\mu$:
if $\{v^{(n)}\}_{n=1}^N \sim \mu$  are i.i.d. then we may form an empirical approximation of $\mu$
as $\frac{1}{N} \sum_{n=1}^N \delta_{v^{(n)}}.$

A \emph{centered random variable}\index{random variable!centered} is a random variable with mean zero. Similarly, a probability density function $\rho$ is called centered if $u \sim \rho$ is a centered random variable. Given probability density function\index{probability!density function} $\rho \in \cP(\R^d)$, and function $g:\R^{d} \to \R^{r}$, $g_\sharp \rho \in \cP(\R^r)$ denotes the probability density function\index{probability!density function} 
of the random variable $g(z)$ where $z \sim \rho.$ We refer to $g_\sharp \rho$ as
the \emph{pushforward}\index{pushforward} of $\rho$ under $g$. 
Formally the pushforward notation $g_\sharp \rho$ means that $\mathbb{P}^{g_\sharp \rho}(A) = \mathbb{P}^\rho(g^{-1}(A))$ for any Borel set $A$. We note that, in some other texts, the pushforward is denoted by $g\#\rho$ or $g_\star \rho$; we
do not use either of these notational conventions. We denote the \emph{convolution}\index{convolution}
of two probability density functions $\rho$ and $\rho'$ as $\rho * \rho'.$ Thus, if $u \sim \rho$ and $u' \sim \rho'$ are independent random variables, then $u+u' \sim \rho * \rho'.$

We sometimes denote by $\Prob (\cdot), \Prob(\cdot \; | \; \cdot )$ the probability density function\index{probability!density function} of
a random variable and its \emph{conditional}\index{probability!conditional} \index{probability!density function}probability density function, respectively. For jointly varying random variables $(u,v)$ we let $u|v$ denote the random variable found by conditioning $u$ on a specific realization of random variable $v.$ Thus $u|v$ has probability density\index{probability!density function} function $\Prob(u|v);$ and we write $\Prob (u)$ and $\Prob (v)$ for the \emph{marginal}
probability\index{probability!marginal} densities on $u$ and $v$, respectively. If the random variables $u$ and $v$ are independent then we write $u \ind v.$

We denote by $F_\rho: \R \to [0,1]$ the \emph{cumulative distribution function}\index{cumulative distribution function} defined by probability density function\index{probability!density function} $\rho$ of a real-valued random variable:
$$F_\rho(u)=\int_{-\infty}^u \rho(w)\,dw.$$
Given $f: \Ru \mapsto  \mathbb{R}$ we denote by
$$\Expect^{u \sim \rho} f(u)=\int_{\Ru}f(u)\rho(u) \, du$$
the expectation of $f$ with respect to probability density function\index{probability!density function} $\rho$ on $\Ru.$ On occasion, given function $f$ defined over $\R^d$, we may denote this expectation by $\Expect^\rho f$ or $\rho(f),$ when the meaning is clear. The 
\emph{characteristic function}\index{characteristic function} of $\rho$ is
$$\varphi_\rho(w)=\Expect^{u \sim \rho} \exp\bigl(i \langle u,w \rangle \bigr).$$

We write $\mathcal{N} \bigl(m,\Sigma \bigr)$ for the distribution of a Gaussian random variable on $\R^d$ with mean $m$ and covariance $\Sigma.$ 
We denote by $\mathcal{N} \bigl(u; m,\Sigma \bigr)$ the probability density function
of this random variable evaluated at $u \in \R^d.$

\item {\bf Calculus} Given function $f:\R^d \to \R$ we denote by $\nabla f: \R^d \to \R^d$ the \emph{gradient}\index{gradient} 
and by $D^2f: \R^d \to \R^{d \times d}$ the \emph{Hessian}\index{Hessian}. 
For more general 
functions $f: \cal{V} \to \R$ acting on elements $v$ of vector space $\cal{V}$
we write $D f$ to denote the derivative. We also directly consider functions 
$f:\R^d \to \R^d$ with \emph{Jacobian matrix}\index{Jacobian matrix} $Df: \R^d \to \R^{d \times d}.$ When there are two potentially varying arguments and we wish to indicate
differentiation with respect to only one of them, say $v$, we will indicate 
this with a subscript: $D_v$ (and similarly for second derivatives). When we contract a derivative
to obtain \emph{divergence}\index{divergence} we write $\div$: for example, for $f:\R^d \to \R^d$, ${\rm Tr}\, Df=\div f.$ For $f:\R \to \R^d$ we write the (typically time-) derivative as $\frac{df}{dt}.$

\item {\bf Functions} 
We use $f\circ g$ to indicate the composition of $f$ and $g$, assuming input dimensions (of $f$) and output dimensions (of $g$) are compatible, allowing composition. We write $\mathsf{id}$ for the identity mapping on $\R^d:$ $\mathsf{id}(u)=u.$
Finally if $A \subset \R^d$ then we define the \emph{indicator function} $\mathbb{1}_A: \R^d \to \R^+$ by $\mathbb{1}_A(x)=1$ for
$x \in A$ and $\mathbb{1}_A(x)=0$ otherwise. For a subset $A\subset \R$ of cardinality $N$, and a point $x\in\R$, the \emph{rank function} 
$\operatorname{rank}(A, x)\in [[0, N]]$,  returns the rank of value $x$ in set $A$: the number of points in $A$ that are less than or equal to $x$.
\end{itemize}

\subsection*{Acknowledgments}
EB acknowledges support from Caltech's Foster and Coco Stanback Postdoctoral Fellowship, the Schmidt Foundation, the Natural Environment Research Council, and the support through US federal funding provided to AMS. RB also gratefully acknowledges support for the von K\'{a}rm\'{a}n Postdoctoral Instructorship from Caltech and AFOSR MURI programs.
DSA is grateful for the support of DOE and NSF (US) and the BBVA Foundation (Spain).
AMS would like to acknowledge the various US government agencies that have 
generously supported research in areas related to this book:  AFOSR,
ARO, DARPA, DoD, NSF and ONR; in particular he would like to acknowledge support
as a Vannevar Bush Faculty Fellow.

All the authors thank the students who took ACM270 at Caltech in Spring 2024 and the TA Yixuan Wang. The lecture notes for this
course constituted an early draft of this book. The students' comments and questions have helped to improve them considerably.
The authors are also grateful to Justin K. Bennett, Jochen Br\"ocker, Bohan Chen, Sangmin Park, and Stephen Yin for helpful input.

\newpage \tableofcontents

\part{Inverse Problems}

\chapter{\Large{\sffamily{Bayesian Inversion}}} 
\label{ch1}
A forward model\index{forward model} $G: \Ru \to \Ry$ 
specifies output $y \in \Ry$ from input 
$u \in \Ru$ by the relationship $y = G(u).$ The 
related inverse problem\index{inverse problem} is to 
recover unknown parameter\index{parameter} $u \in \Ru$ from 
data\index{data}, or observation\index{observation}, 
$y \in \Ry$ defined by
\begin{equation}
\label{eq:jc0}
 y = G(u) + \eta; 
\end{equation}
here $\eta \in \Ry$ denotes \index{observation!noise}observation noise.
In the \index{Bayesian}Bayesian approach to the inverse problem we view 
$(u,y)$ as jointly varying random variables and the solution of the 
inverse problem is the \index{posterior}posterior distribution  
on  parameter\index{parameter} $u$, given a specific instance of the 
data\index{data} $y,$ found by conditioning the joint distribution\index{distribution!joint}. 
Implicit in this framing of the inverse problem is that, 
although $\eta$ is unknown, its distribution is known. 

 In Section~\ref{sec:11} we formulate \index{Bayesian!inverse problem}Bayesian inverse problems and state Bayes Theorem, giving an explicit expression for the \index{posterior}posterior probability density function. In Section~\ref{sec:112} we discuss finite-dimensional summaries of the posterior and we connect Bayesian inversion to the classical optimization-based approach to inverse problems.
 Section~\ref{sec:13} describes the \index{well-posed}well-posedness of the \index{Bayesian}Bayesian formulation, using the Hellinger 
distance\index{distance!Hellinger}
from Chapter~\ref{ch:distances}. In Section~\ref{sec:ME} we discuss
various perspectives on model error. Section~\ref{sec:IPSU} studies the
use of machine-learned surrogate models, serving as an introduction to the potential for
the use of machine learning within inverse problems. We conclude the chapter in
Section~\ref{sec:14}, containing bibliographical remarks.
In this chapter we do not cover classical computational methods for solving Bayesian inverse problems.
However we describe both optimization\index{optimization} and Markov chain Monte Carlo\index{Markov chain Monte Carlo} methods
in Chapter~\ref{ch:optimization}.

\section{\index{Bayesian!inversion}Bayesian Inversion}
\label{sec:11}

We view $(u, y ) \in \Ru \times \Ry$ as a random variable, whose joint distribution\index{distribution!joint} is specified by means of the identity \eqref{eq:jc0} and the following assumption on the distribution of $(u,\eta) \in \Ru \times \Ry:$
\begin{assumption}\label{a:jc1}
The distribution of the random variable $(u,\eta) \in \Ru \times \Ry$ is defined by assuming that $u \ind \eta$, that $u \sim \pr$ and that $\eta \sim \noise$, where $\pr$ and $\noise$ are probability density functions and, in addition, $\noise$ is centered.
\end{assumption}
\noindent We make the following assumption about the provenance of the data:
\begin{dataassumption}\index{Data Assumption}
\label{da:vb}
Data $y \in \R^{\dy}$ is given and is assumed to have come from a realization of the random variable $(u,\eta)$,
under Assumption~\ref{a:jc1},  and identity \eqref{eq:jc0}.
\end{dataassumption}

Combining these assumptions we can define the 
\index{inverse problem!Bayesian}Bayesian inverse problem, a specific and important
subclass of general Bayesian inference\index{inference!Bayesian}.
We refer to $\pr$, the probability density function of $u$,  as the \index{prior}\emph{prior} probability density function. Given
identity \eqref{eq:jc0}, then for fixed $u\in \Ru,$ the distribution of $y$ given $u$
defines the \index{likelihood}\emph{likelihood}:
\begin{equation}
\label{eq:like}
y |  u \sim \like(y|u):=\noise\bigl(y - G(u)\bigr).
\end{equation}
The \index{posterior}\emph{posterior} probability density function is the conditional distribution of $u$ given $y$, that is the distribution of random
variable $u|y$. The posterior density is the solution to the Bayesian
formulation of the inverse problem.

The primary computational challenge associated with Bayesian inversion is that the posterior 
is an infinite-dimensional object, even when $u$ is finite-dimensional.
In particular it is important to appreciate that, although 
\index{Bayes Theorem}Bayes Theorem (which follows) delivers a formula
for the probability density function of $u|y$, the task of obtaining
information from this probability density function, for example by drawing many samples from it, is,
in general, a substantial challenge.

 \begin{theorem}
\label{t:bayes}
 Let Assumption~\ref{a:jc1} and Data Assumption~\ref{da:vb} hold,
 and assume further that
\[
 Z = Z^y :=\int_{\Ru} \noise\bigl( y - G(u)\bigr)\pr(u) \, du  >0.\]
 Then $u | y\sim \post^y,$ where\footnote{When there is no possibility of confusion,
we will  simply write $\post$ for  the \index{posterior}posterior probability density function, rather than
$\post^y.$}
\begin{equation}\label{eq:bayesformula}
 \post^y(u) = \frac{1}{Z} \noise\bigl(y - G(u)\bigr)\pr(u).
\end{equation}
 \end{theorem}
 \begin{proof}
Recall from the Preface that we denote by $\Prob (\cdot)$ the probability density function of a random variable; and we denote 
by $\Prob(\cdot | \cdot )$ the probability density function conditional on the second argument. 
The standard laws of conditional probability give the two identities
 \[
 \begin{aligned}
 \Prob(u , y) &=  \Prob (u | y) \Prob(y),  \textrm{ if } \Prob(y) > 0 ,\\
 \Prob(u, y) &=  \Prob (y | u ) \Prob(u),  \textrm{ if } \Prob(u) > 0 .\\
\end{aligned}
 \]
The marginal probability density function on $y$ is given by
 \begin{align*}
 \Prob(y) &= \int_{\Ru} \Prob( u, y) \, du  \\
 & = \int_{\Ru} \Prob(y|u) \Prob(u) \, du = Z, 
 \end{align*}
and so the assumption that $Z>0$ ensures that the data could indeed have been
obtained from the postulated model. Since $Z>0$ we have, by combining the
two identities above, the desired result
 \begin{equation*}
 \Prob(u | y) = \frac{1}{\Prob(y) } \Prob(y | u) \Prob(u)   =  \frac{1}{\Prob(y)}\noise\bigl(y - G(u )\bigr) \pr(u).
 \end{equation*}
\end{proof}

\begin{remark}
\label{rem:eaudc} 
We note that the \emph{normalization constant}\index{normalization constant} $Z$ can be written as
an expectation with respect to the prior:
\begin{equation}
    \label{eq:yesnorm}
    Z=\bbE^{u \sim \rho} [\noise\bigl(y - G(u)\bigr)].
\end{equation}
Thus Bayes formula \eqref{eq:bayesformula} can be written as
\begin{equation}\label{eq:bayesformula2}
 \post^y(u) =\frac{\noise\bigl(y - G(u)\bigr)\pr(u)}{\bbE^{u \sim \rho} [\noise\bigl(y - G(u)\bigr)]}.
\end{equation}
This makes explicit that Bayes formula may be viewed as a nonlinear map\footnote{Strictly speaking $\An$ is nonlinear when viewed as acting on the vector space $L^1(\R^{\du};\R)$; the space of probability density functions is a subset of this vector space, but is not a linear subspace.}
taking prior $\rho$ into posterior $\post^y:$
\begin{equation}
\label{eq:I2A}
\post^y=\An(\pr;y).
\end{equation}
This notation will be particularly useful when we consider data assimilation\index{data assimilation} 
in Chapter \ref{lecture7}. In that specific context map $\An$ is known as the analysis\index{analysis} map.
We highlight that exact form of the map $\An$ is determined by the likelihood.\index{likelihood}
\end{remark}

\begin{remark}
\label{rem:minus1}
We write the density of the 
joint distribution of $(y,u) \in \Ry \times \Ru$ as $\joint(y,u).$
We write the marginal density on $y \in \Ry$ as $\marg(y).$ Note that
$\marg(y)=Z.$
\end{remark}

\begin{remark}
\label{rem:zero}
If $G$ is linear and $(u,\eta)$ is Gaussian\index{Gaussian} 
then the posterior on $u|y$ 
is Gaussian. This Gaussian setting provides an explicitly solvable
inverse problem in which the posterior\index{posterior} is characterized
by its mean and covariance; it 
reduces the infinite-dimensional problem to
a finite-dimensional problem for a vector (mean) in $\Ru$ and a
positive definite matrix (covariance) in $\Ruu.$

To see this explicitly, let $G(u)=Au$ for matrix $A \in \Ryu$, and
assume that $u \sim \Nc(0,\Cpn)$ and $\eta \sim \Nc(0,\Gamma)$ for positive definite covariance matrices $\Cpn$ and $\Gamma$. 
Using independence of $u$ and $\eta$ to define the likelihood
and completing the square, we see that setting
\begin{subequations}
\label{eq:GEN}
\begin{align}
C^{-1}&=\Cpn^{-1}+A^\top \Gamma^{-1}A,\\
C^{-1}m&=A^\top \Gamma^{-1}y,
\end{align}
\end{subequations}
we have that $\post^{\vct{y}}=\Nc(m ,\mtx C).$  
\end{remark}

\begin{remark}
We have concentrated on formulating Bayes Theorem around equation \eqref{eq:jc0}
in which the noise appears additively and is independent of the unknown $u.$
However Bayes Theorem is not restricted to this setting.  Section~\ref{sec:den} provides 
an explicit example, going beyond this additive noise setting,
to formulate density estimation in a Bayesian fashion.
\end{remark}

We conclude by discussing the following modification of Data Assumption~\ref{da:vb}:

\begin{dataassumption}\index{Data Assumption}
\label{da:vbx}
Assume that $u^{\mathsf{true}} \in \R^{\du}$ is a deterministic true unknown parameter and that
$y \in \R^{\dy}$ is given by
\begin{equation}
\label{eq:jcx}
 y = G(u^{\mathsf{true}}) + \eta,
\end{equation}
where $\eta$ is an unknown realization drawn from given centered noise probability density function $\noise.$
\end{dataassumption}

\begin{remark}\label{rem:avdw}
This assumption is useful for two primary reasons. The first is as the basis for analysis of Bayesian methods from the \emph{frequentist}\index{frequentist} perspective; for instance, it can be important to understand whether the posterior distribution concentrates around true parameter $u^{\mathsf{true}}$ if the variance of the noise $\eta$ is small or if the volume of data is high, and then to quantify the concentration rate.
A second use of the assumption in the Bayesian context is when studying \index{model misspecification}\emph{model misspecification}; in particular it can be useful when the data $y$ is derived from a true unknown $u^{\mathsf{true}} \in \R^{\du}$ that could not have been drawn from the prior $\pr.$  
\end{remark}

\section{Finite-Dimensional Summaries}
\label{sec:112}
The
\index{posterior}posterior $\post^y$ contains all knowledge 
about the parameter $u$, given Assumption~\ref{a:jc1} and Data Assumption~\ref{da:vb}.
It is often useful, however, to extract finite-dimensional information from the \index{posterior}posterior distribution, in order to summarize aspects of the distribution.
Remark~\ref{rem:zero} identifies a specific situation where 
a finite-dimensional summary, comprising the posterior mean and covariance, fully characterizes the posterior, since the posterior is Gaussian. In general, however, we must seek finite-dimensional summaries that will not fully characterize the posterior. These summaries are often computed using optimization and sampling algorithms described in Chapter~\ref{ch:optimization}.

\subsection{Maximum \emph{a Posteriori} Estimator}\label{ssec:MAP}
A natural summary 
is the \index{posterior}{\em posterior mode}\index{mode} or MAP estimator:
\begin{definition}\label{def:map}
A {\em maximum \emph{a posteriori} (\index{MAP estimator}MAP) estimator} of $u$ given data $y$ is  defined as any point $\umap$ satisfying
	$$\umap \in \argmax_{\vct{u}\in\Ru}\post^{\vct{y}}(\vct{u}).$$
\end{definition}

Optimization algorithms are often required to compute the MAP estimator. We now show how MAP estimation connects with classical optimization-based approaches to inversion.
Recall that the \index{posterior}posterior probability density function $\post^y$ on $u|y$ from \index{Bayes Theorem}Theorem~\ref{t:bayes} has the form
\[
\post^y (\vct u) = \frac{1}{Z} \noise \bigl(\vct y- \vct G(\vct u)\bigr) \pr (\vct u).
\]
Recalling \eqref{eq:like} we define a \index{loss!function}\emph{loss function}
\begin{equation}
\label{eq:loss}
\loss (\vct u) = - \log \noise \bigl(\vct y-\vct G(\vct u)\bigr)= - \log \like(y|u),
\end{equation}
and a \emph{regularizer}\index{regularizer}
\begin{equation}
\label{eq:reg}
\reg(\vct u) = - \log \pr(\vct u).
\end{equation}
In fact we will refer to any function which differs from $\loss(\vct u)$
(resp. $\reg(\vct u)$) by a constant independent of $u$ as the 
loss\index{loss} (resp. regularizer\index{regularizer}).
Adding the loss function and regularizer we obtain
an \index{objective}\emph{objective function} of the form
\begin{equation}
\label{eq:MAPO}
\J(\vct u) = \loss(\vct u) + \reg(\vct u).
\end{equation}
Furthermore
\[
\post^y (\vct u) = \frac{1}{Z} \noise \bigl(\vct y-\vct G(\vct u)\bigr) \pr (\vct u) \propto e^{-\J(\vct u)}.
\]
Thus the \index{MAP estimator}MAP estimator can be rewritten as a minimizer\index{minimizer}
of $\J$ as follows:
\begin{subequations}
\label{eq:mapismin}
\begin{align}
\umap & \in \argmax_{\vct{u}\in\Ru}\post^{\vct{y}}(\vct{u}) \\
&= \argmin_{u \in \Ru} \J(u);
\end{align} 
\end{subequations}
minimizing the \index{objective}objective function $\J(\cdot)$
is equivalent to determining a MAP estimator defined by 
maximizing the \index{posterior}posterior probability density function $\post^y(\cdot)$.

\begin{remark}\label{rem:nonorm}
The optimization problem defining the MAP estimator does not depend on the
normalization constant\index{normalization constant} $Z$ in the expression for the posterior in Bayes Theorem~\ref{t:bayes}. This is computationally significant,
as it eliminates the need to compute an integral over $\R^d.$
\end{remark}

\begin{remark} \label{rem:flat}
 Setting the regularizer to zero, known as choosing a \emph{flat prior}\index{prior!flat},
 we obtain the maximum likelihood estimation (MLE)\index{maximum likelihood estimation}
 problem.
\end{remark}

\begin{remark}
The formulation involving minimization of $\J(\cdot)$ is simplest 
when the argument of the logarithm in \eqref{eq:reg}, the prior, is strictly positive 
on all of $\Ru.$ However where the prior $\pr$ is zero we may interpret the regularizer
$\reg$ to take value $+\infty$. The effect of this is to confine the 
minimization problem to the support of the prior.
\end{remark}

\begin{example}[Gaussian\index{Gaussian} Observation Noise\index{observation!noise} and 
Gaussian Prior\index{prior!Gaussian}]
	\label{ex:loss-l2}
	If $\noise = \Nc(\zerovct ,\mtx \Gamma)$, for positive definite covariance $\Gamma$, then
	$$\noise\bigl(\vct y-\vct G(\vct u)\bigr) \propto \exp(- \frac{1}{2} |\vct y-\vct G(\vct u)|_{\mtx \Gamma} ^2).$$
Thus in this case the \index{loss}loss is 
$$\loss(\vct u) = \frac{1}{2} |\vct y-\vct G(\vct u)|_{\mtx \Gamma}^2,$$ 
a \(\mtx \Gamma\)-weighted \(\losss_2\) loss.
Now assume that the \index{prior}prior is a centered Gaussian
\(\pr  = \Nc(\zerovct ,\mtx \Cpn)\), where $\Cpn$ is positive definite.  
Then the corresponding regularizer is given by
$$\reg(\vct u) = \frac{1}{2} |\vct u|_{\mtx \Cpn}^2.$$

Combining these assumptions we obtain a canonical \index{objective}objective function 
\begin{equation}
\label{eq:Phillips}
\J(\vct u) = \frac{1}{2} |\vct y-\vct G(\vct u)|_{\mtx \Gamma}^2 + \frac{1}{2} |\vct  u|_{\mtx \Cpn}^2.
\end{equation}
We refer to minimization of \eqref{eq:Phillips}
as the \emph{Tikhonov--Phillips regularized inverse 
problem}\index{inverse problem!Tikhonov--Phillips regularized}.
If $\Gamma = I_k$ and $\Cpn=\lambda^{-1} I_d$
then this reduces
to the classical \emph{Tikhonov regularized inverse problem} \index{inverse problem!Tikhonov regularized}
with objective function
\begin{equation}
\label{eq:Tikhonov}
\J(\vct u) = \frac{1}{2} |\vct y-\vct G(\vct u)|^2 + \frac{\lambda}{2} |\vct  u|^2.
\end{equation}
Finally, consider the setting of Remark \ref{rem:zero}, where $G(u)=Au$ and
the observation noise and prior are Gaussian. 
Then the MAP estimator\index{MAP estimator} related
to the Tikhonov--Phillips regularized inverse 
problem\index{inverse problem!Tikhonov--Phillips regularized} is unique and
given by point $m$ defined in \eqref{eq:GEN}. 
In this case the posterior is Gaussian
and the posterior mean is equal to the (in this case unique) MAP estimator. 
For general non-Gaussian posteriors, however, 
the MAP estimator\index{MAP estimator} 
and the posterior mean do not coincide.
\end{example}

\begin{example}[\(\losss_1\) Regularizer -- Laplace Prior]
	\label{ex:reg-l1}
Now consider the setting where again $\noise = \Nc(\zerovct ,\mtx \Gamma)$, but
$$\pr(\vct u) \propto \exp\biggl(-\lambda \sum_{i=1}^\du |u_i| \biggr) = \exp(-\lambda |\vct u|_1).$$
This is known as a Laplace\index{prior!Laplace} 
distribution. Then
\(\reg( \vct u ) = \lambda |\vct u|_1\), an \(\losss_1\) regularizer.
Combining this \index{prior}prior with the weighted \(\losss_2\) \index{loss}loss above,
we obtain the \index{objective}objective function 
$$\J(\vct u) = \frac{1}{2} |\vct y - \vct G ( \vct u ) |_{\mtx \Gamma}^2 + \lambda |\vct u |_1.$$ 
It is well known that regularizers of this type promote sparse solutions when $\J(\cdot)$
is minimized. However it is important to appreciate that samples
from the underlying \index{posterior}posterior distribution are typically not sparse. 
\end{example} 

\subsection{Posterior Expectations} \label{ssec:PE1}
In the previous Subsection~\ref{ssec:MAP} we summarized the posterior distribution $\post^y$ using its mode, known as the \index{MAP estimator}MAP estimator.
 Many other finite-dimensional summaries of the posterior can be obtained through posterior expectations of the form 
\begin{equation}\label{eq:posteriorexpectations}
    \Expect^{\post^y} [f(u)] = \int_{\Ru} f(u) \, \post^y(u) \, du ,
\end{equation}
where $f$ is a suitable choice of \emph{test function}, which may take real, vector or matrix values. 
Furthermore, using \eqref{eq:bayesformula2}, this can be rewritten as the ratio of two expectations:
\begin{equation}\label{eq:posteriorexpectations2}
    \Expect^{\post^y} [f(u)] = \frac{\bbE^{u \sim \rho}[\noise\bigl(y - G(u)\bigr)f(u)]}{\bbE^{u \sim \rho} [\noise\bigl(y - G(u)\bigr)]}.
\end{equation}

Both of the preceding formulae can be used as the basis of numerical methods to approximate useful finite-dimensional posterior summaries. Before discussing the numerical methods we first
discuss three important choices of test function and the corresponding posterior summaries.

\begin{example}[Posterior Mean\index{posterior! mean estimator} Estimator] \label{ex:pme}
    Letting $f: \R^d \to \R^d$ be the identity map in $\Ru,$ so that $f(u) = u$ for all $u \in \Ru,$ we obtain the \index{posterior!mean estimator}posterior mean estimator $\upm(y) := \Expect^{\post^y} [u]$ of $u$ given data $y.$ The posterior mean estimator is also sometimes known as the \emph{conditional mean}\index{conditional mean}. In applications such as imaging, the MAP estimator often provides sharper reconstructions than the posterior mean estimator; however, the latter has the advantage of being more stable to perturbations in the model, as will be explained in Section~\ref{sec:13} below. 
\end{example}

An important property of the posterior mean is that it minimizes the squared error in expectation over the posterior, as shown by the following theorem.

\begin{theorem}\label{th:postmean}
Assume that $\Expect^{u\sim \post^y}[|u|^2]<\infty.$
    For all measurable functions $v(y)$ of $y$, we have that
    \begin{equation*}
    \Expect^{u\sim \post^y} \Bigl[ |u - \upm(y) |^2 \Bigr] \leq
        \Expect^{u\sim\post^y}\Bigl[ |u - v(y)|^2 \Bigr];
    \end{equation*}
furthermore, the inequality is strict if $v(y)\neq \upm(y)$.
\end{theorem}
\begin{proof}
    We first note that the left- and right- hand side of the inequality are well defined since
    $\Expect^{u\sim \post^y}[|u|^2]<\infty.$ Straightforward manipulations show that
    \begin{align*}
        \Expect^{u\sim\post^y}\Bigl[ |u - v(y)|^2 \Bigr] =& \Expect^{u\sim\post^y}\Bigl[ |(u - \upm(y)) + (\upm(y) - v(y))|^2 \Bigr]\\
        =& \Expect^{u\sim\post^y}\Bigl[ |u - \upm(y)|^2\Bigr] + 2 \Expect^{u\sim\post^y}\Bigl[(u - \upm(y))^\top(\upm(y) - v(y)) \Bigr]\\
        &\qquad+ \Expect^{u\sim\post^y}\Bigl[ |\upm(y) - v(y)|^2\Bigr].
    \end{align*}
    Now note that the second term is zero, and the expectation is
    redundant in the third term. We therefore find
    \begin{equation*}
       \Expect^{u\sim\post^y}\Bigl[ |u - v(y)|^2 \Bigr] = \Expect^{u\sim\post^y}\Bigl[ |u - \upm(y)|^2\Bigr] + |\upm(y) - v(y)|^2;
    \end{equation*}
    This establishes the desired inequality. Furthermore it demonstrates that
    the inequality is strict if $v(y)\neq \upm(y)$.
\end{proof}

\begin{remark}\label{rem:bayes_estimators}
    Summary statistics that minimize a cost function in expectation over the posterior are known as \emph{Bayes estimators}\index{estimator!Bayes}. Theorem~\ref{th:postmean} shows that the posterior mean is the Bayes estimator for the squared error. Other Bayes estimators correspond to different cost functions. For example, the posterior median is the Bayes estimator when using the $L^1-$norm on $\R$ to define the error (absolute-error\index{loss!absolute-error} objective
    function). For vector-valued states
 on $\R^{\du}$, use of the component-wise $L^1-$norm yields the coordinate-wise posterior medians.
\end{remark}

\begin{example}[Posterior Covariance and Variance]
Letting $f: \Ru \to \Ru \times \Ru$ be the outer product $f(u) = (u - \upm) \otimes (u - \upm),$ we obtain the posterior covariance $\Expect^{\post^y} [ (u - \upm) \otimes (u - \upm) ].$ The trace of the posterior covariance, given by 
    $\Expect^{\post^y} [|u - \upm|^2]$ gives the posterior variance, which quantifies the spread of the posterior distribution. 
\end{example}

\begin{example}[Posterior Probabilities]
    Letting $f: \Ru \to \R$ be the indicator function of a set $A \subset \Ru,$ so that $f(u) = \mathbb{1}_A(u),$ we obtain the posterior probability of $A$    $\Expect^{\post^y} [\mathbb{1}_A(u)]=\Prob^{\post^y}(A)$. In some applications, it is important to understand whether the unknown $u$ lies in a given set $A \subset \Ru$ of the parameter space. In others, it is important to find set $A \subset \Ru$ with a prescribed probability, in order to build \emph{credible intervals}.\index{Credible interval} Credible intervals play a similar role in Bayesian statistics to confidence intervals in frequentist statistics. 
\end{example}

 Finding the MAP estimator\index{MAP estimator} involves solving an optimization problem; in contrast, finding posterior expectations of the form \eqref{eq:posteriorexpectations} involves computing an integral, or integrals, over $\Ru.$ This task can be challenging in high dimension $d$ -- constants entering errors resulting from standard quadrature rules typically depend badly on dimension. Monte Carlo\index{Monte Carlo} methods, and Markov chain 
 Monte Carlo\index{Markov chain Monte Carlo} methods, are often preferred over standard quadrature rules because they have favorable dependence on dimension.

Given $g: \R^d \to \R$ and probability measure $\intp$,
 the idea of Monte Carlo methods is to obtain $N$ independent samples $\{u^{(n)} \}_{n=1}^N$ from $\intp$ 
 and make the approximation\index{empirical measure}
\begin{equation} \label{eq:mca}
    \Expect^{\intp} [g(u)] \approx \frac{1}{N} \sum_{n=1}^N g(u^{(n)}).
\end{equation}
This can be understood as making the approximation
\begin{subequations}
\label{eq:insert99}
\begin{align}
\Expect^{\intp} [g(u)]&\approx\Expect^{\intp^{N}} [g(u)],\\   
\intp^{N} &= \frac{1}{N} \sum_{n=1}^N \delta_{u^{(n)}}.
\end{align} 
\end{subequations}
The error incurred in making this approximation is of size $1/\sqrt{N};$ this idea is made precise
in Theorem \ref{t:MC}. Furthermore the ideas, and approximation theory, can be extended to
functions $g$ taking values in general normed vector spaces.

The Monte Carlo idea can be used to approximate posterior expectations under $\post^y$, as follows.
First consider the expression \eqref{eq:posteriorexpectations2} for posterior expectation
of $f$ and notice that it is the ratio of two expectations under $\pr.$ Using the Monte
Carlo approximation \eqref{eq:mca} with $\intp=\pr$ and with $g(\cdot)=\noise\bigl(y-G(\cdot)\bigr)f(\cdot)$
and with $g(\cdot)=\noise\bigl(y-G(\cdot)\bigr)$ delivers the desired approximation. 
This approach is generally applicable provided that we can obtain independent samples from the prior.

In principle, the same approximation \eqref{eq:mca} can be used directly with $\intp=\post^y$ in \eqref{eq:posteriorexpectations}.
In practice obtaining \emph{independent} samples that are \emph{exactly} distributed according to $\post^y$ is often harder than for the prior, and indeed typically
not possible. However \emph{Markov chain Monte Carlo}\index{Monte Carlo!Markov chain} (MCMC),\index{MCMC}  discussed in Section~\ref{sec:MCMC}, provides a widely applicable approach to obtain \emph{correlated} samples $\{u^{(n)}\}_{n=1}^N$ from a  Markov chain so that  $u^{(n)}$ is \emph{approximately} distributed according to the posterior for large $N$ and \eqref{eq:insert99} applies with $\intp=\post^y$ and
without the independence assumption on samples.
MCMC sampling algorithms are commonly used to compute posterior expectations in Bayesian inverse problems.

\begin{remark}
    \label{rem:nonorma} The pure Monte Carlo approach just outlined, using independent prior samples, can be inefficient.
In particular, whenever $\noise\bigl(y-G(\cdot)\bigr)$ and $\pr(\cdot)$ are largest in different parts of the space $\R^d$, estimation
of the normalization constant\index{normalization constant} $Z$, the denominator in \eqref{eq:posteriorexpectations2}, is inefficient.
A key feature of MCMC methods for Bayesian inference
is that they do not require approximation of the normalization constant\index{normalization constant} $Z$, 
typically making them more
efficient than Monte Carlo approximation from independent prior samples, despite the correlation of the MCMC samples. Recall, also, that the
MAP estimator does not require evaluation of the  normalization constant\index{normalization constant} $Z$ -- see Remark \ref{rem:nonorm}.
\end{remark}

\section{\index{well-posed}Well-Posedness of \index{Bayesian!inverse problem}Bayesian Inverse Problems}\label{sec:13}

The MAP estimator is an unstable quantity in that a small change in the problem specification, such as the forward model $G$ or the data $y,$ can lead to a large change in the MAP estimator. This is because the minimizer of a non-convex 
objective function may change discontinuously with respect to arbitrarily small changes in the objective function itself.
On the other hand, the Bayesian formulation of the inverse problem\index{Bayesian!inverse problem} 
leads to stability in broad generality: small changes in the
problem specification lead to small changes in the posterior density and, consequently, to small changes in posterior expectations of suitable test functions. Here
we prove a representative result of this type.

We consider two different  \index{likelihood}likelihoods
\[\like(y|u)=\noise\bigl(y-G(u)\bigr)\quad \text{and}\quad \like_\delta(y|u)=\noise\bigl(y-G_\delta(u)\bigr)\]
associated with two different forward\index{forward model}
models $G(u)$ and $G_\delta(u).$ 
Assuming the associated Bayesian inverse problems both adopt the same prior
$\rho$ we obtain two different posteriors of the form
\[\post^y(u)=\frac{1}{Z}\like(y|u)\pr(u)\quad \text{and}\quad \post_\delta^y(u)=\frac{1}{Z_\delta}\like_\delta(y|u)\pr(u),\]
where $Z, Z_\delta$ are the corresponding normalizing constants. 
Our aim is to show that if $\like(y|u)$ and  $\like_\delta(y|u)$ are close, then
so are the associated posteriors. To this end we make the following assumptions
about the likelihoods, in which we view the data $y$ as fixed.

\begin{assumption}\label{a:dh1}
The data defined by Assumption~\ref{a:jc1} and Data Assumption~\ref{da:vb}
has positive probability under the resulting joint distribution on $(u,y)$, so that $Z>0.$ There exist $\delta^+>0$ and $K_1,K_2<\infty$ such that, for all $\delta\in (0,\delta^+],$ 
	\begin{itemize}
	\item[(i)] $|\sqrt{\like(y|u)} - \sqrt{\like_\delta(y|u)}|\leq \phi(u)\delta$, for some $\phi(u)$ 
satisfying $\Expect^{\pr} [\phi^2(u)] \le K_1^2$;
	\item[(ii)] $\sup_{u\in\Ru}(|\sqrt{\like(y|u)}|+|\sqrt{\like_\delta(y|u)}|) \le K_2.$
	\end{itemize}
\end{assumption}

Recall the Hellinger distance\index{distance!Hellinger} $\dhell(\cdot,\cdot)$ 
from Chapter~\ref{ch:distances}. The main result of this section is:

\begin{theorem} 
\label{t:wpz}
Under Assumption~\ref{a:dh1}, there exist 
$\Delta,c>0$ such that, for all $\delta\in(0,\Delta)$,
      \[\dhell(\post^y,\post_\delta^y)\leq c \delta.\]
\end{theorem}

\begin{remark}
Before proving it, we make two observations about the implications of Theorem~\ref{t:wpz}:

\begin{itemize}

\item The theorem ensures that expectations of test functions 
of $u$, with $u$ distributed according
to $\post^y$ and $\post_\delta^y$ respectively,
are order $\delta$ apart, provided that sufficient moments of those
test functions are available; this may be shown by use of Lemma \ref{lemmatv}.

\item The theorem is prototypical of a variety of stability results for
Bayesian inversion. Here we have viewed data $y$ as fixed and considered error in the likelihood arising from approximation of the forward model.
In Section~\ref{sec:IPSU} we establish a similar result
in the context of machine learning approximations of the
Bayesian inverse problem. The forward model is replaced by a surrogate found, for example, from
a neural network approximation; this then changes the likelihood.
It is also possible to estimate changes in the posterior, in the Hellinger metric, 
with respect to changes in the data; this, of course, also changes the likelihood.
\end{itemize}
\end{remark}

 \begin{proof}[Proof of Theorem~\ref{t:wpz}]

To prove Theorem~\ref{t:wpz}, we first characterize the stability
of the normalization\index{normalization constant} constants.
We show that, under Assumption~\ref{a:dh1}, there exist 
$\Delta, c_1,c_2 >0$ such that, for all $\delta\in(0,\Delta)$, 
      \[|Z-Z_\delta|\leq c_1\delta\quad \text{and}\quad Z,Z_\delta>c_2.\]
To see this, noting that $Z=\int_{\Ru} \like(y|u)\pr(u) \, du$ and
$Z_\delta=\int_{\Ru} \like_\delta(y|u)\pr(u) \, du$, 
we have
      \begin{align*}
      |Z-Z_\delta|=& \biggl| \int_{\Ru} \bigl(\like(y|u)-\like_\delta(y|u)\bigr)\pr(u) \, du    \biggr|   \\
      \leq&\ \Big(\int_{\Ru} \Bigl|\sqrt{\like(y|u)}- \sqrt{\like_\delta(y|u)}\Bigr|^2\pr(u) \, du\Big)^{1/2}\Big(\int_{\Ru} \Bigl|\sqrt{\like(y|u)}+\sqrt{\like_\delta(y|u)}\Bigr|^2\pr(u) \, du\Big)^{1/2}\\
      \leq&\ \Big(\int_{\Ru} \delta^2 \phi(u)^2\pr(u) \, du\Big)^{1/2}\Big(\int_{\Ru} K_2^2\pr(u) \, du\Big)^{1/2}\\
      \leq&\ K_1 K_2\delta,\quad \delta\in(0,\delta^+). 
      \end{align*}
The Lipschitz stability result follows by taking $c_1=K_1 K_2.$
      Therefore, for $\delta\leq \Delta\coloneqq\min\{\frac{Z}{2K_1 K_2},\delta^+\}$, we have
      \[Z_\delta\geq Z-|Z-Z_\delta|\geq \frac{1}{2}Z.\]
      Since $Z$ is assumed positive we deduce the lower bound on $Z_\delta$ and take $c_2=\frac12 Z.$

   We now study the Hellinger distance between the two posteriors, using
   the preceding estimate on the Lipschitz stability of the normalization constant. To this
   end we break the total error into two contributions, one reflecting
the difference between $Z$ and $Z_\delta$, and the other the difference between $\like$ and $\like_\delta$:
      \begin{align*}
      \dhell(\post^y,\post_\delta^y)=&\ \frac{1}{\sqrt{2}}\Bigl\|\sqrt{\post^y}-\sqrt{\post_\delta^y}\Bigr\|_{L^2}\\
      =&\  \frac{1}{\sqrt{2}}\Bigg\|\sqrt{\frac{\like\pr}{Z}}-\sqrt{\frac{\like\pr}{Z_\delta}}+\sqrt{\frac{\like\pr}{Z_\delta}}-\sqrt{\frac{\like_\delta\pr}{Z_\delta}}\Bigg\|_{L^2}\\
      \leq&\ \frac{1}{\sqrt{2}}\Bigg\|\sqrt{\frac{\like\pr}{Z}}-\sqrt{\frac{\like\pr}{Z_\delta}}\Bigg\|_{L^2}+\frac{1}{\sqrt{2}}\Bigg\|\sqrt\frac{\like\pr}{Z_\delta}-\sqrt{\frac{\like_\delta\pr}{Z_\delta}}\Bigg\|_{L^2}.
      \end{align*}
From the stability estimate on the normalization constants we have, 
for $\delta\in(0,\Delta)$, 
      \begin{align*}
      \Bigg\|\sqrt{\frac{\like\pr}{Z}}-\sqrt{\frac{\like\pr}{Z_\delta}}\Bigg\|_{L^2}=&\ \Bigg|\frac{1}{\sqrt{Z}}-\frac{1}{\sqrt{Z_\delta}}\Bigg|\Bigg(\int_{\Ru} \like(y|u)\pr(u) \, du\Bigg)^{1/2}\\
      =&\ \frac{|Z-Z_\delta|}{(\sqrt{Z}+\sqrt{Z_\delta})\sqrt{Z_\delta}}\\
      \leq&\ \frac{c_1}{2c_2}\delta.
      \end{align*}
From Assumption~\ref{a:dh1} we have
      \[\Bigg\|\sqrt{\frac{\like\pr}{Z_\delta}} - \sqrt{\frac{\like_\delta\pr}{Z_\delta}}\Bigg\|_{L^2}=\frac{1}{\sqrt{Z_\delta}}\Bigg(\int_{\Ru}\Bigl|\sqrt{\like(y|u)}- \sqrt{\like_\delta(y|u)}\Bigr|^2\pr(u) \, du\Bigg)^{1/2} \leq \sqrt{\frac{K_1^2}{c_2}}\delta.\]
      Therefore
      \[\dhell(\post^y,\post_\delta^y)\leq \frac{1}{\sqrt{2}}\frac{c_1}{2c_2}\delta+\frac{1}{\sqrt{2}}\sqrt{\frac{K_1^2}{c_2}}\delta=c \delta,\]
      with $c=\frac{1}{\sqrt{2}}\frac{c_1}{2c_2}+\frac{K_1}{\sqrt{2c_2}}$,
which is independent of $\delta$.
      \end{proof}

       The following corollary of Theorem~\ref{t:wpz} is a straightforward consequence of Lemma \ref{l:dh1}:

       \begin{corollary}[\index{well-posed}Well-Posedness of \index{posterior}Posterior] 
 \label{c:wpz}
 Under Assumption~\ref{a:dh1} there exist 
 $\Delta,c>0$
 such that, for all $\delta\in(0,\Delta)$,
       \[ \dtv(\post^y,\post_\delta^y)\leq c \delta.\]
 \end{corollary}

\section{Evaluating Approximate Posteriors}\label{sec:sbc}

We now describe a method referred to as \emph{simulation-based calibration}\index{simulation-based calibration} for evaluating approximate posterior distributions. These approximations may be obtained using Monte Carlo\index{Monte Carlo},  Markov chain Monte Carlo\index{Markov chain Monte Carlo}, or by use of the machine-learning based methods described in subsequent chapters. The following theorem, true of the exact posterior distribution, motivates the methodology:

\begin{theorem}\label{th:sbc}
    Let $f: \R^\du\to\R$ be a scalar function. Given $(u, y)\sim\Prob(u, y)$ and $N$ samples $\{u^{(n)}\}_{n=1}^N$ drawn i.i.d. from $\pi^y$, then $\operatorname{rank}(\{f(u^{(n)})\}_{n=1}^N, f(u))$ is uniformly distributed over integers $[[0, N]]$.
\end{theorem}
\begin{proof}
    The reference for the proof is provided in the bibliography Section \ref{sec:14}.
\end{proof}

Given this property of the exact posterior, we make the following Data Assumption in order to evaluate an approximate posterior.
\begin{dataassumption}\label{da:sbc}\index{Data Assumption}
    Assume that we can sample $u\sim\rho$ and evaluate the forward model $G(u)$, such that we can obtain $L$ joint samples
    \begin{equation*}
        \{(u^{(\ell)}, y^{(\ell)})\}_{\ell=1}^L,
    \end{equation*}
    where $(u^{(\ell)}, y^{(\ell)})\sim \Prob(u, y)$.
    
    Then, for each $y^{(\ell)}$, assume that we can compute an approximate posterior distribution $\pi^{y^{(\ell)},\mathrm{alg}}$, from which we can sample
    \begin{equation*}
        \{\widetilde{u}^{(\ell, n)}\}_{n=1}^N,
    \end{equation*}
    where $\widetilde{u}^{(\ell, n)}\sim\pi^{y^{(\ell)},\mathrm{alg}}$.
\end{dataassumption}

Given this Data Assumption and Theorem~\ref{th:sbc}, simulation-based calibration\index{simulation-based calibration} evaluates the algorithm that generates the approximate posteriors $\pi^{y^{(\ell)},\mathrm{alg}}$ by verifying whether $\operatorname{rank}(\{f(\widetilde{u}^{(\ell,n)})\}_{n=1}^N, f(u^{(\ell)}))$, for $\ell=1,\ldots,L$, are approximately uniformly distributed on $[[0, N]]$. If the empirical distribution of the ranks deviates significantly from uniformity, this indicates that the approximate posteriors are not well calibrated.

\section{Model Error}\index{model error}
\label{sec:ME}

The underlying concept behind this entire section is that the data $y$, which we use
to determine unknown parameter $u$, is not actually generated by the identity \eqref{eq:jc0}.
This arises, for example, when the forward model $G(\cdot)$ does not exactly represent the physical
reality that gave rise to the data; or when the forward model $G(\cdot)$ is approximated by
a computer model that does not match the forward model exactly.  Indeed, typically both of these
issues arise. In such situation the set-up suffers
from \emph{model error}\index{model error}; it is also another form of \emph{model misspecification}\index{model misspecification}.
We describe three approaches to addressing this issue. 
The reader should note that it is also possible to combine
each of the three approaches to devise more general approaches to model error.

The starting point for all three approaches is that the data $y$ arises
from noisy observation of a physical process $\Gp: \Ru \to \Ry$ so that
\begin{equation}
\label{eq:jcKOH}
 y = \Gp(u) + \eta. 
\end{equation}
The assumption is that $\Gp$ is not available to us, but that we have
a (family of) computational model(s) $\Gs$ which we can use in place of
$\Gp$ to determine $u$ from $y.$

\subsection{Representing Error in Data Space}
\label{ssec:KOH}

This first approach makes the assumption that $\Gp: \Ru \to \Ry$ is related to computer code  $\Gs: \Ru \to \Ry$ in the sense that
\begin{equation}
\label{eq:shift}    
\Gp(u)=\Gs(u)+\bepsilon
\end{equation}
for some unknown vector $\bepsilon \in \Ry$.
Furthermore, random variable $\eta$ is assumed to be drawn from a centered distribution $\nu_\bgamma(\cdot)$, known up to a parameter $\bgamma \in \R^p$.
Thus model error\index{model error} is present both because of the
unknown shift between $\Gp$ and $\Gs$ and because of the unknown parameter
in the distribution of the additive noise.

Rather than just putting a prior $\rho$ on $u$, as in Section~\ref{sec:11}, 
we  put a prior $\rho$ on $(u,\bepsilon,\bgamma).$ The inverse problem is now reformulated 
as determining the distribution $\post^y$ of  $(u,\bepsilon,\bgamma)|y.$
Combining \eqref{eq:jcKOH} and \eqref{eq:shift} we obtain
\begin{equation}
\label{eq:jcKOH2}
 y = \Gs(u) + \bepsilon+ \eta, 
\end{equation}
where $\eta \sim \nu_\bgamma.$
By applying Bayes Theorem~\ref{t:bayes} we obtain
\begin{equation}
 \post^y(u,\bepsilon,\bgamma) = \frac{1}{Z} \noise_\bgamma\bigl(y - \Gs(u)-\bepsilon\bigr)\pr(u,\bepsilon,\bgamma),
\end{equation}
where
\[
 Z = Z^y :=\int_{\Ru \times \Ry \times \R^p} \noise_\bgamma\bigl( y - \Gs(u)-\bepsilon\bigr)\pr(u,\bepsilon,\bgamma) \, du\, d\bepsilon\, d\bgamma  >0.\]

 \begin{remark}
 \label{rem:factor}
     In this setting the prior $\pr(u,\bepsilon,\bgamma)$ is typically factored as an independent
     product of priors on $u$ and on each of the hyperparameters $b$ and $\bgamma.$ 
 \end{remark}

 \begin{example}
     \label{ex:uv} 
     The parameter $\bgamma$ might, for example, be the standard deviation of an isotropic Gaussian density in $\R^k$ and then  $\noise_\bgamma(y)=\cN(y;0,\bgamma^2 I).$
 \end{example}

\subsection{Representing Error in Parameter Space}
\label{ssec:EME}
A different approach is to account for model error in parameter space.
To achieve this, parameter $u$ is assumed to be perturbed by a draw
$\vartheta$ from a random variable with distribution $r(\cdot|\beta)$ defined through an
unknown parameter $\beta \in \R^p.$ Thus we write
\begin{equation}
\label{eq:trep}
y = G(u + \vartheta) + \eta,
\end{equation}
assuming in this setting that $\eta \sim \noise$ and that $\noise$ is completely known.
We also assume that, conditional on $\beta$, parameters $(u,\vartheta,\eta)$ are independent a priori and that
prior on $u$ is $\pr.$ We then put prior $\varrho$ on $\beta$ and
apply Bayes Theorem~\ref{t:bayes} to obtain 
\begin{equation}\label{eq:bayesformula3}
 \post^y(u,\vartheta,\beta) = \frac{1}{Z} \like(y|u,\vartheta) \pr(u) r(\vartheta|\beta)\varrho(\beta),
\end{equation}
where
\[
 Z = Z^y :=\int_{\Ru  \times \Ru \times \R^p} \like(y|u,\vartheta) \pr(u) r(\vartheta|\beta)\varrho(\beta) du\, d\vartheta\,d\beta.
\]

A substantial challenge facing this approach 
is to be able to integrate out $\vartheta$ and learn only the pair $(u,\beta).$ Together $(\vartheta,\eta)$ define
a noise model, but it is not additive. Integrating out $\vartheta$ amounts to seeking
a tractable likelihood $\bbP(y|u,\beta).$
Alternatively, rather than restricting to settings where the likelihood is tractable,
it is possible to use likelihood-free methods such as approximate Bayesian computation\index{approximate Bayesian computation}. 
We will discuss likelihood-free inference in Section~\ref{sec:likelihood-based}. Here, we 
illustrate the representation of error in parameter space in the setting of a linear forward model, in a case
where $\bbP(y|u,\beta)$ is tractable.

\begin{example} For a linear forward model, $G(u) = Au$, the data model
\eqref{eq:trep} becomes
$$y = Au + (A\vartheta + \eta).$$ This can be interpreted as a standard linear forward model with the additive error $A\vartheta + \eta$. If $\vartheta$ and $\eta$ are independent zero-mean Gaussian random variables, the embedded model error corresponds to a likelihood model with inflated variance where the additional variance is related to the forward model. Then $\like(y|u,\beta)$ is determined
by
$$\bbP(y|u,\beta) = \mathcal{N}\bigl(Au, AC(\beta)A^\top + \Sigma\bigr),$$
where
$$C(\beta)=\text{Cov}(\vartheta), \quad \Sigma=\text{Cov}(\eta).$$
\end{example}

\subsection{Parameterizing the Forward Model}
\label{ssec:UTH}

A third approach to model error postulates that the physically realizable
forward map $\Gp$ is related to a class of computational models 
$\Gs:\Ru \times \R^p \to \Ry$ in the sense that, for all $u \in \Ru,$
and for some $\alpha \in \R^p,$
\begin{equation}
\label{eq:trep2}
\Gp(u) = \Gs(u,\alpha).
\end{equation}

It is assumed that $\alpha$ is not known to us. Thus the problem becomes
one of determining the pair $(u,\alpha)$ from the observation $y.$
Combining \eqref{eq:jcKOH} and \eqref{eq:trep2} suggests consideration of
the Bayesian inverse problem of determining the distribution 
of $(u,\alpha)|y$ given that
\begin{equation}
y = \Gs(u,\alpha) + \eta.
\end{equation}
Here $\eta$ again describes additive noise with known distribution $\noise.$ 
Applying Bayes Theorem~\ref{t:bayes} we obtain
\begin{equation}\label{eq:bayesformula4}
 \post^y(u,\alpha) = \frac{1}{Z} \noise\bigl(y - \Gs(u,\alpha)\bigr) \pr(u,\alpha),
\end{equation}
where
\[
 Z = Z^y :=\int_{\Ru  \times \R^p}  
 \noise\bigl(y - \Gs(u,\alpha)\bigr)\pr(u,\alpha) \, du \,d\alpha  >0.
\]
Similarly to Remark \ref{rem:factor} the prior is usually factored as an independent product
of priors on $u$ and on $\alpha.$

\section{Surrogate Modeling}
\label{sec:IPSU}

One of the earliest, and still one of the most prevalent,
uses of machine learning in the field of computational inverse problems is
surrogate\index{surrogate} modeling.
In this section we focus on this topic in the context of solving the inverse problem for $u$ given $y$,
defined by \eqref{eq:jc0}. This serves as an introduction to the use of machine
learning in inverse problems, a topic we expand upon in the next four chapters.

The key idea is to approximate $G: \R^{\du} \to \R^\dy$ with a cheap-to-evaluate surrogate\index{surrogate}, $G_\delta$, using supervised learning from Chapter~\ref{ch:SL}. Thus, to learn this approximation, we make the following assumption. 

\begin{dataassumption}\index{Data Assumption}
    \label{da:surrogate} Data is available in the form 
\begin{align}
\label{eq:datafsm}
\Bigl\{(\un, y^{(n)}) \Bigr\}_{n=1}^N,
\end{align}
where the $\{\un\}_{n=1}^N$ 
are generated \index{i.i.d.}i.i.d. from probability density function $\Upsilon \in \cP(D)$, for some $D \subseteq \R^d,$ and where
$y^{(n)}=G(\un).$
\end{dataassumption}

\begin{remark}
\label{rem:ce}
    A key point to appreciate is that, since we will use approximation of $G$ to solve the Bayesian
    inverse problem for $\pi^y$ given by Bayes Theorem~\ref{t:bayes}, an ideal choice for $\Upsilon$ is that
    it is close to $\pi^y.$ This, of course, leads to a chicken--egg issue. In practice this can be addressed
    by choosing $\Upsilon$ to have generous support, aiming to subsume that of $\pi^y;$ or by 
    generating supervised learning data in tandem with solving the inverse problem, in an iterative fashion.
\end{remark}

In Subsection~\ref{sec:LForward} we discuss the use of surrogate\index{surrogate} forward models
to speed up Bayesian inversion. Subsection~\ref{sec:LForwarda} studies the effect of approximating the forward model on the posterior. In Subsection~\ref{sec:LForward2} we comment on surrogate\index{surrogate} modeling in the context of 
Bayesian inversion in the presence of model error;
this requires a generalization of Data Assumption~\ref{da:surrogate}.

\subsection{\index{forward!model} Accelerating Bayesian Inversion}\label{sec:LForward}

Recall the \index{inverse problem}inverse problem \eqref{eq:jc0} of finding
$u$ from $y$ where
$$y=G(u)+\eta,$$
under the setting of Assumption~\ref{a:jc1} and Data Assumption~\ref{da:vb}.
If the \index{Bayesian}Bayesian approach is adopted then methods such as
\index{MCMC}MCMC, introduced in Subsection \ref{ssec:PE1} and discussed in detail in Section~\ref{sec:MCMC}, will need to be used to sample the \index{posterior}posterior. Generating each new sample typically requires evaluating the likelihood, and hence the forward model $G.$ For instance, in the Metropolis--Hastings Algorithm \ref{algMH} the likelihood needs to be evaluated to compute the acceptance probability \eqref{eq:mhaccept}. When $G$ is computationally 
expensive to evaluate, the multiple evaluations of $G$ may be prohibitively
expensive. We address
this issue by using a cheap computational surrogate\index{surrogate} $G_\delta$. While optimization\index{optimization} algorithms for MAP estimation\index{MAP estimator} often require much fewer forward model evaluations for convergence than MCMC\index{MCMC} sampling algorithms, surrogate models can also be helpful in classical optimization-based approaches to inverse problems. For instance, surrogate models can facilitate the use of gradient-based optimization algorithms (see Chapter \ref{ch:optimization}) in applications where gradients of $G$ are not available but gradients of the surrogate model $G_\delta$ can easily be computed. 

The methods described in Chapter~\ref{ch:SL} 
can be used to approximate the 
(scalar-valued) \index{likelihood}likelihood $\like$  resulting from $G$, 
by an approximation $\like_{\delta}$
with small error $\delta,$ uniformly over bounded open $D  \subset \R^d $; see the discussion
in Section \ref{sec:AP}.
In Chapter~\ref{ch:SL} we focus on learning functions taking values
in $\R;$ however all such methods can
be generalized to approximate vector-valued $G.$
This leads to uniform  approximation over $D  \subset \R^d$ of $G$ by $G_\delta$, which
in turn results in a uniform approximation $\like_{\delta}$  of the \index{likelihood}likelihood $\like$.

Such an approximation $G_\delta$
can be learned under the setting of Data Assumption~\ref{da:surrogate}.
If carefully designed, 
the \index{machine learning}machine learning approximation of the \index{likelihood}likelihood will be much
faster to evaluate than the true \index{likelihood}likelihood and, because of the
approximation properties, will result in accurate \index{posterior}posterior inference,
with errors of size $\delta.$ 
Such results rely on error bounds discussed in Section~\ref{sec:AP}. These approximations
can often be obtained with a number $N$ of evaluations
of the \index{forward!model}forward model which is orders of magnitude smaller than the number of evaluations 
required within \index{MCMC}MCMC, or
even MAP estimation in the context of learning model error (see Subsection \ref{sec:LForward2} below).
The resulting method can thus be very efficient. The next subsection describes
underpinning theory to transfer approximation of the forward model to approximation
of the posterior.

\subsection{\index{forward!model} Posterior Approximation Theorem}\label{sec:LForwarda}

We now establish that the approximate \index{posterior}posterior resulting from approximating
the \index{likelihood}likelihood is 
indeed close to the true posterior. We employ a modification of the 
\index{well-posed}well-posedness theory from Section~\ref{sec:13}.

For notational convenience we drop the dependence of the likelihood\index{likelihood} on $y$ in the remainder of this chapter.
Specifically we let
\[\like(u)=\noise\bigl(y-G(u)\bigr)\quad \text{and}\quad \like_\delta(u)=\noise\bigl(y-G_\delta(u)\bigr)\]
denote the true and approximate \index{likelihood}likelihoods. Thus we may define
the true and approximate \index{posterior}posterior distributions by 
\[\post^y(u)=\frac{1}{Z}\like(u)\pr(u)\quad \text{and}\quad \post_\delta^y(u)=\frac{1}{Z_\delta}\like_\delta(u)\pr(u);\]
here $Z, Z_\delta$ are the corresponding normalizing constants. 
To prove the closeness in Hellinger distance between the true \index{posterior}posterior and the \index{posterior}posterior with machine-learned likelihood, we will rely on the following assumption.

\begin{assumption}\label{a:dh1L}
The \index{prior}prior distribution on $u$ with density $\pr$ is supported on bounded open set $D \subset \Ru$
and the data defined by Assumption~\ref{a:jc1} and Data Assumption~\ref{da:vb}
has positive probability under the resulting joint distribution on $(u,y)$, so that $Z>0.$ Furthermore, there exist $\delta^+>0$ and $K_1,K_2 \in (0,\infty)$ such that, for all $\delta\in (0,\delta^+),$ 
	\begin{itemize}
	\item[(i)] $\sup_{u\in D}|\sqrt{\like(u)} - \sqrt{\like_\delta(u)}|\leq K_1\delta;$
	\item[(ii)] $\sup_{u\in D}(|\sqrt{\like(u)}|+|\sqrt{\like_\delta(u)}|) \le K_2.$
	\end{itemize}
\end{assumption}

\begin{remark} \label{rem:whenap}
Parts (i) and (ii) of Assumption~\ref{a:dh1L} can be verified for surrogate models  obtained using training data of the form in Data Assumption~\ref{da:surrogate}. Approximation-theoretic results for supervised learning are discussed in
Section~\ref{sec:AP}.
\end{remark}

Note also that combining (i) and (ii) gives
\begin{equation}
\label{eq:combine}
\sup_{u\in D}|\like(u) - \like_\delta(u)|\leq K_1 K_2\delta.
\end{equation}

Using Assumption~\ref{a:dh1L}
we may state the following approximation result 
for the machine-learned \index{posterior}posterior.
When combined with Lemma \ref{lemmatv}, the following theorem
guarantees that expectations computed with respect to the machine-learned
\index{posterior}posterior incur an error of order $\delta$, the magnitude of the error
in the approximate \index{likelihood}likelihood.

\begin{theorem}
\label{thm1L}
Under Assumption~\ref{a:dh1L}
there exist $\Delta,c>0$ such that, for all $\delta\in(0,\Delta)$,
      \[\dhell(\post^y,\post_\delta^y)\leq c \delta.\]
\end{theorem}

\begin{proof}
As in the proof of the \index{well-posed}well-posedness Theorem~\ref{t:wpz}, it can be shown that there exist 
$\Delta,c_1,c_2>0$
such that, for all $\delta\in(0,\Delta)$,
\begin{equation}\label{eq:boundconstants}
    |Z-Z_\delta|\leq c_1\delta\quad \text{and}\quad Z,Z_\delta \ge c_2.
\end{equation}
Next, also following proof of the \index{well-posed}well-posedness Theorem~\ref{t:wpz}, we
   decompose the total error into two contributions, reflecting respectively
the difference between $Z$ and $Z_\delta$, and the difference between 
the \index{likelihood}likelihoods $\like$ and $\like_\delta$:
      \begin{align*}
      \dhell(\post^y,\post_\delta^y)
      \leq&\ \frac{1}{\sqrt{2}}\Biggl\|\sqrt{\frac{\like\pr}{Z}}-\sqrt{\frac{\like\pr}{Z_\delta}}\Biggr\|_{L^2}+\frac{1}{\sqrt{2}}\Biggl\|\sqrt\frac{\like\pr}{Z_\delta}-\sqrt{\frac{\like_\delta\pr}{Z_\delta}}\Biggr\|_{L^2}.
      \end{align*}
      Here $L^2=L^2(D;\bbR).$ It then follows from \eqref{eq:boundconstants} that, for $\delta\in(0,\Delta)$, 
      \begin{align*}
      \Biggl\|\sqrt{\frac{\like\pr}{Z}}-\sqrt{\frac{\like\pr}{Z_\delta}}\Biggr\|_{L^2}=&\ \Biggl|\frac{1}{\sqrt{Z}}-\frac{1}{\sqrt{Z_\delta}}\Biggr|\Bigg(\int_{D} \like(u)\pr(u) \, du\Bigg)^{1/2}\\
      =&\ \frac{|Z-Z_\delta|}{(\sqrt{Z}+\sqrt{Z_\delta})\sqrt{Z_\delta}}\\
      \leq&\ \frac{c_1}{2c_2}\delta.
      \end{align*}
      And from Assumption~\ref{a:dh1L} we have that
      \[\Biggl\|\sqrt{\frac{\like\pr}{Z_\delta}} - \sqrt{\frac{\like_\delta\pr}{Z_\delta}}\Biggr\|_{L^2}=\frac{1}{\sqrt{Z_\delta}}\Bigg(\int_{D}\Bigl|\sqrt{\like(u)}- \sqrt{\like_\delta(u)}\Bigr|^2 \pr(u) \, du\Bigg)^{1/2} \leq \sqrt{\frac{K_1^2}{c_2}}\delta.\]
      Therefore
      \[\dhell(\post^y,\post_\delta^y)\leq \frac{1}{\sqrt{2}}\frac{c_1}{2c_2}\delta+\frac{1}{\sqrt{2}}\sqrt{\frac{K_1^2}{c_2}}\delta=c \delta,\]
      with $c=\frac{1}{\sqrt{2}}\frac{c_1}{2c_2}+\frac{1}{\sqrt{2}}\sqrt{\frac{K_1^2}{c_2}}$ independent of $\delta$.
      \end{proof}

      \subsection{\index{forward!model} Accelerating Bayesian Inversion in the Presence of Model Error}\label{sec:LForward2}
  We argued in Subsection~\ref{sec:LForward} for the potential use of 
surrogate\index{surrogate} modeling within
MCMC, where multiple forward evaluations are typically required.
Even when MAP\index{MAP estimator}
estimation is used, multiple evaluations of $G$ may be needed to optimize
the objective function. Furthermore, if model error\index{model error} is being jointly
learned, in the setting of Subsection~\ref{ssec:UTH}, 
then multiple MAP estimators may be
required, leading to many more evaluations.
This subsection is centered on surrogate\index{surrogate} modeling for MAP estimation when model error is also being learned.

Recall the Bayesian inverse problem to jointly estimate unknown parameter $u$ and unknown model error parameter $\alpha$ defined in \eqref{eq:bayesformula4}. 
Thus we have access to a family of computational models $\Gs(\cdot\,; \alpha): \Ru \to \Ry$ parameterized by $\alpha \in \R^p.$ Our data assumption is as follows:

\begin{dataassumption}\index{Data Assumption}
    \label{da:surrogate2} Data is available in the form 
\begin{align}
\label{eq:datafsm2}
\Bigl\{(\un, \alpha^{(n)}, y^{(n)}) \Bigr\}_{n=1}^N,
\end{align}
where the $\{(\un, \alpha^{(n)})\}_{n=1}^N$ 
are generated \index{i.i.d.}i.i.d. from probability density function $\Upsilon \otimes q \in \cP(D \times \R^p)$, for some $D \subseteq \R^d,$ and where
$y^{(n)}=\Gs(\un,\alpha^{(n)}).$
\end{dataassumption}

From this data we may learn a cheap surrogate\index{surrogate} for $\Gs,$ using
the techniques of Chapter~\ref{ch:SL}. The MAP estimation problem,
in the context of Subsection~\ref{ssec:UTH}, requires minimization, over the pair 
$(u,\alpha)$, of
$$\J(u,\alpha)=-\log \noise\bigl(y - \Gs(u,\alpha)\bigr)-\log \pr(u,\alpha).$$
Some methods for this problem work by alternating minimization over
$u$ and over $\alpha$; a prototypical such method is to iterate, for $\ell$
until convergence, starting from initial $\alpha^0 \in \R^p:$
    \begin{subequations}\label{eq:alternatingopt}
        \begin{align}
        u^{\ell +1 }&\in\argmin_{u \in \Ru} \J \bigl(u,\alpha^{\ell} \bigr),\\
        \alpha^{\ell+1}&\in \argmin_{\alpha \in \R^p} \J \bigl(u^{\ell+1},\alpha \bigr).
    \end{align}
    \end{subequations}
Generalizations of this alternating strategy are possible; for example a finite number
of steps (possibly $\ell$-dependent) of gradient descent may be substituted for the two minimizations.
For each $\ell,$ multiple evaluations of $\Gs(\cdot,\cdot)$ may be required
and use of a surrogate\index{surrogate} may make the method more efficient. 

Finally, we point out that similar ideas to accelerate Bayesian inversion in the presence of model error apply when computing posterior summaries via sampling rather than a MAP estimator via optimization. For sampling the joint posterior distribution over the pair $(u,\alpha)$, a natural approach is given by the Gibbs sampler\index{Gibbs sampler} from Subsection \ref{ssec:Gibbs}. As in \eqref{eq:alternatingopt}, the Gibbs sampler would then alternate between sampling the conditional of $u$ given $\alpha$ and the conditional of $\alpha$ given $u.$ Such a Gibbs sampler would also directly benefit from use of cheap surrogate $\Gs.$

\section{Bibliography}\label{sec:14}

For the Bayesian approach to inverse problems\index{Bayesian!inverse problem} see \cite{kaipio2006statistical,tarantola2005inverse,stuart2010inverse,dashti2013bayesian,lasanen2012non,lasanen2012nonb,nickl}. In \cite{knapik2011bayesian} the use of Data Assumption~\ref{da:vbx} underlies the study of posterior consistency for linear inverse problems. In the infinite-dimensional setting considered there, the two motivations for Data Assumption~\ref{da:vbx} given in
Remark \ref{rem:avdw} overlap: the regularity of $u^{\mathsf{true}}$ is such that it is not a draw from
the prior, with probability one, and the effect of this on rates of contraction is studied.
For classical optimization-based\index{optimization} 
approaches to \index{inverse problem}inverse problems, 
we refer to the books and lecture notes \cite{tikhonov1977solutions,engl1996regularization,vogel2002computational,bal2012introduction,miller2003fundamentals}.
The concept of \index{MAP estimator}MAP estimation, which links probability to
\index{optimization}optimization, is discussed in the books  
\cite{kaipio2006statistical,tarantola2005inverse,dashti2013map}.
For generalizations see the papers \cite{helin2015maximum,agapiou2017sparsity,lambley2023order}.
Monte Carlo\index{Monte Carlo} methods are a broad class of approaches for characterizing posterior distributions using a sum of random Dirac measures\index{Dirac measure}; see~\cite{liu2008monte, robert2013monte, sanz2024first} for textbooks on the theoretical and computational aspects of these methods. Quasi Monte Carlo (QMC)\index{QMC} methods blend ideas from deterministic quadrature
with the high dimensional benefits of Monte Carlo; see \cite{caflisch1998monte,owen2013monte}. 
While direct Monte Carlo sampling is possible in some settings, Markov chain Monte Carlo (MCMC)\index{Monte Carlo!Markov chain}\index{MCMC} is a variant that is used in practice when it is not possible to sample exactly from the posterior. 
A general introduction to MCMC\index{MCMC} may
be found in~\cite{gamerman2006markov}. Sequential Monte Carlo (SMC)\index{SMC} \cite{del2006sequential,chopin2020introduction} is a methodology
which proceeds by introducing an interpolation from prior to posterior and incrementally steps 
through a sequence of inverse problems to connect them \cite{beskos2015sequential,beskos2015sequentialb,kantas2014sequential}. We provide a brief overview of MCMC in Section \ref{sec:MCMC} and discuss the use of learned transport maps to accelerate MCMC in Section \ref{sec:lmcmc}. Additional bibliographical remarks on MCMC can be found in Sections \ref{sec:tbib} and \ref{sec:OB}.

The fact that the conditional mean\index{conditional mean} minimizes the mean-square error is a classic result in probability; see, for example, \cite{williams_probability_1991}. A characterization of the class of 
deterministic scoring rules\index{scoring rule!deterministic} which the conditional mean\index{conditional mean} minimizes is given in \cite{banerjee_optimality_2005}.

The stability and \index{well-posed}well-posedness
of \index{Bayesian!inverse problem}Bayesian \index{inverse problem}inverse problems was
first studied in \cite{marzouk2009stochastic}, using 
the Kullback--Leibler divergence (see Definition \ref{def:KL}) and focusing on perturbations in the data.
The articles \cite{stuart2010inverse,dashti2013bayesian} 
study similar stability and \index{well-posed}well-posedness results
in the Hellinger metric, as we do here; application to perturbations arising from numerical
approximation of the forward model $G$ may be found in \cite{cotter2010approximation}.
Related results on stability and \index{well-posed}well-posedness, 
but using other distances and divergences, 
may be found in \cite{latz2020well}.
The papers \cite{hosseini2017well,hosseini2017wellb} discuss generalizations of the \index{well-posed}well-posedness theory to various classes of specific 
\index{Gaussian!non-}non-Gaussian \index{prior}priors. Simulation-based calibration\index{simulation-based calibration} was introduced in \cite{talts_validating_2020}, and the proof of Theorem~\ref{th:sbc} is presented therein.

The approach to model error outlined in Subsection~\ref{ssec:KOH} was introduced
in \cite{kennedy2001bayesian}.
The embedded model error framework of Subsection~\ref{ssec:EME} was introduced in~\cite{sargsyan2019embedded}; for an introduction to approximate
Bayesian computation see \cite{sisson2018handbook}. An example of
the approach to model error described in Subsection~\ref{ssec:UTH} 
may be found in \cite{cotter2012variational}.
A parameterized forward model may be a coarse-scale model where unresolved small-scale physical processes are described by parameters $\alpha$. (For example, $\alpha$ may describe cloud cover in a climate model $\Gs$.) More examples will be provided in Chapter~\ref{ch:DAML}.
 Methodologies to identify parameters of structural error models (e.g., within computational models for dynamical processes), often using indirect data, are described in~\cite{wu2023learning, levine_framework_2022}.

Using emulators\index{emulator} to speed up forward model evaluations, for
example in the context of \index{likelihood}likelihood evaluation in \index{Bayesian!inversion}Bayesian inversion, was first introduced as a systematic methodology in \cite{sacks1989design},
and taken further in the realm of \index{Bayesian}Bayesian \index{model error}model error estimation
in \cite{kennedy2001bayesian}. The paper \cite{stuart2018posterior}
studies the use of \index{Gaussian!process}Gaussian processes for emulation, and derives
error bounds quantifying the effect of emulation error 
on the \index{posterior}posterior. The methodology is developed for a range of applications 
in the geosciences in \cite{cleary2021calibrate}; in particular that paper addresses the issue
of how to determine $\Upsilon$ in tandem with solution of the inverse problem as discussed
in Remark \ref{rem:ce}.
A specific application of the idea in climate science may 
be found in \cite{dunbar2021calibration}. Data-driven discretizations of \index{forward!model}forward models for \index{Bayesian!inversion}Bayesian inversion are studied in \cite{bigoni2020data}.
A recent approach to directly learning parameter to solution (forward) and solution to parameter (inverse)
surrogate\index{surrogate} maps may be found in \cite{vadeboncoeur2023fully}. Surrogate models are also useful in optimization approaches to inverse problems. In this regard, \index{Bayesian optimization}Bayesian optimization techniques \cite{frazier2018tutorial}
replace an expensive-to-evaluate objective function with a surrogate model, and gradually improve this surrogate model along the optimization process by judiciously acquiring new training data. We refer to \cite{kim2024enhancinggaussianprocesssurrogates,Bayesoptgraph} for a discussion of Bayesian optimization for MAP estimation in Bayesian inverse problems.

An important aspect of learning forward models is to choose the supervised training data over which we would like to find an accurate surrogate\index{surrogate} model. While we would ideally like to be accurate over the support of the posterior, finding such a surrogate\index{surrogate} model requires being able to sample the posterior. Instead, it is common to seek a surrogate\index{surrogate} model that is accurate over the support of the prior. To address this, several recent methods use approximate posteriors to iteratively refine the approximation of the surrogate\index{surrogate} model~\cite{cleary2021calibrate, helin2023introduction}.

\chapter{\Large{\sffamily{Variational Inference}}} \label{ch:VI}

A problem has a variational formulation\index{variational!formulation}
if its solution can be written as the minimizer of an objective\index{objective} function.
The variational formulation of any problem in mathematics is useful because it
opens the door to computational methods. Such computational methods seek to minimize the objective
function over a strict subset of the space on which the original variational problem is posed.
If the strict subset is chosen judiciously (this is a problem-dependent
choice) then the resulting computational methodology is both tractable and yields
a useful and interpretable 
approximation of the solution of the original variational problem
posed on the whole space.

This chapter is devoted to formulating Bayes Theorem \ref{t:bayes}
variationally and using this formulation as the basis for approximate inference. 
We continue to work under Data Assumption \ref{da:vb}.
Under this data assumption, and given a prior distribution on $u$, the posterior distribution on $u|y$ is defined by Bayes Theorem \ref{t:bayes}.
In Section \ref{sec:vbay} we show how this theorem can be formulated
variationally as a minimization problem over the set of all probability density 
functions. This leads to the core methodology of variational inference\index{inference!variational}. In
Section \ref{sec:vvar} we present two canonical choices of subsets of all probability density functions over which to approximate the posterior: the mean-field\index{mean-field} and Gaussian\index{Gaussian} subsets. In Section \ref{sec:VIprop} we discuss
some key properties of variational inference, studying other choices of optimization
problem which yield the posterior as a minimizer, and linking to the evidence lower bound\index{evidence! lower bound}. We conclude the chapter, in Section \ref{sec:vbib}, with bibliographical remarks.

\section{Variational Formulation of Bayes Theorem}
\label{sec:vbay}

Our starting point in this section is Bayes Theorem \ref{t:bayes}. Because
 dependence on $y$ is not central to our discussions here, we write the posterior
as $\post:=\post^y.$
A useful formulation of Bayes Theorem arises from viewing the posterior as the solution of an optimization problem over the space of probability density functions. In what follows we recall from the Preface that $\cP:=\cP(\R^d)$ denotes the space 
of all probability density functions\index{probability!density function} over $\R^d.$ We define an objective
$\F: \cP \to \R$ which, when evaluated at a candidate density $q$,  quantifies the discrepancy between the posterior density $\post$ and $q$. 
The objective is constructed to be minimized at $q=\post$. By using
Bayes Theorem \ref{t:bayes} the objective function can be rewritten in terms
of the prior\index{prior}, the likelihood\index{likelihood}
and the normalization\index{normalization constant} 
constant.

One natural class of objective functions $\F$ is defined by the family
of $\mathsf{f}$-divergences\index{divergence!$\mathsf{f}$} $\D_\mathsf{f}(q\|\post):$
$$\F(q):=\D_\mathsf{f}(q\|\post) = \int_{\Ru} \mathsf{f}\left(\frac{q(u)}{\post(u)}\right)\post(u)\, du.$$
See Subsection~\ref{ssec:fdiv} for the definition and properties of this family of
distance-like objects defined on the space of probability densities. 
By the properties of all divergences, detailed at the start of Section \ref{sec:div}, we deduce that $\F(q)$
is minimized at $q=\post$  and that this is the unique minimizer.
A particularly useful choice arises from setting $\mathsf{f}(t)=t\log(t),$ 
leading to definition of the objective
\begin{equation}
\label{eq:KLU}    
\F(q):=\dkl(q\|\post) = \int_{\Ru} \log \left(\frac{q(u)}{\post(u)}\right) q(u) \, du,
\end{equation}
the Kullback--Leibler (KL) divergence\index{divergence!Kullback--Leibler}. We discuss
the reason for making this particular choice in Remark \ref{rem:2.3} below.
Having chosen to work with the KL divergence\index{divergence!Kullback--Leibler} \eqref{eq:KLU}, 
the following theorem establishes a variational formulation of 
Bayes Theorem\index{Bayes Theorem!variational} \ref{t:bayes}.

\begin{theorem}
\label{t:VariationalForm_Bayes} Consider the Bayesian inverse problem defined
by prior $\pr$ and likelihood $\like$ given by \eqref{eq:like}, 
under Data Assumption \ref{da:vb}. Define $\J: \cP \to \R$ by 
\begin{subequations}
\label{eq:KLobj_minusZ}
\begin{align} 
\J(q) &= \dkl(q\|\pr) - \mathbb{E}^q[\log \like(y|\cdot)],\\
\qopt &\in \argmin_{q \in \cP} \J(q).
\end{align}
\end{subequations}
Then $\qopt=\post,$ the posterior distribution.
\end{theorem}

\begin{proof}
By the properties of all divergences, detailed at the start of Section \ref{sec:div},
it follows that $\F(\cdot)=\dkl(\cdot\, \|\post)$ is non-negative for all inputs from the set of probability density functions and has a unique minimizer over $\cP$ at $q=\pi.$ 

Using the form of the   density, we see that
\begin{subequations}
\label{eq:KLobj}
\begin{align} 
\F(q) &= \mathbb{E}^q[\log q - \log \pr - \log \like(y|\cdot) + \log Z] \\
&= \dkl(q \|\pr) - \mathbb{E}^q[\log\like(y|\cdot)] + \log Z\\
&= \J(q) + \log Z. 
\end{align}
\end{subequations}
Because $Z$ is a constant with respect to $q$, the minimizer of $\F(\cdot)$, which
is unique, is equivalent to the minimizer of $\J(\cdot)$ and the proof is complete.
\end{proof}

\begin{remark}
\label{rem:2.3}
The specific choice \eqref{eq:KLU} of $\mathsf{f}$-divergence is convenient as then the objective $\mathsf{F}(q)$ agrees, up to an additive constant independent of $q$, with the objective $\mathsf{J}(q)$ in (\ref{eq:KLobj_minusZ}a), whose definition does not involve the normalization constant of the posterior. This fact is established in equation \eqref{eq:KLobj} while Remark \ref{rem:yifan} discusses why KL is the \emph{only}  $\mathsf{f}$-divergence for which this property holds. 
In Subsection \ref{ssec:modemean}, we discuss reversing the order of $q$ and $\post$ in the arguments of the KL divergence,
corresponding to making the choice $\mathsf{f}(t)=-\log(t).$  

Note that, using the loss function \eqref{eq:loss}, we may write
\begin{align} 
\label{eq:KLobj_minusZ2}
\J(q) &= \mathbb{E}^q[\loss]+ \dkl(q\|\pr).
\end{align}
Using this formulation of $\J(\cdot)$ we comment on the structure of the minimization
problem \eqref{eq:KLobj_minusZ} in relation to the MAP estimation\index{MAP estimator} problem presented in Section \ref{sec:112}. The MAP problem requires minimization of \eqref{eq:MAPO}.
Comparison with \eqref{eq:KLobj_minusZ2} shows that in both cases the objective involves two terms that balance the fit to the data (the first term) with properties of the prior distribution (the second term). Furthermore the first term in \eqref{eq:KLobj_minusZ2} is the
expectation of the first term in \eqref{eq:MAPO}. Note that minimization of~\eqref{eq:KLobj_minusZ2} is more general than minimization of \eqref{eq:MAPO}: the former provides a probability density function rather than the point estimator provided by the latter.

Both the MAP objective function \eqref{eq:MAPO} and the probabilistic objective function \eqref{eq:KLobj_minusZ2} are defined by
the prior and likelihood, and avoid calculation of the normalization constant\index{normalization constant} -- see Remark \ref{rem:nonorm}.
As discussed in Remark \ref{rem:nonorma} avoiding calculation of the normalization constant
is also a key feature of the  Metropolis--Hastings sampling algorithm.
\end{remark}

\section{Canonical Approaches to Variational Inference\index{inference!variational}}
\label{sec:vvar}

In practice, characterizing the posterior distribution is difficult in general.
One approach to doing so is to approximate the posterior using the variational formulation. We replace the set $\cP$ in~\eqref{eq:KLobj_minusZ} by a tractable class of probability measures $\mathcal{Q} \subset \cP$ for the purpose of computations and for the
purpose of revealing an explicit, interpretable form. The variational inference\index{inference!variational}
problem is then given by solving the following optimization problem:
\begin{equation} \label{eq:VariationalForm_Bayes}
\qs \in \argmin_{q \in \mathcal{Q}} \J(q).
\end{equation}

Noting the form (\ref{eq:KLobj_minusZ}a) of $\J(q)$ this optimization problem can be implemented, approximately,
provided we are able to evaluate the prior 
$\pr(u)$ and the likelihood $\like(y|u)$ for any
pair $(u,y) \in \R^\du \times \R^k.$
However, typically some quadrature will be needed, in particular 
to evaluate expectations against $q(\cdot\,\theta)$ from some parameterized class $\mathcal{Q}.$
In the next two subsections we outline two canonical classes of tractable distributions $\mathcal{Q}$: in Subsection \ref{ssec:mf} the products of one-dimensional marginal densities;
and in Subsection \ref{ssec:gauss} multivariate Gaussians. The former case is known as \emph{mean-field variational inference}\index{inference!mean-field variational} and the latter as \emph{Gaussian variational inference.}\index{inference!Gaussian variational}

\subsection{Mean-Field\index{mean-field} Family}
\label{ssec:mf}
Mean-field inference\index{inference!mean-field variational} works with a family $\cQ$ of distributions that factorize into a product of independent components. While this choice does not represent dependencies among its variables, it is often chosen because of the efficiency of methods for finding $\qs$ in this case, and because of the ease
of computing marginal properties once $\qs$ is determined.

\begin{definition}
\label{d:qmf} 
The probability density function $q$ of a random variable $u \in \R^d$ is in the \emph{mean-field}\index{mean-field} family $\cQ$ if the coordinates of $u$ are independent. That is, $q$ can be written in the form
$$ q(u) = \prod_{i=1}^d q_i(u_i),$$
where $q_i \in \cP(\R)$ denotes the probability 
density function of the $i$th coordinate $u_i\in \R$ of $u \in \R^d$.
\end{definition}

With this choice of $\cQ$ the solution of the optimization problem in~\eqref{eq:VariationalForm_Bayes} may be approached by using the
following consistency equations, relating the marginal densities
$\{q_i\}_{i=1}^d$ to one another. 
We employ the notation $u_{-i} \in \R^{d-1}$ described in the Preface.

\begin{proposition} Let $\cQ$ be defined as in Definition \ref{d:qmf}.
Then the optimal 
$\qs(u) = \prod_{i=1}^d \qs_i(u_i)$
solving optimization problem \eqref{eq:VariationalForm_Bayes} satisfies
$$\qs_i(u_i) \propto \exp \left( \int_{\R^{d-1}} \log \post(u) \prod_{j \neq i} \qs_j(u_j) \, du_{-i} \right) \quad \text{ for } i=1,\dots,d.$$
\end{proposition}

\begin{proof} Recall that minimizing $\J(\cdot)$ is equivalent to minimizing
$\dkl(\cdot\|\post)$. The KL divergence $\dkl(q\|\post)$ for a density $q$ in the mean-field family is given by
\begin{align*}
  \dkl(q\|\post) &= \int_{\Ru} \log \left(\prod_{i=1}^d q_i(u_i) \right) \prod_{i=1}^d q_i(u_i) \, du - \int_{\Ru} \log \post(u) \prod_{i=1}^d q_i(u_i) \, du \\
&= \sum_{i=1}^d \int_{\R} \log q_i(u_i) q_i(u_i) \, du_i  - \int_{\Ru} \log \post(u) \prod_{i=1}^d q_i(u_i) \, du,
\end{align*}
where in the last line we used that each integrand only depends on $u_i$ and each $q_i$ is a probability density function that integrates to $1$. Taking the first variation of the KL divergence with respect to each $q_i,$ we have
$$\frac{\delta }{\delta q_i} \dkl(q\|\post) = 1 + \log q_i(u_i) - \int_{\R^{d-1}} \log \post(u) \prod_{j \neq i} q_j(u_j) \, du_{-i}.$$
Here, notice that including a Lagrange multiplier to enforce the normalization that $q_i$ integrates to $1$ changes the right-hand side by an additive constant independent of $u_i$, which is absorbed into the proportionality constant.
Setting the first variation equal to zero and rearranging the terms gives us the un-normalized form for each marginal density $q_i^\star$ at any critical point of $\J(\cdot)$
over $\cQ.$
\end{proof}

\begin{remark}
    We note that although the preceding proposition is formulated in terms of
    the posterior density, its statement remains unchanged if $\pi(u)$ is replaced
    by $\pr(u)\like(y|u)$; this simply changes the constant of proportionality. Thus, to use it,
    we do not need to know the normalization constant $Z$ in Bayes Theorem \ref{t:bayes}.
\end{remark}

The proposition defines a set of coupled equations that must be satisfied by the optimal marginal densities in terms of the other marginal densities. Moreover, it prescribes how to set $q_i$ when keeping all other coordinates fixed. Thus, a natural approach to find the minimizer is to perform coordinate updates on each marginal. This algorithm\index{algorithm} is known as \emph{coordinate-ascent variational inference} (CAVI).\index{inference!coordinate-ascent variational} \emph{Ascent} because the method is traditionally formulated
in terms of maximizing the $\ELBO$ functional (see Definition \ref{d:elbo}) rather than minimizing the KL divergence; we prefer, for consistency with other optimization problems
in the notes, to formulate it in terms of minimizing $\J$, and hence \emph{descent.} 

The updates are thus, sequentially for iteration index $\ell \in \bbZ^+$ until convergence, and for 
$i=1,\dots,d\, ,$
\begin{equation}
\label{eq:updates}
q^{\ell+1}_i(u_i) \propto \exp \left( \int_{\R^{d-1}} \log \post(u) \prod_{j < i} q^{\ell+1}_j(u_j)
 \prod_{j > i} q^{\ell}_j(u_j) \, du_{-i} \right).
\end{equation}

\begin{algorithm}
\caption{Coordinate-Ascent Variational Inference\index{inference!coordinate-ascent variational}\index{algorithm}}
\begin{algorithmic}[1]
\STATE {\bf Input}: Density $\post$ known up to normalizing constant. Initialization $q^0.$ Number $L$ of iterations. \\
    \STATE For $\ell=0, 1,2, \ldots, L-1,$ compute $q^{\ell+1}$ from $q^{\ell}:$ 
    \STATE \hspace{1cm}Update $q_i^{\ell+1}$ for $i = 1, \ldots, d$ using \eqref{eq:updates}.
\STATE {\bf Output}: Approximation $q^L$ of density $\post.$
\end{algorithmic}
\end{algorithm}

\begin{remark}
    The mean-field methodology has the computational advantage of reducing the inference problem
    in (potentially high) dimension $d$ to one of $d$ independent one-dimensional inference problems.
    As such it can potentially accurately learn marginal information on specific coordinates; but it cannot
    learn correlations. In the next subsection we work in a different subset $\cQ$, the set of Gaussian
    measures, in which it is possible to approximate correlation information.
\end{remark}

\begin{remark}
    We make several remarks about the practical computation of the expectation in \eqref{eq:updates}. In the case that the conditional distributions $\Prob(u_i | u_{-i}, y)$ are in the exponential family\index{exponential family}, the update \eqref{eq:updates} can be computed in closed form, facilitating the scalability of the method. See the Bibliography Section \ref{sec:vbib} for more details.
In other cases, if the one-dimensional marginals $q_i$ are easy to sample from, the expectation can be computed via Monte Carlo. That is, at iteration $\ell$ and for component $i$, obtain samples $\{u_j^{(n)}\}_{n=1}^N$, where $u_j^{(n)}\sim q_j^{\ell + 1}$ for $j < i$ and $u_j^{(n)}\sim q_j^{\ell}$ for $j > i$. Then, the update \eqref{eq:updates} can be approximated as 
    \begin{equation*}
        q^{\ell+1}_i(u_i) \propto \exp \left(\int_{\R^{d-1}} \log \post(u) \prod_{j < i} q^{\ell+1}_j(u_j) \prod_{j > i} q^{\ell}_j(u_j) \, du_{-i}\right) \approx \exp\left(\frac{1}{N}\sum_{n=1}^N\log\post(u^{(n)})\right),
    \end{equation*}
    where $u^{(n)} = (u_1^{(n)},\ldots,u_i,\ldots,u_d^{(n)})$. However, note that this method may require many evaluations of the unnormalized density, which will be costly when the likelihood evaluations are expensive.
\end{remark}

\subsection{Gaussian\index{Gaussian} Distributions}
\label{ssec:gauss}

Another tractable variational approach is to seek an approximate distribution $\qs$ 
from within a parametric family $\cQ$ of probability density functions. In many applications, interest
is focused on ultimately estimating the first two moments of $\post$; in
this context a natural family to consider is Gaussian distributions. 
In this subsection we derive optimality conditions for the solution to the resulting optimization problem.  
To find a Gaussian approximation, we seek to minimize $\dkl(\cdot\|\post)$ over the set of distributions 
\begin{subequations}
\label{eq:GVI}
\begin{align}
\cQ :=& \bigl\{q \in \cP: q= \mathcal{N}(m,\Sigma), (m, \Sigma) \in \mathfrak{C}  \bigr\},\\
\mathfrak{C} :=& \bigl\{ (m,\Sigma) \in \R^d \times \bbR^{d \times d}_{\textrm{sym}, >} \bigr\}.
\end{align}
\end{subequations}

\begin{proposition} \label{prof:VI_Gaussians} 
Let $\cQ$ be defined as in \eqref{eq:GVI}.
Then the optimal  
$\qs=\mathcal{N}(\ms,\Sigmast)$ solving \eqref{eq:VariationalForm_Bayes} satisfies
\begin{equation} \label{eq:GaussianVI_objective}
(\ms,\Sigmast) \in \argmin_{(m,\Sigma)\in \mathfrak{C}} \Bigl(\mathbb{E}^q[-\log \post(\cdot)] - 
\frac{1}{2}\log\det(\Sigma)\Bigr),  
\end{equation}
where the expectation is with respect to $q= \mathcal{N}(m,\Sigma).$
\end{proposition}

\begin{proof} Recall that $\qs$ minimizes $\dkl(\cdot\|\post)$ over $\cQ.$
Recall further that the density $q(u)$ of $\mathcal{N}(m,\Sigma)$ is given by
\begin{equation}
    \label{eq:q2}
    q(u)=\frac{1}{(2\pi)^{d/2} \det(\Sigma)^{1/2}}\exp\Bigl(-\frac12|u-m|_{\Sigma}^2\Bigr).
\end{equation}
We may rewrite any expectation under $q$ in terms of a standard Gaussian random variable $\xi \sim \mathcal{N}(0,I_d)$. Indeed, using the relation $u = m + \Sigma^{1/2}\xi$,  it may be shown that
$$\int_{\Ru} \log q(u) \, q(u) \, du  = -\frac{d}{2}-\frac{d}{2}\log(2\pi) - \frac{1}{2}\log\det(\Sigma).$$
Since
    \begin{align*}
      \dkl(q \|\post)  = \int_{\Ru} \log q(u) \, q(u) \, du - \int_{\Ru} \log \post(u) \, q(u) \, du ,\\
    \end{align*}
the desired result follows.
\end{proof}

\begin{remark}
The two terms in the objective function \eqref{eq:GaussianVI_objective} can be interpreted as having a competitive 
behavior and as regularizing the MAP perspective. The first term is minimized by a Gaussian distribution with
zero variance, centered at the MAP estimator\index{MAP estimator} of $\post$, 
the point of highest posterior density: a Dirac measure\index{Dirac measure} at the posterior mode\index{mode}. 
The second term, however, approaches positive infinity if the variance of any marginal approaches zero. 
Hence, the second term in the objective regularizes the optimization problem to ensure the 
approximating Gaussian is not degenerate in any direction.  
\end{remark}

\begin{proposition} Let $(\ms,\Sigmast) \in \mathfrak{C}$ solve the Gaussian variational inference
\index{inference!Gaussian variational} problem defined by Proposition~\ref{prof:VI_Gaussians}. The solution satisfies the first-order optimality 
conditions\footnote{Use of $\nabla$ and $D^2$ here denotes derivatives with respect to $u$, then evaluated
at $u=\ms + (\Sigmast)^{1/2}\xi.$}
$$\mathbb{E}^\xi[\nabla \log \post(\ms + (\Sigmast)^{1/2}\xi)] = 0, \quad \mathbb{E}^\xi[D^2 \log \post(\ms + (\Sigmast)^{1/2}\xi)] = -(\Sigmast)^{-1},$$
where $\xi \sim \mathcal{N}(0,I_d).$ 
\end{proposition}

\begin{proof} 
Throughout the proof, use of $\nabla$ or $D^2$ without a subscript is to be taken to denote
derivative(s) with respect to variable $u;$ in addition, $q$ denotes the Gaussian density 
$\mathcal{N}(m,\Sigma)$ given in \eqref{eq:q2}.  The first-order optimality condition for 
the optimization problem in~\eqref{eq:GaussianVI_objective} is given by
$$\nabla_{(m,\Sigma)} \left. \left[\mathbb{E}^q[-\log \pi(u)] - \frac{1}{2}\log\det(\Sigma) \right] \right|_{(\ms,\Sigmast)} = 0.$$
The gradients of the objective with respect to $m$ and $\Sigma$ are given by
\begin{align*}
\nabla_m &\left[\int_{\Ru} -\log \pi(u) q(u) \, du - \frac12 \log\det(\Sigma)\right] = \int_{\Ru} -\log \pi(u) \nabla_m q(u) \, du, \\
\nabla_\Sigma &\left[ \int_{\Ru} -\log \pi(u) q(u) \, du - \frac12 \log\det(\Sigma) \right] =  \int_{\Ru} -\log \pi(u) \nabla_{\Sigma} q(u) \, du - \frac12 \Sigma^{-1}.
\end{align*}
From the symmetry of the Gaussian density with respect to $u$ and $m$, we have 
\begin{align*}
    \nabla_m q(u) &= -\nabla q(u),\\
    \nabla_\Sigma q(u) &= q(u)\nabla_\Sigma \log q(u) \\
    &= -q(u)\frac{1}{2}[\Sigma^{-1} -  \Sigma^{-1}(u-m)(u-m)^\top\Sigma^{-1}] \\
    &= \frac{1}{2} D^2 q(u).
\end{align*} 
Substituting these expressions above and applying integration by parts gives us 
\begin{align*}
\int_{\Ru} -\log \pi(u) \nabla_m q(u) \, du &= \int_{\Ru} \log \pi(u) \nabla q(u) \, du = -\int_{\Ru} \nabla \log \pi(u) q(u) \, du,\\
\int_{\Ru} -\log \pi(u) \nabla_\Sigma q(u) \, du &= -\frac{1}{2}\int_{\Ru} \log \pi(u) D^2 q(u) \, du = -\frac{1}{2} \int_{\Ru} D^2 \log \pi(u) q(u) \, du.
\end{align*}
Setting the two gradients equal to zero and re-writing the expectations in terms of a standard Gaussian random variable $\xi \sim \mathcal{N}(0,I_d)$ using the relation $u = m + \Sigma^{1/2}\xi$ for a positive definite covariance $\Sigma$ gives us the result.
\end{proof}

\begin{remark} If $-c^- I_d \preceq \nabla^2 \log \pi(u) \preceq -c^+ I_d$ for all $u \in \R^d$, then the optimal solution satisfies $\frac{1}{c^+} I_d \preceq \Sigmast \preceq \frac{1}{c^-} I_d$. 
\end{remark}

\begin{example} \label{ex:GaussianVI_IP}
We work in the setting of the inverse problem defined in Example \ref{ex:loss-l2}: 
the observational noise\index{observational noise} is $\noise = \Nc(\zerovct ,\mtx \Gamma)$, for positive definite covariance $\Gamma$, 
and the \index{prior}prior is a centered Gaussian
\(\pr  = \Nc(\zerovct ,\mtx \Cpn)\), where $\Cpn$ is positive definite.
We consider approximate variational inference in the Gaussian class
and seek $q=\Nc(m,\Sigma).$ In this setting the negative log likelihood has the form, up to an irrelevant additive constant,
$$-\log \like(y|u) = \frac{1}{2} |\vct y-\vct G(\vct u)|_{\mtx \Gamma}^2.$$
Example \ref{ex:klg} shows that 
\begin{align*}
\dkl(q\|\postr) &= \frac{1}{2}\left(|m|^2_{\Cpn}+\text{Tr}(\Cpn^{-1}\Sigma) - d + \log\left(\frac{\det\Cpn}{\det\Sigma}\right) \right).
\end{align*}
Combining these two facts, abusing notation and writing $\J(q)=\J(m,\Sigma)$ and ignoring the irrelevant additive constant,
we find from Theorem \ref{t:VariationalForm_Bayes} that
$$\J(m,\Sigma)=\mathbb{E}^{\xi \sim \Nc(0,I)} \frac12|y-G(m+\sqrt{\Sigma}\xi)|^2_{\mtx \Gamma}+\frac{1}{2}\left(|m|^2_{\Cpn}+\text{Tr}(\Cpn^{-1}\Sigma) - d + \log\left(\frac{\det\Cpn}{\det\Sigma}\right) \right).$$
This should be compared with the objective function in \eqref{eq:Phillips} defining the MAP estimator for the same
inverse problem. Indeed it may be viewed as a form of regularization of that problem.
\end{example}

\section{Properties of Variational Inference\index{inference!variational}}
\label{sec:VIprop}

\subsection{Mode-Seeking Versus Mean-Seeking Variational Inference}
\label{ssec:modemean}

 In the preceding section we seek an approximation to the posterior distribution 
 $\pi$ by minimizing, over $\mathcal{Q} \subset \mathcal{P}$, the
 functional $$q \mapsto \dkl(q\|\pi) = \mathbb{E}_q[\log(q/\pi)];$$ this
 is sometimes referred to as the reverse KL divergence. Using the reverse KL divergence as an objective is computationally convenient in the setting where we can evaluate the likelihood and the prior, but 
 we do not necessarily know the normalization constant, nor do we necessarily have the ability
 to easily sample from the posterior. In comparison, the forward KL divergence 
 $$q \mapsto \dkl(\pi\|q) = \mathbb{E}_\pi[\log(\pi/q)]$$ 
 is more convenient for optimization over $\mathcal{Q}$ when we can sample from $\pi$, but do not necessarily have access to an analytical expression for the target density.

In addition to computational considerations, the two choices of KL divergence lead to different behavior  when minimizing over a subset of probability measures. The reverse KL favors approximate distributions where $\log(q/\pi)$ is small in regions of high probability of $q$, that is when $q \approx \pi$ where $q$ is large, or in regions where $q$ is much smaller than $\pi$. As a result, the minimizers for $q$ tend to fit one mode of multi-modal distributions $\pi$, but miss the other modes. Hence, minimizing the reverse KL is also known as \emph{mode-seeking}\index{mode seeking}. In contrast, the minimizers of the forward KL favor approximate distributions $q$ where $\log(\pi/q)$ is small in regions of high probability of $\pi$. This occurs by having $q$ be non-zero everywhere in the support of $\pi$ so the denominator in the log does not approach zero. Placing mass everywhere in the support of $\pi$ leads to a \emph{mean-seeking}\index{mean seeking} behavior, where the minimizer for $q$ corresponds to matching the moments of the target density $\pi$, for Gaussian variational families.

The following remark shows that mean-seeking behavior is apparent when the set $\mathcal{Q}$ comprises Gaussians. The remark also introduces a useful definition for a form of Gaussian projection\index{Gaussian!projection} from the space of all probability measures on $\R^d$ onto the space of Gaussians on $\R^d$.
\begin{remark} \label{rem:bishop}
We write $\cG=\cG(\R^d)$ for the set 
of all Gaussian probability density functions over $\R^d,$ including degenerate Gaussians
with non-invertible covariance; in particular all Dirac\index{Dirac measure} measures are contained
in $\cG.$ Define $\Gn : \mathcal{P} \to \mathcal{G}$ by
$$\Gn\pi  \in \argmin_{\mu \in \mathcal{G}}\,\, \dkl(\pi\|\mu).$$
It then follows that, in fact,
$$\Gn\pi  = \normal(m_{\pi},\Sigma_{\pi}),$$
where $m_{\pi}$ and $\Sigma_{\pi}$ are the mean and covariance of random variable $u \sim \pi.$
\end{remark}
Further discussion on mode-seeking versus mean-seeking variational inference and references to the literature may be found in the bibliography in Section \ref{sec:vbib}.



\subsection{Evidence Lower Bound}\index{evidence! lower bound}
\label{elb}

We now relate $\J(q)$, defined by (\ref{eq:KLobj_minusZ}a), to an important concept in variational inference,\index{inference!variational} namely the
\emph{evidence lower bound}\index{evidence! lower bound} ($\ELBO$) given in Definition~\ref{d:elbo}:
$$\ELBO(\widetilde{\post},q) = \mathbb{E}^q[\log \widetilde{\post}(u)] - \mathbb{E}^q[\log q(u)].$$
The reason for the terminology evidence lower bound \index{evidence! lower bound}
will be made clear through Remarks \ref{r:elbo1}, \ref{r:elbo2}. 
The evidence lower bound is sometimes also known as the negative variational free energy.
\index{free energy}
The definition is often used in the situation where $\widetilde{\post}(u)$ is found from
the joint distribution of random variable $(u,y)$, with $y$ frozen. This is the setting
of the following proposition, which sheds light on the significance of the $\ELBO$ functional.
Recall prior\index{prior} measure $\pr(u)$, likelihood\index{likelihood} function $\like(y|u)$ given by equation \eqref{eq:like} and posterior\index{posterior} measure $\post:=\post^y$ given by Theorem \ref{t:bayes}.

\begin{proposition}
\label{p:elbo}
Define the unnormalized density $\widetilde{\post}(u)=\pr(u)\like(y|u)$ and define the normalization constant
$Z$ by $\post=Z^{-1}\widetilde{\post},$ assuming that $Z>0.$
Then, the maximizer $$\qopt \in \argmax_{q \in \mathcal{P}} \ELBO(\widetilde{\post},q)$$
is attained at the posterior distribution $\qopt = \post$. Furthermore, the maximum value
is $\ELBO(\widetilde{\post},\qopt)=\log(Z).$
\end{proposition}

\begin{proof} This follows from noting that    
\begin{align*}
    \ELBO(\widetilde{\post},q) & = \mathbb{E}^q[\log \post(u)] - \mathbb{E}^q[\log q(u)] + \mathbb{E}^q[\log \widetilde{\post}(u)] - \mathbb{E}^q[\log \post(u)]\\
    & = \mathbb{E}^q[\log \post(u)] - \mathbb{E}^q[\log q(u)] + \mathbb{E}^q[\log(Z)]\\ 
& = -\dkl(q\|\post) + \log(Z).
\end{align*}
Since $Z$ is independent of $q$, optimality requires that $\dkl(\qopt\|\post) = 0$, and
hence that $\qopt=\post.$ As a consequence $\ELBO(\widetilde{\post},\qopt) = \log(Z)$.
\end{proof}

\begin{remark}
\label{r:elbo1}
In the preceding proposition $\widetilde{\post}(u)=\bbP(u,y)$ is the joint distribution\index{distribution!joint} and the normalizing constant\index{normalizing constant} $Z=\bbP(y)$ is called the \emph{evidence}.\index{evidence} Thus the preceding proposition demonstrates the significance of the quantity $\ELBO(\widetilde{\post},\qopt)$: it delivers the logarithm of the evidence\index{evidence}, namely the log-probability that the observed data $y$ was produced by the statistical model for the joint random variable $(u,y)$ defined by equation \eqref{eq:jc0} and
Assumption \ref{a:jc1}. It is natural to ask what we can learn about the evidence if we perform variational inference over subset $\cQ$ of $\cP;$ this is the subject of the next remark.
\end{remark}

\begin{remark}
\label{r:elbo2}
Variational inference\index{inference!variational}
is performed by solving the following optimization problem over $\mathcal{Q} \subset \mathcal{P}:$
\begin{equation} 
\qs \in \argmin_{q \in \mathcal{Q}} \J(q).
\end{equation}
We continue with the setting of Proposition \ref{p:elbo} in which
$\post=Z^{-1}\widetilde{\post}.$
Recall the definitions of $\F(\cdot)$ from \eqref{eq:KLU} and note that it may be
extended to unnormalized second argument so that we may write
\begin{align*}
\F(q) =\dkl(q\|\post) &= \int_{\Ru} \log \left(\frac{q(u)}{\post(u)}\right) q(u) \, du\\
&= \int_{\Ru} \log \left(\frac{q(u)}{\widetilde{\post}(u)}\right) q(u) \, du + \log(Z)\\
&= - \ELBO(\widetilde{\post},q)+\log(Z).
\end{align*}
But by \eqref{eq:KLobj} we have, for $\J(\cdot)$ defined by \eqref{eq:KLobj_minusZ}, $\F(q)=\J(q)+\log(Z)$. Hence
$$\J(q)=- \ELBO(\widetilde{\post},q)$$
and it follows that
$$\log(Z)=\ELBO(\widetilde{\post},\qopt) \ge \ELBO(\widetilde{\post},q^\star)=-\J(q^\star).$$ 
Hence, after variational inference\index{inference!variational} over ${\cal Q}$ is performed, we can 
estimate the normalizing constant, or the evidence,\index{evidence} by $Z \ge \exp\bigl(-\J(q^\star)\bigr).$ 
The terminology \emph{evidence lower bound}\index{evidence!lower bound} is now made clear.
\end{remark}

\section{Bibliography}
\label{sec:vbib}

The paper~\cite{zellner1988optimal} provided impetus for
variational Bayes, highlighting the fact that the posterior is found by minimizing
$\J(\cdot)$ given in \eqref{eq:VariationalForm_Bayes}. 
Our presentation has focused on finding the posterior as the minimizer of the KL divergence, but there are other functionals that may be considered. For instance, the paper~\cite{trillos2018bayesian} shows that the posterior is found by minimizing a functional of the same structural form as~\eqref{eq:KLobj} based on the $\chi^2$ divergence (see Definition \ref{def:chi2}), rather than KL divergence. Papers~\cite{li2016renyi, hernandez2016black} show 
how to perform variational inference\index{inference!variational} using the family of Renyi-alpha divergences, which include the KL and $\chi^2$ divergences as well as the Hellinger metric. In \cite{cai2024eigenvi} the Fisher 
divergence\index{divergence!Fisher}, based on matching
the score associated with the desired posterior, is used as an objective function
for variational inference. However, whilst there are many divergences\index{divergence} that could be used to define Bayes Theorem via optimization over the space of measures, the specific choice of KL divergence\index{divergence!Kullback--Leibler} in which the measure to be minimized over appears in the first argument, has a special place: amongst all 
$\mathsf{f}$-divergences,\index{divergence!$\mathsf{f}$} it is the
unique choice for which the objective function does not require knowledge
of the normalization constant $Z$~\cite{chen2023gradient}. Generalizations of variational inference, wherein the KL divergence is replaced by a different divergence and the log-likelihood by a different loss function, are explored in \cite{knoblauch_optimization-centric_2022}; in particular, other choices can be more robust to model and prior misspecification. 

Refer to~\cite{hoffman2013stochastic} for a modern approach to variational inference,\index{inference!variational} with applications to topic models for large collections of documents\index{topic models}. 
For applications in graphical models\index{graphical models} see~\cite{wainwright2008graphical}. The paper \cite{blei2017variational} provides an accessible introduction to variational inference, including the implementation of CAVI in exponential family models. 
An extension to the mean-field model is to include interactions~\cite{jordan1999introduction}.
Natural extensions of Gaussians include the Gaussian mixtures considered in~\cite{lambert2022variational}. The derivation of the mean-field equations for Gaussians was first shown in~\cite{opper2009variational}. Variational inference for sparsity-promoting Bayesian models \cite{calvetti2019hierachical} is considered in \cite{agrawal2022variational}; see also \cite{law2022sparse}.
Discussion of mean-seeking versus mode-seeking may be found in \cite{sanz-alonso_inverse_2023};
see Sections 4.2 and 4.3 in particular.
The property of $\Gn$ presented in Remark \ref{rem:bishop} is highlighted in
\cite{bishop} and proved in \cite[Section 4.3]{sanz-alonso_inverse_2023}.

In this chapter we have seen that the problem of minimizing KL divergence over its first argument can be reformulated in terms of maximizing the $\ELBO$ functional. From a conceptual viewpoint, it is often insightful to derive algorithms and estimation procedures using the minimization of KL divergence as a guiding principle; however in practice it is common to work with the $\ELBO,$ whose definition requires only a density known up to normalizing constant. The idea of minimizing KL divergence/maximizing $\ELBO$ is ubiquitous in machine learning and statistics, and it underpins many methods studied in these notes, including learning the prior to posterior map in Section \ref{sec:LPrtoPo}, the variational formulations of data assimilation in Sections \ref{sec:variational_smoothing0} and \ref{sec:variational_filtering0}, variational autoencoders in Section \ref{sec:vauto}, and the \index{expectation maximization} expectation--maximization algorithm in Section \ref{sec:142}, used to learn priors for data assimilation in Section \ref{sec:EMmodel_dynamics}.

The $\ELBO$ is a non-convex functional even for simple potentials; see the example in Appendix G of~\cite{lambert_variational_2023}. Due to this non-convexity, it is often challenging to develop guarantees for variational inference outside restricted settings on the distribution. However, there have been several recent papers studying statistical and algorithmic guarantees for variational inference.\index{inference!variational} We refer to~\cite{lambert2022variational} for analysis using gradient flows and reference to the literature.

\chapter{\Large{\sffamily{Learning the Prior}}}
\label{ch:priorIP}

In Bayesian inference the prior acts as a form of regularization.
This can be seen explicitly for both the optimization and probabilistic approaches
to inverse problems. First, consider the optimization approach and MAP estimation\index{MAP estimator} as described in Subsection \ref{ssec:MAP}. In that setting the objective function is a sum of two terms: the loss function\index{loss!function}, which is the negative log-likelihood $-\log \like(y|u)$; and a regularizer\index{regularizer}, given by the negative logarithm of the prior $\pr$. 
Secondly, consider the probabilistic approach, and the variational form of Bayes Theorem\index{Bayes Theorem!variational} as described in Subsection \ref{sec:vbay}, and Remark \ref{rem:2.3}, in particular.
The posterior is shown to be the minimizer of an objective defined over probability densities. This objective is the sum of two terms: the data mismatch term, which is the expectation, with respect to 
the putative minimizer $q$, of the loss function\index{loss!function}; and the
regularizer\index{regularizer}, which is the KL divergence\index{divergence!Kullback--Leibler} between the putative minimizer $q$ and the prior $\pr.$ Choice of prior, or relatedly a regularizer, is thus an essential component of both classical and probabilistic approaches to inverse problems.

The prior should, of course, be chosen to reflect prior knowledge. But defining a prior, or regularizer, can be challenging. 
Often modellers make simple parametric choices, based on domain-specific knowledge or intuition, within some class of priors 
or regularizers that lead to computational tractability. For example if a Gaussian\index{Gaussian!prior} prior is adopted 
then the mean and covariance are selected using domain knowledge. However the abundance of data that is available 
in many fields suggests the possibility of employing purely data-driven priors for inversion, a major focus of this chapter. 

We start, in Section \ref{sec:BH}, by discussing Bayesian hierarchical modelling,
where parameters of the prior are learned directly from the data available in the Data Assumption \ref{da:vb}
used to define the inverse problem itself. Then, in Section \ref{sec:LP}, we introduce the additional Data Assumption \ref{da:4a}, for settings where the prior is defined through samples. 
We consider learning the prior measure from data $U:=\{ u^{(n)} \}_{n=1}^N$ 
drawn from the prior, relating this problem to the unsupervised learning task studied in Chapter \ref{ch:UL}.
We initially focus on approximation of the prior by use of the empirical\index{empirical} measure (see Definition \ref{def:emp}) defined by the data $U$, and study the effect of this empirical approximation of the
prior on the posterior. Then, in Section \ref{sec:pertp3}, we continue with different ideas from 
unsupervised learning as introduced in Chapter \ref{ch:UL},
representing the prior via a pushforward from a latent space.\index{latent space} Such an
approach arises when the prior is learned from data as a transport\index{transport}, or by using
autoencoders\index{autoencoder}. We highlight that this approach may be useful even if the
prior is known analytically, and not just through data; working in a latent space can improve the capability of
some algorithms for inversion. In Section \ref{sec:pertp3} we also discuss the effect, on the posterior, 
of approximating the prior by using a pushforward. Section \ref{sec:pbib} contains concluding
bibliographic remarks.

\begin{remark}
In this chapter we focus on learning parameterized priors and regularizers; we use either the single observation $y$
defining the inverse problem (hierarchical Bayes\index{Bayes!hierarchical}), or the data $U$ constituting samples from the prior. In 
Chapter \ref{ch:data-dependence} we will study settings where we have data $\{y^{(n)}, u^{(n)} \}_{n=1}^N$ 
defined by the joint distribution of $(y,u) \in \R^k \times \Ru$ and how to learn regularizers and priors in
that context.
\end{remark}

\section{Bayesian Hierarchical\index{Bayes!hierarchical} Modelling}\label{sec:BH}

Recall the Bayesian \index{inverse problem!Bayesian}inverse problem of finding $u \in \R^d$ from $y \in \R^k$ 
when related by \eqref{eq:jc0}, so that $$y=G(u)+\eta.$$ 
The Bayesian hierarchical\index{Bayes!hierarchical} methodology postulates a \emph{parameterized family} of priors\index{prior!family}, $\pr(u;\theta)$ and proposes to learn the pair $(u,\theta)$
from the single piece of data $y$ given by Data Assumption~\ref{da:vb}.
Thus, we learn both the unknown $u$, \emph{and} parameters $\theta$ that define the prior, from $y$.
To formulate this problem we view the parameterized family of priors $\pr(u;\theta)$
as conditional probability densities $\pr(u|\theta)$, put a prior $\pr(\theta)$ on parameter
$\theta$ and construct the prior $\pr(u,\theta)=\pr(u|\theta)\pr(\theta)$ on the pair $(u,\theta)$. The prior $\pr(\theta)$, being a prior on the parameter of a prior, is known as a \emph{hyperprior}\index{hyperprior}.
Under appropriate generalization of the assumptions of Theorem~\ref{t:bayes} to the jointly
varying pair $(u,\theta)$, the \index{posterior}posterior distribution in this case is given by
\begin{equation}
\label{eq:cf0}
 \post(u,\theta) = \frac{1}{Z} \noise\bigl(y - G(u)\bigr)\pr(u,\theta);
\end{equation}
we have dropped the explicit dependence of the \index{posterior}posterior on $y$ 
for notational convenience.

\begin{example}
Consider finding $u$ from $y$ where
$$y=u+\eta$$
and $\eta \sim \Nc(0,\gamma_\eta^2).$ We place a prior on $u|\theta$
with form $\Nc(\theta,\sigma^2);$
if we then assume $\theta \sim \Nc(0,\gamma_\theta^2)$ then this specifies a jointly Gaussian prior $\rho(u,\theta)$
on the pair $(u,\theta)$.
Under the specified prior $\pr(u,\theta)$, $u=\theta+\xi_u$ and $\theta=\xi_\theta$ where $\xi_u$ and $\xi_\theta$ are independent centered Gaussians with variances $\sigma^2$ and
$\gamma_\theta^2$. Thus, under this prior, $u=\xi_u+\xi_\theta$ and hence
has distribution given by $\pr(u):=\Nc(0,\sigma^2+\gamma_\theta^2).$ 

Using these formulae we can identify the mean and covariance of the joint random variable: the mean is $0$ and the covariance is
\begin{eqnarray*}
   \left( 
\begin{array}{cc}
\sigma^2+\gamma_\theta^2 & \gamma_\theta^2 \\
\gamma_\theta^2 & \gamma_\theta^2
\end{array}
\right).
\end{eqnarray*}
By \eqref{eq:cf0} we obtain
$$\post(u,\theta) \propto \noise\bigl(y - u\bigr)\pr(u,\theta),$$
where $\noise$ is the density of the Gaussian noise distribution $\Nc(0,\gamma_\eta^2).$
Since both the likelihood and the prior are Gaussian, the posterior $\post(u,\theta)$ is 
also Gaussian, and its mean and covariance can be readily computed.

By integrating out $\theta$ we arrive at the posterior for the original non-hierarchical problem
$$\post(u) \propto \noise\bigl(y - u\bigr)\pr(u).$$
In general it is not possible to explicitly integrate out the parameter $\theta$ and formulate the problem
in terms of $u$ only; the specific structure on $\pr(u,\theta)$ allows it in this case. In the general setting
it is necessary to solve the inverse problem for pair $(u,\theta).$
\end{example}

Hierarchical Bayesian learning works with the same Data Assumption~\ref{da:vb}
that we employ to define the standard Bayesian inverse problem defined by~\eqref{eq:jc0}.
Each different data instance for $y$ would give rise to a different choice of the parameter $\theta$
(if MAP\index{MAP estimator} estimation is used), or distribution over the parameter $\theta$ 
(if a fully Bayesian approach is used), entering the prior for $u$.
Hence we obtain a different prior on $u$ for each data instance $y.$ 
In the remainder of this chapter, we work in a different scenario where the form of the prior is learned 
from samples of that prior; we thus make a different data assumption.

\section{\index{prior}Prior via Diracs\index{Dirac!measure}}\label{sec:LP}

In the remainder of this chapter we work under the assumption that desired prior $\rho$ is only
known to us through samples:

\begin{dataassumption}\index{Data Assumption}
\label{da:4a} We are given independent samples $U:=\{u^{(n)}\}_{n=1}^N$
from the prior $\rho$ on $u$, which is unknown.
\end{dataassumption}

To clearly  understand the methodology of the remainder of this chapter,
it is important to distinguish between the piece of 
\emph{observational data}\index{observational data}, $y$,  
for which we wish to solve the inverse problem  defined by \eqref{eq:jc0},  
and the \emph{training data}\index{training data}, $U$,
which we use to learn about the prior.

In this section we discuss a direct empirical approximation of the prior (Subsection \ref{ssec:pertp202}),
using the training data, and the properties of the resulting posterior (Subsection \ref{ssec:pertp2}).
In the following Section \ref{sec:pertp3} we look at finding smooth approximations of
the prior, using the same training data.

\subsection{Empirical\index{empirical} Prior}
\label{ssec:pertp202}

As discussed in the preamble to Chapter \ref{ch:UL}, 
Data Assumption \ref{da:4a} immediately yields an empirical\index{empirical} 
approximation for $\rho:$
\begin{align}
\label{eq:emp0}
\rn(u)=\frac{1}{N} \sum_{n=1}^N \delta_{\un}(u)=  \frac{1}{N}  \sum_{n=1}^N \delta(u-\un).
\end{align}
We sometimes refer to $\pr$ as the population\index{population limit} limit of $\pr^N.$
We note that $\rn$ is a \emph{random probability measure}\index{probability!measure!random}
and is a random approximation of the population limit $\pr.$
Recall the following metric on the space of random probability measures, from Subsection \ref{ssec:rpm}:
\begin{equation}
    \label{eq:rpmd0}
\dr(\pr,\pr') = \sup_{|f|_{\infty} \leq 1} \left|\mathbb{E}\left[\left(\pr(f) - \pr'(f)\right)^2\right]\right|^{1/2},
\end{equation}
where, for any probability measure $\intp \in \cP(\R^d)$, $\intp(f):= \mathbb{E}^{u \sim \intp}[f(u)]$. 
Theorem \ref{t:MC} shows that 
$\dr(\pr,\pr^N)^2 =\mathcal{O}(\frac{1}{N}).$
Thus a natural approach to the inverse problem is to simply use $\rn$ as prior in place of $\rho$ in 
Theorem \ref{t:bayes}. We now discuss this idea.

\subsection{Empirical\index{empirical} Posterior}
\label{ssec:pertp2}

Under the assumptions of Bayes Theorem \ref{t:bayes}, the \index{posterior}posterior distribution, given prior $\pr$, is 
\begin{equation}
\label{eq:cf1}
 \post(u) = \frac{1}{Z} \noise\bigl(y - G(u)\bigr)\pr(u);
\end{equation}
we have dropped the explicit dependence of the \index{posterior}posterior on $y$ for notational convenience,
 as $y$ plays no role in the current discussions.
With the finite sample approximation of the population\index{population limit} limit of the prior, and by 
application  of an appropriate generalization of Bayes Theorem \ref{t:bayes} to priors which are sums of Diracs, 
we obtain the approximate posterior
\begin{equation}
\label{eq:cf12}
 \post^N(u) = \frac{1}{Z^N} \noise\bigl(y - G(u)\bigr)\pr^N(u).
\end{equation}
From this it follows that 
\begin{subequations}
\label{eq:emp-post}
\begin{align}
\post^N(u)&=\sum_{n=1}^N w^{(n)}\delta_{\un}(u),\\
 w^{(n)}&= \ell^{(n)}\Big/\Bigl(\sum_{i=1}^N \ell^{(i)}\Bigr),\quad \ell^{(n)} = \like(u^{(n)}),
\end{align}
\end{subequations}
where, from \eqref{eq:like} but dropping explicit $y$-dependence, 
\begin{equation}
 \label{eq:lny}   
\like(u) = \noise\bigl(y - G(u)\bigr).
\end{equation}

We wish to compare the population\index{population limit} limit $\post$ with $\post^N$.
From \eqref{eq:rpmd0} we have
\begin{equation}
    \label{eq:rpmd}
\dr(\post,\post^N) = \sup_{|f|_{\infty} \leq 1} \left|\mathbb{E}\left[\left(\post(f) - \post^N(f)\right)^2\right]\right|^{1/2},
\end{equation}
where $\pi(f):= \mathbb{E}^{u \sim \pi}[f(u)]$ and $\pi^N(f):= \mathbb{E}^{u \sim \pi^N}[f(u)]$. 
We note, because it will be useful in what follows, that 
\begin{equation}
    \label{eq:rpmd2}
\sup_{|f|_{\infty} \leq F} \left|\mathbb{E}\left[\left(\post(f) - \post^N(f)\right)^2\right]\right|^{1/2} = F\,\dr(\post,\post^N).
\end{equation}
Theorem \ref{thm:emp_prior_perturbation} uses this metric on random probability
measures to quantify the distance between the true posterior and the approximate posterior resulting from
empirical approximation of the prior, using samples. We work under the following assumption:

\begin{assumption}\label{a:dh2L} Likelihood $\like:\R^d \to \R^+$ defined by \eqref{eq:lny} is bounded:
there exists $K \in (0,\infty)$ such that $\sup_{u \in \R^d}\like(u)=K.$
\end{assumption}

\begin{theorem} \label{thm:emp_prior_perturbation}  Let Assumption~\ref{a:dh2L} hold and let $\post,\post^N$ be the posteriors given by \eqref{eq:cf1} and \eqref{eq:cf12} respectively, with $\rho_N$ given by \eqref{eq:emp0}
and $\like$ by \eqref{eq:lny}. Then,
\begin{equation} \label{eq:emp_prior_stability}
\dr(\post,\post^N) \leq \frac{2K}{Z}\dr(\pr,\pr^N).
\end{equation}
\end{theorem}

\begin{proof} For the true and approximate posterior, we can write the integrals required to estimate
$\dr(\post,\post^N)$ by
$$\post(f) = \frac{\pr(\like f)}{\pr(\like)},\quad \post^N(f) = \frac{\pr^N(\like f)}{\pr^N (\like)}.$$
Then we have 
\begin{align*}
    \post(f) - \post^N(f) &= \frac{\pr(\like f)}{\pr(\like)} -   \frac{\pr^N(\like f)}{\pr^N (\like)}\\
    &= \frac{\pr(\like f) - \pr^N(\like f)}{\pr(\like)} - \frac{\pr^N(\like f)\bigl(\pr(\like) - \pr^N(\like)\bigr)}{\pr(\like)\pr^N(\like)}\\
    &= \frac{\pr(\like f) - \pr^N(\like f)}{\pr(\like)} - \frac{\post^N(f)\bigl(\pr(\like) - \pr^N(\like)\bigr)}{\pr(\like)}. 
\end{align*}
Using the basic inequality $(a-b)^2 \leq 2(a^2 + b^2)$, \eqref{eq:rpmd2}, that $|\post^N(f)|^2 \leq 1$ for all $|f|_{\infty} \leq 1$ and that $|\like|_{\infty}, |\like f|_{\infty} \leq K$  
\begin{align*}
    \left|\mathbb{E}\left[\left(\post(f) - \post^N(f)\right)^2\right]\right| &\leq \frac{2}{\pr(\like)^2} \Biggl(\mathbb{E}\Bigl[\bigl(\pr(\like f) - \pr^N(\like f)\bigr)^2\Bigr] + \mathbb{E}\left[ \bigl(\post^N(f)\bigr)^2 \bigl(\pr(\like) - \pr^N(\like) \bigr)^2\right] \Biggr) \\
   &\leq \frac{4K^2}{\pr(\like)^2} \dr(\pr,\pr^N)^2.
\end{align*}
Taking the supremum over test functions on the left-hand side and using that $\pr(\like) = Z$ gives us the desired result.
\end{proof}

We note that so far we did not specify how the samples $u^{(n)}$ defining the perturbed prior $\pr^N$ were generated. In fact, the result above holds for any empirical\index{empirical} measure. If $u^{(n)} \sim \pr$ are sampled i.i.d., however, $\pr^N$ is a Monte Carlo approximation of $\pr$. Moreover, we can appeal to convergence results for Monte Carlo to show the convergence rate of $\post^N$ to the true posterior $\post$.

\begin{corollary} Let the assumptions made in Theorem~\ref{thm:emp_prior_perturbation} hold,
and in addition, assume that Data Assumption \ref{da:4a} holds so that
$\pr^N$ is the Monte Carlo estimator \eqref{eq:emp0} of $\pr$. Then 
$$\dr(\post,\post^N) \leq \frac{2K}{Z}\frac{1}{\sqrt{N}}.$$
\end{corollary}

\begin{proof} By Theorem \ref{t:MC},  $\dr(\pr,\pr^N)^2 \leq \frac{1}{N}$. Using this in the right-hand side of~\eqref{eq:emp_prior_stability} gives the desired result.
\end{proof}

\section{Prior via Pushforward\index{pushforward}}
\label{sec:pertp3}

In this section, rather than using an empirical prior, we try to determine a smooth approximation
of the prior. To this end, compare Data Assumption \ref{da:4a} to Data Assumption \ref{assumption:unsupervised} in Chapter \ref{ch:UL}. 
The latter assumption arises in unsupervised learning; 
notice that, taking $\Upsilon := \rho$ shows that the two assumptions are identical.
Thus, when the prior is only given to us through Data Assumption \ref{da:4a}, we may seek to approximate it using the generative modelling\index{generative model} techniques of Chapter \ref{ch:UL}. 

Theorem \ref{t:KM} shows that, under appropriate conditions on $\pr, \zeta$, there exists an exact 
pushforward map $g$ such that $\pr=g_\sharp \zeta.$ In fact there are many such maps.
We may use ideas such as autoencoders\index{autoencoder} and
transport\index{transport} maps from Chapter \ref{ch:UL}, to find an approximation of a map $g$ that pushes forward a given density 
$\zeta$ on latent\index{latent space} space $\R^{\dz}$ into the prior. 

In the context of inverse problems there are several specific reasons why this pushforward
approach to prior modeling can be useful. Firstly, there are settings where it may be of interest to
generate further samples from the prior itself, in addition to those given in Data Assumption \ref{da:4a},
discussed in Subsection \ref{ssec:adl};
it is then desirable that $\zeta$ is easy to sample from. A second setting is one in which the
goal is to solve the inverse problem in the latent space, discussed in Subsection \ref{ssec:latentp}; 
it is then desirable that $\zeta$ has a closed form density to enable application of methods such as MAP 
estimation\index{MAP estimator} and MCMC\index{MCMC}. The third setting, also requiring a closed form density for
$\zeta$ and a closed form for the approximation of map $g$, is one in which methods such as MAP 
estimation\index{MAP estimator} and MCMC\index{MCMC} are applied to solve the inverse problem in the
original space, discussed in Subsection \ref{ssec:pertp1}.
In the second and third settings, once $\zeta$ and an approximation of $g$ are identified they can be re-used for different 
inverse problems for $u$, or for different instances of the realization $y$ used in the inverse problem
under consideration, all provided that the prior remains unchanged.

\subsection{Additional Prior Samples}
\label{ssec:adl}

Throughout this subsection we assume that $\pr = g_\sharp \zeta$ and assume that $d_z=d$ so that $g: \R^d \to \R^d.$ 
Using Data Assumption \ref{da:4a}, and ideas such as transport\index{transport} and autoencoders\index{autoencoders} in Chapter \ref{ch:UL},
we obtain an approximation to $g$, denoted $g'$, and hence obtain an approximation
of $\pr$ given by $\pr' = g'_\sharp \zeta$. If $\zeta$ is easy to sample and $g'$ is
straightforward to evaluate, then we may obtain approximate i.i.d. samples from $\pr$
by sampling i.i.d. from $\zeta$ and applying $g'$, thereby obtaining samples from $\pr'.$

It is natural to ask how to estimate the distance between $\pr$ and $\pr'$ in terms of some measure of
closeness between $g$ and $g'.$ This turns out to be a naturally answered in the 
Wasserstein$-p$\index{distance!Wasserstein} distances
defined in Subsection \ref{sec:OPT}. Recall that, with specific choice of $\dm(u,v)=|u-v|_p$, the $p$-norm on $\R^d,$
we have
$$W_{p}(\pr,\pr')\coloneqq \biggl({\rm inf}_{\intp \in \Pi_{\pr,\pr'}}
\int_{\R^{d} \times \R^d} |u-v|_p^p\,\intp(u,v) \,du dv\biggr)^{1/p};$$
here $\Pi_{\pr,\pr'} \in \cP(\R^d \times \R^d)$ is the set of couplings\index{coupling} 
of measures $\pr, \pr' \in \cP(\R^d).$
Thus if $\intp$ is \emph{any} specific coupling of $\pr, \pr'$ then
$$W_{p}(\pr,\pr')^p \le 
\int_{\R^{d} \times \R^d} |u-v|_p^p\,\intp(u,v) \,du dv=\bbE^{(u,v) \sim \intp}|u-v|_p^p.$$
To upper-bound the right-hand side, we define a coupling $\intp:=(g,g')_{\sharp}\zeta;$ 
clearly the marginals are $\pr,\pr'$ as required. Furthermore
$$\bbE^{(u,v) \sim \intp}|u-v|_p^p=\bbE^{z \sim \zeta}|g(z)-g'(z)|_p^p.$$
We have thus proved: 
\begin{theorem}
    \label{thm:usingwp} Let $g,g': \R^d \to \R^d$ be two Borel measurable maps and let
    $\pr = g_\sharp \zeta, \pr' = g'_\sharp \zeta.$ Then 
$$W_{p}(\pr,\pr') \le   \Bigl(\bbE^{z \sim \zeta}|g(z)-g'(z)|_p^p \Bigr)^{1/p}.$$ 
\end{theorem}

\subsection{Posterior in Latent Space}
\label{ssec:latentp}

Throughout this subsection we again assume that $\pr = g_\sharp \zeta.$ 
We discuss the idea of solving the inverse problem in the latent space
where $\zeta$ is defined.
In practice, of course, $g$ will only be known approximately. This approximation may be
obtained, as in the preceding subsection, using samples given in Data Assumption \ref{da:4a}. But
the methodology in this subsection may also be useful in settings where prior $\pr$ is known analytically; this
is because solution of the inverse problem may be easier in a latent space.

We consider the inverse problem of finding $z$ from $y$, where
$$y=G\bigl(g(z)\bigr)+\eta.$$
Since $g$ is a deterministic map Assumption \ref{a:jc1} can be reinterpreted as:
\begin{assumption}\label{a:jc12}
The distribution of the random variable $(z,\eta) \in \R^{\dz} \times \Ry$ is defined by assuming that $z \ind \eta$, that $z \sim \zeta$ and that $\eta \sim \noise$, where $\zeta$ and $\noise$ are probability density functions and, in addition, $\noise$ is centered.
\end{assumption}
\noindent Under this assumption we may apply Bayes Theorem \ref{t:bayes} to define the posterior distribution
\begin{equation}
\label{eq:cf2}
\postg(z) = \frac{1}{Z_g} \noise\Bigl(y - G\bigl(g(z)\bigr)\Bigr)\zeta(z).
\end{equation}
If map $g$ is given by the decoder of an autoencoder, as in Section \ref{sec:auto}, then typically
the latent space dimension $\dz$ is smaller than $\du$. In the case of transport\index{transport}, as described
in Section \ref{sec:transport2}, $\dz=\du$ and we have the following:

\begin{proposition}
\label{prop:push}
Let $ \dz  =d$ and assume that  $\pr=g_\sharp \zeta$ and that $g$ is invertible
and that the inverse has positive determinant everywhere on $\R^d.$
Then $\postg$ given by \eqref{eq:cf2} and the desired posterior $\post$
are related by the identity $\post=g_\sharp \postg.$ 
\end{proposition}

\begin{remark}
\label{rem:push}
The significance of Proposition \ref{prop:push} is that it implies that we can
solve the Bayesian inverse problem \eqref{eq:cf2}, posed in the latent space\index{latent space!inverse problem}, 
and push forward under
$g$ to find solutions of the original Bayesian inverse problem defined
by \eqref{eq:cf1}. For example if we generate samples under $\postg$ in the latent space, then application of $g$ to those samples will generate samples
under $\post$ in the original space. Hence,
if pair $(g,\zeta)$ is known, then the original \index{Bayesian!inverse problem}Bayesian 
inverse problem for $\post$ on $\R^d$
may be converted to one for $\postg$ on $\R^{\dz}$;
pushforward\index{pushforward} of $\postg$ under $g$ yields solution to the original problem. Thus we have designed a method for solving Bayesian inverse problems when prior $\rho$ is known only through samples by converting to an inverse problem
for $z$. 

We may apply MAP\index{MAP estimator} or MCMC\index{MCMC} methods
because the posterior $\postg(z)$ has an explicit
formula for its density when $G,g$ and density for $\zeta$ are known analytically. 
Furthermore, if the proposal distribution for MCMC requires samples
from $\zeta$ then it is desirable that $\zeta$ is easy to sample from.
We point out that performing MCMC in latent space can be useful even in cases where the prior $\rho$ is fully known and not only accessed through samples, as in some cases the posterior in latent space may be easier to sample from. 
A reference that illustrates this idea can be found in the bibliography Section \ref{sec:pbib}.
We also discuss data assimilation in a latent space\index{latent space!data assimilation} in Section \ref{sec:latent_DA}.
\end{remark}

\begin{proof}[Proof of Proposition \ref{prop:push}]

From Lemma \ref{lem:cov} with $u=g(z)$, and \eqref{eq:cov} in particular, 
\begin{align*}
(g_\sharp \postg)(u)&= \bigl(\postg \circ g^{-1}\bigr)(u) {\rm det} 
D\bigl(g^{-1}\bigr)(u)\\
& \propto \noise\bigl(y - G(u)\bigr) \bigl(\zeta \circ g^{-1}\bigr)(u) 
{\rm det} D\bigl(g^{-1}\bigr)(u)\\ 
&= \noise\bigl(y - G(u)\bigr) g_\sharp \zeta(u)\\ 
&= \noise\bigl(y - G(u)\bigr) \pr(u)\\
& \propto \post(u).
\end{align*}
Thus $(g_\sharp \postg)(\cdot) \propto \pi(\cdot)$ and since both are probability density functions, they are in fact equal.
\end{proof}

Remark \ref{rem:push} shows that, if we learn an exact transport map $g$ satisfying $\rho = g_\sharp \zeta,$ then we 
can exactly sample the true posterior with prior $\pr$ by sampling
the posterior distribution associated with a (potentially simpler) inverse problem with prior $\zeta$, 
and applying $g$ to those samples. In practice, however,
we learn $g$ from $\pr^N \approx \pr$ and so we recover an approximate transport map $g'$; the issue of lack of exactness is further compounded by only learning the transport map over a parametric family of densities and by not iterating the optimization solver to the global minimizer. In Subsection \ref{ssec:pertp1} 
we prove Theorem \ref{thm:prior_stability} which addresses the effect of approximate $g'$:
we assume that the approximation of the exact transport map leads to an approximate smooth prior, in the space of $u$, and look at the effect on the posterior of the approximation error between this prior and the true prior. In Theorem \ref{thm:emp_prior_perturbation} of Subsection \ref{ssec:pertp2} we have already studied a related problem, namely the
effect on the posterior of replacing the prior by an empirical\index{empirical} approximation.

\subsection{Effect of Approximate Pushforward}
\label{ssec:pertp1}

Throughout this subsection we again assume that $\pr = g_\sharp \zeta.$ However, as discussed
in Subsection \ref{ssec:adl},
in practice we will not have an \emph{exact} pushforward $g$ from $\zeta$ to $\rho,$ but rather will have
an approximation $g'$.
We will thus have an approximate prior $\pr'$ in place of $\pr.$
Consider the inverse problem arising from use of a smooth approximate prior $\pr' \approx \pr.$
This gives rise to an approximate posterior, $\post':$
\begin{equation}
\label{eq:cf11}
 \post'(u) = \frac{1}{Z'} \noise\bigl(y - G(u)\bigr)\pr'(u).
\end{equation}
It is thus important to know that a small change in the prior leads
to a small change in the posterior. We prove this in the following
theorem which exhibits conditions under which the prior to posterior
map is Lipschitz in the Hellinger metric. In the proof of the theorem,
and conditions that precede it, we let $\like(u)=\noise\bigl(y - G(u)\bigr)$, suppressing dependence on $y$,
as in Section \ref{sec:LP}.

\begin{theorem} \label{thm:prior_stability} Let Assumption \ref{a:dh2L} hold. Consider posteriors
$\post, \post'$ in \eqref{eq:cf1}, \eqref{eq:cf11} corresponding, respectively, to priors
$\pr, \pr'$. Then, we have
\begin{equation}
\dhell(\post,\post') \leq \left(\frac{2K}{(\sqrt{Z} + \sqrt{Z'}) \sqrt{Z'}} + \frac{\sqrt{K}}{\sqrt{Z'}} \right) \dhell(\pr,\pr').
\end{equation}
\end{theorem}

The preceding theorem, and proof that follows, are readily modified to the setting where the \index{prior}prior distributions $\pr$ and $\pr'$ are both supported on bounded open set $D \subset \Ru,$ in which case the bound on the likelihood in Assumption \ref{a:dh2L}
may be replaced by a bound over $D$.

\begin{proof}[Proof of Theorem \ref{thm:prior_stability}] We have 
\begin{align*}
    \dhell(\post,\post') &= \frac{1}{\sqrt{2}}\left\|\sqrt{\frac{\like\pr}{Z}} - \sqrt{\frac{\like\pr'}{Z'}} \right\|_{L^2} \\
    &\leq \frac{1}{\sqrt{2}}\left\|\sqrt{\frac{\like\pr}{Z}} - \sqrt{\frac{\like\pr}{Z'}} \right\|_{L^2} + \frac{1}{\sqrt{2}}\left\|\sqrt{\frac{\like\pr}{Z'}} - \sqrt{\frac{\like\pr'}{Z'}} \right\|_{L^2} \\
    &= \frac{1}{\sqrt{2}}\left|\frac{1}{\sqrt{Z}} - \frac{1}{\sqrt{Z'}} \right| \sqrt{Z}+ \frac{1}{\sqrt{2}\sqrt{Z'}}\|\sqrt{\like\pr} - \sqrt{\like\pr'}\|_{L^2}.
\end{align*}
The first term can be rewritten using the identity $$\left|\frac{1}{\sqrt{Z}} - \frac{1}{\sqrt{Z'}} \right| \sqrt{Z} = \frac{|Z - Z'|}{(\sqrt{Z} + \sqrt{Z'})\sqrt{Z'}}.$$ The 
difference between the normalization constants can be bounded, using Lemma
\ref{l:dh1}, by 
\begin{align*}
    |Z - Z'| &\le \int_{\R^d} |\like(u)\pr(u) - \like(u)\pr'(u)| \, du \\
    &\leq K \int_{\R^d} |\pr(u) - \pr'(u)| \, du \\
    &= 2K \dtv(\pr,\pr') \\
    &\leq 2K\sqrt{2} \dhell(\pr,\pr').
\end{align*}
For the second term, we have 
$$\|\sqrt{\like \pr} - \sqrt{\like \pr'} \|_{L^2} \leq \sqrt{K}\|\sqrt{\pr} - \sqrt{\pr'}\|_{L^2} = \sqrt{2K}\dhell(\pr,\pr').$$
Collecting the bounds for the two terms, we obtain the result.
\end{proof}

\section{Bibliography}
\label{sec:pbib}
Bayesian hierarchical modeling for inverse problems is reviewed in \cite{calvetti2018inverse}. Hierarchical models for level-set inversion were introduced in \cite{dunlop2017hierarchical} while hierarchical models to promote sparsity in the reconstructions are studied in \cite{calvetti2019hierachical,kim2022hierarchical,waniorek2023bayesian,sanz2025hierarchical}. 
The idea of learning \index{prior}prior probabilistic models from data is
overviewed in \cite{asim2020invertible}. An application is
described in \cite{patel2021gan} where a \index{generative adversarial network}generative adversarial
network  (GAN) is used to determine a mapping from a \index{Gaussian}Gaussian
to the space of \index{prior}prior data samples. The paper \cite{gao2023image} also implicitly learns
a prior, through simultaneous consideration of multiple inverse problems; but it also learns the
posterior distribution for each of these inverse problems at the same time, linking to 
the two subsequent Chapters \ref{ch:transport} and \ref{ch:data-dependence}.
Many methods of this type
reduce the dimension of the unknown parameter space, determining
a \index{latent space}latent space of low effective dimension. A different approach to learning this mapping is to use invertible maps
\cite{kingma2018glow} and the work on
\index{normalizing flow}normalizing flows \cite{tabak2010density,song2020score};
see \cite{asim2020invertible} for application of invertible
maps to \index{prior}prior construction for inversion. 
This idea can be combined with \index{variational!Bayesian method}variational inference to solve
\index{sampling}sampling problems in general, and \index{inverse problem}inverse problems in particular
\cite{rezende2015variational,sun2021deep,gabrie2021efficient,ongie2020deep}.
The use of triangular \index{transport}transport maps for \index{Bayesian!inference}Bayesian inference was introduced in \cite{el2012bayesian}. The paper \cite{glaubitz2025efficient} considers transforming a sparsity promoting hierarchical prior into a standard Gaussian prior with the goal of obtaining a posterior distribution in latent space that is easier to sample from.

The idea of learning regularizers from data, to define \index{objective}objective functions
for the optimization approach to inversion, is overviewed in
\cite{arridge2019solving} and~\cite{benning2018modern}. The paper \cite{soh2019learning} provides
a framework for the subject which employs structured factorizations of 
data matrices to learn semidefinite regularizers.
The paper \cite{lunz2018adversarial} develops an adversarial approach
to the problem. Inspired by the Neumann series, the work \cite{gilton2019neumann} proposes an end-to-end learning approach that utilizes data-driven nonlinear regularizers. Score-based generative priors for inverse problems in imaging are considered, for instance, in  \cite{feng2023score,sun2024provable}.

A collection of stability results relating the posterior error to perturbations in the prior measure with respect to various metrics and divergences (e.g., KL divergence, $\chi^2$ divergence and Wasserstein-1 metric) can be found in~\cite{garbuno2023bayesian, sprungk2020local}. We refer to Chapter \ref{ch:distances} for background on these and other distances and divergences. While the stability results presented in this chapter are with respect to distance between two priors, the results can also be translated to stability with
respect to the distance between the transport maps that define the prior. A set of these stability results for various metrics and divergences can be found in~\cite{baptista2023approximation}.

The field of image-processing has developed a range of machine-learned prior modeling or regularization techniques,
known collectively as ``Plug-and-Play'' methods~\cite{romano2017little,kamilov2023plug}. The original motivation is to use image 
de-noisers, trained on paired data that is independent of the observations defining the inverse problem, to learn the effect of regularization in an iterative optimization technique that alternates between minimizing the loss and applying the regularizer.
These ideas are generalized to Bayesian inversion, rather than MAP estimation, in \cite{wu2024principled}, where the denoising
step is often undertaken with a diffusion-based generative model~\cite{song2020score}. The idea of using diffusion-based
generative modeling to solve Bayesian inverse problems first appears in~\cite{chung2022diffusion}; recent work in this area
may be found in~\cite{zhang2025improving}. Benchmark problems for diffusion-based ``Plug and Play'' methods are proposed in
\cite{zheng2025inversebench}. In addition to prior modeling based on diffusions, similar ideas are being developed using flow-matching techniques~\cite{lipman2022flow, zhang2024flow}.

\chapter{\Large{\sffamily{Transport to the Posterior}}} \label{ch:transport}

In this chapter we introduce various transport\index{transport} approaches to determining the posterior distribution. Recalling the Definition \ref{def:transportmap} of a transport
map, we build on the transport methodologies introduced in Chapter \ref{ch:UL}. Section \ref{sec:EPrtoPo} is devoted to the question of whether a transport map, taking prior\index{prior} to posterior\index{posterior} under pushforward, exists.
In Section \ref{sec:LPrtoPo} we formulate the problem of mapping the prior to the posterior
as an optimization problem over a parameterized class of transports. We discuss a population formulation
of the objective function, as well as the empirical approximation implemented in practice. 
Furthermore, we demonstrate that the methodology is a special
case of variational inference and we obtain insights into the approach by considering a nonparametric
formulation. Section \ref{sec:BVI}
generalizes the approach beyond variational inference\index{variational!inference}, using an objective function
defined by divergences other than KL\index{divergence!Kullback--Leibler}, leading to a methodology which is
usable given a prior which is known only through samples.
In Section \ref{sec:LOtoPo} we generalize the transport approach: we consider maps from general latent spaces, not restricted to the prior, to the posterior. Section \ref{sec:lmcmc} is devoted to using transport to enhance 
Markov chain Monte Carlo (MCMC)\index{MCMC}. 
Bibliographic remarks are contained in Section \ref{sec:tbib}.

\section{Existence of the \index{prior}Prior-to-\index{posterior}Posterior Map}\label{sec:EPrtoPo}

Consider \index{Bayes Theorem}Bayes Theorem \ref{t:bayes} 
written as map relating \index{prior}prior $\pr$
to \index{posterior}posterior $\post$ via the relation
\begin{subequations}
\label{eq:thisagain}
\begin{align}
 \post(u) &= \frac{1}{Z} \like(u) \pr(u),\\
Z &= \bbE^{u \sim \pr}\, \bigl[ \like(u) \bigr].
\end{align}
\end{subequations}
We drop explicit dependence on $y$ since we do not exploit it in this chapter.
Equation~\eqref{eq:thisagain} defines a nonlinear map $\pr \mapsto \post$ on the space of probability density functions,
as discussed in Remark \ref{rem:eaudc}.
We now ask whether we can realize this map via a transport map $T$ on $\R^d$ with the property $\post=T_\sharp \pr$.  We say that map $T$ with this property is a \emph{prior-to-posterior transport map.}\index{transport!prior-to-posterior} Furthermore, if $T$ is invertible then we may write $\pr = (T^{-1})_\sharp \post$. Although \emph{optimal} 
transport\index{transport!optimal} will not play a significant role in the methods 
developed in this chapter, it can be used to provide
a straightforward existence theorem for the desired transport maps:

\begin{theorem} \label{t:P2P}
Let $\rho \in L^1(\R^d;\R)$ 
be a probability density function. 
Assume that 
$\mathbb{E}^{z \sim \pr} |z|_r^r <\infty$ for some $r \in (1,\infty)$, that $\like(\cdot)$ 
satisfies Assumption \ref{a:dh2L} and that $Z$ given by (\ref{eq:thisagain}b)
is strictly positive. Then there is a
Borel measurable map $T$ with property that $\pi=T_\sharp \pr.$ 
\end{theorem}


\begin{proof}
 The result is a direct corollary of Theorem \ref{t:KM} if we prove that $\mathbb{E}^{u \sim \post} |u|_r^r <\infty.$
 This follows from noting that $$\mathbb{E}^{u \sim \post} |u|_r^r \le \frac{K}{Z} \mathbb{E}^{z \sim \pr} |z|_r^r<\infty.$$
\end{proof}

\begin{remark}
The optimal transport map $T$, whose existence is asserted in the preceding theorem,
is unique within the Monge class defined by Theorem \ref{t:KM}. However, it
is not the unique map from $\rho$ to $\pi.$ Indeed the methodology we develop in this
chapter will not invoke optimality in the sense of optimal transport maps.
\end{remark}

\section{Learning the \index{prior}Prior-to-\index{posterior}Posterior Map}\label{sec:LPrtoPo}
In practice an exact prior-to-posterior transport must be approximated, and in this and the following sections we do so within a class of parameterized maps $\cTt,$ where $ \Theta \subset \R^p$ denotes the parameter space. 
In Subsection \ref{ssec:PL} we consider the idealized setting of infinite data, and describe how to find an optimal value of the parameter $\theta \in \Theta$ by optimizing a population-based objective function. Subsection \ref{ssec:EL} builds on this idealized setting to define the empirical objective function used in practice. 
In Subsection \ref{ssec:adderr} we quantify the effect of using an approximate transport map on the posterior
approximation and in Subsection \ref{ssec:cvi} we show that the methodology is a special case of variational inference\index{variational inference}. 


\subsection{Population Objective}
\label{ssec:PL}

In this subsection we assume that an explicit formula for $\rho$ is available
\emph{and} that we can evaluate expectations under $\rho$ exactly; we
define an optimization problem for $\theta$ under these two assumptions. In
the following subsection we relax the second assumption, and instead evaluate
expectations approximately using samples from $\rho$ and Monte Carlo integration.
As in Section \ref{sec:transport2}, we work with function classes $\cTt$ that satisfy the inclusion
\begin{align}
\label{eq:thisclass55}
\begin{split}
\cTt \subset
\Bigl\{T(\cdot\,;\theta):\R^d\to\R^d , \, \theta \in \Theta \Big|\,\,
& T(\cdot\,;\theta) \text{ is a } C^1 \text{diffeomorphism},\\
&\det D_u T(u;\theta) > 0,\quad \forall u \in \R^d, \forall \theta \in \Theta
\Bigr\}.
\end{split}
\end{align}
We aim to find an element $T^\star \in \cTt$ such that,
typically only approximately, $T^\star_\sharp \rho=\pi.$

\begin{example} \label{ex:55}
Examples of classes of smooth invertible functions $T(\cdot\,;\theta): \R^d \to \R^d$ are discussed in Section \ref{sec:NORM}
where normalizing flows\index{flow!normalizing} are introduced.
In particular neural ODEs\index{neural ODEs}, 
introduced in Subsection \ref{sec:nODEs} provide an example of $\cTt.$
We first introduce a family of neural networks 
$h:\R^d \times \Theta \to \R^d$ -- see Remark \ref{rem:NNG}, 
and the material preceding it. We then consider the ODE
\begin{align}
\label{eq:nfode999}
\begin{split}
\frac{dv}{dt}&=h(v;\theta),\\
v(0)&=u.
\end{split}
\end{align}
Noting that $v(t)$ depends on $(u,\theta)$ we set 
 $T(u;\theta)=v(1).$ This defines class $\cTt.$ Crucially, the discussion in Subsection \ref{sec:nODEs}  demonstrates
how the inverse of $T(\cdot;\theta)$ can also be calculated via a differential equation, and shows how the
determinant of the Jacobian of $T(\cdot\,;\theta)$ can be evaluated.
\end{example}

Using the KL divergence\index{divergence!Kullback--Leibler} to measure discrepancy 
between the pushforward of the prior and the posterior
we define our goal as being determination of parameter $\theta$
to make the following objective function small:
\begin{equation}
\label{eq:way0}
\mathsf{F}(\theta)=\dkl\bigl(T(\cdot\,; \theta)_\sharp \pr \| \post\bigr).
\end{equation}
By Theorem \ref{t:P2P} we know that there exists map $g^\star$ such that 
$\dkl(g^\star_\sharp \pr \| \post)=0.$ Thus minimizing $\mathsf{F}(\cdot)$ forms a natural
objective. Note, however, that Theorem \ref{t:P2P} asserts existence of
$g^\star$ by use of optimal transport.\index{transport!optimal} In contrast the objective function \eqref{eq:way0}
does not appeal to optimal transport in any way; there is hence no reason to expect an optimizer,
found by minimizing over $\theta$, to be close to an \emph{optimal} transport.

By Theorem \ref{thm:invariance}, which ensures the invariance of KL divergence under invertible and differentiable transformations such as those in the class $\cTt$, minimizing \eqref{eq:way0} is equivalent to minimizing 
\begin{equation}
\label{eq:waya}
\mathsf{F}(\theta)=\dkl\bigl(\pr \| T^{-1}(\cdot\,; \theta)_\sharp \post\bigr).
\end{equation}
This is exactly the same problem considered in Section~\ref{sec:transport2}, with
$g \mapsto T^{-1}$, $\zeta \mapsto \post$ and
$\Upsilon \mapsto \pr$.\footnote{See Remark \ref{rem:invparam}.} Thus, since $\bbE^{u \sim \pr} \log \pr(u)$ is
independent of $\theta$, or directly from \eqref{eq:OT}, we deduce that minimizing $\mathsf{F}(\cdot)$ 
is equivalent to minimizing
\begin{equation*}
-\bbE^{u \sim \pr} \Bigl[\log \post \circ T(u;\theta) + \log {\rm det}  
D_u T(u;\theta) \Bigr].
\end{equation*}
Using the expression \eqref{eq:thisagain} for the \index{posterior}posterior in terms of 
the \index{prior}prior, and noting that $\log Z$ is an additive constant with respect to $\theta$, 
we may define the optimization problem
\begin{subequations}
\label{eq:subeqp}
\begin{align}
\J(\theta) &= -\bbE^{u \sim \pr} \Bigl[\log \pr \circ T(u;\theta)+ 
\log \like \circ T(u;\theta)+\log {\rm det}  D_u T(u;\theta) \Bigr], \label{eq:OT50}\\
\thetas & \in \argmin_{\theta \in \Theta} \J(\theta). \label{eq:optimal_parameters_paramprior2}
\end{align}
\end{subequations}
Once we have (approximately) solved this problem we may employ \index{transport}transport map $T^\star=T(\cdot\,;\thetas)$
to approximately map the prior to the posterior: $\post \approx T^\star_{\sharp} \pr$.

\begin{remark}
We observe that, as for the Metropolis-Hastings\index{Metropolis-Hastings} MCMC\index{MCMC} of
Section \ref{sec:MCMC}, it is not necessary to know the normalization constant
for the posterior in order to apply the methodology of this section. This may be seen from Remark \ref{rem:2.3} after
noting that the transport map methodology for Bayesian inversion is simply a specific instance of variational
inference; we show this latter fact in Subsection \ref{ssec:cvi}.

The methodology in this subsection 
is not specific to inverse problems; rather it applies to the sampling of
any measure $\post$ which is defined via \eqref{eq:thisagain}. The methodology requires knowledge
of $\like(\cdot)$ and reference measure $\pr(\cdot)$. No data point $y$ is explicitly required. If, however, the density $\like(u)$ is defined 
by a Bayesian inverse problem, then Data Assumption~\ref{da:vb} is needed to provide the single data 
point $y$ entering the likelihood. 
\end{remark}

\subsection{Empirical Objective}
\label{ssec:EL}

To implement the optimization problem defined by \eqref{eq:subeqp} it is necessary to empiricalize the objective function, in order to
avoid evaluating expectations over $\rho$. To this end, here we work under the following generalization of Data Assumption\index{Data Assumption} \ref{da:4a}:

\begin{dataassumption}\index{Data Assumption}
\label{da:4aGEN} We are able to evaluate prior $\pr(u)$ and likelihood $\like(u)$ for all $u \in \R^d.$ 
We are given samples $U:=\{u^{(n)}\}_{n=1}^N$ drawn i.i.d. from $\rho$. 
\end{dataassumption}

Using the \emph{training data}\index{training data} 
$U$ from $\rho$ we define
\begin{align}
\label{eq:emp1}
\rn(u)=\frac{1}{N} \sum_{n=1}^N \delta_{\un}(u).
\end{align}
We may then define the following minimization problem:
\begin{subequations}\label{eq:minempirical}
\begin{align}
\label{eq:OT50N}
\J^N(\theta)
 &= -\bbE^{u \sim \pr^N} \Bigl[\log \pr \circ T(u;\theta)+ 
\log \like \circ T(u;\theta)+\log {\rm det} D_u T(u;\theta) \Bigr],\\
\thetas & \in \argmin_{\theta \in \Theta} \J^N(\theta). 
\end{align}
\end{subequations}
Note that we use Data Assumption\index{Data Assumption} \ref{da:4aGEN} to define this objective function:
samples from $\pr$ are used to compute expectation under $\pr^N$; and the ability to evaluate prior and likelihood are used to compute the terms $\log \pr \circ T(u;\theta)$ and $\log \like \circ T(u;\theta).$
Once we have (approximately) solved the minimization problem \eqref{eq:minempirical} 
we may employ \index{transport}transport map $T^\star=T(\cdot\,;\thetas)$
to (approximately) map the prior to the posterior: $\post \approx T^\star_{\sharp} \pr$.

\subsection{Effect of Transport Error on the Posterior} \label{ssec:adderr}

In the preceding subsection we explain how to identify map $T^\star$ that approximates an exact transport map $T$,
leading to an approximate posterior $\post^\star = T^\star_{\sharp} \pr$.
It is natural to ask what effect the error in approximating $T$ has on the posterior. 
By using identical ideas to those leading to Theorem \ref{thm:usingwp} we may prove:

\begin{theorem}
    \label{thm:usingwp2}
     Let $T,T^\star$ be two Borel measurable maps on $\R^d$ and let
    $\post = T_\sharp \pr, \post^\star= T^\star_\sharp \pr$.  Then 
 $$W_{p}(\post,\post^\star) \le  \Bigl(\bbE^{u \sim \pr}|T(u)-T^\star(u)|_p^p \Bigr)^{1/p}.$$ 
\end{theorem}

\subsection{Connection to Variational Inference}
\label{ssec:cvi}

In this subsection we connect the learning of a transport\index{transport!prior-to-posterior}  
from prior to posterior, to a  variational inference\index{variational inference} 
problem of the type introduced in Chapter \ref{ch:VI}.
We work in the population loss setting of Subsection \ref{ssec:PL}. Consider a parametric class of maps
$\mathcal{T}$ satisfying  \eqref{eq:thisclass55}. We drop explicit $\Theta$
dependence as we will now formulate optimization over the class $\mathcal{T}$
itself, rather than over the parameters defining the class.
Given $\mathcal{T}$ we may define the variational class  $\mathcal{Q} := \{q: q = T_\sharp \pr, T \in \mathcal{T} \}$. And
given the solution $\theta^\star$ defined by the prior-to-posterior transport\index{transport}
optimization problem \eqref{eq:subeqp} we may define $\qs \coloneqq T(\cdot\,;\thetas)_\sharp \pr$.

\begin{theorem} 
Probability density $\qs \in {\mathcal{P}(\R^d)}$ 
solves the variational inference problem 
$$\qs \in \argmin_{q \in \mathcal{Q}} \dkl(q\|\pi).$$
\end{theorem}
\begin{proof}
We first note that the class $\mathcal{T}$ necessarily comprises only invertible maps. We showed above that solving the optimization problem \eqref{eq:optimal_parameters_paramprior2},
with $\J(\cdot)$ defined by \eqref{eq:OT50}, is equivalent to minimizing $\mathsf{F}(\cdot)$ from \eqref{eq:waya}.
But from Theorem~\ref{thm:invariance} we have that, for invertible and differentiable transformations, 
$$\dkl(\pr\|T^{-1}(\cdot\,;\theta)_\sharp \pi) = \dkl(T(\cdot\,;\theta)_\sharp \pr \|\pi).$$ 
Thus, 
$$\argmin_{\theta \in \Theta} \J(\theta) = \argmin_{\theta \in \Theta} \dkl(T(\cdot\,;\theta)_\sharp \pr \|\pi).$$
Lastly, $T(\cdot\,;\thetas): \Ru \to \Ru$ is a differentiable map, and its derivative has positive determinant; it is therefore an invertible map as a consequence of being an element of
a parametric class of maps $\mathcal{T}$ satisfying \eqref{eq:thisclass55}. 
Thus $q^\star=T(\cdot ;\thetas)_\sharp \pr \in \mathcal{Q}$. 
\end{proof}

\begin{example}
\label{ex:cholGVI}
Consider the setting where the prior is a standard Gaussian $\cN(0,I).$
Now we assume that $T$ is affine. Specifically, we let $$\Theta:= \Bigl \{ \theta:=(m,A) \in \R^d \times \R^{d\times d} \Big| \det(A)>0 \Bigr\}$$ 
and set
\begin{equation}
\label{eq:thisclass2}
\cTt := \Bigl\{T(u;\theta) = Au + m \, \Big| (m,A) \in \Theta \quad \theta=(m,A) \in \Theta \Bigr\}.
\end{equation}
Then, for $T \in \cTt$ it holds that $D_u T(u;\theta) = A \in \mathbb{R}^{d \times d}$ is an invertible matrix with positive determinant. An affine transformation of a standard Gaussian yields a Gaussian random variable. In fact, the class of approximate posterior distributions then consists of multivariate Gaussians of the form 
$$\mathcal{Q} = \Bigl\{q = \mathcal{N}(m, AA^\top) \Big|\; m \in \R^d,\,\, A \in \R^{d \times d},\,\, \det(A)>0 \Bigr\}.$$
Hence, solving the optimization problem for $\J(\cdot)$ is a subclass of the general Gaussian variational inference problem considered in Subsection \ref{ssec:gauss}. Here we
are parameterizing the Gaussian covariance 
through a square root. It is possible to parameterize the covariance using its Cholesky\index{Cholesky} factorization by
further restricting $A$ to the class of lower triangular matrices 
$A \in \R^{d\times d}$  with positive diagonal entries.
\end{example}

\begin{example} Assume that prior\index{prior} $\pr$ is a probability density function in the mean-field family introduced in Definition~\ref{d:qmf}:
$$ \pr(u) = \prod_{i=1}^d \pr_i(u_i).$$
Here each $\pr_i$ is a probability distribution over $\R.$
We now seek probability distribution $q$ over $\R^d$ in the same form:
$$ q(u) = \prod_{i=1}^d q_i(u_i).$$
Again  each $q_i$ is a one-dimensional probability distribution over $\R.$  
We seek each of these $q_i$ as the pushforward of $\pr_i$, under map $T_0(\cdot\,;\theta_i): \R \to \R$ for some judicious choice of parameter $\theta_i \in\R^p.$  To this end we define 
\begin{align}
\label{eq:thisclass3}
\begin{split}
\cTt &:= \Bigl\{T(u;\theta) = \bigl(T_0(u_1;\theta_1),\dots,T_0(u_d;\theta_d)\bigr),
\theta=(\theta_1,\dots,\theta_d) \in \R^{p \times d},\\ &\qquad T_0(\cdot\,;\theta_0) \in C^1(\R;\R)\,\,\textrm{diffeomorphism,} \\
&\qquad \det D_u T_0(u;\theta)>0, \, \forall (u,\theta_0) \in \R^d \times \R^p\Bigr\}.
\end{split}
\end{align}
Then, the corresponding variational class is
$$\mathcal{Q} = \left\{q = T(\cdot\,;\theta)_{\sharp} \pr,\quad T(\cdot\,;\theta) \in \cTt \right\}.$$
\end{example}

\section{Sample-Based Objective Functions} \label{sec:BVI}

Recall that the method described in Subsection \ref{ssec:EL} requires knowledge of the density $\pr$, as well
as samples from it; in particular an explicit
expression for $\log \rho$ is required. We now describe an alternative approach, which is purely sample-based;
to achieve this we use the energy distance introduced in Definition \ref{def:energy}, rather than KL divergence, to define an objective 
function. In doing so we avoid the need to have a formula for the prior: 
all that is required is samples. This is similar to the situation in 
Section \ref{sec:LP} where we constructed an approximate prior from samples.
We start, however, with a population level objective function, defined through the energy distance:
\begin{equation}
\label{eq:way0E}
\mathsf{F}(\theta)=\den^2\bigl(T(\cdot\,; \theta)_\sharp \pr \| \post\bigr).
\end{equation}
We show that it is possible to empirically approximate this objective 
function under the following generalization of 
Data Assumption\index{Data Assumption} \ref{da:4a}: 

\begin{dataassumption}\index{Data Assumption}
\label{da:4alike} We are given samples $\{u^{(n)}\}_{n=1}^N$ assumed
to be drawn i.i.d. from prior measure $\rho$ on $u$ which is unknown.
We are able to evaluate likelihood $\like(u)$ for any $u \in \R^d.$ 
\end{dataassumption}

Recall $\rho^N$ defined by \eqref{eq:emp1}, repeated here for expository purposes:
\begin{align*}
\rn(u)=\frac{1}{N} \sum_{n=1}^N \delta_{\un}(u).
\end{align*}
We note that this measure is available to us
under Data Assumption\index{Data Assumption} \ref{da:4alike}.
From this measure we may define 
\begin{align*}
 \post^N(u) &= \frac{1}{Z^N} \like(u) \pr^N(u),\\
Z^N &= \bbE^{u \sim \pr^N}\, \bigl[ \like(u) \bigr].
\end{align*}
Since $\pr^N$ is an empirical measure, constructed from
samples from $\rho$, we may explicitly construct $\post^N$ by evaluating the likelihood
$\like(\cdot)$ at the samples and, by renormalizing, define weights that lie on the
probability simplex -- see equation \eqref{eq:emp-post}. In the notation prevailing here this
yields 
\begin{align*}
\post^N(u)&= \sum_{n=1}^N w^{(n)}\delta_{\un}(u),\\
 w^{(n)}&= \ell^{(n)}\Big/\Bigl(\sum_{i=1}^N \ell^{(i)}\Bigr),\quad \ell^{(n)} = \like(u^{(n)}).
\end{align*}
We may, then, approximate \eqref{eq:way0E} empirically, under
Data Assumption\index{Data Assumption} \ref{da:4alike}, by
$$\mathsf{F}^N(\theta)=\den^2\bigl( T(\cdot\,; \theta)_\sharp\pr^N,\post^N\bigr).$$
Following the ideas outlined in Remark \ref{rem:mmd&e} we then obtain the approximation
\begin{align*}
  & \den^2\bigl(T(\cdot\,; \theta)_\sharp \pr \| \post\bigr)
  \approx  \frac{2}{NM} \sum_{i,j} |T(u^{(i)}; \theta) - v^{(j)}|\\
  &\qquad\qquad\qquad\qquad\qquad - \frac{1}{N^2} \sum_{i,j} |T(u^{(i)}; \theta) - T(u^{(j)}; \theta)| - \frac{1}{M^2} \sum_{i,j} |v^{(i)} - v^{(j)}|,
\end{align*}
where the $\{v^{(i)}\}_{i=1}^M$ are drawn i.i.d. from $\pi^N$ by resampling.

\begin{remark}
The ideas in this section are readily generalized to the use of the maximum mean discrepancy 
$\dmmd$ from Definition \ref{def:dmmd}, with other choices of kernel, rather than the energy 
distance specifically. Furthermore, the discussion in Remark \ref{rem:POF} is relevant to 
understanding the different choices of objective function 
that we use throughout this book.
\end{remark}

\section{Learning Other Maps to the \index{posterior}Posterior}\label{sec:LOtoPo}

In the preceding two subsections we sought a pushforward map from prior to posterior. This is advantageous when we would like the posterior to inherit certain properties of the prior. For example, if the prior and posterior have the same tail behavior, then the transport map $T$ only needs to depart from an identity map in the bulk of the distribution. Similarly, if the posterior and prior only differ on a low-dimensional subspace of the parameters $u \in \R^d,$ then the map $T$ can be represented using a {\emph{ridge function}} \index{ridge function} that is constant for inputs orthogonal to the subspace of interest. Pushing forward from the prior can also be advantageous because priors that are easily sampled are often deployed in practice. However
in some settings it may be computationally expedient to seek $\post$
which is the pushforward\index{pushforward} of a different, easy-to-sample, reference density\index{reference distribution} $\refT(u)$ on $\Ru$, (such as a multivariate Gaussian\index{Gaussian}) rather than the prior $\pr(u)$ (when it is not Gaussian). Choosing a different reference measure may thus be convenient when it is challenging to sample from the prior density or when the posterior is very different from the prior -- for example having different tail behavior.

To consider this setting we replace \eqref{eq:waya} by
\begin{equation}
\label{eq:waya2}
\mathsf{F}(\theta)=\dkl\bigl(\refT \| T^{-1}(\cdot\,; \theta)_\sharp \post\bigr).
\end{equation}
Similar analysis to that leading to~\eqref{eq:OT50} in the case of mapping the prior to the
posterior shows that we may learn the parameters $\theta \in \Theta$ solving \eqref{eq:waya2} by minimizing
\begin{equation}
\J(\theta)
 =  -\bbE^{u \sim \refT} \Bigl[\log \pr \circ T(u;\theta)+ 
\log \like \circ T(u;\theta)+\log {\rm det}  
D_u T(u;\theta) \Bigr].
\end{equation}
Letting $\thetas$ denote the optimal parameters, the resulting posterior approximation is given by $\pi \approx T(\cdot ;\thetas)_\sharp \refT$.
Again, as in Subsection \ref{sec:LPrtoPo}, the expectation 
over the reference density must be evaluated empirically\index{empirical} and, for this, we generalize Data Assumption\index{Data Assumption} \ref{da:4aGEN} as follows:

\begin{dataassumption}\index{Data Assumption}
\label{da:4aGEN2}
We are able to evaluate prior $\pr(u)$ and likelihood $\like(u)$ for all $u \in \R^d.$ 
We are given samples $\{u^{(n)}\}_{n=1}^N$ drawn i.i.d. from $\refT$. 
\end{dataassumption}

\begin{remark} Unlike the variational inference problem in Chapter~\ref{ch:VI}, we now have two degrees of freedom: the reference measure \index{reference distribution} and the map. In principle, one may even parameterize the reference measure $\refT(\cdot,\vartheta)$ and learn its parameters $\vartheta$ simultaneously with learning the map $T(\cdot\,;\theta)$. For example, we might parameterize $\refT$ using a Gaussian mixture\index{Gaussian!mixture} with unknown means, covariances, and mixture weights.
\end{remark}

\section{Transport-Assisted MCMC\index{MCMC}}
\label{sec:lmcmc}
Given a reference distribution\index{reference distribution} $\postv$ on $\Ru$ and parameter space $\Theta \subseteq \R^p,$ the approach considered in the previous Section~\ref{sec:LOtoPo} yields the posterior approximation  $\post \approx T(\cdot\,; \thetas)_\sharp \postv.$ This approximation can naturally be used to produce $N$ independent approximate posterior samples as follows: 
\begin{enumerate}
    \item Draw  $z^{(n)} \overset{\rm i.i.d.} {\sim} \postv$ for $1 \le n \le N.$
    \item Set $u^{(n)} : = T(z^{(n)}; \thetas)$ for $1 \le n \le N.$
\end{enumerate}
The extent to which the samples $\{u^{(n)} \}_{n=1}^N$ are approximately distributed like the posterior $\post$ is fully determined by the quality of the approximation $\post \approx T(\cdot\,; \thetas)_\sharp \postv.$ 

This section describes a \index{MCMC}MCMC transport-assisted\index{MCMC!transport!assisted} approach for asymptotically-exact posterior sampling; we refer to Section \ref{sec:MCMC} for background on MCMC, a general methodology to obtain approximate and correlated samples from a desired target distribution. To introduce the idea of transport-assisted MCMC recall first that since the map $T(\cdot\,; \theta^\star)$ from Section \ref{sec:LOtoPo} is invertible, we also have the approximation $\postv \approx T^{-1}( \cdot\,; \thetas)_\sharp \post.$ 
The idea behind transport-assisted MCMC is that, since $\postv$ is straightforward to sample from, then its approximation $T^{-1}( \cdot\,; \thetas)_\sharp \post$ will also be straightforward to sample from using MCMC. For example, if $\postv$ is a standard multivariate Gaussian,
and if $T^{-1}( \cdot\,; \thetas)_\sharp \post$ approximates this measure well, then sampling it using MCMC will be straightforward. Moreover, the reference distribution $\postv$ can be used to design efficient proposal distributions for the MCMC algorithm.

\begin{algorithm}[!ht]
\caption{\label{alg:transportMCMC} Transport-assisted MCMC}
\begin{algorithmic}[1]
    \STATE {\bf Input}: Target distribution $\pi$, Number of samples $N.$ 
    \STATE Define preconditioned target distribution $T^{-1}( \cdot\,; \thetas)_\sharp \post.$
    \STATE Run MCMC 
    to obtain $N$ approximate samples $\{z^{(n)} \}_{n=1}^N$ from $T^{-1}( \cdot\,; \thetas)_\sharp \post.$
    \STATE Set $u^{(n)} := T(z^{(n)}; \thetas)$ for $1 \le n \le N.$
\end{algorithmic}
\end{algorithm}

Applying this approach leads to the methodology in Algorithm~\ref{alg:transportMCMC} to generate approximate and correlated posterior samples.
In comparison with directly running \index{MCMC}MCMC with the target $\post,$ transport-assisted MCMC  has the advantage that MCMC algorithms with target $T^{-1}( \cdot\,; \thetas)_\sharp \post$ may converge faster. In other words, the transport map is used to 
 \emph{precondition}\index{precondition} the posterior for MCMC sampling; for instance map $T^{-1}(\cdot\,; \thetas)$ may pushforward\index{pushforward} multimodal posterior $\pi$ into an approximately Gaussian distribution that is easy to sample from. 
 
 \begin{remark}
     It is important to notice that  the extent to which transport-assisted MCMC samples $\{u^{(n)} \}_{n=1}^N$ are approximately distributed like the posterior is determined by the convergence of the Markov chain $\{u^{(n)} \}_{n=1}^N$ to its limit distribution $T^{-1}( \cdot\,; \thetas)_\sharp \post$, rather than by the quality of the approximation $\post \approx T(\cdot\,; \thetas)_\sharp \postv.$ In other words, provided that we run MCMC with target $T^{-1}( \cdot\,; \thetas)_\sharp \post$ for long enough, transport-assisted MCMC\index{MCMC!transport-assisted} will yield samples whose distribution is arbitrarily close to the posterior, even if the approximation $\post \approx T(\cdot\,; \thetas)_\sharp \postv$ is poor. A good choice of $\thetas$ will be manifest in fast convergence of the
     transport-assisted MCMC.
 \end{remark}

\section{Bibliography}
\label{sec:tbib}

The approach to seek prior-to-posterior transport maps\index{transport!prior-to-posterior} 
was proposed in~\cite{el2012bayesian}. In particular that paper parameterized $T$ as a monotone-triangular map known as the Knothe--Rosenblatt rearrangement~\cite{villani2009optimal}; the advantage of this choice of map is that the determinant Jacobian, required to define the
minimization, is easy to evaluate. The approach was extended to compositions of triangular maps, as used in normalizing flows,\index{normalizing flow} in~\cite{rezende2015variational}. An approach to seek prior-to-posterior maps that are optimal with respect to minimizing a transportation distance (see Subsection \ref{sec:OPT} for a discussion on optimal transport) was proposed in~\cite{taghvaei2022optimal}. Transport from a latent random variable to the posterior, within the context of variational Bayes, is used to solve computational imaging problems in \cite{sun2021deep}.

An alternative to seeking the transport map directly is to construct the map incrementally by learning a map that is a small perturbation of the identity and that minimizes the distance between the current approximation and the actual posterior distribution. Stein variational gradient descent is one algorithm\index{algorithm} for finding such maps; it proceeds by seeking the map in a reproducing kernel Hilbert space in which the distance is measured by the KL divergence~\cite{liu2016stein}. Another approach learns the dynamics of a (possibly stochastic) differential equation to define the transport from its flow map as in~\cite{kerrigan2024dynamic, chemseddine2024conditional, albergostochastic}.

To handle certain classes of target distributions, it is sometimes desirable to consider objectives other than the KL divergence; examples include use of $\alpha$-divergences~\cite{hernandez2016black}, Wasserstein distances (see Subsection~\ref{sec:OPT})~\cite{ambrogioni2018wasserstein} or the development of tailored transport map approximations for heavy-tailed posteriors~\cite{liang2022fat}.

The development of transport-assisted\index{transport!assisted} MCMC sampling algorithms is an active research area, see for example~\cite{hoffman2019neutra,peherstorfer2019transport,wu2020stochastic,matthews2022continual,hagemann2022stochastic,arbel2021annealed,grenioux2023sampling,gabrie2022adaptive,parno2018transport}. We refer to \cite{bouchard2024mcmc} for an overview of transport-based techniques for enhanced MCMC\index{MCMC} sampling. While some methods first learn the transport map and then apply MCMC in the reference space as described in Section \ref{sec:lmcmc}, other approaches learn the transport map on the fly using the output of a Markov chain.  
These techniques are closely related to \emph{adaptive MCMC}\index{MCMC!adaptive} algorithms where the proposal kernel is suitably parameterized and learned online~\cite{atchade2005adaptive,roberts2009examples}. 
Transports learned from samples of a target posterior distribution have also been used to learn proposal distributions for importance sampling~\cite{muller2019neural}.

\chapter{\Large{\sffamily{Data Dependence of the Posterior}}}\label{ch:data-dependence}

Our focus in this chapter is on learning the dependence of the posterior distribution $\post^y(u)$, for parameter\index{parameter} $u$, on the observation\index{observation} $y$. 
Recall that, in the setting of Chapter \ref{ch1}, the posterior is given by
\begin{subequations}
\label{eq:thisagain2}
\begin{align}
 \post^y(u) &= \frac{1}{Z^y} \like(y|u) \pr(u),\\
Z^y &= \bbE^{u \sim \pr}\, \bigl[ \like(y|u) \bigr].
\end{align}
\end{subequations}
Here, in contrast to the previous chapter, we retain explicit $y$-dependence in the notation $\post^y$ and $Z^y$, since our goal is to learn 
parametric dependence of $\post^y$ on $y$. The resulting learned model can then be repeatedly used for different realizations of $y$, rather than having to relearn a different approximate posterior for each $y$. This reduces the cost of multiple inference procedures to training a single model. Hence, the approach is known as \emph{amortized inference}\index{inference!amortized} or simply \emph{amortization}.\index{amortization} 
In this chapter we consider amortized variational inference,\index{variational inference!amortized} amortized MAP and posterior mean estimation,\index{MAP!estimator} learning regularizers, and learning priors. The unifying theme is to exploit knowledge about the $y$-dependence of the inverse problem. A similar approach will be applied to learn the $y$-dependence of data assimilation problems in Chapter~\ref{ch:LG10}.

It is helpful to recall the notation, from Remark~\ref{rem:minus1}, for the probability density function $\gamma(y,u)$ characterizing the joint distribution of the observations\index{observations} and parameter\index{parameter} random variable $(y,u)$, and the marginal $\marg(y) = \int \joint(y,u) \, du$ on the observations.\footnote{Note that $Z^y=\kappa(y).$} The density may be factorized as $\gamma(y,u)=\like(y|u) \pr(u)$, formed from the product of the likelihood $\like(y|u)$, for the distribution on $y|u$, and the prior density $\pr(u)$ on parameter\index{parameter} $u$. It may also be factorized as $\pi^y(u)\kappa(y).$

In Section~\ref{sec:MAPd} we start by discussing the concept of amortization\index{amortization}, 
with respect to observation\index{observation}
dependence, in the context of MAP and posterior mean estimators.\index{MAP estimator}
Section~\ref{sec:likelihood-based} studies amortization in the context of variational inference, and includes the use of
prior-to-posterior transport\index{transport!prior-to-posterior} as a running example.
Section~\ref{sec:LLM!}  presents a framework for learning of likelihood models on the basis of paired parameter\index{parameter} and observation data;
it may be viewed as a probabilistic generalization of Section~\ref{sec:IPSU} where we discuss  
the learning of forward models for the purpose of inference. 
To understand the proposed methodology introduced in these three sections,
it is important to distinguish between the one piece of data $y$
for which we wish to solve the inverse problem defined by 
\eqref{eq:jc0} and the \emph{training data}\index{training data} $\{y^{(n)}\}_{n=1}^N$ 
which we assume is available here. Furthermore, it is important to notice that
we make a variety of assumptions, of increasing complexity, about the
information available to us, in addition to the training data, as we progress through the
subsections comprising Sections \ref{sec:MAPd}, \ref{sec:likelihood-based} and \ref{sec:LLM!}. The
reader will benefit from comparing Data Assumptions \ref{da:wnl}, \ref{da:map},  \ref{da:4b} and \ref{a:LBI} before reading Sections \ref{sec:MAPd}--\ref{sec:LLM!} in detail.

Everything in Sections \ref{sec:MAPd} and \ref{sec:likelihood-based} is concerned with learning
about data dependence of the posterior, and amortizing the cost of learning by doing so.
Section \ref{sec:LLM!} uses similar ideas to learn the likelihood.
But it is also possible to use modified versions of Data Assumptions \ref{da:4b} and \ref{a:LBI}
to learn about regularizers and prior models, respectively. We have already
discussed learning  regularizers and prior models in Chapter \ref{ch:priorIP} assuming 
the availability of data samples from the prior; here, we learn regularizers and 
priors by additionally assuming access to $y$-training data\index{training data}.
In Section~\ref{sec:LRP} we study the learning of regularizers for
MAP estimation and learning of priors\index{prior} for general
Bayesian inversion. The goal is to learn regularizers and priors 
that perform well, on average, for inversion with data $y \sim \kappa$. 
Thus our goal in Section~\ref{sec:LRP} is not amortization with respect to $y$-dependence, 
the goal of the preceding sections. But since the data assumptions employed overlap,
the learning of regularizers and priors is included in this chapter. 
We end with Section \ref{sec:lfbib} containing bibliographic remarks.

\section{Amortized Point Estimators\index{point estimator!amortized}}
\label{sec:MAPd}

In this section we discuss amortization of point estimators, in particular, the MAP estimator
\index{MAP estimator!amortized} and the posterior mean\index{posterior!mean estimator}.
\index{Bayes Theorem}Theorem \ref{t:bayes} delivers the posterior given in \eqref{eq:thisagain2}.
The \index{MAP estimator}MAP estimator of $u$ given data $y$ is defined, in Subsection~\ref{ssec:MAP},
as any point solving the maximization problem
$$\umap(y) \in \argmax_{\vct{u}\in\Ru}\post^{\vct{y}}(\vct{u}).$$
The posterior mean estimator\index{posterior!mean estimator} of $u$ given data $y$ is defined, in Subsection~\ref{ssec:PE1}, as
\begin{equation*}
    \upm(y) = \Expect^{u\sim \post^y}[u].
\end{equation*}
The MAP estimator and posterior mean estimator coincide for Gaussian $\post^{\vct{y}}$, but are otherwise typically distinct point estimators.

In this chapter we emphasize that $\umap$ and $\upm$ are functions of $y$: we write $\umap: \R^\dy \to \R^\du$ and denote
the MAP point at a given observation $y$ as $\umap(y)$, and similarly write $\upm: \R^\dy \to \R^\du$ and $\upm(y)$.
We discuss two approaches to learn the dependence of the MAP estimator on data; that is, to amortize MAP estimation. In addition, we describe one way to amortize posterior mean estimation.
To this end, we introduce the parameter set $\Theta \subseteq \R^p$ and let $\cRt \subset C(\R^k,\R^d)$ denote
a parameterized set of functions $u(\cdot \, ; \theta)$ for $\theta \in \Theta.$
We seek a function $u \in \cRt$ to approximate
the dependence of the \index{MAP estimator}MAP or posterior mean\index{posterior!mean estimation} estimator on data $y$.

\begin{example} \label{ex:crt}
We might take $\cRt$ to comprise a family of neural networks, with specified depths and widths, with $\Theta$ comprising the parameter space for the weights\index{weights} and biases\index{biases} 
which define a specific element in $\cRt.$ See Remark \ref{rem:NNG}, and the material preceding it,
for discussion of neural networks mapping Euclidean spaces into one another.
\end{example}

In Subsection~\ref{ssec:amortized_map_obs} we make Data Assumption \ref{da:wnl}: we assume that we have training data comprising
only observation realizations, and learn an amortized MAP estimator that minimizes the standard MAP objective from Chapter
\ref{ch:VI} in expectation over the observations.
In Subsection~\ref{ssec:amortized_map} we employ Data Assumption \ref{da:map}: we assume that we have pairs of observations and numerically computed MAP estimates, and treat the problem of finding an amortized MAP estimator as a supervised learning problem. In Subsection~\ref{ssec:CME} we study posterior mean estimation,
using Data Assumption~\ref{da:4b}: we assume we have samples from the joint distribution on parameters and observations.

\subsection{Amortized MAP Estimation\index{MAP estimator} Using Only Observations}\label{ssec:amortized_map_obs}

We start by discussing an approach that uses only the observations to learn an amortized MAP estimator. We make the following data assumption:
\begin{dataassumption}\index{Data Assumption}
\label{da:wnl}
We are able to evaluate the prior $\pr(u)$ and the likelihood $\like(y|u)$ for any
pair $(y,u) \in \R^\dy \times \R^\du.$ We are given data in the form of independent samples $\{y^{(n)}\}_{n=1}^N$ 
from the marginal distribution $\marg$.
\end{dataassumption}

Recall that $\umap(y)$ minimizes the MAP objective function $\J_{\mathsf{MAP}}(\cdot\,;y): \R^d \to \R$ defined, for each $y \in \R^k$, 
by \eqref{eq:loss}, \eqref{eq:reg}, and \eqref{eq:MAPO}:
\begin{equation*}
\J_{\mathsf{MAP}}(u;y) = - \log \like(y|u) - \log \pr(\vct u).
\end{equation*}
In Chapter \ref{ch1} this method is applied for a given fixed observation $y.$
Here our goal is to learn the mapping $y \mapsto \umap(y)$. We approximate this mapping by a function $u^\star \in \cRt$. 
By minimizing the average of the MAP objective function over $y$ from its marginal $\kappa$ we seek a good approximation, in $\cRt$, of the mapping $y \mapsto \umap(y)$ over the support of $\kappa$.
To determine the value $\thetas \in \Theta$ which characterizes the function $u^\star$ we 
consider the optimization problem
\begin{subequations}
\begin{align}\label{eq:MAP_amortized}
\J(\theta) &=  -\bbE^{y\sim\kappa} \bigl[\log \like(y|u(y; \theta)) + \log \pr(\vct u(y; \theta)) \bigr],\\
\theta^\star &\in \argmin_{\theta\in\Theta} \J(\theta),
\end{align}
\end{subequations}
for $u(\cdot\,; \theta) \in \cRt.$ Then $u^\star=u(\cdot\,;\theta^\star).$
In practice we approximate the expectation defining the optimization problem by
using the empirical measure $\kappa^N$ defined by $\{y^{(n)}\}_{n=1}^N$.

\begin{remark} \label{rem:zy}
When we derived the objective function to be minimized for MAP estimation in Subsection \ref{ssec:MAP}, we
dropped the term $\log Z^y$ which appears in the expression for the negative log posterior. This term remains irrelevant
here, even though we are now computing an expectation over $y$, because it is again independent of $\theta.$
\end{remark}

\subsection{Amortized MAP Estimation\index{MAP estimator} Using Supervised Learning}\label{ssec:amortized_map}

In this subsection we consider the same problem as in the preceding subsection,
but under the following variation on Data Assumption \ref{da:wnl}:
\begin{dataassumption}\index{Data Assumption} \label{da:map}
We are given multiple pairs of observations and numerically computed MAP estimators 
$\bigl\{ (y^{(n)},\umap^{(n)})\bigr\}_{n=1}^N$, drawn i.i.d. from a probability measure $\mu$ on $\Ry \times \Ru$.
\end{dataassumption}

\begin{remark} \label{rem:mapl}
The canonical example of measure $\mu$ is simply $(\mathsf{id},\umap)_\sharp \kappa$ so that the data are
simply $\bigl\{ \bigl(y^{(n)},\umap(y^{(n)})\bigr)\bigr\}_{n=1}^N$ with the $y^{(n)}$ drawn i.i.d. from probability 
measure $\kappa$ on $\Ry.$ We can hence apply the ideas of supervised learning\index{supervised learning} from Chapter \ref{ch:SL} to learn the
map $y \mapsto \umap(y)$ from class $\cRt.$
The existence of such paired data will typically have required knowledge of the likelihood and prior,
as assumed explicitly in Data Assumption \ref{da:wnl}, but the methodology may be applied simply by working
with pairs $(y,\umap)$  drawn from any measure $\mu.$ 
\end{remark}

\begin{example} \label{ex:mapl}
As just discussed, when the MAP estimator is a deterministic function of
the data, the joint measure is $\mu=(\mathsf{id},\umap)_\sharp\kappa$. 
However, since the output of the MAP estimation algorithm\index{algorithm!random} will be random if computed with stochastic optimization algorithms such as stochastic gradient descent\index{stochastic gradient descent} (Chapter \ref{ch:optimization}), we work with a general measure $\mu.$ Indeed it is possible to generalize away from using the MAP estimator, and simply use the joint measure $\gamma$ underlying the inverse problem, 
as we will do later in Data Assumption \ref{da:4b}. 
\end{example}

Given measure $\mu$, we may try and learn the dependence of the \index{MAP estimator}MAP estimator on observations as a supervised learning problem; see Chapter~\ref{ch:SL} and Remark \ref{rem:slp}. To this end we seek
a parameterized function $u \in \cRt$. Recall Definition \ref{def:det_score} from Subsection \ref{ssec:dldsc}
of distance-like deterministic scoring rules $\dd$\index{scoring rule!distance-like deterministic}. 
We use the scoring rule to compare the point estimator $u(\cdot\,;\theta) \in \cRt$, evaluated
at point $y$, to $\umap$.
We then define the optimal value $\thetas \in \Theta$ for $u(\cdot\,; \theta) \in \cRt$, by
\begin{align}\label{eq:MAPL}
\begin{split}
\J(\theta) &= \bbE^{(y,\umap) \sim \mu} \, \bigl[\dd\bigl(u(y;\theta),\umap\bigr) \bigr],\\
\thetas &\in \argmin_{\theta \in \Theta} \J(\theta),
\end{split}
\end{align}
and set $u^\star=u(\cdot\,;\theta^\star).$ Example \ref{ex:crt} provides a class $\cRt$ in which to seek solutions capturing the
$\theta$-dependence. In practice we replace the expectation over $\mu$ in \eqref{eq:MAPL} with an expectation with respect to $\mu^N,$ the empirical\index{empirical} measure defined by $\bigl\{ \bigl(y^{(n)},\umap^{(n)} \bigr)\bigr\}_{n=1}^N$.

\begin{example}
Assume that the algorithm to find the MAP estimator from observed data is defined by a deterministic function $\mathsf{map}:\R^k \to \R^d.$ 
Then $\mu=(\mathsf{id},\mathsf{map})_\sharp\kappa$ is the pushforward of
$\kappa$ under the map $y\mapsto\bigl(y,\mathsf{map}(y)\bigr)$.
The
optimization problem becomes
\begin{align}\label{eq:MAPLE}
\begin{split}
\J(\theta) &= \bbE^{y \sim \kappa} \, \bigl[\dd\bigl(u(y;\theta),\mathsf{map}(y)\bigr) \bigr],\\
\thetas &\in \argmin_{\theta \in \Theta} \J(\theta).
\end{split}
\end{align}

\end{example}

\subsection{Amortized Posterior Mean\index{posterior mean} Estimation}
\label{ssec:CME}

In this subsection we consider amortizing the posterior mean estimator.\index{posterior!mean estimation} We assume that we have samples from the joint distribution on observations and parameters:
\begin{dataassumption}\index{Data Assumption}
\label{da:4b} We are given samples $\{(y^{(n)},u^{(n)})\}_{n=1}^N$ assumed to be drawn i.i.d. from the joint probability measure $\joint$ on $(y,u) \in \Ry \times \Ru$. 
\end{dataassumption}

Recall that we denote the marginal of $\gamma$ on $y \in \Ry$ by $\marg$.
The following theorem shows that the posterior mean is the minimizer of the squared error, in expectation over the joint distribution $\joint$.

\begin{theorem}\label{th:postmean_opt}\index{posterior!mean estimation}
Let $v: \Ry \to \Ru$ denote any $\marg$-measurable function. Then
\begin{equation}
    \Expect^{(y, u)\sim \gamma} \left[|v(y) - u|^2\right] \geq \Expect^{(y, u)\sim \gamma} \left[|\upm(y) - u|^2\right].
\end{equation}
Furthermore, the inequality is strict if $v(y)\neq \upm(y)$ for $\kappa$-almost every $y$.
\end{theorem}
\begin{proof}
From Theorem~\ref{th:postmean}, for any $y \in \R^\dy$ we have that 
\begin{equation*}
    \mathbb{E}^{u\sim \post^y} \left[|v(y) - u|^2\right]\geq \mathbb{E}^{u\sim \post^y} \left[|\upm(y) - u|^2\right]
\end{equation*}
Taking the expectation with respect to $y\sim\kappa$ of both sides of this inequality, we obtain the desired inequality. It also follows that the minimizer $\upm(y)$ is unique.
\end{proof}

We now consider finding a parameterized function $\upm(\cdot\,; \theta) \in \cRt$ to approximate the posterior mean\index{posterior!mean estimation}. Theorem~\ref{th:postmean_opt} suggests the following optimization problem:
\begin{subequations} \label{eq:slo}
    \begin{align}
        \J(\theta) &= \Expect^{(y, u)\sim \gamma} \left[|\upm(y; \theta) - u|^2\right],\\
        \thetas &\in \argmin_{\theta\in\Theta} \J(\theta).
    \end{align}
\end{subequations}
In practice, the optimization can be carried out by empirically approximating the expectation using the samples $\{(y^{(n)},u^{(n)})\}_{n=1}^N$ from Data Assumption~\ref{da:4b}.

\begin{remark}
In fact, \eqref{eq:slo} is the standard supervised learning\index{supervised!learning} objective to learn a function mapping $y$ to $u$, given data from a joint distribution $\gamma$ on the pair $(y,u).$ Supervised learning is the topic of Chapter \ref{ch:SL}; but note that in that chapter we learn functions mapping $u$ to $y$ and that we work under a more restrictive assumption on the structure of the joint distribution.
\end{remark}

\section{Amortizing Variational Inference\index{variational inference!amortized}} \label{sec:likelihood-based}

In this section we go beyond amortized point estimators, and generalize variational 
inference, introduced in Chapter \ref{ch:VI}, to allow for dependence of the approximate posterior 
distribution on the data $y$. This leads to the concept of amortized variational 
inference.\index{variational inference!amortized}
To this end we introduce parameter set $\Theta \subseteq \R^p$ and
let $\cQt \subset \cP(\R^d)$ denote
a parameterized set of probability density functions with property that 
$q(\cdot\,; y, \theta) \in \cP(\R^d)$ for all $(y,\theta) \in \R^k \times \Theta.$
Thus we aim to learn dependence of the approximate posterior on the observed data $y$. To do this
we need to parameterize not only the probability measure, but its dependence on $y$.

\begin{example}
\label{ex:AVIG}
The amortized variational inference problem can be implemented by defining $\cQt$ to contain Gaussians,
where the parameters defining the Gaussian are themselves learnable functions of the observation $y$. 
This is a generalization of the family of Gaussians in Subsection~\ref{ssec:gauss}, to include $y$-dependence of the mean and covariance. 
It leads to a density of the form
\begin{equation}
q(u;y,\theta)  = \frac{1}{\sqrt{(2\pi)^d \det \Sigma(y;\theta)}}\exp \left( - \frac{1}{2} \bigl|u - m(y;\theta)\bigr|_{\Sigma(y;\theta)}^2 \right).
\end{equation}
The dependence of mean and covariance on $y$ will itself need to be parameterized,
for example  using neural networks, random features, or Gaussian processes (see Chapter \ref{ch:SL}).
Thus parameterizations will need to be constrained to ensure the covariance is a positive definite 
matrix for all $y \in \R^k$; see Example \ref{ex:cholGVI} for an illustration of how this may be
achieved.
\end{example}

We study variational inference in a general class $\cQt$, noting that this defines a set of probability measures depending on $y$; the parameter $\theta \in \Theta$ is selected to learn the best approximation from within this class. 
To learn $y$-dependence we require information about the inverse problem for multiple realizations of the data. 
In Subsection \ref{ssec:myo} we return to work under Data Assumption \ref{da:wnl}, assuming only access to
observational data $\{y^{(n)}\}_{n=1}^N$ from the marginal $\kappa.$ 
We make Data Assumption \ref{a:LBI} in Subsection \ref{ssec:myu2}: we assume that, in addition to the samples from $\kappa$, 
we have samples $\{u^{(n)}\}_{n=1}^N$ from the prior $\pr.$ 
Finally, in Subsections \ref{ssec:sjoint} and \ref{ssec:sjoint_energy}, we work under Data Assumption \ref{da:4b} in a likelihood-free setting, assuming samples
$\{\bigl(y^{(n)},u^{(n)}\bigr)\}_{n=1}^N$ from the joint distribution $\gamma.$
In Subsection \ref{ssec:sjoint} we use an objective function defined using
KL divergence\index{divergence!Kullback–Leibler} and in Subsection \ref{ssec:sjoint_energy} we use strictly proper
scoring rules.\index{scoring rule!strictly proper}

\subsection{Samples From the Marginal on $y$}
\label{ssec:myo}

In this subsection we work again under Data Assumption \ref{da:wnl}.
With the goal of generalizing variational inference to learn the dependence of the optimal approximation 
on the observational data $y$, we define the optimization problem, for $q(\cdot\,; y, \theta) \in \cQt,$ by
\begin{subequations}
\label{eq:gvi}
\begin{align}
    \F(\theta) & = \mathbb{E}^{y \sim \marg} \bigl[\dkl\bigl(q(\cdot\,;y, \theta\bigr)\|\post^y)\bigr],\\
    \thetas & \in \argmin_{\theta \in \Theta} \F(\theta).
\end{align}  
\end{subequations}
By averaging the standard divergence\index{divergence} used to define variational inference\index{variational inference}
over $y$ drawn from $\kappa$ we aim to find an element in $\cQt$ that approximates the $y$-dependence of a variational
minimizer, over the support of $\kappa$.
Using the calculations underlying the proof of Theorem \ref{t:VariationalForm_Bayes}
we see that this optimization problem is equivalent to
\begin{subequations}
\label{eq:gvix}
\begin{align}
    \J(\theta) & =
\mathbb{E}^{y\sim\marg}
\left[
\dkl\bigl(q(\cdot\,;y,\theta)\|\pr\bigr)
-
\mathbb{E}^{u\sim q(\cdot\,;y,\theta)}
\bigl[\log\like(y|u)\bigr]
\right],
    \\
    \thetas & \in \argmin_{\theta \in \Theta} \J (\theta).
\end{align}  
\end{subequations}
Remark \ref{rem:zy}, suitably modified, also applies here: the
normalization constant for the posterior depends on $y$, however, it is independent of $\theta$ and hence not relevant to the optimization.

The objective function $\J(\cdot)$ can be evaluated approximately, using samples to approximate the expectation with respect to $y$,
under Data Assumption \ref{da:wnl}. 
We refer to Example \ref{ex:AVIG} for an explicit actionable choice of $\cQt.$
However, typically some quadrature will be needed, in particular 
to evaluate expectations against $q(\cdot\,;y, \theta)$, as discussed in the introduction to
standard variational inference in Section \ref{sec:vvar}.
This is the case for the Gaussian expectations arising from Example \ref{ex:AVIG}.

\subsection{Samples From the Marginals on $y$ and on $u$}
\label{ssec:myu2}

We now make further demands on our available data, extending Data Assumption \ref{da:wnl}
to work under the assumption that we also have prior samples available:

\begin{dataassumption}\index{Data Assumption}
\label{a:LBI}
We are able to evaluate the likelihood $\like(y|u)$ and prior $\pr(u)$ for any
pair $(y,u) \in \R^\dy \times \R^\du.$ We are given data in the form of independent samples $\{y^{(n)}\}_{n=1}^N$ 
from the marginal distribution $\marg$ and independent samples $\{u^{(n)}\}_{n=1}^N$ from the prior distribution $\pr.$
\end{dataassumption}

Generalizing the transport approach to variational inference described in Section \ref{sec:LPrtoPo}, we can define the family of approximating distributions by using a transport map that pushes forward the prior $\pr$ to the posterior $\post^y$, and is also parameterized by $y.$ In this case, we seek a class $\cTt$ of transport maps $T \colon \R^{\du} \times \R^{\dy} \times \Theta \rightarrow \R^d$ that depend on both input parameters and observations so that $u \mapsto T(u;y,\theta)$ defines a transport map that is a $C^1$-diffeomorphism\footnote{Recall that a diffeomorphism is a $C^1$ map whose inverse is also $C^1$.}
with positive Jacobian determinant, which (approximately) pushes forward the prior 
$\pr$ to the posterior $\pi^y$, for each choice of the observations $y \in \R^k$ and $\theta \in \Theta$:
\begin{equation}
\begin{aligned}\label{eq:thisclass}
&\cTt \subset \Bigl\{T(\cdot\,;y,\theta) \in C^1(\R^d;\R^d),\,\forall(y,\theta) \in \R^k \times \Theta \,\\
&\hspace{1.0in}\Bigl| \det D_u T(u;y,\theta) > 0, \,\forall(u,y,\theta) \in \R^d \times \R^k \times \Theta \Bigr\}.
\end{aligned}
\end{equation}

\begin{example}
Examples of such classes of maps may be found in Section \ref{sec:NORM} where normalizing flows\index{flow!normalizing} are introduced.
In particular, neural ODEs\index{neural ODEs}, 
introduced in Subsection \ref{sec:nODEs}, provide an example of $\cTt.$ In the current context they need to be
generalized to include parametric dependence of the defining vector fields on $y$.
We introduce a family of neural networks $h:\R^d \times \R^k \times \Theta \to \R^d$, as discussed in Remark \ref{rem:NNG} and
the material preceding it. We then consider the ODE
\begin{align}
\label{eq:nfode9999}
\begin{split}
\frac{dv}{dt}&=h(v;y,\theta),\\
v(0)&=u.
\end{split}
\end{align}
Noting that $v(t)$ depends on $(u,y,\theta)$, we set 
 $T(u;y,\theta)=v(1).$ This defines class $\cTt.$ The discussion in Subsection~\ref{sec:nODEs} 
on inversion of the map defined by the neural ODE also applies here.
\end{example}

We seek this transport in $\cTt$ for $\Theta \subseteq \R^p$ as the minimizer of the following objective
\begin{subequations}
\label{eq:thetaVIy}
\begin{align}
    \F(\theta) 
    &=\mathbb{E}^{y \sim \marg}\bigl[\dkl\bigl(T(\cdot\, ;y,\theta)_\sharp \pr\|\pi^y\bigr)\bigr]\\
    &=  \mathbb{E}^{y \sim \marg}\bigl[\dkl\bigl(\pr \|T^{-1}(\cdot\, ;y,\theta)_\sharp\pi^y\bigr)\bigr],\\
\thetas &\in \argmin_{\theta \in \Theta} \F(\theta).
\end{align}  
\end{subequations}
In going from (\ref{eq:thetaVIy}a) to (\ref{eq:thetaVIy}b) we have used the
invariance of the KL divergence under invertible and differentiable transformations from Theorem~\ref{thm:invariance}. 
By specifying a class of transport maps, we have implicitly defined a class of $y$-dependent probability
measures $\cQt$ over which the amortized variational inference\index{variational inference!amortized}
is performed. This is the amortized analog of the discussion in Subsection \ref{ssec:cvi}.

Following the approach in Chapter~\ref{ch:transport}, and equation \eqref{eq:subeqp} in particular, we have 
\begin{align*}
\dkl\bigl(\pr \|T^{-1}(\cdot\, &;y,\theta)_\sharp\pi^y\bigr) 
=\mathbb{E}^{u\sim\pr}[\log\pr(u)] + \log Z^y \\
&- \bbE^{u \sim \pr} \Bigl[\log \pr\bigl(T(u;y,\theta)\bigr) + 
\log \like\bigl(y|T(u;y,\theta)\bigr)+\log {\rm det}  D_u T(u;y,\theta) \Bigr],
\end{align*}
where the first and second terms on the right-hand side of the equality, the negative entropy of $\rho$ and $\log Z^y$, are constant with respect to $\theta$.
Thus, we can rewrite the optimization problem \eqref{eq:thetaVIy} as a problem that may be 
approximated\index{empirical} empirically under Data Assumption \ref{a:LBI}:
\begin{subequations}
\label{eq:thetaVIy2}
\begin{align}
    \J(\theta) 
    &=-\mathbb{E}^{(y,u) \sim \marg \otimes \pr}\Bigl[\log\pr\bigl(T(u;y,\theta)\bigr) + \log\like\bigl(y|T(u;y,\theta)\bigr) + \log\det D_u T(u;y,\theta) \Bigr],\\
   \thetas &\in \argmin_{\theta \in \Theta} \J(\theta).
\end{align}  
\end{subequations}
Note that this optimization problem is the analog of~\eqref{eq:subeqp}, arising in the case of a fixed single instance of data $y.$

\subsection{Samples From the Joint Distribution on $(y,u)$; Kullback--Leibler\index{divergence!Kullback--Leibler}}
\label{ssec:sjoint}

In this subsection we assume that samples are available from the joint distribution underlying the Bayesian inverse problem defined by equation \eqref{eq:jc0}; that is, we make Data Assumption~\ref{da:4b}, which does not require
explicit knowledge of the prior or the likelihood.
We have already shown that it is possible to
learn the posterior mean, and its dependence on observation $y$, from this
data set (Subsection~\ref{ssec:CME}); we now show that it is in fact possible
to learn the posterior distribution from such data.

We do not assume that the likelihood is defined by equation~\eqref{eq:jc0} and Assumption \ref{a:jc1}. While the methodology in this subsection also applies in that specific setting, it is much more general. Indeed, since the methodology in this subsection depends only on having paired $(y,u)$ samples, it is a form of \emph{likelihood-free inference}\index{likelihood!free}. This setting often arises when the forward model involves complex simulations; such methods are thus also called \emph{simulation-based inference}\index{simulation-based inference}. In particular, the methodological approach introduced in this subsection will be used for simulation-based inference problems appearing in the context of \emph{data assimilation} in Subsection~\ref{sec:TA}.

\subsubsection{General Formulation}

Recall the parameterized subset $\cQt$ of $\cP(\R^d)$ in which 
we seek to solve the variational Bayesian problem. We define the optimization problem, for
$q(\cdot\,;y, \theta) \in \cQt$,
\begin{subequations}
\label{eq:gvi22}
\begin{align}
    \F(\theta) & = \mathbb{E}^{y \sim \marg} \bigl[\dkl\bigl(\post^y\|q(\cdot\,;y, \theta)\bigr)\bigr],\\
    \thetas & \in \argmin_{\theta \in \Theta} \F(\theta).
\end{align}  
\end{subequations}
Note that this is identical to optimization problem \eqref{eq:gvi}, 
which we employ when making Data Assumption \ref{da:wnl}, except that we have used the reversed
Kullback--Leibler divergence\index{divergence!Kullback--Leibler}. This reversal has profound effects 
on when this optimization problem can serve as the basis for computational methods.
To see this, note that
\begin{align*}
\F(\theta) & = \mathbb{E}^{y \sim \marg} \mathbb{E}^{u \sim \post^y}   \bigl[\log\bigl(\post^y(u)\bigr)-\log\bigl(q(u;y, \theta)\bigr)\bigr]\\
 & = \mathbb{E}^{(u,y) \sim \gamma} \bigl[\log\bigl(\post^y(u)\bigr)-\log\bigl(q(u;y, \theta)\bigr)\bigr].
\end{align*}
Since the first term in the resulting expression is independent of $\theta$,
we see that optimization problem~\eqref{eq:gvi22} is equivalent to
\begin{subequations}
\label{eq:gvix22}
\begin{align}
    \J(\theta) & = - \mathbb{E}^{(u,y) \sim \gamma} \bigl[\log\bigl(q(u;y, \theta)\bigr)\bigr],\\
    \thetas & \in \argmin_{\theta \in \Theta} \J(\theta).
\end{align}  
\end{subequations}

\begin{remark}
As mentioned following Data Assumption \ref{da:4b}, the methodology in this subsection does not require knowledge of a likelihood.
However it is possible that the samples in Data Assumption~\ref{da:4b} may have been generated by sampling $u^{(n)}$ from the prior $\rho$ 
and then sampling synthetic observations from a likelihood model $y^{(n)} \sim \like(\cdot \;|u^{(n)})$. 
Even if the samples are generated from the likelihood model, this does not necessarily imply that the likelihood is available in closed form, as we explain in the next remark.
\end{remark}

\begin{remark} \label{rem:integrated_like}
One setting in which the likelihood $\like(y|u)$ is not analytically available or tractable to evaluate, but is straightforward
to sample, arises when the likelihood is defined through marginalization with respect 
to a latent random variable $z.$ That is,
\begin{equation} \label{eq:likelihood_latent}
  \like(y|u) = \int_{\R^{\dz}} \bbP(y|u,z)\bbP(z|u)\,dz.  
\end{equation}
When $\dz \gg 1$, it will be difficult to evaluate the integral over $z.$ But
we can simulate from the joint distribution $\bbP(y,u) = \like(y|u)\bbP(u),$ with likelihood 
as in \eqref{eq:likelihood_latent}, as follows. First sample $u$ from the prior $\pr$; 
then sample latent state $z$ from $\bbP(z|u)$;  and finally sample $y$ from 
$\bbP(y|u,z)$. The collection $(y,z,u)$ is a sample from the distribution $\Prob(y,z,u)$. The subset of pairs $(y,u)$ are then samples from the distribution $\like(y|u)\pr(u)$. 

Note that in \eqref{eq:likelihood_latent}, the object of interest is obtained by integrating out a (potentially high dimensional) latent variable. This same structure underlies
the expectation maximization\index{expectation maximization} methodology, 
described in Chapter \ref{ch:optimization}, 
and applied in Chapter \ref{ch:DAML}. The next example illustrates this structure.

\end{remark}

\begin{example} \label{ex:integrated_like}
A likelihood of the form in~\eqref{eq:likelihood_latent}, with high-dimensional latent variable, arises when performing parameter inference in a \emph{hidden Markov model}\index{hidden!Markov model}; noisy observations $y_{j} \in \R^\dy$ are given of
a stochastic dynamical system for hidden state $v_j \in \R^{\du_v}$ as: 
\begin{align*}
    v_{j+1} & \sim  p_{\mathsf{dyn}}(\cdot\,|v_{j};u), \\
    y_{j+1} & \sim  p_{\mathsf{obs}}(\cdot\,|v_{j+1}).
\end{align*}
Here $v_0 \sim p_{\mathsf{0}}$ is drawn at random from a known probability density 
and
$p_{\mathsf{dyn}}$ is a Markov kernel, parameterized by unknown $u$, defining the evolution of
the hidden variable $V = (v_0,\dots,v_J)$. The observations $Y = (y_1,\dots,y_J)$ are defined by the
known Markov kernel $p_{\mathsf{obs}}.$ We are interested in determining $u$ from $Y$, and the 
likelihood of $Y|u$ has the form 
\begin{equation}
\label{eq:cpf}    
\like(Y|u) = \int_{\R^{\du_v(J+1)}} \bbP(Y|u,V)\bbP(V|u)\,dV.
\end{equation}
This likelihood is typically not available in explicit form, as we now illustrate in a concrete setting.

Hidden Markov models are the common framework for \emph{data assimilation}\index{data!assimilation}, which is explored in Chapters~\ref{lecture7}--\ref{ch:LG10}. We follow the formulation in Chapter~\ref{lecture7},  but include an unknown parameter $u$, to be determined, to define the dynamics. The states $v_j \in \R^{\du_v}$ and observations $y_{j} \in \R^\dy$  follow the dynamics and observation models:
\begin{align*}
    v_{j+1} &= \Psi(v_{j};u) + \xi_j, \quad \xi_j \sim \mathcal{N}(0,\Sigma), \\
    y_{j+1} &= h(v_{j+1}) + \eta_{j+1}, \quad \eta_{j+1} \sim \mathcal{N}(0,\Gamma), 
\end{align*}
for $j=0,\dots,J-1$ starting from initial condition $v_0 \in \R^{d_v}$ drawn from a known Gaussian
$\mathcal{N}(m_0,C_0).$
Here, $\Psi \colon \R^{\du_v} \times \R^{\du} \rightarrow \R^{\du_v}$ is a forward model depending on the unknown parameter $u \in \R^{\du}$. Function $h \colon \R^{\du_v} \rightarrow \R^{\dy}$ is a known observation operator. 
All random variables in the noise sequences $\{\xi_j\}$ and $\{\eta_{j}\}$ are assumed to be independent.

The conditional probabilities required to define~\eqref{eq:cpf} have the closed forms
\begin{align*} \label{eq:likelihood_latent_elements}
   \bbP(Y|u,V) &= \prod_{j=1}^{J} \mathcal{N} \bigl(y_{j}; h(v_j),\Gamma \bigr), \\
   \bbP(V|u) &= \Bigl(\prod_{j=0}^{J-1} \mathcal{N} \bigl(v_{j+1}; \Psi(v_j;u),\Sigma \bigr)\Bigr) \mathcal{N}
   \bigl(v_{0}; m_0, C_0\bigr).
\end{align*}
The desired likelihood for $Y|u$ is given by the integral in~\eqref{eq:cpf}. 
Comparing to \eqref{eq:likelihood_latent}, we see that the hidden states $V$ are the latent variables for such a problem. The integration over $V$ that is required to render the likelihood in explicit form is not analytically tractable when either $\Psi$ or $h$ are nonlinear. Hence, such settings require numerical approximations that are computationally expensive for large $J$. This motivates the use of likelihood-free\index{likelihood!free} inference methods that do not require (approximate) evaluations of $\like$. A different approach to learn parameters in hidden Markov models\index{hidden!Markov model}, based on \index{expectation maximization} expectation maximization, is presented in Chapter~\ref{ch:DAML}.
\end{example}

\subsubsection{Transport-Based Formulation}

Here we study the solution of optimization problem~\eqref{eq:gvix22} through the use of a prior-to-posterior 
transport\index{transport!prior-to-posterior} map. As in Subsection \ref{ssec:myu2}, we generalize the transport approach: we seek parameter $\theta \in \Theta$ in map $T \colon \R^{\du} \times \R^{\dy} \times \Theta \rightarrow \R^d$ so that $u \mapsto T(u;y,\theta)$ defines an invertible transport map approximating the pushforward of prior $\pr$ to the posterior $\pi^y(u)$, for each choice of the observations\index{observation} $y$. 
With the selected choice of $\theta$, the resulting map depends on the input state $u$
and the observations $y$.\index{observation} 

We seek this transport as the minimizer of the following objective, obtained from equation~\eqref{eq:gvix22} by taking the distribution $q(\cdot;y,\theta) \in \cQt$ to be the pushforward of $\rho$ through $T(\cdot;y,\theta)$:
\begin{subequations}
\label{eq:NNL_map_posterior_approx}
\begin{align}
\J(\theta) & = - \mathbb{E}^{(u,y) \sim \gamma} \bigl[\log \bigl(T(\cdot\, ;y,\theta)_\sharp \pr(u) \bigr) \bigr]\\
&=  - \mathbb{E}^{(u,y) \sim \gamma} \Bigl[\log\pr\bigl(T^{-1}(u;y,\theta)\bigr) + \log\det D_u (T)^{-1}(u;y,\theta)\Bigr],\\
\thetas &\in \argmin_{\theta \in \Theta} \J(\theta),
\end{align}  
\end{subequations}
where $\Theta \subseteq \R^p.$ To express the objective, we have used (\ref{eq:cov}b), which provides an expression for the
density of $T(\cdot\, ;y,\theta)_\sharp \pr.$ If we make
Data Assumption~\ref{da:4b} this optimization problem can be approximated by sampling, provided that we also have access 
to a formula for the prior density $\rho.$  The empirical optimization problem takes the form
$$\thetas \in \argmin_{\theta\in\Theta} \frac{1}{N} \sum_{n=1}^N -\Bigl[\log\pr\bigl(T^{-1}(u^{(n)};y^{(n)},\theta)\bigr) + \log\det D_u (T)^{-1}(u^{(n)};y^{(n)},\theta)\Bigr].$$
The optimal solution $\thetas$ computed in practice corresponds to solving the \emph{maximum likelihood}\index{maximum!likelihood} problem for the parameters under the model $T(\cdot\, ;\cdot,\theta)_\sharp \rho$. See Section~\ref{sec:den} for more discussion on maximum likelihood estimation. 

\begin{remark} When a formula for the prior density is not available, we can seek the transport $T(\cdot;y,\theta)$ that approximately pushes forward a general reference measure $\refT \in \cP(\R^\du)$ to the posterior $\post^y$ for each $y \in \R^\dy$. That is, $\post^y(u) \approx (T(\cdot;y,\theta)_\sharp \refT)(u)$. The resulting optimization problem for $T$ has the form in~\eqref{eq:NNL_map_posterior_approx} with $\pr$ replaced by $\refT$. This setting will be considered in Subsection~\ref{sec:TA}.
\end{remark}

\begin{remark}
    \label{rem:worthnoting}
The objective in optimization problem~\eqref{eq:NNL_map_posterior_approx} only depends on the inverse map $T^{-1}(u; y,\theta)$. Hence, rather than parameterizing $T$ and evaluating the inverse, we can directly parameterize the inverse map $S(\cdot\, ;y,\theta) \coloneqq T^{-1}(\cdot\, ;y,\theta)$. This choice avoids inversion during learning; however, it requires inverting the map $S$ to sample from the posterior $\pi^y$ after learning. 
\end{remark}

The following theorem provides a closed-form solution for optimization problem~\eqref{eq:NNL_map_posterior_approx} when optimizing over a space of affine maps in $u$ and $y$.
Recall that $\bbR^{d \times d}_{\textrm{sym},>}$ denotes the subset of
symmetric positive definite matrices.

\begin{theorem} \label{thm:ClosedFormAffineTransportMaps} 
Let $\refT(u) = (2\pi)^{-d/2}\exp(-\frac{1}{2}|u|^2)$ be the standard Gaussian reference 
density and let the matrices $\Sigma_{uu}$, $\Sigma_{yy}$ and $\Sigma_{uy}$ denote the covariance of $u$, covariance of $y$ and cross-covariance of $(u,y)$ under $\joint$, respectively. Assume that $\Sigma_{yy}$ is invertible and that $\Sigma_{u|y} \coloneqq \Sigma_{uu} - \Sigma_{uy}\Sigma_{yy}^{-1}\Sigma_{yu}$
is strictly positive definite. 

Consider the family of affine maps given by
$$\cTt = \Bigl\{T(u;y,\theta) = Au + By + c, \; A \in \bbR^{d \times d}_{\textrm{sym},>}, B \in \R^{\du \times \dy}, c \in \mathbb{R}^\du \Bigr\},$$
parameterized by $\theta := (A,B,c)$. Then, the optimal map solving~\eqref{eq:NNL_map_posterior_approx} in this family has the form
$$T(u;y;\thetas) = \Sigma_{u|y}^{1/2}u + \Sigma_{uy}\Sigma_{yy}^{-1}(y - \mathbb{E}[y]) + \mathbb{E}[u].$$
\end{theorem}
\begin{proof} 
Optimization problem~\eqref{eq:NNL_map_posterior_approx} for affine maps $T \in \cTt$ has the form 
\begin{align*}
(A^\star,B^\star,c^\star) \in \argmin_{A,B,c} \mathbb{E}^{(y,u) \sim \joint}\left[\frac12(u - By - c)^\top A^{-\top} A^{-1}(u - By - c) + \log\det A \right]. 
\end{align*}
For fixed $A$ and $B$, taking the gradient of the objective with respect to $c$ and setting it equal to zero, we have 
\begin{align*}
    c^\star &= \mathbb{E}[u] - B\mathbb{E}[y].
\end{align*}
Substituting the optimal $c^\star$ in the objective, the loss for $B$ given a fixed $A$ is
\begin{align*}
\loss(B;A)
\coloneqq{}&
\frac12 \mathbb{E}_{(y,u)\sim\joint}\!\left[
\bigl(u-\mathbb{E}[u]-B(y-\mathbb{E}[y])\bigr)^\top
A^{-\top}A^{-1}
\bigl(u-\mathbb{E}[u]-B(y-\mathbb{E}[y])\bigr)
\right] \\
& \qquad \qquad + \log\det A .
\end{align*}
Taking the gradient with respect to $B$ and setting it equal to zero, we have $\nabla_B \loss(B;A) = A^{-\top} A^{-1}(B \Sigma_{yy} - \Sigma_{uy}) = 0.$ 
Rearranging for $B$ gives us
$$B^\star = \Sigma_{uy}\Sigma_{yy}^{-1}.$$
Substituting the optimal $B^\star$ in the loss $\loss$, we notice that the optimization problem for $A^{-\top} A^{-1}$ corresponds to the Gaussian variational inference problem in~\eqref{eq:GaussianVI_objective}. The optimal solution for $A^\star$ is given by:
$$(A^\star)^{-1}(A^\star)^{-\top} = \Bigl(\mathbb{E}\Bigl[ \bigl(u - \mathbb{E}[u] - \Sigma_{uy}\Sigma_{yy}^{-1}(y - \mathbb{E}[y]) \bigr)\bigl(u - \mathbb{E}[u] - \Sigma_{uy}\Sigma_{yy}^{-1}(y - \mathbb{E}[y])\bigr)^\top \Bigr]\Bigr)^{-1}.$$
Expanding the squared terms yields
\begin{align*}
   (A^\star)^{T} (A^\star)  &= \Sigma_{uu} - \Sigma_{uy}\Sigma_{yy}^{-1}\Sigma_{yu} - \Sigma_{uy}\Sigma_{yy}^{-1}\Sigma_{yu} + \Sigma_{uy}\Sigma_{yy}^{-1}\Sigma_{yy}\Sigma_{yy}^{-1}\Sigma_{yu} \\
   &=\Sigma_{uu} - \Sigma_{uy}\Sigma_{yy}^{-1}\Sigma_{yu} \\
   &= \Sigma_{u|y},
\end{align*}
which gives us the final result, $A^\star = \Sigma_{u|y}^{1/2},$ in terms of the square root of the conditional covariance matrix, after noting that we
seek $A$ as an element of $\bbR^{d \times d}_{\textrm{sym},>}$.
\end{proof}

\subsection{Samples From the Joint Distribution on $(y, u)$; Strictly Proper Scoring Rules}\label{ssec:sjoint_energy}

As in the preceding subsection, we again seek an amortized approximation to the posterior distribution in the setting of Data Assumption~\ref{da:4b}. However,
we now make use of strictly proper scoring rules\index{scoring rule!strictly proper} from Definition~\ref{def:proper_score}, 
rather than the KL divergence,\index{divergence!Kullback–Leibler}  to define the
objective function. Strictly speaking this takes us outside the standard setting of variational inference,
which is traditionally based on use of the KL divergence.\index{divergence!Kullback–Leibler}

\subsubsection{General Formulation}

Consider a joint distribution $\joint$, with underlying marginal distribution $\marg$ on the observations $y$, and conditional (posterior) $\pi^y$ on the parameter $u$. The following theorem shows that the posterior is the minimizer of a proper scoring rule, averaged through an expectation over the joint distribution $\joint$.

\begin{theorem}\label{th:scoring_rule_joint}
     Consider a strictly proper scoring rule\index{scoring rule!proper}  $\mS(\cdot, \cdot): \cP(\R^{\du}) \times \R^{\du} \to \R.$  Let $q(\cdot\,; y)$ denote any distribution
 on $\R^{\du}$ parameterized by $y$. Then, we have
\begin{equation}
\mathbb{E}^{(u, y)\sim\joint} \Bigl[\mS(q(\cdot\,;y), u)\Bigr]\geq \mathbb{E}^{(u, y)\sim\joint} \Bigl[\mS(\post^y, u)\Bigr].
\end{equation}
Furthermore, the inequality is strict if $q(\cdot\,; y)\neq \post^y$.
\end{theorem}
\begin{proof}
From the definition of a proper scoring rule, we have that
\begin{equation*}
    \mathbb{E}^{u\sim \post^y} \Bigl[\mS(q(\cdot\,; y), u)\Bigr]\geq \mathbb{E}^{u\sim \post^y} \Bigl[\mS(\post^y, u)\Bigr].
\end{equation*}
Taking the expectation with respect to $y\sim\marg$ of both sides of this inequality, we obtain the desired inequality. From strict propriety, we obtain uniqueness of the minimizer.
\end{proof}

As in Subsection~\ref{ssec:sjoint}, we consider a parameterized subset of $y$-dependent approximate distributions $\cQt$. Then, we can use the cost function in Theorem~\ref{th:scoring_rule_joint} to define an optimization problem over $\theta$, for any given strictly proper scoring rule $\mS$:
\begin{subequations}
\begin{align}
    \J(\theta) &= \Expect^{(u, y)\sim\joint}[\mS(q(\cdot\,; y, \theta), u)],\label{eq:amortized_score}\\
    \thetas &\in \argmin_{\theta\in\Theta} \J(\theta).
\end{align}
\end{subequations}
We may then define $\qs(\cdot\,;y):=q(\cdot\,;y,\thetas) \in \cQt$.
In practice, to approximate $\thetas$, the objective $\J(\theta)$ is
empiricalized using samples from Data Assumption~\ref{da:4b}.

\begin{example}
    If we take the scoring rule $\mS$ to be the logarithmic score\index{score!logarithmic} from Definition~\ref{d:logs}, $\LS(\rho, v) = -\log \rho(v)$, we recover the variational inference optimization problem \eqref{eq:gvix22} from the preceding subsection.
\end{example}

\begin{example}
    If we take the scoring rule $\mS$ to be the energy score\index{score!energy} from 
    Subsection~\ref{ssec:ES}, we obtain
    \begin{equation}\label{eq:energy_amortized}
        \J(\theta) = \Expect^{(u, y)\sim\joint}\Expect^{v\sim q(\cdot\,; y, \theta)} |v - u| - \frac{1}{2}\Expect^{y\sim\marg}\Expect^{(v, v')\sim q(\cdot\,; y, \theta)\otimes q(\cdot\,; y, \theta)} |v - v'|.
    \end{equation}
    The objective $\J$ can then be evaluated by expressing the expectations over distributions that do not depend on the $\theta$ parameter. This is desirable
    when it is necessary to compute the gradient of the objective function
    with respect to $\theta$, for example if using gradient-based
    optimization as in Section \ref{sec:grad}. This is often referred to as the \emph{reparameterization trick}\index{reparameterization trick}; see Subsection~\ref{ssec:reparameterization}.
\end{example}
The following example shows how to implement the optimization problem for the Gaussian variational family introduced in 
Example~\ref{ex:AVIG}.

\begin{example} \label{eq:energy_amortized_Gaussian}
When $\cQt$ is the family of multivariate Gaussian distributions $q(\cdot\,;y, \theta) = \mathcal{N}(m(y,\theta),\Sigma(y,\theta))$, we can express a sample from $q$ as 
$$m(y,\theta) + \Sigma(y,\theta)^{\frac{1}{2}} \xi \sim q(\cdot\, ;y,\theta), \qquad \xi \sim \mathcal{N}(0,I_d),$$
where $m \colon \R^k \times \Theta \rightarrow \R^d$ and $\Sigma \colon \R^k \times \Theta \rightarrow \R^{d \times d}$ are a $y$-dependent mean and covariance matrix;
note that it is desirable that parameterization of the covariance delivers a positive semidefinite matrix so that we can compute its (unique positive) square root. If we again take $\mS$ to be the energy score\index{score!energy} from Subsection~\ref{ssec:ES}, we can write the objective function \eqref{eq:energy_amortized} as
\begin{align*}       
    \J(\theta) &=  \Expect^{(u, y)\sim\joint}\mathbb{E}^{\xi \sim \mathcal{N}(0,I_d)}|m(y,\theta) + \Sigma(y,\theta)^{\frac12} \xi - u|\\
    &\qquad\qquad -\frac{1}{2}\Expect^{y\sim\marg}\mathbb{E}^{(\xi,\xi') \sim \mathcal{N}(0,I_d) \otimes \mathcal{N}(0,I_d)}|\Sigma(y,\theta)^{\frac12}(\xi - \xi')|.
\end{align*}
Given samples from Data Assumption~\ref{da:4b}, an empirical approximation of the objective function is given by
\begin{align*}       
    \J^N(\theta) &\coloneqq \frac{1}{N} \sum_{i=1}^N \left[\mathbb{E}^{\xi \sim \mathcal{N}(0,I_d)}|m(y^{(i)},\theta) + \Sigma(y^{(i)},\theta)^{\frac{1}{2}} \xi - u^{(i)}| \right. \\
    &\qquad\qquad\qquad\qquad \left. -\frac{1}{2}\mathbb{E}^{(\xi,\xi') \sim \mathcal{N}(0,I_d) \otimes \mathcal{N}(0,I_d)}|\Sigma(y^{(i)},\theta)^{\frac12}(\xi - \xi')| \right].
\end{align*}
\end{example}
The preceding example provides a methodology for amortizing inference over a Gaussian family, in which the energy score is used to define
the objective and for which only samples from the joint distribution are
required. The following subsubsection provides a methodology to extend beyond the Gaussian family of distributions.

\subsubsection{Transport-Based Formulation}

We now study the amortized posterior approximation by minimization of a strictly proper scoring rule\index{scoring rule!proper} through the use of a prior-to-posterior transport\index{transport!prior-to-posterior} map. As in Subsection \ref{ssec:sjoint}, we seek a transport map $T \colon \R^{\du} \times \R^{\dy} \times \Theta \rightarrow \R^d$ depending on the observation $y$ that defines elements of the family $\cQt$ as
\begin{equation}
    q(u;y,\theta) = \bigl(T(\cdot\, ;y,\theta)_\sharp \rho\bigr)(u).
\end{equation}
Using the pushforward expression in~\eqref{eq:amortized_score},
the objective function can be expressed as
\begin{align*}
  \J(\theta)=
\Expect^{(u,y)\sim\joint}
\Bigl[
\mS\bigl(T(\cdot\,;y,\theta)_\sharp\rho,u\bigr)
\Bigr].  
\end{align*}
In order to compute this objective under Data Assumption \ref{da:4b}, we must use a scoring rule that can be evaluated for an empirical measure. Kernel scores (Definition~\ref{def:kernel_score}), which include the energy score and the CRPS, are examples. For a choice of kernel $c: \Ru \times \Ru \to \R$, we can write the optimization problem as
\begin{subequations}
\begin{align}
    \J(\theta) &= \Expect^{(u, y)\sim\joint}\Expect^{z\sim \postr}[c(T(z; y, \theta), u)] - \frac{1}{2}\Expect^{y\sim\marg}\Expect^{(z, z')\sim \postr\otimes \postr}[c(T(z; y, \theta), T(z'; y, \theta))],\\
    \thetas &\in \argmin_{\theta\in\Theta} \J(\theta).
\end{align}
\end{subequations}
The objective is readily empiricalized, under Data Assumption \ref{da:4b}, noting that samples from $\postr$ and $\marg$ can be
found by marginalization.

\begin{remark} For a standard Gaussian prior distribution $\rho = \mathcal{N}(0,I_d)$, writing the transport as $T(z;y,\theta) = m(y,\theta) + \Sigma(y,\theta)^{\frac{1}{2}}z$  in terms of a mean $m$ and covariance $\Sigma$, and choosing the kernel to be $c(u, v) = |u - v|$, reduces to the objective in Example~\ref{eq:energy_amortized_Gaussian}. 
\end{remark}

\section{Learning Likelihood Models}
\label{sec:LLM!}

Everything in Subsections~\ref{ssec:sjoint} and \ref{ssec:sjoint_energy} is performed
under Data Assumption~\ref{da:4b}, which is symmetric with respect to the roles of $u$ and $y$.
We have used this data to learn the posterior, which is one conditional defined by the joint;
but the same ideas can also be used to learn the other conditional\index{conditional}, the likelihood defined by the joint. Learning the likelihood may be useful in inverse problems where the likelihood is unknown analytically or intractable to evaluate, as in Remark \ref{rem:integrated_like} and Example~\ref{ex:integrated_like}. It may also be natural to learn the prior (as explained in Chapter \ref{ch:priorIP}) and the likelihood (as explained here)
as the basis for sampling the posterior using MCMC\index{MCMC}; in particular this may be more
effective than learning the posterior directly from the joint, in some circumstances. Another
approach to learning the likelihood, in cases where it is unknown or intractable, is to use surrogate\index{surrogate} modelling in inference, as discussed in Section~\ref{sec:IPSU}.

To approximate the likelihood function, we introduce the parameterized subset $\cQt$ of $\cP(\R^k)$, the set of distributions on the observation space, which also depend on an unknown $u \in \R^d$. The following examples consider two families of distributions $\cQt$. First, Example~\ref{ex:map_Gaussian_lik} defines a family of Gaussian likelihood functions with a parameter-dependent forward model. Second, Example~\ref{ex:map_likelihood_approximation} defines a family of conditional distributions on the observation space via transport. 
Example \ref{ex:combine} combines elements of the two preceding examples.

\begin{example} \label{ex:map_Gaussian_lik} 
Given a family of forward models, $G \colon \Ru \times \Theta \rightarrow \Ry$, and a positive definite matrix $\Gamma \in \R^{\dy\times\dy}$, a Gaussian conditional distribution for $y|u$ with $u$-dependent mean is
\begin{align*}
    q(y;u,\theta) = \frac{1}{\sqrt{(2\pi)^k \det{\Gamma}}} \exp \left(-\frac{1}{2}|y - G(u;\theta)|_\Gamma^2 \right).
\end{align*}
That is, $q(\cdot\, ;u,\theta) = \mathcal{N}(G(u;\theta),\Gamma)$. Thus we are learning, from a parameterized class, the forward model
underlying the Bayesian inverse problem from Section \ref{sec:11}. Within this class it might also be of interest to learn the pair
$(\theta,\Gamma)$, rather than just $\theta.$
\end{example}

\begin{example} \label{ex:map_likelihood_approximation} Let $\refT(y)$ be the standard Gaussian reference measure on $\Ry$ and let $T \colon \R^k \times \R^d \times \Theta \rightarrow \R^k$
be a parameter-dependent transport that pushes forward $\refT(y)$ to $q(y;u,\theta)$ for each $u$. Then, a conditional distribution for $y|u$ is given by
\begin{equation} \label{eq:likelihood_transport_density}
q(y;u,\theta) = \refT \bigl(T^{-1}(y;u,\theta)\bigr)\det{\nabla_y T^{-1}(y;u,\theta)},
\end{equation}
where $T^{-1}(\cdot\, ;u,\theta)$ denotes the inverse of the function $y \mapsto T(y;u,\theta)$. By choosing $T(y;u,\theta) = \Gamma^{1/2}y + G(u;\theta)$, the distribution in~\eqref{eq:likelihood_transport_density} reduces to the Gaussian likelihood in Example~\ref{ex:map_Gaussian_lik}. 
\end{example}

To learn the $\theta$ parameters, we follow the approach in Subsection~\ref{ssec:sjoint} with the order of the variables $(u,y)$ reversed. Our objective is the KL divergence\index{divergence!Kullback--Leibler} from the 
likelihood function $\like(\cdot|u)$ to the approximate distribution $q(\cdot\, ;u,\theta)$:
\begin{subequations}
\begin{align} \label{eq:KL_transportmap_lik}
    \F(\theta) & = \mathbb{E}^{u \sim \rho} \bigl[\dkl\bigl(\like(\cdot|u)\|q(\cdot\,;u, \theta)\bigr)\bigr],\\
    \thetas & \in \argmin_{\theta \in \Theta} \F(\theta).
\end{align}
\end{subequations}
As in Subsection~\ref{ssec:sjoint}, this optimization problem is equivalent to minimizing an objective that only depends on the joint distribution $\joint$:
\begin{subequations}
\begin{align} \label{eq:KL_transportmap_lik_loss}
    \J(\theta) & = -\mathbb{E}^{(u,y) \sim \gamma}\bigl[\log \bigl(q(y;u, \theta)\bigr)\bigr],\\
    \thetas & \in \argmin_{\theta \in \Theta} \J(\theta).
\end{align}
\end{subequations}
Approximating the expectation using the empirical measure associated with
Data Assumption~\ref{da:4b},
we obtain an implementable algorithm to learn the likelihood. 
The following example \ref{ex:combine} combines the two preceding examples to illustrate the general methodology.

\begin{example} \label{ex:combine} For the family of parameterized transport maps in Example~\ref{ex:map_likelihood_approximation}, the objective defined by the optimization problem~\eqref{eq:KL_transportmap_lik_loss} is equivalent to
\begin{subequations}
    \begin{align} \label{eq:map_likelihood_loss}
    \argmin_{\theta \in \Theta} & \, -\mathbb{E}^{(y,u) \sim \joint} \left[ \log \refT \circ T^{-1}(y;u,\theta) + \log\det D_y T^{-1}(y;u,\theta) \right].
\end{align}
\end{subequations}
In particular, for the family of Gaussian likelihood models defined by transports of the form $T(y;u,\theta) = \Gamma^{1/2}y + G(u;\theta)$ 
with known $\Gamma \succ 0$ and a standard Gaussian distribution $\refT$, 
minimizing the loss in~\eqref{eq:map_likelihood_loss} corresponds to solving the problem 
\begin{align*}
\argmin_{\theta \in \Theta} & \,\, \mathbb{E}^{(y,u) \sim \joint} \left[\frac{1}{2} \left|T^{-1}(y;u,\theta)\right|^2 - \log \det D_y T^{-1}(y;u,\theta) \right] \\
&= \argmin_{\theta \in \Theta}  \,\, \mathbb{E}^{(y,u) \sim \joint} \left[\frac{1}{2}\left|\Gamma^{-1/2}\bigl(y - G(u;\theta)\bigr)\right|^2 - \log \det \Gamma^{-1/2} \right] \\
&= \argmin_{\theta \in \Theta}  \,\, \mathbb{E}^{(y,u) \sim \joint} \left[\frac{1}{2}\left|y - G(u;\theta)\right|_\Gamma^2 \right].
\end{align*}
Hence, finding the transport is equivalent to finding an approximate forward model by the solution of a supervised learning\index{supervised learning} problem similar
to that defined in Chapter \ref{ch:SL}. Broader families of transports then generalize the approach of learning the forward model in Section~\ref{sec:IPSU} to learning the likelihood function.
\end{example}

\section{Learning Regularizers and Priors}
\label{sec:LRP}

The three preceding sections have considered a range of data assumptions, all centered on multiple observational
data $\{y^{(n)}\}_{n=1}^N$ from the inverse problem defined by \eqref{eq:jc0}:
$$y=G(u)+\eta.$$
We have used this data to learn about the posterior distribution or likelihood. 
In this section we use such data
to learn about regularizers or priors for the inverse problem. 
This is different from Chapter \ref{ch:priorIP} where we discussed
learning the prior, or regularizer for MAP estimation; in that
chapter the training data from which we learn or define the prior was 
unsupervised data $\{u^{(n)}\}_{n=1}^N$ or, in the case of Bayesian 
hierarchical modelling, simply the single observation $y$. Here the use of
multiple observational
data realizations $\{y^{(n)}\}_{n=1}^N$ can be thought of as limiting over-fitting
to one specific observed outcome.
In Subsection \ref{ssec:RMAP} we consider regularizers for MAP estimation
and in Subsection \ref{ssec:carGB} we study priors for Bayesian inversion.

\subsection{Learning Regularizers for MAP Estimation\index{MAP estimator}}
\label{ssec:RMAP}

In this section  we describe the idea of \emph{bilevel optimization}\index{optimization!bilevel} to determine
parameter $\theta \in \Theta \subseteq \R^p$ defining a regularizer in MAP\index{MAP estimator} estimation. 
We return to work under a modification of Data Assumption \ref{da:4b} to include evaluation of likelihood
and a parameterized prior -- Data Assumption \ref{da:4c}.

Paired $(y,u)$ data, as in Data Assumption \ref{da:4b}, is exactly what is required for
the ideas of supervised learning from Chapter \ref{ch:SL}, where
we use it to learn map between $\R^d$ and $\R$, and this is readily
generalized to vector-valued output in $\R^q$ -- see Remark \ref{rem:NNG}.
Note that, to solve the inverse problem, we would like 
to learn the inverse map from the data  $y \in \R^\dy$ to
the parameter\index{parameter}  $u \in \R^\du.$ We could try and do this directly using supervised
learning. However, for inverse problems in which $\dy<\du,$
learning such a map may be difficult because of a lack of uniqueness; furthermore there is
noise present in the $y^{(n)}$ which needs to be carefully accounted for.
For these reasons it is arguably more informative to learn, from
the supervised data defined by Data Assumption \ref{da:4b}, how to regularize the inverse problem. 
This is the viewpoint we take here.

We recall the \index{posterior}posterior probability density function $\post^y(u)$ on $u|y$ from \index{Bayes Theorem}Theorem \ref{t:bayes}, but assume that prior $\pr$
depends on an unknown parameter $\theta \in \Theta \subseteq \R^p$ so that the posterior has form 
\begin{equation} \label{eq:param_posterior}
\post^y (\vct u;\theta) = \frac{1}{Z^y(\theta)} \noise \bigl(\vct y- \vct G(\vct u) \bigr) \pr (\vct u;\theta).
\end{equation}
We will determine $\theta$ by choosing it
so that MAP estimators for the inverse problem defined by data $y^{(n)}$ best match the paired data point $u^{(n)}$. 
To this end recall the  \index{loss}\emph{loss function} $\loss$
and the now $\theta$-dependent \emph{regularizer}\index{regularizer}
$\reg$ defined by
\begin{equation}
\label{eq:lossandreg}
\loss (\vct u;y) = - \log \noise \bigl(\vct y-\vct G(\vct u)\bigr)= - \log \like(y|u), \quad \reg(\vct u;\theta) = - \log \pr(\vct u;\theta),
\end{equation}
leading to
an \index{objective}\emph{objective function} of the form
\begin{equation}
\label{eq:MAPP}
\J(\vct u;y,\theta) = \loss(\vct u;y) + \reg(\vct u;\theta).
\end{equation}
Note that we have also emphasized the parametric dependence of the loss and the objective function on $y \in \Ry$, 
something we did not do in Section \ref{sec:112}.
The generalization of Data Assumption \ref{da:4b} that we now work with is

\begin{dataassumption}\index{Data Assumption}
\label{da:4c} 
We are able to evaluate the likelihood $\like(y|u)$ and prior $\pr(u;\theta)$ for any
pair $(y,u,\theta) \in \R^\dy \times \R^\du \times \Theta.$ 
We are given samples $\{(y^{(n)},u^{(n)})\}_{n=1}^N$, assumed to be drawn i.i.d. from
the joint probability measure $\joint$ on $(y,u) \in \Ry \times \Ru$ defined by equation \eqref{eq:jc0} and Assumption \ref{a:jc1}.
\end{dataassumption}

The \index{MAP estimator}MAP estimator can be viewed as a function, for each $\theta,$ 
$\umap(\cdot\, ; \theta): \Ry \to \Ru$:
\begin{align*}
\umap(y;\theta) & \in \argmin_{u \in \Ru} \J(u;y,\theta).
\end{align*} 
(For simplicity we assume that a unique point is determined by the minimization.)
We now introduce a loss function defined through a distance-like deterministic
scoring rule  $\dd: \Ru \times \Ru \to \R^+,$  as defined in Definition \ref{def:det_score}; the canonical example is the squared Euclidean norm.  We then define the choice of $\theta$ through the optimization problem
\begin{align*}
\thetas &\in \argmin_{\theta \in \Theta} \bbE^{(y,u) \sim \joint}\, \dd\bigl(u,\umap(y;\theta)\bigr).
\end{align*}

This is referred to as bilevel optimization\index{optimization!bilevel} because of the optimization to
find $\umap$, which is used within the optimization
to find $\thetas$. This procedure corresponds to hyperparameter\index{hyperparameter} tuning of $\theta$ on the validation set $\{(y^{(n)}, u^{(n)})\}_{n=1}^N$. In practice, the determination of $\thetas$ is implemented using $\joint^N$, the approximation of $\joint$ found from empirical\index{empirical} samples $\{(y^{(n)},u^{(n)})\}_{n=1}^N$ drawn i.i.d. from $\gamma$, as assumed in Data Assumption~\ref{da:4c}.

\subsection{Learning Priors for Posterior Approximation}
\label{ssec:carGB}

In this subsection we generalize the setting from the previous one to the more general problem of learning parameters in the prior to best approximate the posterior. We work under the following data assumption, which differs from that in the previous subsection:
we assume only marginal samples from $\kappa$ and $\pr$,
but this is compensated for by employing evaluation of the likelihood.
We return to work under the following modification of Data Assumption \ref{a:LBI}:

\begin{dataassumption}\index{Data Assumption}
\label{a:LBIMinus}
We are able to evaluate the likelihood $\like(y|u)$ and prior $\pr(u;\theta)$ for any
pair $(y,u,\theta) \in \R^\dy \times \R^\du \times \Theta.$ 
We are given data in the form of independent samples $\{y^{(n)}\}_{n=1}^N$ 
from the marginal distribution $\marg$ and independent samples $\{u^{(n)}\}_{n=1}^N$ from the prior distribution $\pr.$
\end{dataassumption}

Note the difference between Data Assumption \ref{a:LBI} and Data Assumption \ref{a:LBIMinus}:
the point of this subsection is to learn the prior density, and we extend the assumption to allow
for evaluation of a parametrically dependent family of priors.

\begin{remark}
    In  Data Assumption \ref{a:LBI} and in Data Assumption  \ref{a:LBIMinus} the set of independent samples $\{y^{(n)}\}_{n=1}^N$
    and the set of independent samples $\{u^{(n)}\}_{n=1}^N$ do not need to be independent of one another,
    and the same number of each is not required. In practice they may be found by marginalizing a data set of joint samples $\{(y^{(n)},u^{(n)})\}_{n=1}^N$  from $\joint$. It is then natural to have the same number of samples of each, and for each set of samples to be correlated. Thus, in this subsection, we may also work under Data Assumption \ref{da:4b} provided that, additionally, we can evaluate the likelihood $\like(y|u)$ and prior $\pr(u; \theta)$ for any pair $(y,u,\theta) \in \R^\dy \times \R^\du\times \Theta.$
\end{remark}

It is again helpful to recall the notation from Section \ref{sec:11} and Remark \ref{rem:minus1}. The goal is to choose parameters $\theta$ in the prior so that the resulting posterior $\post^y(u;\theta)$ in~\eqref{eq:param_posterior} is as close as possible to the true posterior $\post^y(u)$ corresponding to the joint distribution $\joint(y,u) = \like(y|u)\pr(u)$ underlying the data given in Data Assumption \ref{a:LBIMinus}. We identify the parameters by minimizing the posterior error with respect to the KL divergence\index{divergence!Kullback--Leibler}, 
in expectation over the marginal on $y$ which, we recall, has probability distribution $\marg$: 
\begin{subequations}
\begin{align}
    \F(\theta)&:=\mathbb{E}^{y \sim \kappa}\Bigl[ \dkl\bigl(\post^y\|\post^y(\cdot\,;\theta)\bigr)\Bigr], \label{eq:expected_KLdiv}\\
    \thetas & \in \argmin_{\theta \in \Theta} \F(\theta). \label{eq:optimal_parameters_paramprior}
\end{align}
\end{subequations}

The following result shows that the optimal parameters can be computed without needing to evaluate the true posterior density. 
\begin{theorem} \label{t:sss} The optimal parameter $\thetas$ in~\eqref{eq:optimal_parameters_paramprior} corresponds to the solution of the optimization problem
\begin{subequations}
\label{eq:addback2}
\begin{align}
\J(\theta) & := \biggl(\mathbb{E}^{u \sim \pr} \left[-\log\pr(u;\theta)\right] + \mathbb{E}^{y \sim \marg}\left[\log \int \like(y|u') \pr(u';\theta)\, du' \right]\biggr),\\
\thetas &\in \argmin_{\theta \in \Theta} \J(\theta).
\end{align}
\end{subequations}
\end{theorem}

\begin{proof} 
We first note that $\joint(y,u)=\post^y(u)\marg(y).$ Thus
\begin{subequations}
\label{eq:addback}
\begin{align}\mathbb{E}^{y \sim \kappa}\Bigl[\dkl\bigl(\post^y\|\post^y(\cdot\,;\theta)\bigr)\Bigr]&=\int_{\R^k} \left[\int_{\R^d}  \log \left(\frac{\post^y(u)}{\post^y(u;\theta)}\right) \post^y(u) du \right] \marg(y)\, dy\\
&=\int_{\R^k \times \R^d} \log \left(\frac{\post^y(u)}{\post^y(u;\theta)}\right) \gamma(y,u)\,dy\,du.
\end{align}
\end{subequations}

By \eqref{eq:addback}, the expected KL divergence in~\eqref{eq:expected_KLdiv} can be decomposed into two terms as 
\begin{align} \label{eq:KLdecomposition_approxprior}
    \mathbb{E}^{y}\Bigl[\dkl\bigl(\post^y \|\post^y(\cdot\, ;\theta)\bigr)\Bigr] &=  \int \joint(y,u)\left[ \log\left(\frac{\like(y|u)\pr(u)}{Z^y}\right) \nonumber - \log\left(\frac{\like(y|u)\pr(u;\theta)}{Z^y(\theta)}\right)\right] \, du\,dy \\ 
    &= \int \joint(y,u)\Bigl[\bigl(\log \pr(u) - \log Z^y\bigr) - \bigl(\log \pr(u;\theta) - \log Z^y(\theta)\bigr) \Bigr] \, du\,dy.
\end{align}
The first two terms are constant with respect to $\theta$ and are thus irrelevant for the minimization task that is the object of this theorem. Thus we need only minimize
\begin{equation}
\label{eq:abi}\thetas \in \argmin_{\theta \in \Theta} \int \joint(y,u) \left[-\log\pr(u;\theta) + \log Z^y(\theta) \right] \, du\, dy.
\end{equation}
Using the form of the normalizing constant $Z^y(\theta) = \int \like(y|u')\pr(u';\theta) \, du'$, we arrive at the objective $\J(\theta)$, after noticing that in the first term integration over $y$ is redundant, and in the
second integration over $u$ is redundant.
\end{proof}

Note that the objective function in Theorem~\ref{t:sss} can be approximated
empirically\index{empirical} using Data Assumption \ref{a:LBIMinus}. In particular, this requires
only marginal samples from $\gamma$, and does not require knowledge of the
posterior -- evaluation of the likelihood is sufficient.

\begin{remark}
In the preceding chapters we have differentiated between methods which require evaluation of
the normalization constant\index{normalization constant} and those which do not. This is because,
for problems for which the likelihood and prior do not have significant overlap, evaluation of the
normalization constant can be expensive. The method in this section does require evaluation of the normalization constant. This is a potentially severe limitation on the methodology in this subsection,
and typically limits it to problems where the parameter $u$ is of low dimension.
\end{remark}

\begin{remark} The expected KL divergence in~\eqref{eq:KLdecomposition_approxprior} can also be written as 
\begin{align*}
    \mathbb{E}^{y}\Bigl[ \dkl\bigl(\post^y \|\post^y(\cdot\, ;\theta)\bigr) \Bigr] &= \int \pr(u)\left[\log \pr(u) - \log \pr(u;\theta)\right] \, du + \int \marg(y) \log \Biggl(\frac{Z^y(\theta)}{Z^y} \Biggr) \, dy\\
    &= \dkl \bigl(\pr \|\pr(\cdot\, ;\theta)\bigr) - \dkl \bigl(\marg \|\marg(\cdot\, ;\theta)\bigr),
\end{align*}
where in the last line we recognize that the normalizing constants $Z^y, Z^y(\theta)$ are equivalent to the marginal distributions of the observations\index{observation} $\marg(y) = \int \like(y|u')\pr(u') \, du'$ and $\marg(y;\theta) = \int \like(y|u')\pr(u';\theta) \, du'$, respectively.
Thus, the objective in~\eqref{eq:expected_KLdiv} involves two competing terms, the first relating only to the prior, the marginal distribution on the states, and the second to the marginal distribution on the observations.
Note that when this objective is to be
approximated by means of empirical\index{empirical} samples from $\rho$ and $\kappa$, undefined terms arise in both divergences; these terms are $\theta$-independent and can be removed as we did to arrive at \eqref{eq:abi}. See Remark \ref{rem:discuss0} for discussion of this point in the canonical context of maximum likelihood density estimation.
\end{remark}

\section{Bibliography}
\label{sec:lfbib}

Amortized variational inference\index{variational inference!amortized} was introduced in a number of specific contexts within the field
of variational inference. Variational autoencoders\index{variational!autoencoder|see{autoencoder, variational}}\index{autoencoder!variational}, the subject of Section \ref{sec:vauto} and which were proposed concurrently in  
\cite{kingma2013auto} and \cite{rezende2014stochastic}, implicitly introduced
the concept of amortized inference. In particular the encoder may be thought of
as defining a family of posterior distributions on the latent variable, conditioned
on the different samples in the original space; see Remark \ref{rem:319}. This
family may be approximated, for example, as a parameterized family of Gaussians,
as in Example \ref{ex:AVIG}. Amortized inference\index{inference!amortized} is overviewed in \cite{margossian2023amortized,zhang2018advances} and is now 
widely used in the context of latent variable models~\cite{dayan1995helmholtz, blei2017variational}.
The idea of iterating amortized inference was introduced in
\cite{marino2018iterative}; the iterative approach is designed to close
the amortization gap\index{amortization!gap} caused by failing to reach
optimality when training.  Amortized variational inference is
applied to filtering problems in \cite{marino_general_2018}.
The methodology has also been used to avoid expensive per-instance computation across a range of tasks in meta-learning~\cite{finn2017model} and to learn parameters for optimization algorithms~\cite{andrychowicz2016learning}. Amortized MAP estimation\index{MAP estimation!amortized} of the type discussed in Subsection~\ref{ssec:amortized_map_obs} is used in \cite{sonderby_amortised_2017} for image super-resolution.

The concept of amortization to perform likelihood-free or simulation-based inference has origins in statistics with high-dimensional latent variables. The latent variable $z$ over which we marginalize is often referred to as a \emph{nuisance} 
or auxiliary random variable; this is often used in the setting where $z$ is not the 
primary parameter of interest in the inverse problem. An alternative approach to those
described here seeks the joint posterior of $(u,z)$, and then marginalizes after the 
computation. While the corresponding likelihood for the joint posterior doesn't involve a marginal or integrated likelihood, it can lead to a challenging problem for high or even infinite-dimensional latent variables. We refer to~\cite{basu1977nuissance} for more details on integrated likelihoods. In MCMC algorithms\index{algorithm!MCMC}, these integrated likelihoods are commonly addressed using pseudo-marginal methods that work with unbiased estimators of the likelihood~\cite{andrieu2009pseudo}.
Amortized posterior estimation using the energy score is explored in \cite{kaveh_amortized_2026}.

Approximate Bayesian computation (ABC)\index{approximate Bayesian computation} is a classic inference method for performing likelihood-free inference with latent variables or other intractable likelihoods; see~\cite{sisson2018handbook} for a comprehensive overview on ABC. These approaches define a distance function that compares simulated observations\index{observation} to the true observation and reject parameter samples that are not consistent with the true observation based on a small tolerance. ABC methods can be shown to be consistent in the limit of the tolerance approaching zero, but typically require large sample sizes with high-dimensional observations\index{observation}.

Learning approaches have appeared as alternatives to ABC; machine learning-based approaches for simulation-based inference\index{simulation-based inference} are outlined in~\cite{cranmer2020frontier}. Methodologies for both posterior and  likelihood approximation in this setting are implemented using various unsupervised learning architectures (see Chapter~\ref{ch:UL})  including: conditional normalizing flows~\cite{winkler2019learning, papamakarios2019sequential}, conditional generative adversarial networks~\cite{ray2023solution}, conditional diffusion models~\cite{batzolis2021conditional}. 
Triangular transport maps are a core element of these architectures for solving inverse problems. They are related to the Knothe--Rosenblatt\index{transport!Knothe--Rosenblatt} transport, which has been used for conditional density estimation in~\cite{marzouk2016sampling, baptista2023representation}. The design and analysis of block-triangular transport maps to solve likelihood-free inference problems on function space is investigated in~\cite{baptista2020conditional, hosseini2023conditional}.


\part{Data Assimilation}

\chapter{\Large{\sffamily{Filtering and Smoothing Problems}}}\label{lecture7}

This chapter provides an introduction to \index{data assimilation}\emph{data assimilation}. 
We study both the \index{filtering}\emph{filtering} and the \index{smoothing}\emph{smoothing} problems. Consider the \index{stochastic dynamics model}{\em stochastic dynamics model} 
given by 
\begin{subequations}
\label{eq:sdm}
\begin{align}
\vd_{j+1} &= \Psi(\vd_j) + \xid_j, \quad j \in \Z^+,\\
\vd_0 &\sim \Nc(m_0, C_0), \quad \xid_j \sim \Nc(0, \Sigma) \: \,\, \index{i.i.d.}\text{i.i.d.}\,, 
\end{align}
\end{subequations}
combined with the \index{data model}{\em data model} given by
\begin{subequations}
\label{eq:dm}
\begin{align}
\yd_{j+1} &= h(\vd_{j+1}) + \etad_{j+1} , \quad j \in \Z^+,\\
\etad_{j+1} &\sim \Nc(0, \Gamma) \: \,\,\index{i.i.d.}\text{i.i.d.}
\end{align}
\end{subequations}
Broadly speaking, \index{data assimilation} data assimilation seeks to
find the \emph{state}\index{state}, or \emph{signal}\index{signal}, $\{\vd_j\}$ over index set $j \in \{0,1,\dots,J\}$
based on realized \emph{data}\index{data}, or \emph{observations}\index{observations} $\{\yd_j\}$ 
over index set $j \in \{1,\dots,J\}.$
Various different inverse problems\index{inverse problem}, with
this aim, can be defined; we will introduce them in this chapter. 
Indeed, the goal of this chapter is to define the key problems arising in the field of data assimilation,
and to outline the primary methodologies used, prior to the current influx of new ideas from machine learning. These new ideas are then described in the four subsequent chapters.

Unless otherwise noted, we will work throughout this chapter, and the four subsequent chapters 
devoted to data assimilation, under the following two assumptions on the stochastic dynamics and data models. 
Recall, from the preface, that $\ind$ denotes independence between random variables. The first assumption 
concerns the independence structure of
the random variables entering the signal--observation\index{signal}\index{observations} model:
\begin{assumption} \label{a:noise}
$\{ \xid_j \}_{j \in \bbZ^+} \ind \{\etad_j\}_{j \in \bbN} \ind \vd_0.$
\end{assumption}
\noindent The second assumption concerns the matrices and maps defining the signal--observation model:
\begin{assumption}
\label{a:fas} 
The matrices $C_0,$ $\Sigma$ and $\Gamma$ are \index{positive definite}positive definite.
Furthermore, the nonlinear maps $\Psi$ and $h$ are continuously differentiable: $\Psi \in C^1(\Ru, \Ru)$ and $h \in C^1(\Ru, \Ry)$. 
\end{assumption}

Section \ref{sec:int} contains introductory remarks that will orient the reader in the remainder
of the chapter.
In Section \ref{sec:DA} we formulate the filtering 
and smoothing 
problems in detail. In Sections \ref{sec:filtering} and \ref{sec:smoothing}, respectively, we describe algorithms for the filtering and smoothing approaches to data assimilation. In Section \ref{sec:meda} we discuss model error in the context of filtering and smoothing problems. Section \ref{sec:ml_approx_da} is
devoted to the study of surrogate\index{surrogate} modeling, for both filtering and
smoothing; this serves as an introduction to the potential for the use of ideas from
machine learning within data assimilation. In Section~\ref{sec:gen} we describe various generalizations of the setting
in this chapter and, in particular, relaxations of the Assumptions \ref{a:noise} and \ref{a:fas}. Section \ref{sec:dabib} contains bibliographic notes.

\section{Introduction} \label{sec:int}

Subsection~\ref{ssec:note} is devoted to notation and, in particular, explains
the use of the $\dagger$ on the signal--observation models \eqref{eq:sdm}, \eqref{eq:dm}; Subsection~\ref{ssec:fvs} contrasts the filtering and smoothing approaches to data assimilation; and Subsection~\ref{ssec:pvse} contrasts the probabilistic and state estimation approaches to data assimilation.

\subsection{Notation} \label{ssec:note}
To state the inverse problems of interest precisely, we make the following
definitions: for a given and fixed integer $J,$ and for $1\le j \le J,$
\begin{subequations}
\label{eq:VdYd}
\begin{align}
\Vd :=& \{\vd_0, \ldots, \vd_J\}, \: \, \Vd_j := \{\vd_0, \ldots, \vd_j\},\\
\: \,\Yd :=& \{\yd_1, \ldots, \yd_J\}, 
\: \, \Yd_j := \{\yd_1, \ldots, \yd_j\}. 
\end{align}
\end{subequations}
The sequence $\Vd$ is often termed the \index{signal}{\em signal} or {\em state}
and the sequence $\Yd$ the {\em data} or {\em observation}. It is sometimes helpful 
to define $\Yd_0$ as the empty set $\varnothing.$ 
We will occasionally consider the infinite time horizon limit where $J = \infty.$

We use the  notation $\dagger$ throughout Part II of the notes, devoted to data assimilation,
to denote all random variables
associated with the underlying model for signal and data. We adopt this 
convention because many of the algorithms that we develop for data assimilation will use the
signal and/or data models \eqref{eq:sdm}, \eqref{eq:dm} as part of their backbone. We will
use symbols $v_j$, or for ensemble algorithms $v_j^{(n)}$ with $n$ indexing particles in the ensemble, for the
outputs of algorithms to find the state 
$\{\vd_j\}$; 
these algorithms will operate over some set of time indices $j \in \{0,1,\dots,J\}$,
based on realized observations $\Yd$. We will also use the
notation $(\widehat{v}_{j+1},\widehat{y}_{j+1})$, or for ensemble algorithms $(\widehat{v}_{j+1}^{(n)},\widehat{y}_{j+1}^{(n)}),$ for 
predictions made as part of algorithms; in particular for predictions of the state $\vd_{j+1}$, and  predictions of the observation $\yd_{j+1}$, at time $j+1$ using only 
observed data $\Yd_j$ up to time $j.$
The state predictions $\widehat{v}_{j+1}$ or $\widehat{v}_{j+1}^{(n)}$,
possibly together with $\widehat{y}_{j+1}$ or $\widehat{y}_{j+1}^{(n)}$,
are then corrected using the observed data $\yd_{j+1}$ to yield the output
$v_{j+1}$ or $v_{j+1}^{(n)}$ of algorithms at time $j+1.$

By adopting this notational convention we distinguish
between the output of the underlying signal and/or data models \eqref{eq:sdm}, \eqref{eq:dm}
($v$ and $y$ symbols with $\dagger$) and algorithms to estimate the signal 
($v$ and $y$ symbols without $\dagger$); this serves to avoid
confusion between the source of data, and the algorithms for data assimilation, when the
latter also use the underlying signal and/or data models as part of their definition. 
We do not use the $\dagger$ convention in Part I of the notes, devoted to inverse problems, because the algorithms we study do not have the same potential for confusion with output of
the underlying inverse problem model \eqref{eq:jc0}.

\subsection{Filtering Versus Smoothing} \label{ssec:fvs}

There are two core problems in data assimilation,
one known as \index{filtering}\emph{filtering}, 
and the other as \index{smoothing}\emph{smoothing}. Both use: 

\begin{dataassumption}\index{Data Assumption}
\label{da:8}
Data $\Yd \in \R^{\dy J}$ is given and is assumed to have come from
the signal-observation\index{signal}\index{observation} model defined by
\eqref{eq:sdm} and \eqref{eq:dm}.
\end{dataassumption}

The \index{filtering}\emph{filtering distribution} at time $j$ is $\mathbb{P}(\vd_j | \Yd_j)$: the distribution of the state $\vd_j \in \R^{\du} $ at time $j$, conditioned on data $\Yd_j \in \R^{\dy j}$. This distribution at time $j$ is defined without knowledge of future observations at times $j'>j.$ We say that this distribution is \emph{time-causal.}\index{time-causal} The objective of filtering is to find $\mathbb{P}(\vd_j | \Yd_j)$ for each $j \in \{1, \ldots, J\},$ sequentially.

The \index{smoothing}\emph{smoothing distribution} is  $\mathbb{P}(\Vd | \Yd)$, a probability density function on an entire sequence of states $\Vd$ given the entire dataset $\Yd$; note that $\Vd \in \R^{\du(J+1)}$ and $\Yd \in \R^{\dy J}$ represent sequences over a window of length $J+1$ and $J$, respectively. For  $0 \le j \le J-1,$ the marginal distribution of the smoothing distribution $\mathbb{P}(\Vd | \Yd)$, corresponding to time $j,$ is $\mathbb{P}(\vd_j|\Yd);$ this depends
on observations at times $j'>j$, highlighting an essential difference between \index{filtering}filtering and \index{smoothing}smoothing.

The smoothing problem is an example of a single Bayesian inverse problem
of the type defined in Chapter \ref{ch1}. The filtering problem comprises $J$ Bayesian inverse problems, interleaved with predictions of the model.
We now describe the Bayesian perspective on filtering and on smoothing.
In smoothing the prior on $\Vd$ is a probability density function defined by
\eqref{eq:sdm}; the posterior is found by conditioning this prior on data $\Yd$ defined by \eqref{eq:dm}. In filtering, at time $j+1$, the prior is given by $\mathbb{P}(\vd_{j+1} | \Yd_{j})$ which itself is defined by combining $\mathbb{P}(\vd_j | \Yd_j)$ with the dynamics model \eqref{eq:sdm};  thus we see that the prior for the filtering distribution
at time $j+1$ is defined by combining the filtering distribution at time $j$
with the dynamics model \eqref{eq:sdm}. The posterior  $\mathbb{P}(\vd_{j+1} | \Yd_{j+1})$ is then found by conditioning the prior $\mathbb{P}(\vd_{j+1} | \Yd_{j})$ on observation $\yd_{j+1}$ using the data model \eqref{eq:dm}.

\begin{figure}
    \centering
    \includegraphics[width=0.7\linewidth]{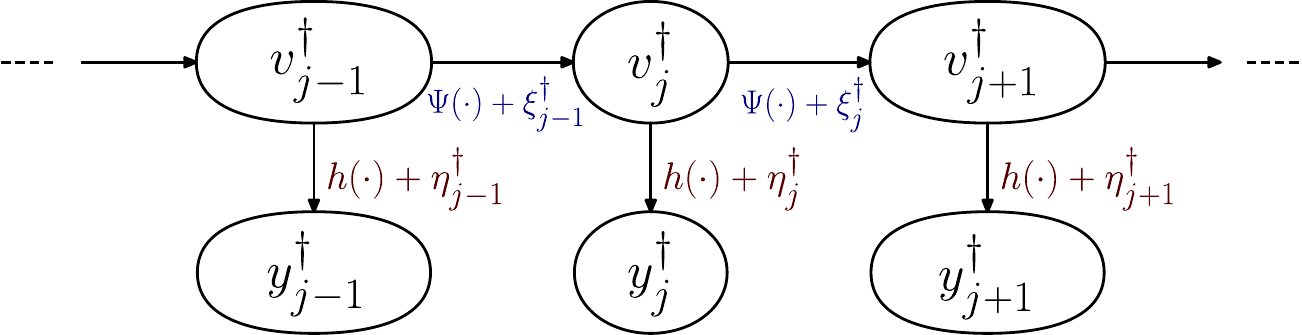}
    \caption{Dynamics and observation models underlying data assimilation problems.}
    \label{fig:hmm}
\end{figure}

\subsection{Probabilistic Versus State Estimation} \label{ssec:pvse}  
When our goal is to approximate or characterize the filtering and smoothing distributions arising in these Bayesian formulations of the filtering and smoothing problems,
we will refer to \emph{probabilistic estimation.}\index{probabilistic estimation} 
The underlying inverse problems are solved in a Bayesian fashion.\index{inverse problem!Bayesian} 
On the other hand many optimization-based filtering and smoothing algorithms\index{algorithm!filtering}\index{algorithm!smoothing} 
aim only to provide a point estimate for the state,  given the data. 
We refer to this as \emph{state estimation.}\index{state estimation} State estimation
may be thought of as being analogous to MAP estimation, from Subsection~\ref{ssec:MAP},
in the general Bayesian inverse problem\index{inverse problem!Bayesian} setting.
Indeed, for the smoothing problem, the MAP estimator associated with the posterior
distribution on $\Vd$ is frequently employed as a point estimator of the state. The situation is
more complicated for filtering; this is because the sequential nature of the problem means that we only have access to approximate priors, often in the form of point masses, or a Gaussian, for example; for this reason iterative computation of the MAP estimator is not always natural. Nonetheless, useful algorithms for state estimation do exist in filtering and they will play an important role in what follows. In many cases they are closely related to classical algorithms from control theory.

\section{Formulation of Data Assimilation\index{data assimilation}}
\label{sec:DA}

In Subsection~\ref{ssec:filtering} we formulate the filtering problem
and in Subsection~\ref{ssec:smoothing} the
smoothing problem.

\subsection{Formulation of the \index{filtering}Filtering Problem}
\label{ssec:filtering}

\begin{definition}
\label{def:FP}
The \index{filtering}{\em filtering problem} is to find, and update sequentially in $j$, the probability densities 
$\post_j(\vd_j):=\Prob(\vd_j|\Yd_j)$ on $\Ru$ for $j=1, \dots, J.$
We refer to $\post_j$ as the {\em \index{filtering!distribution}filtering distribution at time $j.$}
\end{definition}

Filtering may be understood as the sequential interleaving of prediction, using the
\index{stochastic dynamics model} stochastic dynamics model \eqref{eq:sdm}, with inversion,
using the \index{data model}data model \eqref{eq:dm}.
To explain this perspective it is helpful to introduce\footnote{Notice that $\hat{\post}_1 := \Prob(\vd_{1})$ with the convention that $\Yd_0 = \varnothing.$} 
$\hat{\post}_{j+1}=\Prob(\vd_{j+1}|\Yd_j)$.
We also define the Gaussian likelihood $\like(y|u) = \Nc(y; h(u),\Gamma)$, 
dictated by the data model \eqref{eq:dm}. 
Combining $\hat{\post}_{j+1}$ as prior with this likelihood, evaluated at time $j+1$, Bayes theorem delivers the formula 
$$\pi_{j+1}(\vd_{j+1}) \propto \like(\yd_{j+1}|\vd_{j+1})\widehat\pi_{j+1}(\vd_{j+1}).$$

\begin{figure}[h]
  \begin{center}
    \includegraphics[width=0.8\textwidth]{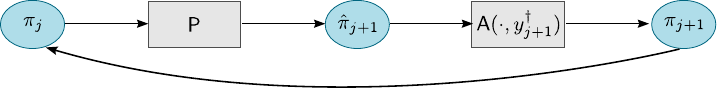}
  \end{center}
  \caption{Prediction and analysis steps combined.} \label{fig:interaction}
\end{figure}

It is useful to decompose the sequential updating 
$\post_j \mapsto \post_{j+1}$ into the following two steps, expressed abstractly using maps on
probability measures:
\begin{align}\label{eq:pna}
\begin{split}
&{ \text{ \bf  Prediction Step:}} ~~~~\;  \hat{\post}_{j+1} = \Pred \post_j.  \index{prediction} \\
& {\text{  \bf Analysis Step:}} ~~~~~\,\,\,\,\,  \post_{j+1} = \An_j (\hat{\post}_{j+1}):=\An(\hat{\post}_{j+1};\yd_{j+1}). \index{analysis} 
\end{split}
\end{align}
The recursion starts at $\pi_0$, assumed to be the Gaussian $\Nc(m_0, C_0)$ from (\ref{eq:sdm}b). 
The combination of the \index{prediction}prediction and \index{analysis}analysis steps is shown schematically in Figure \ref{fig:interaction} and leads to the update 
\begin{equation}
\label{eq:An}
\post_{j+1} = \An_j(\Pred\post_j). 
\end{equation}
Here $\Pred$ is a linear map,\footnote{Strictly speaking $\Pred$ is linear when viewed as acting on the vector space $L^1(\R^{\du};\R)$; the space of probability density functions is a subset of this vector space, but is not a linear subspace.} defining the Markov process underlying the 
\index{stochastic dynamics model} stochastic dynamics model. Map $\An_j=\An(\cdot\,;\yd_{j+1})$ is a 
nonlinear map defined by application of \index{Bayes Theorem}Bayes Theorem \ref{t:bayes} to solve the inverse problem
specified by the likelihood $\like(\yd_{j+1}|\vd_{j+1})$, with prior $\hat{\post}_{j+1}$ -- see Remark \ref{rem:eaudc}.

The linear prediction operator $\Pred$ is defined as follows:
\begin{subequations}
\label{eq:pred_operator}
\begin{align}
    \Pred \pi(u) & = \int_{\R^d} \Prob(u|v) \pi(v)\, dv \\
    & = \frac{1}{\sqrt{(2\pi)^d \det \Sigma}}\int_{\R^d} \exp \left( - \frac{1}{2} \bigl|u - \Psi(v)\bigr|_{\Sigma}^2 \right) \pi(v) \, dv.
\end{align}
\end{subequations}
The expression for the Gaussian conditional probability density function $\Prob(u|v) = \Nc(u; \Psi(v),\Sigma) $ is dictated by the given stochastic dynamics model  \eqref{eq:sdm}.   
If the state is distributed according to $\pi$ at time $j,$ then $\Pred \pi$ represents the distribution of the state at time $j+1$ when the state evolution is defined by the Markovian dynamics \eqref{eq:sdm}.

To define the analysis operator, notice that any probability density function $\pi$ in $\cP(\R^d)$ can be extended to a joint probability density $\gamma_\pi$ in $\cP(\R^{d} \times \R^k)$ as follows:
\begin{subequations}
\label{eq:n4ukf}
\begin{align}
\gamma_\pi(u, y) & = \like(y|u) \pi(u)\\
& = \frac{1}{\sqrt{(2\pi)^k \det \Gamma}}
\exp \left( - \frac{1}{2} \bigl\lvert y - h(u) \bigr\rvert_{\Gamma}^2 \right) \pi(u).
\end{align}
\end{subequations}
Here, recall, $\like(y|u)$ is dictated by the data model \eqref{eq:dm}. Map $\pi \to \gamma_\pi$
is linear. However, using it we may define the nonlinear operator $\An(\cdot;y)$ by
\begin{equation}
  \An(\pi;y)(u) = \frac{\gamma_\pi(u,y)}{\int_{\R^d} \gamma_\pi(u,y) \, d u}.\label{eq:filter_bayes_inference}
\end{equation}
Given a prior $\pi$ on the state at time $j+1,$ $\An_j(\pi) = \An(\pi;\yd_{j+1})$ represents the posterior distribution of the state at time $j+1$ determined by data model  \eqref{eq:dm} and observed data $\yd_{j+1}.$

\subsection{Formulation of the \index{smoothing}Smoothing Problem}
\label{ssec:smoothing}

Recall the definitions of $\Vd$ and $\Yd$ from \eqref{eq:VdYd}.
\begin{definition}
The \index{smoothing}{\em smoothing problem} is to find the probability density 
$\Pi^{\Yd}(\Vd):=\Prob(\Vd|\Yd)$ on
$\R^{\du(J+1)}$ 
for some fixed integer $J.$ We refer to $\Pi^{\Yd}$ as the \index{smoothing!distribution}{\em  smoothing distribution}.
\end{definition}

Using Bayes Theorem \ref{t:bayes}, the density for the smoothing distribution $\mathbb{P}(\Vd | \Yd)$ is given by
\begin{equation}
\label{eq:spost2}
\Pi^{\Yd}\!(\Vd) \propto \like(\Yd|\Vd)\rho(\Vd).
\end{equation}
The signal--observation model defined by \eqref{eq:sdm} and \eqref{eq:dm}, together with the standing Assumptions \ref{a:noise} and \ref{a:fas}, determine the likelihood\index{likelihood} $\like(\Yd|\Vd)$ and the prior\index{prior}
$\rho(\Vd)$ entering this expression. We obtain
\begin{subequations}
\label{eq:sdef}
\begin{align}
    \like(\Yd|\Vd) &= \prod_{j=1}^{J}\like_j(\vd_j),\quad \rho(\Vd) = \Bigl(\prod_{j=1}^J \mathsf{t}(\vd_j,\vd_{j-1})\Bigr)\rho_0(\vd_0),\\
    \like_j(\vd) &= Z_\Gamma^{-1}\exp\Bigl(-\frac{1}{2} |\yd_{j} - h(\vd)|^2_\Gamma\Bigr),\quad Z_\Gamma = (2\pi)^{k/2} \det(\Gamma)^{1/2},\\
    \mathsf{t}(\vd,\dw) &= 
    Z_\Sigma^{-1}\exp\Bigl(-\frac{1}{2}|\vd - \Psi(\dw)|^2_\Sigma\Bigr), \quad Z_\Sigma  = (2\pi)^{d/2} \det(\Sigma)^{1/2},\\
    \rho_0(\vd) &=Z_{C_0}^{-1}\exp\Bigl(-\frac{1}{2}|\vd - m_0|_{C_0}^2\Bigr), \quad Z_{C_0}  = (2\pi)^{d/2} \det(C_0)^{1/2}.
\end{align}
\end{subequations}

Note that the prior on $\Vd$ is given by the dynamics model \eqref{eq:sdm} and is not Gaussian (unless $\Psi(\cdot)$ is linear). The likelihood, defined by \eqref{eq:dm}, is Gaussian.
For both the likelihood and the prior we have used the independence assumptions encapsulated in
Assumption \ref{a:noise}, and the Markovian structure arising from the i.i.d. nature of the sequences
$\{ \xid_j \}_{j \in \bbZ^+}$ and $\{\etad_j\}_{j \in \bbN}$,
to derive the formulae for likelihood and prior as given.
It is also insightful to write the solution of this Bayesian inverse problem in a notation that
removes the indices, to clarify the high-level structure. 
To this end we define $\mathsf{\Psi}: \R^{\du J} \times \R^{\du} \to \R^{\du(J+1)}$ by 
\begin{align*}
\mathsf{\Psi}(\Vd_{J-1};m_0):=\bigl\{m_0, \Psi(\vd_0), \ldots, \Psi(\vd_{J-1}) \bigr\},
\end{align*}
and $\Sigmas$ to be a block diagonal matrix with $C_0$ in the first diagonal block and with $\Sigma$ in 
all remaining diagonal blocks. From \eqref{eq:sdm} we have
\begin{equation}
\label{eq:sprior}
\rho(\Vd) \propto \exp\Bigl(-\frac12|\Vd-\mathsf{\Psi}\bigl(\Vd_{J-1};m_0\bigr)|_{\Sigmas}^2\Bigr).
\end{equation}
We define $\etad := \{\etad_1, \ldots, \etad_J\} \in \R^{\dy J}$
and then $\etad \sim \Nc(0, \Gammas)$, where $\Gammas$ is block diagonal with $\Gamma$ in 
each diagonal block. If we then define $\hs:\R^{\du(J+1)} \to \R^{\dy J}$ by
$$\hs(\Vd):=\bigl\{h(\vd_1), \ldots, h(\vd_J) \bigr\},$$
then the \index{data model}data model \eqref{eq:dm} may be written as
$$\Yd=\hs(\Vd)+\etad.$$
The likelihood\index{likelihood} thus has the form
\begin{equation}
    \label{eq:likes}
    \like(\Yd|\Vd) \propto \exp\Bigl(-\frac12|\Yd-\hs(\Vd)|_{\Gammas}^2\Bigr).
\end{equation}

We are interested in finding $\Prob(\Vd|\Yd)=\Pi^{\Yd}\!(\Vd).$ We may apply Bayes Theorem \ref{t:bayes}
with prior $\rho$ and likelihood $\like$, noting that, under Assumption \ref{a:noise}, $\Vd \ind \etad,$ the standard setting for Bayesian inversion from Chapter~\ref{ch1}. We may then write \eqref{eq:spost2} in the form 
\begin{subequations}
\label{eq:spost}
\begin{align}
\Pi^{\Yd}\!(\Vd) &\propto \exp\Bigl(-\frac12|\Yd-\hs(\Vd)|_{\Gammas}^2\Bigr)\rho(\Vd)\\
&\propto \exp\Bigl(-\frac12|\Yd-\hs(\Vd)|_{\Gammas}^2-\frac12|\Vd-\mathsf{\Psi}\bigl(\Vd_{J-1};m_0\bigr)|_{\Sigmas}^2\Bigr).
\end{align}
\end{subequations}

\begin{remark}\label{rem:fixed_interval_smoothers}
Because filtering is defined by conditioning on data arriving sequentially,
such algorithms may be used \emph{online}\index{algorithm!online}, the probability distribution over states being updated every time
a new data point $\yd_j$ arrives. Smoothing gives rise to  methodologies that are most 
naturally used, in their most basic form, in an \emph{offline}\index{algorithm!offline} fashion. 
However, smoothing algorithms may also be used sequentially, in block form with respect to discrete time index $j$, as we
now explain.

In the above we assumed data $\yd_j$ given on the
index set $j \in \{1,\ldots, J\}$ and solved the smoothing problem on
$j \in \{0,\ldots, J\}.$ However this can be shifted to solve the inverse problem 
on any interval, for example containing $K+1$ points for some integer $K$, using data
on an interval excluding the first point, but provided with an initial Gaussian distribution
at the first point. Let $K \ll J$ be such an integer. We first solve the
smoothing problem defined on the index set
$j \in \{0,\ldots, K\}.$
Taking the solution at time $K$ as starting point we
may then solve a smoothing problem 
on index set $j \in \{K,\ldots, 2K\}.$
This idea may be iterated, working on index set $j \in \{\ell K,\ldots, (\ell+1)K\}$, then on  index set $j \in \{(\ell+1)K,\ldots, (\ell+2)K\}$ and so on. 
If $\ell=0,\dots, L-1$ then this will deliver a solution to the data assimilation problem
for $\vd_j$ defined on $j \in \{0,\ldots, J\}$ where $J=LK.$
Such methods are known as \emph{fixed-interval smoothers} and may be thought of as blending aspects
of smoothing and filtering, when iterated over $\ell.$ At overlap points, which are a multiple
of $K$, some form of Gaussian projection\index{Gaussian!projection} (see Remark \ref{rem:bishop}) will be needed to restart the process on the next time interval as the assumption made above is that the initial condition is Gaussian; however, this assumption on the initial distribution can be relaxed.
The case $K=1$ delivers the filtering distribution if the Gaussian projection is omitted.
\end{remark}

\section{Filtering Algorithms}\index{filtering}\index{algorithm!filtering}
\label{sec:filtering}

In this section we formulate various filtering algorithms.
Subsection~\ref{ssec:kalman} recalls the
Kalman filter\index{Kalman filter}, applicable in the linear and Gaussian setting, and in Subsection~\ref{ssec:sskalman} we study the Kalman filter when the covariance is in steady state.
Subsections~\ref{ssec:3dvar}, \ref{ssec:exkf}, \ref{ssec:ukf55} and \ref{ssec:enkf} recall 3DVar\index{3DVar}, the extended Kalman filter\index{Kalman filter!extended} (ExKF), the unscented Kalman filter (UKF) and the ensemble Kalman filter\index{Kalman filter!ensemble} (EnKF) respectively. Subsections~\ref{ssec:pf} and \ref{ssec:opf} concern the bootstrap particle
filter\index{particle filter!bootstrap} and the optimal particle
filter\index{particle filter!optimal} approaches to filtering. We discuss
the issue of weight collapse, which can degrade performance of the
bootstrap and particle filters in high dimensions $d$, in Subsection~\ref{ssec:collapse};
this degradation underpins the widespread adoption of the EnKF in
the high-dimensional setting as it is a particle filter
with equal weights.\index{particle filter!equal-weights}
We conclude the section, in Subsection~\ref{ssec:eval},
with discussion of a commonly used heuristic methodology for evaluating filtering algorithms.

The filtering distribution is, in general, an infinite dimensional object. In order
to approximate it (probabilistic estimation)\index{probabilistic estimation} some 
form of finite-dimensionalization is needed. Most
approximation methods use either empirical\index{empirical} particle or 
Gaussian\index{Gaussian} approximations, or a combination of both ideas.
The algorithms we study comprise nonautonomous dynamical systems\index{dynamical system!nonautonomous},
defined on state spaces of varying dimension. 
They are nonautonomous because they are driven by the observation
sequence $\{\yd_j\}_{j \in \bbN}$; the dimension varies depending on whether a
state estimate is evolved (3DVar\index{3DVar}), mean and covariance
are evolved (Kalman filter\index{Kalman filter} or the extended Kalman 
filter\index{Kalman filter!extended}), or an ensemble is evolved
(ensemble Kalman filter\index{Kalman filter!ensemble}, bootstrap
particle filter\index{particle filter!bootstrap}, optimal particle filter\index{particle filter!optimal}).

\subsection{Kalman Filter}\index{Kalman filter}
\label{ssec:kalman}

Here we consider the dynamics and observation models \eqref{eq:sdm} and \eqref{eq:dm} in which the dynamics and observation operators are linear; furthermore we generalize to make them non-autonomous:
\begin{equation}
\label{eq:rep}
\Psi(\cdot) \mapsto A_j \cdot\,, \quad h(\cdot) \mapsto H_j \cdot\,.
\end{equation}
Then the solution to the filtering problem is Gaussian and $\pi_j=\cN(v_j, C_j)$,
$\hat{\post}_{j+1}=\cN(\hat{v}_{j+1},\hat{C}_{j+1}).$ The update rules for the 
mean\footnote{We use symbol $v_j$ for the mean here, and not $m_j$, because of the links with
the 3DVar\index{3DVar} algorithm, which uses $v_j$ as state estimator; indeed, in the linear setting
of the Kalman filter considered in this subsection, it is natural to view the mean $v_j$ as a state estimator. This is related to the discussion of posterior mean estimators\index{posterior! mean estimator}
for general Bayesian inversion;\index{Bayesian!inversion} see Example \ref{ex:pme}.}
$v_j$ and covariance $C_j$ are given by the \emph{Kalman filter}\index{Kalman filter}: the mean is initialized at $v_0 = m_0$ and then updated, for $j \in \bbZ^+$, according to
\begin{subequations}\label{eq:kalman_mean}
\begin{align}
\hat{v}_{j+1} &= A_j v_j,\\ 
v_{j+1} &= \hat{v}_{j+1} + K_{j+1} \bigl(\yd_{j+1}-H_{j+1}\hat{v}_{j+1}\bigr).
\end{align}
\end{subequations}
Here the \emph{Kalman gain}\index{gain!Kalman} $K_{j+1}$ is determined as part 
of the following update rule $C_j \mapsto \hat{C}_{j+1} \mapsto C_{j+1}$ for the covariances
\begin{subequations} \label{eq:Kalman_recursions}
\begin{align}
\label{eq:udc}
\hat{C}_{j+1} &= A_j C_{j} A_j^\top + \Sigma,\\
S_{j+1} &= H_{j+1} \hat{C}_{j+1} H_{j+1}^\top + \Gamma,\\
K_{j+1} &= \hat{C}_{j+1} H_{j+1}^\top S_{j+1}^{-1}, \label{eq:udc_kalmangain}\\
C_{j+1} &= (I - K_{j+1}H_{j+1})\hat{C}_{j+1}.
\end{align}
\end{subequations}
The gain is thus computed using the predictive covariance which, when suitably
combined with noise covariance and the observation operator, facilitates the transfer 
of information from observation\index{observations} space to state\index{state} space.
Indeed
$$K_{j+1}=\hat{C}_{j+1}^{vy}\bigl(\hat{C}_{j+1}^{yy} \bigr)^{-1},$$
where $\hat{C}_{j+1}^{vy}:=\hat{C}_{j+1} H_{j+1}^\top$ 
is the predictive covariance between state and observation,
and $\hat{C}_{j+1}^{yy}:=S_{j+1}$ is the predictive covariance of the observation.

Together (\ref{eq:kalman_mean}a) and (\ref{eq:Kalman_recursions}a) define
a mapping from mean and covariance, at time $j$, to predictive mean and covariance, 
at time $j+1$, representing the outcome of
the prediction step $\Pred$
in \eqref{eq:pna}:
\begin{equation}\label{eq:Gaussianpred0}
\Prob (\vd_{j+1}| \Yd_j) = \Nc \bigl(\vd_{j+1};\widehat{v}_{j+1},\widehat{C}_{j+1}\bigr).
\end{equation}
The remaining equations (\ref{eq:kalman_mean}b) and (\ref{eq:Kalman_recursions}b)--(\ref{eq:Kalman_recursions}d) then define the analysis step $\An_j$ in \eqref{eq:pna}:
\begin{equation}\label{eq:Gaussianan0}
\Prob (\vd_{j+1}| \Yd_{j+1}) = \Nc \bigl(\vd_{j+1}; v_{j+1},C_{j+1}\bigr)
\end{equation}
mapping the predictive mean and covariance at time $j+1$ to the mean and covariance
of the filter at time $j+1.$ Both $\Pred$ and $\An_j$ map  Gaussians to
Gaussians in this setting.

The update for the covariance evolves independently of the update for the mean and is independent of the data; in contrast, the update for the mean depends on the update for the covariance 
and on the data.
Once the covariance is known, $v_j$ provides a state estimator\index{state estimation} for
$\vd_j$, given the specific observations $\Yd_j$; the Kalman filter updates this estimator sequentially.
Furthermore, since the Gaussian $\cN(v_j,C_j)$ is actually equal to the filtering distribution $\pi_j$
the methodology constitutes exact \emph{probabilistic estimation.}\index{probabilistic estimation}

\subsection{Steady State Kalman Filter}\index{Kalman filter!steady state}
\label{ssec:sskalman}

In this subsection we consider the Kalman filter\index{Kalman filter} in the setting where
$A_j = A$ and $H_j = H$ are constant in time. Then, under appropriate controllability and observability assumptions,  
 \index{gain!steady state}the covariance\index{covariance!steady state} 
 converges to a unique steady state. The resulting
 steady state covariance and gain\index{gain!Kalman}\index{gain!steady state} can be obtained by finding the unique solution $(\hC_\infty, S_{\infty}, K_{\infty})$ to the equations
\begin{subequations}
\label{eq:ss_kalman}
\begin{align}
    \hC_\infty &= A (I - K_{\infty}H)\hC_{\infty} A^\top + \Sigma,\\
    S_\infty &= H \hC_\infty H^\top + \Gamma,\\
    K_\infty &= \hC_\infty H^\top S_{\infty}^{-1} ,\label{eq:ss_kalman_gain}
\end{align}
\end{subequations}
and setting 
\begin{equation}
\label{eq:ss_kalman2}
C_{\infty} = (I - K_{\infty}H)\hC_{\infty}.
\end{equation}
If we assume that $C_0=C_{\infty}$ then, 
using the steady-state Kalman gain\index{gain!steady state} from (\ref{eq:ss_kalman}c) in the 
update formulae \eqref{eq:kalman_mean} we obtain
state estimator update equations in the form
\begin{subequations}
\label{eq:ssK}
\begin{align}
\hat{v}_{j+1} &= A v_j,\\ 
v_{j+1} &= \hat{v}_{j+1} + K_\infty \bigl(\yd_{j+1}-H\hat{v}_{j+1}\bigr).
\end{align}
\end{subequations}
We note the form of this update of the mean $v_j \mapsto v_{j+1}$, viewed as a state estimator: 
it comprises a prediction\index{prediction} step $v_j \mapsto \hv_{j+1},$ in the form of the stochastic dynamics model \eqref{eq:sdm} with the noise dropped, and an analysis\index{analysis} step $\hv_{j+1} \mapsto v_{j+1}$ which incorporates data given  by the data model \eqref{eq:dm}.
Combining the prediction and data incorporation steps we obtain
\begin{equation} \label{eq:ss3}
v_{j+1} = (I - K_\infty H)Av_j + K_\infty\yd_{j+1}.  
\end{equation}
Updating the mean according to this formula, and fixing the covariance at $C_{\infty}$ is
sometimes referred to as the steady state Kalman filter.\index{Kalman filter!steady state}

\subsection{3DVar}\index{3DVar}\index{variational!3-dimensional|see{3DVar}}
\label{ssec:3dvar}

The 3DVar algorithm does not attempt to approximate the filtering distribution; rather
it just attempts to provide a state estimator. However aspects of the algorithm
can be motivated by the form of the steady-state Kalman filter\index{Kalman filter!steady state} 
from the preceding subsection. To motivate 3DVar\index{3DVar} we consider the form of 
the Kalman filter when the covariance is in steady state.
Motivated by the mean update equations \eqref{eq:ssK}
for the steady state Kalman filter,\index{Kalman filter!steady state}
when the gain\index{gain!steady state} is in steady state, we propose the following generalization
to the setting of nonlinear $(\Psi,h):$
\begin{subequations}
\label{eq:pred3DVAR}
\begin{align}
\hat{v}_{j+1} &= \Psi(v_j)+s\xi_{j},\\ 
v_{j+1} &= \hat{v}_{j+1} + K \bigl(\yd_{j+1}-h(\hat{v}_{j+1})\bigr).	
\end{align}
\end{subequations}
The algorithm is initialized at $v_0$, typically sampled from Gaussian
$\cN(m_0,C_0)$, the known distribution of $\vd_0.$
In the setting where $h(\cdot) := H\cdot$ is linear
the algorithm can be written in the form
\begin{equation} \label{eq:3DVARL}
v_{j+1} = (I - KH)\Psi(v_j) + K\yd_{j+1} + s (I-KH)\xi_{j}.
\end{equation}

\begin{remark} \label{rem:CTOPF}
The standard form of \eqref{eq:pred3DVAR} employs $s=0$ and the resulting 
algorithm should be viewed as a state estimator\index{state estimation}.
Note also that, when $s=0$ and $\Psi(\cdot)=A\cdot$ is linear, \eqref{eq:3DVARL} recovers the 
same form as the steady state Kalman mean update \eqref{eq:ss3}.\index{Kalman filter!steady state}
We have included the case $s=1$ primarily so that we can make a connection
to the optimal particle filter\index{particle filter!optimal} introduced
in Subsection~\ref{ssec:opf}. In that context we will discuss the distribution
of the i.i.d. sequence $\{\xi_j\}_{j \in \bbZ^+.}$
\end{remark}

\begin{remark}
\label{rem:cycled}
    The methodology encapsulated in \eqref{eq:pred3DVAR} is in fact more properly termed \emph{cycled 3DVar}\index{3DVar!cycled}. 
    In this context the \emph{3DVar} component of the nomenclature refers to the analysis step (\ref{eq:pred3DVAR}b), which is \emph{cycled} with the prediction step (\ref{eq:pred3DVAR}a) by iterating over $j.$ For
    linear observation operator we show below that the analysis step has an optimization, or \emph{variational}, formulation; this is the source of the ``Var'' terminology. The use of  ``3D'' refers to the fact that, in weather forecasting where the methodology was introduced, the optimization problem is for a field in three physical space dimensions.
\end{remark}
    
We conclude this discussion of 3DVar by formulating the analysis step via an optimization problem.
To appreciate this connection to optimization we consider the case where $\Psi$ is allowed to be nonlinear, but $h(\cdot)=H\cdot$ is linear. The predict-then-optimize viewpoint then leads to 3DVar\index{3DVar} above if $v_{j+1}$ is computed from $(\hv_{j+1},\yd_{j+1})$ by solving the optimization problem
    \begin{subequations}\label{eq:3dvar_opt}
\begin{align} 
\J_{j+1}(v) &= \frac{1}{2}|v-\hv_{j+1}|^2_{\hC}+\frac12|\yd_{j+1}-h(v)|^2_{\Gamma},\\
v_{j+1} &\in \argmin_{v \in \R^d} \J_{j+1}(v).
\end{align}
\end{subequations}

\begin{remark}\label{rem:3dvar_map}
    The solution $v_{j+1}$ of \eqref{eq:3dvar_opt} is equivalent to the maximum \emph{a posteriori}\index{maximum \emph{a posteriori} estimation} solution of the Bayesian inverse problem with prior $\rho(v) = \mathcal{N}(\hv_{j+1}, \hC)$, and likelihood $\like(\yd_{j+1}|v) = \mathcal{N}(h(v), \Gamma)$; see Example~\ref{ex:loss-l2}.
\end{remark}

The minimization problem \eqref{eq:3dvar_opt} delivers (\ref{eq:pred3DVAR}b), in the case $h(\cdot)=H\cdot$, provided that $K$ in (\ref{eq:pred3DVAR}b) and $\hC$ in (\ref{eq:3dvar_opt}a) 
are related by
\begin{equation}
\label{eq:gain3}
    K = \hC H^\top (H \hC H^\top + \Gamma)^{-1}.
\end{equation}
When $h(\cdot)$ is nonlinear, an explicit solution of the optimization problem \eqref{eq:3dvar_opt} is not available in closed form. However, numerical optimization may be used to find $v_{j+1}$ from $\hv_{j+1}$ and $\yd_{j+1}.$

Equation \eqref{eq:gain3} should be compared with formula 
(\ref{eq:ss_kalman}a)--(\ref{eq:ss_kalman}c) defining
the steady-state Kalman gain.\index{gain!steady state}
In particular this comparison suggests a methodology for choosing $K$ in 3DVar: instead of
choosing $K$ directly,  choose an estimate of the uncertainty in the prediction, $\hC$, and use this to
define $K$ from \eqref{eq:gain3}. This approach to modeling $\hC$, and then deducing $K$, is natural because the uncertainty in the prediction is an interpretable quantity.

\subsection{Extended Kalman Filter}\index{Kalman filter!extended}\label{ssec:exkf}

The Kalman filter from Subsection~\ref{ssec:kalman} gives an exact expression
for the filtering distribution, which is Gaussian, in the setting of linear dynamics and observation operators. It is structured around
formulae for the prediction and analysis updates defined in equation \eqref{eq:pna}.
The extended Kalman filter (ExKF)\index{Kalman filter!extended}\index{ExKF|see{Kalman filter, extended}} generalizes the Kalman filter to nonlinear dynamics map $\Psi$ and nonlinear observation map $h$
by linearizing these functions, leading to \emph{approximate} Gaussian
representation of the filtering distribution. 
Specifically we seek the prediction and analysis
distributions as approximations of the form
\begin{equation}\label{eq:Gaussianpred99}
\Prob (\vd_{j+1}| \Yd_j) \approx \Nc \bigl(\vd_{j+1}; \widehat{v}_{j+1},\widehat{C}_{j+1}\bigr)
\end{equation}
for the predictive distribution; and
\begin{equation}\label{eq:Gaussianan99}
\Prob (\vd_{j+1}| \Yd_{j+1}) \approx \Nc \bigl(\vd_{j+1}; v_{j+1}, C_{j+1}\bigr)
\end{equation}
for the analysis distribution.
We now seek update rules to map $(v_j,C_j)$ to $(\widehat{v}_{j+1},\widehat{C}_{j+1})$, the
approximate prediction step; and to map $(\widehat{v}_{j+1},\widehat{C}_{j+1})$ to
$(v_{j+1},C_{j+1})$, the approximate analysis step. Doing this involves
computation of the extended Kalman filter gain\index{gain!ExKF}. 

The output $v_j$ from ExKF provides state estimation\index{state estimation} for $\vd_j,$ given $\Yd_j.$ The Gaussian $\cN(v_j,C_j)$ enables probabilistic estimation\index{probabilistic estimation} to be undertaken. For this reason we will
refer, for this algorithm, to $v_j$ as the mean, as for the Kalman filter.
We define the Jacobians of the functions and evaluate them at the outputs $v_j$ and $\hat{v}_j$ of a putative filtering algorithm:
\begin{subequations}
\label{eq:lin}
\begin{align}
    A_j &:= D\Psi(v_{j}),\\
    H_j &:= Dh(\hat{v}_j).
\end{align}
\end{subequations}
These Jacobians may be known analytically; otherwise, auto-differentiation\index{auto-differentiation} (Section \ref{sec:auto-differentiation}) can be used to obtain them.

The mean is initialized at $v_0=m_0$ and the update is then performed using the nonlinear $\Psi$ and $h$,
similarly to 3DVar\index{3DVar} \eqref{eq:pred3DVAR} with $s=0$, 
but with an evolving gain\index{gain!3DVar}:
\begin{subequations}\label{eq:exkf_mean}
\begin{align}
\hat{v}_{j+1} &= \Psi(v_j),\\ 
v_{j+1} &= \hat{v}_{j+1} + K_{j+1} \bigl(\yd_{j+1}-h(\hat{v}_{j+1})\bigr).
\end{align}
\end{subequations}
The computation of $K_{j+1}$, and the covariances involved in its
definition, uses \eqref{eq:Kalman_recursions}, but with evolving $(A_j,H_j)$
defined by \eqref{eq:lin}, \eqref{eq:exkf_mean}:
\begin{subequations}
\label{eq:wnrefd}
\begin{align}
\hat{C}_{j+1} &= A_j C_{j} A_j^\top + \Sigma,\\
S_{j+1} &= H_{j+1} \hat{C}_{j+1} H_{j+1}^\top + \Gamma,\\
K_{j+1} &= \hat{C}_{j+1} H_{j+1}^\top S_{j+1}^{-1},\label{eq:exkf_gain}\\
C_{j+1} &= (I - K_{j+1}H_{j+1})\hat{C}_{j+1}.
\end{align}
\end{subequations}
As for the Kalman filter, the gain facilitates transfer of information from observation\index{observations} 
space to state\index{state} space.

\begin{remark}
\label{rem:cycled2}
    The ExKF\index{Kalman filter!extended}\index{ExKF|see{Kalman filter, extended}} recovers the true filtering distribution in the case of linear $\Psi$ and $h$, where it reduces to the standard Kalman filter from Subsection~\ref{ssec:kalman}. It is often used beyond the linear Gaussian setting:
    for example when $\Psi$ and $h$ are close to linear; or when small covariances are assumed in both the dynamics and data models. In both these settings linearization may be used to justify  use of the ExKF.
\end{remark}

\subsection{Unscented Kalman Filter}\index{Kalman filter!unscented}
\label{ssec:ukf55}

The Kalman filter solves the filtering problem when the exact distribution on
state given observations is Gaussian; thus only a mean and covariance need
to be computed. The ExKF\index{Kalman filter!extended}\index{ExKF|see{Kalman filter, extended}} approximates the general filtering distribution by
a Gaussian, using linearization to derive the joint evolution of mean
and covariance. The unscented Kalman filter\index{Kalman filter!unscented} 
(UKF\index{UKF|see{Kalman filter, unscented}}) 
also makes a Gaussian ansatz, but does not require
linearization of the dynamics and observation maps;
instead it uses the projection $\Gn$ onto Gaussian measures, defined in Remark \ref{rem:bishop}.

The UKF\index{UKF|see{Kalman filter, unscented}} algorithm may be written
in the following form, which approximates the exact prediction--analysis cycle from \eqref{eq:pna}:
\begin{align}\label{eq:pnaU}
\begin{split}
&{ \text{ \bf  UKF Prediction Step:}} ~~~~\;  \hat{\post}_{j+1} = \Pred \post_j.  \index{prediction} \\
& {\text{ \bf UKF Analysis Step:}} ~~~~~\,\,\,\,\,  \post_{j+1} =\Un(\hat{\post}_{j+1};\yd_{j+1}). \index{analysis} 
\end{split}
\end{align}
Here, for $\gamma_\pi(u, y)$ defined by \eqref{eq:n4ukf},
we define the unscented approximation of the analysis operator $\An$ by 
\begin{equation}
  \Un(\pi;\yd)(u) = \frac{\Gn \gamma_\pi(u,\yd)}{\int_{\R^d} \Gn \gamma_\pi(u,\yd) \, d u}.\label{eq:filter_bayes_inferenceU}
\end{equation}

    The preceding gives an iteration with property that $\post_{j}$ is Gaussian
    for all $j$, but $\hat{\post}_{j+1}$ is not (unless $\Psi$ is linear).
    The algorithm thus produces an approximation of the filtering distribution
    which remains in the manifold of Gaussians; it can hence
    be represented in terms of propagation of means and covariances. As for the ExKF\index{ExKF}, it reduces to the Kalman filter if the dynamics and observation maps are linear. In practice the
    integrations required to define the Gaussian projection are approximated by quadratures; it is the resulting method, after quadrature, that is typically referred to as the unscented Kalman filter\index{unscented Kalman filter}.

\subsection{Ensemble Kalman Filter (EnKF)}\index{Kalman filter!ensemble}\index{EnKF|see{Kalman filter, ensemble}}
\label{ssec:enkf}

Despite the simplicity of the ExKF,\index{Kalman filter!extended}\index{ExKF|see{Kalman filter, extended}} computing and storing covariances when $\du$ or $\dy$ is large can be prohibitively
expensive. This motivates the ensemble methods we describe next. These methods carry an ensemble of particles which are used to compute low-rank approximations of covariances which enable, via the resulting gain, transfer of information from the observations\index{observations} to the state\index{state}. In this subsection our first, and primary, goal is to introduce the ensemble Kalman filter (EnKF) algorithm, a specific methodology of this type.
In addition, we also discuss inflation and localization, two
important practical modifications of the methodology.

\paragraph{The EnKF\index{EnKF|see{Kalman filter, ensemble}} Algorithm}

The content of Remark \ref{rem:cycled}, concerning 3DVar\index{3DVar}, is to shift 
the problem of choosing $K$ to one of choosing $\hC$, an estimate of the uncertainty in the predictions.
In the ExKF\index{Kalman filter!extended}\index{ExKF|see{Kalman filter, extended}} estimation of the uncertainty in predictions is achieved via a linearization. The EnKF\index{Kalman filter!ensemble}\index{EnKF} instead forms a low-rank\index{covariance!low-rank} approximation of the covariance based on an ensemble in state space. The algorithm operates by running an ensemble of $N$ interacting copies of 3DVar-like\index{3DVar!-like} algorithms. These are
then used to estimate the filtering distribution, and means and covariances in particular. The covariances
are used to define interactions between the particles.

Specifically we seek an approximation of the filtering distribution in the form
\begin{equation}
\label{eq:empEnKF}
\pi_j \approx \pie_j = \frac{1}{N}\sum_{n=1}^N \delta_{v_j^{(n)}}.
\end{equation}
The prediction and analysis distributions are sought as approximations of the form
\begin{equation}\label{eq:Gaussianpred88}
\hat{\post}_{j+1} \approx \pieh_{j+1} = \frac{1}{N} \sum_{n=1}^N \delta_{\,\hat{v}_{j+1}^{(n)}} 
\end{equation}
and
\begin{equation}\label{eq:enkf_emp_an}
\pi_{j+1} \approx  \pie_{j+1} = \frac{1}{N}\sum_{n=1}^N \delta_{v_{j+1}^{(n)}}.
\end{equation}

We now describe the update rules for the prediction step  $\{v_{j}^{(n)}\}_{n=1}^N \mapsto 
\{\hat{v}_{j+1}^{(n)}\}_{n=1}^N$ and the analysis 
step $\{\hat{v}_{j+1}^{(n)}\}_{n=1}^N
\mapsto \{v_{j+1}^{(n+1)}\}_{n=1}^N,$ approximating \eqref{eq:pna}.
These updates will involve independent copies of the random variables $\xid_j$ and $\etad_j$ entering the signal and data models \eqref{eq:sdm} and \eqref{eq:dm}. Specifically,
we make the following assumption:
\begin{assumption} \label{a:enkf-noise}
Here $\xi_{j}^{(\sam)} \sim \Nc(0,\Sigma), \, \eta_{j+1}^{(\sam)} \sim \Nc(0,\Gamma)$
are independent arrays of i.i.d. random vectors with respect to both $j$ and $n$, and
the two arrays themselves are independent of one another. The initial conditions for the
ensemble at $j=0$ are also chosen independently from one another and from these i.i.d. arrays.
\end{assumption}

Unlike 3DVar, the updates for $\{\hat{v}_{j+1}^{(n)}\}_{n=1}^N$ employ a time-dependent gain\index{gain!EnKF} $K_{j+1}$; this gain is estimated
by computing the empirical\index{empirical} covariances of the predicted states
$\{\hat{v}_{j+1}^{(n)}\}_{n=1}^N$ and 
$\{h\bigl(\hat{v}_{j+1}^{(n)}\bigr)\}_{n=1}^N$, 
their mappings under $h$. The algorithm takes the form
\begin{subequations}
\label{eq:predEnKF}
\begin{align}
\widehat{v}_{j+1}^{(\sam)} &= \Psi(v_{j}^{(\sam)})+\xi^{(\sam)}_{j}, \quad \sam=1,\ldots,\Sam, \label{eq:predEnKF_model}\\
v_{j+1}^{(\sam)} &= \widehat{v}_{j+1}^{(\sam)}+K_{j+1}\bigl(\yd_{j+1}-\eta_{j+1}^{(\sam)}-h(\widehat{v}_{j+1}^{(\sam)})\bigr),\quad \sam=1,\ldots,\Sam. 
\end{align}
\end{subequations}
The gain\index{gain} matrix $K_{j+1}$ is calculated according to
\begin{subequations}
\label{eq:gainEnKF}
\begin{align}
\widehat{m}_{j+1} &= \frac{1}{\Sam}\sum^{\Sam}_{\sam=1} \widehat{v}_{j+1}^{(\sam)}, \\
\widehat{h}_{j+1} &= \frac{1}{\Sam}\sum^{\Sam}_{\sam=1} h(\widehat{v}_{j+1}^{(\sam)}), \\
\widehat{C}_{j+1}^{vh} &= \frac{1}{\Sam}\sum^{\Sam}_{\sam=1}\bigl(\widehat{v}^{(\sam)}_{j+1}-\widehat{m}_{j+1}\bigr)\otimes \bigl(h(\widehat{v}^{(\sam)}_{j+1})-\widehat{h}_{j+1}\bigr),\\
\widehat{C}_{j+1}^{hh} &= \frac{1}{\Sam}\sum^{\Sam}_{\sam=1}\bigl(h(\widehat{v}^{(\sam)}_{j+1})-\widehat{h}_{j+1}\bigr)\otimes \bigl(h(\widehat{v}^{(\sam)}_{j+1})-\widehat{h}_{j+1}\bigr),\\
\widehat{C}_{j+1}^{yy} &= \widehat{C}_{j+1}^{hh}+\Gamma,\\
K_{j+1} &= \hat{C}_{j+1}^{vh}\bigl(\hat{C}_{j+1}^{yy}\bigr)^{-1}.
\end{align}
\end{subequations}

As well as the empirical approximation of the filtering distribution,
defined by \eqref{eq:empEnKF}, it is also possible, and indeed natural, to approximate $\pi_j$ with a Gaussian computed
using empirical moments. Recall the notation
for $\Gn$, the projection onto Gaussian measures, defined in Remark \ref{rem:bishop}. Then note that
\begin{equation}
\label{eq:gauss_proj}    
\Gn\Bigl(\frac{1}{N}\sum_{n=1}^N \delta_{v_j^{(n)}}\Bigr)=\Nc(m_j,C_j),
\end{equation}
where the ensemble mean $m_{j}$ and covariance $C_j$ are given by 
\begin{equation}
\label{eq:mcenkf}
{m}_{j} = \frac{1}{\Sam}\sum^{\Sam}_{\sam=1} {v}_{j}^{(\sam)}, \quad
{C}_{j} = \frac{1}{\Sam}\sum^{\Sam}_{\sam=1}\bigl({v}^{(\sam)}_{j}-{m}_{j}\bigr)\otimes \bigl({v}^{(\sam)}_{j}-{m}_{j}\bigr).
\end{equation}
Whilst \eqref{eq:empEnKF} provides probabilistic estimation\index{probabilistic estimation}, 
mean $m_j$ may be viewed as providing state estimation\index{state estimation}.

\begin{remark}
\label{rem:enkf}
Conditions under which the EnKF\index{filter!ensemble Kalman} provides a good approximation of
the true filtering distribution are discussed in the bibliography Section \ref{sec:dabib}, and revolve around Gaussian approximations. When the true filtering and predictive distributions are close to Gaussian then the EnKF
accurately approximates these distributions if sufficient particles are used.
When the true filtering distribution is far from Gaussian, however, the method will typically fail to 
 provide accurate probabilistic estimation. In this case, the EnKF can still provide accurate state estimation under observability conditions that ensure the data is sufficiently informative, as discussed in the bibliography Section \ref{sec:dabib}.   
\end{remark}

\begin{remark}
    \label{rem:lh}
We note that in the case where $h(\cdot)=H\cdot$ (and is hence linear) we may calculate the gain\index{gain} through estimation of a single covariance, as follows:
\begin{subequations}
\begin{align}
\label{eq:lh}
\widehat{C}_{j+1} &= \frac{1}{\Sam}\sum^{\Sam}_{\sam=1}\bigl(\widehat{v}^{(\sam)}_{j+1}-\widehat{m}_{j+1}\bigr)\otimes \bigl(\widehat{v}^{(\sam)}_{j+1}-\widehat{m}_{j+1}\bigr),\\
K_{j+1} &= \hat{C}_{j+1} H^\top \bigl(H \hat{C}_{j+1} H^\top + \Gamma\bigr)^{-1}.\label{eq:single_cov}
\end{align}
\end{subequations}
This should be compared with the gain\index{gain!3DVar}~\eqref{eq:gain3}, resulting from 3DVar when applied in the linear case $h(\cdot)=H\cdot$.
The analysis mean and covariance can be estimated using \eqref{eq:mcenkf} with index $j \mapsto j+1.$
\end{remark}

\begin{remark} \label{rem:fake}
The particle system \eqref{eq:predEnKF} is sometimes referred to as the \emph{perturbed
observation}\index{observation!perturbed} form of the EnKF;\index{Kalman filter!ensemble, perturbed observations}\index{EnKF|see{Kalman filter, ensemble}} 
this is because of the presence of the noise
$\eta^{(\sam)}_{j+1}$ which is sometimes viewed by practitioners as perturbing the
observation $\yd_{j+1}$. We present an alternative
way of thinking about the algorithm, in terms of \emph{simulated observations}\index{observation!simulated}. Equations \eqref{eq:predEnKF} can be
written in the equivalent form
\begin{subequations}
\label{eq:fakeobs}
\begin{align}
\widehat{v}_{j+1}^{(\sam)} &= \Psi(v_{j}^{(\sam)})+\xi^{(\sam)}_{j}, \quad \sam=1,\ldots,\Sam,\\
\widehat{y}_{j+1}^{(\sam)} &= h(\hv_{j+1}^{(\sam)})+\eta^{(\sam)}_{j+1}, \quad \sam=1,\ldots,\Sam, \label{eq:predEnKF_model2}\\
v_{j+1}^{(\sam)} &= \widehat{v}_{j+1}^{(\sam)}+K_{j+1}\, i_{j+1}^{(\sam)}, 
\quad \sam=1,\ldots,\Sam. 
\end{align}
\end{subequations}
Here define the \emph{innovation}\index{innovation} by
\begin{equation}
\label{eq:inn}
i_{j+1}^{(\sam)}  =\yd_{j+1}-\widehat{y}_{j+1}^{(\sam)}.
\end{equation}
This way of writing the EnKF leads to an interpretation of the
algorithm as predicting both the state $\widehat{v}_{j+1}^{(\sam)}$ and the observation $\widehat{y}_{j+1}^{(\sam)}$,
and then correcting the predicted state by a quantity linearly related to
the discrepancy between the observation $\yd_{j+1}$ and the predicted observation.
\end{remark}

\begin{remark} \label{rem:glandmand}
Consider the setting of linear $\Psi$ and $h$. With the addition of the perturbations to the observations, as described in the preceding remark, the predictive and filter covariances in \eqref{eq:lh} and \eqref{eq:mcenkf} match those of the Kalman filter, in the large ensemble limit $N=\infty$. See the bibliography Section \ref{sec:dabib}. 
\end{remark}

EnKFs are widely used in high-dimensional problems because they do not suffer from the weight
collapse\index{weight collapse} that plagues the particle filters introduced in Subsection~\ref{ssec:pf}.
Weight collapse is discussed in Subsection~\ref{ssec:collapse}. 
We refer to EnKFs as \emph{equal weights particle filters}.\index{particle filter! equal weights}
However, EnKFs often need inflation and localization to perform well for high-dimensional systems, for chaotic dynamical systems, and for small ensemble sizes.  Inflation and localization are discussed in the following two paragraphs.

\paragraph{Inflation}\index{inflation}

For simplicity we confine the discussion to the case of linear
observation operator $h(\cdot)=H\cdot.$ The EnKF gain\index{gain!EnKF}
is then determined by $\widehat{C}_{j+1}$, as defined in Remark \ref{rem:lh}. Because we
are interested here in dependence on sample size we write $\widehat{C}_{j+1}(N).$
We assume that the covariance at infinite sample size $N=\infty$, $\widehat{C}_{j+1}(\infty)$,
leads to the optimal choice of gain\index{gain!optimal}. This assumption is provably true, 
utilizing a precise definition of ``optimal'', in the case of linear $\Psi$; see the
discussion in the bibliography. From this perspective, the EnKF suffers 
from sampling error due to finite ensemble size $N$. 
When sampling error in the forecast covariance matrix $\widehat{C}_{j+1}(N)$ leads to underestimation of the
optimal analysis covariance $C_{j+1},$ this can lead to filter divergence,\index{filter divergence} whereby repeated underestimation of the forecast covariance causes the filter to put increasingly more weight on the forecasts than observations, eventually becoming unresponsive to observations. We provide two examples to illustrate the underestimation of the analysis covariance.

\begin{example}\label{ex:sampling_error}
Suppose that we assimilate a scalar observation of the $b^\text{th}$ variable into the $a^\text{th}$ variable. Thus $H$ is a $1\times d$ vector with entry $1$ in the $b^\text{th}$ position and zeros elsewhere.  Since
the phenomenon we wish to illustrate occurs at any fixed time, we simply let $C$ and $\widehat{C}$ represent the desired
infinite sample size covariances, and $C(N)$ and $\widehat{C}(N)$ the finite sample size estimators coming
from the EnKF. Now, we claim that
    \begin{equation}\label{eq:enkfexample}
       \Expect[\bigl({C}(N)\bigr)_{aa}] - \bigl(\widehat{C}(N)\bigr)_{aa} =  \frac{-(\widehat{C}(N))_{ab}^2}{\bigl(\widehat{C}(N)\bigr)_{bb} + (\Gamma)_{bb}},
    \end{equation}
    where the expectation is taken with respect to the observation perturbations $\eta_j^{(n)}$. References that substantiate this claim can be found in the bibliography Section \ref{sec:dabib}. Next, suppose that variables with indices $a$ and $b$ are in fact uncorrelated, so that $\bigl(\widehat{C}\bigr)_{ab} = 0.$ When estimated with a finite ensemble,  in an implementation of the EnKF,  the estimated $\bigl(\widehat{C}(N)\bigr)_{ab} \neq 0$, due to sampling error resulting from a finite ensemble. This leads to a spurious reduction in the analysis covariance, with respect to its optimal value that is defined by the infinite-ensemble case.
    \end{example}

\begin{example}\label{ex:sampling_error2}
Inflation is also used to compensate for an unknown, or ignored, $\Sigma$, a form of model error\index{model error}. If the $\xi^{(\sam)}_{j}$ are excluded in 
\eqref{eq:predEnKF_model}, then the covariance 
$\widehat{C}_{j+1}$ will be underestimated in (\ref{eq:lh}).
\end{example}

The underestimation of the analysis covariance can be mitigated by inflating the covariance of the forecast ensemble, although inflation can also be applied to the analysis ensemble instead. The most common form of inflation is \emph{multiplicative inflation}, whereby the forecast ensemble is modified as
\begin{equation}\label{eq:inflation}
    \widehat{v}_{j+1}^{(\sam)} \to \widehat{m}_{j+1} + \alpha (\widehat{v}_{j+1}^{(\sam)} - \widehat{m}_{j+1}),
\end{equation}
with $\alpha \geq 1$ being the inflation parameter. This corresponds to scaling the covariance $\widehat{C}_{j+1}$ by $\alpha^2$.
Theoretical results on the impact of sampling error on EnKFs, including results on the optimal value of the inflation parameter under assumptions on the model dynamics, are referenced in the bibliography Section \ref{sec:dabib}.

\paragraph{Localization}\index{localization}

In addition to the undersampling effects from the previous paragraph,
further negative impacts of sampling errors on EnKFs can arise in spatially extended systems. In such physical systems there is typically decay of correlations with distance; but sampling error\index{sampling error} can induce spurious long-range correlations. In the empirical\index{empirical} covariance matrices computed by the EnKFs, nearby points in space are likely to be truly correlated, while ones that are farther apart may only have spurious correlations dominated by sampling error. These spurious correlations not only lead to an underestimation of the analysis covariance, as discussed in Example \ref{ex:sampling_error}, but also lead to degraded state estimation due to spurious information transfer between uncorrelated variables.

\emph{Localization}\index{localization} is introduced to damp covariances, within the empirical\index{empirical} covariance matrices computed by the EnKF, according to physical distance
(for discretization of PDE problems) or index distance (for large ODEs with notion of index locality). A common way of implementing localization is by taking the Hadamard\index{Hadamard product} product\footnote{The Hadamard
product computes the product of two matrices
of the same dimensions elementwise; it is also known as the Schur\index{Schur product} product.} of the empirical\index{empirical} covariance with a localization matrix $L$:
\begin{equation}\label{eq:loc_cov}
    \widehat{C}_{j+1}\to L \circ \widehat{C}_{j+1}.
\end{equation}
Typically, $L$ is constructed such that the covariance between variables indexed
by integers $a$ and $b$ is exponentially damped according to a measure of 
distance\footnote{Computed, for example, using a deterministic scoring rule\index{scoring rule!distance-like deterministic} from Definition \ref{def:det_score}.} $\dd(a,b)$ between them, scaled by characteristic length scale $\ell$:
\begin{equation}\label{eq:loc}
    (L)_{ab} = e^{-\dd(a,b)^2/\ell^2}.
\end{equation}
The choice of $\ell$ will depend on the ensemble size, the correlation scales in the system, and the time between analysis steps (since this will control how far information has had time to propagate). Localization can also be implemented in other ways that scale better to high-dimensional problems; these include domain localization and observation localization, discussed in the bibliography Section \ref{sec:dabib}. In the main body of the text,
we restrict our discussion to covariance localization \eqref{eq:loc_cov}.

\begin{remark}
\label{r:loc1}    
If $h(\cdot)$ is linear, the localized $\widehat{C}_{j+1}$ is used to compute the 
gain\index{gain} according to \eqref{eq:single_cov}. If $h(\cdot)$ is nonlinear, then localization needs to be applied to both $\widehat{C}_{j+1}^{vh}$ and $\widehat{C}_{j+1}^{hh}$ to compute the gain according to (\ref{eq:gainEnKF}f). If function $h(\cdot)$ is such that each observation has a corresponding spatial location, then localization can be applied as discussed above. If some observations correspond to, for example, spatially integrated quantities, then localization is not straightforward to apply; see the bibliography 
Section \ref{sec:dabib}.
\end{remark}

\begin{remark}
\label{r:loc2}  
Besides the sampling error considerations discussed above, localization is also important in increasing the rank of the forecast covariance matrix. When the ensemble size $N$ is less than the system dimension $d$, the forecast covariance will be rank deficient. This implies that the analysis ensemble will lie in the span of the forecast ensemble, and thus that analysis increments are restricted to an $N$-dimensional subspace. Localization typically increases the rank of the forecast covariance, bypassing this subspace property.
\end{remark}

\subsection{Bootstrap Particle Filter}\index{particle filter}\index{particle filter!bootstrap}
\label{ssec:pf}

The starting point for the bootstrap particle filter (BPF) is the factorization
\eqref{eq:pna} of the filtering update into two steps: prediction composed with analysis. We summarize this factorization here:
\begin{equation}\label{eq:pnaB}
\hat{\post}_{j+1} = \Pred \post_j,\,\,\,\,\,  \post_{j+1} = \An_j (\hat{\post}_{j+1}).
\end{equation}
The method starts with an empirical\index{empirical} approximation to the filtering distribution
$\pi_j \approx \pip_j$. Here $\pip_0=\pi_0=\cN(m_0,C_0)$
and, for $j \in \mathbb{N}$,
\begin{equation}
\label{eq:pip2}    
\pip_j = \sum_{n=1}^N w_{j}^{(n)}\delta_{\widehat v_{j}^{(n)}}.
\end{equation}
To define this approximation we evolve the particles $\widehat v_{j}^{(n)}$
(the prediction step) and the weights $w_{j}^{(n)}$ (the analysis step).
This is done according to the following update formulae, which hold for $\sam=1,\ldots,\Sam$:
\begin{subequations}
\label{eq:bpf2}
\begin{align}
    \widehat v_{j+1}^{(n)} &= \Psi \bigl(v_j^{(n)}\bigr) + \xi_j^{(n)},\qquad v_{j}^{(n)} \overset{\rm i.i.d.} {\sim} \pip_j,\\
    \ell_{j+1}^{(n)} &= \exp \left( - \frac{1}{2} \bigl\lvert \yd_{j+1} - h\bigl(\widehat v_{j+1}^{(n)}\bigr) \bigr\rvert_{\Gamma}^2 \right),\\
w_{j+1}^{(n)}&= \ell_{j+1}^{(n)}\Big/\Bigl(\sum_{m=1}^N \ell_{j+1}^{(m)}\Bigr).
\end{align}
\end{subequations}
Here $\xi_{j}^{(n)} \sim \cN(0, \Sigma)$  are Gaussian random variables,
i.i.d.\ with respect to both $n$ and $j,$ generalizing the Assumption \ref{a:noise} 
concerning the underlying signal--observation model.

With this we obtain
\begin{equation}
\label{eq:pip22_pred}    
\widehat v_{j+1}^{(n)} \sim \Pred \pip_j,\, \text{i.i.d.\quad and}  \quad \sum_{n=1}^N \frac1N\delta_{\widehat v_{j+1}^{(n)}} \approx \Pred \pip_j.
\end{equation}
The approximation error in this Monte Carlo approximation of $\Pred \pip_j$ may be quantified using Theorem \ref{t:MC}. We may then note that
\begin{equation}
\label{eq:pip22_an}    
\pip_{j+1} = \sum_{n=1}^N w_{j+1}^{(n)}\delta_{\widehat v_{j+1}^{(n)}}=
\An_j\Bigl( \sum_{n=1}^N \frac1N\delta_{\widehat v_{j+1}^{(n)}} \Bigr) \approx \An_j\bigl(\Pred \pip_j\bigr).
\end{equation}
Notice that the approximation in \eqref{eq:pip22_an} arises from applying Bayes formula (the analysis map $\An_j$) with a Monte Carlo approximation of the prior (here $\Pred \pip_j$). Thus, the error in this approximation can be quantified using Theorem \ref{thm:emp_prior_perturbation}  in conjunction with Theorem \ref{t:MC}.
The resulting algorithm that implements the update $\pip_{j} \mapsto \pip_{j+1}$ is known as the BPF. It performs
probabilistic estimation\index{probabilistic estimation}: \eqref{eq:pip2}
delivers an approximation $\pip_{j}$ to the true filtering distribution $\pi_j.$

\begin{remark}
\label{rem:pf}
The BPF\index{particle filter!bootstrap}
is provably convergent to the true filtering distribution
as $N \to \infty$. In particular this convergence holds for
filtering distributions that are far from Gaussian. This fact should be contrasted
with ensemble methods as discussed in Remark \ref{rem:enkf} -- they only provide accurate probabilistic estimation in the near-Gaussian setting. On the other hand, the BPF behaves poorly in many high-dimensional problems -- see Subsection~\ref{ssec:collapse}.
\end{remark}

\subsection{Optimal Particle Filter}\index{particle filter!optimal}
\label{ssec:opf}

The optimal particle filter\index{particle filter!optimal} (OPF) is also a probabilistic
estimation\index{probabilistic estimation} methodology and like the BPF it is based
on a particle approximation of the filter $\pi_j.$ The OPF differs from the BPF
by reversing the order of the Bayesian inference step and the sampling from a Markovian kernel based on the dynamics.
It is directly implementable when the observation operator $h(\cdot)$ is linear. We describe
the general formulation, followed by details in the linear observation setting.

\subsubsection{OPF: General Setting}
The starting point to understand the OPF is the following identity: 
\begin{equation} \label{eq:aoptd}
	\Prob(\vd_{j+1}|\Yd_{j+1}) 
 = \int_{\Ru} \Prob(\vd_{j+1}|\vd_j,\Yd_{j+1}) \Prob(\vd_j|\Yd_{j+1}) \, d\vd_j.
\end{equation}
Recall that $\pi_j=\Prob(\vd_j|\Yd_{j}).$ Now note that, using Bayes Theorem \ref{t:bayes} \index{Bayes Theorem} and Assumption \ref{a:noise}
\begin{equation}
    \label{eq:aint}
\intp_{j+1}:=\Prob(\vd_{j}|\Yd_{j+1})=\frac{\Prob(\yd_{j+1}|\vd_j,\Yd_j)}{\Prob(\yd_{j+1}|\Yd_{j})}  \Prob(\vd_j|\Yd_{j})=\frac{\Prob(\yd_{j+1}|\vd_j)}{\Prob(\yd_{j+1}|\Yd_{j})}  \Prob(\vd_j|\Yd_{j})=:\mathsf{A}_j^{\textrm{OPF}}(\pi_j).
\end{equation}
Note also that, using again Assumption \ref{a:noise},
which defines the structure of a hidden Markov model,\index{hidden!Markov model} we have
\begin{equation}
\Prob(\vd_{j+1}|\vd_j,\Yd_{j+1})=\Prob(\vd_{j+1}|\vd_j,\yd_{j+1}).
\end{equation}
These identities show that we may factorize  
the filtering update into: 
(i) $\post_j \mapsto \intp_{j+1}$, defined by \eqref{eq:aint}; 
and (ii) $\intp_{j+1} \mapsto \post_{j+1}$, where 
\begin{equation} \label{eq:pint}
\post_{j+1}(\vd_{j+1})=\int_{\Ru} \Prob(\vd_{j+1}|v,\yd_{j+1})\intp_{j+1}(v)  \, dv
=: \mathsf{P}_j^{\textrm{OPF}}\intp_{j+1}.
\end{equation}
The two steps in this factorization may be summarized as
\begin{equation}
\label{eq:fact02}    
\intp_{j+1}= \mathsf{A}_j^{\textrm{OPF}}(\pi_j), \quad \pi_{j+1} = \mathsf{P}_j^{\textrm{OPF}}\intp_{j+1}.
\end{equation}
Note that both steps depend on the data $\yd_{j+1}$.
This factorization is the basis of the OPF\index{particle filter!optimal}. 
The factorization should be compared with \eqref{eq:pnaB}, the factorization underlying the
BPF.\index{particle filter!bootstrap}

Assume that we have a particle approximation of the true filter at time $j$, 
in the form
\begin{equation} 
\label{eq:pop}
\pop_j = \sum_{n=1}^N w_{j}^{(n)}\delta_{\widehat v_{j}^{(n)}}.
\end{equation}
We wish to update the weights and particles to mimic the two steps in the factorization
\eqref{eq:fact02}. To update the weights we proceed as follows
\begin{subequations}
\label{eq:pzp222} 
\begin{align}
v_{j}^{(n)} & \sim  \pop_j\,\,\mathrm{i.i.d.}\, ,\\
w_{j+1}^{(n)} & \propto \Prob(\yd_{j+1}|v_{j}^{(n)}),
\end{align}
\end{subequations}
where the weights $w_{j+1}^{(n)}$ are normalized to sum to $1$ over $n,$ 
defining the constant of proportionality. We then have
\begin{equation}
\label{eq:pzp22}    
\sum_{n=1}^N w_{j+1}^{(n)}\delta_{v_{j}^{(n)}} \approx \An_j^{\textrm{OPF}}(\pop_j),
\end{equation}
and we have mimicked step (i) in the factorization
\eqref{eq:fact02}. To mimic step (ii)
we define updated particles 
$$\widehat v_{j+1}^{(n)} \sim \Prob(\vd_{j+1}|v_{j}^{(n)},\yd_{j+1})\,\,\mathrm{i.i.d.}\, ,$$
and note that then
\begin{equation} 
\label{eq:pop2}
\pop_{j+1} = \sum_{n=1}^N w_{j+1}^{(n)}\delta_{\widehat v_{j+1}^{(n)}}
\approx \Pred_j^{\textrm{OPF}}\sum_{n=1}^N w_{j+1}^{(n)}\delta_{v_{j}^{(n)}}.
\end{equation}
Similarly to the BPF, the Monte Carlo approximation may be quantified using Theorem \ref{t:MC}.
Together \eqref{eq:pzp22} and \eqref{eq:pop2} approximate \eqref{eq:fact02}.

\subsubsection{OPF: Linear Observation Operator}

In general nonlinear settings, it may not be possible to implement the OPF exactly. This arises for two reasons. First, the likelihood weights in (\ref{eq:pzp222}b) are naturally expressed in terms of the integral
$$\Prob(\yd_{j+1}|v_{j}^{(n)}) = \int \Prob(\yd_{j+1}|\vd_{j+1})\Prob(\vd_{j+1}|v_{j}^{(n)}) \, d\vd_{j+1},$$
but this may not have a closed form. Second, sampling exactly from 
$\Prob(\vd_{j+1}|v_{j}^{(n)},\yd_{j+1})$, as required to define \eqref{eq:pop2}, may not be possible. 
However, one
(frequently occurring) setting where it is tractable to evaluate the likelihood weights and sample from $\Prob(\vd_{j+1}|v_{j}^{(n)},\yd_{j+1})$ is when the observation model is linear: $h(\cdot) = H\cdot$ for some $H \in \R^{k \times d}$. The state and observation models \eqref{eq:sdm} and \eqref{eq:dm}
then reduce to the form
\begin{subequations}
\label{eq:linearobs}
\begin{align}
\vd_{j+1} &= \Psi(\vd_j) + \xid_j , \\
\yd_{j+1} &= H\vd_{j+1} + \etad_{j+1}\,.
\end{align}
\end{subequations}
We make the same Assumptions \ref{a:noise} and \ref{a:fas} as made for equations \eqref{eq:sdm} and \eqref{eq:dm}.
In the algorithm we are about to define
the covariance $C$, gain $K$ and matrix $S$ are defined by
\begin{subequations}
    \label{eq:covopt} 
\begin{align}
C &= (I - KH)\Sigma, \\
K &= \Sigma H^\top S^{-1}, \\
S &= H\Sigma H^\top + \Gamma.
\end{align}
\end{subequations}
Note that then
$$\yd_{j+1} =H\Psi(\vd_j)+\zeta^\dagger_{j+1},$$
where $\zeta^\dagger_{j+1} \sim \Nc(0,S).$

Using that $\Prob(\yd_{j+1}|v_{j}^{(n)})$ and $\Prob(\vd_{j+1}|v_{j}^{(n)},\yd_{j+1})$ are Gaussian
leads to the following OPF algorithm. First set $\pop_0=\pi_0=\cN(m_0,C_0)$.
Then recall, for $j \in \mathbb{N}$, the desired approximate filtering distribution $\pop_{j}$ given by \eqref{eq:pop}. The particles $\widehat v_{j}^{(n)}$ and weights $w_{j}^{(n)}$ evolve according to, for $\zeta_{j+1}^{(n)} \sim\text{i.i.d.}\thinspace \mathcal{N}(0,C)$, and for $\sam=1,\ldots,\Sam$,
\begin{subequations}
\label{eq:opf99}
\begin{align}
    \widehat{v}_{j+1}^{(n)} &= (I - KH)\Psi(v_j^{(n)}) + K\yd_{j+1} + \zeta_{j+1}^{(n)},
    \qquad v_{j}^{(n)} \overset{\rm i.i.d.} {\sim} \pop_j,\\
    \ell_{j+1}^{(n)} &= \exp\left(-\frac12|\yd_{j+1} - H\Psi(v_j^{(n)})|_S^2\right),\\
    w_{j+1}^{(n)} &= \ell_{j+1}^{(n)} / \sum_{m=1}^N \ell_{j+1}^{(m)}.
\end{align}
\end{subequations}
Now define $\pop_{j+1}$ according to \eqref{eq:pop2}, set $j \mapsto j+1$, and repeat the above steps.
The map from \eqref{eq:pop} to \eqref{eq:pop2} approximates the true filter
update \eqref{eq:An}. We again emphasize that the ensemble of predictions (\ref{eq:opf99}a)
are  3DVar-like\index{3DVar!-like}; see Remark \ref{rem:CTOPF}. Thus the complete
algorithm is a form of ensemble 3DVar\index{3DVar!ensemble}.

\begin{remark}
\label{rem:opt}
\emph{Optimality} here refers to minimizing the variance
of the weights, in one step of the filter, over a wide
class of possible particle-based methods. Once
cycled through multiple steps $j$, this optimality property
is lost. Nonetheless the approach of using the data to predict is
natural. Indeed the OPF has a clear potential advantage over the BPF: the predictive kernel in the OPF, $\intp_{j+1}$, incorporates knowledge of the current observation $\yd_{j+1}$. In contrast, the predictive kernel in the BPF simply predicts with the unconditioned forecast kernel $\Pred,$ drawing samples from $\Prob(v_{j+1}|v_j)$ without knowledge of $\yd_{j+1}$, and then reweights  
them to reflect the observation.
\end{remark}

\subsection{Weight Collapse for Bootstrap and Optimal Particle Filters}
\label{ssec:collapse}
The BPF\index{particle filter!bootstrap} can perform poorly in high dimensions,
suffering from weight collapse\index{weight collapse}, a phenomenon whereby, for $j$ large, all weights 
$\{w_j^{(n)}\}_{n=1}^{N}$ 
are close to zero, with the exception of one weight close to one. As a consequence
the ensemble approximation $\pip_j$ is dominated by one particle. Although the OPF obviates this slightly by 
minimizing weight variance, the problem still persists. Ensemble methods, on the other hand,
have been designed to perform robustly in high dimensions, and have equal weights on all
particles; again see Remark \ref{rem:enkf} for further discussion of
methods used to impart robustness on the EnKF in high dimensions.

On the other hand, the BPF and OPF are provably convergent to the true
filtering distribution, as the number of particles $N \to \infty.$ In
contrast the EnKF and UKF are only provably convergent in the linear
Gaussian setting. These convergence issues are discussed in the bibliography
Section~\ref{sec:dabib}. In Subsections~\ref{ssec:PLA1} and \ref{ssec:PLA2} we propose a methodology
which aims to improve the accuracy of equal weights particle filters, with
respect to the true filtering distribution, using ideas from the BPF and the OPF.

\subsection{Evaluating Probabilistic Estimation}
\label{ssec:eval}

In this subsection we describe two widely used methodologies for 
evaluating probabilistic estimation of the filtering distribution, the spread--error ratio and rank histograms. The approximate probability distributions, and the corresponding true trajectory, are assumed to be available, as per the following Data Assumption.

\begin{dataassumption}\label{da:eval_prob}\index{Data Assumption}
    Assume we have a sequence of probabilistic forecasts $\{\pia_j\}_{j=1}^J$ generated using an approximate filtering methodology, along with the corresponding true trajectory $\{\vd_j\}_{j=0}^J$ generated from \eqref{eq:sdm}.
\end{dataassumption}

\begin{remark}
    Typically the probabilistic forecasts will have used observations $\{\yd_j\}_{j=1}^J$, from \eqref{eq:dm}, generated
    in concert with $\{\vd_j\}_{j=0}^J$ from \eqref{eq:sdm}, but we do not make this explicit in Data Assumption \ref{da:eval_prob} as the methodology for \emph{evaluating}  probabilistic forecasts does not require it. However, the optimality properties that underpin the evaluation methodologies that we now describe refer to the true filter which will have used $\{\yd_j\}_{j=1}^J$. For this reason, of course, any reasonable approximate filter will also have used 
    $\{\yd_j\}_{j=1}^J$.
\end{remark}

\subsubsection{Spread--Error Ratio}

The use of the \emph{spread--error ratio} as a diagnostic of probabilistic calibration is based on the following theorem which
shows that the filtering distribution has an
interesting property: its variance, averaged over the marginal distribution on
observations, equals the squared bias in its mean, 
averaged over the joint distribution of the state and observations.

\begin{theorem}
\label{t:ser}
In the following, the pair $(\vd_j,\Yd_j)$ is distributed according
to the joint distribution under the dynamics/data model \eqref{eq:sdm}, \eqref{eq:dm}, with expectation denoted $\mathbb{E}^{(\vd_j,\Yd_j)}$, and the  expectation under the marginal distribution for $\Yd_j$ is denoted by $\mathbb{E}^{\Yd_j}$. Furthermore, 
$v_j$ is distributed according to the 
filtering distribution for $\vd_j | \Yd_j,$ with 
expectation denoted by $\mathbb{E}$. Then
   \begin{equation*}
        \frac{\mathbb{E}^{\Yd_j}\mathbb{E}\Bigl|v_j - \mathbb{E}v_j\Bigr|^2}{\mathbb{E}^{(\vd_j,\Yd_j)}\Bigl|\vd_j-\mathbb{E}v_j\Bigr|^2} = 1.
    \end{equation*}
\end{theorem}
\begin{proof}
    This is a consequence of the properties of conditional probability,
    using that $\mathbb{P}(\vd_j,\Yd_j)$ can be factored as 
    $\mathbb{P}(\vd_j|\Yd_j)\mathbb{P}(\Yd_j).$  Thus
    $$\mathbb{E}^{(\vd_j,\Yd_j)}\Bigl|\vd_j-\mathbb{E}v_j\Bigr|^2=\mathbb{E}^{\Yd_j}\Bigl(\mathbb{E}^{\vd_j|\Yd_j}\Bigl|\vd_j-\mathbb{E}v_j\Bigr|^2\Bigr).$$
    But, since $v_j$ is distributed according to the 
conditional $\vd_j | \Yd_j,$ for frozen $\Yd_j$ we have
$$\mathbb{E}^{\vd_j|\Yd_j}\Bigl|\vd_j-\mathbb{E}v_j\Bigr|^2=\mathbb{E}\Bigl|v_j-\mathbb{E}v_j\Bigr|^2.$$
Thus the result follows.
\end{proof}

\begin{remark} The preceding statement in Theorem~\ref{t:ser}, and method of proof, can be generalized to the smoothing distribution $\mathbb{P}(\vd_j | \Yd)$ if $\Yd_j$ is replaced by $\Yd$.
\end{remark}

The result above motivates the use of the spread--error ratio\index{spread--error ratio},
which is defined next, to study probabilistic predictions.

\begin{definition} \label{def:spread-error}
Consider a sequence of probabilistic forecasts $\{\pia_j\}_{j=1}^J$, as given by Data Assumption~\ref{da:eval_prob}.
    The \emph{spread--error ratio}\index{spread--error ratio} of this
    sequence with respect to the true trajectory $\{\vd_j\}_{j=1}^J$ is the ratio
    \begin{equation}\label{eq:se_ratio}
        r := \frac{\frac{1}{J}\sum_{j=1}^J\mathbb{E}^{v_j\sim \pia_j}\bigl[|v_j - \mathbb{E}^{v_j\sim \pia_j}[v_j]|^2 \bigr]}{\frac{1}{J}\sum_{j=1}^J \bigl|\vd_j-\mathbb{E}^{v_j\sim \pia_j}[v_j] \bigr|^2}.
    \end{equation}
\end{definition}

\begin{remark} \label{rem:fr}
This terminology for the spread--error ratio\index{spread--error ratio} is used because
the denominator is the square \emph{error} of the mean under
the filtering distribution, with respect to the true signal underlying
the data; the numerator is
the variance under the filtering distribution (the \emph{spread}). The ratio in Theorem~\ref{t:ser} averages over the marginal on the data and over the joint distribution of the state and observation. The spread--error ratio, however, replaces these expectations by averaging over the observations  
along a single trajectory; this reflects the fact that only a single trajectory is available in practice.  Nonetheless, the ratio serves as a heuristic for evaluating an algorithm.
If $r < 1$, we say that the approximate filtering distributions $\{\pia_j\}_{j=1}^J$ are \emph{underdispersed}. If $r > 1$, we say that they are \emph{overdispersed}. Ratios other than $1$ can also be used, reflecting the heuristic derivation of $r$ as an approximation of the exact value $1$ that is obtained
in Theorem~\ref{t:ser}.
\end{remark}

\begin{remark}
    Another common metric for evaluating the spread--error relationship, which can also be motivated using Theorem~\ref{t:ser}, is the spread--error correlation\index{spread--error correlation}, computed as the sample Pearson correlation coefficient between the \emph{spreads} (the terms of the summation in the numerator of \eqref{eq:se_ratio}) and \emph{errors} (the terms of the summation in the denominator of \eqref{eq:se_ratio}). If the correlation coefficient is 1, we say that the approximate filtering distributions $\{\pia_j\}_{j=1}^J$ have a \emph{perfect spread--error relationship}.\index{spread--error relationship}
\end{remark}

\subsubsection{Rank Histograms}\index{rank histogram}

Another common methodology for evaluating approximate filters is the rank histogram. We motivate this methodology by the following theorem.

\begin{theorem}\label{th:rh}
    Let $f: \R^\du\to\R$ be a scalar function. Assume that pair $(\vd_j,\Yd_j)$ is distributed according
to the joint distribution under the dynamics/data model \eqref{eq:sdm}, \eqref{eq:dm}. Then, $\operatorname{rank}(\{f(v^{(n)}_j)\}_{n=1}^N, f(v^\dag_j))$, where $\{v^{(n)}_j\}_{n=1}^N$ are drawn from the filtering distribution $\mathbb{P}(v^\dag_j | Y^\dag_j)$, is uniformly distributed over the integers $[[0, N]]$.
\end{theorem}
\begin{proof}
    Replacing $u$ with $v_j$ and $y$ with $Y_j$, the proof follows directly from Theorem~\ref{th:sbc}.
\end{proof}

This theorem suggests the following methodology.

\begin{definition}\label{def:rh}
    Let $f: \R^\du\to\R$ be a scalar function. The \emph{rank histogram}\index{rank histogram} of a series of ensemble forecasts $\{\pia_j\}_{j=1}^J$, where $\pia_j = \sum_{n=1}^N \delta_{v^{(n)}}$, is the histogram of the rank values
    $\left\{\operatorname{rank}(\{f(v_j^{(n)})\}_{n=1}^N, f(\vd_j))\right\}_{j=1}^J$.
\end{definition}

By Theorem~\ref{th:rh}, the rank of the true trajectory with respect to samples from the true filtering distribution is uniformly distributed. Thus, deviation of the rank histogram from uniformity can indicate probabilistic miscalibration.

\section{Smoothing Algorithms}\index{smoothing}
\label{sec:smoothing}

In this section we describe a variety of smoothing algorithms.
Subsection~\ref{ssec:rtss} introduces the Rauch--Tung--Striebel smoother\index{smoother!Rauch--Tung--Striebel}, the solution to the smoothing problem in the linear and Gaussian setting. Subsection~\ref{ssec:4dvar} is devoted to the 4DVar\index{4DVar} approach
to smoothing, Subsection~\ref{ssec:4dvar_strong} discusses the strong constraint variant, Subsection~\ref{ssec:enks} discusses the ensemble Kalman smoother\index{smoother!ensemble Kalman}, and the subject of reanalysis\index{reanalysis} is overviewed in
Subsection~\ref{ssec:rean}.

\subsection{Rauch--Tung--Striebel Smoother}\index{smoother!Rauch--Tung--Striebel}\label{ssec:rtss}

As in Subsection~\ref{ssec:kalman},
here we consider the dynamics and observations models \eqref{eq:sdm} and \eqref{eq:dm} in which the dynamics and observation operator are linear, and generalize to make them non-autonomous, so that
\eqref{eq:rep}, repeated here for convenience, holds:
\begin{equation*}
\Psi(\cdot) \mapsto A_j \cdot, \quad h(\cdot) \mapsto H_j \cdot\,.
\end{equation*}
Then the marginals of the smoothing distribution, at each time $j$, are Gaussian: $\Prob(\vd_j|\Yd)=\cN(v_j^s, C_j^s)$. Since the $j$-marginals of the filter and smoother
agree at $j=J$ we also have $v_J^s=v_J$ and $C_J^s=C_J,$ where $v_J$ and $C_J$ are the end-time mean and covariance given by the Kalman filter\index{Kalman filter} from Subsection \ref{ssec:kalman}.

The \emph{Rauch--Tung--Striebel (RTS) smoother}\index{smoother!Rauch--Tung--Striebel} computes the means and covariances of these marginal smoothing distributions recursively by first carrying out a forward pass from $j=1$ to $j=J$ using the Kalman filter (Subsection~\ref{ssec:kalman}), and subsequently a backward pass to update the distribution at indices from $J-1$ down to $0$. The backward pass modifies the Kalman filter solutions at time $j$ by introducing the influence of future observations at times $j'>j$.

The backward pass starts at $j=J$ and uses the $\{v_j\}_{j=1}^J$, $\{\widehat{v}_j\}_{j=1}^J$, $\{C_j\}_{j=1}^J$, and $\{\widehat{C}_j\}_{j=1}^J$ computed by the Kalman filter\index{filter!Kalman} to compute the smoothed means and covariances, decreasing $j$ from $J-1$ to $0.$ The 
resulting algorithm has form
\begin{subequations}\label{eq:rtss}
    \begin{align}
        v_J^s &= v_J,\quad C_J^s = C_J,\\
        v_j^s &= v_j + G_j(v_{j+1}^s - \widehat{v}_{j+1}),\\
        C_j^s &= C_j + G_j(C_{j+1}^s - \widehat{C}_{j+1})G_j^\top,\\
        G_j &= C_j A_j^\top \widehat{C}_{j+1}^{-1}.
    \end{align}
\end{subequations}

\subsection{4DVar}\index{4DVar}\index{variational!4-dimensional|see{4DVar}}
\label{ssec:4dvar}

We introduce the 4DVar\index{4DVar} methodology, a nomenclature which
abbreviates the specific \emph{weak-constraint 4DVar}\index{4DVar!weak-constraint} formulation that we focus on in this subsection. In short,
this method computes a MAP estimator\index{MAP estimator} (see Definition \ref{def:map}) for the smoothing problem from Subsection~\ref{ssec:smoothing}.  
It is a state estimator.\index{state estimator}

To describe the methodology we recall the notation $\Vd  \in \R^{\du(J+1)}, \Vd_j  \in \R^{\du(j+1)}$ and $\Yd \in \R^{\dy J}$ from \eqref{eq:VdYd}.
Here $\Vd := \{\vd_0, \ldots, \vd_J\}$ is  the dummy variable used in defining our optimization
problem to estimate the state. We then set
\begin{subequations}
\label{eq:regS}
\begin{align}
\reg(\Vd) &:= \frac{1}{2}|\vd_0-m_0|^2_{C_0} + \frac{1}{2}  \sum_{j=0}^{J-1} |\vd_{j+1} - \Psi(\vd_j)|^2_{\Sigma}\\
&= \frac12|\Vd-\mathsf{\Psi}\bigl(\Vd_{J-1};m_0\bigr)|_{\Sigmas}^2,
\end{align}
\end{subequations} 
where the second expression involving $\mathsf{\Psi}$ is defined in the text preceding \eqref{eq:sprior}, defining the prior for the smoothing problem. We also define
\begin{subequations}
\label{eq:lossSA}
\begin{align}
\loss(\Vd;\Yd) &:=  \frac{1}{2} \displaystyle{\sum_{j=0}^{J-1}}|\yd_{j+1} - h(\vd_{j+1})|_{\Gamma}^2\\
&= \frac12|\Yd-\hs(\Vd)|_{\Gammas}^2,
\end{align}
\end{subequations}
where the second expression involving $\hs$ is defined in the text preceding \eqref{eq:likes}, defining the likelihood
for the smoothing problem. We may add the expressions in \eqref{eq:regS} and \eqref{eq:lossSA} to obtain
\begin{align}
\label{eq:objS}
\J(\Vd;\Yd)&  =\reg(\Vd)+\loss(\Vd;\Yd),
\end{align}
noting that $\exp\bigl(-\J(\Vd;\Yd)\bigr)$ is proportional to the posterior density
given in \eqref{eq:spost2}. The MAP estimator\index{MAP estimator} for this posterior is given by
\begin{equation} \label{eq:4DVarasMAP}
\Vmap \in \argmin_{\Vd \in \R^{\du(J+1)}} \J(\Vd;\Yd).
\end{equation}
We then take $\Vmap$ as our estimate of $\Vd.$ 
In the linear-Gaussian setting this recovers the mean of the RTS smoother\index{smoother!Rauch--Tung--Striebel}.

\begin{remark}
\label{eq:ws}
Minimization \eqref{eq:4DVarasMAP} is often implemented using Newton\index{Newton's method}
or Gauss--Newton\index{Gauss--Newton} methodologies (see Section \ref{sec:NGN}). Ensemble approximations of Gauss--Newton
are also used -- see Remark \ref{rem:ensg}. The problem  \eqref{eq:4DVarasMAP} is the weak-constraint 4DVar method.\index{4DVar!weak-constraint} The use of ``4D'' refers to the fact that, in weather
forecasting, the optimization problem is for a field in three physical
space dimensions and one time dimension; this should be contrasted with the reasoning for the nomenclature described in Remark \ref{rem:cycled} for 3DVar. As in  3DVar, the ``Var'' refers to the
optimization framing of the methodology. Weak constraint can be understood by comparing
with strong-constraint 4DVar\index{4DVar!strong-constraint} in the next subsection.

As with all MAP estimators, the result of applying 4DVar is simply a point estimator. Obtaining additional information about the posterior with density proportional to $\exp\bigl(-\J(\Vd;\Yd)\bigr)$  requires the use of methods such as Markov chain Monte Carlo (MCMC)\index{MCMC} and sequential Monte Carlo (SMC)\index{SMC}, or variational approximations; for the latter, see Section \ref{ssec:4DVARgoesGaussian}. Hybrid ensemble--4DVar methods are also used to give uncertainty estimates.
\end{remark}

\subsection{Strong-Constraint 4DVar}\index{4DVar!strong-constraint}
\label{ssec:4dvar_strong}

Strong-constraint 4DVar\index{4DVar!strong-constraint} is found by letting $\Sigma \to 0$ in \eqref{eq:4DVarasMAP}. This results in a minimization problem defined with respect to
variable $\vd_0 \in \R^d,$ the initial condition of the deterministic dynamics model
arising from \eqref{eq:sdm} by setting $\Sigma=0.$ We then obtain point estimator 
\begin{equation} \label{eq:4DVarasMAP0}
\vzmap \in \argmin_{\vd_0 \in \R^{\du}} \J_0(\vd_0;\Yd).
\end{equation}
Here
\begin{align}
\label{eq:objS0}
\J_0(\vd_0;\Yd)&  :=\reg_0(\vd_0)+\loss_0(\vd_0;\Yd),
\end{align}
where
\begin{equation}
\label{eq:regS0}
\reg_0(\vd_0) := \frac{1}{2}|\vd_0-m_0|^2_{C_0}
\end{equation}
and 
\begin{equation}
\label{eq:lossSA0}
\loss_0(\vd_0;\Yd) :=  \frac{1}{2} \displaystyle{\sum_{j=0}^{J-1}}\big|\yd_{j+1} - h\bigl(\Psi^{(j+1)}(\vd_0)\bigr)\big|_{\Gamma}^2. 
\end{equation}
In the preceding definition $\Psi^{(j)}$ is the $j$-fold composition of $\Psi$ with itself.
Strong-constraint 4DVar takes $v_0$ as point estimate of $\vd_0;$ it is a state 
estimator.\index{state estimator} Other techniques, such as MCMC\index{MCMC} or SMC\index{SMC} sampling algorithms, can be used for probabilistic estimation of the posterior on $\vd_0|\Yd,$ which has density proportional to $\exp\bigl(-\J_0(\vd_0;\Yd)\bigr).$

\begin{remark}
    The nomenclature \emph{strong constraint}, as opposed to \emph{weak constraint}, comes from the fact that the solution of \eqref{eq:4DVarasMAP0} can then be used to initialize the deterministic model, found from \eqref{eq:sdm} with $\Sigma=0$, leading to sequence $v_j=\Psi^{(j)}(v_0)$. Then sequence $v_j$ is said to be ``strongly constrained'' by the deterministic model. In contrast the trajectory in \eqref{eq:4DVarasMAP} is not constrained to satisfy the deterministic model. Strong-constraint 4DVar is appealing when the deterministic model conserves certain quantities, or has qualitative attributes, that the stochastic model does not share.
\end{remark}

\subsection{Ensemble Kalman Smoother (EnKS)}\label{ssec:enks}\index{smoother!ensemble Kalman}

The \emph{ensemble Kalman smoother (EnKS)} is an ensemble extension of the Rauch--Tung--Striebel  smoother\index{smoother!Rauch--Tung--Striebel} (RTS smoother, Subsection~\ref{ssec:rtss}), which can be applied to nonlinear and non-Gaussian settings. This is similar to how the ensemble Kalman filter (EnKF, Subsection~\ref{ssec:enkf})\index{Kalman filter!ensemble} extends the Kalman filter\index{Kalman filter} and can similarly be applied beyond the linear Gaussian setting. Here, the task is to obtain at each time $j$ an ensemble which approximates the marginal smoothing distribution; that is,
\begin{equation*}
    \Prob (\vd_{j}| \Yd) \approx \frac{1}{N} \sum_{n=1}^N \delta_{{v}_{j}^{(\sam),\text{s}}}(\vd_{j}),
\end{equation*}
where $v_{j}^{(\sam),\text{s}}$ denotes the $n$th smoothed ensemble member for time $j$. Furthermore, the sequence of smoothed states from time $j=0$ to $j=J$, for ensemble index $n$, is an approximate sample from the joint smoothing distribution; that is,
\begin{equation}\label{eq:enks_joint}
    \{v_{j}^{(\sam),\text{s}}\}_{j=0}^J\underset{\footnotesize\text{approx.}}{\sim} \Prob (\Vd| \Yd).
\end{equation}

As with the RTS smoother, the EnKS solution can be computed by carrying out a forward and subsequently a backward pass. The forward pass is defined by the EnKF: equation
\eqref{eq:predEnKF} defines ensembles $\{v^{(n)}_j\}_{n=1}^N$ and 
$\{\widehat{v}^{(n)}_j\}_{n=1}^N$, and equations \eqref{eq:gainEnKF}, \eqref{eq:mcenkf} define
the resulting ensemble means $(m_j, \widehat{m}_j)$ and covariances $(C_j,\widehat{C}_j)$, for $j=1,\ldots,J$. The EnKS computes the backward pass, running $j$ from $J-1$ down to $0$, through the iteration
\begin{subequations}
    \begin{align}
        v_J^{(\sam),\text{s}} &= v_J^{(n)},\quad \sam=1,\ldots,\Sam,\\
        v_j^{(\sam),\text{s}} &= v_j^{(n)} + G_j(v_{j+1}^{(\sam),\text{s}} - \widehat{v}_{j+1}^{(n)}),\quad \sam=1,\ldots,\Sam,\\
        G_j &= \widetilde{C}_{j} \widehat{C}_{j+1}^+,\\
        \widetilde{C}_{j} &= \frac{1}{N}\sum_{n=1}^N (v_j^{(n)} - m_j)\otimes(\widehat{v}_{j+1}^{(n)} - \widehat{m}_{j+1}).
    \end{align}
\end{subequations}
Note the similarity to the RTS smoother backward pass \eqref{eq:rtss}.


\subsection{Reanalysis}\index{reanalysis}
\label{ssec:rean}

\emph{Reanalysis}\index{reanalysis}, also known as retrospective analysis, is concerned with retrospectively producing an estimate of the trajectory of a system, or the conditional probability density function of the states given the observations. Since it is done retrospectively, it is offline, in the sense described in Remark~\ref{rem:fixed_interval_smoothers}, making this task amenable to smoothing. However, filtering algorithms are often used to produce reanalysis for reasons of computational cost.

An important feature of reanalysis is that it consolidates possibly sparse, irregular observations to give, at every time step, a complete state estimate of the system on the model state space, 
one which also incorporates information from the model dynamics. The extent to which the information from observed parts of the system can transfer to unobserved parts is formalized under the concept of \emph{observability}; see the bibliography Section \ref{sec:dabib} for references.\index{observability} 

Reanalysis datasets, due to the impact of the observations, are often assumed to 
exhibit lower effects from model error (see Section \ref{sec:meda}), in contrast
to the raw forecasts produced by the model. This is discussed in Subsection~\ref{ssec:analysis_increments}.
Both because the output of reanalysis is produced on a regular grid, and because it is thought to
exhibit lower model error, reanalyses are often used to train machine learning forecast models such as those described in Chapter~\ref{ch:ts_forecasting}.

\section{Model Error}\index{model error}
\label{sec:meda}

We now consider the \index{data assimilation} data assimilation problem in the setting where parts of the model are unknown, building on discussions
in Section~\ref{sec:ME} for the general inverse problem. To describe \index{model error!forecast}\emph{forecast model error}, we introduce parameter
$\vartheta$ that captures unknown aspects of the systematic part of the dynamical model.
The stochastic dynamics model\index{stochastic dynamics model} becomes
\begin{subequations}
\label{eq:sdm_err}
\begin{align}
\vd_{j+1} &= \Psi_\vartheta(\vd_j) + \xid_j, \: j \in \Z^+,\\
\vd_0 &\sim \Nc\bigl(m_0, C_0\bigr), \: \xid_j \sim \Nc\bigl(0, \Sigma\bigr) \: \index{i.i.d.}\text{i.i.d.} 
\end{align}
\end{subequations}
To describe \index{observation model error} \emph{observation model error} we introduce parameter
$\varphi$ which captures unknown aspects of the observation
operator.
The \index{data model} data model becomes
\begin{subequations}
\label{eq:dm_err}
\begin{align}
\yd_{j+1} &= h_\varphi(\vd_{j+1}) + \etad_{j+1} , \: j \in \Z^+,\\
\etad_j &\sim \Nc\bigl(0, \Gamma\bigr) \: \index{i.i.d.}\text{i.i.d.}
\end{align}
\end{subequations}
In the preceding we again invoke Assumption \ref{a:noise}.

\begin{remark}\label{rem:paramcov}
    In the following we work with unknown parameter $\theta=(\vartheta,\Sigma,\varphi,\Gamma)$,
    or choose $\theta$ to be a subset of these four parameters, as we recognize that
    this  will often be natural in applications. Furthermore, whilst we have general parameterization of the vector fields, the covariance matrices $\Sigma$ and
    $\Gamma$ may themselves be parameterized; for example we might write a covariance matrix as an
    unknown scalar non-negative real parameter multiplying the identity, or we might
    parameterize it in terms of a Cholesky factor.
\end{remark}

In the next two subsections we formulate
model error in the filtering and smoothing contexts.
Chapter \ref{ch:DAML} is devoted to detailed algorithms
concerned with learning model error.

\subsection{Model Error: Filtering}\index{model error}
\label{ssec:medaf}
It is possible to learn about model error using filtering. To illustrate
this idea we consider the setting in which $\Sigma$ and $\Gamma$ are known and the unknown $\theta$ comprises only
$\vartheta$ and $\varphi$ appearing in $\Psi_\vartheta$   and $h_\varphi$, respectively.  Consider the
following dynamical system, holding for all $j \in \Z^+:$
\begin{subequations}
\label{eq:sdm_err2}
\begin{align}
\vd_{j+1} &= \Psi_{\vartheta_j^\dagger}(\vd_j) + \xid_j,\\
\vartheta^\dagger_{j+1} &= \vartheta^\dagger_j,\\
\varphi^\dagger_{j+1} &= \varphi_j^\dagger,\\
\yd_{j+1} &= h_{\varphi_j^\dagger}(\vd_{j+1}) + \etad_{j+1}.
\end{align}
\end{subequations}
This can now be viewed as a filtering problem for 
$\{(\vd_j,\vartheta^\dagger_j,\varphi^\dagger_j)\}_{j \in \mathbb{Z}^+}$ given the data
generated by $\{\yd_j\}_{j \in \mathbb{N}}.$ 

\begin{remark}
As it stands it is slightly out of the scope
of problems studied in Section \ref{sec:filtering}, because no noise
appears in the evolution of the unknown parameters. However, as discussed in more detail in
Section \ref{sec:gen}, filtering can be extended to this setting. Alternatively
it is possible to replace equations (\ref{eq:sdm_err2}b, \ref{eq:sdm_err2}c)
by stochastic processes and to use an average over index $j$ of the filtering solution to provide a parameter estimate; in this context linear autoregressive models,\index{autoregressive!linear model} as developed in Section \ref{sec:autoregressive}, are natural replacements 
for (\ref{eq:sdm_err2}b, \ref{eq:sdm_err2}c).
\end{remark}

\subsection{Model Error: Smoothing}\index{model error}
\label{ssec:medas}

It is also possible to consider model error in the context of smoothing. 
To formulate learning of model error in the context of a smoothing problem we put a prior on $(\Vd,\theta)$, defined by the stochastic dynamics model \eqref{eq:sdm_err} 
for $\Vd|\theta$, and a prior on $\theta$; we then
condition on a likelihood defined by the data model. As in the preceding subsection, we consider the setting in which $\Sigma$ and $\Gamma$ are known and the unknown $\theta$ comprises only
$\vartheta$ and $\varphi$ appearing in $\Psi_\vartheta$ and $h_\varphi$. 
Building on Section \ref{sec:smoothing} we define prior
\begin{equation}
\label{eq:spriorME}
\rho(\Vd,\theta) \propto \exp\Bigl(-\frac12|\vd_0-m_0|_{C_0}^2-\frac12\sum_{j=0}^{J-1}|\vd_{j+1}-\Psi_\vartheta\bigl(\vd_j\bigr)|_{\Sigma}^2\Bigr)\rho_\theta(\theta),
\end{equation}
where we have written the prior on $\Vd|\theta$ 
and then multiplied by prior
$\rho_\theta(\theta)$ on $\theta.$
As in  Section \ref{sec:smoothing} we define
$$\etad := \{\etad_1, \ldots, \etad_J\},$$
noting that $\etad \sim \Nc(0, \Gammas)$, where $\Gammas$ is block diagonal with $\Gamma$ in 
each diagonal block. If we then define
$$\hs_\varphi(\Vd):=\bigl\{h_\varphi(\vd_1), \ldots, h_\varphi(\vd_J) \bigr\},$$
then the \index{data model}data model may be written as
$$\Yd=\hs_\varphi(\Vd)+\etad.$$
We are interested in finding $\Prob(\Vd,\theta|\Yd)=\Pi^{\Yd}(\Vd,\theta).$ We may apply Bayes Theorem \ref{t:bayes}
with prior $\rho$, noting that, under this prior, $\Vd \ind \etad,$ the standard setting for Bayesian inversion
from Chapter \ref{ch1}. We obtain
\begin{equation}
\label{eq:spost3}
\Pi^{\Yd}(\Vd,\theta) \propto \exp\Bigl(-\frac12|\Yd-\hs_\varphi(\Vd)|_{\Gammas}^2\Bigr)\rho(\Vd,\theta).
\end{equation}

\section{Surrogate Modeling}\label{sec:ml_approx_da}\index{surrogate}

As for inverse problems, one of the earliest, and still one of the most prevalent,
uses of machine learning in the field of data assimilation is
surrogate\index{surrogate} modeling.
This section, then, introduces the use of machine
learning in data assimilation, a topic we expand upon in the next four chapters.

We are interested in applications where evaluating $\Psi$ is computationally expensive, but where we have a surrogate\index{surrogate} model $\Psi'$ that approximates $\Psi$ and can be cheaply evaluated. We would like to use such a surrogate model to accelerate data assimilation, similar to the use of surrogate models in the general inverse problem setting in Section~\ref{sec:IPSU}. To obtain such a surrogate model we make the following assumption about available data:

\begin{dataassumption} \label{da:daaa}
    Assume that we have $\Vd = \{\vd_n\}_{n=0}^N$ generated from the dynamics model \eqref{eq:sdm}.
\end{dataassumption}

We use index $n$ within the data assumption, for consistency with the chapters on inverse
problems, and with Chapter~\ref{ch:ts_forecasting} on time-series forecasting.
Imagine a situation where $\Psi$ is not known, but a model of the form \eqref{eq:sdm}
is assumed to hold, for known $\Sigma.$ Given the preceding Data Assumption \ref{da:daaa},
we can then train $\Psi'$ as a supervised learning problem on pairs $\{(\vd_n, \vd_{n+1})\}_{n=0}^{N-1}$ using variants of the time-series forecasting methods discussed in Chapter~\ref{ch:ts_forecasting}.
The resulting trained model $\Psi'$ will then approximate $\Psi.$ 

\begin{remark}
Data Assumption \ref{da:daaa} may hold, for example, if $\Psi$ represents output of a differential
equation solve over a fixed time interval; $\Psi'$ will then represent a neural network surrogate,
which may be cheaper to evaluate. However, in many situations Data Assumption \ref{da:daaa} may not be
reasonable: we only have access to a trajectory through data assimilation.
For example it may be advantageous to take $\Vd$ to be data obtained by reanalysis\index{reanalysis}; see Subsection~\ref{ssec:rean}. Another alternative is to train a surrogate model using observations $\Yd$, employing the maximum likelihood methods described in Chapter~\ref{ch:DAML}.
\end{remark}

It is clearly of importance to understand how the use of $\Psi'$, in place of $\Psi$, affects the
filtering or smoothing distributions. This is the subject of the next two subsections.

\subsection{Filtering with Surrogate Dynamics}\index{surrogate}

Consider the filtering problem of estimating  signal 
$\{\vd_j\}_{j\geq 0}$ given observations $\{\yd_j\}_{j\geq 1}$ in the setting where the dynamics--observation models \eqref{eq:sdm}, \eqref{eq:dm} reduce to
\begin{subequations}
\label{eq:detlin}   
\begin{align}
    \vd_{j+1} &= \Psi(\vd_j),\\
    \yd_{j+1} &= H \vd_{j+1} + \etad_{j+1};
\end{align}
\end{subequations}
that is, we assume the dynamics are noiseless ($\Sigma = 0$) and that the observation map $h(\cdot) = H \cdot$ is linear. 
We make the same Assumption \ref{a:noise}, on the noise and initial condition, as detailed at the start of this chapter. Consider the 3DVar\index{3DVar} method \eqref{eq:pred3DVAR}, with $s=0$, 
applied in the setting \eqref{eq:detlin}; we obtain
$$v_{j+1} = (I - KH)\Psi(v_j) + K\yd_{j+1}.$$ 
The following theorem proves long-time accuracy for a 3DVar\index{3DVar!accuracy} filtering algorithm that uses the surrogate\index{surrogate} dynamics $\Psi'$ rather than the true dynamics $\Psi$, provided the surrogate model is sufficiently accurate and the observations sufficiently informative. Thus
we consider deploying the algorithm
\begin{equation}
\label{eq:3DVarS}
v_{j+1} = (I - KH)\Psi'(v_j) + K\yd_{j+1}. 
\end{equation}

\begin{theorem}
\label{t:ques}
Consider the 3DVar\index{3DVar} method \eqref{eq:3DVarS}, using approximate forecast model $\Psi',$
and driven by observations $\yd_{j+1}$ from \eqref{eq:detlin}.
Assume that $\Psi'$ is close to $\Psi$ in the sense that, for some
$\delta \in [0,\infty)$,
\begin{align*} 
\sup_{v \in \R^{\du}} \bigl|(I - KH)\bigl(\Psi(v) - \Psi'(v)\bigr)\bigr| &= \delta.
\end{align*}
Assume, further, the observability condition that there exists constant $\lambda \in (0,1)$ so that
\begin{align}\label{eq:Kalman_gain_conditions_ME}
    \sup_{v \in \R^{\du}} |(I - KH)D\Psi(v)| &\leq \lambda.
\end{align}
Denote $\epsilon := \mathbb{E}|K\etad_j|.$
Then, the 3DVar\index{3DVar} estimate $v_j$ based on \eqref{eq:3DVarS} satisfies
$$\limsup_{j \rightarrow \infty} \mathbb{E}|v_j - \vd_j| \le \frac{\epsilon + \delta}{1 - \lambda},$$
where $\vd_j$ is given by \eqref{eq:detlin}.
\end{theorem}

\begin{proof} The 3DVar update with the approximate forecast model\index{surrogate} from \eqref{eq:3DVarS} 
can be written as 
$$v_{j+1} = (I - KH)(\Psi(v_j) + b_j) + K\yd_{j+1},$$
where $b_j:=\Psi'(v_j)- \Psi(v_j)$ 
represents bias in the forecast model at time $j$. 
From \eqref{eq:detlin} we obtain
$$\yd_{j+1} = H\Psi(\vd_j)+\etad_{j+1}$$
so that
\begin{align*}
v_{j+1} - \vd_{j+1} &= (I - KH)\bigl(\Psi(v_j) + b_j\bigr) + K\bigl(H\Psi(\vd_{j}) + \etad_{j+1}\bigr) - \Psi(\vd_{j}) \\
&= (I - KH)(\Psi(v_j) - \Psi(\vd_{j})) + K\etad_{j+1} + (I - KH)b_j \\
&= \int_0^1 (I - KH)D\Psi(s v_j + (1 - s)\vd_{j})(v_j - \vd_{j}) \, ds + K\etad_{j+1} + (I - KH)b_j,
\end{align*}
where in the last line we used the mean value theorem for $\Psi$ on the line segment $s v_j + (1-s)\vd_{j}$ for $s \in [0,1]$. By the triangle inequality, the error is bounded by
\begin{align*}
|v_{j+1} - \vd_{j+1}| &\leq \Bigl(\int_0^1 |(I - KH)D\Psi(s v_j + (1 - s)\vd_{j})| \, ds\Bigr)|v_j - \vd_{j}| + |K\etad_{j+1}| + \delta\\
&\leq \Bigl(\int_0^1 \lambda \, ds\Bigr) |v_j - \vd_{j}| + |K\etad_{j+1}| + \delta.
\end{align*}
Taking an expectation over the measurement errors, we have
$$\mathbb{E}|v_{j+1} - \vd_{j+1}| \leq \lambda \mathbb{E}|v_{j} - v_{j}^\dagger| + \epsilon + \delta.$$
 Letting $e_j \coloneqq \mathbb{E}|v_{j} - \vd_{j}|$ and applying the discrete Gronwall inequality we have
$$e_{j} \leq \lambda^j e_0 + (\epsilon + \delta)\frac{1 - \lambda^j}{1 - \lambda}.$$
Noting that $\lambda \in (0,1)$ it follows that $\lambda^j \rightarrow 0$ as $j \rightarrow \infty$ and so the result follows.
\end{proof}

\begin{remark} Theorem~\ref{t:ques} highlights that running the 3DVar\index{3DVar} algorithm for long times can prevent small model error from accumulating. Moreover, this model error only needs to be controlled in the unobserved directions. That is, the overall error of the recovered state will be small if $(I-KH)b_j$ is small for all $j$, rather than the overall model error $b_j$ being small. Ideally, the gain\index{gain!3DVar} $K$ in 3DVar\index{3DVar} is chosen such that the conditions in the assumption of Theorem~\ref{t:ques} hold. See the bibliography Section \ref{sec:dabib} for one such guarantee under an observability condition.
\end{remark}

\begin{remark}
As discussed in the bibliography Section \ref{sec:dabib}, similar accuracy results can be established for other filtering algorithms.  These results  rely on   
observability conditions on the true dynamics and observation model $(\Psi, H)$ and on accuracy of the surrogate\index{surrogate} model $\Psi'$ in the unobserved part of the state-space. Notice that in Theorem \ref{t:ques} we do not assume long-time accuracy or stability of the surrogate\index{surrogate} dynamics defined by $\Psi'$, or of the true dynamics defined by $\Psi$, but we can nevertheless obtain long-time accuracy of filtering estimates by leveraging the 
stability properties of the map $(I - KH)\Psi$; this is a form of observability.
\end{remark}

\subsection{Smoothing with Surrogate Dynamics}\index{surrogate}
Recall definitions of $\Vd, \Yd$, and $\Yd_j$ from \eqref{eq:VdYd}.
From \eqref{eq:sprior} we obtain the prior on $\Vd$ from the dynamics model, 
a probability density function in $\R^{\du(J+1)}$ given by
\begin{equation}
\label{eq:sprior2}
\rho(\Vd) \propto \exp\Bigl(-\frac12|\vd_0-m_0|_{C_0}^2-\frac12\sum_{j=0}^{J-1}|\vd_{j+1}-\Psi\bigl(\vd_j\bigr)|_{\Sigma}^2\Bigr).
\end{equation}
Immediately after \eqref{eq:sprior} we define
\begin{align*}
\etad &= \{\etad_1, \ldots, \etad_J\} \in \R^{\dy J},\\
\hs(\Vd) &=\bigl\{h(\vd_1), \ldots, h(\vd_J) \bigr\},
\end{align*}
where $\etad \sim \Nc(0, \Gammas)$, and $\Gammas$ is block diagonal with $\Gamma$ in 
each diagonal block. We then obtain the posterior, an element of $\cP\bigl(\R^{\du(J+1)}\bigr)$,
\begin{equation}
\label{eq:spost22}
\Pi(\Vd) \propto \like(\Yd|\Vd)\rho(\Vd),
\end{equation}
where
\begin{equation}
    \label{eq:like22}
    \like(\Yd|\Vd) \propto \exp\Bigl(-\frac12|\Yd-\hs(\Vd)|_{\Gammas}^2\Bigr).
\end{equation}
(We drop the superscript $\Yd$ on the posterior as it is not central to
the following discussion.)
If we use a surrogate\index{surrogate} model $\Psi'$ to accelerate computations we will change the
prior dynamics model to have the form 
\begin{equation}
\label{eq:sprior222}
\rho'(\Vd) \propto \exp\Bigl(-\frac12|\vd_0-m_0|_{C_0}^2-\frac12\sum_{j=0}^{J-1}|\vd_{j+1}-\Psi'\bigl(\vd_j\bigr)|_{\Sigma}^2\Bigr).
\end{equation}
The resulting posterior is
\begin{equation}
\label{eq:spost222}
\Pi'(\Vd) \propto \like(\Yd|\Vd)\rho'(\Vd).
\end{equation}

We are interested in what effect the use of a surrogate\index{surrogate} model has on the posterior.
As a first step we simply assume that $\rho'$ is close to $\rho$ and ask what can be said about the closeness of $\Pi'$ and $\Pi.$ In the following theorem and proof, all integrals are over $\R^{\du(J+1)}$.

\begin{theorem}
Assume that $Z:=\int \like(\Yd|\Vd)\rho(\Vd) \, d\Vd>0.$ Then, there is a constant $C>0$ such that, for all
$\dtv(\rho,\rho')$ sufficiently small,
$$\dtv(\Pi,\Pi') \le C \dtv(\rho,\rho').$$
\end{theorem}

\begin{proof} In this proof $Z':=\int \like(\Yd|\Vd)\rho'(\Vd) \, d\Vd>0$ and $\like(\Vd)=\like(\Yd|\Vd).$
We have
\begin{align*}
2\dtv(\Pi,\Pi')&=\int \Bigl| \frac{1}{Z}\like(\Vd)\rho(\Vd)-\frac{1}{Z'}\like(\Vd)\rho'(\Vd)\Bigr| \, d\Vd\\
& \le I_1+I_2,\\
I_1 & := \int \Bigl| \frac{1}{Z}\like(\Vd)\rho(\Vd)-\frac{1}{Z}\like(\Vd)\rho'(\Vd)\Bigr| \, d\Vd,\\
I_2 & := \int \Bigl| \frac{1}{Z}\like(\Vd)\rho'(\Vd)-\frac{1}{Z'}\like(\Vd)\rho'(\Vd)\Bigr| \, d\Vd.
\end{align*}
Since $Z>0$ and $\like(\Vd)$ is uniformly bounded with respect to $\Vd$ we deduce that, for some $C_1>0,$
$$I_1 \le C_1 \dtv(\rho,\rho').$$
It also follows that, for some $C_2>0,$
$$|Z-Z'| \le \int \like(\Vd)|\rho(\Vd)-\rho'(\Vd)| \, d\Vd \le C_2 \dtv(\rho,\rho').$$
Hence, for all $\dtv(\rho,\rho')$ sufficiently small, $Z'>\frac12 Z>0.$ Finally
we deduce that, for $C_3=Z'>0$,
$$I_2 = \left|\frac{1}{Z} - \frac{1}{Z'}\right|Z \le C_3 |Z-Z'|.$$
The result follows.
\end{proof}

\begin{remark}
    The sense in which $\rho'$ is close to $\rho$ depends on details of surrogate\index{surrogate} models that we do not get into in these notes. In particular the approximation
    theorems we allude to in Section \ref{sec:AP} are valid, in their simplest form, 
    over compact sets $D$; in contrast, the prior distribution here is supported on the whole Euclidean space. However, by making assumptions about the behaviour of $\Psi$ and $\Psi'$ at infinity, which controls errors outside $D$, and by choosing $D$ large enough, it is possible to deduce that $\rho'$ is close to $\rho$.
\end{remark}

\section{Generalizations} \label{sec:gen}

    Here we conclude with some remarks on generalizations
    of the setting we adopt in this chapter. First we observe that the stochastic
    dynamics model \eqref{eq:sdm} and data model \eqref{eq:dm} are readily generalized
    to settings in which the maps $\Psi(\cdot)$ and $h(\cdot),$ as well as the covariances
    $\Sigma$ and $\Gamma$, are dependent on the time-index $j.$ It is also possible
    to allow degenerate noises (positive semidefinite covariances), 
    and the case where no noise is present in the
    dynamics ($\Sigma \equiv 0$) arises frequently. Furthermore, all the
    algorithms in this chapter, except for the Kalman filter\index{Kalman filter} which leverages Gaussian structure, 
    admit generalizations to settings in which the noises
    $\{\xid_j\}_{j \in \Z^+}, \{\etad_j\}_{j \in \N}$ are independent i.i.d. centered but \emph{non-Gaussian}\index{Gaussian!non-} sequences.
    Adding \emph{correlation}\index{correlation} to the noises is also possible.
    For example it is possible to consider correlation across the discrete time index $j$. 
    This significantly complicates filtering, but it is possible to accommodate it if
    the noise itself is generated by a Markov process.
    Adding correlation to the noises across the discrete time index $j$ 
    also complicates smoothing, but can also be handled, typically at additional
    computational cost. In addition, allowing for correlation between 
    the dynamics and observational noise is also possible. Finally we note
    that assuming differentiability, or even continuity, of the maps $\Psi(\cdot)$ and 
    $h(\cdot)$ is also not necessary; however
    assuming differentiability of $\Psi(\cdot)$ and $h(\cdot)$ facilitates specific
    filtering methods such as the ExKF\index{Kalman filter!extended} and gradient-based
    algorithms for smoothing such as 4DVar\index{4DVar}.

\section{Bibliography}
\label{sec:dabib}

For overviews of the subjects of data assimilation and filtering/smoothing, see the
textbooks \cite{jazwinski_stochastic_1970,law2015data,reich2015probabilistic,asch2016data,sarkka2013bayesian,bain2008fundamentals,crisan2011oxford,evensen2022data} and the review paper \cite{reich2019data}. Furthermore the
books \cite{kalnay2003atmospheric,majda2012filtering} and the review paper \cite{carrassi2018data}
comprise pedagogical introductions to \index{data assimilation}data assimilation in the context of weather forecasting, turbulence modeling, and geophysical sciences, respectively. The importance of data assimilation for numerical weather forecasting was articulated in \cite{panofsky_objective_1949}, then known as \emph{objective analysis}, and connected to sequential estimation theory in \cite{ghil_applications_1981}. The terminology of \emph{prediction} and \emph{analysis} to define
filtering was introduced in the weather forecasting community \cite{panofsky_objective_1949}.

The Kalman filter\index{Kalman filter} was introduced in \cite{kalman1960new}.
Discussion of the steady state covariance, and results concerning
convergence to the steady state, may be found in \cite{lancaster1995algebraic}.
The ideas of 3DVar\index{3DVar} for the solution of the analysis step were introduced in
the context of weather forecasting in \cite{lorenc1986analysis}.
Using the formula \eqref{eq:gain3} directly, rather than carrying out the minimization \eqref{eq:3dvar_opt} numerically, is known as \emph{optimal interpolation}\index{optimal interpolation} in the weather forecasting literature. Optimal interpolation\index{optimal interpolation} was proposed earlier by Eliassen (1954) and Gandin (1963); see \cite{bengtsson_dynamic_1981} for bibliographical notes.
For definition of the (cycled) 3DVar\index{3DVar!cycled} 
algorithm as employed here, see \cite{law2015data}.
For analysis of the (cycled) 3DVar algorithm see \cite{law2012analysis,moodey2013nonlinear}.
The ExKF\index{Kalman filter!extended}
\index{ExKF|see{Kalman filter, extended}}
is described and analyzed in \cite{jazwinski_stochastic_1970} and its use in numerical weather prediction is considered in \cite{ghil_applications_1981}. For theoretical results concerning Gaussian approximations for evolving probability distributions with small variance, see \cite{sanz2016gaussian}.

The EnKF\index{Kalman filter!ensemble} is introduced in \cite{evensen1994sequential}; see
\cite{asch2016data,evensen2022data,law2015data,reich2015probabilistic}
for more recent discussion. 
The methodology based around optimization has a probabilistic interpretation
in the Gaussian setting, and this can be used to justify the empirical\index{empirical}
approximation \eqref{eq:empEnKF} (and also Remark \ref{rem:glandmand}); in the Gaussian setting see
\cite{le2009large,mandel2011convergence,kwiatkowski2015convergence} and 
in the near-Gaussian setting see \cite{carrillo2022ensemble,calvello2024accuracy}. The paper \cite{sanz2024long} establishes long-time accuracy of EnKF, as a state estimator, for a wide class of partially observed chaotic dynamical systems; moreover, \cite{sanz2024long} also shows that long-time accuracy still holds when the true dynamics are replaced with a sufficiently accurate surrogate model.  

The impact of sampling error on EnKFs\index{Kalman filter!ensemble} in the case that the signal is a Gaussian process is addressed in \cite{sacher_sampling_2008}. Localization and inflation are reviewed in \cite{evensen2022data}. Insights into localization are given in \cite{furrer2007estimation,al-ghattas_non-asymptotic_2024,bickel2008regularized,al2023covariance,al2023optimal,al2024covariance,vishny_high-dimensional_2024}. Localization in the context of using the EnKF to solve inverse problems
is discussed in \cite{aanonsen2009ensemble}. For further background on the derivation of \eqref{eq:enkfexample}, see \cite[Page 158]{asch2016data} and also \cite{furrer2007estimation,al-ghattas_non-asymptotic_2024}.

There are many variants of the EnKF, including ensemble square-root filters that are more computationally suited for high-dimensional problems than the formulation presented in this chapter \cite{tippett2003ensemble,hunt_efficient_2007}. Since these variants avoid the construction of the covariance matrices, localization is often implemented in these filters through \emph{domain localization}, where the spatial domain is divided into multiple areas, and analyses are done locally in each. A popular EnKF variant that uses domain localization is the local ensemble transform Kalman filter (LETKF) \cite{hunt_efficient_2007}.

We note that besides the considerations about spurious correlations and forecast covariance rank discussed in the chapter, localization can also be interpreted in a dynamical systems context. For an EnKF\index{Kalman filter!ensemble}, one generally needs enough ensemble members to span the unstable--neutral subspace, corresponding to the number of non-negative Lyapunov exponents\index{Lyapunov exponents}, in order to prevent filter divergence.\index{filter divergence} This observation is supported by the fact that, in the case of linear dynamics, the forecast covariance matrix will collapse onto the unstable--neutral subspace; see the early work of \cite{trevisan_assimilation_2004} and the review of data assimilation for chaotic dynamics \cite{carrassi_data_2022} for references to these results. It has been observed, however, that dynamical systems such as the atmosphere are locally low dimensional \cite{patil_local_2001,oczkowski_mechanisms_2005}; that is, the dynamics within a spatial region may have a significantly lower dimension (as quantified by the dimension of the subspace spanned by the fastest growing modes) than that of the entire system, implying that filtering may be successful with a small ensemble when localization is applied.

For an introduction to the particle filter see \cite{doucet2001introduction}.
For the OPF see \cite{doucet2000sequential}. Details of the derivation of the OPF, using the same
notation that we have here, and in particular when the observation operator is linear, may be found in \cite{sanz-alonso_inverse_2023}.
Particle filters typically suffer from weight collapse in high-dimensional problems:
all weights become zero, except one \cite{bickel2008sharp,snyder2008obstacles}.

The Rauch--Tung--Striebel (RTS) smoother\index{smoother!Rauch--Tung--Striebel} was introduced in \cite{rauch1965maximum}. For a discussion of 4DVar, in the context of weather forecasting, see
\cite{fisher2001developments}. As with all MAP estimators, 
the result of applying 4DVar is simply a point
estimator. To obtain information about the posterior with density proportional
to $\exp\bigl(-\J(V)\bigr)$ requires the use of methods such as MCMC\index{MCMC} \cite{brooks2011handbook}, SMC\index{SMC} \cite{del2006sequential} or variational
methods\index{variational!formulation of Bayes Theorem} \cite{jordan1999introduction}. Hybrid methods that combine 4DVar with ensembles are also popular ways to estimate uncertainty for 4DVar \cite{carrassi2018data}.

The ensemble Kalman smoother (EnKS)\index{smoother!ensemble Kalman} was introduced in \cite{evensen2000ensemble}. The form of the EnKS that we discuss here is from \cite{raanes_ensemble_2016}; this work also discusses the relationship between the EnKS and the RTS smoother. See \cite{ramgraber_ensemble_2023} for a discussion of the EnKS and the ensemble RTS smoother in the more general framework of ensemble transport smoothers.

The concept of reanalysis originates in atmospheric science \cite{uppala2005era}. Although operational weather forecasting has been done for decades, there have been considerable changes in forecast models and data assimilation methods, necessitating reanalysis to have a consistent trajectory estimate over decadal timescales.
One of the first reanalysis datasets for the atmosphere was produced by the National Centers for Environmental Prediction (NCEP) and the National Center for Atmospheric Research (NCAR) \cite{kalnay_ncepncar_1996}.
The subject of observability is reviewed in \cite{maybeck_stochastic_1982}.

The spread--error relationship for ensemble forecasts is discussed in \cite{fortin_why_2014}. Rank histograms\index{rank histogram}, also called Talagrand diagrams, were introduced in the context of ensemble forecasts by several independent groups; see \cite{talagrand_evaluation_1997,hamill_interpretation_2001}. Theorem~\ref{th:rh} was stated in \cite{bach_learning_2026}.
Model error in data assimilation is often modeled by Gaussian noise, and encapsulated in the model noises $\xi^\dagger$. Deterministic formulations of model error are reviewed in \cite{carrassi_deterministic_2016}. The special case of model bias is considered in \cite{dee_data_1998}. Time-correlated model error is considered in \cite{amezcua_time-correlated_2018}. Methods for correcting model error are discussed in Chapter \ref{ch:DAML}, and a bibliography provided there.
While the effect of model bias on data assimilation is an open research topic, \cite{dee_data_1998} analyzed Kalman filtering in the presence of model bias. Under observability conditions,~\cite{maybeck_stochastic_1982} showed that the Kalman 
gain\index{gain!Kalman} $K$ can be designed to satisfy the condition in~\eqref{eq:Kalman_gain_conditions_ME} for stability of a data assimilation scheme with linear dynamics; 
this is also related to the design of Luenberger observers in control theory.

The study of surrogate\index{surrogate} modeling, for both filtering and
smoothing, is an area where machine learning has a simple and direct
potential for impact on data assimilation, and is an area where substantial
initial and continuing activity at the data assimilation and machine learning
interface is taking place. 
Machine learning models for numerical weather forecasting have recently received significant attention; see, e.g., \cite{pathak2022fourcastnet,Bi2023,keisler2022forecasting,chen2023fuxi,chen2023fengwu,lam2022graphcast}. 
Data assimilation using learned forecast models has been considered in \cite{hamilton_ensemble_2016,lguensat_analog_2017,chen_bamcafe_2021,chattopadhyay2022towards,penny_integrating_2022,xiao_fengwu-4dvar_2023,adrian2024data,kotsuki_integrating_2024,slivinski_assimilating_2025,tian_evaluating_2026}. Theorem \ref{t:ques} was also proved in \cite{adrian2024data}. Continuous data assimilation with surrogate models, and analysis of the amount of training data needed for accurate nudging, is explored in \cite{li_continuous_2026}. The importance of these forecast models correctly reproducing the Lyapunov spectrum and forecast error covariance was explored in \cite{penny_integrating_2022}. Including the Lyapunov spectrum and attractor dimension into the training process was explored in \cite{platt_constraining_2023}. Instead of learning a full forward model, learning an adjoint model for use in 4DVar\index{4DVar} was considered in \cite{hatfield_building_2021}.

Hybrid methods combining a numerical and learned forecast model in data assimilation and ensemble forecasting have been considered in \cite{bach_ensemble_2021,bach_improved_2024,chattopadhyay_deep_2023}. The latter used a large ensemble of a learned forecast model, along with a smaller ensemble of a more expensive numerical model solving the equations of motion, in an EnKF\index{Kalman filter!ensemble}, mitigating the need for localization.
Combining a small high-fidelity ensemble with a large ensemble of reduced-order models in an EnKF using control variates was considered in \cite{popov_multifidelity_2021,silva_adaptive_2025}. A multilevel EnKF was introduced in \cite{hoel_multilevel_2016}, and a multi-model EnKF in \cite{bach_multi-model_2023}.
The generation of ensembles from a machine-learned forecast model is considered in \cite{scher_ensemble_2021,li_seeds_2023,price_gencast_2023}.

\chapter{\Large{\sffamily{Variational Inference for Data Assimilation}}}
\label{ch:LG}

In this chapter we use the variational\index{variational!formulation of Bayes Theorem} formulation of Bayes Theorem\index{Bayes Theorem}, introduced in
Chapter \ref{ch:VI}, to frame data assimilation problems
as optimization\index{optimization} problems over probability measures; both smoothing\index{smoothing}
and filtering\index{filtering} problems are considered. In Section~\ref{sec:variational_smoothing0} we introduce the variational formulation of the smoothing problem, in analogy to the procedures in Chapter~\ref{ch:VI} for inverse problems; Section~\ref{sec:variational_filtering0} extends the variational formulation to filtering. 
In both of these first two sections we also introduce algorithmic frameworks\index{algorithm!variational}, stemming from the variational formulations.
In Section~\ref{sec:fss} we show how filtering may be linked to smoothing by imposing a filtering structure on a variational formulation of smoothing. Sections~\ref{ssec:4DVARgoesGaussian} and~\ref{sec:filteringuq} introduce examples of variational algorithms\index{algorithm!variational}, based on the formulations in Sections \ref{sec:variational_smoothing0} and
\ref{sec:variational_filtering0} respectively. We conclude the chapter with bibliographic remarks in Section~\ref{sec:VSFB}.

\begin{remark}
\label{rem:smooth-note}    
We are focused on the problem of finding certain distributions on
the state of the stochastic dynamical system \eqref{eq:sdm} conditioned on
data from \eqref{eq:dm}. Throughout we employ the notational conventions
established in Subsection \ref{ssec:note}, in particular those displayed in \eqref{eq:VdYd}
and the definition of $\Yd_0$ as the empty set $\varnothing.$
We write $\mathcal{P}:=\mathcal{P}(\R^d)$ for the space of probability
measures on $\R^d$; then, the probability distribution of the state of the system at any fixed time belongs to $\mathcal{P}.$
It is also convenient to denote by $\mathcal{P}^j:=\mathcal{P}(\R^{\du(j+1)})$ the space of probability measures on $\R^{\du(j+1)}$; then, the probability distribution of $\Vd_j$  belongs to $\mathcal{P}^j,$ and, in particular, the distribution of $\Vd$ belongs to $\mathcal{P}^J$.

Consider the smoothing problem defined in Subsection~\ref{ssec:smoothing}; in this context we will adapt the notation from Remark \ref{rem:minus1} to allow definition of marginal $\kappa$ on $\Yd$ determined by the joint $\gamma$ on $(\Yd,\Vd)$.
Recall, also, the filtering problem defined in Subsection~\ref{ssec:filtering},  and the notation $\hat{\post}_{j+1}(\vd_{j+1})=\Prob(\vd_{j+1}|\Yd_j)$ for the predictive distribution; here, and in other chapters, it will be useful to define
the joint distribution $\joint_{j+1}(\vd_{j+1},\yd_{j+1}) \coloneqq \Prob(\vd_{j+1},\yd_{j+1} | \Yd_j) = \like(\yd_{j+1}|\vd_{j+1})\widehat\pi_{j+1}(\vd_{j+1})$, on state and observation at 
time $j+1$, and its marginal $\kappa_{j+1}(\yd_{j+1})= \Prob(\yd_{j+1} | \Yd_j)$ on the observation coordinate.
\end{remark}

Throughout the chapter we make the following assumption 
about data available to define the filtering and smoothing problems variationally,
and for learning the parameters of models for filtering and smoothing through these variational
formulations.

\begin{dataassumption}\index{Data Assumption}
\label{da:9a}
The available data consists of $\Yd := \{\yd_1, \ldots, \yd_J\}$, a single draw from the marginal $\marg$
on the observed data.
\end{dataassumption}

\section{Variational Formulation of Smoothing}\index{variational!formulation of smoothing}
\label{sec:variational_smoothing0}

From \eqref{eq:spost2}, the density for the smoothing\index{smoothing} distribution $\mathbb{P}(\Vd | \Yd)$ is given by
\begin{align*}
    \Pi^{\Yd}(\Vd) &\propto \like(\Yd|\Vd)\rho(\Vd).
\end{align*}
Detailed formulae for the prior $\rho$ and likelihood $\like$ are given in Subsection \ref{ssec:smoothing}.
The following theorem shows that the smoothing density arises as the solution of a variational problem.
The following result is a direct application of the variational
formulation of Bayes Theorem\index{Bayes Theorem!variational}
in Theorem~\ref{t:VariationalForm_Bayes} to the smoothing problem
defined in~\eqref{eq:spost2}.

\begin{theorem}
    Consider the likelihood $\like(\Yd|\Vd)$ and prior $\rho(\Vd)$ given in~\eqref{eq:sdef}, and let  
\begin{subequations}
\label{eq:var_smoothing}
\begin{align} 
\J(q) &= \dkl(q\|\rho) - \mathbb{E}^q \bigl[\log \like(\Yd|\cdot) \bigr],\\
\qopt &\in \argmin_{q \in \cP^J} \J(q).
\end{align}
\end{subequations}
Then, the minimizer of $\J$ over $\mathcal{P}^J$ is the smoothing density: $\qopt=\Pi^{\Yd}.$
\end{theorem}

This formulation of the smoothing problem can be used as the basis for algorithms to learn approximations for the smoothing distribution from data: we replace $\cP^J$ in \eqref{eq:var_smoothing}
by some $\cQ^J \subset \cP^J$ and solve
\begin{subequations}
\label{eq:var_smoothinga}
\begin{align} 
\J(q) &= \dkl(q\|\rho) - \mathbb{E}^q \bigl[\log \like(\Yd|\cdot) \bigr],\\
\qs &\in \argmin_{q \in \cQ^J} \J(q).
\end{align}
\end{subequations}
In Section \ref{ssec:4DVARgoesGaussian} we give an explicit example of a choice of $\cQ^J$
that leads to actionable algorithms.

\begin{remark} \label{rem:amor1}
In this section we have focused on 
the idea of solving the smoothing problem for a given realization
$\Yd$ -- we work under Data Assumption \ref{da:9a}.
However, it is of interest to learn the dependence of the solution on the data instance $\Yd.$
This is known as \emph{amortization}\index{amortization} and is typically tackled by considering learning problems
defined over multiple realizations of the data $\Yd$; see Subsection~\ref{ssec:AVS} for the amortized version of the methodology introduced here.
\end{remark}

\section{Variational Formulation of Filtering}\index{variational!formulation of filtering}
\label{sec:variational_filtering0}

Recall the prediction ($\Pred$) and analysis ($\An_j:=\An(\cdot\,;\yd_{j+1})$) maps
that define the evolution of the filtering distribution in \eqref{eq:pna}, \eqref{eq:An}. The mapping from the filter at time $j$, $\post_j$, to the filter at time $j+1$, $\post_{j+1}$, is comprised of two steps. The forecast step 
$\widehat{\post}_{j+1}=\Pred \pi_j$ is defined by \eqref{eq:pred_operator}, and the analysis step, as detailed in \eqref{eq:filter_bayes_inference}, is an application of Bayes Theorem \ref{t:bayes} with likelihood $\mathbb{P}(y^\dagger_{j+1} | \vd_{j+1})$ and prior $\widehat{\post}_{j+1}$. The analysis density $\post_{j+1}$ can be written as the solution to an optimization problem using the variational formulation of Bayes Theorem \ref{t:VariationalForm_Bayes}:
\begin{subequations}
\label{eq:var_filtering}
\begin{align}
\J_{j+1}(q) &= \dkl(q\|\widehat{\post}_{j+1}) - \mathbb{E}^{q} \bigl[\log \mathbb{P}(y^\dagger_{j+1} | \cdot) \bigr],\\
\post_{j+1} &\in \argmin_{q \in \mathcal{P}} \J_{j+1}(q).
\end{align}
\end{subequations}
The preceding formulation assumes that we have access to the true prior $\widehat{\post}_{j+1}$. We now consider densities $q'_j$ recursively defined by the prediction--analysis cycle
\begin{subequations}\label{eq:online}
    \begin{align}
    q'_0 &= \post_0,\\
    \J_{j+1}(q) &= \dkl(q\|\Pred q'_j) - \mathbb{E}^{q} \bigl[\log \mathbb{P}(\yd_{j+1} | \cdot) \bigr],\label{eq:online_cost}\\
    q'_{j+1} &\in \argmin_{q\in\mathcal{P}} \J_{j+1}(q).\label{eq:theta_star2}
\end{align}
\end{subequations}
      
\begin{lemma}
Applying the iteration \eqref{eq:online} for $j=0, \ldots, J-1$ yields $q'_j=\pi_j$ for $j=0, \ldots, J.$ 
\end{lemma}

\begin{proof} The result holds for $j=0$ by (\ref{eq:online}a). The inductive step $j \mapsto j+1$ follows by
noting that, since solution of the optimization problem solves the Bayesian inverse problem that defines the analysis step
$\An$ defined in \eqref{eq:filter_bayes_inference}, we have
\begin{equation}
\label{eq:pran0}
    q_{j+1}' = \An(\Pred q_j'; \yd_{j+1}) = \An_j(\Pred q_j'),\quad q'_0 = \pi_0;
\end{equation}
this delivers the true filtering update equations \eqref{eq:pna}, \eqref{eq:An}.
\end{proof}

Performing the minimization \eqref{eq:theta_star2} over a 
restricted class $\cQ_{j+1}$ of probability densities, potentially defined
differently at each step $j$, leads to a sequence of approximate filtering densities. 
To provide detail on this idea we introduce a parameterized family of candidate approximate filters $q_{j+1}(\theta)$ to develop the basis of actionable algorithms, implicitly defining $\cQ_{j+1}.$ Assume that we have available to us the approximate filter at time $j$, $\qs_j \approx \pi_j.$ Consider a family of approximate filters at time $j+1$, $q_{j+1}(\theta),$ defined via
\begin{equation}
\label{eq:pran}
    q_{j+1}(\theta) = \An_{\theta}(\Pred \qs_j; \yd_{j+1}).
\end{equation}
Here $\An_{\theta}$ is a $\theta$-parameterized family of maps, and our goal is to choose $\theta \in \Theta \subseteq \R^p$ so that $q_{j+1}(\theta)$ approximates the filtering distribution $\pi_{j+1}$. Using the variational formulation of filtering in \eqref{eq:online}, with $q_j' \mapsto \qs_j$, we define the cost function (abusing notation to now view $\J_{j+1}$ as a function
of $\theta$)
\begin{equation} \label{eq:Jj}
    \J_{j+1}(\theta) = \dkl\bigl(q_{j+1}(\theta)\|\Pred \qs_j\bigr) - 
    \mathbb{E}^{{q_{j+1}(\theta)}}[\log \mathbb{P}(\yd_{j+1} | \cdot)].
\end{equation}
We may then consider the following
optimization problem to define the best choice of $\theta$:
\begin{equation*}
    \thetas_{j+1} \in \argmin_{\theta \in \Theta} \J_{j+1}(\theta).
\end{equation*}
Iterating on this idea leads to the algorithm:
\begin{subequations}
\label{eq:var_smoothinga99}
\begin{align} 
\qs_0 &=\pi_0,\\
\thetas_{j+1} &\in \argmin_{\theta \in \Theta} \J_{j+1}(\theta),\\
\qs_{j+1} & = \An_{\thetas_{j+1}}(\Pred \qs_j; \yd_{j+1}).
\end{align}
\end{subequations}

The following result establishes that the approximate filtering algorithm does indeed preserve filtering
structure.

\begin{lemma} \label{lem:willneed}
The approximate filtering algorithm defined by \eqref{eq:var_smoothinga99} has the
property that $\qs_j$ depends only on $\Yd_j$, and not on any other components of $\Yd.$
\end{lemma}

\begin{proof}
The result holds for $j=0$ because $\Yd_0=\varnothing$. Suppose that
$\qs_j$ depends only on $\Yd_j$. Then the objective
$\J_{j+1}$, and hence $\thetas_{j+1}$, depends only on $\Yd_j$ and
$\yd_{j+1}$. It follows from~\eqref{eq:var_smoothinga99} that
$\qs_{j+1}$ depends only on $\Yd_{j+1}$. The result follows by
induction.
\end{proof}

\begin{remark} \label{rem:amor2}
In Section \ref{sec:variational_smoothing0} we showed how to learn the solution of the
smoothing problem, under Data Assumption \ref{da:9a}, using variational methods.
In Remark \ref{rem:amor1} we point out that this methodology does not enable learning
of dependence, on the observed data, of the solution of the smoothing problem, because
only one instance of the observed data is provided by Data Assumption \ref{da:9a}. We draw
a sharp distinction between the variational approach described here, and the \emph{amortized}\index{variational inference!amortized} variational approach to smoothing that is considered 
(along with other amortized methodologies) in Chapter \ref{ch:LG10}: Subsection \ref{ssec:AVS} 
demonstrates how, when multiple realizations of the observation sequence are available,
we may learn dependence of the solution to the smoothing problem on that observation sequence.

The situation for \emph{filtering} is more complicated. In this chapter we work, as for
smoothing, under Data Assumption \ref{da:9a}. But in terms of filtering we can view this as providing
\emph{multiple} data instances, one at each filtering step $j \in \{1, \ldots, J\}.$ 
Recall that each update of the filtering distribution,
as defined in \eqref{eq:pna}, depends, through the analysis step $\An_j$, on an observation $\yd_{j+1}$.
This is also true of the approximate filter \eqref{eq:var_smoothinga99}. Indeed, the proposed form of models 
that we learn for filtering do learn
explicit dependence on the observations $\yd_{j+1}$ used to approximate each analysis step $\An_j$,
as is seen in \eqref{eq:var_smoothinga99}. We are thus using a form of amortization over time index $j$, 
even within this chapter, and under the same Data Assumption \ref{da:9a} used for smoothing.

It is arguable, however, that the parameter choices $\thetas_j$ are overfit to the specific data $\Yd_{j}$,
and especially to $\yd_j$, in the proposed methodology. One \emph{ad hoc} approach to ameliorate this
overfitting is to average the observed $\thetas_j$:
\begin{equation}
\label{eq:st1}    
\thetas  = \frac{1}{J} \sum_{j=1}^J  \thetas_j
\end{equation}
When the algorithm, coupled to the observation data
generation process, is ergodic (see Subsection \ref{ssec:ergf} for discussion of ergodicity for filtering 
algorithms) then this will lead, for large $J$, to a single parameter estimate for
$\theta$ which can, in principle, be used to define a fixed approximation to $\An:$
\begin{equation}
\label{eq:pran2}
    q_{j+1} = \An_{\thetas}(\Pred q_j; \yd_{j+1}),\quad q_0 = \pi_0.
\end{equation}
This approximate filter may then be used on new data sequences
$\Yd$, different from the specific sequence appearing in Data Assumption \ref{da:9a}
and used to determine $\thetas.$ The algorithm retains the desired filtering structure.

Although averaging $\thetas_j$ is a plausible approach to finding a single parameter in $\An_{\theta}(\cdot\,;\yd)$, 
the selected value of $\thetas$ is not derived directly from a single optimization principle.
A preferable approach is to find a single $\theta$ by solving the following averaged minimization problem
\begin{subequations}
\label{eq:var_smoothinga9}
\begin{align} 
\thetas &\in \argmin_{\theta \in \Theta} \frac{1}{J} \sum_{j=0}^{J-1} \J_{j+1}(\theta),\\
\J_{j+1}(\theta) &= \dkl\bigl(q_{j+1}(\theta)\|\Pred q_j(\theta)\bigr) - 
    \mathbb{E}^{{q_{j+1}(\theta)}}[\log \mathbb{P}(\yd_{j+1} | \cdot)],\\
    q_{j+1}(\theta) &= \An_{\theta}(\Pred q_j(\theta); \yd_{j+1}),\quad q_0 = \pi_0.
\end{align}
\end{subequations}
Note that for such a learning algorithm parameter
$\theta$ does depend on the training data in a manner that breaks the filtering structure, since it sees all the training
data from Data Assumption \ref{da:9a}. However the point of this approach is to learn a method that is usable
with new data. The algorithm \eqref{eq:pran2}, with $\thetas$ computed as in \eqref{eq:var_smoothinga9},
retains the filtering structure when applied to a new observation sequence $\Yd_j$, different from that used in training.

In Subsection~\ref{ssec:AVF} we go beyond amortization for filtering through multiple
time indices, and consider use of multiple observation \emph{sequences} $\Yd$ which can
enhance learning of the dependence of each analysis step $\An(\cdot\,;\yd)$ on $\yd$.
\end{remark}

\begin{remark} \label{rem:mklok}
To be able to employ algorithms based on the objective function \eqref{eq:Jj} it is necessary that
the KL divergence\index{divergence!Kullback--Leibler} between $q_{j+1}(\theta)$, given by \eqref{eq:pran}, 
and $\Pred q_j(\theta)$, is well-defined and that we are able to compute it.
In Section~\ref{sec:filteringuq} we give an example in which
$q_j(\theta)$, together with a Gaussian approximation
$\Pred_{\mathsf G}q_j(\theta)$ to its predictive distribution, are maintained in Gaussian form.
In Subsection \ref{ssec:ensemble_vi} we show how the ideas may be used for ensemble\index{ensemble} methods where the KL divergence is not defined; we extract Gaussian approximations from the empirical distributions defined by the ensemble method.
\end{remark}

\section{Imposing Filtering Structure on Smoothing}
\label{sec:fss}

An alternative way to use variational Bayes to derive filtering algorithms is
to start from the variational formulation of smoothing and then impose a temporal structure on the resulting density, to enforce filtering; specifically to impose that the probability density function for state $\vd_j$ at time $j$ only depends on past observations $\Yd_j$, as defined in \eqref{eq:VdYd}, and not future observations.
Any probability distribution  can be factorized autoregressively in the form
\begin{equation*}
    q(\Vd) = \mathfrak{q}_0(\vd_0)\prod_{i=1}^J \qf_i(\vd_i|\Vd_{i-1});
\end{equation*}
furthermore, if $q(\Vd)$ solves the smoothing\index{smoothing} problem then
\begin{equation*}
    q(\Vd) = \mathfrak{q}_0(\vd_0)\prod_{i=1}^J \qf_i(\vd_i|\Vd_{i-1}; \Yd).
\end{equation*}
Thus each component $\qf_i$ of the autoregressive factorization depends, in principle, on the entire sequence
of observed data, $\Yd.$ In contrast,
if $q \in \cP^J$ solves the filtering problem\index{filtering} then
\begin{equation}\label{eq:filter_density_factorized}
    q(\Vd) = \mathfrak{q}_0(\vd_0)\prod_{i=1}^J \qf_i(\vd_i|\Vd_{i-1};\Yd_i).
\end{equation}
Now each component $\qf_i$ of the autoregressive factorization depends only on the sequence
of observed data up to time $i$, $\Yd_i.$

In order to approximate a filtering distribution it is natural to seek minimizers of $\J(\cdot)$
defined by (\ref{eq:var_smoothing}a) which factorize in the form \eqref{eq:filter_density_factorized}.
For later use we also define $q_j \in \cP^j$ by
\begin{equation}\label{eq:factb}
    q_j(\Vd_j) = \mathfrak{q}_0(\vd_0)\prod_{i=1}^j \qf_i(\vd_i|\Vd_{i-1};\Yd_i).
\end{equation}
It is important to note the distinction between $\qf_j$ and $q_j$.
We let $\cC^J \subset \cP^J$ denote probability density functions of the form \eqref{eq:filter_density_factorized}.
The use of symbol $\cC$ is to invoke the conditional structure in terms of the dependence of $\qf_i$ on only $\Yd_i$ 
The smoothing distribution is, in general, contained in $\cP^J\backslash \cC^J$. This is because the conditional distribution of $\vd_i|V_{i-1}$ will, for smoothing, depend on $\Yd$ and not just on $\Yd_i.$

\begin{remark}
\label{rem:CJ}
Formally, we would like to minimize the objective $\J$ from
(\ref{eq:var_smoothing}a) over $\cC^J$. Under Data
Assumption~\ref{da:9a}, however, only one realization of $\Yd$ is
available. Direct optimization for this realization cannot ensure that
each fitted factor depends on the observations only through $\Yd_i$:
even if $\qf_i$ is written as a function of $\Yd_i$, its fitted
parameters may depend on the full sequence $\Yd$. The sequential
construction below circumvents this issue by ensuring that the
objective used to determine the factor at time $i$ only sees $\Yd_i$.
Alternatively, the dependence can be learned using amortization and
multiple observation sequences; see Subsection~\ref{ssec:AVF}.
\end{remark}

We now describe a way of circumventing the issue highlighted in
Remark \ref{rem:CJ}, different from the amortization\index{amortization} approach. The resulting methodology
involves minimization of a sequence of objective functions $\J_j(\cdot)$, each of which only sees $\Yd_j.$ The approach is based on the following theorem, which employs the notation in \eqref{eq:sdef}, \eqref{eq:filter_density_factorized}, \eqref{eq:factb} and in which, we reiterate, it is important to note the distinction between $\qf_j$ and $q_j$:

\begin{theorem}\label{theorem:filter_factorized}
Consider $q \in \cP^J$ factorized in the form \eqref{eq:filter_density_factorized}, \eqref{eq:factb}
so that $q \in \cC^J$. Define the objective function $\J': \cC^J \to \R$ by
\begin{subequations}
\label{eq:seqf}
\begin{align}
\J_j\Bigl(\qf_j(\cdot\,|\Vd_{j-1}; \Yd_{j})\Bigr)
&= \dkl\Bigl(\qf_j(\cdot|\Vd_{j-1}; \Yd_j) \|\mathsf{t}(\cdot,\vd_{j-1})\Bigr) - 
\mathbb{E}^{\qf_j} \bigl[\log \like_j \bigr],\\
\J'(q)
&= \dkl(\mathfrak{q}_0 \|\rho_0) + 
\sum_{j=1}^J\mathbb{E}^{\Vd_{j-1}\sim q_{j-1}}
\Bigl[ \J_j\Bigl(\qf_j(\cdot\,|\Vd_{j-1}; \Yd_{j})\Bigr) \Bigr],
\end{align}
\end{subequations}
where
$$
\like_j(\vd_j):=\mathbb{P}(\yd_j | \vd_j),
\qquad
\mathsf{t}(\vd_j,\vd_{j-1}):=\mathbb{P}(\vd_j|\vd_{j-1}), \qquad \rho_0 \coloneqq \Prob(\vd_0).
$$
Then the minimizers of $\J$  given by (\ref{eq:var_smoothing}a), and of $\J'$, both over $\cC^J$, agree.
   \end{theorem}
\begin{proof}
First we note that minimizing $\J(\cdot)$ is the same as minimizing
$\dkl(\cdot\|\Pi^{\Yd})$, as shown in the proof of
Theorem~\ref{t:VariationalForm_Bayes}.
We will use the $\ELBO$ introduced in Definition~\ref{d:elbo} and the
relationship between $\ELBO$, KL divergence, and likelihood established
in Theorem~\ref{th:EMlikecharacterization}.
To this end, we note that Definition~\ref{d:elbo} implies that
\begin{equation*}
\dkl(q\|\Pi^{\Yd})
=
\log\mathbb{P}(\Yd)
-
\ELBO\bigl(\mathbb{P}(\Vd,\Yd),q\bigr),
\end{equation*}
because $\mathbb{P}(\Yd)$ is the normalization constant that turns
$\mathbb{P}(\Vd,\Yd)$ into the conditional
$\mathbb{P}(\Vd|\Yd)$.
Since the first term does not depend on $q$, minimizing
$\dkl(q\|\Pi^{\Yd})$ over $q$ is equivalent to maximizing
\begin{equation*}
\ELBO\bigl(\mathbb{P}(\Vd,\Yd),q\bigr)
=
\mathbb{E}^{\Vd\sim q}
\left[
\log\frac{\mathbb{P}(\Vd,\Yd)}{q(\Vd)}
\right].
\end{equation*}
The joint density $\mathbb{P}(\Vd,\Yd)$ can be factorized
autoregressively as
\begin{align*}
\mathbb{P}(\Vd,\Yd)
&=
\rho_0(\vd_0)
\prod_{j=1}^J
\mathbb{P}(\vd_j,\yd_j|\Vd_{j-1},\Yd_{j-1})\\
&=
\rho_0(\vd_0)
\prod_{j=1}^J
\mathbb{P}(\yd_j|\Yd_{j-1},\Vd_j)
\mathbb{P}(\vd_j|\Yd_{j-1},\Vd_{j-1}).
\end{align*}
Using this, as well as the factorized structure
\eqref{eq:filter_density_factorized} of $q$ defining membership in
$\cC^J$, we obtain
\begin{align*}
\ELBO(\mathbb{P}(\Vd,\Yd),q)
&=
\mathbb{E}^q
\left[
\log\left(
\frac{\rho_0(\vd_0)}{\mathfrak{q}_0(\vd_0)}
\prod_{j=1}^J
\frac{
\mathbb{P}(\yd_j|\Yd_{j-1},\Vd_j)
\mathbb{P}(\vd_j|\Yd_{j-1},\Vd_{j-1})
}{
\qf_j(\vd_j|\Vd_{j-1};\Yd_j)
}
\right)
\right]\\
&=
-\dkl(\mathfrak{q}_0\|\rho_0)
+
\sum_{j=1}^J
\mathbb{E}^q
\left[
\log\left(
\frac{
\mathbb{P}(\yd_j|\Yd_{j-1},\Vd_j)
\mathbb{P}(\vd_j|\Yd_{j-1},\Vd_{j-1})
}{
\qf_j(\vd_j|\Vd_{j-1};\Yd_j)
}
\right)
\right].
\end{align*}
The $j$th summand depends only on $\Vd_j$, so its expectation
under $q$ may be replaced by expectation under $q_j$. Furthermore,
by~\eqref{eq:factb},
\[
q_j(\Vd_j)
=
q_{j-1}(\Vd_{j-1})
\qf_j(\vd_j|\Vd_{j-1};\Yd_j),
\]
and hence this expectation may be written as an iterated expectation
under $q_{j-1}$ and $\qf_j$. Thus,
\begin{align*}
\ELBO(\mathbb{P}(\Vd,\Yd),q)
&=
-\dkl(\mathfrak{q}_0\|\rho_0)
+
\sum_{j=1}^J
\mathbb{E}^{q_j}
\left[
\log\left(
\frac{
\mathbb{P}(\yd_j|\Yd_{j-1},\Vd_j)
\mathbb{P}(\vd_j|\Yd_{j-1},\Vd_{j-1})
}{
\qf_j(\vd_j|\Vd_{j-1};\Yd_j)
}
\right)
\right]\\
&\hspace{-3cm}
=
-\dkl(\mathfrak{q}_0\|\rho_0)
+
\sum_{j=1}^J
\mathbb{E}^{q_{j-1}}
\mathbb{E}^{\qf_j}
\left[
\log\left(
\frac{
\mathbb{P}(\yd_j|\Yd_{j-1},\Vd_j)
\mathbb{P}(\vd_j|\Yd_{j-1},\Vd_{j-1})
}{
\qf_j(\vd_j|\Vd_{j-1};\Yd_j)
}
\right)
\right]\\
&\hspace{-3cm}
=
-\dkl(\mathfrak{q}_0\|\rho_0)
-
\sum_{j=1}^J
\mathbb{E}^{\Vd_{j-1}\sim q_{j-1}}
\left[
\dkl\Bigl(
\qf_j(\vd_j|\Vd_{j-1};\Yd_j)
\|
\mathsf{t}(\vd_j,\vd_{j-1})
\Bigr)
-
\mathbb{E}^{\vd_j\sim\qf_j}
\bigl[\log\like_j(\vd_j)\bigr]
\right],
\end{align*}
where the last line follows from noticing that
$\mathbb{P}(\yd_j|\Yd_{j-1},\Vd_j)
=
\mathbb{P}(\yd_j|\vd_j)
=
\like_j(\vd_j)$
is the observation likelihood at time $j$, and
$\mathbb{P}(\vd_j|\Yd_{j-1},\Vd_{j-1})
=
\mathbb{P}(\vd_j|\vd_{j-1})
=
\mathsf{t}(\vd_j,\vd_{j-1})$.
The final display shows that
\[
\ELBO\bigl(\mathbb{P}(\Vd,\Yd),q\bigr)=-\J'(q).
\]
On the other hand, the definition of $\J$ and the identity
$\mathbb{P}(\Vd,\Yd)=\like(\Yd|\Vd)\rho(\Vd)$ give
\[
\ELBO\bigl(\mathbb{P}(\Vd,\Yd),q\bigr)=-\J(q).
\]
Consequently, $\J(q)=\J'(q)$ for every $q\in\cC^J$, and the result
follows.
\end{proof}

Methods that exploit this representation of the variational problem for filtering are
cited in the bibliography Section \ref{sec:VSFB}. In particular, the methods exploit the possibility of iteratively
optimizing the averaged $\J_j$, similarly to the methodology developed in the preceding section.

\section{Variational Algorithm\index{algorithm!smoothing} for Smoothing: Example}
\label{ssec:4DVARgoesGaussian}

We may apply the ideas of Chapter \ref{ch:VI} to the smoothing problem, working under
Data Assumption \ref{da:9a}, to develop algorithms. The purpose of this section is to
introduce an example of such an approach to smoothing, using variational inference.
To this end we recall \eqref{eq:var_smoothinga} and introduce a specific choice of $\cQ^J$,
the class of probability distributions within which we approximate the smoothing distribution.

    Recall the weak-constraint 4DVar\index{4DVar} solution \eqref{eq:4DVarasMAP} which is the MAP estimator for the smoothing problem,
    as introduced in Subsection \ref{ssec:4dvar}. Precompute this solution of an optimization problem, and call it $\Vs$. Now let 
    $$\cQ^J=\Bigl\{q \in \cP^J: q=\cN(\Vs,\Acs\Acs^\top) \Big|\; \Acs \in \R^{d(J+1) \times d(J+1)}, \,\, \det(\Acs)>0\Bigr\}.$$
    Thus in solving minimization problem \eqref{eq:var_smoothinga} with this
    choice of variational space we seek a Gaussian approximation of the
    smoothing problem, centered on a pre-computed weak-constraint 4DVar solution.
    In this formulation we thus seek to optimize over parameter $\Acs,$ 
    a square-root of the covariance, as in Example \ref{ex:cholGVI}; it may be natural
    to impose further triangular structure on $\Acs$ in order to obtain a Cholesky
    factorization. An amortized\index{amortization} version of the methodology from this section is described in Example \ref{ex:4DVARgoesGaussianAM}.

\section{Variational Algorithm\index{algorithm!filtering} for Filtering: Example}
\label{sec:filteringuq}

Here we give an example of the methodology introduced in Section \ref{sec:variational_filtering0}, 
centered around optimizing parameter $\theta$ appearing in \eqref{eq:pran}. 
We formulate the algorithm class around the idea of learning a gain matrix, $K$, that
defines a family of updates for the approximate filtering distributions $q_j(\theta);$
thus, in the notation of Section \ref{sec:variational_filtering0}, $\theta=K.$
Although we will ultimately work with a Gaussian approximation of the true filtering
distribution, we start by considering a more complicated approximate model for the evolution
of the filtering distribution; we then derive the Gaussian approximation from it.

Let $v_j$ denote a random vector whose law is intended to approximate
the filtering distribution $\pi_j$; in general, this law is non-Gaussian. 
We assume that $v_j$ is defined through the following 3DVar-like recursion:
\begin{subequations}\label{eq:3dvar_linear}
\begin{align}
    \widehat{v}_{j+1} &= \Psi(v_{j}) + \xi_j, \quad \xi_j \sim \mathcal{N}(0, \Sigma) \: \text{i.i.d.},\\
    v_{j+1} &= \widehat{v}_{j+1} + K\bigl(y^\dagger_{j+1}  - h(\widehat{v}_{j+1}) - \eta_{j+1}\bigr),
    \quad \eta_{j+1}\sim\mathcal{N}(0, \Gamma) \: \text{i.i.d.},\\
    v_0&\sim \mathcal{N}(m_0, C_0).
\end{align}
\end{subequations}
Now we postulate that the distribution of $v_j$ is approximately Gaussian. We
derive the Gaussian approximation by a linearization procedure; 
indeed the following Gaussian approximation
is exact if $\Psi$ and $h$ are linear. We seek
the filtering distribution in form $q_j(K) = \mathcal{N}(m_j, C_j)$,
and define the predictive distribution $\widehat{q}_{j+1}(K)$ by the Gaussian $\mathcal{N}(\widehat{m}_{j+1}, \widehat{C}_{j+1})$. A linearization procedure applied to the law of $v_j$
evolving according to \eqref{eq:3dvar_linear} then yields
the evolution $q_j(K) \mapsto q_{j+1}(K)$ as follows:
\begin{subequations}\label{eq:kalman_fixed_cov2}
    \begin{align}
        \widehat{m}_{j+1} &= \Psi(m_j),\quad \widehat{C}_{j + 1} = A_j C_{j} A_j^\top + \Sigma, \quad A_j = D\Psi(m_j)\label{eq:kalman_fixed_cov2_pred}\\
        m_{j+1} &= \widehat{m}_{j+1} + K(\yd_{j+1} - h(\widehat{m}_{j+1})),\label{eq:kalman_fixed_cov2_mean_ana}\\
                    C_{j+1} &= (I - KH_{j+1})\widehat{C}_{j+1}(I-KH_{j+1})^\top + K\Gamma K^\top, \quad H_j = D h(\widehat{m}_j). \label{eq:kalman_fixed_cov2_cov_ana}
\end{align}
\end{subequations}

The recursion resembles the extended Kalman filter\index{Kalman
filter!extended} from Subsection~\ref{ssec:exkf}, with two differences.
First, we learn the gain $K$ rather than compute it
using~\eqref{eq:exkf_gain}. Second, because $K$ is not necessarily the
Kalman gain, we use the covariance update in
\eqref{eq:kalman_fixed_cov2_cov_ana}; the source of this formula is
cited in the bibliography Section~\ref{sec:VSFB}.

We now generalize the optimization problems from Section \ref{sec:variational_filtering0}
to this Gaussian approximation.
 Map ${q}_{j}(K) \to \hat{q}_{j+1}(K)$ maps the Gaussian  manifold into a subset of
 itself and is defined by (\ref{eq:kalman_fixed_cov2}a)).
 This mapping is not equal to $\Pred$, but rather is a Gaussian approximation of it,
 which we denote by $\Pred_{\mathsf{G}}.$ Map $\Pred_{\mathsf{G}}$ is only defined
 on the Gaussian manifold; on that manifold it has the property that $\Pred_{\mathsf{G}}=\Pred$
 if $\Psi$ is linear. Map $\hat{q}_{j+1}(K) \mapsto {q}_{j+1}(K)$ also maps the Gaussian manifold into a subset of itself and is defined by (\ref{eq:kalman_fixed_cov2}b) and(\ref{eq:kalman_fixed_cov2}c); this is the map $\An_K(\cdot;\yd_{j+1})$. We may generalize \eqref{eq:Jj} to this 
 approximate Gaussian setting to obtain
\begin{equation*} 
    \J_{j+1}(K) = \dkl\bigl(q_{j+1}(K)\|\Pred_{\mathsf{G}} \, q_j(K)\bigr) - \mathbb{E}^{q_{j+1}(K)}[\log \mathbb{P}(\yd_{j+1} | \cdot)].
\end{equation*}
Dropping an additive constant independent of $K$, the resulting
optimization objective is
\begin{align}\label{eq:Jj_gaussian}
        \J_{j+1}(K) =& \dkl\bigl(\mathcal{N}(m_{j+1}, C_{j+1})\|\mathcal{N}(\widehat{m}_{j+1}, \widehat{C}_{j+1})\bigr)+\frac{1}{2}\mathbb{E}^{\mathcal{N}(m_{j+1}, C_{j+1})}
        |\yd_{j+1} - h(\cdot)|_{\Gamma}^2.
\end{align}
Recalling the formula (\ref{eq:kl_gaussians}a) for the KL divergence between two $d$-dimensional Gaussians we see the first term of \eqref{eq:Jj_gaussian} can be evaluated in closed form.
Evaluation of the second term, for nonlinear $h$, will require quadrature or sampling.

We can then consider the online learning problem \eqref{eq:var_smoothinga99}, in which we learn a time-dependent gain $K_j^\star$ at each analysis time:
\begin{equation*}
    K_j^\star \in \argmin_{K\in \mathbb{R}^{d\times k}} \J_j(K);
\end{equation*}
or the offline problem \eqref{eq:var_smoothinga9} in which a single fixed gain $K^\star$ is learned for all analysis times:
\begin{equation*}
    K^\star \in \argmin_{K\in \mathbb{R}^{d\times k}} \sum_{j=1}^J\J_j(K).
\end{equation*}

\section{Bibliography}
\label{sec:VSFB}

Several works have studied variational formulations of smoothing and filtering problems. 
The use of Gaussian approximations in filtering, including via variational Bayes, is studied in the papers \cite{csato2002data,vrettas2011estimating,archambeau2007variational}. Combinations with ensemble methods are discussed in \cite{vrettas2015variational}. The paper
\cite{galy2021flexible} employs mean-field\index{mean-field} models.
Variational filtering is discussed in \cite{marino_general_2018}, and the proof of Theorem~\ref{theorem:filter_factorized} is given there. Using variational methods in the context of sequential Monte Carlo\index{Monte Carlo!sequential} is
discussed in \cite{naesseth2018variational}. Learning parameterized filters using the variational objective introduced in Section \ref{sec:variational_filtering0} is considered in  \cite{luk_learning_2024}; in particular that
work introduces and deploys the learnable Gaussian filter described in
Section~\ref{sec:filteringuq}. An alternative variational objective for jointly learning a parameterized filter and the system dynamics is given in \cite{boudier2023data}. A variational formulation of the filtering and smoothing problems in continuous time, where the posterior is restricted to be in the exponential family\index{exponential family}, is provided in \cite{sutter_variational_2016}. The variational formulation of the optimal continuous-time filter (the Kushner--Stratonovich equation\index{Kushner--Stratonovich equation}) with no dynamics is derived in \cite{laugesen_poissons_2015}. The latter is formally combined with a variational formulation of the Fokker--Planck equation\index{Fokker--Planck equation} for gradient systems in \cite{lambert_variational_2023}, and it is shown that the solution coincides with the Kalman--Bucy filter\index{Kalman filter!Kalman--Bucy} in the linear case. The connection between 4DVar\index{4DVar} and variational inference is discussed in \cite{filoche_variational_2022}, and the idea of using variational inference and Gaussian approximations to extend 4DVar to posterior estimation is explored in \cite{filoche_variational_2025}.

\chapter{\Large{\sffamily{Learning the Prior for Data Assimilation}}}
\label{ch:DAML}
This chapter is devoted to learning priors for both smoothing and filtering problems, paralleling the material in
Chapter \ref{ch:priorIP} concerning the learning of priors for inverse problems.
In the smoothing\index{smoothing} problem, the prior is determined by the \index{stochastic dynamics model}stochastic dynamics model.
Thus, here, we will consider learning parameters defining the dynamics model, along with the state,\index{state} 
from observations\index{observations}. This is analogous to Section~\ref{sec:BH} in which we consider hierarchical modeling for
a Bayesian inverse problem:
the unknown parameter, and the parameters of the prior, are jointly learned from the single observation\index{observation} available.
In the filtering problem, there is a prior at each time point, found by predicting with the dynamics model,
starting from an approximation of the filtering distribution; this prior is
then combined with a likelihood to incorporate the next observation. In this chapter, we  consider 
learning priors for the filtering step, using Gaussian and particle-based empirical approximations. 
In Section~\ref{sec:pertp3} we showed how to represent the prior as a pushforward from a latent space.
In the final part of this chapter we consider data assimilation in a latent space, for both smoothing and filtering.

We introduce the problem of learning priors for smoothing and for filtering in Section \ref{sec:DAlearningmodels_setting}. 
The following two sections discuss general methodologies for solving this problem in the context of smoothing.
We develop two approaches both based on maximum likelihood estimation of the parameters of the prior dynamics model.
Section \ref{sec:EMmodel_dynamics} describes the expectation--maximization (EM) framework, developed
in a general setting in Chapter \ref{ch:optimization}, and applied here to estimate parameters entering the prior (dynamics)
model; as a by-product we obtain a method for joint parameter and state estimation.
In particular, we will present practical algorithmic instances of the EM framework to estimate parameters defining 
the dynamics map  and/or the error covariance.
Section~\ref{sec:143} discusses auto-differentiable Kalman filters, an approach to maximum likelihood estimation 
that can be much cheaper than the EM framework, but applies primarily
when a linear observation operator is used since it invokes a Gaussian approximation to estimate the state.
The method relies on Kalman-type algorithms to approximate the likelihood function, and on 
auto-differentiation to approximate its gradient.  In Section \ref{sec:POS} we discuss the learning of prior distributions 
for each step of the filtering problem, considering the use of empirical\index{empirical} priors and Gaussian priors 
constructed from the outcome of the preceding step of the filtering process. Section~\ref{sec:latent_DA} discusses 
the idea of working in a latent space for data assimilation algorithms.  Section \ref{sec:144} closes with 
extensions and bibliographical remarks.

\begin{remark} \label{rem:refback}
Learning priors for data assimilation is intimately related to the important problem of jointly estimating the \index{state}state and parameters of a stochastic dynamics model;  this problem is ubiquitous in data assimilation, and
arises in a variety of situations. 
For example, the model form may be known from first principles, but specific values of parameters may need to be learned from data.
Or the model form may not be known perfectly and machine learning may be used to find a correction of the model which enhances
its accuracy. Finally, when a model is expensive, it may be desirable to learn a cheap-to-evaluate machine-learned surrogate\index{surrogate}, as discussed in Section~\ref{sec:ml_approx_da}. Learning the model also arises when the dynamics of the system that generates the data are totally unknown
and a data-driven model is sought. A correction, a surrogate or a data-driven model may be represented by
a \index{neural network}neural network, \index{random features} random features, 
or \index{Gaussian!process}Gaussian process model, for example; see Chapter~\ref{ch:SL} for more detail on these types of
function approximation classes; see also Chapter~\ref{ch:ts_forecasting} for classes of functions designed for data-driven temporal forecasting.
\end{remark}

\section{Learning Priors for Smoothing and Filtering: Problem Setting}\label{sec:DAlearningmodels_setting} 
Consider the \index{stochastic dynamics model}stochastic dynamics and \index{data model}observation models given by
\begin{subequations}
   \label{eq:SSM1}  
\begin{align}
        \vd_{j+1} &= \Psi_\vartheta(\vd_j) + \xid_j, && \hspace{0.35cm}\xid_j \sim \Nc(0, \Sigma) \index{i.i.d.} \text{ i.i.d.}, \label{eq:SSM11} \\
        \yd_{j+1} &= h(\vd_{j+1}) + \etad_{j+1}, && \etad_{j+1} \sim \Nc(0, \Gamma) \index{i.i.d.}\text{ i.i.d.}, \label{eq:SSM12}
\end{align}
\end{subequations}
with $\vd_0 \sim \Nc(m_0,C_0)$ and the noise/initial data independence structure summarized in Assumption \ref{a:noise}.
Here, $\vartheta$ are unknown parameters entering the definition of the dynamics map $\Psi_\vartheta$ and we assume the error covariance $\Sigma$ to be also, in general, unknown.  
We denote by $\theta \coloneqq \{\vartheta,\Sigma\}$  the collection of unknown parameters; it is possible to extend to parameterizations of
the covariance $\Sigma$, rather than to learn the entire covariance matrix, as discussed in Remark \ref{rem:paramcov}, but we do not pursue this here. For a given and fixed integer $J,$ we recall the following notation, from Subsection \ref{ssec:note}, for the signal\index{signal}
and observations\index{observations}:
\begin{equation*}
\Vd := \{\vd_0, \ldots, \vd_J\}, \quad \quad  \Yd := \{\yd_1, \ldots, \yd_J\}.
\end{equation*}
Note that $\Vd \in \R^{(J+1)\du}$ and $\Yd \in \R^{J\dy}$. 
We make the following data assumption:
\begin{dataassumption}\index{Data Assumption}\label{da:truemodel}
Data $\Yd$ is given and assumed to come from the model \eqref{eq:SSM1}.
\end{dataassumption}

\subsection{Smoothing}
\label{ssec:smoot}

Given data $\Yd$ we wish to recover the parameter $\theta$ defining the
dynamics model. In Chapter~\ref{ch:SL} we study the supervised learning \index{supervised learning} 
problem of determining a map such as $\Psi_\vartheta$ from data of the form
$\bigl\{ u^{(n)}, \Psi_\vartheta(u^{(n)}) \bigr\}_{n=1}^N$
where the inputs $\{u^{(n)}\}_{n=1}^N$ are generated i.i.d. from an unknown probability density function; see Data Assumption \ref{data:sl}.
Here we only have access to observations $\Yd$ obtained from indirect and noisy measurement of an unknown trajectory $\Vd.$
We aim to choose the parameters of the dynamics model, such as $\vartheta$ and also $\Sigma$, which will explain an unknown trajectory $\Vd$ consistent, according to the observations model, with the observations $\Yd$ that we are given.
We are hence going beyond the standard supervised learning setting with i.i.d. input data, considered in Chapter \ref{ch:SL}.
This idea underlies the EM and auto-differentiable Kalman filtering algorithms introduced in the next two sections.

\subsection{Filtering}
\label{ssec:filt}

Recall that the filtering distribution cycles between predicting with the (stochastic) dynamics model \eqref{eq:sdm} and solving an inverse problem defined by the 
data model \eqref{eq:dm}. The prediction and
analysis steps of the filtering cycle are summarized in \eqref{eq:pna}. The analysis step
itself corresponds to solving the Bayesian inverse problem\index{inverse problem!Bayesian} for $\vd_{j+1}$, given $\yd_{j+1}$, where
\begin{equation} \label{eq:oba}
\yd_{j+1}=h(\vd_{j+1})+\etad_{j+1}.
\end{equation}
The prior on $\vd_{j+1}$ is $\widehat{\post}_{j+1}$ and the noise is $\etad_{j+1} \sim \Nc(0, \Gamma).$
Filtering is intrinsically harder than a standard Bayesian inverse problem, however, in the sense
that the prior $\widehat{\post}_{j+1}$ is typically not available exactly. For this reason, many algorithms used in practice employ either learned priors on the state to facilitate the analysis step, or learned joint distributions on the state and observation, also to facilitate the analysis step.

\subsection{Motivating Examples}

Before presenting the algorithms to learn priors for smoothing and for filtering, we consider three motivating examples,
building on the discussion in Remark \ref{rem:refback}.

\begin{example}[Parameterized Dynamics]
The state dynamics may be governed by the flow of a parameterized system of differential equations, and we may be interested in estimating a parameter $\vartheta$ entering the definition of the vector field in the differential equations from data $\Yd.$
\end{example}

\begin{example}[Model Correction]\label{ex:correction}
The dynamics may be unknown, but we may still have access to an inaccurate model $\Psi^{\text{approx}}$. Here $\vartheta$ can represent, for example, the parameters of a \index{neural network}neural network, \index{random features}random features, or \index{Gaussian!process}Gaussian process $\Psi^{\text{correction}}_\vartheta$ used to correct the inaccurate model $\Psi^{\text{approx}}$. The goal is to learn $\vartheta$ so that $\Psi_\vartheta \triangleq \Psi^{\text{approx}} + \Psi^{\text{correction}}_\vartheta$ approximates the true dynamics of the state. Learning model corrections is important in applications where available models have moderate predictive accuracy.  
For instance, fine scales of the \index{state}state may not be resolved accurately due to computational constraints, and we may be interested in learning a surrogate\index{surrogate} model from data which accounts for the unresolved scales of the system.
More generally, non-additive model corrections may also be learned.
\end{example}

\begin{example}[Fully Unknown Dynamics]
The state dynamics may be fully unknown and $\vartheta$ may represent the parameters of, for example, a \index{neural network}neural network, \index{random features} random features, 
or \index{Gaussian!process}Gaussian process surrogate\index{surrogate} model $\Psi_\vartheta;$ see Chapter~\ref{ch:SL} for background on these parameterizations of functions. The goal is to find an accurate surrogate\index{surrogate} model $\Psi_\vartheta$. Even if the true dynamics are  known, a \index{surrogate}surrogate model, or \index{emulator}\emph{emulator}, can be cheaper to evaluate, enabling the use of large sample sizes for particle filters or ensemble Kalman methods; see Section~\ref{sec:ml_approx_da}. Similar computational considerations motivate learning the forward map for inverse problems in
Section~\ref{sec:IPSU}. 
\end{example}

\section{Expectation Maximization} \label{sec:EMmodel_dynamics}
In this section we work under Data Assumption \ref{da:truemodel}, and so have access only to data $\Yd$ obtained from indirect and noisy measurement of a signal trajectory.
We view the problem of estimating $\theta \coloneqq \{\vartheta,\Sigma\} \in \Theta \subset \R^p$ from 
data $Y^\dagger$ as a \emph{missing data}\index{missing data} problem.
Ideally we estimate $\theta$ from both $\Vd$ and $\Yd$, but the data $\Vd$ is missing. So it is natural that
any algorithm used will also estimate $\Vd$ from $\Yd.$

We adopt a \index{maximum likelihood estimation}
\emph{maximum likelihood} approach to the problem of estimating $\theta$ from $\Yd$: we aim to maximize $\Prob(\Yd | \theta).$ The premise of the methodology we now introduce is that evaluation of the \index{likelihood}likelihood function $\Prob(\Yd | \theta)$ is challenging due to the missing, or  \index{latent variable}\emph{latent}, variable $\Vd$, but that the joint distribution $\Prob (\Yd, \Vd|\theta)$ 
is readily available. With this in mind we seek to address the problem of missing data by employing 
the identities
\begin{subequations} \label{eq:int_likelihood_filtering}
\begin{align}
\Prob(\Yd | \theta) &= \int_{\R^{(J+1)\du}} \Prob (\Yd, \Vd|\theta)\,  d\Vd ,\\
\Prob (\Yd, \Vd|\theta) &= \Prob (\Yd | \Vd, \theta) \Prob (\Vd | \theta).
\end{align}
\end{subequations}
  
\begin{remark} Likelihood functions that are computationally expensive to evaluate, such as the integral in~\eqref{eq:int_likelihood_filtering}, 
motivate the introduction of likelihood-free methods. These are discussed for Bayesian inference in Chapter~\ref{ch:data-dependence}.
The methods use only samples of the joint distribution on state and observation, rather than evaluations of $\Prob(\Yd | \theta)$; see Subsection~\ref{ssec:sjoint}. In this chapter, we will focus on optimization algorithms, rather than Bayesian methods, to find point estimators for the parameter $\theta$.
\end{remark}

Our working assumption, then, is that the likelihood $\Prob(\Yd | \theta)$ is intractable; 
but that it is expressible by marginalization of the tractable extended likelihood $\Prob (\Yd, \Vd|\theta)$ as 
in (\ref{eq:int_likelihood_filtering}a); and that, furthermore,
for the dynamics-observation model \eqref{eq:SSM1}, this extended likelihood is readily computed using conditioning as in (\ref{eq:int_likelihood_filtering}b).
In this setting the \index{maximum likelihood estimation}maximum likelihood estimator of $\theta$ can be computed by employing the \emph{expectation--maximization}\index{expectation maximization} (EM) algorithm\index{algorithm!EM} described in Section \ref{sec:142}. Specifically, in the notation of that section, we consider parameter $u:= \theta,$ missing variable $z:= \Vd$ and data $y:= \Yd.$ 

We employ the EM Algorithm~\ref{alg:EM},  using $\ell$ to index the algorithmic iterations because $j$, the notation used in Chapter \ref{ch:optimization}, is used here to index time in the dynamics-observation model \eqref{eq:SSM1}. The extended likelihood $\Prob (\Yd, \Vd|\theta)$
needed in order to implement the EM algorithm can be readily computed from the dynamics and observation models in~\eqref{eq:SSM1}, using
the identity (\ref{eq:int_likelihood_filtering}b):
\begin{align}\label{eq:joint}
\begin{split}
\log  \Prob(\Yd,\Vd| \theta)  &= \log\Prob (\Yd|\Vd,\theta)  + \log \Prob (\Vd|\theta) =  \log\Prob (\Yd|\Vd)  + \log \Prob (\Vd|\theta)\\
&= \mathsf{const}  -\frac{1}{2} \displaystyle{\sum_{j=0}^{J-1}} \bigl|\yd_{j+1} - h(\vd_{j+1}) \bigr|_{\Gamma}^2- \frac{J}{2}\log\det(\Gamma) \\
 &\qquad\qquad- \frac{1}{2} \sum_{j=0}^{J-1} \bigl|\vd_{j+1} - \Psi_\vartheta(\vd_j)\bigr|^2_{\Sigma}
  - \frac{J}{2}\log\det(\Sigma)
 -\frac{1}{2}|\vd_0-m_0|^2_{C_0},
\end{split}
\end{align}
where $\mathsf{const}$ is a constant independent of $\theta$.

We recall that the E-step in the EM Algorithm \ref{alg:EM}
integrates the \emph{full} log-likelihood $\log  \Prob(\Yd,\Vd| \theta)$ in~\eqref{eq:joint} with respect to $q^\ell(\Vd)=\Prob(\Vd|\Yd,\theta^\ell)$. 
Note that $q^\ell(\Vd)$ is the solution of the smoothing problem, with parameter of the stochastic dynamics set at $\theta^\ell.$
Then $\theta^{\ell+1}$ is computed  in the M-step by maximizing the expectation resulting from this integration, with respect to
$\theta:$
$$\theta^{\ell+1} \in \argmax_{\theta \in \Theta} \int_{\R^{(J+1)\du}} \log \Prob (\Yd , \Vd | \theta) q^\ell(\Vd) \, d\Vd.$$
The resulting maximization problem, dropping from the objective terms that do not depend on $\theta$, is as follows:
\begin{subequations}\label{eq:Erisk}
\begin{align}
\loss(\theta;q^\ell) &=   -\frac{J}{2} \log \det(\Sigma)  - \frac{1}{2} \int_{\R^{(J+1)\du}}  \sum_{j=0}^{J-1} \bigl|\vd_{j+1} - \Psi_\vartheta(\vd_j) \bigr|^2_{\Sigma}\, q^\ell(\Vd) \, d\Vd,  \\ 
\theta^{\ell+1} &\in \argmax_\theta \loss(\theta;q^\ell).
\end{align}
\end{subequations}
Noting that $q^\ell(\Vd)=\Prob(\Vd|\Yd,\theta^\ell)$ depends on $\theta^\ell$, we see that this algorithm defines
a mapping $\theta^\ell \mapsto \theta^{\ell+1}.$ The mapping is defined by solving the smoothing problem, with stochastic
dynamics parameter set at $\theta^\ell$ and then optimizing (\ref{eq:Erisk}a) to find $\theta^{\ell+1}.$

To illustrate the application of the \index{expectation maximization}EM algorithm, in the next Subsection~\ref{ssec:LearningCov} 
we discuss a simplified setting in which we learn only the model error covariance $\Sigma,$ assuming a known dynamics map $\Psi.$
This enables closed-form solution of \eqref{eq:Erisk}. Then, in  Subsection~\ref{ssec:EM_graddescent} we return to the problem of learning $\theta = \{\vartheta, \Sigma\}.$ Finally, in Subsection~\ref{ssec:analysis_increments} we consider the application of the EM algorithm in a model correction setting with known error covariance $\Sigma,$ and make a connection to supervised learning. 

\subsection{Learning Model Error Covariance} \label{ssec:LearningCov}

In this subsection we consider the setting where the dynamics map $\Psi$ is known, but the dynamics error covariance $\Sigma$ is unknown. That is, we have $\theta =\Sigma$. 
In this setting, we notice that the loss function~\eqref{eq:Erisk} in the M-step
has the same form as the objective in Proposition~\ref{prof:VI_Gaussians} for an 
unknown covariance in a Gaussian variational inference problem. Thus, the loss can be 
explicitly maximized over $\Sigma$. Setting $\theta^\ell =\Sigma^\ell$,
and using the explicit optimizer, we find the covariance matrix updates 
\begin{align} \label{eq:CovUpdatesEM}
\Sigma^{\ell+1} &= \frac{1}{J} \int_{\R^{(J+1)\du}} \sum_{j=0}^{J-1} \Big( \vd_{j+1} - \Psi(\vd_{j}) \Big) \otimes \Big( \vd_{j+1} - \Psi(\vd_{j}) \Big) \, q^\ell(\Vd) \, d\Vd;
\end{align}
recall that $q^\ell(\Vd)=\Prob(\Vd|\Yd,\Sigma^\ell)$ solves the smoothing problem,  
here with stochastic dynamics noise covariance $\Sigma^\ell.$ The following result shows that this update 
for the parameter $\Sigma$ leads to a monotonic increase of the log-likelihood for the observed data $\Yd$.

\begin{theorem}
Suppose that the parameter space for $\Sigma$ is the set of symmetric positive-definite matrices and that, for every $\ell\geq 0,$
\begin{equation}
\label{eq:EM_residual_covariance}
A^\ell
\coloneqq
\int_{\R^{(J+1)\du}}
\sum_{j=0}^{J-1}
\Bigl(\vd_{j+1}-\Psi(\vd_j)\Bigr)
\otimes
\Bigl(\vd_{j+1}-\Psi(\vd_j)\Bigr)
q^\ell(\Vd)\,d\Vd
\end{equation}
is positive definite. Then the M-step has the unique solution
\begin{equation}
\Sigma^{\ell+1}
=
\frac{1}{J}A^\ell,
\end{equation}
and the resulting EM iterates satisfy
$$
\log\Prob(\Yd|\Sigma^{\ell+1})
\geq
\log\Prob(\Yd|\Sigma^\ell).
$$
\end{theorem}
\begin{proof}
For fixed $q^\ell,$ the part of the M-step objective that depends on $\Sigma$ is
$$
\loss(\Sigma;q^\ell)
=
-\frac{J}{2}\log\det(\Sigma)
-\frac{1}{2}\Tr\bigl(\Sigma^{-1}A^\ell\bigr).
$$
Introducing the precision matrix $\Omega=\Sigma^{-1}$ gives
$$
\loss(\Omega;q^\ell)
=
\frac{J}{2}\log\det(\Omega)
-\frac{1}{2}\Tr\bigl(\Omega A^\ell\bigr).
$$
This function is strictly concave in $\Omega,$ and its first-order optimality condition is
$$
J\Omega^{-1}=A^\ell.
$$
Hence its unique maximizer is
$$
\Sigma^{\ell+1}
=
\Omega^{-1}
=
\frac{1}{J}A^\ell.
$$
The conclusion now follows from the monotonicity property of the EM algorithm in Theorem~\ref{th:EMmonotone}.
\end{proof}



\begin{remark} See the discussion in Remark \ref{rem:em} highlighting the fact that
additional assumptions are required for convergence of the EM iteration
to a local maximizer of the likelihood. Furthermore, in practice we will only have samples from an empirical\index{empirical} approximation of $q^\ell(\Vd)=\Prob(\Vd|\Yd,\Sigma^\ell)$ and so we may update the covariances using Monte Carlo approximations of the covariance matrices in~\eqref{eq:CovUpdatesEM}. With these Monte Carlo approximations the monotonic increase in the likelihood is no longer guaranteed.
\end{remark}

\subsection{Monte Carlo EM} \label{ssec:EM_graddescent}
Now we return to the general setting of finding parameter $\theta = \{\vartheta, \Sigma\}$. Given (approximate) samples $\{(\Vd)^{(n)}\}_{n=1}^N$ from the smoothing distribution $q^\ell(\Vd) = \Prob(\Vd|\Yd, \theta^\ell),$ we may use 
\index{Monte Carlo}Monte Carlo to approximate the expectation in the loss function \eqref{eq:Erisk}. We thus 
 obtain the empirical\index{empirical} approximation 
\begin{equation}\label{eq:Eempiricalrisk}
\loss^N(\theta;q^\ell)=  -\frac{J}{2} \log \det(\Sigma)  -  \frac{1}{2N} \sum_{n=1}^N \sum_{j=0}^{J-1} \bigl|(\vd_{j+1})^{(n)} - \Psi_\vartheta\bigl((\vd_j)^{(n)}\bigr) \bigr|^2_{\Sigma}.
\end{equation}
Note that the objective function depends on $\theta^\ell$ through the smoothing samples $\{(\Vd)^{(n)}\}_{n=1}^N.$
As in the preceding subsection, we notice that, given $\vartheta,$ the expression~\eqref{eq:Eempiricalrisk} can be explicitly maximized over $\Sigma.$ Indeed, the optimal solution is given by the covariance matrix 
\begin{equation}\label{eq:Eempiricalrisk2}
\Sigma^{\ell+1} = \frac{1}{NJ} \sum_{n=1}^N \sum_{j=0}^{J-1} \Big( (\vd_{j+1})^{(n)} - \Psi_{\vartheta} \bigl((\vd_{j})^{(n)}\bigr) \Big) \otimes \Big( (\vd_{j+1})^{(n)} - \Psi_{\vartheta}\bigl((\vd_{j})^{(n)}\bigr) \Big);
\end{equation}
this expression may also be viewed as a Monte Carlo approximation of
identity \eqref{eq:CovUpdatesEM}.
We note that this covariance depends on the parameters $\vartheta$ via
the model $\Psi_{\vartheta}.$ To maximize the loss $\loss(\theta;q^\ell)$ over $\theta=\{\vartheta,\Sigma\}$, we can employ iterative \index{optimization}optimization methods: alternate between optimization over $\Sigma$,
for fixed $\vartheta$, as given in the preceding subsection, and optimization
over $\vartheta$. Algorithm~\ref{alg:EM2} summarizes the resulting \index{expectation maximization}EM-type algorithm, for learning both \index{state}states and
parameters of a \index{dynamical system}dynamical system, using \index{Monte Carlo}Monte Carlo and \index{gradient ascent}gradient ascent updates and, in particular,
equations \eqref{eq:Eempiricalrisk} and \eqref{eq:Eempiricalrisk2}. 
We notice that this implementation of the EM framework treats the
parameters $\vartheta$ and $\Sigma$ differently in the M-step in order
to exploit the closed-form covariance update. Specifically, for each
fixed $\vartheta,$ we maximize the empirical M-step objective explicitly
over $\Sigma,$ obtaining $\Sigma^{\ell+1}(\vartheta).$ Substituting this
expression into the objective defines the profiled objective
$\J^N(\vartheta),$ which we then approximately maximize using gradient
ascent. Thus, the M-step is solved exactly with respect to $\Sigma$ for
each fixed $\vartheta,$ but only approximately with respect to
$\vartheta.$ Finally, the number of gradient ascent steps $\mathcal{I}$
may depend on $\ell,$ although it is fixed in
Algorithm~\ref{alg:EM2} for ease of presentation.

\begin{algorithm}
\caption{\label{alg:EM2} \index{expectation maximization}Expectation--Maximization with \index{Monte Carlo}Monte Carlo and \index{gradient ascent}Gradient Ascent}
\begin{algorithmic}[1]
\STATE {\bf Input}: Initialization $\theta^0 = \{\vartheta^0, \Sigma^0\}.$ Rule for \index{gradient ascent}gradient ascent step-sizes $\{\alpha^i\}_{i = 1}^\mathcal{I}$.
\STATE For $\ell = 0, 1, \ldots, L-1$ do the following expectation and maximization steps:

\STATE {\bf E-Step}: Obtain (approximate) samples  $\{(\Vd)^{(n)}\}_{n=1}^N$ from $\Prob(\Vd|\Yd, \theta^\ell)$. 

\STATE {\bf M-Step}: Using the samples obtained in the E-step, define
$$
\Sigma^{\ell+1}(\vartheta)
\triangleq
\frac{1}{NJ}
\sum_{n=1}^N\sum_{j=0}^{J-1}
\Bigl((\vd_{j+1})^{(n)}
-\Psi_\vartheta\bigl((\vd_j)^{(n)}\bigr)\Bigr)
\otimes
\Bigl((\vd_{j+1})^{(n)}
-\Psi_\vartheta\bigl((\vd_j)^{(n)}\bigr)\Bigr)
$$
and
$$
\J^N(\vartheta)
\triangleq
-\frac{J}{2}\log\det\bigl(\Sigma^{\ell+1}(\vartheta)\bigr)
-\frac{1}{2N}
\sum_{n=1}^N\sum_{j=0}^{J-1}
\Bigl|(\vd_{j+1})^{(n)}
-\Psi_\vartheta\bigl((\vd_j)^{(n)}\bigr)
\Bigr|_{\Sigma^{\ell+1}(\vartheta)}^2.
$$

\STATE Initialize $\vartheta^{\ell,0}=\vartheta^\ell.$
\FOR{$i=0,\dots,\mathcal{I}-1$}
\STATE $\vartheta^{\ell,i+1} = \vartheta^{\ell,i} + \alpha^i \nabla_{\vartheta} \J^N(\vartheta^{\ell,i}).$ 
\ENDFOR
\STATE Set $\vartheta^{\ell+1} = \vartheta^{\ell,\mathcal{I}}$ and $\Sigma^{\ell+1} = \Sigma^{\ell+1}(\vartheta^{\ell+1}).$
\STATE{\bf Output}: Approximation $\theta^L$ to the \index{maximum likelihood estimation}maximum likelihood estimate for $\theta$.
\end{algorithmic}
\end{algorithm}

\begin{remark} \label{rem:whereq2}
The smoothing distribution $\Prob(\Vd|\Yd,\theta^\ell)$ for the \index{signal}signal $\{\vd_j\}_{j=0}^{J}$ can be approximated by performing \index{inverse problem!Bayesian}Bayesian inversion or \index{data assimilation}data assimilation methods,
using the \index{stochastic dynamics model}stochastic dynamics and \index{data model}data models in~\eqref{eq:SSM1} with learned parameter $\vartheta^\ell$ and covariance model $\Sigma^\ell.$ These samples may be obtained, for instance, using Markov chain Monte Carlo methods. 
Alternatively, the ensemble Kalman smoother\index{smoother!ensemble Kalman} described in Subsection~\ref{ssec:enks} may be employed,
noting, however, that it is only guaranteed to
yield accurate posterior samples in approximately \index{linear--Gaussian setting}linear--Gaussian settings.
Note that approximate filters\index{filtering distribution} could also be used; however, the theory underpinning 
\index{expectation maximization}EM does require use of the smoothing\index{smoothing distribution} distribution. Nonetheless 
the computational expediency of filtering can be appealing and does allow for a biased approximation of the required missing data.
\end{remark}

\begin{remark} 
In analogy to the supervised learning task studied in Chapter \ref{ch:SL}, the objective function introduced in equation \eqref{eq:Erisk} can be conceptually interpreted as defining a risk\index{risk}, that is, an expected loss. Here, expectation is with respect to the 
smoothing distribution $\Prob(\Vd|\Yd,\theta^\ell)$ which represents our available knowledge of the latent variable $\Vd$ 
given data $\Yd$ and parameter estimate $\theta^\ell.$ Then, \eqref{eq:Eempiricalrisk} can be interpreted as an empirical\index{risk!empirical}\index{empirical} risk,  defined via approximate samples $(\Vd)^{(n)} \sim \Prob(\Vd|\Yd, \theta^\ell).$ 
\end{remark}

\begin{remark}\label{rem:grad_ascent}While we present Algorithm~\ref{alg:EM2} using gradient ascent, in practice we may employ other optimization methods such as accelerated first-order methods or second-order methods, like Gauss--Newton,\index{Gauss--Newton} to optimize the parameter $\vartheta$; we refer to Chapter~\ref{ch:optimization} for more details. The ideal choice of algorithm will depend on computational constraints and specific large- and fine-scale properties of the loss function.
\end{remark}

\subsection{Correcting Model Error Using Analysis Increments}\label{ssec:analysis_increments}
Here we assume the model error covariance $\Sigma$ to be known, and consider the problem of learning the parameter $\theta := \vartheta$  in the dynamics map model correction setting introduced in Example \ref{ex:correction}. That is, we consider a parameterization of the dynamics map of the form
\begin{equation}
\label{eq:pcorr}    
\Psi_\vartheta = \Psi^{\text{approx}} + \Psi_\vartheta^{\text{correction}},
\end{equation}
where $\Psi^{\text{approx}}$ is a known approximation to the dynamics map and we seek to find a
correction $\Psi_\vartheta^{\text{correction}}.$ In this setting, there is a natural connection between the EM framework and supervised learning\index{supervised learning}. Specifically, approximating $q^\ell(\Vd)$ in the E-step using trajectories  $\{(\Vd)^{(n)}\}_{n=1}^{N}$ computed with a data assimilation algorithm (filtering or smoothing, see Remark \ref{rem:whereq2}), we obtain an empirical approximation of the loss function in~\eqref{eq:Erisk} of the form 
\begin{equation} \label{eq:lossParameterCorrection}
\loss^N(\theta;q^\ell) = c - \frac{1}{2N}\sum_{n=1}^N\sum_{j=0}^{J-1} \Bigl|(\vd_{j+1})^{(n)} - \Psi^{\text{approx}}\bigl((\vd_j)^{(n)}\bigr) - \Psi_{\vartheta}^{\text{correction}}\bigl((\vd_j)^{(n)}\bigr) \Bigr|_\Sigma^2;
\end{equation}
here $c$ is a constant that is independent of the parameter $\theta = \vartheta$ to be estimated. 
We recognize that maximizing~\eqref{eq:lossParameterCorrection} over $\theta$ corresponds to the supervised learning problem of estimating the mapping from $\vd_{j}$ to the \emph{analysis increments} $\vd_{j+1} - \Psi^{\text{approx}}(\vd_j)$ given (approximate) analysis samples $(\Vd)^{(n)}$.

\begin{remark}
The Monte Carlo samples $\{(\Vd)^{(n)}\}_{n=1}^{N}$ of the time-series may have been created without the model error correction; this amounts to assuming $\Psi^{\text{approx}}$ is the true dynamics model for the purpose of the data assimilation, in which case the loss in
\eqref{eq:lossParameterCorrection} is simply $\loss^N(\theta)$ (no dependence on $q^\ell$) and EM is no longer needed. Alternatively, they
may have been created from the corrected model for a setting  $\vartheta^\ell$ that was found in an earlier iteration of the EM algorithm. In the latter case, the maximization
of $\loss^N(\cdot;q^\ell)$ yields $\theta^{\ell+1}.$
\end{remark}

\section{Auto-Differentiable Kalman Filters}
\label{sec:143}
Consider again the state--observation model defined by \eqref{eq:SSM1}, and make the additional assumption that
the observation operator is linear: $h(v)=Hv$ for some matrix $H \in \R^{\dy \times \du}$. We again consider the
problem of estimating the unknown parameters of the dynamics model, $\theta = \{\vartheta, \Sigma\}$, under Data Assumption \ref{da:truemodel},
as in the preceding section. But here we consider a different algorithmic framework, based on two ideas: firstly approximating the log-likelihood $\log \Prob(\Yd|\theta)$ using Kalman-based filters; and secondly computing its derivatives via automatic differentiation.
The goal is to perform (approximate) gradient ascent on the log-likelihood function; we refer to  Section \ref{sec:grad} for background on gradient descent (in this section \emph{ascent} as we seek to \emph{maximize} the likelihood) and to Section  \ref{sec:auto-differentiation} for background on automatic differentiation\index{auto-differentiation}. We will illustrate the idea of approximating the log-likelihood function using Kalman-based filters by focusing on the \index{Kalman filter!extended}extended Kalman filter (ExKF). However
the methodology directly generalizes to the \index{Kalman filter!ensemble}ensemble Kalman filter (EnKF) 
and other Kalman-based filtering algorithms -- see Remark \ref{rem:NVEX}. In Subsection~\ref{ssec:rexkf} we recall the ExKF; 
and in Subsection~\ref{ssec:autodexkf} we show how this filter may be used to execute approximate maximum likelihood estimation
for $\theta.$

\subsection{The Extended Kalman Filter}
\label{ssec:rexkf}

Recall from Subsection \ref{ssec:exkf} that the ExKF\index{Kalman filter!extended} is based on a \index{Gaussian!approximation}Gaussian approximation of the prediction\index{prediction} and analysis\index{analysis} distributions given by \eqref{eq:Gaussianpred99} and 
\eqref{eq:Gaussianan99}. Here we apply the same ExKF\index{Kalman filter!extended} methodology to the $\theta$-dependent
dynamics--observation model \eqref{eq:SSM1}, resulting in the approximations
\begin{equation}\label{eq:Gaussianpred}
\Prob (\vd_{j+1}| \Yd_j, \theta) \approx \Nc \bigl(\widehat{m}_{j+1}(\theta),\widehat{C}_{j+1}(\theta) \bigr)
\end{equation}
for the predictive distribution and
\begin{equation}\label{eq:Gaussianan}
\Prob (\vd_{j+1}| \Yd_{j+1}, \theta) \approx \Nc \bigl( m_{j+1}(\theta), C_{j+1}(\theta)\bigr)
\end{equation}
for the analysis distribution. Comparing with \eqref{eq:Gaussianpred99} and 
\eqref{eq:Gaussianan99}, the reader will note that we have (intentionally) switched to $m_j$ and
$\widehat{m}_{j+1}$ for the ExKF\index{Kalman filter!extended} mean and predictive mean, 
and we have made explicit the dependence of computed means and covariances on the parameter 
$\theta.$ We now describe the resulting algorithm.

We set
\begin{align}\label{eq:lin55}
    A_j(\theta) &:= D\Psi_\vartheta\bigl(m_{j}(\theta)\bigr).
\end{align}
The mean and covariance update according to the formulae
\begin{subequations}\label{eq:ekf_mean55}
\begin{align}
\widehat{m}_{j+1}(\theta) &= \Psi_\vartheta\bigl(m_j(\theta)\bigr),\\ 
m_{j+1}(\theta) &= \widehat{m}_{j+1}(\theta) + K_{j+1}(\theta) \bigl(\yd_{j+1}- H\widehat{m}_{j+1}(\theta)\bigr).
\end{align}
\end{subequations}
Here the gain $K_{j+1}(\theta)$ is computed as part of the computation of
the covariance evolution, as in \eqref{eq:wnrefd}:
\begin{subequations} \label{eq:ekf_cov55}
\begin{align}
\widehat{C}_{j+1}(\theta) &= A_j(\theta) C_{j}(\theta) A_j(\theta)^\top + \Sigma,\\
K_{j+1}(\theta) &= \widehat{C}_{j+1}(\theta) H^\top \bigl(H \widehat{C}_{j+1}(\theta) H^\top + \Gamma\bigr)^{-1},\\
C_{j+1}(\theta) &= \bigl(I - K_{j+1}H\big)\widehat{C}_{j+1}(\theta).
\end{align}
\end{subequations}

\subsection{Approximate Maximum Likelihood Estimation}
\label{ssec:autodexkf}

In this subsection we show how the \index{Gaussian!ansatz}Gaussian ansatz \eqref{eq:Gaussianpred}, and the formulae \eqref{eq:lin55}, \eqref{eq:ekf_mean55},
and \eqref{eq:ekf_cov55}, can be leveraged to produce an ExKF-based\index{Kalman filter!extended} approximation of the \index{likelihood}log-likelihood function $\log \Prob(\Yd|\theta)$. This leads to an algorithm for \index{maximum likelihood estimation}maximum likelihood estimation of parameters $\theta$ in the stochastic dynamics model. 
For the derivation presented below, we assume that the observation operator is linear, so that the predictive distribution of each observation is Gaussian whenever the predictive state distribution is Gaussian. Specifically,
the derivation rests on the following characterization of the \index{likelihood}log-likelihood function under a \index{Gaussian!ansatz}Gaussian ansatz.

\begin{theorem}
Assume that the observation
operator is linear: $h(\cdot)=H \cdot.$
Suppose that, for each $0 \le j \le J-1,$ the predictive distribution 
$\Prob (\vd_{j+1}| \Yd_j, \theta)$ of the stochastic dynamics and data models \eqref{eq:SSM1} is \index{Gaussian}Gaussian with mean $\widehat{m}_{j+1}(\theta)$ and covariance $\widehat{C}_{j+1}(\theta).$ Then the \index{likelihood}log-likelihood function admits the following characterization:
\begin{subequations}
\begin{align}
\label{eq:kalmanlikelihood}
\log \Prob (\Yd | \theta) =& \,\J(\theta), \quad S_{j+1}(\theta) = H \widehat{C}_{j+1}(\theta) H^\top + \Gamma,\\
\label{eq:thisJ}
    \J(\theta) \coloneqq & \mathsf{const} -\frac12 \sum_{j = 0 }^{J-1} \bigl|\yd_{j+1} - H\widehat{m}_{j+1}(\theta) \bigr|_{S_{j+1}(\theta)}^2 - \frac12 \sum_{j=0}^{J-1} \log \det \bigl(S_{j+1}(\theta)\bigr),
\end{align}
\end{subequations}
where $\mathsf{const}$ is independent of $\theta$.
\end{theorem}

\begin{proof}
We have  
\begin{align*}
\log \Prob(\Yd|\theta) &= \sum_{j=0}^{J-1} \log  \Prob(\yd_{j+1}|\Yd_{j},\theta),
\end{align*}
where we use the convention that  $\Yd_0 := \emptyset$
so that conditioning on $\Yd_0$ does not provide any information. 
Now conditioning in the \index{data model}data model
$$\yd_{j+1}  = H \vd_{j+1} + \etad_{j+1},$$
we see that
\begin{align*}
\Expect \bigl[\yd_{j+1} |\Yd_j,\theta \bigr] &= \Expect \bigl[H \vd_{j+1} + \etad_{j+1} |\Yd_j,\theta\bigr] = H \widehat{m}_{j+1}(\theta), \\
\text{Cov}\bigl[\yd_{j+1} | \Yd_j,\theta \bigr] & = \text{Cov} \bigl[ H \vd_{j+1} + \etad_{j+1}|\Yd_j,\theta \bigr] = H \widehat{C}_{j+1}(\theta) H^\top + \Gamma.
\end{align*}
Moreover, $\Prob(\yd_{j+1} |\Yd_j,\theta)$ is \index{Gaussian}Gaussian. The result follows. 
\end{proof}

Thus, for any value of $\theta$, we may obtain an \emph{approximation} to the log-likelihood $\log \Prob(\Yd | \theta)$ by running an
extended Kalman filter, obtaining predictive means and covariances $\bigl(\widehat{m}_{j+1}(\theta),\widehat{C}_{j+1}(\theta)\bigr),$ for $0 \le j \le J-1$, and using \eqref{eq:kalmanlikelihood}. Moreover, an approximation of the log-likelihood gradient at $\theta$ can be obtained 
by auto-differentiating\index{auto-differentiation} through our extended Kalman-based likelihood approximation. Auto-differentiable Kalman filters use these estimates of the log-likelihood gradients to conduct gradient ascent. The procedure is summarized in Algorithm \ref{alg:autodiff}.

\index{Kalman filter!ensemble}\index{EnKF|see{Kalman filter, ensemble}}

\begin{algorithm}
\caption{\label{alg:autodiff} Auto-Differentiable Kalman Filter}
\begin{algorithmic}[1]
\STATE {\bf Input}:  Initialization $\theta^0,$ rule to choose step-sizes $\{\alpha^\ell\}_{\ell = 0}^{L-1}.$
\STATE For $\ell = 0, 1, \ldots, L-1$ do the following Kalman \index{filtering}filtering and \index{gradient ascent}gradient ascent steps:
\STATE {\bf Kalman Filtering}: Run the extended Kalman \index{filtering}filtering algorithm defined in Subsection \ref{ssec:rexkf} to obtain predictive means and covariances $\widehat{m}_{j+1}(\theta^\ell),\widehat{C}_{j+1}(\theta^\ell),$ for $0 \le j \le J-1.$ 
\STATE {\bf \index{gradient ascent}Gradient Ascent:} \index{auto-differentiation}Auto-differentiate the map $\theta \mapsto \J(\theta)$ defined
in \eqref{eq:thisJ} to obtain a gradient estimate $\nabla \J(\theta^\ell).$  Set 
	\begin{align}
	\theta^{\ell + 1} = \theta^\ell + \alpha^\ell \nabla \J(\theta^\ell).
	\end{align}
\STATE{\bf Output}: Approximation $\theta^L$ to the \index{maximum likelihood estimation}maximum likelihood estimate for $\theta$. 
\end{algorithmic}
\end{algorithm}
\FloatBarrier

\begin{remark} \label{rem:NVEX}
    The algorithm may be generalized from the ExKF\index{Kalman filter!extended}\index{ExKF|see{Kalman filter, extended}} to the EnKF\index{Kalman filter!ensemble}\index{EnKF|see{Kalman filter, ensemble}}, using (\ref{eq:gainEnKF}a) and \eqref{eq:lh} for the predicted mean and covariance of the EnKF.\index{Kalman filter!ensemble}
If desired, at the conclusion of the algorithm, the \index{signal}signal $\{\vd_j\}$ can be estimated by performing \index{data assimilation}data assimilation (ExKF or EnKF) with \index{stochastic dynamics model}the stochastic dynamics and \index{data model}data models in \eqref{eq:SSM1} with learned parameters $\theta^L.$ As with Remark~\ref{rem:grad_ascent}, although we present Algorithm \ref{alg:autodiff} using gradient ascent, other optimization methods may be used instead.
\end{remark}

\begin{remark}
Notice that in the EM Algorithm \ref{alg:EM2} each posterior sample $\{(\Vd)^{(n)} \}_{n=1}^N$ is used to perform $\mathcal{I}$ gradient ascent 
steps in the M-Step, with the goal of maximizing the lower bound $\mathcal{L}(q^\ell, \theta).$  In contrast, in the auto-differentiable Kalman filter Algorithm \ref{alg:autodiff}, each run of a Kalman filtering algorithm is used to produce an estimate of the log-likelihood gradient, and a gradient ascent step is taken. 
This is possible because we can analytically integrate out the state variable,
given a linear observation operator and Gaussian predictive distribution.
\end{remark}

\section{Learning Priors for the Filtering\index{filtering} Step from Samples}
\label{sec:POS}

The preceding two sections have focused on learning the prior
for the smoothing\index{smoothing} problem,  namely the stochastic dynamics model, 
in the setting described in Subsection~\ref{ssec:smoot}. In this section we study
learning the prior, and generalizations, for the filtering\index{filtering} problem,
in the setting described in Subsection \ref{ssec:filt}. The previous sections also focus on 
learning a fixed set of parameters defining the dynamics model, while here we discuss learning a prior distribution much more generally
and at every analysis time.

We start by describing how particle-based filtering methods can be seen as learning a prior for the analysis\index{analysis} step in the filtering cycle, in both
Subsections \ref{ssec:bpf55} and \ref{ssec:enkf55}. These methods are all based on
learning, from samples, either a non-parametric\index{non-parametric} or 
parametric\index{parametric} distribution for the prior used in the analysis step; 
the resulting prior is either an empirical measure (non-parametric) or
a Gaussian (parametric) measure and in either case is found from a form of density estimation\index{density!estimation} (see Subsection~\ref{sec:den}). We also show, in Subsection~\ref{ssec:enkf55}, that certain
algorithms are based not on learning the prior, but on learning the joint distribution
on the state--observation pair.

\subsection{Bootstrap Particle Filter} \label{ssec:bpf55}
For particle-based methods the prior, or the joint distribution on state and observation,  must be reconstructed from the predicted ensemble of particles $\{\widehat v_{j+1}^{(n)}\}_{n=1}^N.$
For the bootstrap particle filter\index{particle filter!bootstrap} we reconstruct a particle approximation to $\widehat{\post}_{j+1}$, namely
\begin{equation}  
\hapost_{j+1} = \frac{1}{N}\sum_{n=1}^N \delta_{\widehat v_{j+1}^{(n)}}.
\end{equation}
This is an elementary form of learning the prior from data, as explained in Chapter \ref{ch:UL},
and equation \eqref{eq:emp} in particular.
The Bayesian inverse problem defined by \eqref{eq:oba} may be solved exactly with this prior, to obtain
the posterior
\begin{equation}
\label{eq:pap33}    
\apost_{j+1} = \sum_{n=1}^N w_{j+1}^{(n)}\delta_{\widehat v_{j+1}^{(n)}}.
\end{equation}
The weights, defined by the likelihood from \eqref{eq:oba}, are given in \eqref{eq:bpf2}.
Thus we have shown that the bootstrap particle filter\index{particle filter!bootstrap} has at its
core a form of learning the prior, in this case construction of an empirical measure\index{probability!measure!empirical} from the predicted samples.

\subsection{Ensemble Kalman Filter}\index{Kalman filter!ensemble} \label{ssec:enkf55}
Now consider the setting where the particles $\{\widehat v_{j+1}^{(n)}\}_{n=1}^N$ are derived from
the prediction step of the EnKF. Initially we consider the setting in which $h(\cdot)=H\cdot$ is linear, then generalize to nonlinear $h$.
In the setting of linear observation map, the inverse problem defining the analysis step is to find $\vd_{j+1}$ given $\yd_{j+1}$ where
\begin{equation} \label{eq:oba2}
\yd_{j+1}=H\vd_{j+1}+\etad_{j+1}.
\end{equation}
Recall the notation for $\Gn$, the projection onto Gaussian measures, defined in Remark~\ref{rem:bishop}. 
To define the EnKF we learn the Gaussian prior\index{Gaussian!prior}
\begin{equation}
\hapost_{j+1} = \Gn\Bigl(\frac{1}{N}\sum_{n=1}^N \delta_{\widehat v_{j+1}^{(n)}}\Bigr).
\end{equation}
We obtain $\hapost_{j+1}=\cN\bigl(\widehat{m}_{j+1},\widehat{C}_{j+1}\bigr)$, where
\begin{subequations} \label{eq:mav}
\begin{align}
    \widehat{m}_{j+1} &= \frac{1}{\Sam}\sum^{\Sam}_{\sam=1} \widehat{v}_{j+1}^{(\sam)},\\
    \widehat{C}_{j+1} &= \frac{1}{\Sam}\sum^{\Sam}_{\sam=1}\bigl(\widehat{v}^{(\sam)}_{j+1}-\widehat{m}_{j+1}\bigr)\otimes \bigl(\widehat{v}^{(\sam)}_{j+1}-\widehat{m}_{j+1}\bigr).
\end{align}
\end{subequations}
Thus we have adopted a different form of learning the prior, in comparison to the
bootstrap particle filter, since we have formed a Gaussian. If we solve the linear inverse problem defined by \eqref{eq:oba2}, for $\vd_{j+1}$ given $\yd_{j+1}$, using this Gaussian prior\index{Gaussian!prior}, we obtain a Gaussian posterior\index{Gaussian!posterior}. Approximate samples from this Gaussian posterior are given by
\begin{subequations}
\begin{align}
v_{j+1}^{(\sam)} &= \widehat{v}_{j+1}^{(\sam)}+K_{j+1}\bigl(\yd_{j+1}-\eta_{j+1}^{(\sam)}-H\widehat{v}_{j+1}^{(\sam)}\bigr),\quad \sam=1,\ldots,\Sam,\\
K_{j+1} &= \widehat{C}_{j+1} H^\top \bigl(H \widehat{C}_{j+1} H^\top + \Gamma\bigr)^{-1},
\end{align}
\end{subequations}
where $\eta_{j+1}^{(n)} \sim \Nc(0, \Gamma)$ are drawn i.i.d. with respect to particle index $n$, as well as time-index $j.$
This simply defines the EnKF analysis step from Subsection \ref{ssec:enkf}, in the linear case
described in Remark \ref{rem:lh}. Thus, in the case of
linear $H$, the EnKF can be  thought of as being derived from using a learned Gaussian prior in every analysis step, and then sampling the posterior, before cycling back to the prediction step.

Similar ideas may be applied when the observation operator $h$ is
nonlinear, but instead of learning the \emph{prior} in the
Gaussian\index{Gaussian!prior} class, we learn a Gaussian approximation
of the \emph{joint distribution} of the predicted state and observation.
To this end, we define the joint distribution
\begin{equation}
\label{eq:joi22}
\widehat{\gamma}_{j+1}
=
\Gn\left(
\frac{1}{\Sam}
\sum_{\sam=1}^{\Sam}
\delta_{\widehat v_{j+1}^{(\sam)}}
\otimes
\Nc\bigl(h(\widehat v_{j+1}^{(\sam)}),\Gamma\bigr)
\right).
\end{equation}
We have again used the notation $\Gn,$ the projection onto Gaussian
measures defined in Remark~\ref{rem:bishop}. The mean of
$\widehat{\gamma}_{j+1}$ is
\begin{equation}
\begin{pmatrix}
\widehat{m}_{j+1}\\
\widehat{h}_{j+1}
\end{pmatrix},
\end{equation}
and its covariance is
\begin{equation}
\begin{pmatrix}
\widehat{C}_{j+1}
&
\widehat{C}_{j+1}^{vh}\\
\widehat{C}_{j+1}^{hv}
&
\widehat{C}_{j+1}^{hh}+\Gamma
\end{pmatrix},
\end{equation}
where
\begin{subequations}
\label{eq:00gainEnKF}
\begin{align}
\widehat{m}_{j+1}
&=
\frac{1}{\Sam}
\sum_{\sam=1}^{\Sam}
\widehat{v}_{j+1}^{(\sam)},\\
\widehat{h}_{j+1}
&=
\frac{1}{\Sam}
\sum_{\sam=1}^{\Sam}
h\bigl(\widehat{v}_{j+1}^{(\sam)}\bigr),\\
\widehat{C}_{j+1}^{vh}
&=
\frac{1}{\Sam}
\sum_{\sam=1}^{\Sam}
\bigl(\widehat{v}_{j+1}^{(\sam)}-\widehat{m}_{j+1}\bigr)
\otimes
\bigl(h(\widehat{v}_{j+1}^{(\sam)})-\widehat{h}_{j+1}\bigr),\\
\widehat{C}_{j+1}^{hh}
&=
\frac{1}{\Sam}
\sum_{\sam=1}^{\Sam}
\bigl(h(\widehat{v}_{j+1}^{(\sam)})-\widehat{h}_{j+1}\bigr)
\otimes
\bigl(h(\widehat{v}_{j+1}^{(\sam)})-\widehat{h}_{j+1}\bigr).
\end{align}
\end{subequations}
Here $\widehat{C}_{j+1}$ is given by (\ref{eq:mav}b), and
$\widehat{C}_{j+1}^{hv}=(\widehat{C}_{j+1}^{vh})^\top.$

Conditioning the Gaussian approximation $\widehat{\gamma}_{j+1}$ on
the observed value $\yd_{j+1}$ motivates the perturbed-observation
EnKF update
\begin{subequations}
\label{eq:00EnKF}
\begin{align}
\widehat{y}_{j+1}^{(\sam)}
&=
h\bigl(\widehat{v}_{j+1}^{(\sam)}\bigr)
+
\eta_{j+1}^{(\sam)},
\qquad
\eta_{j+1}^{(\sam)}
\sim
\Nc(0,\Gamma),\\
v_{j+1}^{(\sam)}
&=
\widehat{v}_{j+1}^{(\sam)}
+
K_{j+1}
\bigl(
\yd_{j+1}-\widehat{y}_{j+1}^{(\sam)}
\bigr),
\qquad
\sam=1,\ldots,\Sam,\\
K_{j+1}
&=
\widehat{C}_{j+1}^{vh}
\bigl(
\widehat{C}_{j+1}^{hh}+\Gamma
\bigr)^{-1}.
\end{align}
\end{subequations}
The variables $\eta_{j+1}^{(\sam)}$ are drawn independently across
the ensemble index $\sam$ and time index $j.$ The update
\eqref{eq:00EnKF} is the EnKF analysis step from
Subsection~\ref{ssec:enkf}, as given by equations
(\ref{eq:predEnKF}b) and \eqref{eq:gainEnKF}. Thus, when $h$ is
nonlinear, the EnKF can be viewed as learning a Gaussian approximation
of the joint distribution of the predicted state and observation,
conditioning this approximation on the observed data, and using the
resulting conditional Gaussian structure to update the ensemble before
cycling back to the prediction step.

\section{Data Assimilation in a Latent Space}\index{latent space}\label{sec:latent_DA}

In this section we perform data assimilation in a latent space representation of the state space $\Ru$. Let $f: \R^d  \to \R^{\dz}$ and  $g: \R^{\dz} \to \R^d$
be an encoder--decoder pair, as discussed in Section \ref{sec:auto}.
Thus $g \circ f \approx \Id$ on $\Ru.$ The pair $(f,g)$ define an autoencoder\index{autoencoder}.
If $g$ is invertible, we can take $f = g^{-1}$.
We now express the likelihood and prior, and consequently the posterior, in terms of the latent space variables. There are some similarities with the ideas underlying Section~\ref{sec:pertp3} where we reformulate the inverse problem in a latent space for variable $z$; in that section, however, we primarily focus on the setting $\dz=d$ whereas here our primary motivation is the setting where $\dz \ll d.$ 

We start by making the following assumption concerning available data:
\begin{dataassumption}\index{Data Assumption}\label{da:latent_DA}
 The dynamics model (\ref{eq:sdm}a)  has an invariant measure $\mu.$ 
We are given a trajectory $\{v^{\rm train}_1,\ldots,v^{\rm train}_N\}$ 
generated by the dynamics model (\ref{eq:sdm}a)
with $v^{\rm train}_0 \sim \mu$.
\end{dataassumption}
For each integer $n$ in $\{1,\cdots,N\}$, the point $v^{\rm train}_n$ is drawn from $\mu$, albeit not independently.  Using this data for training,
and following techniques such as the variational autoencoder\index{autoencoder!variational} described in Chapter~\ref{ch:UL}, we obtain a map $g$ that approximately pushes forward a latent space distribution $\zeta$ on $\R^{\dz}$ to $\mu$ on $\R^d$; that is, $g_\sharp\zeta\approx \mu$. We also assume that map $f$ approximately pushes forward $\mu$ to the latent space: $f_\sharp\mu\approx \zeta$.

\begin{remark}
We note that although $\zeta$ will not directly be used as a prior in the data assimilation problem, it is a natural latent 
reference distribution for the model \eqref{eq:SSM1}, since it represents the model statistics through its invariant measure.
\end{remark}

We now proceed to show how $g$ and $f$ can be used to perform smoothing and filtering in the latent space.

\subsection{Smoothing in a Latent Space}\index{latent space}

Adapting the idea of latent space representation contained in
Equation~\ref{eq:cf2}, and the definition of the smoothing distribution
in Subsection~\ref{ssec:smoothing}, we define $Z^\dagger \in \R^{(J+1)\dz}$
and recall $\Vd \in \R^{(J+1)\du}:$
\begin{align*}
Z^\dagger & :=\{z^\dagger_0,z^\dagger_1,\ldots,z^\dagger_J\},\\
\Vd & := \{\vd_0, v^\dagger_1, \ldots, \vd_J\}. 
\end{align*}
We then also define
$\mathsf{g}:\R^{(J+1)\dz}\to\R^{(J+1)\du}$ and
$\hs:\R^{(J+1)\du}\to\R^{J\dy}$ by
\begin{align*}
\mathsf{g}(Z^\dagger)
&:=
\bigl\{
g(z^\dagger_0),g(z^\dagger_1),\ldots,g(z^\dagger_J)
\bigr\},\\
\hs(\Vd)
&:=
\bigl\{
h(\vd_1),\ldots,h(\vd_J)
\bigr\}.
\end{align*}
The likelihood for the smoothing problem in the latent space is then
given by
\begin{equation*}
\like_g(\Yd|Z^\dagger)
\propto
\exp\Bigl(
-\frac12
\bigl|
\Yd-\hs(\mathsf{g}(Z^\dagger))
\bigr|_{\Gammas}^2
\Bigr).
\end{equation*}


We then define a prior in the latent space as\index{latent space}
\begin{equation*}
\rho_g(Z^\dagger) \propto \exp\Bigl(-\frac12|g(z^\dagger_0)-m_0|_{C_0}^2-\frac12\sum_{j=0}^{J-1}|g(z^\dagger_{j+1})-\Psi\bigl(g(z^\dagger_{j})\bigr)|_{\Sigma}^2\Bigr),
\end{equation*}
leading to the latent space posterior\index{latent space}
\begin{equation*}
\Pi^{\Yd}_g\!(Z^\dagger) \propto \like_g(\Yd|Z^\dagger)\rho_g(Z^\dagger).
\end{equation*}

We then have the posterior in the original space given approximately by
\begin{equation*}
    \Pi^{\Yd}\!(\Vd) \approx \mathsf{g}_\sharp\left(\Pi^{\Yd}_g\!(Z^\dagger)\right).
\end{equation*}

\begin{remark}
    We note that the latent-space prior is not the pullback of the original prior, even if $g$ is invertible. 
    This is because the change of variables from $\Vd$ to $Z^\dagger$ introduces determinants of the Jacobian of the change of variables. So, in the posterior approximation,  there are two approximations: one arising from the fact that $g$ is only an approximate push-forward; and a second, modeling, approximation that may be harder to justify. This observation also applies to the discussion of MAP estimation (4DVar)\index{4DVar} in what follows below, since MAP\index{MAP estimator} estimators are not invariant under nonlinear reparameterization.
\end{remark}

\subsubsection{Example: 4DVar in a Latent Space}\index{4DVar}\index{latent space}

4DVar is derived as the maximum \emph{a posteriori} (MAP)\index{MAP estimator} estimate of the smoothing distribution; see Subsection~\ref{ssec:4dvar}. 
To reduce the complexity of solving the maximization problem in $\R^{(J+1)d}$ we introduce the idea of finding a MAP estimator for the latent space posterior, in the smaller space $\R^{(J+1)d_z}.$  This leads to the following optimization problem:
\begin{subequations}
\begin{align}
    \J(Z^\dagger;\Yd) &= \frac12|g(z^\dagger_0)-m_0|_{C_0}^2+\frac12\sum_{j=0}^{J-1}|g(z^\dagger_{j+1})-\Psi\bigl(g(z^\dagger_{j})\bigr)|_{\Sigma}^2
    + \frac{1}{2} \displaystyle{\sum_{j=0}^{J-1}}|\yd_{j+1} - h(g(z^\dagger_{j+1}))|_{\Gamma}^2;\\
    \Zmap &\in \argmin_{Z^\dagger \in \R^{(J+1)\dz}} \J(Z^\dagger;\Yd).\label{eq:4dvar_latent}
\end{align}
\end{subequations}
To obtain the state estimate $\Vmap$ in the original space $\Ru$, we define
\begin{equation}
  \Vmap = \mathsf{g}(\Zmap).
\end{equation}

\begin{remark}
    If $\dz\ll \du$, there may be significant computational gains from carrying out the 4DVar optimization problem \eqref{eq:4dvar_latent} in the latent space rather than in $\Ru$ as in \eqref{eq:4DVarasMAP}, which motivates this approach. Note, however, that the cost of evaluating 
    the decoder map $g(\cdot)$ must be factored in.
\end{remark}

\subsection{Filtering in a Latent Space}

Thus far we have used only the decoder $g$; this suffices for the smoothing problem
and for its attendant MAP\index{MAP estimator} estimator. For filtering we will
also need the encoder $f$, allowing us to perform prediction in the original state space,
then encode to perform the analysis step in the latent space.
Recalling the definition of the filtering distribution in Subsection~\ref{ssec:filtering}, we may define the latent space analysis operator $\An_g(\cdot;y)$ by
\begin{equation*}
  \An_g(\pi;y)(z) = \frac{\like_g(y|z)\pi(z)}{\int_{\R^{\dz}} \like_g(y|z)\pi(z) \, dz},
\end{equation*}
where
\begin{equation*}
    \like_g(y|z)\propto\exp \left( - \frac{1}{2} \bigl\lvert y - h(g(z)) \bigr\rvert_{\Gamma}^2 \right).
\end{equation*}

We now define the modified prediction--analysis cycle, wherein the analysis step is performed in the latent space while the prediction step \eqref{eq:pred_operator} is unchanged:
\begin{align*}
\begin{split}
{ \text{ \bf  Prediction Step:}} ~~~~\;  &(\widehat{\post}_g)_{j+1} = \Pred (\post_g)_j.  \index{prediction} \\
 {\text{  \bf Latent Analysis Step:}} ~~~~~  &(\post_g)_{j+1}=g_\sharp\An_g(f_\sharp(\widehat{\post}_g)_{j+1};\yd_{j+1}). \index{analysis} 
\end{split}
\end{align*}
Then, $(\pi_g)_j\approx \pi_j$, where $\pi_j = \Prob(\vd_j|\Yd_j)$ is the true filtering distribution. Note that $\pi_g$, like $\pi$, is
in $\cP(\R^\du).$

\subsubsection{Example: EnKF in a Latent Space}\index{Kalman filter!ensemble}\index{latent space}
We start by defining an observation operator that maps from the latent space to the observation space, $h_g: \R^{\dz}\to\R^{\dy}$:
\begin{equation*}
    h_g = h\circ g.
\end{equation*}

To obtain the ensemble $\{v_{j+1}^{(n)}\}_{n=1}^N$ from $\{v_j^{(n)}\}_{n=1}^N$, we first perform the forecast step, then map into the latent space, perform the EnKF update, and map back into the original space:
\begin{subequations}
\label{eq:predEnKF_latent}
\begin{align}
\widehat{v}_{j+1}^{(\sam)} &= \Psi(v_{j}^{(\sam)})+\xi^{(\sam)}_{j}, \quad \sam=1,\ldots,\Sam,\\
\widehat{z}_{j+1}^{(\sam)} &= f(\widehat{v}_{j+1}^{(\sam)}),\quad \sam=1,\ldots,\Sam,\\
z_{j+1}^{(\sam)} &= \widehat{z}_{j+1}^{(\sam)}+K_{j+1}\bigl(\yd_{j+1}-\eta_{j+1}^{(\sam)}-h_g(\widehat{z}_{j+1}^{(\sam)})\bigr),\quad \sam=1,\ldots,\Sam,\\
v_{j+1}^{(\sam)} &= g(z_{j+1}^{(\sam)}),\quad \sam=1,\ldots,\Sam.
\end{align}
\end{subequations}
Here $\xi_{j}^{(\sam)} \sim \Nc(0,\Sigma), \eta_{j+1}^{(\sam)} \sim \Nc(0,\Gamma)$, as before.

The gain\index{gain} matrix $K_{j+1}$ is calculated according to
\begin{subequations}
\label{eq:gainEnKF_latent}
\begin{align}
\widehat{m}_{j+1} &= \frac{1}{\Sam}\sum^{\Sam}_{\sam=1} \widehat{z}_{j+1}^{(\sam)}, \\
\widehat{h}_{j+1} &= \frac{1}{\Sam}\sum^{\Sam}_{\sam=1} h_g(\widehat{z}_{j+1}^{(\sam)}), \\
\widehat{C}_{j+1}^{zh} &= \frac{1}{\Sam}\sum^{\Sam}_{\sam=1}\bigl(\widehat{z}^{(\sam)}_{j+1}-\widehat{m}_{j+1}\bigr)\otimes \bigl(h_g(\widehat{z}^{(\sam)}_{j+1})-\widehat{h}_{j+1}\bigr),\\
\widehat{C}_{j+1}^{hh} &= \frac{1}{\Sam}\sum^{\Sam}_{\sam=1}\bigl(h_g(\widehat{z}^{(\sam)}_{j+1})-\widehat{h}_{j+1}\bigr)\otimes \bigl(h_g(\widehat{z}^{(\sam)}_{j+1})-\widehat{h}_{j+1}\bigr),\\
\widehat{C}_{j+1}^{yy} &= \widehat{C}_{j+1}^{hh}+\Gamma,\\
K_{j+1} &= \widehat{C}_{j+1}^{zh}\bigl(\widehat{C}_{j+1}^{yy}\bigr)^{-1}.
\end{align}
\end{subequations}

\begin{remark}
    As with the latent space 4DVar, the latent space EnKF may be computationally beneficial when $\dz\ll \du$. In this setting, moreover, it may suffer less from the problems of sampling error discussed in Subsection~\ref{ssec:enkf}.
\end{remark}

\begin{remark}
    To avoid having to map back into $\Ru$ after the analysis step and apply the possibly costly forward dynamics $\Psi$, one can instead learn a surrogate $\Psi'_g$ and noise covariance matrix $\Sigma_g'$ for the dynamics in the latent space. This can be done by mapping the data given in Data Assumption~\ref{da:latent_DA} to the latent space, using
    the encoder, and then using the methods discussed in Chapter~\ref{ch:ts_forecasting} for time-series forecasting to learn a dynamics model in the latent space; see also Section~\ref{sec:ml_approx_da} for the use of surrogates in data assimilation. Then, the latent forecast ensemble can be given directly by
    \begin{align*}
        \widehat{z}_{j+1}^{(\sam)} &= \Psi'_g(z_{j}^{(\sam)})+(\xi_g')^{(\sam)}_{j},\quad \sam=1,\ldots,\Sam,\\
        (\xi_g')^{(\sam)}_{j} &\sim \mathcal{N}(0, \Sigma'_g),\quad \sam=1,\ldots,\Sam.
    \end{align*}
    
\end{remark}

\section{Bibliography}\label{sec:144}

In this chapter we have introduced two computational frameworks for joint state and parameter estimation: the EM algorithm and \index{Kalman filter!auto-differentiable}auto-differentiable filters. Both frameworks find the parameters in the \index{dynamical system}dynamical system by approximating the \index{maximum likelihood estimation}maximum likelihood estimator, and then use a \index{filtering}filtering or \index{smoothing}smoothing algorithm to recover the \index{state}state. We have focused on learning the dynamics model, but both frameworks directly generalize to learning the observation model; thus, the EM algorithm and auto-differentiable filters can be used not only to learn the prior distribution for the smoothing problem, but also the likelihood function. This final section provides bibliographical context for the algorithms considered in this chapter, and briefly discusses other computational methods that do not stem from a \index{maximum likelihood estimation}maximum likelihood formulation.

Embedding of the \index{Kalman filter!ensemble}EnKF and the \index{Kalman smoother!ensemble}ensemble Kalman smoother (EnKS) into the \index{expectation maximization}EM algorithm was proposed in \cite{tandeo2015offline,ueno2014iterative, dreano2017estimating, pulido2018stochastic}, with a focus on estimation of error covariance matrices.  The E-step is approximated with an EnKS under the \index{Monte Carlo}Monte Carlo EM framework \cite{wei1990monte}. In addition, \cite{brajard2020combining,nguyen2019like,wikner_using_2021} incorporate machine learning techniques in the M-step to train \index{neural network}neural network surrogate\index{surrogate} models or corrections. The paper \cite{bocquet2020bayesian} proposes \index{Bayesian}Bayesian estimation of model error statistics, together with a \index{neural network}neural network emulator\index{emulator} for the \index{dynamical system}dynamical system. On the other hand, \cite{ueno2016bayesian,tandeo_review_2020,cocucci2021model} consider online \index{expectation maximization}EM methods for error covariance estimation with \index{Kalman filter!ensemble}EnKF. Online methods aim to reduce computation by not reprocessing the smoothing distribution for each new observation. Although gradient information is used during the M-step to train the surrogate\index{surrogate} models in~\cite{brajard2020combining, nguyen2019like,bocquet2020bayesian}, these methods do not \index{auto-differentiation}auto-differentiate through the \index{Kalman filter!ensemble}EnKF. Other methods to learn observation and model error covariances are considered in \cite{mehra_identification_1970,berry_adaptive_2013,waller_theoretical_2016,menard_error_2016}.

Auto-differentiable Kalman filters were introduced in \cite{chen2021auto}, which proposes and analyzes an approach for joint \index{state}state and parameter estimation that leverages gradient information of an \index{Kalman filter!ensemble}EnKF estimate of the \index{likelihood}likelihood. A unified framework for co-learning state, dynamics, and filtering algorithms in data assimilation using auto-differentiable Kalman methods was developed in \cite{adrian2026auto}. The paper \cite{chen2023reduced} considers learning a latent low-dimensional surrogate\index{surrogate} model for the dynamics and a decoder that maps from the latent space to the state space using auto-differentiable Kalman filters. 
\index{Kalman filter!ensemble}EnKFs for derivative-free \index{maximum likelihood estimation}maximum likelihood estimation are studied in \cite{stroud2010ensemble,pulido2018stochastic}. An empirical comparison of the \index{likelihood}likelihood computed using the  \index{Kalman filter!ensemble}EnKF and other \index{filtering}filtering algorithms is made in \cite{carrassi2017estimating}; see also \cite{hannart2016dada,metref2019estimating}. The paper \cite{drovandi2021ensemble} uses \index{Kalman filter!ensemble}EnKF \index{likelihood}likelihood estimates to design a pseudo-marginal \index{MCMC}MCMC method  for \index{Bayesian!inference}Bayesian inference of model parameters. The works \cite{stroud2007sequential,stroud2018bayesian} propose online \index{Bayesian}Bayesian parameter estimation using the \index{likelihood}likelihood computed from the \index{Kalman filter!ensemble}EnKF under a certain family of conjugate distributions.

While our discussion has focused on ensemble Kalman methods, \index{particle filter}particle filters can also be employed for joint \index{state}state and parameter estimation. \index{particle filter}Particle filters give an unbiased estimate of the data \index{likelihood}likelihood \cite{del2004feynman,andrieu2010particle}. Based upon this \index{likelihood}likelihood estimate, a particle \index{MCMC}MCMC \index{Bayesian}Bayesian parameter estimation method is designed in \cite{andrieu2010particle}. Although \index{particle filter}particle filter \index{likelihood}likelihood estimates are unbiased, they suffer from two potential drawbacks that limit their applicability to some problems. First, their variance can be large, as they inherit the weight degeneracy of \index{importance sampling}importance sampling in high dimensions \cite{snyder2008obstacles,agapiou2017importance,sanz2018importance,sanz2020bayesian}---see Chapter~\ref{lecture7} for further background on this subject. Second, while the forecast and analysis steps of \index{particle filter}particle filters can be \index{auto-differentiation}auto-differentiated, the \index{resampling}resampling steps involve discrete distributions that cannot be handled by the reparameterization trick, as discussed in \cite{chen2021auto}. For this reason, previous differentiable \index{particle filter}particle filters omit \index{auto-differentiation}auto-differentiation of the \index{resampling}resampling step \cite{naesseth2018variational,maddison2017filtering,le2017auto}, introducing a bias.

An alternative to \index{maximum likelihood estimation}maximum likelihood estimation is to optimize a lower bound on the data \index{likelihood}log-likelihood with \index{variational!Bayesian method}variational inference \cite{bishop,kingma2013auto,ranganath2014black}. The \index{posterior}posterior distribution over the \index{latent state}latent \index{state}states is approximated with a parametric distribution and is jointly optimized with model parameters defining the \index{dynamical system!stochastic}stochastic dynamics model. In this direction, \index{variational!Bayesian method}variational sequential \index{Monte Carlo}Monte Carlo methods  \cite{naesseth2018variational,maddison2017filtering,le2017auto} construct the lower bound using a \index{particle filter}particle filter. Moreover, the proposal distribution of the \index{particle filter}particle filter is parameterized and jointly optimized with model parameters. 
Although \index{variational!Bayesian method}variational sequential \index{Monte Carlo}Monte Carlo methods provide consistent data \index{likelihood}log-likelihood estimates, they suffer from the same two potential drawbacks as \index{likelihood}likelihood-based \index{particle filter}particle filter methods. A recent work \cite{ishizone2020ensemble} proposes blending \index{variational!Bayesian method}variational sequential \index{Monte Carlo}Monte Carlo and \index{Kalman filter!ensemble}EnKF with an \index{importance sampling}importance sampling-type lower bound estimate, which is effective if the \index{state}state dimension is small. Other works that build on the \index{variational!Bayesian method}variational inference framework include \cite{krishnan2017structured,rangapuram2018deep,fraccaro2017disentangled, marino_general_2018}. An important challenge is to obtain suitable parameterizations of the \index{posterior}posterior, especially when the \index{state}state dimension is high. For this reason, a restrictive \index{Gaussian}Gaussian parameterization with a diagonal covariance matrix is often used in practice \cite{krishnan2017structured,fraccaro2017disentangled}. The topic of variational approximations to filtering and smoothing problems is discussed in Chapter~\ref{ch:LG}.

Another alternative approach to maximum likelihood estimation is to concatenate state and parameters into an augmented \index{state-space}state-space and employ the data assimilation methods, from other
chapters in this Part II, in the augmented state-space. This approach requires one to design a pseudo-dynamic for the parameters, which can be challenging when certain types of parameters (e.g., error covariance matrices) are involved \cite{stroud2007sequential,delsole2010state} or if the dimension of the parameters is high.
The use of the \index{Kalman filter!ensemble}EnKF for jointly learning the \index{state}state and model parameters by \index{state augmentation}\emph{state augmentation} was introduced in~\cite{anderson2001ensemble}.

Beyond the problem of joint parameter and state estimation, the development of data-driven \index{machine learning} frameworks for learning \index{dynamical system}dynamical systems is a very active research area. We refer to \cite{levine_framework_2022} for a framework and to \cite{brunton2019data,gottwald2021supervised,gottwald_combining_2021,harlim2021machine,raissi2018multistep} for recent methods that do not rely on the \index{expectation maximization}EM algorithm, \index{auto-differentiation}auto-differentiation of \index{filtering}filtering methods, or \index{variational!Bayesian method}variational inference. In addition, data-driven machine learning frameworks have gained popularity for combining physics-based models with model error corrections. These hybrid models may avoid the over-smoothing or unphysical behavior of predictive models that replace the entire forecast dynamics~\cite{bonavita2023limitations}. Two common representations for model error include  additive corrections~\cite{farchi2021using, bonavita2020machine} and data-driven subgrid scale parameterizations~\cite{rasp2018deep, bolton2019applications}; we refer to Section~\ref{sec:ME} for other forms of model error. We note that the approach discussed in this chapter learns the corrections offline, as online methods typically require the adjoint operator of the physics-based model, which may be challenging to obtain for some physics-based models, e.g., in numerical weather prediction and for climate modeling~\cite{lopez2022training}.

Model error correction using analysis increments is considered in \cite{danforth_estimating_2007,farchi2021using,chen_correcting_2022}. Alternatively, 
machine learning can be used to account for model error in data assimilation by learning a mapping between the attractor defined by the model and the attractor of the true system; see~\cite{agarwal_cross-attractor_2024} for one approach. Another approach to correcting for model error in weather forecasting is known as \emph{post-processing}; in particular, it is common to post-process ensemble forecasts. See \cite{vannitsem_statistical_2018} for a survey of approaches, as well as \cite{hohlein_postprocessing_2024} for an approach using permutation-invariant neural networks.

Performing data assimilation in the latent space of an autoencoder\index{autoencoder} (Section~\ref{sec:auto}) or variational autoencoder\index{autoencoder!variational} (Section~\ref{sec:vauto}) is considered in 
\cite{peyron_latent_2021,chen2023reduced,melinc_3d-var_2024,pasmans_ensemble_2025}.
Recent application to problems with shocks may be found in \cite{zhou2026neural,chandravamsi2026feature}.
\chapter{\Large{\sffamily{Transport Perspective on Filters}}}
\label{ch:LG2}

This chapter is devoted to the transport\index{transport} perspective on filtering; its content is therefore related to
Chapter~\ref{ch:transport} concerning transport methods for 
inverse problems\index{inverse problem}. Here, the term \emph{transport} refers broadly to a map designed to represent aspects of the analysis\index{analysis} step within the filtering update \eqref{eq:pna}. To see the connection with Chapter~\ref{ch:transport}, recall from Subsection~\ref{ssec:filtering} that the analysis step can in fact be formulated as an inverse problem. The transport framework for filtering developed in this chapter encompasses methods for both \index{probabilistic estimation}probabilistic and \index{state estimation}state estimation as defined in Subsection~\ref{ssec:pvse}. In  probabilistic estimation\index{probabilistic estimation} the aim is to approximate the analysis map $\An$ itself; in state estimation\index{state estimation} the aim is to determine a single point estimate for the solution of the inverse problem defined by $\An.$

We start with the dynamics and data model~\eqref{eq:sdm} and~\eqref{eq:dm}, repeated here for convenience. Consider initialization $\vd_0 \sim \Nc(m_0, C_0)$ and, for  $j \in \Z^+,$ let
\begin{subequations}
    \label{eq:ddm99}
\begin{align}
\vd_{j+1} &= \Psi(\vd_j) + \xid_j, \label{eq:ddm99_dynamics}\\
\yd_{j+1} &= h(\vd_{j+1}) + \etad_{j+1}. \label{eq:ddm99_obs}
\end{align}
\end{subequations}
The assumptions on the initial condition and the noise sequences are as detailed in Assumption \ref{a:noise}.
Recall Definition~\ref{def:FP} of the filtering problem\index{filtering problem}
and the notation for $\Vd, \Yd, \Yd_j$ introduced in Subsection~\ref{ssec:note}. We are interested in designing algorithms to learn about $\vd_j$, given all the data observed up to time $j$, $\Yd_j.$  In this chapter we will study algorithms for estimating the state,\index{state estimation} $\{\vd_j\}_{j \in \mathbb{Z}^+}$, or for approximating the filtering distribution on the state,\index{probabilistic estimation} 
$\{\pi_j\}_{j \in \mathbb{Z}^+}.$ In both cases we consider sequential algorithms with respect to $j$, and refer to these collectively as \emph{filters}\index{filter}.
The transport algorithms we consider take the form of mappings from $\R^{\du}$ (or multiple copies thereof) into itself. We will parameterize these mappings and learn the parameters from training data. 
These learnable mappings are
based on the prediction\index{prediction}--analysis\index{analysis} 
cycle~\eqref{eq:pna}; we also briefly consider the
modified prediction\index{prediction}--analysis\index{analysis} structure underlying
the optimal particle filter\index{particle filter!optimal} from Subsection \ref{ssec:opf}.

\paragraph{Generalizing 3DVar}
We start by formulating a class of state estimation algorithms of the form, 
for $j \in \Z^+$,
\begin{subequations}
 \label{eq:pred3DVAR2add}   
\begin{align}
\hat{v}_{j+1} &= \Psi(v_j)+s\xi_j,\\
v_{j+1} &= T\bigl(\hat{v}_{j+1};\yd_{j+1},\theta\bigr).	
\end{align}
\end{subequations}
The algorithm is initialized at $v_0$, typically sampled from Gaussian
$\cN(m_0,C_0)$, the known distribution of $\vd_0.$
We allow for both $s=0$ and $s=1$, as in our formulation of 3DVar in \eqref{eq:pred3DVAR}. We will primarily use this form of state estimation algorithm with $s=0$ but use of $s=1$, with $\xi_{j} \sim \Nc(0,\Sigma)$ an i.i.d. sequence, may also be considered. In the context of state estimation, the objective is to choose $\theta \in \Theta \subseteq \R^p$ 
so that $V = \{v_j\}_{j=0}^J$ determined by \eqref{eq:pred3DVAR2add} approximates well the signal $\Vd = \{v_j\}_{j=0}^J$ given by the dynamics model \eqref{eq:ddm99}. Notice that $v_j$ defined by \eqref{eq:pred3DVAR2add} depends only on  observations $\Yd_j$ up to time $j.$ Consequently, the sequence $V = \{v_j\}_{j=0}^J$ provides a \emph{filter} for state estimation.\index{state estimation}

\paragraph{Generalizing EnKF}
We now introduce ensemble filters that can be used for either
state estimation\index{state estimation}
or probabilistic estimation.\index{probabilistic estimation}
Motivated by the EnKF in form \eqref{eq:fakeobs}, we generalize 
\eqref{eq:pred3DVAR2add} and consider a particle-based map of the form,
for $j \in \Z^+$,
\begin{subequations}
\label{eq:fakeobs2}
\begin{align}
\widehat{v}_{j+1}^{(\sam)} &= \Psi(v_{j}^{(\sam)})+\xi^{(\sam)}_{j}, \quad \sam=1,\ldots,\Sam, \\
\widehat{y}_{j+1}^{(\sam)} &= h(\hv_{j+1}^{(\sam)})+\eta^{(\sam)}_{j+1}, \quad \sam=1,\ldots,\Sam, \\
v_{j+1}^{(\sam)} &= T\Bigl(\widehat{v}_{j+1}^{(\sam)},\widehat{y}_{j+1}^{(\sam)},\yd_{j+1},
\{\widehat{v}_{j+1}^{(\ell)},\widehat{y}_{j+1}^{(\ell)}\}_{\ell=1}^\Sam;
\theta\Bigr),\quad \sam=1,\ldots,\Sam. \label{eq:analysis_map_ensemble_dependence}
\end{align}
\end{subequations}
The recursion may be initialized, for instance, by independent samples $v_0^{(n)} \sim \cN(m_0,C_0).$
The noise defining this random dynamical system satisfies Assumption \ref{a:enkf-noise}.
For any given integer $n$ the resulting
noise sequences over $j$ are thus distributionally equivalent
to the noise sequences appearing in \eqref{eq:ddm99},
as detailed in Assumption \ref{a:enkf-noise}. 
The collection $\{{v}_{j}^{(\ell)}\}_{\ell=1}^\Sam$
defines the empirical\index{empirical} measure
\begin{equation}
\label{eq:empEnKF99}
    \pia_j(\theta) = \frac{1}{N}\sum_{\ell=1}^N \delta_{v_j^{(\ell)}}.
\end{equation}
We may define a state estimator $v_j(\theta)$ by taking the mean with respect to this
probability measure:
\begin{equation}
\label{eq:SE}
    v_j(\theta) = \frac{1}{N}\sum_{\ell=1}^N v_j^{(\ell)}.
\end{equation}

For each integer $1 \le n \le N$, the step (\ref{eq:fakeobs2}a) defines a Markov operator $\Pred$, 
identical to the one defining the prediction step in~\eqref{eq:pna}. 
Furthermore, under (\ref{eq:fakeobs2}a), $\mathbb{P}(\widehat{v}_{j+1}^{(n)}|{v}_{j}^{(n)})$ is Gaussian. As a consequence of \eqref{eq:empEnKF99} it follows
that, conditioned on a given ensemble at time $j,$ $\Pred \pia_j$ is a Gaussian mixture. 
The points $\{\widehat{v}_{j+1}^{(\ell)}\}_{\ell=1}^\Sam$ then comprise samples
from this Gaussian mixture. Later it will be useful to consider the empirical measure 
defined by these samples:
\begin{equation}
\label{eq:empEnKF88}
     \piaw_{j+1}(\theta)=\frac{1}{N}\sum_{\ell=1}^N \delta_{\widehat{v}_{j+1}^{(\ell)}}.
\end{equation}

\begin{remark}
\label{rem:perm} 
Recall mapping $\pi \to \gamma_\pi$ defined by \eqref{eq:n4ukf}.
Then the points 
$\{\widehat{v}_{j+1}^{(\ell)},\widehat{y}_{j+1}^{(\ell)}\}_{\ell=1}^\Sam$
comprise samples from $\gamma_{\Pred \pia_j}$ and
can be used to define the empirical measure 
\begin{equation}
\label{eq:empEnKF9889}
     \frac{1}{N}\sum_{\ell=1}^N \delta_{
(\widehat{v}_{j+1}^{(\ell)},\widehat{y}_{j+1}^{(\ell)})}.
\end{equation}
\end{remark}

The empirical measures \eqref{eq:empEnKF88} and \eqref{eq:empEnKF9889} 
are invariant under permutations of the ensemble from
which they are formed. We assume henceforth 
that dependence of the map $T(\cdot)$ on 
$\{\widehat{v}_{j+1}^{(\ell)},\widehat{y}_{j+1}^{(\ell)}\}_{\ell=1}^\Sam$
is also invariant under permutations in index $\ell.$ To be specific, the map in~\eqref{eq:analysis_map_ensemble_dependence} is assumed to take the form (abusing notation) 
\begin{equation} \label{eq:permref}
T\Bigl(\widehat{v}_{j+1}^{(\sam)},\widehat{y}_{j+1}^{(\sam)},\yd_{j+1},
\{\widehat{v}_{j+1}^{(\ell)},\widehat{y}_{j+1}^{(\ell)}\}_{\ell=1}^\Sam;
\theta\Bigr)=T\Bigl(\widehat{v}_{j+1}^{(\sam)},\widehat{y}_{j+1}^{(\sam)},
\yd_{j+1},\frac{1}{N} \sum_{\ell=1}^\Sam 
\delta_{(\widehat{v}_{j+1}^{(\ell)},\widehat{y}_{j+1}^{(\ell)})};\theta\Bigr);
\end{equation}
thus, the map depends on the joint empirical measure \eqref{eq:empEnKF9889} 
of the states and observations.

\begin{remark}
\label{rem:nonstoch} 
It may also be of interest to work with a non-stochastic form of generalized
ensemble Kalman filter. We describe several examples of this, adopting the
idea illustrated in \eqref{eq:permref} to enforce ensemble permutation invariance.
For instance we might modify \eqref{eq:fakeobs2} and consider 
\begin{subequations}
\label{eq:nofakeobs0}
\begin{align}
\widehat{v}_{j+1}^{(\sam)} &= \Psi(v_{j}^{(\sam)})+\xi^{(\sam)}_{j}, \quad \sam=1,\ldots,\Sam, \\
v_{j+1}^{(\sam)} &= T\Bigl(\widehat{v}_{j+1}^{(\sam)},h\bigl(\widehat{v}_{j+1}^{(\sam)}\bigr),
\yd_{j+1}, \sum_{\ell=1}^\Sam 
\delta_{\bigl(\widehat{v}_{j+1}^{(\ell)},h(\widehat{v}_{j+1}^{(\ell)})\bigr)};
\theta\Bigr),\quad \sam=1,\ldots,\Sam.
\end{align}
\end{subequations}
The noise defining this random dynamical system again satisfies Assumption \ref{a:enkf-noise}.
The following simpler special case of \eqref{eq:nofakeobs0} is also useful for expository purposes:
\begin{subequations}
\label{eq:nofakeobs}
\begin{align}
\widehat{v}_{j+1}^{(\sam)} &= \Psi(v_{j}^{(\sam)})+\xi^{(\sam)}_{j}, \quad \sam=1,\ldots,\Sam, \\
v_{j+1}^{(\sam)} &= T\Bigl(\widehat{v}_{j+1}^{(\sam)},\yd_{j+1},
\sum_{\ell=1}^\Sam 
\delta_{\widehat{v}_{j+1}^{(\ell)}};
\theta\Bigr),\quad \sam=1,\ldots,\Sam. 
\end{align}
\end{subequations}
\end{remark}

\begin{remark}
\label{rem:eqwe}
We refer to an algorithm such as \eqref{eq:fakeobs2}, \eqref{eq:nofakeobs0} or \eqref{eq:nofakeobs}, leading to a candidate approximation of the filtering distribution in the form \eqref{eq:empEnKF99}, as an \emph{equal weights ensemble filter}\index{particle filter! equal weights}. Recall from Subsection \ref{ssec:collapse} that weighted particle filters, such as the 
bootstrap\index{particle filter!bootstrap}
and optimal particle\index{particle filter!optimal} filters, are prone to weight collapse\index{weight collapse}; this motivates the wish to find equally weighted filters.
\end{remark}

\begin{remark} \label{rem:amor3}
The proposed learnable algorithms \eqref{eq:pred3DVAR2add}, \eqref{eq:fakeobs2}, \eqref{eq:nofakeobs0}
and \eqref{eq:nofakeobs} all allow for the possibility of learning dependence on the observed data 
$\yd_{j+1}$ at each analysis\index{analysis} step. Learning such dependence is feasible
when the stochastic dynamical system defined by coupling the underlying signal\index{signal} dynamics to
the filtering algorithm, through the observations,\index{observations} is ergodic.\index{ergodic}
(See Subsection \ref{ssec:ergf} for discussion of ergodicity for filtering algorithms.)
This is feasible because of a form of amortization
over time, exploiting ergodicity present in the filtering problem.
In Chapter \ref{ch:LG10} we go beyond amortization for filtering through multiple
time indices, and consider use of multiple observation \emph{sequences} $\Yd$ which can
enhance learning of the dependence of each analysis step on $\yd_{j+1}$. The discussion in this
remark is closely related to that for variational inference in Remark \ref{rem:amor2}.
\end{remark}

We may choose $\theta$ either so that the resulting algorithm performs well at
state estimation\index{state estimation} or so that it 
matches the filtering distribution\index{filtering distribution}.
We may think of the map $T$ as a transport\index{transport!map} map, 
as introduced in Section~\ref{sec:transport2}. This nomenclature
is most commonly used in the mathematics literature to describe maps
that, through the pushforward operation, map one probability measure into another.
We will use this terminology, however, in the context of both state
estimation and filtering. We think of the map $T$ as transporting states,
predicted by the underlying stochastic dynamical system\index{dynamical system!stochastic}, so that they match the data.  
In Section~\ref{sec:selo} we describe a variety of
learning objectives that may be used so that this transport performs well at state estimation. Section~\ref{sec:sea}
then deploys these objectives to learn optimal parameters in specific
problems classes of the form \eqref{eq:pred3DVAR2add} or \eqref{eq:fakeobs2}, and their generalizations, such as \eqref{eq:nofakeobs0} and \eqref{eq:nofakeobs}, for the
purposes of state estimation.\index{state estimation}
In Section~\ref{sec:flo} we describe various objectives
that may be used so that these transport algorithms perform well at matching the filtering distribution -- probabilistic estimation.\index{probabilistic estimation}
Section \ref{sec:fa} describes algorithm classes that can be learned using 
these objectives defined for probabilistic estimation; we
concentrate on ensemble-based methods. 
Finally, Section \ref{sec:Bib_learning_filters} 
contains bibliographical remarks.

\section{State Estimation: Learning Objectives}\label{sec:selo}
This section defines a variety of learning objectives that are relevant
for state estimation. We work primarily under the following data assumption
concerning a data stream in $\R^{\du} \times \R^{\dy}$.
\begin{dataassumption}\index{Data Assumption}
\label{da:9d}
We have access to the data set $\{\vd_{j},\yd_{j}\}_{j=1}^{J}$, 
generated as a realization of \eqref{eq:ddm99}. 
\end{dataassumption}

Notice that here $\{\vd_j,\yd_{j}\}_{j=1}^{J}$ represents \emph{training data}\index{training data} used to learn the parameters of filtering algorithms. The interpretation of these sequences is therefore slightly different than in previous chapters, in which  $\{\yd_{j}\}_{j=1}^{J}$ represents given observations used to estimate unknown signal $\{\vd_{j}\}_{j=0}^{J}.$ 
In this chapter we will use training data from Data Assumption \ref{da:9d} to learn filtering algorithms; these algorithms can be deployed on a new set of observations to estimate the unknown signal that underlies them. The training data from Data Assumption \ref{da:9d} 
enables us, potentially, to learn good choices of the parameters defining the filters. 
Recall that we discussed training data\index{training data} in the context of 
amortized inference,\index{amortized inference} 
for the Bayesian inverse problem, in Chapter \ref{ch:data-dependence}. 
In the next remark we discuss the sense
in which amortization\index{amortization} applies in this setting where we simply have access
to a single realization from the joint state and observation model.

\begin{remark}
\label{rem:slvam}
If the filter is viewed as an architecture that maps the entire sequence of observations 
$\Yd$ to the entire sequence of states $\Vd$, while respecting the causal structure of these sequences in time, then utilizing Data Assumption \ref{da:9d} could be viewed as solving a supervised learning\index{supervised learning} problem with only one data pair, a data-poor learning environment. However, embedded in the solution to this problem is the learning of a single transport map that uses the observation at each step $j$ to perform filtering sequentially: the analysis\index{analysis} step
from \eqref{eq:pna}. Since the methodology uses only one element of the observation
sequence at a time then if $J$ is large we may view this as a data-rich learning environment for this single transport map. For this to indeed be a data-rich learning environment requires some form
of statistical similarity in the transport problem, independently of $j$; then
for large $J$ the learning problem will indeed be data rich. This data rich environment may arise, for instance, under ergodicity of the signal and observation processes; we will discuss ergodicity of filtering algorithms in 
Remark \ref{rem:ergo} and in Subsection \ref{ssec:ergf}.
\end{remark}

As an alternative to Data Assumption \ref{da:9d} we may simply assume that we have access to data 
$\{\yd_j\}_{j=1}^J$ and that this data is related to an unobserved signal process $\{\vd_j\}$ with unknown dynamics. Because the dynamics (\ref{eq:ddm99}a) are unknown we do not have access to 
$\{\vd_{j}\}_{j=0}^J$. In this setting we will use the observation operator $h:\R^{\du} \to \R^{\dy}$ to map the states into the observation space, to define a loss from which to learn parameters of the filtering algorithm. In this setting we work under the following:

\begin{dataassumption}\index{Data Assumption}
\label{da:9e}
We have access to data set $\{\yd_{j}\}_{j=1}^J$,
generated as a realization of \eqref{eq:ddm99}. 
\end{dataassumption}

The remainder of this section is organized as follows. Subsection~\ref{ssec:mts} focuses on state estimation using deterministic 
scoring rules.\index{scoring rule!deterministic}
Subsection~\ref{ssec:sru} goes beyond
this setting by working with probabilistic scoring rules\index{scoring rule!probabilistic} 
that compare the distribution defined by an ensemble algorithm with the true 
state. In both Subsection~\ref{ssec:mts} and Subsection~\ref{ssec:sru} we employ Data Assumption~\ref{da:9d}.  In Subsection \ref{ssec:lme} we discuss objective functions defined \index{covariance} in observation space rather than state space; this is useful when model error is significant and we work under Data Assumption~\ref{da:9e}.

\subsection{Matching the State: Deterministic Scoring Rule}
\label{ssec:mts}

Consider an algorithm, such as \eqref{eq:pred3DVAR2add} or \eqref{eq:fakeobs2}, which gives rise to an estimator $v_j=v_j(\theta)$ for $\vd_j$; for example $v_j(\theta)$ may be $v_j$ itself for algorithm~\eqref{eq:pred3DVAR2add} or the ensemble average $v_j$ in~\eqref{eq:SE}, for algorithm~\eqref{eq:fakeobs2}. Given Data Assumption \ref{da:9d} we may then apply a deterministic scoring rule $\dd(\cdot, \cdot)$, satisfying Definition~\ref{def:det_score}, to measure the distance between
sequences $\{\vd_j\}_{j=1}^J$ and $\{v_j(\theta)\}_{j=1}^J$. An optimal $\theta^\star$ may then
be chosen to minimize this distance:
\begin{subequations}\label{eq:state_estimation_obj}
    \begin{align}
        \J^J(\theta) &= \frac{1}{J}\sum_{j=1}^{J} \dd\bigl(v_j(\theta), \vd_j\bigr),\\
        \theta^\star&\in \argmin_{\theta\in\Theta} \J^J(\theta).
    \end{align}
\end{subequations}
Note that $v_j(\theta)$ will, in order to respect the causal structure of filtering, depend on the observed data $\Yd_j=\{\yd_{i}\}_{i=1}^j$; thus when run to $J$ steps the state estimator
has seen the entire observation sequence $\Yd=\{\yd_{j}\}_{j=1}^{J}$. Consequently $\theta^\star$ will, through the optimization procedure, depend on 
the entirety of the data specified in Data Assumption~\ref{da:9d}. The optimization problem defined by \eqref{eq:state_estimation_obj} may be performed using auto-differentiation with respect to $\theta$\index{auto-differentiation}; see Section \ref{sec:auto-differentiation} for details on this methodology. If $J$ is large we may exploit
some form of ergodicity\index{ergodic} to learn an algorithm which can be deployed 
with different observed data streams.

\begin{example}\label{ex:sqeu}
The canonical example of a distance-like deterministic scoring 
rule\index{scoring rule!distance-like deterministic} 
is $\dd(v,w)=|v-w|^2$, the squared Euclidean distance. If this choice is made, 
then (\ref{eq:state_estimation_obj}a) is called the \emph{mean-square error}, 
and its square root the \emph{root-mean-square error} (RMSE).\index{root-mean-square error (RMSE)}
See Remark \ref{rem:rmse}, and the material preceding
it, connecting the squared Euclidean distance to probabilistic scoring rules.\index{scoring rule!probabilistic}
\end{example}

\begin{remark}
In Subsection~\ref{ssec:squared_error_min} we show how working with an \emph{amortized}\index{amortization} version of~\eqref{eq:state_estimation_obj}, in the specific case of the squared Euclidean distance and averaged over multiple realizations of the state and data,
leads to an objective function that is minimized at the true filter mean.
\end{remark}

\subsection{Matching the State: Probabilistic Scoring Rule}
\label{ssec:sru}

We now go beyond simply matching a state estimator (from an algorithm) with the true state of the system (from (\ref{eq:ddm99}a)). Instead, we look at algorithms that output a probability distribution over states, and use a scoring rule\index{scoring rule} to measure
the distance of that distribution from the true state of the system. 
Assume that the algorithm class of interest produces an estimator of the filtering
distribution $\pia_j(\theta) \approx \pi_j.$ 
An example of such an approximation is $\pia_j(\theta)$ defined by \eqref{eq:fakeobs2}, \eqref{eq:empEnKF99}. We aim to choose $\theta$ so as to minimize the score
between $\pia_j(\theta)$ and the true signal $\vd_j$ underlying the data.

To this end, we employ a probabilistic scoring rule\index{scoring rule!probabilistic} $\mS(\cdot, \cdot)$, 
introduced in Section \ref{sec:psr}, to obtain
\begin{subequations}\label{eq:state_estimation_obj3}
    \begin{align}
        \J^J(\theta) &= \frac{1}{J}\sum_{j=1}^{J} \mS(\pia_j(\theta), \vd_j),\\
        \theta^\star&\in \argmin_{\theta\in\Theta} \J^J(\theta).
    \end{align}
\end{subequations}
This may be implemented under Data Assumption \ref{da:9d}.
We emphasize that it is important, in order to be able to implement this
optimization problem, to use scoring rules which can be evaluated when $\pia_j(\theta)$ is given as an empirical measure\index{empirical measure}. Scoring rules with this
property include the energy score\index{score!energy}, from Definition \ref{d:energy}.
On the other hand, the logarithmic score\index{score!logarithmic}, from Definition
\ref{d:logs}, cannot be used with
empirical measures. In summary, the optimization problem in~\eqref{eq:state_estimation_obj3} can be implemented with ensemble-based methods if the scoring rule is chosen appropriately.

\begin{remark} \label{rem:SPSR}
In Section~\ref{sec:SPSR} we show how working with an amortized\index{amortization} version of~\eqref{eq:state_estimation_obj3}, averaged over multiple realizations of the state and data,
leads to objective functions that are minimized at the true filtering distribution.
\end{remark}

\subsection{Matching the Data}
\label{ssec:lme}

We now assume access only to data 
$\{\yd_j\}_{j=1}^J$, a natural setting when the dynamics (\ref{eq:ddm99}a) are unknown 
so we do not have access to  $\{\vd_{j}\}_{j=0}^J$. We work under Data Assumption \ref{da:9e} and
consider the following objective function defined in the observation space:
\begin{equation}\label{eq:lossnewref}
    \J^J(\theta) = \frac{1}{J}\sum_{j=1}^{J} \dd \Bigl(h\bigl(v_j(\theta)\bigr), \yd_j\Bigr).
\end{equation}
Once again we typically choose $v_j(\theta)$ to be $v_j$ itself, given by \eqref{eq:pred3DVAR2add}, or the ensemble average given by \eqref{eq:fakeobs2}, \eqref{eq:SE}.

We note that $\J^J(\theta)$ is then a proxy for the loss with respect to the true trajectory in the observation space:
\begin{equation}\label{eq:dist_true_traj_obs_space}
    \frac{1}{J}\sum_{j=1}^{J} \dd\Bigl(h\bigl(v_j(\theta)\bigr), h(\vd_j)\Bigr).
\end{equation}
The difference between the two losses is that~\eqref{eq:lossnewref} contains observational noise in the observation, which may be substantial, and there is a risk to overfitting to this noise. The following remark addresses this issue.

\begin{remark} 
Optimizing the cost function $\J^J(\theta)$ in \eqref{eq:lossnewref} does not always lead to good performance. The root cause is that the objective function can overfit to the specific observations, which contain unknown observational errors. For example, if $h$ is uniquely invertible then a perfect score can be achieved by simply setting $v_j(\theta) = h^{-1}(\yd_j)$. When the signal-to-noise ratio is small, 
this will lead to a very noisy estimate $v_j(\theta)$ of the signal. If only a single realization of the observed process is available, then this is a fundamental obstacle. However, if it is possible to obtain multiple independent realizations, 
then the overfitting issue can be mitigated. In this setting, we describe a method based on this observation to avoid overfitting. We take the squared Euclidean distance $\dd(u_1,u_2) = |u_1 - u_2|^2$  as the scoring rule and analyze a single time $j$. Similar analyses could be performed for other scoring rules and multiple times.
 
Recall that $\yd_j = h(v_j^\dagger) + \eta_j^\dagger$.
Note that the data assimilation algorithm produces an estimator $v_j(\theta)$
for $\vd_j$ which will depend on $\yd_j$ and hence on $\etad_j.$ We consider the contribution to the
objective function in~\eqref{eq:lossnewref} at $j$ and take expectation
$\mathbb{E}$ over independent realizations of the observation noise $\eta^\dagger_j\sim\cN(0, \Gamma)$. This yields
\begin{align*} 
    \mathbb{E} \bigl|h\bigl(v_j(\theta)\bigr) - \yd_j \bigr|^2 = \mathbb{E} \bigl|h(v_j(\theta)) - h(\vd_j) \bigr|^2 + \Tr(\Gamma) - 2\mathbb{E}\Bigl[h\bigl(v_j(\theta)\bigr)^\top \eta^\dagger_j\Bigr].
\end{align*}
Rearranging we obtain
\begin{align} \label{eq:unbiased_objective}
    \mathbb{E} \bigl|h(v_j(\theta)) - h(\vd_j) \bigr|^2 + \Tr(\Gamma)=
    \mathbb{E} \bigl|h\bigl(v_j(\theta)\bigr) - \yd_j \bigr|^2 +2\mathbb{E}\Bigl[h\bigl(v_j(\theta)\bigr)^\top \eta^\dagger_j\Bigr].
\end{align}
Minimizing the left-hand side is desirable as it matches the output of the
analysis, $v_j(\theta)$, with the true signal, $\vd_j,$ in observation
space; note that it does
not depend on the observational noise explicitly, only through the
dependence of $v_j(\theta)$ on the data. The right-hand side
can be approximated through sampling the observational noise $\etad_j$, 
and then optimizing  over $\theta$ to approximately minimize the left-hand side.\footnote{The term $\mathbb{E}\Bigl[h(v_j(\theta))^\top \eta^\dagger_j\Bigr]$ is 
sometimes referred to as the \emph{optimism}\index{optimism}.}

The natural setting in which the idea explained here may be used is when
Data Assumption~\ref{da:9e} is generalized to multiple realizations of an
observation stream defined over $j \in \{1,\ldots, J\}$.
\end{remark}

\section{State Estimation\index{algorithm!state estimation}: 
Algorithms}\label{sec:sea}

Our interest in this section is in learning transport-based\index{transport} 
filtering algorithms for state estimation.\index{algorithm!state estimation}
We study learning within the context of specific forms of
 algorithms \eqref{eq:pred3DVAR2add} and \eqref{eq:fakeobs2}, or their generalizations. In particular, we consider algorithms designed by modifying three specific algorithms from Chapter \ref{lecture7}: 3DVar, the EnKF, and the optimal particle filter. We work under
Data Assumption \ref{da:9d}. Subsection \ref{ssec:3gain} is devoted to learning the
gain\index{gain!3DVar} in 3DVar\index{3DVar}. In Subsection \ref{ssec:enkf_gain_learning} we study the same question,
but in the context of generalizations of the EnKF\index{Kalman filter!ensemble}. Subsection~\ref{ssec:loci} 
also studies generalizations of the EnKF, but focuses on learning localization and
inflation parameters. In Subsection~\ref{ssec:opflf} we go beyond ensemble algorithm classes of the form in \eqref{eq:fakeobs2},
showing how the optimal particle filter\index{particle filter!optimal} 
may be used as the basis to learn new
algorithms for equally weighted ensemble methods.

\subsection{Learning the Gain\index{gain!3DVar} in 3DVar}
\label{ssec:3gain}

Recall the 3DVar\index{3DVar} algorithm \eqref{eq:pred3DVAR}, repeated here for convenience:
\begin{subequations}
 \label{eq:pred3DVAR2}   
\begin{align}
\hat{v}_{j+1} &= \Psi(v_j),\\
v_{j+1} &= \hat{v}_{j+1} + K \bigl(\yd_{j+1}-h(\hat{v}_{j+1})\bigr).	
\end{align}
\end{subequations}
This is a specific subclass of the algorithm class defined in \eqref{eq:pred3DVAR2add}. To fully specify the 3DVar\index{3DVar} algorithm we need to choose the gain\index{gain!3DVar} matrix $K$. We investigate learning $\theta := K$ on the basis of Data Assumption \ref{da:9d}. (We note that, alternatively, $K$ could be parameterized as $K = K(\theta)$, and the parameter $\theta$ could be learned, but we do not pursue this here.) For the dynamics/data model underlying Data Assumption \ref{da:9d} and for the 3DVar algorithm \eqref{eq:pred3DVAR2}, we make the same Gaussian and independence assumptions on the initialization and noise as made in equations \eqref{eq:sdm}, \eqref{eq:dm} and Assumption \ref{a:fas}. In particular, we assume that $\vd_0$ and $v_0$
are drawn from the same distribution $\cN(m_0,C_0)$, but independently.

\subsubsection{Linear $h$: The Original Derivation of 3DVar}

Consider the linear setting where $h(v)=Hv$ for some matrix $H \in \R^{\dy \times \du}$. Motivated by the form of the Kalman gain\index{gain!Kalman} itself, and equations~\eqref{eq:Kalman_recursions} in particular, but
invoking a steady-state hypothesis on the dynamics, we seek $K$ in the form 
\begin{equation}
\label{eq:udcr}
K = \hat{C} H^\top \bigl(H \hat{C} H^\top + \Gamma\bigr)^{-1},
\end{equation}
where $\hat{C}$ is a time-independent state covariance. 
See Subsection~\ref{ssec:3dvar} for details of the
derivation of this relationship between $K$ and $\hat{C}.$
In the original derivation of 3DVar\index{3DVar}, $\hat{C}$ itself is estimated  and $K$ is formed using \eqref{eq:udcr}. Estimation of $\hat{C}$ is achieved using data relating to
the underlying dynamics model (\ref{eq:ddm99}a), and estimates of the covariance in forecasts made by this model. However, in the remainder of this subsection we adopt a different approach, aiming to determine $K$ directly,
under Data Assumption \ref{da:9d}.

\subsubsection{General $\Psi$ and $h$: A Machine Learning Approach to 3DVar}

We now move away from the assumption that $H$ is linear and we consider general $\Psi$ and $h$. We note that $v_j$ arising from algorithm class \eqref{eq:pred3DVAR2} depends on the gain\index{gain} matrix: $v_j=v_j(K).$ We use the loss function in \eqref{eq:state_estimation_obj} 
to determine the optimal choice of $\theta$: 
\begin{align*}
        \J^J(K) &= \frac{1}{J}\sum_{j=1}^{J} \dd\bigl(v_j(K), \vd_j\bigr),\\
        K^\star&\in \argmin_{K \in\R^{\du \times \dy}} \J^J(K).
\end{align*}

\begin{remark} \label{rem:ergo}
In the learning frameworks of Chapter \ref{ch:transport} we introduced the
use of transport maps to solve inverse problems. There we started from the population loss, defined as an expectation over a measure with Lebesgue density, and then noted that in practice we approximate this empirically\index{empirical}.
Here, because the analog of the population loss is more complicated to write down, we work the other way around: we have started with empirical\index{empirical} loss \eqref{eq:state_estimation_obj} defined by a time series of length $J$; and now proceed to derive the analog of a population 
loss by letting $J \to \infty.$ To this end, let us view \eqref{eq:ddm99}, \eqref{eq:pred3DVAR2} as defining a coupled stochastic dynamical system for $(\vd_j,v_j)$; this is obtained by substituting the expression for $\yd_{j+1}$ into that for $v_{j+1}$. 
The resulting stochastic dynamical system has the form 
\begin{subequations}
\label{eq:pred3DVAR22}   
\begin{align}
\vd_{j+1} &= \Psi(\vd_j) + \xid_j,\\
v_{j+1} &= \Psi(v_j) + K \Bigl(h\bigl(\vd_{j+1}\bigr)-h\bigl(\Psi(v_j)\bigr)+ \etad_{j+1}\Bigr).	
\end{align}
\end{subequations}
Recall the discussion of ergodicity in Section \ref{sec:ERG}, and in particular in Remark \ref{rem:erg-stoc}.  We assume that the dynamical system \eqref{eq:pred3DVAR22} is ergodic\index{ergodic} on $\R^d \times \R^d$,  with invariant measure $\mu(d\vd,dv;K).$  We may then view $\J^J(\cdot)$ as approximation of the objective function  $\J(\cdot)$
    \begin{equation*}
        \J(K)=\int_{\R^d \times \R^d} \dd\bigl(v(K), \vd\bigr)\, \mu(d\vd,dv;K).
    \end{equation*}
By ergodicity, for any $K,$
\begin{equation*}
        \J(K)=\lim_{J \to \infty} \J^J(K).
\end{equation*}
The objective function  $\J(\cdot)$ is the analog of the population loss in this setting.
This ergodicity\index{ergodic} observation also suggests that, rather than using one long trajectory, Data Assumption~\ref{da:9d} could be modified to deploy a set of independently generated dynamics--data pairs from \eqref{eq:ddm99}, rather than a single one. 
\end{remark}

\subsubsection{Linear $\Psi$ and $h$: Expected Squared Error Loss Function for 3DVar}

We now consider using an expected form of \eqref{eq:state_estimation_obj}, with the deterministic scoring rule chosen to be the squared error, to determine the gain\index{gain!3DVar} $K$. In this case, the expectation can be evaluated in closed form, leading to an implementable cost function.
We show that, in the setting of linear dynamics and observations, the learned gain $K$
in the 3DVar algorithm \eqref{eq:pred3DVAR2} converges to the steady-state Kalman 
gain\index{gain!steady state}\index{gain!Kalman} as $J\to\infty$. To this end, consider the 3DVar algorithm~\eqref{eq:pred3DVAR2} 
with $\Psi(\cdot) = A\cdot$ and $h(\cdot) = H\cdot$. Then \eqref{eq:ddm99}, 
\eqref{eq:pred3DVAR2}  show that $(\vd_j,v_j)$ evolve according to the recursion
\begin{subequations}\label{eq:kalman_fixed_mean}
    \begin{align}
    \vd_{j+1} &= A\vd_{j}+\xid_j, \\
        \hat{v}_{j+1} &= A v_j, \\
        v_{j+1} &= (I - KH)\hat{v}_{j+1} + K\bigl(H\vd_{j+1}+\etad_{j+1}\bigr).
    \end{align}
\end{subequations}
Define the error covariances $\hat{C}_j = \mathbb{E}\bigl[(\vd_j - \hat{v}_j)\otimes(\vd_j - \hat{v}_j) \bigr]$  and $C_j = \mathbb{E}\bigl[(\vd_j - v_j)\otimes(\vd_j - v_j)\bigr]$, which we note are not the covariances of the forecast and filtering distributions. These covariances obey the recursions
    \begin{subequations}\label{eq:kalman_fixed_cov}
        \begin{align}
            \hat{C}_{j + 1} &= A C_{j} A^\top + \Sigma, \\
            C_{j+1} &= (I - KH)\hat{C}_{j+1}(I-KH)^\top + K\Gamma K^\top.
        \end{align}
    \end{subequations}
Identity (\ref{eq:kalman_fixed_cov}a) holds because
   \begin{align*}
        \hat{C}_{j + 1} &= \mathbb{E}   \Bigl[\bigl(\vd_{j+1} - \hat{v}_{j+1}\bigr)\otimes \bigl(\vd_{j+1} - \hat{v}_{j+1}\bigr)\Bigr]\\
        &= \mathbb{E}\Bigl[\bigl(A\vd_{j} + \xid_j - A{v}_{j} \bigr)\otimes \bigl(A\vd_{j} + \xid_j - A{v}_{j}\bigr) \Bigr]\\
        &= \mathbb{E}\Bigl[\bigl(A(\vd_{j} - {v}_{j})\bigr)\otimes \bigl(A(\vd_{j} - {v}_{j})\bigr)\Bigr] + \Sigma\\
        &= AC_jA^\top + \Sigma.
    \end{align*}
Identity (\ref{eq:kalman_fixed_cov}b) holds because, from (\ref{eq:kalman_fixed_mean}c),
    $$v_{j+1}-\vd_{j+1} = (I - KH)\bigl(\hat{v}_{j+1}-\vd_{j+1}\bigr)+K\etad_{j+1}.$$

In the following theorem, we show that the minimizer of an expected squared error objective function, 
with $\theta:=K$, is the steady-state Kalman gain\index{gain!steady state}\index{gain!Kalman}\index{Kalman filter!steady state} found from the Kalman filter in the limit $J\to\infty$.  
We will use the preceding covariances and also use the definitions of the  covariance 
and gain arising from the Kalman filter itself, given in~\eqref{eq:Kalman_recursions},
with a constant-in-time dynamics and observation operators $(A,H).$
Here we recall the update formulae for the Kalman covariance 
and the Kalman gain for notational convenience, introducing a new notation to distinguish from that in \eqref{eq:kalman_fixed_mean}:
\begin{subequations}
\label{eq:ofgs}
\begin{align}
    \hat{C}^{\sf{kalman}}_{j+1} &= A C^{\sf{kalman}}_{j} A^\top + \Sigma, \\
    K^{\sf{kalman}}_{j+1} &= \hat{C}_{j+1}^{\sf{kalman}} H^\top \bigl(H \hat{C}_{j+1}^{\sf{kalman}} H^\top + \Gamma\bigr)^{-1}, \\
    C^{\sf{kalman}}_{j+1} &=  (I - K^{\sf{kalman}}_{j+1} H) \hat{C}^{\sf{kalman}}_{j+1}.
\end{align}
\end{subequations}
Note the resemblance of the covariance update equations \eqref{eq:kalman_fixed_cov} to those of \eqref{eq:ofgs}; in fact, (\ref{eq:kalman_fixed_cov}b) reduces to (\ref{eq:ofgs}c) if the gain $K$ is chosen to be the Kalman gain (\ref{eq:ofgs}b).
    
\begin{theorem} \label{thm:sc}
    Consider the 3DVar algorithm \eqref{eq:pred3DVAR2} with $\Psi(\cdot) = A\cdot$ and $h(\cdot) = H\cdot$. Assume that the analysis covariance and gain of the Kalman filter, given by \eqref{eq:ofgs},
    converge to their steady states in \eqref{eq:ss_kalman}, \eqref{eq:ss_kalman2}:
    $$C_j^{\sf{kalman}} \rightarrow C_\infty, \qquad K_j^{\sf{kalman}} \rightarrow K_\infty, \qquad j \rightarrow \infty.$$
    For $K \in \R^{\du \times \dy}$ define $v_j(K)$ via the 
    recursion \eqref{eq:kalman_fixed_mean} and define 
    \begin{equation*}
        \J(K) := \limsup_{J \to \infty } \frac{1}{J} \sum_{j=1}^J \Expect[|\vd_j - v_j(K)|^2|\Yd_j]. 
    \end{equation*}
    Equivalently, defining $C_j=C_j(K)$ by the recursion \eqref{eq:kalman_fixed_cov}, we have a closed form expression for the expectation,
    \begin{equation*}
        \J(K) := \limsup_{J \to \infty } \frac{1}{J} \sum_{j=1}^J \Tr \bigl(C_j(K)\bigr). 
    \end{equation*}
    
    Then, the minimizer of $\J$ is the steady-state Kalman 
    gain\index{gain!steady state}\index{gain!Kalman}\index{Kalman filter!steady-state}
    in~\eqref{eq:ss_kalman_gain}, i.e.,
    $$K_\infty \in \argmin_{K\in\R^{\du \times \dy}} \J(K).$$
\end{theorem}

\begin{proof}
    Consider the following recursion for a state estimator $v_j$, using a time-varying gain\index{gain!time-varying} $K_{j+1}$, taking the form
    \begin{subequations}\label{eq:kalman_rec}
        \begin{align}
            v_{j+1} &= (I - K_{j+1}H)A v_j + K_{j+1} \yd_{j+1},\\
            \hat{C}_{j + 1} &= A C_{j} A^\top + \Sigma,\\
            C_{j+1} &= (I - K_{j+1} H)\hat{C}_{j+1}(I-K_{j+1}H)^\top + K_{j+1}\Gamma K_{j+1}^\top.\label{eq:analysis_cov}
        \end{align}
    \end{subequations}
    Next note that the update for the state in~\eqref{eq:kalman_fixed_mean} and the covariance in~\eqref{eq:kalman_fixed_cov} can be written in this form with
    $K_{j+1}=K.$

    We now use the fact that the Kalman filter provides the exact filtering distribution for this system. Therefore the Kalman filter mean $v_j = \mathbb{E}[\vd_j|\Yd_j]$, being the conditional expectation, provides the minimum mean-squared error estimate of the state at each step; see Theorem~\ref{th:postmean}. That is, let $v_j$ be the sequence given by the Kalman filter mean in~\eqref{eq:kalman_mean}. Then, for any measurable function $z_j$ of the observations $\Yd_j$ we have
$$\mathbb{E}\left[|v_j^\dagger - z_j|^2\,|\,\Yd_j \right] \geq \mathbb{E}\left[|v_j^\dagger - v_j|^2\,|\,\Yd_j \right].$$
The right-hand side corresponds to the trace of the error covariance $C_j(K_j^{\sf{kalman}})$, thereby showing that the Kalman filter gain minimizes the trace of the covariance among all gains $K$; that is,
    \begin{align} \label{eq:trace_lower_bound_stepwise}
        \Tr \bigl(C_{j+1}(K)\bigr) \geq \Tr \bigl(C_{j+1}(K^{\sf{kalman}}_{j+1})\bigr)
    \end{align}
    for all $j$. From the assumption that the  Kalman filter covariance and gain converge to their steady states $C_\infty$ and $K_\infty$, respectively, and the continuity of the trace operator, we have
    $$\Tr \bigl(C_{j+1}(K^{\sf{kalman}}_{j+1}) \bigr) \rightarrow \Tr(C_{\infty}), \qquad j \rightarrow \infty.$$
    Therefore, by the Stolz--Ces\`aro-mean theorem (see bibliography), the average of the trace of the analysis covariances over the sequence also converges. That is,
    \begin{equation} \label{eq:limit_convergence}
    \lim_{j \rightarrow \infty} \frac{1}{J} \sum_{j=1}^J \Tr \bigl(C_{j+1}(K^{\sf{kalman}}_{j+1})\bigr) \rightarrow \Tr(C_{\infty}).
    \end{equation}
    Lastly, averaging~\eqref{eq:trace_lower_bound_stepwise} over $J$ steps and taking the limit $J \rightarrow \infty$ we have  
    \begin{equation} \label{eq:objective_lower_bound}
    \liminf_{J \to\infty} \frac{1}{J} \sum_{j=1}^{J}\Tr \bigl(C_{j}(K)\bigr) \geq \lim_{J \to\infty} \frac{1}{J} \sum_{j=1}^{J}\Tr \bigl(C_{j}(K_{j}^{\sf{kalman}}) \bigr) = \Tr(C_\infty),
    \end{equation}
    where we have used the convergence of the limit in~\eqref{eq:limit_convergence}. The inequality in~\eqref{eq:objective_lower_bound} holds for all $K$ and the lower bound is attained by $K = K_\infty.$ From~\eqref{eq:objective_lower_bound}, we have $\J(K) \geq \J(K_\infty)$, and hence the minimum value of the objective is attained with choice $K=K_\infty$.
\end{proof}

\subsection{Learning the Gain\index{gain!EnKF} in EnKF}\label{ssec:enkf_gain_learning}
The ensemble Kalman filter\index{Kalman filter!ensemble}
was designed as a Monte Carlo\index{Monte Carlo} method, 
in which the ensemble is viewed as providing an
approximation of the filtering distribution. However it is often used as a state estimator, with the ensemble mean providing the estimate of the state.
This is the perspective we adopt in this subsection. We continue to work under
Data Assumption \ref{da:9d}
and recall the EnKF algorithm \eqref{eq:fakeobs}\index{Kalman filter!ensemble}. Rather than calculating $K_{j+1}$ from empirical covariances as in \eqref{eq:gainEnKF}, we instead try and learn dependence on the ensemble. To this end we modify \eqref{eq:fakeobs} to read
\begin{subequations}
\label{eq:predEnKF2}
\begin{align}
\widehat{v}_{j+1}^{(\sam)} &= \Psi(v_{j}^{(\sam)})+\xi^{(\sam)}_{j}, \quad \sam=1,\ldots,\Sam, \\
\widehat{y}_{j+1}^{(\sam)} &= h(\hv_{j+1}^{(\sam)})+\eta^{(\sam)}_{j}, \quad \sam=1,\ldots,\Sam, \\
v_{j+1}^{(\sam)} &= \widehat{v}_{j+1}^{(\sam)}+K_{j+1}\bigl(\yd_{j+1}-\widehat{y}_{j+1}^{(\sam)}\bigr),\quad \sam=1,\ldots,\Sam,\\
K_{j+1}&=\ks\bigl(\{\widehat{v}_{j+1}^{(\ell)},\widehat{y}_{j+1}^{(\ell)}\}_{\ell=1}^\Sam;\theta\bigr), \label{eq:parameterized_gain_3DVar}
\end{align}
\end{subequations}
where $\xi_{j}^{(\sam)} \sim \Nc(0,\Sigma)$, $\eta_{j+1}^{(\sam)} \sim \Nc(0,\Gamma)$
are independent sequences of i.i.d.\thinspace random vectors with respect to both $j$ and $n$, and
the two sequences themselves are independent of one another.
This is then a specific subclass of the algorithm class defined in \eqref{eq:fakeobs2}.
It is useful to define a state estimator by taking the mean of 
the ensemble at time $j$ as defined in \eqref{eq:SE}. We may then use 
the loss function in \eqref{eq:state_estimation_obj} to determine $\theta,$
with $v_j(\theta)$ given by the ensemble mean in~\eqref{eq:SE}.

As explained in the text preceding  \eqref{eq:permref} it is desirable that $\ks(\cdot \, ;\theta)$ be invariant with respect to permutation of the ensemble; this may be achieved by viewing the dependence on
the ensemble as being defined through allowing dependence only on the empirical measure \eqref{eq:empEnKF9889}. The following example highlights a specific way of achieving this.

\begin{example}
\label{ex:thetaform}
It may be of interest to replace~\eqref{eq:parameterized_gain_3DVar} by
$K_{j+1}=\ks\bigl(\widehat{C}_{j+1}^{vh},\widehat{C}_{j+1}^{yy};\theta\bigr),$
where the covariance matrices are as defined in \eqref{eq:gainEnKF} and
$\ks(\cdot \, ;\theta)$ is a parameterized family of gain\index{gain!parameterized} functions to be learned by optimizing over $\theta;$ this automatically builds in the
desired permutation invariance from Remark \ref{rem:perm}. Function
$\ks(\cdot;\theta)$ may be parameterized, for example, as a neural network.
Recalling definition \eqref{eq:inn} of the innovation\index{innovation},
it may also be of interest to replace
(\ref{eq:predEnKF2}c, \ref{eq:predEnKF2}d) by an update of the form
\begin{align*}
v_{j+1}^{(\sam)} &=\ks\bigl(\widehat{v}_{j+1}^{(\sam)},\widehat{m}_{j+1},i_{j+1}^{(\sam)},\widehat{C}_{j+1}^{vh},\widehat{C}_{j+1}^{yy},\widehat{C}_{j+1}^{vv};\theta\bigr), \quad \sam=1,\ldots,\Sam,\\
i_{j+1}^{(\sam)} & =\yd_{j+1}-\widehat{y}_{j+1}^{(\sam)}, \quad \sam=1,\ldots,\Sam,
\end{align*}
where 
$$\widehat{m}_{j+1}=\frac{1}{N}\sum_{n=1}^N \widehat{v}_{j+1}^{(n)}$$
is the mean of the predicted forecast ensemble.

Recall from Remark \ref{rem:lh} that if $h$ is linear then the two covariances
$\widehat{C}_{j+1}^{vh},\widehat{C}_{j+1}^{yy}$ can be expressed in terms of 
$\widehat{C}_{j+1}^{vv}$ and $H$; thus, in this case we might seek to learn the
gain in the form $\ks\bigl(\widehat{C}_{j+1}^{vv}, H;\theta\bigr).$
The reader will be able to suggest many other variants on the preceding formulations of a learning problem for an EnKF-like data assimilation algorithm. One such variant is described in Subsection \ref{ssec:pft}, where the learning
objective is to match the filtering distribution.
\end{example}

\subsection{Learning Localization and Inflation in EnKF}
\label{ssec:loci}

Inflation\index{inflation} and localization\index{localization}, discussed in Subsection~\ref{ssec:enkf}, are essential for the performance of the EnKF, as discussed in Remark \ref{rem:enkf}. Learning can be used to determine appropriate parameters for these features of the EnKF. Again we continue to work under Data Assumption \ref{da:9d}.

Recall that the multiplicative approach to inflation\index{inflation} given by \eqref{eq:inflation} has the form
\begin{equation*}
    \widehat{v}_{j+1}^{(\sam)} \to \widehat{m}_{j+1} + \alpha (\widehat{v}_{j+1}^{(\sam)} - \widehat{m}_{j+1}),
\end{equation*}
where $\widehat{v}_{j+1}^{(\sam)}$ are predicted states in the ensemble and $\widehat{m}_{j+1}$ is their mean. Hadamard product covariance localization\index{localization} 
given by \eqref{eq:loc_cov}, \eqref{eq:loc} has the form 
\begin{align*}
    \widehat{C}_{j+1}&\to L \circ \widehat{C}_{j+1},\\
    (L)_{ab} &= e^{-\mathsf{d}(a,b)^2/\ell^2},
\end{align*}
where $\circ$ denotes the Hadamard product, $a,b \in \{1,\dots,d\}$ represent  variable indices, and $\mathsf{d}(a,b)$ measures the distance between the variables corresponding to the indices.  If both inflation and localization are to be learned
simultaneously, then the inflation parameter $\alpha$ and localization radius $\ell$ can be considered to form parameter $\theta: = (\alpha, \ell)$. We may again use 
the loss function in \eqref{eq:state_estimation_obj} to determine $\theta,$
with $v_j(\theta)$ given by the ensemble mean in~\eqref{eq:SE}.

\begin{remark}
It may be of interest to separately learn either inflation or localization, and
generalization to do this is straightforward. Furthermore,
other ways of parameterizing localization\index{localization} could also be considered, such as taking $\theta = L$ and learning the entire matrix, or parameterizing $L(\theta)$ with a given structure different from the one given above. Alternatives to Hadamard product localization can also be considered, such as ones that learn a nonlinear map $\mathfrak{L}(\cdot\,; \theta)$  that localizes the gain\index{gain!localized}, $K_j\to\mathfrak{L}(K_j; \theta).$
\end{remark}

\subsection{Optimal Particle Filter}
\label{ssec:opflf}
Here we work in the setting of the linear observation operator~\eqref{eq:linearobs},
leading to the following state--observation dynamics model, repeated here for convenience: 
\begin{align*}
\vd_{j=1} &= \Psi(\vd_j) + \xid_j , \\
\yd_{j+1} &= H\vd_{j+1} + \etad_{j+1}\,.
\end{align*}
We again make the same Gaussian and independence 
assumptions on the initialization and noise as 
are made in equations \eqref{eq:sdm}, 
\eqref{eq:dm} and Assumption \ref{a:fas}. 
Recall that the optimal particle filter, from Subsection \ref{ssec:opf}, 
works by proposing
particles through an ensemble of equally weighted noisy 3DVar estimators and then reweighting them:
see equation \eqref{eq:opf99}. Here, recognizing that reweighting often leads to particle
collapse (see citations in Section \ref{sec:dabib}), 
we instead seek to learn a desirable combination of the ensemble of 3DVars:
\begin{subequations}
\label{eq:opf999}
\begin{align}
    \widehat{v}_{j+1}^{(n)} &= (I - KH)\Psi(v_j^{(n)}) + K\yd_{j+1} + \zeta_{j+1}^{(n)}, \quad \sam=1,\ldots,\Sam, \\
    v_{j+1}^{(\sam)} &= \widehat{v}_{j+1}^{(\sam)}+
\ks\Bigl(\widehat{v}_{j+1}^{(\sam)},\yd_{j+1},i_{j+1}^{(\sam)},\{\widehat{v}_{j+1}^{(\ell)}\}_{\ell=1}^N;\theta\Bigr), \quad \sam=1,\ldots,\Sam,\\
i_{j+1}^{(\sam)} & =\yd_{j+1}-H\widehat{v}_{j+1}^{(\sam)},\quad \sam=1,\ldots,\Sam.
\end{align}
\end{subequations}
Thus the innovation\index{innovation} is defined analogously to
\eqref{eq:inn}, measuring how closely the predicted state
matches the data.
Random variables $\zeta_{j+1}^{(n)}$ are i.i.d.\thinspace in $j$ and $n$ and distributed according to $\mathcal{N}(0,C),$ where, as in \eqref{eq:covopt}:
\begin{align}
\label{eq:covopt2}
C = (I - KH)\Sigma, \quad
K = \Sigma H^\top S^{-1}, \quad
S = H\Sigma H^\top + \Gamma.
\end{align}
Again, as explained in Remark \ref{rem:perm}, it is desirable that $\ks(\widehat{v}_{j+1}^{(\sam)},\yd_{j+1},i_{j+1}^{(\sam)},\cdot\,;\theta)$ 
is invariant with respect to permutation of the ensemble. The proposed form
of the algorithm is a generalization of \eqref{eq:fakeobs2}.

We continue to work under Data Assumption \ref{da:9d}. 
To emphasize dependence of the algorithm on the parameters $\theta$ to be learned,
we again write $v_j^{(n)}(\theta)$ and consider the state estimator in~\eqref{eq:SE}. 
We may then determine $\theta$ from \eqref{eq:state_estimation_obj}.

\begin{remark}
As in the context of the EnKF, other forms of learning problems may be
postulated; see Example \ref{ex:thetaform}. One generalization specific 
to this optimal particle filter--based approach
relates to the fact that it may be desirable to learn
the fixed gain $K$ appearing in~(\ref{eq:opf999}a) alongside $\theta$, 
rather than fix $K$ according to \eqref{eq:covopt2}. 
We also note that the update step in (\ref{eq:opf999}b)  contains multiple
redundancies and could be simplified to
$$v_{j+1}^{(\sam)} = 
\ks\Bigl(\widehat{v}_{j+1}^{(\sam)},i_{j+1}^{(\sam)},\{\widehat{v}_{j+1}^{(\ell)}\}_{\ell=1}^N;\theta\Bigr), \quad \sam=1,\ldots,\Sam.$$
\end{remark}

\subsection{Ergodicity of Ensemble and Particle Filters}
\label{ssec:ergf}

In Remark \ref{rem:ergo} we discussed how our proposed learning framework
for generalizations of 3DVar can be viewed in terms of ergodicity of
the underlying filter.
In the ensemble and particle filters described in preceding subsections,
the notion of ergodicity is also relevant and it can be
extended to these interacting particle systems; we do not write down
the details as it is notationally heavy to do so and as the details are
not exploited in this book.
As in Remark \ref{rem:ergo}, ergodicity may be used to define the large $J$
limit of the optimization problem \eqref{eq:state_estimation_obj}.
Beyond the state estimation\index{state estimation} setting considered in this section, the idea of ergodicity is also relevant to define the large $J$
limit of the optimization problem \eqref{eq:state_estimation_obj33}
that we define in the next section for
probabilistic estimation. \index{probabilistic estimation}

\section{Probabilistic Estimation: Learning Objectives}\label{sec:flo}

In Section \ref{sec:selo} we introduced learning objectives for state
estimation\index{state estimation}; 
this is followed in Section \ref{sec:sea} in which we
study algorithms learned according to state estimation objectives. 
In this section we introduce probabilistic learning objectives, 
aimed at probabilistic estimation;\index{probabilistic estimation}  
in the following Section \ref{sec:fa}, we study 
algorithms learned to perform well at probabilistic estimation\index{probabilistic estimation}.

Thus our aim in this section is to introduce learning objectives that
may be used to optimize a specified class of algorithms 
with respect to their ability to approximate
the filtering distribution. 
In Subsection \ref{ssec:mfd} we set out a desirable learning objective, but
highlight the fact that implementation is difficult precisely because the
true filtering distribution is not known. 
Subsection \ref{ssec:PLA1} proposes using components of the 
bootstrap particle filter\index{particle filter!bootstrap} to
train an equal weights particle filter.\index{particle filter!equal weights}
In Subsection \ref{ssec:PLA2} we describe the same idea, but in the context of
the optimal particle filter\index{particle filter!optimal}.
In Subsection \ref{ssec:mvb} we provide links to others parts of the book where
ideas from variational inference\index{variational inference} 
are used to learn approximate filters from probabilistic learning objectives. 

The learning objectives we consider for probabilistic estimation are based on various ``distance-like'' ways to quantify closeness between
probability distributions studied in Chapter \ref{ch:distances}. In our filtering context, it is essential to consider metrics, divergences and scoring rules that are implementable given only samples, since pointwise evaluation of the underlying density is often infeasible. We refer to Remark \ref{rem:POF} for a discussion of choices of distance-like measures with this attribute.

\subsection{Matching the Filtering Distribution}
\label{ssec:mfd}

Let $\D\colon \cP(\Ru) \times \cP(\Ru) \rightarrow \mathbb{R}$ be a divergence\index{divergence}, as defined in Section \ref{sec:div}.
A desirable objective function for matching the approximate filtering distribution $\pia_j(\theta)$ to the true filtering distribution $\pi_j$
defined in Subsection \ref{ssec:filtering}, 
over $j \in \{1, \ldots, J \},$ is 
\begin{subequations}\label{eq:state_estimation_obj33}
    \begin{align}
        \J^J(\theta) &= \frac{1}{J}\sum_{j=1}^{J} \D(\pia_j(\theta)\|\pi_j),\\
        \theta^\star&\in \argmin_{\theta\in\Theta} \J^J(\theta).
    \end{align}
\end{subequations}
However, implementation of such an objective is difficult because the
true filtering distribution $\pi_j$ cannot be constructed explicitly,
except in simple cases such as the linear Gaussian setting, leading
to the Kalman filter\index{Kalman filter};
thus creation of training data $\{\yd_j,\pi_j\}_{j=1}^J$ to enable
solution of the minimization problem \eqref{eq:state_estimation_obj33} is
difficult. In contrast, when state estimation is 
the goal of the learning problem, as detailed in Subsection \ref{ssec:mts},
it is possible to simulate training data $\{\vd_j,\yd_j\}_{j=1}^J$ 
from the stochastic dynamics model, as in Data Assumption \ref{da:9d}. 

A variety of approaches exist to address the difficulty of not having  
training data for probabilistic estimation\index{probabilistic estimation}.
We detail these in the next three subsections. The first two build on the use 
of ideas underlying the bootstrap and optimal particle filters; 
the third uses ideas from variational inference.
Furthermore we note that, in the context of amortization\index{amortization}, probabilistic  scoring rules\index{scoring rule!probabilistic} may be
used, circumventing the need to be able to simulate 
from the filtering distribution; see Remark \ref{rem:SPSR} and
Section~\ref{sec:SPSR}.

\begin{remark}
An essential difficulty to be confronted when learning approximate filters using a probabilistic 
learning objective is that filtering involves repeated solution
of an inverse problem, at each step $j$; the operator $\mathsf{A}_j$
appearing in \eqref{eq:An} is an operator representing the map from prior
$\mathbb{P}(v_{j+1}|\Yd_{j})$ to posterior 
$\mathbb{P}(v_{j+1}|\Yd_{j},\yd_{j+1}).$  
The prior used to solve this inverse problem does 
not have a closed form and differs at each step $j$ of the
filtering procedure. The issue of having a prior 
only known through samples  is addressed for inverse problems 
in Subsection \ref{ssec:EL}, but the assumption there is that only
one prior is considered, and exact samples from it are known. Neither
assumption applies in the filtering setting.
\end{remark}

\subsection{Matching the Filtering Distribution: Transport}\label{ssec:PLA1}
The aim of the methodology proposed in this subsection is to learn equal weights
filters\index{particle filter! equal weights} that benefit from both the robustness of methods with
equal weights and the accuracy of the bootstrap and optimal particle filters; the discussion
in Subsection \ref{ssec:collapse} serves as motivation.

We consider learning parameter $\theta$ appearing in an
equal weights ensemble filter\index{particle filter! equal weights}
algorithm of the form \eqref{eq:nofakeobs}. 
In what follows we employ the empirical measures $\pia_j$ and 
$\piaw_{j+1}$ which are defined in \eqref{eq:empEnKF99} and 
\eqref{eq:empEnKF88} respectively. 
Recall that the filtering update may be factorized
as in \eqref{eq:pnaB} into composition of prediction and
analysis steps:
\begin{equation*}
\hat{\post}_{j+1} = \Pred \post_j,\,\,\,\,\,  \post_{j+1} = \An_j (\hat{\post}_{j+1}).
\end{equation*}
The step (\ref{eq:nofakeobs}a) generates $\piaw_{j+1} \approx \Pred \pia_j$, 
approximating the prediction step $\Pred$. The step (\ref{eq:nofakeobs}b) maps
$\piaw_{j+1}$ to $\pia_{j+1}$, approximating the analysis step $\An_j$.
Using transport, as in the step (\ref{eq:nofakeobs}b), to 
approximate Bayesian inversion when the prior is only known through 
samples is the subject of Section \ref{sec:BVI}; we use
those ideas, now in the context of data assimilation.

We work under the following data assumption:
\begin{dataassumption}\index{Data Assumption}
\label{da:10a}
We are given data $\{\yd_{j}\}_{j=1}^J$ generated through a realization of \eqref{eq:ddm99}; furthermore we are able to evaluate $\like(\yd_{j+1}|  \cdot\,).$
\end{dataassumption}
For compactness of notation we write the step (\ref{eq:nofakeobs}b) as
\begin{align}
\label{eq:fakeobs3}
v_{j+1}^{(\sam)} &= T_j\Bigl(\widehat{v}_{j+1}^{(\sam)};\yd_{j+1},\theta\Bigr),\quad \sam=1,\ldots,\Sam.
\end{align}
We have removed explicit dependence on the 
ensemble $\{\widehat{v}_{j+1}^{(\ell)}\}_{\ell=1}^\Sam$ and use subscript $j$ on $T$ to indicate it.
To define the optimization problem for $\theta$ we proceed as follows.
The empirical\index{empirical} measure $\piaw_{j+1}$ given in \eqref{eq:empEnKF88} may itself be reweighted by likelihood information in order to obtain an 
approximation\footnote{$\mathsf{walg}$ for ``weighted algorithm''.} $\piw_{j+1}$ of the analysis distribution at time $j+1$:
\begin{equation}
\label{eq:empEnKFpred2}
\piw_{j+1} = \sum_{n=1}^N w_{j+1}^{(n)} \delta_{\widehat{v}_{j+1}^{(n)}},\quad
\ell_{j+1}^{(n)} = \like(\yd_{j+1}|\widehat{v}_{j+1}^{(n)}), \quad w_{j+1}^{(n)} = \frac{\ell_{j+1}^{(n)}}{\sum_{i=1}^N\ell_{j+1}^{(i)}}, \quad  
\end{equation}
where the likelihood weights are normalized to sum to one so that $\piw_{j+1}$ is a probability measure. Now recall the energy distance 
$\den^2\colon \cP(\Ru) \times \cP(\Ru) \rightarrow \mathbb{R}$
introduced in Definition \ref{def:energy}.
Using construction \eqref{eq:empEnKFpred2} we may estimate an optimal $\theta^\star$ that minimizes the distance at each analysis step, averaged over the time series:
\begin{subequations}\label{eq:prob_estimation_obj}
    \begin{align}
        \J^{N}(\theta) &= \frac{1}{J}\sum_{j=0}^{J-1} 
        \den^2\Bigl(T_j(\cdot \, ;\yd_{j+1},\theta)_\sharp \piaw_{j+1}(\theta),\piw_{j+1}(\theta) \Bigr),\\
        \theta^\star&\in \argmin_{\theta\in\Theta} \J^{N}(\theta).
    \end{align}
\end{subequations}
This is a time-averaged version of the loss function introduced, for the solution of 
inverse problems, in Section \ref{sec:BVI}.
After finding $\theta^\star$, and hence  $T_j(\cdot\, ;\yd_{j+1}, \theta^\star)$, 
we can evaluate the map at samples from the prediction to generate analysis samples.

The resulting learning approach implied by the objective \eqref{eq:prob_estimation_obj} 
 may be viewed as a form of \emph{unsupervised learning}\index{unsupervised learning} -- 
 we do not use explicit information
from the prior. As a result the performance can be poor. 
The proposed learning objective may have most value
when employed in conjunction with other learning objectives, 
that anchor the particles to the true signal, and hence anchor
the prior particle distribution encountered at each step of
the filtering process to be close to the truth. 
To this end we combine Data Assumptions
\ref{da:9d} and \ref{da:10a} to obtain
\begin{dataassumption}\index{Data Assumption}
\label{da:100a}
We are given data $\{\vd_{j},\yd_{j}\}_{j=1}^J$ generated through a realization of \eqref{eq:ddm99}; furthermore we are able to evaluate $\like(\yd_{j+1}|\cdot\,).$
\end{dataassumption}
We may then combine the objectives \eqref{eq:prob_estimation_obj}
and \eqref{eq:state_estimation_obj} to obtain, for $\lambda \in (0,1)$, 
the optimization problem
\begin{subequations}\label{eq:prob_estimation_obj_anchor}
    \begin{align}
        \J^{N}(\theta) &= \lambda \frac{1}{J}\sum_{j=0}^{J-1}
        \den^2\Bigl(T_j(\cdot;\yd_{j+1},\theta)_\sharp \piaw_{j+1}(\theta),\piw_{j+1}(\theta) \Bigr)
+(1-\lambda)\frac{1}{J}\sum_{j=1}^{J} \dd\bigl(v_j(\theta), \vd_j\bigr),\\
        \theta^\star&\in \argmin_{\theta\in\Theta} \J^{N}(\theta).
    \end{align}
\end{subequations}

\begin{remark}
\label{rem:resl}
By resampling $\piw_{j+1}$ we may obtain approximate
i.i.d.\thinspace samples $\{\mathfrak{v}_{j+1}^{(m)}\}_{m=1}^M$ from the analysis
distribution $\pi_{j+1}$ at time $j+1$ arising from the analysis step in \eqref{eq:pna}.
We then have the alternate approximation of the analysis distribution given by \begin{equation}
    \label{eq:empEnKFpred}
\piap_{j+1}  = \frac{1}{M}\sum_{m=1}^M \delta_{{\mathfrak{v}}_{j+1}^{(m)}}.
 \end{equation}
Using $\piap_{j+1}$ in~\eqref{eq:empEnKFpred} instead of $\pia_{j+1}$ in~\eqref{eq:prob_estimation_obj}, we may estimate an optimal $\theta^\star$ by solving the following problem:
\begin{subequations}\label{eq:prob_estimation_obj2}
    \begin{align}
        \J^{N,M}(\theta) &= 
        \frac{1}{J}\sum_{j=0}^{J-1} 
        \den^2\Bigl(T_j(\cdot \, ;\yd_{j+1},\theta)_\sharp \piaw_{j+1}(\theta),\piap_{j+1}(\theta) \Bigr),\\
        \theta^\star&\in \argmin_{\theta\in\Theta} \J^{N,M}(\theta).
    \end{align}
\end{subequations}
Methods for numerically approximating the objective function through sampling and resampling
are outlined in Section \ref{sec:BVI}.
We may also anchor this objective function to the true signal, similar to
\eqref{eq:prob_estimation_obj_anchor}.
\end{remark}

\subsection{Matching the Filtering Distribution: Generalized Transport}
\label{ssec:PLA2}

Recall the OPF\index{particle filter!optimal} from 
Subsection \ref{ssec:opf}. In this subsection we generalize 
the transport ideas from the previous subsection to 
incorporate the OPF proposal, rather than the 
BPF\index{particle filter!bootstrap} proposal. 
As in the previous subsection, our main motivation is 
avoiding the problem of weight collapse described in 
Subsection \ref{ssec:collapse}. Recall, from \eqref{eq:fact02},
that the two steps in the OPF factorization may be written as 
\begin{subequations}
\label{eq:fact022}
\begin{align}
\intp_{j+1}& = \mathsf{A}_j^{\textrm{OPF}}(\pi_j), \quad \pi_{j+1}  = \mathsf{P}_j^{\textrm{OPF}}\intp_{j+1},\\
\pi_j &=\Prob(\vd_j|\Yd_{j}), \quad \intp_{j+1}:=\Prob(\vd_{j}|\Yd_{j+1}).
\end{align}
\end{subequations}
Importantly for the benefits of the method over the BPF, 
both steps depend on the data $\yd_{j+1}$.

In the preceding subsection we showed how to make an equal weights 
particle filter\index{particle filter!equal weights} by
defining a transport that is close to the analysis step in the 
BPF.\index{particle filter!bootstrap}
We now consider an analogous approach, based on proposing particles 
using the optimal particle filter.\index{particle filter!optimal} Let
\begin{equation}
\label{eq:poppy0}
\pia_j=\frac{1}{N} \sum_{n=1}^N \delta_{v_j^{(n)}}
\end{equation}
be an approximation of the true filtering distribution.
Note that in Subsection \ref{ssec:opf} the approximation is
different, being based on a non-equally weighted set of
points $\{\hv_j^{(n)}\}_{n=1}^N$ and given in \eqref{eq:pop} as:
\begin{equation} 
\label{eq:poppy}
\pop_j = \sum_{n=1}^N w_{j}^{(n)}\delta_{\widehat v_{j}^{(n)}}.
\end{equation}
This is sampled, in the step (\ref{eq:pzp222}a) of the
OPF algorithm from Subsection \ref{ssec:opf},
to obtain equally weighted points $\{v_j^{(n)}\}_{n=1}^N$; these
particles  could have been used to state an approximate 
filtering distribution with form \eqref{eq:poppy0}.

We now describe learnable particle evolutions for the $\{v_j^{(n)}\}_{n=1}^N$. 
Our use of the variables with and without hats differs from the
useage in our initial exposition of the OPF\index{particle filter!optimal}
contained in Subsection \ref{ssec:opf}. This is because, in that context,
it is possible to commute the order in which mappings
$\mathsf{A}_j^{\textrm{OPF}}$  and $\mathsf{P}_j^{\textrm{OPF}}$
are undertaken; this is because application of $\mathsf{A}_j^{\textrm{OPF}}$
corresponds, in Subsection \ref{ssec:opf}, to a simple reweighting. Here we
replace it by a transport map that must be executed before
application of $\mathsf{P}_j^{\textrm{OPF}}$.
Recall matrices $K$, $S$, and $C$ defined in \eqref{eq:covopt} and consider the following algorithm class: for $n=1, \ldots, \Sam$, 
\begin{subequations}
\label{eq:sboe}
\begin{align}
\widehat{v}_{j}^{(\sam)} &= T_j\Bigl(v_j^{(\sam)};\yd_{j+1},
\theta\Bigr), \\
{v}_{j+1}^{(n)} &= (I - KH)\Psi(\widehat{v}_j^{(n)}) + K\yd_{j+1} + \zeta_{j+1}^{(n)},
\end{align}
\end{subequations}
where $\zeta_{j+1}^{(n)} ~\text{i.i.d.}\thinspace \mathcal{N}(0,C)$. 
Step (\ref{eq:sboe}a) is a particle, transport-based, approximation of the action of $\mathsf{A}_j^{\textrm{OPF}}$, resulting in a particle approximation 
of $\intp_{j+1};$ step (\ref{eq:sboe}b) gives samples from an approximation 
of $\Pred_j^{\textrm{OPF}}\intp_{j+1}.$ 
We view the resulting method as an equal weights ensemble filter\index{particle filter!equal weights} and thus obtain the approximate filtering distribution at time $j+1$ from formula \eqref{eq:poppy} with $j \mapsto j+1:$
$$\pia_{j+1}=\frac{1}{N} \sum_{n=1}^N \delta_{v_{j+1}^{(n)}}.$$

We note that
$$T_j(\cdot\, ;\yd_{j+1},\theta)_\sharp \pia_j \approx \intpw_{j+1}:=\mathsf{A}_j^{\textrm{OPF}}(\pia_j).$$ 
On the other hand $\intpw_{j+1}$ may be obtained exactly
by employing the OPF weightings (\ref{eq:pzp222}b), resulting in the
linear observations setting in \eqref{eq:opf99}: 
\begin{equation}
\label{eq:sbpoe2}
\intpw_{j+1} = \sum_{n=1}^N w_{j+1}^{(n)} \delta_{\widehat{v}_{j+1}^{(n)}},\quad
\ell_{j+1}^{(n)} = \exp\left(-\frac12|\yd_{j+1} - H\Psi(v_j^{(n)})|_S^2\right), \quad w_{j+1}^{(n)} = \frac{\ell_{j+1}^{(n)}}{\sum_{i=1}^N\ell_{j+1}^{(i)}}.
\end{equation}
The methods of the preceding subsection may now be generalized,
leading to the following optimization problem, defined for any
$\lambda \in (0,1):$
\begin{subequations}\label{eq:prob_estimation_obj_anchor2}
    \begin{align} 
        \J^{N}(\theta) &= \lambda \frac{1}{J}\sum_{j=0}^{J-1}
        \den^2\Bigl(T_j(\cdot;\yd_{j+1}, \theta)_\sharp \pia_{j}(\theta),\intpw_{j+1}(\theta) \Bigr)
+(1-\lambda)\frac{1}{J}\sum_{j=1}^{J} \dd\bigl(v_j(\theta), \vd_j\bigr).\\
        \theta^\star&\in \argmin_{\theta\in\Theta} \J^{N}(\theta).
    \end{align}
\end{subequations}

\subsection{Matching the Filtering Distribution: Variational Bayes}
\label{ssec:mvb}

In Sections \ref{sec:variational_filtering0} and \ref{sec:fss} we demonstrate how variational  Bayes may be used to design objective functions from which to learn approximate filters that are defined implicitly by the action of transports. 
The reader is encouraged to study the resulting learning objectives implied by these approaches.
We include an example, based on the ideas in Section \ref{sec:variational_filtering0}, in
Subsection \ref{ssec:ensemble_vi}.

\section{Probabilistic Estimation: Algorithms\index{algorithm!filtering}}\label{sec:fa}

In this section we provide explicit instances of the general ideas developed in the
preceding section. Subsection \ref{ssec:pft} contains a specific example of the
transport methodology defined in Subsection \ref{ssec:PLA1}.
In Subsection \ref{ssec:pfgt} we include a specific example of the
generalized transport methodology defined in Subsection \ref{ssec:PLA2}.
Subsection \ref{ssec:ensemble_vi} is devoted to learning the analysis\index{analysis} map,
in the ensemble setting, using variational inference as outlined in Subsection \ref{ssec:mvb}.

\subsection{Transport}\index{transport}
\label{ssec:pft}

We describe an explicit example of the methodology introduced in Subsection \ref{ssec:PLA1}. 
In particular we describe a specific class of equal weight particle 
filters\index{particle filter!equal weights}.
However, rather than using \eqref{eq:nofakeobs} as we did for simplicity in Subsection \ref{ssec:PLA1}, 
we use the slightly more general form \eqref{eq:nofakeobs0}. Within this class we consider an algorithm of the form,
for $\sam=1,\ldots,\Sam$,
\begin{align*}
\widehat{v}_{j+1}^{(\sam)} &= \Psi(v_{j}^{(\sam)})+\xi^{(\sam)}_{j},  \\
v_{j+1}^{(\sam)} &= \widehat{v}_{j+1}^{(\sam)}+K_{j+1}\bigl(\yd_{j+1}-\eta_{j+1}^{(\sam)} -h(\widehat{v}_{j+1}^{(\sam)})\bigr),\\
K_{j+1}&=\ks\Bigl(\{\widehat{v}_{j+1}^{(\ell)},h\bigl(\widehat{v}_{j+1}^{(\sam)}\bigr)\}_{\ell=1}^\Sam,\Gamma;\theta\Bigr).
\end{align*}
The noise entering these particle updates satisfy Assumption \ref{a:enkf-noise}.
We generalize the form of Kalman gain $K_{j+1}$
used in the EnKF\index{Kalman filter!ensemble}\index{EnKF|see{Kalman filter, ensemble}}
and given by \eqref{eq:gainEnKF}.
To this end we assume the following form for $\ks$: 
\begin{equation}\label{eq:k-theta}
    \ks = \Kt^{(1)} \left( \Kt^{(2)} + \Gamma \right)^{-1}.
\end{equation}
Here $\Kt^{(1)}$ and $\Kt^{(2)}$ are modified versions of $\hat{C}^{vh}_{j+1}$ and $\hat{C}^{hh}_{j+1}$ from \eqref{eq:gainEnKF}, with modifications parameterized by $\theta.$
Specifically we have, recalling \eqref{eq:empEnKF88},
\begin{subequations}\label{eq:full_K}
\begin{align}
    \Kt^{(1)} &= \bbE^{\hv\sim {\piaw}_{j+1}} \left[ \left(\hv - \widehat{m} + \hw_\theta\right) \otimes \left(h(\hv) - \widehat{h} + \hz_\theta\right)\right],\quad \widehat{m}= \bbE^{\hv\sim {\piaw}_{j+1}}\bigl(\hv\bigr),\label{eq:full_K1}\\
    \Kt^{(2)} &= \bbE^{\hv\sim {\piaw}_{j+1}} \left[ \left(h(\hv) - \widehat{h} + \hz_\theta\right) \otimes \left(h(\hv) - \widehat{h} + \hz_\theta\right)\right],
    \quad \widehat{h}= \bbE^{\hv\sim {\piaw}_{j+1}}\bigl(h(\hv)\bigr).\label{eq:full_K2}
\end{align}
\end{subequations}
Here $\hw_\theta$ and $\hz_\theta$ are learnable functions, parameterized by $\theta$, that  correct the basic form of the EnKF. They are assumed to depend on the state $\hv$, observation $h(\hv)$, true observation $\yd_{j+1}$ and the distribution ${\piaw}_{j+1}$.

\subsection{Generalized Transport}\index{transport}
\label{ssec:pfgt}

Here we consider application of the ideas proposed in Subsection \ref{ssec:PLA2}, exploiting the
OPF\index{particle filter!optimal}. We work in the setting of linear observation operator
where the method is tractable, thus considering the data assimilation problem defined
by \eqref{eq:linearobs}. Recall the Kalman gain $K$ and covariances $C$, $S$ 
defined by \eqref{eq:covopt}.

We study a particular subclass of the equal weights particle filter\index{particle filter! equal weights} \eqref{eq:sboe} and consider particles evolving according to, for $n=1, \ldots, \Sam$,
\begin{subequations}
\label{eq:sbo2}
\begin{align}
\widehat{v}_{j}^{(\sam)} &= v_j^{(\sam)} + K_{j}\Bigl(\yd_{j+1}-\varsigma_{j+1}^{(n)}-H\Psi\bigl(v_j^{(\sam)}\bigr)\Bigr),\\ 
K_{j} &=\ks\Bigl(\{v_j^{(\ell)},\Psi\bigl(v_j^{(\ell)}\bigr)\}_{\ell=1}^\Sam,H,S;\theta\Bigr), \\
{v}_{j+1}^{(n)} &= (I - KH)\Psi(\widehat{v}_j^{(n)}) + K\yd_{j+1} + \zeta_{j+1}^{(n)},
\end{align}
\end{subequations}
where $\zeta_{j+1}^{(n)}, \varsigma_{j+1}^{(n)}$ are distributed $\text{i.i.d.}$\thinspace respectively,
according to $\mathcal{N}(0,C)$ and $\mathcal{N}(0,S)$, in $j$ and $n$, analogous to Assumption \ref{a:enkf-noise}.
Remark \ref{rem:perm} concerning permutation invariance is also relevant here.

\subsection{Variational Bayes}\index{variational!Bayes}
\label{ssec:ensemble_vi}
In this section we present a formulation for learning parameters based on the variational inference objectives introduced in Section~\ref{sec:variational_filtering0}. To do so, we consider a class of variational approximations defined by
$$q_{j+1}(\theta) = T_j(\cdot;\yd_{j+1},\theta)_\sharp \mathsf{P}q_j(\theta), \quad q_0 = \pi_0.$$
As in Subsection \ref{ssec:PLA1} we use symbol $T_j$ to denote dependence of the transport on the predicted
ensemble. And, in particular, by building in permutation invariance this may be viewed as dependence on 
$\piaw_{j+1}$ defined in \eqref{eq:empEnKF88}.

If $q_j$ is an empirical measure
\begin{equation*}
    q_j = \frac{1}{N}\sum_{n=1}^N \delta_{v_j^{(n)}},
\end{equation*}
then $q_{j+1}$ is also an empirical measure whose samples are defined by first sampling from the forecast model, followed by evaluating the transport map. That is,
\begin{align*}
    q_{j+1} &= \frac{1}{N} \sum_{n=1}^N \delta_{v_{j+1}^{(n)}}, &&\widehat{v}_{j+1}^{(n)} = \Psi(v_{j}^{(n)}) + \xi_j^{(n)},\\
    v_{j+1}^{(n)} &= T_j(\widehat{v}_{j+1}^{(n)}; y^\dagger_{j+1},\theta), && v_0^{(n)}\sim \pi_0.
\end{align*}
Our goal is to identify an element of this class of measures $q_j$, by selecting $\theta$ that approximates the filtering distribution $\pi_j$ at step $j$ of filtering using the variational Bayes objective in~\eqref{eq:Jj}. Recall the likelihood  $\mathbb{P}(y^\dagger_{j+1} | \vd_{j+1})$ and
the objective function~\eqref{eq:Jj}:
\begin{equation*}\label{eq:variational_obj_term}
    \J_{j+1}(\theta) = \dkl(q_{j+1}(\theta)\|\Pred q_j(\theta)) - \mathbb{E}^{\vd_{j+1}\sim q_{j+1}(\theta)}[\log \mathbb{P}(\yd_{j+1} | \vd_{j+1})].
\end{equation*}
Since we have access to the likelihood in the filtering setting, the second term in the preceding
objective can be directly evaluated for the empirical measure to obtain
\begin{equation}
    - \mathbb{E}^{\vd_{j+1}\sim q_{j+1}(\theta)}[\log \mathbb{P}(\yd_{j+1} | \vd_{j+1})] = -\frac{1}{N}\sum_{n=1}^N \log \mathbb{P}(\yd_{j+1} | v^{(n)}_{j+1}).
\end{equation}
To differentiate $\J_{j+1}(\theta)$ with respect to $\theta$  the reparameterization trick\index{reparameterization trick}, defined in Subsection~\ref{ssec:reparameterization},
can be used to obtain the gradient with respect to $\theta$.
However, the first term is problematic since one cannot evaluate the KL divergence\index{divergence!Kullback--Leibler} between two empirical measures. To
address this issue we make an approximation by first projecting into the space of Gaussian measures. Using the Gaussian projection\index{Gaussian!projection} $\mathsf{G}$ introduced in Remark~\ref{rem:bishop}, we make the replacement
\begin{equation}\label{eq:gaussian_approx_kl}
    \dkl(q_{j+1}(\theta)\|\mathsf{P}q_j(\theta)) \mapsto
\dkl\Bigl(\mathsf{G}(q_{j+1}(\theta))\|\mathsf{G}\bigl(\mathsf{P}q_j(\theta)\bigr)\Bigr),
\end{equation}
and then use the formula for the KL divergence between two Gaussian measures, \eqref{eq:kl_gaussians}.
One can then either minimize the sum over $j$ of the objectives $\J_j$, as in \eqref{eq:var_smoothinga9}, or minimize each $\J_j$ individually, as in \eqref{eq:var_smoothinga99}.

\begin{remark}
    The substitution \eqref{eq:gaussian_approx_kl} does not lead to a viable method if $N < d$, since the covariance matrix of the resulting Gaussian will be rank-deficient, leading to a degenerate distribution for which the KL divergence cannot be computed. The covariance matrix can be modified to be full rank by a technique such as localization\index{localization}; see Subsection~\ref{ssec:enkf}.\ as{We may also wish to learn localization, as discussed in 
    Subsection~\ref{ssec:loci}.}
\end{remark}

\begin{remark}\label{rem:alternative_cost_gaussian}
    Using the invariance of the KL divergence\index{divergence!Kullback--Leibler} under invertible transformations, it can be shown that an alternative to \eqref{eq:gaussian_approx_kl} can be formulated that uses the Gaussian projection $\mathsf{G}$ only once rather than twice; see the bibliography.
\end{remark}

\section{Bibliography} \label{sec:Bib_learning_filters}

This chapter presents various methodologies for learning algorithms to solve both the
state\index{state estimation} and probabilistic estimation\index{probabilistic estimation} 
problems in data assimilation.
We adopt a unifying framework based on transport maps. We show how several learning problems for classical algorithms, such as learning the gain in 3DVar\index{3DVar}, can be viewed through the
lens of machine learning; and we generalize the classical EnKF\index{Kalman filter!ensemble}\index{EnKF|see{Kalman filter, ensemble}} algorithms to create new equal weights particle filters.\index{particle filter! equal weights}  Whilst some of the proposed algorithms have already been implemented and tested, demonstrating favourable behaviour,  not all of the proposed methods have been implemented. It is thus of interest to examine the performance of variants proposed here, such as the methods in Section \ref{sec:flo}, which have not yet been tested. In this section, we provide references to the current state-of-the-art.

Learning approaches for ensemble filtering and smoothing based on transportation of measure have shown recent promise in generalizing classic Kalman approaches and reducing the error. The paper \cite{spantini2022coupling} introduces the idea of learning transports 
to build prior-to-posterior maps, while~\cite{al2023optimal} seeks optimal transport maps by solving adversarial learning problems. A related framework known as Gaussian anamorphosis seeks invertible transformations of the state and observations to a space where Kalman algorithms apply. See~\cite{grooms2022comparison} for an approach based on nonlinear diagonal transformations and a generalization that uses invertible neural networks in~\cite{chipilski2023exact}. Other nonlinear generalizations of ensemble Kalman filtering methods include~\cite{hoang2021machine, anderson2010non,lei2011moment}. A nonlinear filter that is derived using variational inference, based on transports that lie in reproducing kernel Hilbert spaces (see Example~\ref{ex:kernel_trick}) is~\cite{pulido2019sequential}. Approaches based on diffusion models are provided in \cite{bao_score-based_2024,hodyss_using_2026}. An operator learning perspective on learning smoothers is provided in \cite{calvello_operator_2026}.

Approaches to parameterize and learn the gain to minimize state estimation error has been studied in various settings. 3DVar gain\index{gain!3DVar} learning was considered in \cite{hoang_simple_1994,mallia-parfitt_assessing_2016}, and a lower-dimensional parameterization was considered in \cite{hoang_adaptive_1998}; geophysical applications motivated this work.
The form of 3DVar, based on estimation of $\hat{C}$ and use of \eqref{eq:udc} to form the
gain, originated in numerical weather prediction \cite{lorenc1986analysis,lorenc2000met}. It was proven in \cite{hoang_simple_1994,mallia-parfitt_assessing_2016} that the fixed gain\index{gain} which minimizes the expected error with respect to observations, in the asymptotic limit, is the steady-state Kalman gain.\index{gain!steady state}\index{gain!Kalman} However, the cost functions considered were different and \cite{hoang_simple_1994} required assumptions on the rank of $H$. Learning the gain was also considered in \cite{levine_framework_2022} with nonlinear dynamics. 
For the statement and proof of the Stolz--Ces\`aro theorem, used in the proof of 
Theorem \ref{thm:sc}, see \cite[Theorem~1.23]{muresan_concrete_2009}.

The learning of a neural network analysis step by minimizing a state estimation loss is considered in \cite{mccabe_learning_2021}. The analysis step there is EnKF-like in the sense that the ensemble members interact during the analysis step only through the ensemble mean and covariance. The learning of an analysis step is also considered in \cite{bocquet_accurate_2024}. Permutation-invariant neural networks were used for postprocessing of an ensemble in \cite{hohlein_postprocessing_2024}, and for learning  equal weight ensemble filters in \cite{bach2025learning}. The architecture described in Subsection \ref{ssec:pft}
is closely to that proposed in \cite{bach2025learning}. The paper
\cite{bach2025learning} considers a general framework for learning, based on minimizing state estimation error, and generalizing the ensemble Kalman filter; the set transformer is used to
encode ensembles of arbitrary size with a single set of learned parameters, and fine-tuning is
used for localization\index{localization} and inflation\index{inflation} parameters which are
dependent on ensemble size.
The issue of estimating optimism\index{optimism} and out-of-sample performance of DA algorithms is discussed in \cite{brocker_sensitivity_2012,mallia-parfitt_assessing_2016}.
Various approaches to learning inflation are discussed in \cite{anderson_spatially_2009,miyoshi_gaussian_2011}. Learning localization by minimizing the squared Euclidean distance between the analysis and the true trajectory is considered in \cite{wang_convolutional_2023}, while a similar approach with a different scoring rule is considered in \cite{moosavi_tuning_2019}. Other approaches to learning localization are discussed in \cite{popov_bayesian_2019,cheng_graph_2021,kalnay_review_2023,le_provost_adaptive_2023,vishny_high-dimensional_2024}. 
Emulating the analysis step of a Kalman filter using recurrent neural networks is considered in \cite{harter_data_2012}.
An ensemble estimator for the energy score is introduced in \cite{gneiting_assessing_2008}.

Several works have proposed to learn the filtering distribution directly without using knowledge of the model dynamics \cite{brocker_probabilistic_2009,boudier2023data,rozet_score-based_2023}. This is sometimes referred to as ``end-to-end''\index{end-to-end prediction} learning. This has also been considered for state estimation in~\cite{tsiamis_online_2023,du_can_2023,vaughan_aardvark_2024,calvello_operator_2026}.

The methodology presented in Subsection~\ref{ssec:ensemble_vi} for learning ensemble filters using variational inference is introduced in \cite{luk_learning_2024}. The reformulation of the objective function that avoids one of the Gaussian projections, mentioned in Remark \ref{rem:alternative_cost_gaussian}, is discussed therein.

\chapter{\Large{\sffamily{Data Dependence of Filters and Smoothers}}}
\label{ch:LG10}

In this chapter we use learning to solve the filtering and smoothing problems introduced in Chapter \ref{lecture7}. While we also considered learning filters and smoothers in Chapters~\ref{ch:LG} and \ref{ch:LG2},
the key distinguishing concept in this chapter is {\em amortization}\index{amortization}: the idea that, if we learn dependence of an algorithm on the observations, it may be re-used multiple times and hence the initial cost of training 
is amortized. The success of such an approach depends on the ability of the proposed method to learn observation dependence that generalizes well. 

Recall that the idea of learning based on 
multiple instances of observational data, amortization,\index{amortization} is introduced in earlier chapters. In Sections~\ref{sec:MAPd}--\ref{sec:LRP} of Chapter~\ref{ch:data-dependence} we introduced amortization to approximate observed data dependence in the MAP estimator, the posterior distribution, and the likelihood function; we also used related data to learn regularizers in the context of inverse problems. Furthermore, we note that in Chapter \ref{ch:LG2} we have also learned models that take different observation sequences as input; there we use long time-trajectories and ergodicity, coupled with inductive bias in the form of the learned filters, to address the issue of learning dependence on different observation sequences. In this chapter our approach is more akin to 
Chapter \ref{ch:data-dependence}, using multiple data sets, rather than one long trajectory and the appeal to ergodicity. We focus on learning observational data dependence for filtering and smoothing algorithms.

We consider both state estimation: finding a representative sequence of states from the observations and knowledge of the model; and probabilistic estimation: approximating the posterior distributions for sequences of states given observations.
The material is organized as follows. Section~\ref{sec:SE} is devoted to state estimation, and to learning observational  data dependence
in 3DVar\index{3DVar} and 4DVar\index{4DVar}-type algorithms, as well as learning conditional mean estimators. In Section~\ref{sec:VA} we study
variational approaches from Chapter~\ref{ch:LG}, showing how to introduce amortization in both filtering and smoothing approaches. In Section~\ref{sec:SPSR} we introduce a learning paradigm for amortized data assimilation\index{data assimilation!amortized} problems, based on strictly proper scoring rules\index{scoring rule!strictly proper}. Section~\ref{sec:TA} is devoted to transport-based methods, generalizing those of Chapter~\ref{ch:LG2} to the amortized setting; transport methodology is presented for learning ensemble filters by employing objectives based on strictly proper scoring rules or maximum likelihood. We also include a concrete example of learning transports for smoothing, again employing scoring rules. A key distinction between Section~\ref{sec:TA} and the earlier 
Sections~\ref{sec:SE}--\ref{sec:SPSR} is that the former develops online methods, whilst the other sections
learn offline. The chapter concludes with bibliographic remarks in Section \ref{sec:Bib_learning_filters2}. 
Throughout, we use the notational conventions summarized in Subsection \ref{ssec:note} and in Remark~\ref{rem:smooth-note}; in particular, we write
$\Vd:=\{\vd_0, \ldots, \vd_J\}, \: \Yd := \{\yd_1, \ldots, \yd_J\}$.
We recommend that the reader review these conventions.

\section{State Estimation}\label{sec:SE}

In Subsections~\ref{ssec:A3} and \ref{ssec:A4}, respectively, we show how the 3DVar\index{3DVar} and 4DVar\index{4DVar} 
algorithms may be amortized. We employ amortized versions of the objective from Subsection~\ref{ssec:mts}, using distance-like deterministic scoring rules $\dd$\index{scoring rule!distance-like deterministic}; see Definition \ref{def:det_score} from Subsection \ref{ssec:dldsc}. In Subsection~\ref{ssec:squared_error_min} we study amortization for estimators of the filter or smoother mean. We employ the expected squared error, which is a specific example of a
distance-like deterministic scoring rule, to define the objective function for amortization.

\subsection{Amortized 3DVar\index{3DVar!amortized}}\label{ssec:A3}

We consider \index{3DVar}3DVar\footnote{More properly \emph{cycled 3DVar}\index{3DVar!cycled}; see Remark \ref{rem:cycled}.} with the analysis step written in the optimization form \eqref{eq:3dvar_opt}, leading to the following algorithm:
\begin{subequations}
\label{eq:pred3DVARopt}
\begin{align}
\hat{v}_{j+1} &= \Psi(v_j),\\
v_{j+1} &\in \argmin_{v \in \R^d} \J_{j+1}(v),\\
\J_{j+1}(v) &= \frac{1}{2}|v-\hv_{j+1}|^2_{\hC}+\frac12|\yd_{j+1}-h(v)|^2_{\Gamma}.	
\end{align}
\end{subequations}
Our goal in this subsection is to make choices within the algorithm class
\eqref{eq:pred3DVARopt}, or approximations of this algorithm class, 
to improve recovery of the state $\Vd$ from the observations $\Yd$ (using the notation from \eqref{eq:VdYd}).

Recall from Subsection~\ref{ssec:3dvar} that, for linear $h(\cdot)=H\cdot$, 
this algorithm reduces to
\begin{subequations}
 \label{eq:pred3DVAR33}
\begin{align}
\hat{v}_{j+1} &= \Psi(v_j),\\
v_{j+1} &= \hat{v}_{j+1} + K \left(\yd_{j+1}-H\hat{v}_{j+1}\right),
\end{align}
\end{subequations}
where $ K = \hC H^\top (H \hC H^\top + \Gamma)^{-1}$. This suggests learning
$\hC$, or $K$, to optimize the algorithm; indeed, the idea of learning $K$ is pursued in Subsection~\ref{ssec:3gain}. In this subsection, 
motivated by the setting where $h$ is nonlinear,
we describe learning strategies that work within generalizations
of the class of algorithms \eqref{eq:pred3DVAR33}.

\subsubsection{Amortized 3DVar: Nonlinear Dependence On Innovation\index{3DVar}}

Here we postulate an explicit form of the map $\hat{v}_{j+1} \mapsto v_{j+1},$ side-stepping the solution of the
optimization problem~\eqref{eq:pred3DVARopt}. To  this end, we introduce a learnable \emph{nonlinear} 
gain\index{gain!nonlinear} function, 
which takes the innovation\index{innovation} as input:
\begin{subequations}
 \label{eq:pred3DVAR202}   
\begin{align}
\hat{v}_{j+1} &= \Psi(v_j),\\
v_{j+1} &= \hat{v}_{j+1} + \ks \Bigl(\yd_{j+1}-h\bigl(\hat{v}_{j+1}\bigr);\theta\Bigr).	
\end{align}
\end{subequations}
Thus, we drop the optimization formulation of 3DVar in \eqref{eq:pred3DVARopt}, and instead seek to learn $\ks(\cdot\,;\theta): \R^{\dy} \to \R^{\du}$ in \eqref{eq:pred3DVAR202}. Function $\ks(\cdot\,;\theta)$ is a (possibly nonlinear) map for each fixed $\theta \in \Theta \subseteq \R^p$ that applies the state correction at each assimilation time based on the observed data and, in particular, the innovation.\index{innovation} 
This is a specific subclass of the algorithm class defined in \eqref{eq:pred3DVAR2add}. Moreover, it reduces to the classical ansatz made for 3DVar in~\eqref{eq:pred3DVAR} when the map $\ks$ is a linear function of the innovation.
We aim to learn $\theta$ given the following assumption about available data.

\begin{dataassumption}\index{Data Assumption}
    \label{d:9c3dvar}
We are given $\{\big((\Vd)^{(n)},(\Yd)^{(n)}\big)\}_{n=1}^N$, 
comprising multiple independent realizations from the joint 
distribution on $(\Vd,\Yd)$ defined by \eqref{eq:ddm99}.
\end{dataassumption}

Recall the objective function in~\eqref{eq:state_estimation_obj} that measures the distance between the estimated state $v_j(\theta)$ and true state $\vd_j$, both depending on  specific realizations of data. Here we define 
$\bigl\{  \{v_j^{(n)}(\theta)\}_{j=0}^J \bigr\}_{n=1}^N$
to be the set of state sequences from the algorithm \eqref{eq:pred3DVAR202}, when driven by the set of observation sequences $\{(\yd_j)^{(n)}\}_{n=1}^N$, for each $j \in \{1,\dots,J\}$; this set of state sequence estimators
depends on the gain parameter $\theta$, as made explicit in the notation. The initial conditions $\{v_0^{(n)} \}_{n=1}^N$ may be chosen, for instance, by sampling the Gaussian initial distribution $\Nc(m_0,C_0).$ 
We average the objective function in~\eqref{eq:state_estimation_obj} over the multiple realizations of sequences indexed by $n$; minimization defines the optimal $\theta^\star.$ Thus, we obtain the optimization problem
\begin{subequations}
\begin{align}
\label{eq:NobjSF}
        \J^{J,N}(\theta) &= \frac{1}{JN}\sum_{n=1}^N \sum_{j=1}^{J} 
        \dd\bigl(v_j^{(n)}(\theta), (\vd_j)^{(n)}\bigr),\\
        \theta^\star&\in \argmin_{\theta\in\Theta} \J^{J,N}(\theta).
\end{align}
\end{subequations}

\begin{remark}
\label{rem:threed}
Here we are solving the supervised learning\index{supervised learning} problem of recovering the state $\Vd$ from the 
observations $\Yd.$ We have imposed a sequential filtering structure, via the architecture
\eqref{eq:pred3DVAR202}, on the learned solution. It is natural to ask whether a smoothing-type
architecture could also be learned, in the context of 4DVar\index{4DVar}.
The next subsection considers such approaches but first, in this subsection, we consider another
approach to amortizing 3DVar\index{3DVar}.
\end{remark}

\subsubsection{Amortized 3DVar: Learning Optimization Step\index{3DVar!optimization amortized}}

We now return to the optimization formulation \eqref{eq:pred3DVARopt} of 3DVar.
We consider fixed $\hC$; see the original derivation as described in 
Subsection~\ref{ssec:3gain}. We then explicitly learn the solution of the optimization
problem embedded in algorithm \eqref{eq:pred3DVARopt}. To this end, we isolate the optimization step and define
\begin{subequations}
\label{eq:pred3DVARopt99isolate}
\begin{align}
\vmap &\in \argmin_{v \in \R^d} \J(v),\\
\J(v) &= \frac{1}{2}|v-w|^2_{\hC}+\frac12|\yd-h(v)|^2_{\Gamma}.	
\end{align}
\end{subequations}
Noting that the solution of this optimization problem, for fixed $\hC$, depends
on $(w,\yd)$, we write $\vmap=\vmap(w,\yd).$ We now discuss learning the function $\vmap.$

\begin{remark}
    The subscript ${\mathrm{MAP}}$ denotes the fact that the optimization problem \eqref{eq:pred3DVARopt99isolate} defines a maximum \emph{a posteriori}\index{maximum \emph{a posteriori}} estimator for a Bayesian inverse problem  with prior $\rho(v) = \mathcal{N}(w, \hC)$ and likelihood $\like(\yd|v) = \mathcal{N}(h(v), \Gamma)$; see Remark~\ref{rem:3dvar_map}.
\end{remark}

\begin{dataassumption}\index{Data Assumption}
\label{da:opt3d}
We are given $\{(w^{(n)},(\yd)^{(n)},\vmap^{(n)})\}_{n=1}^N$ drawn i.i.d. from a probability measure on $\R^{\du} \times \R^{\dy} \times \R^{\du}.$
\end{dataassumption}

We use this data to learn an approximation to the function $\vmap(w,\yd).$
With this goal in mind, we introduce parameter set $\Theta \subseteq \R^p$ 
and let $\cRt \subset C(\R^d \times \R^k,\R^d)$ denote
a parameterized set of functions $\vmap^{\mathsf{a}}(\cdot \, , \cdot\, ; \theta)$ for $\theta \in \Theta.$
The comments in Remark~\ref{rem:mapl} generalize to the situation here,
as does Example~\ref{ex:mapl}. The main issue is the choice of the probability measure from which the training data in Data Assumption~\ref{da:opt3d} are drawn. 
Once the training data for $(w,\yd)$ 
are well chosen, the corresponding $\vmap$ data may be found by solving the optimization problem~\eqref{eq:pred3DVARopt99isolate}. Then, supervised learning may be employed to determine $\thetas$ leading to the approximation 
$\vmap^{\mathsf{a}}(\cdot \, ,\cdot\, ;\thetas) \approx \vmap(\cdot \, ,\cdot)$ from within $\cRt.$
This leads to the following variant on the (cycled) 3DVar\index{3DVar!cycled} algorithm:
\begin{subequations}
 \label{eq:pred3DVAR303}   
\begin{align}
\hat{v}_{j+1} &= \Psi(v_j),\\
v_{j+1} &= \vmap^{\mathsf{a}}\Bigl(\hat{v}_{j+1},\yd_{j+1};\thetas\Bigr).	
\end{align}
\end{subequations}

\subsection{Amortized 4DVar\index{4DVar!amortized}}\label{ssec:A4}

Recall the 4DVar approach to the smoothing problem defined in Section \ref{ssec:4dvar}. 
The methodology defines an estimate of the state $\Vm := \{(\vmap)_0, \ldots, (\vmap)_J\}$
from the observed data $\Yd := \{\yd_1, \ldots, \yd_J\},$ through the solution of an
optimization problem. Emphasizing the dependence of the objective function $\J$ and the solution $\Vm$ on the available data $\Yd$ we have
\begin{subequations}
\label{eq:NobjS}   
\begin{align}
\J(V;\Yd)&  =\reg(V)+\loss(V;\Yd), \label{eq:4dvar_JVY}\\
\Vm(\Yd) & \in \argmin_{V \in \R^{\du(J+1)}} \J(V;\Yd),
\end{align}
\end{subequations}
where $\reg(V)$ and $\loss(V;\Yd)$ are defined in~\eqref{eq:regS} and~\eqref{eq:lossSA}, respectively.
The following two  subsections parallel Subsections
\ref{ssec:amortized_map_obs} and \ref{ssec:amortized_map}, where we discuss
two different approaches to learning the MAP estimator\index{MAP estimator} 
defined by a Bayesian inverse problem.

\subsubsection{Amortized 4DVar Using Only Observations}

Here we discuss an approach to amortize 4DVar based on the methodology of Subsection~\ref{ssec:amortized_map_obs} for amortizing MAP estimation. To do so we make the following data assumption\footnote{Again recall the notation from Remark~\ref{rem:smooth-note}.}:
\begin{dataassumption}\index{Data Assumption}
\label{da:9bb}
We are given $\{(\Yd)^{(n)}\}_{n=1}^N$, which are multiple independent realizations from the marginal distribution $\marg$
on the observed data $\Yd := \{\yd_1, \ldots, \yd_J\}$
defined by \eqref{eq:ddm99}.
\end{dataassumption}

Using the 4DVar objective function~\eqref{eq:4dvar_JVY}, we learn an amortized 4DVar estimator $\Vs(\cdot) := \V(\cdot\,; \thetas)$ by finding the $\thetas$ that minimizes this objective in expectation over the observation realizations:
\begin{subequations}\label{eq:amortized_4DVar}
\begin{align} 
\J_{\mathsf{obs}}(\theta) &= \bbE^{\Yd \sim \marg}  \Bigl[ \J(\V(\Yd;\theta); \Yd) \Bigr],\\
\thetas &\in \argmin_{\theta \in \Theta} \J_{\mathsf{obs}}(\theta).
\end{align}
\end{subequations}
In practice, we use the empirical measure defined by $\{(\Yd)^{(n)}\}_{n=1}^N$
to approximate the expectation defining $\J_{\mathsf{obs}}(\cdot)$ and minimize the resulting
approximate objective function.

\subsubsection{Amortized 4DVar\index{4DVar} Using Supervised Learning}
Here we assume that we have multiple solutions of the 4DVar MAP estimator\index{MAP estimator}
for multiple realizations of the data $\Yd.$ Our goal is to learn the dependence of
$\Vm(\cdot)$ on $\Yd.$ 

\begin{dataassumption}\index{Data Assumption}
    \label{d:9c}
We are given $\{(\Yd)^{(n)}\}_{n=1}^N$, 
comprising multiple independent realizations from the marginal distribution $\kappa$ on the observed data $\Yd := \{\yd_1, \ldots, \yd_J\}$ defined by \eqref{eq:sdm}, \eqref{eq:dm}, together with solutions $(\Vm)^{(n)}=\Vm\bigl((\Yd)^{(n)}\bigr)$ defined by \eqref{eq:NobjS}.
\end{dataassumption}

If the optimization problem \eqref{eq:NobjS} has a unique solution then the $\{\big(\Vm)^{(n)}\big)\}_{n=1}^N$  are distributed according to the pushforward of $\kappa$ under the map from observation $\Yd$ to minimizer $\Vm$.
Given the data $\bigl\{\big((\Yd)^{(n)},(\Vm)^{(n)}\big)\bigr\}_{n=1}^N$ from Data Assumption \ref{d:9c} 
we may apply the methodology of Subsection~\ref{ssec:amortized_map} to
learn the mapping from $\Yd$ to $\Vm$. Let $\thetas$ solve 
(in practice an empirical\index{empirical} approximation of) the minimization problem
\begin{subequations}
\label{eq:MAPL2}
\begin{align} 
\J(\theta) &= \bbE^{\Yd \sim \marg}  \Bigl[ \dd\bigl(\V(\Yd;\theta),\Vm(\Yd)\bigr) \Bigr],\\
\thetas &\in \argmin_{\theta \in \Theta} \J(\theta).
\end{align}
\end{subequations}
Then, we identify the function $\Vs(\cdot) := \V(\cdot\,;\thetas)$ as our solution to the amortized MAP estimation problem.


\subsection{Amortized Posterior Mean\index{posterior!mean estimation} Estimation}\label{ssec:squared_error_min}

In this subsection we consider learning an approximation of the map from observed data to the posterior mean of
either the filter or the smoother. We do this by  minimizing the squared error between a state estimate and a trajectory, in expectation over the joint distribution of states and observations. We will show that this results in a cost function that is minimized at the mean of the filtering or smoothing distribution. This is an amortized\index{amortization} version of the methodology introduced in Subsection~\ref{ssec:mts}, with a particular choice of deterministic scoring rule. It is also an application of the methodology described in Subsection~\ref{ssec:sjoint_energy} for amortized posterior mean estimation in Bayesian inverse problems, applied to filtering and smoothing\index{posterior!mean estimation}.

\subsubsection{Learning the Filter Mean}
We make the following assumption about the available data.

\begin{dataassumption}\index{Data Assumption}
\label{da:9f}
For some given $j \in \{1,\ldots, J\}$ we have access to data set $\bigl\{ \bigl( (\Vd_{j})^{(\sam)}, (\Yd_{j})^{(\sam)} \bigl)  \bigr\}_{\sam \in \{1, \ldots,\Sam\}}$ 
generated as a realization of \eqref{eq:ddm99}.
\end{dataassumption}

The following theorem provides the basis for learning state estimators using the squared error, using Data Assumption \ref{da:9f}.

\begin{theorem}\label{th:squared_error}
 Let $m_j=\Expect[\vd_j|\Yd_j]$ be the mean of the filtering distribution, and let $v(\Yd_j)$ denote any state estimator taking $\Yd_j$ as input. Then, with $\Expect^{(\vd_j, \Yd_j)}$
 denoting expectation under the dynamics/observation model \eqref{eq:sdm}, \eqref{eq:dm}, we have
\begin{equation}
\Expect^{(\vd_j, \Yd_j)} \Bigl[|v(\Yd_j) - \vd_j|^2\Bigr]\geq \Expect^{(\vd_j, \Yd_j)} \Bigl[|m_j - \vd_j|^2\Bigr].
\end{equation}
Furthermore, the inequality is strict if
$\mathbb{P}\bigl(v(\Yd_j)\neq m_j\bigr)>0,$
where probability is taken with respect to the marginal distribution of
$\Yd_j$.
\end{theorem}
\begin{proof}
This is a direct application of Theorem~\ref{th:postmean_opt}.
\end{proof}

\begin{remark}
As mentioned in Remark~\ref{rem:bayes_estimators}, the posterior mean
is the Bayes estimator\index{estimator!Bayes} under squared-error objective;\index{objective!squared-error}
that is, it minimizes the posterior expected squared error. Other objective functions
lead to other Bayes estimators. For example, an absolute-error\index{objective!absolute-error}
objective yields the posterior median for scalar-valued states, whilst a
component-wise absolute-error objective yields the coordinate-wise posterior
medians for vector-valued states. Theorem~\ref{th:squared_error} extends
directly to these settings.
\end{remark}

Thus, having only the information $Y^\dagger_j$, the filter mean minimizes the expected squared error. Hence the squared error may be used as the basis for learning algorithms designed to match the true filter mean. We now minimize over a parameterized class of state estimators $v(\Yd_j; \theta)$, for $\theta\in\Theta \subseteq \R^p$,
\begin{subequations}
    \begin{align}
        \J(\theta) &= \Expect^{(\vd_j, \Yd_j)} \Bigl[|v(\Yd_j; \theta) - \vd_j|^2\Bigr],\\
        \theta^\star &\in \argmin_{\theta\in\Theta} \J(\theta).
    \end{align}
\end{subequations}

\begin{remark}\label{rem:implement_conditional_mean}
    To implement the minimization of $\J(\theta)$, we use an empirical\index{empirical} approximation. In particular, using Data Assumption~\ref{da:9f}
    we may approximate the expectation $\mathbb{E}^{(\vd_j,\Yd_j)}$ using the pairs $\{\big((\Vd_j)^{(n)}, (\Yd_j)^{(n)}\big)\}_{n=1}^N$ generated from the dynamical system \eqref{eq:ddm99}. Assuming ergodicity\index{ergodic} (see Section~\ref{sec:ERG}), moreover, implies that the expectation can instead be approximated using time averages over a long trajectory rather than multiple realizations of the data.
\end{remark}

Note that, in general, the parameterized models we learn will not recover the exact filter mean unless the class used is sufficiently rich to include it. Furthermore empiricalization, as discussed in the preceding remark,
leads to further discrepancy from the true posterior mean.

\subsubsection{Learning the Smoother Mean}

Theorem~\ref{th:squared_error} is readily modified to apply to
the smoothing case. This also gives rise to a Bayes estimator.\index{estimator!Bayes}

\begin{corollary}\label{cor:smoother_squared_error}
    Let $V(\Yd)$ be any state estimator taking $\Yd$ as input. Then
    \begin{equation*}
        \Expect^{(\Vd, \Yd)} \Bigl[|V(\Yd) - \Vd|^2\Bigr]\geq \Expect^{(\Vd, \Yd)} \Bigl[|\Expect[\Vd|\Yd] - \Vd|^2\Bigr].
    \end{equation*}
Furthermore, the inequality is strict if
$\mathbb{P}\Bigl(V(\Yd)\neq\Expect[\Vd|\Yd]\Bigr)>0,$
where probability is taken with respect to the marginal distribution of
$\Yd$.
\end{corollary}

We now again make Data Assumption~\ref{d:9c3dvar}.
Using Corollary~\ref{cor:smoother_squared_error} suggests learning
state estimators $V(\Yd; \theta)$ to estimate the sequence of states 
$\Vd$, for $\theta\in\Theta$, using the minimization problem
\begin{subequations}
    \begin{align}
        \J(\theta) &= \Expect^{(\Vd, \Yd)} \Bigl[|V(\Yd; \theta) - \Vd|^2\Bigr],\\
        \theta^\star &\in \argmin_{\theta\in\Theta} \J(\theta).
    \end{align}
\end{subequations}
As for learning the filter mean, an empirical minimization of $\J$ may be implemented in practice using the data from Data Assumption~\ref{d:9c3dvar}, or through time averages over long trajectories under the assumption of ergodicity.

\section{Variational\index{filtering!variational} Inference Approaches}\label{sec:VA}
In this section we build on the variational inference approach to data assimilation described
in Chapter \ref{ch:LG}. We study amortized\index{filtering!variational} versions of variational approaches 
to both filtering, in Subsection \ref{ssec:AVF}, and smoothing\index{smoothing!variational} 
in Subsection \ref{ssec:AVS}.




\subsection{Amortized Variational Filtering\index{filtering!amortized}\index{filtering!variational}}
\label{ssec:AVF}

Here we build on the material in Subsection \ref{sec:variational_filtering0} to include the amortization afforded
by working under Data Assumption~\ref{da:9bb}. To that end we let $\mkappa_{j+1}$ denote the marginal of $\kappa$ on
$\Yd_{j+1}.$ As in Subsection \ref{sec:variational_filtering0} we let $\An_{\theta}(\cdot\,;\yd)$ denote a $\theta$-parameterized 
family of approximate analysis maps, recalling prediction and analysis maps $\Pred$ and $\An(\cdot\,;\yd)$ defined 
in Subsection \ref{ssec:filtering}.  Using prediction and approximate analysis steps we may define a sequence, 
in $j$, of approximate filtering distributions $q_j(\theta)$. Arguing identically to Lemma \ref{lem:willneed}
shows that the algorithms we study have the property that $q_j(\theta)$ depends only on $\Yd_j,$ the desired filtering
structure. Throughout this subsection, we leave this dependence implicit in the notation. Again, as in Subsection \ref{sec:variational_filtering0}, our goal is to choose $\theta \in \Theta \subseteq \R^p$ so that $q_j(\theta)$ approximates the filtering distribution $\pi_j$.

Generalizing \eqref{eq:var_smoothinga99} to include amortization over multiple data instances, we obtain
\begin{subequations}
\label{eq:var_smoothinga99_AM}
\begin{align} 
\qs_0 &=\pi_0,\\
\J'_{j+1}(\theta; \Yd_{j+1}) &= \dkl\bigl(q_{j+1}(\theta; \Yd_{j+1})\|\Pred \qs_j(\Yd_j)\bigr) - 
    \mathbb{E}^{{q_{j+1}(\theta; \Yd_{j+1})}}[\log \mathbb{P}(\yd_{j+1} | \cdot)] \\
    q_{j+1}(\theta; \Yd_{j+1}) &= \An_{\theta}(\Pred \qs_j(\Yd_j); \yd_{j+1}),\\
\J_{j+1}(\theta) &=  \mathbb{E}^{\Yd_{j+1} \sim \mkappa_{j+1}}\bigl[\J'_{j+1}(\theta; \Yd_{j+1})\bigr],\\
\thetas_{j+1} &\in \argmin_{\theta \in \Theta} \J_{j+1}(\theta),\\
\qs_{j+1}(\Yd_{j+1}) & = \An_{\thetas_{j+1}}(\Pred \qs_j(\Yd_j); \yd_{j+1}).
\end{align}
\end{subequations}
In practice, the expectation over $\mkappa_{j+1}$ will be approximated empirically, using Data Assumption~\ref{da:9bb}.

Again, as in Subsection \ref{sec:variational_filtering0}, we may invoke an ergodic hypothesis and define
\begin{equation}
\label{eq:st1_AM}    
\thetas  = \frac{1}{J} \sum_{j=1}^J  \thetas_j
\end{equation}
and consider the approximate filter
\begin{equation}
\label{eq:pran2_AM}
    q_{j+1} = \An_{\thetas}(\Pred q_j; \yd_{j+1}),\quad q_0 = \pi_0.
\end{equation}
for $q_j \approx q_j'.$ This approximate filter may then be used on new data sequences
$\Yd$, different from the specific sequences appearing in Data Assumption~\ref{da:9bb}
and used to determine $\thetas.$ Because of the amortization across both time $j$ and
data realizations $n$ this is likely to be an improvement, in terms of generalization
to new observation sequences, over the methodology in Subsection \ref{sec:variational_filtering0}.

A variant on the methodology following from \eqref{eq:var_smoothinga99_AM} is to use the 
following averaged minimization problem
\begin{subequations}
\label{eq:var_smoothinga9_AM}
\begin{align} 
\thetas &\in \argmin_{\theta \in \Theta} \frac{1}{J} \sum_{j=0}^{J-1} \J_{j+1}(\theta),\\
\J_{j+1}(\theta) &=  \mathbb{E}^{\Yd_{j+1} \sim \mkappa_{j+1}} \bigl[\dkl\bigl(q_{j+1}(\theta)\|\Pred q_j(\theta)\bigr) - 
    \mathbb{E}^{{q_{j+1}(\theta)}}[\log \mathbb{P}(\yd_{j+1} | \cdot)]\bigr],\\
    q_{j+1}(\theta) &= \An_{\theta}(\Pred q_j(\theta); \yd_{j+1}),\quad q_0 = \pi_0.
\end{align}
\end{subequations}

\begin{example}
Consider making the choice that $q_j$ is Gaussian $\Nc(m_j,C_j)$.  To implement either of the methodologies outlined above
we can proceed as in Section \ref{sec:filteringuq}, using $\Pred_{\mathsf{G}}$, as defined by (\ref{eq:kalman_fixed_cov2}a).
This forces the predicted filtering distribution to remain
on the manifold of Gaussians, after each prediction step. We then replace $\Pred$ in \eqref{eq:var_smoothinga99_AM} or (\ref{eq:var_smoothinga9_AM}c) by $\Pred_{\mathsf{G}}$.

We can then define $\An_{\theta}$ appearing in (\ref{eq:var_smoothinga9_AM}c) by the Gaussian update
rules (\ref{eq:kalman_fixed_cov2}b), (\ref{eq:kalman_fixed_cov2}c); parameter $\theta$ is the
gain matrix $K$. In the case of $\An_{\theta_{j+1}}$ appearing in (\ref{eq:var_smoothinga99_AM}d) 
we may again use the Gaussian updates  (\ref{eq:kalman_fixed_cov2}b), (\ref{eq:kalman_fixed_cov2}c), but
with $K \mapsto \theta_{j+1}:=K_{j+1}.$ 
\end{example}

\subsection{Amortized Variational Smoothing\index{smoothing!amortized}\index{smoothing!variational}} 
\label{ssec:AVS}
We may also be interested in learning a parametric model to capture dependence of an approximate smoothing distribution on the data $\Yd$, an amortized version of the ideas in  Section~\ref{sec:variational_smoothing0}. To this end, in this subsection we work under Data Assumption~\ref{da:9bb}. We generalize the previous subsection and seek to find smoothing
distributions, dependent on data $\Yd$, from the set $\cQ^J\subset\cP^J$. To define $\cQ^J$ we consider minimizing over the $\theta$-parameterized family of maps
\[
\mathcal{S}_\theta:\R^{Jk}\to\cQ^J,
\qquad
\mathcal{S}_\theta(\Yd)=q(\cdot\,;\Yd,\theta),
\]
where $\theta\in\Theta\subseteq\R^p$:
\begin{subequations}
\label{eq:var_smoothingb2}
\begin{align} 
\J(q) &= \mathbb{E}^{\Yd \sim \kappa} \Bigl[\dkl(q(\cdot\,; \Yd)\|\rho) - \mathbb{E}^{q(\cdot\,; \Yd)} \bigl[\log \like(\Yd|\cdot) \bigr] \Bigr],\\
\thetas &\in \argmin_{\theta \in \Theta} \J\bigl(\mathcal{S}_\theta\bigr).
\end{align}
\end{subequations}
Then, for new, previously unseen data $(\Yd)'$, we choose our approximate smoothing distribution to be $\qs=\mathcal{S}_{\thetas}((\Yd)').$ We emphasize that the resulting  smoothing distribution
does not enforce the temporal dependence in filtering where each state marginal depends on past observations. In particular, 
in the notation of Section \ref{sec:fss}, we typically expect that $\qs \in \cP^J\backslash \cC^J.$ 

We consider an example of amortized variational smoothing which is similar
in approach to that outlined for Bayesian inversion in Section~\ref{sec:vbay}; see Example~\ref{ex:GaussianVI_IP} for Gaussian variational inference with a specific likelihood.
However, we propose a different two-stage approach to the optimization problem;
this two-stage approach could have been adopted in Example~\ref{ex:GaussianVI_IP} and likewise
the one-stage approach of Example~\ref{ex:GaussianVI_IP} could be adopted here.

\begin{example}
\label{ex:4DVARgoesGaussianAM}
In Section~\ref{ssec:4DVARgoesGaussian} we discuss the use of
variational smoothing to find a Gaussian approximation centered on the
MAP estimator. We now generalize this approach to the amortized setting.
We use a two-stage procedure. First, using the ideas of
Subsection~\ref{ssec:A4}, we determine an amortized MAP estimator
$\V(\Yd;\thetas)$ that captures its dependence on the observations
$\Yd$. We then use amortized variational smoothing to determine an
observation-dependent covariance for the Gaussian approximation.

To this end, let $\Theta_{\mathsf A}\subseteq\R^r$ and introduce
functions
\[
\Acs(\cdot\,;\vartheta):
\R^{Jk}\to
\R^{(J+1)d\times(J+1)d},
\qquad
\vartheta\in\Theta_{\mathsf A}.
\]
For example, $\Acs(\Yd;\vartheta)$ may be restricted to be lower
triangular. We require that
\[
\det\bigl(\Acs(\Yd;\vartheta)\bigr)>0
\]
for every $\Yd\in\R^{Jk}$ and $\vartheta\in\Theta_{\mathsf A}$. Define
the map $\mathcal{S}_\vartheta:\R^{Jk}\to\cP^J$ by
\[
\mathcal{S}_\vartheta(\Yd)
=
\cN\Bigl(
\V(\Yd;\thetas),
\Acs(\Yd;\vartheta)\Acs(\Yd;\vartheta)^\top
\Bigr).
\]
We identify the variational approximation by solving
\begin{equation}
\label{eq:var_smoothingb22}
\vartheta^\star
\in
\argmin_{\vartheta\in\Theta_{\mathsf A}}
\J\bigl(\mathcal{S}_\vartheta\bigr),
\end{equation}
where $\J$ is defined in~\eqref{eq:var_smoothingb2}. Then, for new,
previously unseen data $(\Yd)'$, we define the approximate smoothing
distribution by
$\qs=\mathcal{S}_{\vartheta^\star}\bigl((\Yd)'\bigr).$
\end{example}

\section{Learning via Scoring Rule Minimization}
\label{sec:SPSR}

In this section we use objectives based on expected strictly proper
scores to learn the filtering distributions. 
The methodology may be viewed as an application to filtering of the methodology for amortizing posterior estimation in Bayesian inverse problems that is described in Subsection~\ref{ssec:sjoint_energy}. By invoking ergodicity to approximate the
expectation we will make a connection to the methodology introduced in Subsection~\ref{ssec:sru}, which does not
involve amortization over multiple realizations from the dynamics-observation model \eqref{eq:sdm}, \eqref{eq:dm}, 
but just uses one long time-series. In Subsection \ref{ssec:general} we introduce the general formulation,
specifying to transport maps in Subsection \ref{ssec:spsr_transport}. 

\subsection{General Formulation} \label{ssec:general}

The following theorem provides the basis for learning filters using strictly proper scoring rules\index{scoring rule!proper} from Definition \ref{def:proper_score}.

\begin{theorem}\label{theorem:strict_proper_filter}
 Consider a strictly proper scoring rule\index{scoring rule!proper}  $\mS(\cdot, \cdot): \cP(\R^{\du}) \times \R^{\du} \to \R.$  Let $\pi_j(\vd_j)=\mathbb{P}(\vd_j|\Yd_j)$ be the filtering distribution
 on $\R^{\du}$ and let $q(\cdot\,;\Yd_j)$ denote any distribution
 on $\R^{\du}$ parameterized by $\Yd_j.$  Then, with $\mathbb{E}^{(\vd_j, \Yd_j)}$
 denoting expectation under the dynamics/observation model \eqref{eq:sdm}, \eqref{eq:dm}, we have
\begin{equation}\label{eq:scoreOBJ}
\mathbb{E}^{(\vd_j, \Yd_j)} \Bigl[\mS(q(\cdot\,;\Yd_j), \vd_j)\Bigr]\geq \mathbb{E}^{(\vd_j, \Yd_j)} \Bigl[\mS(\pi_j, \vd_j)\Bigr].
\end{equation}
Furthermore, the inequality is strict if
$\mathbb{P}\bigl(q(\cdot\,;\Yd_j)\neq\pi_j\bigr)>0,$
where probability is taken with respect to the marginal distribution of
$\Yd_j$.
\end{theorem}
\begin{proof}
This is a direct application of Theorem~\ref{th:scoring_rule_joint}.
\end{proof}

Thus, having only the information $Y^\dagger_j$, the filtering distribution minimizes the expected score. Hence this expected score may be used as the basis for learning algorithms designed to match
the true filter, using only samples from the joint distribution of $\vd_j$ and $\Yd_j$. 
Certain scoring rules are amenable to sample-based implementation, which will be convenient in what follows.

We now define a learning problem for the filtering distributions. Motivated by \eqref{eq:pran2}, we parameterize a set of approximate filters $q$ of the form:
\begin{equation} \label{eq:pran3}
    q_{j+1}(\theta) = \An_\theta(\Pred q_j(\theta); \yd_{j+1}),\quad q_0 = \pi_0,
\end{equation}
where $\An_\theta$ denotes an approximation of the true analysis\index{analysis} map $\An$, parameterized by $\theta \in \Theta$; 
map $\An_\theta$ depends on the approximate forecast distribution and the observation $\yd_{j+1}$ and computes the filtering distribution. The following result shows that if set of parameters $\Theta$ is large enough to represent the true analysis map given in~\eqref{eq:filter_bayes_inference}, then 
the learning problem that minimizes the scoring rule will recover the true filtering distribution.

\begin{corollary} \label{cor:scoring_rule_objective}
    Consider the distribution $q_j$ evolving under the $\theta$-parameterized algorithm
    \eqref{eq:pran} and define the optimization problem
\begin{align*}
    \J(\theta) &= \mathbb{E}^{(\vd_j,\Yd_j)} \Bigl[\mS(\An_\theta(\Pred \pi_{j-1}(\theta); y^\dagger_{j}), \vd_j)\Bigr],\\
    \thetas &\in \argmin_{\theta\in\Theta} \J(\theta).
\end{align*}
Then, if there is some $\theta \in \Theta$ for which $\An_\theta = \An$ and $j>0$, we have that
\begin{equation*}
    \J(\thetas) = \mathbb{E}^{(\vd_j,\Yd_j)} \Bigl[\mS(\mathbb{P}(\vd_j|\Yd_j), \vd_j)\Bigr].
\end{equation*}
\end{corollary}
\begin{proof}
This follows from Theorem~\ref{theorem:strict_proper_filter} by using \eqref{eq:pran3} to replace $q(\cdot\,;\Yd_j)$
by $q_j(\cdot\,;\Yd_j),$ computed from the true filtering distribition $\pi_{j-1}$ at time $j-1.$
\end{proof}

In principle we could implement an empirical approximation of the objective function implied
by the preceding corollary, under Data Assumption~\ref{da:9f}. In practice this is complicated
by the dependence of $q_j(\theta)$ on all preceding filtering steps. Implementation of the methodology requires empirical minimization. To this end we introduce:

\begin{dataassumption}\index{Data Assumption}
\label{da:9femp}
For some $J \in \mathbb{N}$ we have access to one trajectory pair $\{(\Vd_{J}, \Yd_{J})\}$, 
generated as a realization of \eqref{eq:ddm99}.
\end{dataassumption}

Under this data assumption, invoking an ergodic hypothesis as discussed in Remark~\ref{rem:implement_conditional_mean} 
leads to the empirical optimization problem
\begin{align*}
    \J^J(\theta) &= \frac{1}{J}\sum_{j=1}^{J} \Bigl[\mS(\An_\theta(\Pred q_{j-1}(\theta); y^\dagger_{j}), \vd_j)\Bigr],\\
    \thetas &\in \argmin_{\theta\in\Theta} \J^J(\theta).
\end{align*}
This is precisely the form we have already introduced in Subsection~\ref{ssec:sru}. In the bibliographic remarks in 
Section \ref{sec:Bib_learning_filters2} we refer to theory justifying this form of objective function.

\subsection{Transport-Based Formulation}\label{ssec:spsr_transport}

We now consider using strictly proper scoring rules to learn a parameterized transport map $T \colon \R^\du \times \R^\dy \times \Theta \rightarrow \R^\du$ that approximately maps a sample from the forecast distribution and an observation to a sample from the filtering distribution. This is a special case of the transport-based methodology for learning amortized prior-to-posterior maps using strictly proper scoring rules\index{score!proper} from Subsection~\ref{ssec:sjoint_energy}.

This can be achieved by writing the action of the parameterized forecast-to-analysis map by pushing forward the forecast distribution under an observation-dependent transport map: 
\begin{equation}
    \An_\theta(\widehat{\pi}, y) = T(\cdot, y; \theta)_\sharp \widehat{\pi}.
\end{equation}
This results in 
the following sequential ensemble algorithm, which alternates the forecast step with an analysis step (similarly to \eqref{eq:fakeobs3}) that is implemented using transport:
\begin{subequations}
\begin{align}
\widehat{v}_{j+1}^{(\sam)} &= \Psi(v_{j}^{(\sam)})+\xi^{(\sam)}_{j}, \quad \sam=1,\ldots,\Sam, \\
v_{j+1}^{(\sam)} &= T\Bigl(\widehat{v}_{j+1}^{(\sam)},\yd_{j+1};\theta\Bigr),\quad \sam=1,\ldots,\Sam.
\end{align}
\end{subequations}
If we assume that the $\{v_j^{(n)}\}_{n=1}^N$ are drawn 
from an approximate filter $\pi_j^N$ at time $j$,
then writing the analysis ensemble at time $j+1$ as
\begin{equation*}
    \pi^N_{j+1} = \frac{1}{N} \sum_{n=1}^N \delta_{v_{j+1}^{(n)}},
\end{equation*}
the optimization problem in Corollary~\ref{cor:scoring_rule_objective} becomes
\begin{align*}
    \J(\theta) &= \Expect^{(\vd_{j+1}, \Yd_{j+1})}[\mS(\pi_{j+1}^N, \vd_{j+1})],\\
    \thetas &\in \argmin_{\theta \in \Theta} \J(\theta).
\end{align*}
Implementing this optimization problem requires a scoring rule that can be evaluated for empirical measures; kernel scores\index{score!kernel} (Definition~\ref{def:kernel_score}), which include the energy score\index{score!energy} (Definition~\ref{d:energy}), are examples of such scoring rules. In practice, the expectation over $(\vd_{j+1}, \Yd_{j+1})$ can be approximated using ergodicity as discussed at the end of the preceding subsection. 

\section{Online Transport Approaches to Filtering and Smoothing\index{transport!amortized}}
\label{sec:TA}


In this section we seek transports that map to each filtering distribution $\pi_{j+1}$, as in Subsections~\ref{ssec:PLA1} and~\ref{ssec:PLA2}. In this case, however, we seek a transport for each assimilation step that is computed \emph{online}\index{online!algorithm}, meaning that it only uses data available at the given assimilation step. This is in contrast to earlier Sections~\ref{sec:SE}--\ref{sec:SPSR} in this chapter, which approximate the filtering distribution \emph{offline}\index{offline!algorithm} -- using variational inference in Section~\ref{sec:VA} and strictly proper scoring rules in Section~\ref{sec:SPSR} --
by amortizing over multiple sequences of observations. These latter approaches update the filtering distribution on the basis of online observations. They assume that observations given online to evaluate the approximation are drawn from the same distribution as the data given offline for learning. On the other hand, the approximations we construct in this subsection only use online data and do not make this assumption. This may be particularly desirable in generalizations of \eqref{eq:ddm99} where the observation likelihood or model dynamics are time-dependent.

We start, in Subsection \ref{ssec:general2}, with some general considerations.
In Subsection~\ref{sec:ED_datadependence_opt} we minimize general strictly proper scoring rules, with the energy score\index{energy!score} as an example, 
as in Subsection~\ref{ssec:PLA1} where dependence on the observed data is not learned; in Subsection~\ref{sec:MLE_datadependence_opt} 
we leverage the transport methodology developed in Chapter~\ref{ch:transport}, based on maximum likelihood.\index{maximum likelihood}
The ideas in Subsection~\ref{sec:ED_datadependence_opt} are also related to Subsection~\ref{ssec:sjoint_energy}, where amortized transports are learned to sample the posterior distributions of Bayesian inference problems, using strictly proper scoring rules.\index{scoring rule!strictly proper} 
Lastly, we extend this framework of learning transports to smoothing in Subsection~\ref{sec:OPFT}.

\subsection{General Considerations} \label{ssec:general2}

Here, we will focus on approaches to perform inference at each assimilation step using a $j$-dependent map 
$T_{j+1}:\R^\du\times\R^\dy\times\Theta\to\R^\du$ 
that \emph{explicitly} depends on both the predicted state $\widehat{v}_{j+1} \in \R^\du$ and the true observation $y_{j+1}^\dagger \in \R^\dy$. A separate map is estimated at each time step, and thus we emphasize this dependence on $j$ with the subscript.   
This results in the following sequential algorithm, which alternates the forecast step with an analysis step (similarly to \eqref{eq:fakeobs3}) that is implemented using transport:
\begin{subequations}
\label{eq:fakeobs4}
\begin{align}
\widehat{v}_{j+1}^{(\sam)} &= \Psi(v_{j}^{(\sam)})+\xi^{(\sam)}_{j}, \quad \sam=1,\ldots,\Sam, \\
v_{j+1}^{(\sam)} &= T_{j+1}\Bigl(\widehat{v}_{j+1}^{(\sam)},\yd_{j+1};\theta_{j+1}\Bigr),\quad \sam=1,\ldots,\Sam.\label{eq:fakeobs42}
\end{align}
\end{subequations}
where $T_{j+1}$ depends on parameters $\theta_{j+1} \in \Theta \subset \R^p$.
Our goal is to find the parameters of the map $T_{j+1}$, so that
\begin{equation} \label{eq:sample_pushforward_map_data_dependence}
\widehat{v}_{j+1} \sim \widehat\pi_{j+1} \Rightarrow T_{j+1}(\widehat{v}_{j+1},\yd_{j+1};\theta_{j+1}) \sim \pi_{j+1},
\end{equation}
where we have emphasized the dependence of the map and its (separate) parameters $\theta_{j+1}$ on the index $j+1$.

To achieve this goal without samples from the filtering distribution, we will learn a map $T_{j+1}(\cdot,\cdot;\theta) \colon \R^\du \times \R^\dy \rightarrow \R^\du$ 
that amortizes over the observations and can sample from the conditional distribution $\mathbb{P}(v_{j+1}|\Yd_j, y_{j+1})$
corresponding to the prior defined by the forecast distribution $\widehat\pi_{j+1}(v_{j+1})$ and the likelihood defined by $\like(y_{j+1}|\widehat{v}_{j+1})$ for \emph{any} observation $y_{j+1} \in \R^{\dy}$. That is,
we wish to approximately satisfy
$$\widehat{v}_{j+1} \sim \widehat\pi_{j+1} \Rightarrow T_{j+1}\Bigl(\widehat{v}_{j+1},y_{j+1};\theta_{j+1}\Bigr) \sim \mathbb{P}(v_{j+1}|\Yd_j, y_{j+1}),$$
for any $y_{j+1} \in \R^{\dy}.$ 
We note that this is a different purpose for amortization than earlier sections in this chapter, which use amortization to learn a map that can generalize to new true observations $y_{j+1}^\dagger$ across time steps $j$. Here, we learn the map's dependence under the following data assumption which assumes only access to an ensemble from the joint distribution of state $\widehat{v}_{j+1}$ and synthetic observations $\widehat{y}_{j+1}$ at step $j+1$. The synthetic observations are generated from the state using the observation model $\widehat{y}_{j+1}^{(\sam)} = h(\hv_{j+1}^{(\sam)})+\eta^{(\sam)}_{j+1}$ for $\sam=1,\ldots,\Sam$. In this section, the ability to make likelihood evaluations is not assumed, thereby making the approach \emph{likelihood-free}\index{likelihood!free}. 


\begin{dataassumption}\index{Data Assumption}\label{da:99b} We are given pairs of i.i.d.\thinspace samples $\{(\hv_{j+1}^{(n)},{\hy}_{j+1}^{(n)})\}_{n=1}^N$ from the joint distribution $\joint_{j+1}(v_{j+1},y_{j+1}) \coloneqq \widehat\pi_{j+1}(v_{j+1})\like(y_{j+1}|v_{j+1})$.
\end{dataassumption}

\begin{remark}
In practice we will estimate the map's parameters, meaning that analysis step~\eqref{eq:sample_pushforward_map_data_dependence} will only generate approximate posterior samples. As a result, the forecast samples at the next step will not be exact forecast samples from $\joint_{j+1}(v_{j+1},y_{j+1})$ because $\widehat\pi_{j+1}(v_{j+1})$ will have been approximated.
This will typically produce an error accumulation across time. We do not study this error
propagation here. For simplicity we will assume, in this section, 
that we can sample $\joint_{j+1}(v_{j+1},y_{j+1})$ exactly, as in Data Assumption \ref{da:99b}.
\end{remark}

In the following two subsections we will present two formulations to approximate map $T_{j+1}$ and hence
to approximately enforce \eqref{eq:sample_pushforward_map_data_dependence}. 

\subsection{Online Minimization of Scoring Rules} 
\label{sec:ED_datadependence_opt}

The filtering distribution $\pi_{j+1}$ is found from
Bayes Theorem~\ref{t:bayes} with the prior given by
the forecast $\widehat\pi_{j+1}$ and the likelihood model given by $\like(\yd_{j+1}|\cdot)$. 
Using the pushforward notation, our goal is to choose $\theta_{j+1}$ to achieve the pushforward condition
\begin{equation}
 \label{eq:www}   
T_{j+1}(\cdot,\yd_{j+1};\theta_{j+1})_\sharp \widehat\pi_{j+1} = \pi_{j+1}.
\end{equation}

We introduce the approach of this subsection with the following theorem.
\begin{theorem}\label{th:spsr_transport}
 Consider a strictly proper scoring rule\index{scoring rule!proper}  $\mS(\cdot, \cdot): \cP(\R^{\du}) \times \R^{\du} \to \R.$  Let $\pi_{j+1}(v_{j+1}|y_{j+1})=\mathbb{P}(v_{j+1}|\Yd_{j},y_{j+1})$ be the filtering distribution
 on $\R^{\du}$ given an observation $y_{j+1} \in \R^{\dy}$ at time $j+1$ and let $q(\cdot\,;y_{j+1})$ denote any distribution
 on $\R^{\du}$ parameterized by $y_{j+1}.$ Then,
\begin{equation}
\mathbb{E}^{(v_{j+1}, y_{j+1})\sim \gamma_{j+1}} \Bigl[\mS(q(\cdot\,;y_{j+1}), v_{j+1})\Bigr]\geq \mathbb{E}^{(v_{j+1}, y_{j+1})\sim \gamma_{j+1}} \Bigl[\mS(\pi_{j+1}(\cdot|y_{j+1}), v_{j+1})\Bigr].
\end{equation}
Furthermore, equality is achieved if $q(\cdot\,;y_{j+1}) = \pi_{j+1}(\cdot|y_{j+1})$ for $\kappa_{j+1}$-almost every $y_{j+1}$.
\end{theorem}
\begin{proof}
Noting that $\joint_{j+1}(v_{j+1},y_{j+1}) =
\pi_{j+1}(v_{j+1}|y_{j+1})\marg_{j+1}(y_{j+1})$, 
this follows similarly to the proof of Theorem~\ref{th:scoring_rule_joint}.
\end{proof}

By parameterizing the class of approximate distributions as
\begin{equation*}
    q(\cdot; y_{j+1}, \theta) = T_{j+1}(\cdot,y_{j+1};\theta)_\sharp \widehat{\pi}_{j+1},
\end{equation*}
we obtain the following objective function for the transport map parameters, motivated by Theorem~\ref{th:spsr_transport}:
\begin{align*}
    \J_{j+1}(\theta) &= \mathbb{E}^{(v_{j+1},y_{j+1})\sim \gamma_{j+1}} \Bigl[\mS(T_{j+1}(\cdot,y_{j+1};\theta)_\sharp \widehat{\pi}_{j+1}, v_{j+1})\Bigr].
\end{align*}
This allows us to use the joint ensemble for $\gamma_{j+1}$ arising from Data Assumption~\ref{da:99b} to define the following empirical\index{empirical} optimization problem for the map parameters:
\begin{subequations}
    \begin{align}
    \J^N_{j+1}(\theta) &= \frac{1}{N}\sum_{n=1}^N
    \mS(\pi^N_{j+1}(\widehat{y}_{j+1}^{(n)};\theta),
\widehat{v}_{j+1}^{(n)}),
    \label{eq:spsr_JN}\\
    \thetas_{j+1} &\in \argmin_{\theta\in\Theta} \J^N_{j+1}(\theta),
    \end{align}
\end{subequations}
where $\pi^N_{j+1}$  represents an empirical measure for the pushforward samples, which depends on the map parameters and an observation $\widehat{y}_{j+1}^{(n)}$. That is, 
\begin{equation} \label{eq:emp_measure_pushforward_filtering}
\pi^N_{j+1}(\widehat y_{j+1}^{(n)};\theta)
=
\frac{1}{N-1}\sum_{\substack{m=1 \\ m \neq n}}^N
\delta_{
T_{j+1}(\widehat v_{j+1}^{(m)},
\widehat y_{j+1}^{(n)};\theta)}
.
\end{equation}

\begin{remark}
   The exclusion of $n=m$ in~\eqref{eq:emp_measure_pushforward_filtering} is needed because $\widehat v_{j+1}^{(m)}$ and $\widehat y_{j+1}^{(m)}$ are paired and therefore dependent, whereas the forecast sample entering the pushforward distribution should be independent of the conditioning pair. 
\end{remark}

As in Subsection~\ref{ssec:spsr_transport} a scoring rule that can be evaluated on empirical measures,
defined by Data Assumption~\ref{da:99b}, must be used to carry out the minimization.

\begin{example}\label{ex:online_spsr}
We employ ideas similar to those underlying empirical approximation of the energy distance and MMD in
Remarks \ref{rem:mmd&e} and \ref{rem:jic}.
    Picking $\mS$ to be the energy score (Definition~\ref{d:energy}),\index{score!energy} we obtain the objective for~\eqref{eq:spsr_JN} given by
    \begin{align*} 
    \J_{j+1}^N(\theta) &= \frac{1}{N(N-1)}\sum_{\substack{m,n=1 \\ m \neq n}}^N |T_{j+1}(\widehat{v}_{j+1}^{(n)},\widehat{y}_{j+1}^{(m)};\theta) - \widehat{v}_{j+1}^{(m)}| \\
    &\quad - \frac{1}{2N(N-1)(N-2)}\sum_{\substack{l,m,n=1 \\ l \neq m, l \neq n, m \neq n}}^N |T_{j+1}(\widehat{v}_{j+1}^{(n)},\widehat{y}_{j+1}^{(l)};\theta) - T_{j+1}(\widehat{v}_{j+1}^{(m)},\widehat{y}_{j+1}^{(l)};\theta)|.
    \end{align*}
    The expectations in each term of the objective function involve a tensor product of distributions for the states and observations. These expectations are estimated from a single ensemble of paired states and observations by permuting the pairings. For instance, $(\widehat{v}_{j+1}^{(n)}, \widehat{y}_{j+1}^{(l)}) \sim \widehat\pi_{j+1} \otimes \marg_{j+1}$ for $n \neq l$.
\end{example}

\subsection{Maximum Likelihood Estimation} \label{sec:MLE_datadependence_opt}\index{maximum likelihood estimation}

In this subsection we consider an alternative approach to construct an observation-dependent transport. To do so, we first identify a transport map  
that pushes forward 
the joint distribution $\gamma_{j+1}(y,v)$ to the filtering distribution $\pi_{j+1}(v)$ 
by composing transport maps that can be estimated using the maximum likelihood approach introduced in Chapter~\ref{ch:data-dependence}. We continue to work under Data Assumption~\ref{da:99b}.
See Remark~\ref{rem:discuss0} for the connection between KL  divergence\index{divergence!Kullback--Leibler} minimization and maximum likelihood\index{maximum!likelihood} and Section~\ref{sec:transport2} for more information on density estimation via transport.

Recall from Remark~\ref{rem:smooth-note} that $\kappa_{j+1}(y)$ 
denotes the marginal of joint distribution $\gamma_{j+1}(y,v)$.
Let $\refT(v)$ be a reference density on $\Ru$, such as a standard Gaussian. The following theorem shows how to construct a transport that pushes forward the joint distribution
on the state and observation, $\joint_{j+1}$,
to the filtering distribution $\pi_{j+1}$, based on a map that pushes forward any conditional distribution $\mathbb{P}(v|y)$ of $\gamma_{j+1}$ to the same reference distribution.

\begin{theorem}
\label{thm:FilteringMapComposition}
Let
$S_{j+1}:\R^\dy\times\R^\du\to\R^\du$
be a transport map such that
$S_{j+1}(y,\cdot):\R^\du\to\R^\du$
is invertible for every $y\in\R^\dy$. We denote its inverse by
$S_{j+1}(y,\cdot)^{-1}$.
Suppose that, for every $y\in\R^\dy$, the map
$S_{j+1}(y,\cdot)$ pushes forward the conditional distribution
$\mathbb{P}(v|y)$ under $\gamma_{j+1}$ to the reference distribution
$\refT$. Then, for every $\yd_{j+1}\in\R^\dy$, the map
\begin{equation}
\label{eq:ComposedMapFiltering}
T_{j+1}(v,y,\yd_{j+1})
\coloneqq
S_{j+1}(\yd_{j+1},\cdot)^{-1}
\bigl(S_{j+1}(y,v)\bigr)
\end{equation}
pushes forward $\gamma_{j+1}$ to the filtering distribution
$\pi_{j+1}$ corresponding to the observation $\yd_{j+1}$.
\end{theorem}

\begin{proof}
Conditional on $y$, the random variable $S_{j+1}(y,v)$ has
distribution $\refT$ given $v$ drawn from the conditional distribution $\mathbb{P}(v|y)$ under $\gamma_{j+1}$. Applying
$S_{j+1}(\yd_{j+1},\cdot)^{-1}$ therefore produces a random variable
with distribution $\mathbb{P}(v|\yd_{j+1})$, which is the desired
filtering distribution.
\end{proof}


 
Using the procedure in Subsection~\ref{ssec:sjoint},
and noting Remark~\ref{rem:worthnoting} in particular, we identify the transport map $S(\cdot,\cdot\,;\theta)$, parameterized by $\theta \in \Theta \subseteq \R^p$ as the solution of the maximum likelihood problem
\begin{subequations}\label{eq:MAP_MLE_Population}
    \begin{align} 
    \J(\theta) &= \mathbb{E}^{(v_{j+1},y_{j+1}) \sim \gamma_{j+1}} \left[\log S_{j+1}(y_{j+1},\cdot\,;\theta)^{-1}_\sharp \refT(v_{j+1}) \right] \\
    \thetas_{j+1} &\in \argmax_{\theta \in\Theta} \J(\theta).
    \end{align}
\end{subequations}
Given the joint ensemble of states and observations, as defined in Data Assumption~\ref{da:99b}, we can define our desired transport map as the maximizer of the empirical objective:
\begin{equation} \label{eq:MAP_MLE_Composed}
\thetas_{j+1} \in \argmax_{\theta \in\Theta} \frac{1}{N} \sum_{n=1}^N \log S_{j+1}(\widehat{y}_{j+1}^{(n)},\cdot\,;\theta)^{-1}_\sharp \refT(\widehat{v}_{j+1}^{(n)}).
\end{equation}
Using Theorem~\ref{thm:FilteringMapComposition}
we define the transport
$$T_{j+1}(v,y,\yd_{j+1};\thetas_{j+1}) \coloneqq  S_{j+1}(\yd_{j+1},\cdot\,;\thetas_{j+1})^{-1}(S_{j+1}(y,v;\thetas_{j+1})).$$
The transport is used for the sample update in (\ref{eq:fakeobs2}c):
$$v_{j+1}^{(n)} =  S_{j+1}(\yd_{j+1},\cdot\,;\thetas_{j+1})^{-1}(S_{j+1}(\widehat{y}_{j+1}^{(n)},\widehat{v}_{j+1}^{(n)};\thetas_{j+1})), \qquad n = 1,\dots,N.$$
We note that the dependence of each particle update on the entire ensemble is through
the parameter $\thetas_{j+1}.$

\begin{remark} The joint-to-posterior map depends on both the true observation $\yd_{j+1}$ and synthetic observations $\widehat{y}_{j+1}^{(n)}$ at time $j+1$. For a fixed $\yd_{j+1}$, the transformation from the forecast to analysis state can be seen as a stochastic transport due to the randomness in the synthetic observations $\widehat{y}_{j+1}^{(n)}$. 
In contrast, for a fixed observation $\yd_{j+1}$, the transport in
Subsection~\ref{sec:ED_datadependence_opt} is a deterministic map that
pushes forward $\widehat\pi_{j+1}$ to $\pi_{j+1}$.
\end{remark}

While the framework above is quite general, in practice we may want to seek transports within a parameterized family of functions with a particular structure. The following result shows that these filters based on composed maps are related to a mean-field version of the ensemble Kalman filter, discussed in the bibliography Section~\ref{sec:Bib_learning_filters2}.

\begin{theorem} \label{thm:mean-field-enkf}
Assume that the reference distribution is
$\refT=\Nc(0,I_{\du})$. Let
\[
S(y,v;\theta)=A(v+By+c)
\]
be an affine transport map that is invertible with respect to $v$ for
every $y$, where $\theta=(A,B,c)$. Assume that the optimal parameter is
found by solving~\eqref{eq:MAP_MLE_Population} over
\[
\Theta
=
\left\{
A\in\R^{\du\times\du}_{>},
\;
B\in\R^{\du\times\dy},
\;
c\in\R^\du
\right\}.
\]

Let $C_{\hv_{j+1},\hy_{j+1}}$ and
$C_{\hy_{j+1},\hy_{j+1}}$ denote, respectively, the cross-covariance
of $(\hv_{j+1},\hy_{j+1})$ and the covariance of $\hy_{j+1}$ under
$\joint_{j+1}$. Assume that
$C_{\hy_{j+1},\hy_{j+1}}$ is invertible and that the residual
covariance
\[
C_{\mathrm{res}}
\coloneqq
C_{\hv_{j+1},\hv_{j+1}}
-
C_{\hv_{j+1},\hy_{j+1}}
C_{\hy_{j+1},\hy_{j+1}}^{-1}
C_{\hy_{j+1},\hv_{j+1}}
\]
is positive definite. Then the composed map
in~\eqref{eq:ComposedMapFiltering} is the population analogue of the
ensemble Kalman filter\index{filter!ensemble Kalman}
update~(\ref{eq:predEnKF}b):
\[
T(v,y,\yd_{j+1})
=
v+
C_{\hv_{j+1},\hy_{j+1}}
C_{\hy_{j+1},\hy_{j+1}}^{-1}
(\yd_{j+1}-y).
\]
\end{theorem}

\begin{proof}
For conciseness, we suppress the time subscripts in what follows.
By Theorem~\ref{thm:ClosedFormAffineTransportMaps}, an optimal affine
transport is
\[
S(y,v;\thetas)
=
C_{\mathrm{res}}^{-1/2}
\left(
v-\mathbb{E}[v]
-
C_{\hv,\hy}C_{\hy,\hy}^{-1}
\bigl(y-\mathbb{E}[y]\bigr)
\right).
\]
Its inverse, evaluated at the observation $\yd_{j+1}$, is
\[
S(\yd_{j+1},\cdot\,;\thetas)^{-1}(z)
=
C_{\mathrm{res}}^{1/2}z
+
\mathbb{E}[v]
+
C_{\hv,\hy}C_{\hy,\hy}^{-1}
\bigl(\yd_{j+1}-\mathbb{E}[y]\bigr).
\]
Composing these maps cancels
$C_{\mathrm{res}}^{1/2}$ with
$C_{\mathrm{res}}^{-1/2}$, as well as the mean terms, and gives
\[
T(v,y,\yd_{j+1})
=
v+
C_{\hv,\hy}C_{\hy,\hy}^{-1}
(\yd_{j+1}-y).
\]
\end{proof}

\subsection{Amortized Smoothing\index{smoothing, amortized}}\label{sec:OPFT}


In this subsection, we generalize the approach in Subsection~\ref{ssec:PLA2} by seeking the transport map\index{transport map} $T$ that explicitly depends on the observation and pushes forward a distribution $\pi_{j}(v_j) = \mathbb{P}(v_j|\Yd_j)$ for the state at time $j$ to the  smoothing distribution $\mathbb{P}(v_j|\Yd_{j},\yd_{j+1})$ given a new observation $\yd_{j+1}$ at time $j+1$.

\begin{remark}
    The distribution for the state given an observation at one time step in the future is often referred to as the lag-1 smoothing distribution. It is also an essential part of the construction of the optimal\index{particle filter!optimal} 
    particle filter; see~\eqref{eq:aint} in Section~\ref{ssec:opf}. That is, we would like to find map parameters $\theta_{j} \in \Theta \subset \R^p$ such that
\begin{equation}
    T_{j}(v_j,\yd_{j+1};\theta_{j}) \sim \mathbb{P}(v_j|\Yd_j,\yd_{j+1}) =:\pop_j, \quad v_j \sim \pi_j.
\end{equation}
\end{remark}

We will describe a similar methodology to Subsection~\ref{sec:ED_datadependence_opt} that identifies the map by minimizing a strictly proper scoring rule\index{score!proper}. The main difference is that we will identify the transport that matches the smoothing distribution $\pop_j$ at time $j$ rather than the filtering distribution $\pi_{j+1}$. We start by defining the joint distribution
\begin{equation}
    \gamma^1_{j}(v_j,y_{j+1}) \coloneqq \mathbb{P}(v_j,y_{j+1}|\Yd_j) = \int \pi_{j}(v_j)\mathbb{P}(\hv_{j+1}|v_j)\like(y_{j+1}|\hv_{j+1}) d\hv_{j+1}. 
\end{equation}
We now introduce the following theorem, which will motivate our choice of objective function.

\begin{theorem}
Consider a strictly proper scoring rule\index{scoring rule!proper}  $\mS(\cdot, \cdot): \cP(\R^{\du}) \times \R^{\du} \to \R.$  Let $\pop_{j}(v_{j}|y_{j+1})=\mathbb{P}(v_{j}|\Yd_{j},y_{j+1})$ be the lag-1 smoothing distribution
 on $\R^{\du}$ given an observation $y_{j+1} \in \R^{\dy}$ at time $j+1$ and let $q(\cdot\,;y_{j+1})$ denote any distribution
 on $\R^{\du}$ parameterized by $y_{j+1}.$ Then,
    \begin{equation}
\mathbb{E}^{(v_{j}, y_{j+1})\sim \gamma^1_{j}} \Bigl[\mS(q(\cdot\,;y_{j+1}), v_{j})\Bigr]\geq \mathbb{E}^{(v_{j}, y_{j+1})\sim \gamma^1_{j}} \Bigl[\mS(\pop_{j}(\cdot|y_{j+1}), v_{j})\Bigr].
\end{equation}
Furthermore, equality is achieved if $q(\cdot\,;y_{j+1}) = \pop_{j}(\cdot|y_{j+1})$ for $\kappa_{j+1}$-almost every $y_{j+1}$, where $\kappa_{j+1}$ denotes the $y_{j+1}$-marginal of $\gamma_j^1$.
\end{theorem}
\begin{proof}
    First, note that $\gamma^1_j(v_j,y_{j+1}) = \kappa_{j+1}(y_{j+1}) \Prob(v_j | \Yd_j, y_{j+1})$. 
    Then,
    \begin{align*}
        \mathbb{E}^{(v_{j}, y_{j+1})\sim \gamma^1_{j}} \Bigl[\mS(q(\cdot\,;y_{j+1}), v_{j})\Bigr] &= \Expect^{y_{j+1}\sim \kappa_{j+1}}\mathbb{E}^{v_{j}\sim \pop_{j}(\cdot|y_{j+1})} \Bigl[\mS(q(\cdot\,;y_{j+1}), v_{j})\Bigr].
    \end{align*}
    The result then follows similarly to the proof of Theorem~\ref{th:scoring_rule_joint}.
\end{proof}

Based on this theorem, to identify the transport $T$ parameterized by $\theta \in \R^p$, we introduce the objective function for a strictly proper scoring rule $\mS$:
\begin{align*}
    \J_{j}(\theta) &= \mathbb{E}^{(v_j,y_{j+1})\sim\gamma^1_{j}} \Bigl[\mS(T_{j}(\cdot,y_{j+1};\theta)_\sharp \pi_{j}, v_{j})\Bigr].
\end{align*}

We generalize Data Assumption~\ref{da:99b}, from Subsection~\ref{sec:ED_datadependence_opt}, and assume access to the following data pairs for the state at time $j$ and the observation at time $j+1$ to implement this objective. This data structure is integral
to learning the lag-$1$ smoothing distribution.

\begin{dataassumption}\index{Data Assumption}\label{da:99_smoothing} We are given pairs of i.i.d.\thinspace samples $\{(v_{j}^{(n)},{\hy}_{j+1}^{(n)})\}_{n=1}^N$ from the joint conditional distribution $\gamma_{j}^1$.
\end{dataassumption}

Given a collection of analysis samples $\{v_j^{(n)}\}_{n=1}^N \sim \pi_{j}$, we can generate the pairs required in Data Assumption~\ref{da:99_smoothing} by sampling synthetic observations using the dynamics and observation model \eqref{eq:sdm}, \eqref{eq:ddm99}. That is,
\begin{subequations} \label{eq:Lag1_sampling}
\begin{align}
    \widehat{v}_{j+1}^{(n)} &= \Psi(v_j^{(n)}) + \xi_j^{(n)}, \quad \xi_j^{(n)} \sim \mathcal{N}(0,\Sigma), \\
    \widehat{y}_{j+1}^{(n)} &= h(\widehat{v}_{j+1}^{(n)}) + \eta_{j+1}^{(n)}, \quad \eta_{j+1}^{(n)} \sim \mathcal{N}(0,\Gamma).
\end{align}
\end{subequations}
We can then define the following empirical\index{empirical} optimization problem for the map parameters:
\begin{subequations}
    \begin{align}
    \J^N_{j}(\theta) &= \frac{1}{N}\sum_{n=1}^N 
    \mS(\pi^{\mathrm{OPF},N}_{j}
(\widehat{y}_{j+1}^{(n)};\theta),
v_j^{(n)}),
    \\
    \thetas_{j} &\in \argmin_{\theta\in\Theta} \J^N_{j}(\theta),
    \end{align}
\end{subequations}
where $\pi^{\mathrm{OPF},N}_{j}$ 
represents an empirical measure for the pushforward samples, which depends on the map parameters and an observation $\widehat{y}_{j+1}^{(n)}$. That is, 
\begin{equation*}
\pi^{\mathrm{OPF}, N}_{j}(\widehat y_{j+1}^{(n)};\theta)
=
\frac{1}{N-1}\sum_{\substack{m=1 \\ m \neq n}}^N
\delta_{
T_{j}(v_{j}^{(m)},
\widehat y_{j+1}^{(n)};\theta)}.
\end{equation*}
The objective may then be implemented via a scoring rule that admits evaluation on empirical measures, as in Example~\ref{ex:online_spsr}.

\begin{remark} The procedure in Subsection~\ref{ssec:opflf} for sampling from the lag-1 smoothing distribution  
requires a closed-form expression for the likelihood weights, because it
uses the optimal particle filter from Subsection \ref{ssec:opf} to approximately sample $\mathbb{P}(v_j|\Yd_{j+1})$. In the linear–Gaussian observation setting considered there, the required weights are available in closed form. 
On the other hand, the approach 
in this subsection can be implemented for general nonlinear observation 
operators, as long as we can sample from the  dynamics and 
observation model in~\eqref{eq:Lag1_sampling}.
\end{remark}

\section{Bibliography} \label{sec:Bib_learning_filters2}

In addition to the approaches outlined in Section~\ref{sec:Bib_learning_filters} for learning ensemble filters and smoothers, the concept of learning from different realizations of the observed data to amortize the cost of inference has become popular in various time-series models. Starting from the variational auto-encoding framework~\cite{kingma2013auto} that operationalized amortization in the machine learning community, neural architectures specifically for nonlinear filtering and smoothing via amortized variational inference were considered in~\cite{krishnan2017structured, karl2016deep}. Applications of this approach to hierarchical decision-making and model-based reinforcement learning were explored in~\cite{becker2019switching, haarnoja2018latent}. In addition, the related neural process framework in~\cite{garnelo2018conditional, garnelo2018neural} learns to map sets of time-series observations to variational distributions for probabilistic inference after training on a family of stochastic processes. Amortizing 3DVar was considered in \cite{keller_ai-based_2024}. Amortizing 4DVar was considered in~\cite{filoche2023learning,zinchenko_combined_2024}.

Several probabilistic approaches based on transportation of measure have been recently developed for learning filters. These procedures often generalize classical Kalman approaches and reduce their bias when applied to non-Gaussian problems. In particular, the paper \cite{spantini2022coupling} introduces the approach of composing triangular transports as in Theorem~\ref{thm:FilteringMapComposition} to build prior-to-posterior maps that correspond to the mean-field EnKF~\cite{carrillo2022ensemble} with linear maps and generalize the EnKF with nonlinear maps. 
The concept of mean-field EnKF is overviewed in~\cite{calvello22} and analyzed, in terms of
error in relation to the true filter, in~\cite{carrillo2022ensemble}.

This framework was extended to smoothing in~\cite{ramgraber_ensemble_2023}. Ensemble filtering methods that parameterize the map using optimal transport were considered in~\cite{al2024nonlinear}, while using a strictly proper scoring rule cost function to learn the map was explored in \cite{bach_learning_2026, zeng2025ensemble}. In particular \cite{bach_learning_2026} contains analysis justifying
the invocation of the ergodic hypothesis in Subsection \ref{ssec:general}.

For work on learning controllers, a subject closely related
to learning data assimilation algorithms in the presence
of model error, we refer the reader to~\cite{fiechter1997pac,dean2020sample}.
Lastly, we note that use of reinforcement learning\index{reinforcement learning} within data assimilation is an area likely to
see considerable growth; see \cite{mowlavi2024reinforcement, hammoud_data_2024}. 

We conclude with some remarks on implementation issues that may arise when solving the optimization problems in this chapter. In order to evaluate the variational inference objective $\J(\theta)$ in Section~\ref{sec:VA}, the expectations can be replaced by empirical\index{empirical} means. Additionally, the KL divergence must be computed. In some cases, such as with Gaussian densities, a closed form of the divergence is available: see Example \ref{ex:klg}, and
identity \eqref{eq:kl_gaussians} in particular. Indeed Gaussian approximation
is, in general, popular both because of the interpretability of the mean
and covariance, and because of the closed form for the divergence. When only samples from $q_\theta^{\Yd}$ are available (such as in an ensemble smoother), the divergence may be approximated by first applying density estimation to these samples. Finally, with all of the considered learning problems, the derivatives of $\J(\theta)$ will generally be hard to obtain analytically. In this case, auto-differentiation\index{auto-differentiation} can be applied, and then gradient-based methods used to minimize it. An example of using auto-differentiation in a similar context is provided in \cite{levine_framework_2022}.

\part{Fundamentals}
\chapter{\Large{\sffamily{Metrics, Divergences, and Scoring Rules}}}
\label{ch:distances}  
In this chapter we introduce various metrics\index{metric} and divergences\index{divergence} to quantify closeness between probability measures\index{probability!measure} in $\cP(\Ru)$, the set of probability measures on $\Ru$. 
We also introduce scoring rules,\index{scoring rule} which quantify
closeness of a probability measure\index{probability!measure} in $\cP(\Ru)$ to a point in the underlying sample space $\Ru$,
and deterministic\index{scoring rule!deterministic} scoring rules, which quantify the discrepancy between two points in $\Ru$. 
To be consistent with the rest of the book the presentation focuses mainly on probability density functions\index{probability!density function}; but the definitions can all be extended to general probability measures. Indeed we will employ Dirac\index{Dirac measure} measures in several instances. We start with metrics\index{metric} in Section \ref{sec:metric}; we continue with 
divergences\index{divergence} in Section \ref{sec:div}; and finally we consider scoring rules\index{scoring rule} in Section \ref{sec:psr}. Throughout the chapter we will make connections between metrics, divergences and scoring rules; these connections
are summarized in Figure~\ref{fig:nesting_distances}.

Our motivation for studying this topic is twofold. 
First, some metrics and divergences provide a language in which to
state theoretical results about inverse problems and data assimilation. Second,
other metrics and divergences are used to define loss\index{loss!function} functions for probabilistic machine learning models based on data, and to design and assess the convergence of machine learning algorithms\index{algorithm!machine learning}. As an example of the first motivation, the Hellinger distance\index{distance!Hellinger} is useful for stating stability results for measures; see, for instance, Theorems \ref{t:wpz} and \ref{thm1L}. As an example of the second, the energy distance\index{distance!energy} is a metric that is used as a loss function for tasks such as learning
the prior to posterior map (Section \ref{sec:BVI}) and probabilistic filters (Section \ref{sec:flo}). 
The Kullback--Leibler\index{divergence!Kullback--Leibler} divergence is
useful both for stating theoretical results for inverse problems and data assimilation (see discussion in bibliography Section \ref{sec:14}) and for defining loss functions in machine learning (see Chapter \ref{ch:transport}). Scoring rules have  traditionally been used for evaluating probabilistic forecasts against samples and hold potential to define objectives for machine learning tasks. 

In this chapter, when comparing two probability density functions,\index{probability!density function} we will generically denote them by $\postr$ and $\postv$. When considering scalar-valued random variables we denote the cumulative distribution functions\index{cumulative distribution function} associated with these two probability density functions by $\Fr$ and $\Fv$. Whenever a metric, divergence, or score involves moments or other expectations, we tacitly restrict attention to probability measures for which the displayed quantities are finite.

\begin{figure}
\centering
\begin{tikzpicture}
    \draw [thick, fill=gray!50] (0,-0.5) arc (-90:270:3cm and 2cm);
    \draw [thick, fill=gray!80] (0,0) arc (-90:270:2cm and 1cm);
    \draw [thick, fill=gray!10, opacity=0.4] (4,-0.5) arc (-90:270:3cm and 2cm);
    \node [yshift=1cm] (0,0) {Metrics};
    \node [yshift=2.5cm] (0,0) {Divergences};
    \node [xshift=5cm, yshift=1.5cm, text width=2.4cm] (0,0) {Expected scoring rules};
\end{tikzpicture}
\caption{Diagram representing relationships between three different ways of quantifying closeness between probability measures.\index{probability!measure} 
Metrics\index{metric} impose more restrictive conditions than divergences\index{divergence}: only some divergences are metrics. Furthermore, a subset of expected scoring rules\index{scoring rule!expected}---those based on strictly proper scoring rules---\index{scoring rule!strictly proper}lead to divergences,
and in some cases to metrics. 
\label{fig:nesting_distances}} 
\end{figure}

\section{Metrics}\index{metric}
\label{sec:metric}

\subsection{Metrics on the Space of Probability Measures\index{probability!measure}}\label{ssec:metrics_measures}

A \emph{metric}\index{metric} between probability measures is a function $\D\colon \cP(\Ru) \times \cP(\Ru) \rightarrow \mathbb{R}$ that satisfies the following four properties for all $\postr,\postv \in \cP(\Ru)$: 
\begin{enumerate} \itemsep0pt
    \item Non-negative: $\D(\postr,\postv) \geq 0.$
    \item Positive: $\D(\postr,\postv) = 0$ if and only if $\postr = \postv.$
    \item Symmetric: $\D(\postr,\postv) = \D(\postv, \postr).$
    \item Triangle inequality\index{triangle inequality}: 
    \footnote{Satisfaction of the triangle inequality\index{triangle inequality} is also referred to as \emph{sub-additivity}.\index{sub-additivity}}
    $\D(\postr,\postv) \leq \D(\postr,\post) + \D(\post,\postv)$ for all $\pi \in \cP(\Ru).$
\end{enumerate}

A metric\index{metric} is also sometimes referred to as a \emph{distance}\index{distance}.

\subsection{Total Variation and Hellinger Metrics}

\begin{definition}\label{def:hellinger}
The \emph{total variation distance}\index{distance!total variation} between two probability density functions $\postr$ and $\postv$ in $\cP(\Ru)$ is defined by
\[\dtv (\postr,\postv)\coloneqq\frac{1}{2}\int_{\R^d}|\postr(u)-\postv(u)| \, du=\frac{1}{2}\|\postr-\postv\|_{L^1}.\]
The \index{distance!Hellinger}\emph{Hellinger distance} between two probability density functions $\postr$ and $\postv$ is defined by
\[\dhell(\postr,\postv)\coloneqq\biggl(\frac{1}{2}\int_{\R^d} \Bigl|\sqrt{\postr(u)}-\sqrt{\postv(u)}\Bigr|^2 \, du\biggr)^{1/2}=\frac{1}{\sqrt{2}}\|\sqrt{\postr}-\sqrt{\postv}\|_{L^2}.\]
\end{definition}

We now establish bounds between the \index{distance!Hellinger}Hellinger and \index{distance!total variation}total variation distance.

      \begin{lemma} \label{l:dh1}
For any probability density functions $\postr$ and $\postv$ in $\cP(\Ru)$,
the following hold.  
\begin{itemize}
\item (i) The Hellinger metric satisfies
\begin{equation}
    \label{eq:useH}
\dhell(\postr,\postv)^2=1-\big\langle \sqrt{\postr},\sqrt{\postv} \big\rangle_{L^2}.
\end{equation}

\item (ii) The total variation and Hellinger metrics are uniformly bounded; indeed
\[0\leq \dtv(\postr,\postv)\leq 1,\quad 0\leq \dhell(\postr,\postv)\leq 1.\]
\item (iii) The total variation and Hellinger metrics  bound one another; indeed
       \[\frac{1}{\sqrt{2}}\dtv(\postr,\postv)\leq \dhell(\postr,\postv)\leq \sqrt{\dtv(\postr,\postv)}.\]
       \end{itemize}
      \end{lemma}
      
      \begin{proof} Part(i) follows from the fact that all densities are non-negative and have norm one in $L^1$.
      For part (ii) we note that the lower bounds follow immediately from the 
definitions, so we only need to prove the upper bounds. For \index{distance!total variation}total variation distance 
      \[\dtv(\postr,\postv)=\frac{1}{2}\int_{\R^d}|\postr(u)-\postv(u)| \, du\leq \frac{1}{2}\int_{\R^d}\postr(u) \, du+\frac{1}{2}\int_{\R^d}\postv(u) \, du=1. \]
      For the \index{distance!Hellinger}Hellinger distance the upper bound follows from \eqref{eq:useH}.

For the lower bound in part (iii) we use the \index{Cauchy--Schwarz inequality}Cauchy--Schwarz inequality 
      \begin{align*}
      \dtv(\postr,\postv)=&\ \frac{1}{2}\int_{\R^d}\Big|\sqrt{\postr(u)}-\sqrt{\postv(u)}\Big|\Big|\sqrt{\postr(u)}+\sqrt{\postv(u)}\Big| \, du\\
      \leq&\ \biggl(\frac{1}{2}\int_{\R^d}\Big|\sqrt{\postr(u)}-\sqrt{\postv(u)}\Big|^2 \, du\biggr)^{1/2}\biggl(\frac{1}{2}\int_{\R^d}\Big|\sqrt{\postr(u)}+\sqrt{\postv(u)}\Big|^2 \, du\biggr)^{1/2}\\
      \leq&\ \dhell(\postr,\postv)\biggl(\frac{1}{2}\int_{\R^d}\big(2\postr(u)+2\postv(u)\big) \, du\biggr)^{1/2}\\
      =&\ \sqrt{2}\dhell(\postr,\postv).
      \end{align*}
      For the upper bound in part (iii) first notice that 
      $$|\sqrt{\postr(u)}-\sqrt{\postv(u)}|\leq |\sqrt{\postr(u)}+\sqrt{\postv(u)}|$$ since $\sqrt{\postr(u)},\sqrt{\postv(u)}\geq0$. Thus we have
      \begin{align*}
      \dhell(\postr,\postv)=&\ \biggl(\frac{1}{2}\int_{\R^d}\Big|\sqrt{\postr(u)}-\sqrt{\postv(u)}\Big|^2 \, du\biggr)^{1/2}\\
      \leq&\ \biggl(\frac{1}{2}\int_{\R^d}\Big|\sqrt{\postr(u)}-\sqrt{\postv(u)}\Big|\Big|\sqrt{\postr(u)}+\sqrt{\postv(u)}\Big| \, du\biggr)^{1/2}\\
      \leq&\ \biggl(\frac{1}{2}\int_{\R^d}\big|\postr(u)-\postv(u)\big| \, du\biggr)^{1/2}\\
      =&\ \sqrt{\dtv(\postr,\postv)}.
      \end{align*}
      \end{proof}

We now show that, if two densities are close in \index{distance!total variation}total variation or in \index{distance!Hellinger}Hellinger distance, then expectations computed with respect to both densities are also close. In addition, the following lemma also provides a useful characterization of the \index{distance!total variation}total variation distance. 

      \begin{lemma} \label{lemmatv} 
      Let $f:\Ru \to \R.$ 
\begin{itemize}

\item (i) Assume furthermore that $f$ satisfies
$|f|_\infty:=\sup_{u\in \Ru}|f(u)|<\infty.$ It holds that
      \[\bigl|\Expect^\postr[f]-\Expect^{\postv}[f]\bigr|\leq 2|f|_\infty \dtv(\postr,\postv).\]
In fact, the following \index{variational!characterization of total variation}variational characterization of the \index{distance!total variation}total variation distance holds:
\begin{equation} \label{eq:TV_VariationalCharacterization}
\dtv(\postr, \postv) = \frac12 \sup_{|f|_\infty \le 1} \bigl|\Expect^\postr[f]-\Expect^{\postv}[f]\bigr|.
\end{equation}

\item (ii)  If, instead, $f$ is such that 
$f_2:=\bigl(\Expect^\rho[|f|^2]+\Expect^{\postv}[|f|^2]\bigr)^{1/2}< \infty$, then
      \[\big|\Expect^\rho[f]-\Expect^{\postv}[f]\big|\leq 2 f_2 \dhell(\postr,\postv).\]
\end{itemize}
\end{lemma}

      \begin{proof}
      For part (i) we start by noting that
     	\begin{align*}
     	\bigl|\Expect^\postr[f]-\Expect^{\postv}[f]\bigr|=&\ \Big|\int_{\R^d} f(u)\bigl(\postr(u)-\postv(u)\bigr) \, du\Big|\\
     	\leq&\ |f|_\infty\cdot\int_{\R^d} |\postr(u)-\postv(u)| \, du\\
     	=&\ 2|f|_\infty\dtv(\postr,\postv).
     	\end{align*}
Thus, for any $f$ with $|f|_\infty =1$, we obtain 
     	 $$\dtv(\postr, \postv) \ge \frac12 \big|\Expect^\postr[f]-\Expect^{\postv}[f]\big|.$$
     We complete the proof of part (i) by exhibiting a choice of $f$ with $|f|_\infty =1$ that achieves equality.	To this end we choose $f(u):= \text{sign} \Bigl( \postr(u)- \postv(u) \Bigr)$, so that $f(u) \Bigl( \postr(u)- \postv(u) \Bigr) = |\postr(u) - \postv(u)|.$ Then $|f|_\infty =1,$ and 
     	\begin{align*}
     	\dtv(\postr, \postv) &= \frac12 \int_{\R^d} |\postr(u) - \postv(u)| \, du \\
     	&= \frac12 \int_{\R^d} f(u) \Bigl( \postr(u) - \postv(u) \Bigr) \, du \\
     	& = \frac12 \bigl|\Expect^\postr[f]-\Expect^{\postv}[f]\bigr|.
     	\end{align*}

      For part (ii) we use the \index{Cauchy--Schwarz inequality}Cauchy--Schwarz inequality 
      to show that
     	\begin{align*}
     	\bigl|\Expect^\postr[f]-\Expect^{\postv}[f]\bigr|=&\ \biggl|\int_{\R^d} f(u)\Bigl(\sqrt{\postr(u)}-\sqrt{\postv(u)}\Bigr)\Bigl(\sqrt{\postr(u)}+\sqrt{\postv(u)} \, \Bigr) \,du\biggr|\\
     	\leq&\ \biggl(\frac{1}{2}\int_{\R^d} \Big|\sqrt{\postr(u)}-\sqrt{\postv(u)}\Big|^2 \, du\biggr)^{1/2} \biggl(2\int_{\R^d}|f(u)|^2\Big|\sqrt{\postr(u)}+\sqrt{\postv(u)}\Big|^2 \,du\biggr)^{1/2}\\
     	\le &\ \dhell(\postr,\postv)\Big(4\int_{\R^d}|f(u)|^2\bigl(\postr(u)+\postv(u)\bigr) \, du\Big)^{1/2}\\
     	= &\ 2f_2 \,\dhell(\postr,\postv).
     	\end{align*}
      \end{proof}

      \begin{remark}
        When working with general probability measures, not necessarily restricted to measures on $\Ru$ with Lebesgue density,
        the characterization \eqref{eq:TV_VariationalCharacterization} should be taken as the \emph{definition} of total variation distance.
        Definition \ref{def:hellinger} then exhibits a \emph{property} of the total variation distance when the measures are
        defined over $\Ru$ and possess Lebesgue densities.
      \end{remark}

      \begin{example} \label{ex:gtv}
          Consider the two Gaussians\index{Gaussian} $\postr = \cN( m_\postr,\Sigma_\postr)$ and $\postv = \cN( m_\postv,\Sigma_\postv).$
          Using these Gaussians we present examples demonstrating the different manner in which total variation distance behaves, 
          depending on properties of means and covariances. 
          
          First consider the case where
          $\Sigma_\postv=0$, so that $\postv$ is a Dirac\index{Dirac measure} measure at $m_\postv$. Then, if $\Sigma_\postr > 0$, it
          follows that $\dtv(\postr, \postv)=1$; we may show this using the characterization \eqref{eq:TV_VariationalCharacterization}, 
          and the choice of test functions
          $$f_\epsilon=1-2\mathbb{1}_{B(m_\postv,\epsilon)}.$$ 
          To see this note that then $\Expect^\postv[f_\epsilon]=-1$. On the other hand $\Expect^\postr[f_\epsilon] \to 1$ as $\epsilon \to 0.$
          Taking the limit $\epsilon \to 0$ delivers the desired supremum. 
          
          Now consider the case where $\Sigma_\postv=\Sigma_\postr=0$.
          Then $\dtv(\postr, \postv)=0$ if $m_\postr=m_\postv$; however if
          $m_\postr \ne m_\postv$ then $\dtv(\postr, \postv)=1$ a fact which can be seen
          by again using characterization \eqref{eq:TV_VariationalCharacterization}
          and now making choice of test function equal to the difference of two indicator functions with disjoint support, and supported on each of  
          $m_\postr$ and $m_\postv.$
      \end{example}

\begin{remark}
\label{rem:etic}
Consider the setting where $\postr$ and $\postv$ are probability density functions. Then the Hellinger \index{distance!Hellinger} distance between $\postr$ and $\postv$ is equal to one if and only if the measures have disjoint supports. This follows from \eqref{eq:useH}: the Hellinger distance is $1$ if and only if $\sqrt{\postr}\sqrt{\postv}=0$ in a Lebesgue a.e. sense, since the square-roots of the densities are non-negative functions; this implies that their supports must be disjoint. A similar result holds for total variation distance.\index{distance!total variation} More generally,
for probability measures that may include atoms, both statements are
most naturally expressed in terms of mutual singularity:
$\dhell(\postr,\postv)=1$ if and only if $\postr$ and $\postv$ are
mutually singular, and the same equivalence holds for
$\dtv(\postr,\postv)=1$. Example \ref{ex:gtv} illustrates this issue.

\end{remark}

\subsection{\index{transport!optimal} Transportation Metrics\index{metric!transportation}}\label{sec:OPT}
We start by defining the notions of \index{coupling}{\em coupling} and \emph{transport map}\index{transport}.

\begin{definition}
  Let  $\postr,$ $\postv \in \cP(\R^{\du}).$ A \index{coupling}{\em coupling} of $\postr$ and $\postv$ is $\pi \in \cP(\R^d \times \R^d)$ with the property that
$$\int_{\R^d} \pi(z,u) \, du=\postr(z), \quad \int_{\R^{d}} \pi(z,u) \, dz=\postv(u);$$
thus the $z$ and $u$ marginals of $\pi$ deliver $\postr$ and $\postv$ respectively.
We denote the set of all such \index{coupling}couplings by $\Pi_{\postr,\postv}.$ 
\end{definition}

\begin{definition} \label{def:transportmap} Let  $\postr,$ $\postv \in \cP(\R^{\du}).$ A \emph{transport map}\index{transport} $g: \R^d \to \R^d$ between $\postr$ and $\postv$ is
a map with the property that $\postv=g_\sharp \postr.$
\end{definition}

We now link these two concepts through two distinct formulations of
\emph{optimal transport}\index{transport!optimal}.

\subsubsection{Kantorovich Formulation}\index{transport!optimal, Kantorovich}
Given a cost function $\co: \R^d \times \R^d \to \R^+$ we
define the Kantorovich formulation of the 
optimal transport\index{transport!optimal, Kantorovich}
problem as follows:
\begin{align}
\label{eq:KOT}
\pist& \in \arginf_{\pi \in \Pi_{\postr,\postv}} 
\int_{\R^{d} \times \R^d} \co(z,u)\pi(z,u) \, dz du.
\end{align}
The reason for the ``optimal'' terminology is clear from the infimization.
The connection of \index{transport!optimal}optimal transport to transport maps on $\R^d$
is made clear in the Monge formulation of optimal 
transport\index{transport!optimal, Monge} which we now introduce.

\subsubsection{Monge Formulation}\index{transport!optimal, Monge}

In this formulation we explicitly link $\postv$ with a pushforward\index{pushforward} of $\postr$ through a transport map\index{transport!map} $g$ on $\R^d$. We do this by identifying the map which solves the following minimization problem:
\begin{align}
\label{eq:MOT}
\gs&\in \arginf_{g: g_\sharp \postr = \postv}
\int_{\R^{d}} \co\bigl(z,g(z)\bigr) \postr(z)\, dz.
\end{align}

In the following $|\cdot|_r: \R^d \to \R^+$ is the norm defined by $|u|_r^r=\sum_{j=1}^d |u_j|^r.$
Under certain smoothness assumptions, the Kantorovich\index{transport!optimal, Kantorovich} formulation
has solution within the Monge class\index{transport!optimal, Monge}; see the bibliography 
Section \ref{sec:12bib} for detailed discussion of this topic. We present a specific setting in
which such a result holds:

\begin{theorem} \label{t:KM}
Assume that $\co\bigl(z,u\bigr)=|u-z|_{r}^r$ for some $r \in (1,\infty)$ and that $\rho \in L^1(\R^d;\R)$ is a probability density function.
Assume furthermore that $\mathbb{E}^{z \sim \postr} |z|_r^r, \mathbb{E}^{u \sim \postv} |u|_r^r <\infty.$ Then the
Monge pushforward optimal transport problem has a unique solution $\gs.$ Furthermore
the optimal Kantorovich \index{coupling}coupling is also unique and is
constructed using the Monge pushforward\index{pushforward}: 
\begin{subequations}
\begin{align}   
\label{eq:CouplingMap}
\pist &=(\mathsf{id},\gs)_\sharp \postr,\\
  \pist(z,u) &= \postr(z) \delta\bigl(u-{\gs(z)}\bigr). 
\end{align}
\end{subequations}
\end{theorem}

\subsubsection{Metrics From Optimal Transport}\label{ssec:metricstransport}

The notion of \index{transport!optimal}optimal transport may be used to define
families of probability metrics\index{metric} by choosing the cost function in
the Kantorovich\index{transport!optimal, Kantorovich} formulation of optimal transport
to be a power of a metric on $\R^d.$

\begin{definition}
  Given a metric $\dm(\cdot,\cdot): \Ru \times \Ru \to \R^+$ and $p \geq 1,$  the \emph{Wasserstein$-p$ distance\index{distance!Wasserstein}} between two probability density functions $\postr$ and $\postv$ is defined by 
  \begin{align}
\label{eq:WP}
W_{p}(\postr,\postv)&\coloneqq \biggl(\inf_{\pi \in \Pi_{\postr,\postv}}
\int_{\R^{d} \times \R^d} \dm(z,u)^p\pi(z,u) \,dz\, du\biggr)^{1/p}.
\end{align}
\end{definition}

Assume that there is a reference point $u_0$ such that $\int_{\R^d} \dm(u,u_0)^p \postr(u) du <\infty$; by
the triangle inequality for $\dm(\cdot,\cdot)$, the existence of one such point $u_0$ implies that the result holds
for any choice of $u_0.$ We say, then, that $\postr$ has finite $p$-th moments. The Wasserstein$-p$ distance $W_p$ defined above
is indeed a metric in the space of probability measures with finite $p$-th moments.\index{metric} 
The metric has two key aspects: first, it derives from an underlying metric on
$\R^d$; and second, it allows for a meaningful \index{distance}distance to
be calculated between measures which are mutually singular. 

\begin{example}
\label{ex:enosiht}
Recall Example \ref{ex:gtv} and again consider the Gaussians
$\postr = \cN( m_\postr,\Sigma_\postr)$ and $\postv = \cN( m_\postv,\Sigma_\postv).$
It may be shown that, if the metric $\dm$ induced by the Euclidean
norm $|\cdot|$ is used, then 
$$W_{2}(\postr,\postv)^2= |m_\postr-m_\postv |^2+{\rm Tr}\Bigl(\Sigma_\postr+\Sigma_\postv-
2\bigl(\Sigma_\postr^{1/2}\Sigma_\postv \Sigma_\postr^{1/2}\bigr)^{1/2}\Bigr).$$
In particular, if $\postv $ is a \index{Dirac measure}Dirac measure at $m_\postv$ then
$$W_{2}(\postr,\postv)^2= | m_\postr-m_\postv |^2+{\rm Tr}\bigl(\Sigma_\postr\bigr).$$
Thus, in the \index{distance!Wasserstein}Wasserstein$-2$ distance, \index{Gaussian}Gaussian $\postr$ is close to
a \index{Dirac measure}Dirac at $m_\postv$ if the mean of the Gaussian\index{Gaussian} $\postr$ is close
to $m_\postv$ and if its covariance is small. In contrast, Example \ref{ex:gtv}
shows that the total variation distance\index{distance!total variation}
between Gaussian\index{Gaussian} $\postr$ and a Dirac\index{Dirac measure} at $m_\postv$ is maximal, and equal to $1$,  unless the two measures coincide ($m_\postr=m_\postv$, $\Sigma_\postr=0$) in which case it is $0.$ Notice also that if $\postr$ and $\postv$ are \index{Dirac measure}Diracs at $m_\postr$ and $m_\postv$, then $W_2(\postr,\postv) = |m_\postr - m_\postv|;$ thus closeness of the mass locations $m_\postr$ and $m_\postv$ in Euclidean space translates into closeness of $\postr$ and $\postv$ in \index{distance!Wasserstein}Wasserstein distance. 
Again, Example \ref{ex:gtv} shows that two Diracs\index{Dirac measure} are maximally far apart in the total variation  distance\index{distance!total variation} unless they coincide.
\end{example}

\subsubsection{One-Dimensional Setting}

Optimal transport metrics have as an advantage that they
directly link a metric on $\R^d$ to the metric on $\mathcal{P}(\R^d).$ However it is an implicit definition, via an optimization.
Notwithstanding the explicit Gaussian\index{Gaussian} formula in Example
\ref{ex:enosiht} and the Monge formulation via a map on $\R^d$,
it can be hard to develop intuition about the metric in general.
In one dimension, however, there are a number of explicit formulae for the optimal transport metric which are insightful and build intuition; we describe them here.

\begin{lemma}\label{lemma:wasserstein1d} 
Let $\postr,\postv \in \cP(\R)$ have invertible 
cumulative distribution functions\index{cumulative distribution function} $F_\postr,F_\postv \colon \R \rightarrow [0,1]$, respectively. Then,  for $p \ge 1$, the Wasserstein-$p$ distance\index{distance!Wasserstein} with metric $\dm$ induced by the Euclidean norm $|\cdot|$ has the form
$$W_p(\postr,\postv)^p = \int_0^1 |F_\postr^{-1}(q) - F_\postv^{-1}(q)|^p \, dq.$$
\end{lemma}

\begin{proof} Recall that the pushforward of a random variable under its own cumulative distribution 
function\index{cumulative distribution function} $F$ is a uniform random variable on $[0,1].$ 
Moreover, applying the (generalized) inverse of $F$ to a uniform random variable on $[0,1]$ produces a random variable with cumulative distribution function $F$. As a consequence,
the map $g(z) = F_\postv^{-1} \circ F_\postr(z)$ pushes forward $\postr$ to $\postv$. Furthermore, this pushforward
leads to the optimal Kantorovich coupling in the one-dimensional setting; this follows from the Monge formulation as detailed in
the bibliography Section \ref{sec:12bib}.
By substituting this map in the objective and performing the change of variables $q = F_\postr(z)$ we have,
using $dq=F_\postr'(z) \, dz = \postr(z) \, dz$,
\begin{align*} W_p(\postr,\postv)^p&=\inf_{g_\sharp \postr=\postv} \int_\R |z-g(z)|^p \postr(z) \, dz \\
    &= \int_\R |z-F_\postv^{-1} \circ F_\postr(z)|^p \postr(z)\, dz\\
    &= \int_0^1 |F_\postr^{-1}(q)-F_\postv^{-1}(q)|^p \, dq.
\end{align*}
\end{proof}

The Wasserstein-$p$ distance\index{distance!Wasserstein} for $p=1$ has some additional computational advantages. It is an integral probability metric (IPM)\index{metric!IPM}, a concept introduced in Subsection \ref{ssec:IPM} below; see Example~\ref{ex:IPM_W1}. 
Furthermore, the closed-form expression in one dimension can be rewritten without requiring the inverse of cumulative distribution functions:

\begin{lemma} \label{lem:wcdf} In the setting of Lemma \ref{lemma:wasserstein1d},
    the Wasserstein-$1$ distance\index{distance!Wasserstein} can be expressed as 
$$W_1(\postr,\postv) = \int_0^1 |F_\postr^{-1}(q) - F_\postv^{-1}(q)| \, dq = \int_\R |F_\postr(z) - F_\postv(z)| \, dz.$$
\end{lemma}

\begin{figure}[!htp]
    \centering    \includegraphics[width=0.9\textwidth]{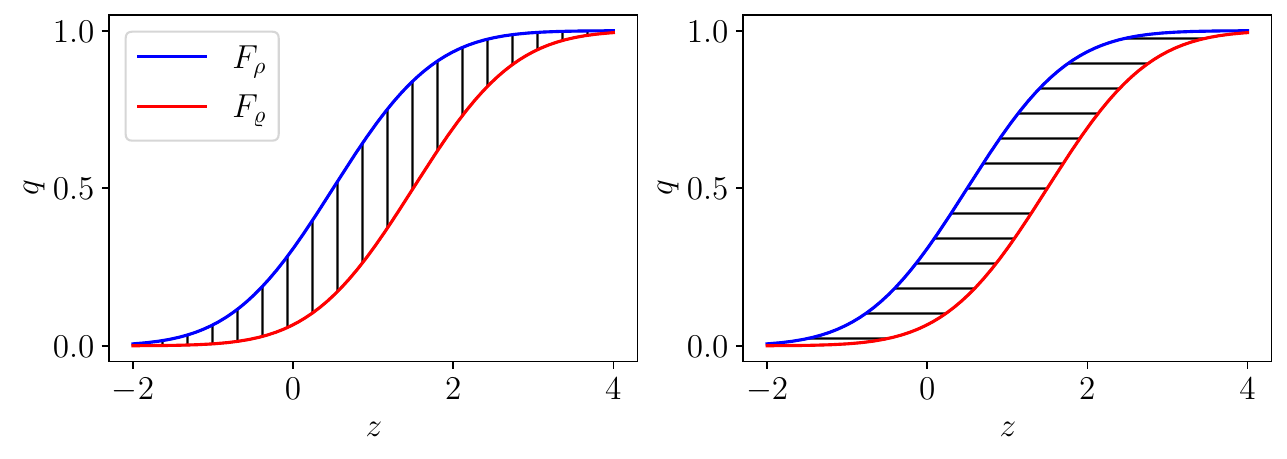}
    \caption{Integration of the difference between cumulative distribution functions (left) and their inverses (right)}
    \label{fig:cdf_integration}
\end{figure}

\begin{proof}
The idea behind the proof of this result is shown in Figure~\ref{fig:cdf_integration}. It illustrates that integrating the difference between the cumulative distribution functions vertically (left) is equivalent to the horizontal integration (right).
\end{proof}

\subsubsection{Sliced Wasserstein Distance\index{distance!Wasserstein!sliced}}

The fact that the Wasserstein distance\index{distance!Wasserstein} in $\R$
is explicit, and defined by quadrature,
may be used as a building block to construct a computable metric in $\Ru.$
The \emph{sliced Wasserstein distance}\index{distance!Wasserstein!sliced} is a readily
computable distance that avoids explicit solution of an optimal transport problem and is defined by 
$${SW}_p^p(\mu,\nu)=\int_{\mathbb{S}^{d-1}} {W}_p^p(P^\theta_\sharp\mu,P^\theta_\sharp\nu) \, d\theta, \quad P^\theta u=\langle u,\theta \rangle.$$
The integral over the unit sphere may be approximated by Monte Carlo\index{Monte Carlo} sampling leading to an implementable methodology based on computation of multiple one-dimensional Wasserstein distances which, in turn, may be approximated by quadrature.
      
\subsection{Integral Probability Metrics}\index{metric!integral probability} \label{ssec:IPM}

We now build on the variational characterization \eqref{eq:TV_VariationalCharacterization} 
of the total variation distance\index{distance!total variation} 
to describe a wide class of probability metrics\index{metric}. 
Rather than taking the supremum over functions with a bound on their maximum value,
we consider taking supremum over other classes of functions. To this end,
let $\mathcal{F}$ be a set of real-valued functions 
$f \colon \Ru \rightarrow \R$ known as discriminators\index{discriminator}.
These are used to define a metric as follows:

\begin{definition} Let $\mathcal{F}$ be a set of discriminator\index{discriminator} functions. An {\em integral probability metric}\index{metric!integral probability} 
(IPM)\index{metric!IPM} $\mathsf{D}_\mathcal{F}:
\cP(\Ru) \times \cP(\Ru) \rightarrow \mathbb{R}$ is defined by setting, for measures\index{probability!measure}
$\postr,\postv \in \cP(\R^{\du})$, 
\begin{equation} \label{eq:IPM}
\mathsf{D}_\mathcal{F}(\postr,\postv) \coloneqq \sup_{f \in \mathcal{F}}\,\big|\Expect^\postr[f]-\Expect^{\postv}[f]\big|.
\end{equation}
\end{definition}

It is natural to ask, then, under what conditions on $\mathcal{F}$ the preceding definition 
yields a metric. Citation to proof of the following may be found in the bibliography Section \ref{sec:12bib}.

\begin{lemma}
\label{lem:sep} Function $\mathsf{D}_\mathcal{F}$
given by \eqref{eq:IPM} defines a metric\index{metric} if and only if
$\mathcal{F}$ separates points\index{separates points} on the space of probability measures: for any two different probability measures\index{probability!measure}
$\postr,\postv$ there is an element in $\mathcal{F}$ with
different expectations under $\postr$ and $\postv.$
\end{lemma}

\begin{example}
The total variation distance\index{distance!total variation}, given in~\eqref{eq:TV_VariationalCharacterization}, corresponds to choosing
$\mathcal{F}$ to be the scaled unit ball 
of functions bounded by $\frac12$ in $L^\infty$. Furthermore, the idea of separating points in the space of probability measures
is central to the calculations concerning total variation distance\index{distance!total variation} in Example \ref{ex:gtv}.
\end{example}

\begin{example} \label{ex:IPM_W1}
    A second important example of an IPM\index{metric!IPM} is the Wasserstein-$1$ distance\index{distance!Wasserstein}. This metric is reproduced as an
   \index{metric!IPM} IPM by choosing $\mathcal{F}$ to be the
space of Lipschitz continuous functions with Lipschitz constant less than or equal to one; see the bibliography Section \ref{sec:12bib}.
\end{example}

\begin{example}
\label{ex:kernel_trick} In the following subsection we introduce the concept of RKHS.\index{reproducing kernel Hilbert space!RKHS} 
A third example of an IPM arises when $\mathcal{F}$ is the unit ball of an RKHS\index{reproducing kernel Hilbert space!RKHS}. 
The resulting metric, known as the \index{metric!maximum mean discrepancy}maximum mean discrepancy, will also be described 
in the following subsection.
\end{example}

\subsection{Maximum Mean Discrepancy and Energy Distance}
\label{ssec:mmded}

We start by defining the concept of kernel, in which we employ the notation $\mathbf{1}$ to denote the vector of all ones:

\begin{definition}
\label{def:kernel}
Let $D \subseteq \R^d.$ A \emph{kernel}\index{kernel} 
is a function $c: D \times D \to \R$. 
\label{d:mk} 
It is called \emph{symmetric}\index{kernel!symmetric} 
if $c(u,u')=c(u',u)$ for all $(u,u') \in D \times D.$ Let $N \in \mathbb{N}$ and $U=\{u^{(i)}\}_{i=1}^N  \subset D$. Define the \emph{Gram matrix}\index{Gram matrix} $C \in \R^{N \times N}$ by its entries
$C_{ij}= c\bigl(u^{(i)}, u^{(j)}\bigr).$ A symmetric kernel is said to be \emph{positive definite}\index{kernel!positive definite} 
if, for all $N \in \mathbb{N}$ and $U \in D^{\otimes N}$, $C \in \bbR^{N \times N}_{\textrm{sym},\ge}.$ 
If matrix $C \in \bbR^{N \times N}_{\textrm{sym},>}$ the kernel is called \emph{strictly} positive definite.\index{kernel!strictly positive definite} If, for all $N \in \mathbb{N}$ and $U=\{u^{(i)}\}_{i=1}^N  \subset D$, 
matrix $C$ has the property that, for all $v$ such that $\langle v, \mathbf{1} \rangle = 0$, $\langle v, C v \rangle \leq 0$, the kernel is called \emph{negative definite}\index{kernel!negative definite}.
\end{definition}

\begin{definition}
\label{def:dmmd}
Let $c: \Ru \times \Ru \to \R$ be a symmetric and positive definite kernel. The 
{\em maximum mean discrepancy}\index{metric!maximum mean discrepancy} (MMD)\index{metric!MMD} with kernel $c$ between  $\postr$ and  $\postv$ in $\cP(\R^{\du})$ is defined by 
$$\dmmd^2(\postr,\postv) \coloneqq \bbE^{(u,u') \sim \postr \otimes \postr} \bigl[c(u,u')\bigr]+\bbE^{(v,v') \sim \postv \otimes \postv} \bigl[ c(v,v')\bigr] - 2\bbE^{(u,v) \sim \postr \otimes \postv} \bigl[c(u,v)\bigr].$$
\end{definition}

We will now show that MMD is an example of an IPM. To that end, we recall first the notion of an RKHS:

\begin{definition}\label{def:RKHS}
    A Hilbert space $\mathcal{H}$ of functions $f:D \subseteq \Ru \to \R$ is called a \emph{reproducing kernel Hilbert space} (RKHS) if there is a function $c: D \times D \to \R$, with $c(u, \cdot) \in \mathcal{H}$ for all $u \in D$, and with the property that, for all $f \in \mathcal{H}$ and $u \in D,$
\begin{equation}\label{eq:reproducingprop}
    f(u) = \langle f, c(u,\cdot) \rangle_{\mathcal{H}}.
\end{equation}
     This is known as the \index{reproducing property}\emph{reproducing property} and 
     the function $c$ is called the \emph{reproducing kernel}.
\end{definition}

An important classical result, the Moore--Aronszajn theorem, ensures that given a symmetric and positive definite kernel $c: D \times D \to \R$, there is a unique Hilbert space of functions on $D$ for which $c$ is a reproducing kernel. Citation to this result can be found in the bibliography Section \ref{sec:12bib}.

\begin{lemma} \label{lem:MMDasIPM}
Let $c: \R^d \times \R^d \to \R$ be a symmetric and positive definite kernel. Let $\mathcal{H}$ be the RKHS with reproducing kernel $c.$
Define $\mathcal{F} = \{f \in \mathcal{H}: \|f\|_{\mathcal{H}} \leq 1\}.$ Then the MMD\index{metric!MMD} with kernel $c$ is an IPM\index{metric!IPM} with set of discriminator functions $\mathcal{F}.$ 
\end{lemma}

\begin{proof}  Let $m_\rho(\cdot) := \Expect^{u\sim \rho} [c(u, \cdot)]$
denote the mean embedding of a distribution $\postr$ with respect to the kernel $c.$
By the reproducing property\index{reproducing property} \eqref{eq:reproducingprop} for functions $f$ in RKHS $\mathcal{H}$ we have
$$\mathbb{E}^\postr[f] = \int_{\R^d} f(u) \postr(u) \, du = \int_{\R^d} \langle f, c(u,\cdot) \rangle_{\mathcal{H}}\, \postr(u) \, du = \left\langle f, \int_{\R^d} c(u,\cdot) \postr(u) \, du \right\rangle_{\mathcal{H}} = \langle f, m_\postr \rangle_{\mathcal{H}}.$$
Using the definition of $\mathcal{F}$, and definition of the Hilbert space norm through duality, $\|g\|_{\mathcal{H}} = \sup_{f \in \mathcal{F}}\, \langle f, g \rangle_{\mathcal{H}}$, we have

\begin{align*}
 \sup_{f \in \mathcal{F}}\, \bigl| \mathbb{E}^\postr[f] - \mathbb{E}^\postv[f]\bigr| 
&= \sup_{f \in \mathcal{F}}\, \bigl| \langle f, m_\postr - m_\postv \rangle_{\mathcal{H}} \bigr| \\
&= \sup_{f \in \mathcal{F}}\,  \langle f, m_\postr - m_\postv \rangle_{\mathcal{H}} \\
&= \|m_\postr - m_\postv\|_{\mathcal{H}} \\
&= \Bigl( \langle m_\postr, m_\postr \rangle_{\mathcal{H}} - 2 \langle m_\postr, m_\postv \rangle_{\mathcal{H}} + \langle m_\postv, m_\postv \rangle_{\mathcal{H}} \Bigr)^{1/2}.
\end{align*}
The right-hand side is identical to $\dmmd(\postr,\postv).$ To see this, we can apply for each term the reproducing property\index{reproducing property} to compute the inner product of the mean embedding. 
In particular we have
\begin{align}
    \langle m_\postr, m_\postv \rangle_\mathcal{H} &= \Bigl\langle \mathbb{E}^{u \sim \postr}[c(u,\cdot)], \mathbb{E}^{u' \sim \postv}[c(u',\cdot)] \Bigr\rangle_{\mathcal{H}} \nonumber \\
    &= \mathbb{E}^{(u,u') \sim \postr \otimes \postv} \Bigl[ \bigl\langle c(u,\cdot), c(u',\cdot) \bigr\rangle_{\mathcal{H}} \Bigr] \nonumber \\ 
    &= \mathbb{E}^{(u,u') \sim \postr \otimes \postv} \bigl[ c(u,u')\bigr].\label{eq:kernel_trick}
\end{align}
and, similarly,
$$\langle m_\postr, m_\postr \rangle_\mathcal{H}= \mathbb{E}^{(u,u') \sim \postr \otimes \postr} \bigl[ c(u,u')\bigr].$$
\end{proof}

\begin{remark} The reproducing property\index{reproducing property} in~\eqref{eq:kernel_trick} permits the computation of inner products with respect to the kernel (or an associated infinite-dimensional feature map) using evaluations of the kernel alone. In many machine learning tasks, this property is referred to as the \index{kernel!trick}\emph{kernel trick.}
\end{remark}

\begin{definition}
The kernel\index{kernel!characteristic} of an RKHS $\mathcal{H}$ defined by $c(\cdot,\cdot)$  \index{reproducing kernel Hilbert space!RKHS}
is \emph{characteristic} if the property 
$\mathbb{E}^\postr [f] = \mathbb{E}^\postv [f]$  for all
$f \in \mathcal{H}$  implies that $\postr = \postv$.
\end{definition}

\begin{remark}
\label{rem:aatc}
We have shown that MMD is an IPM with   $\mathcal{F} = \{f \in \mathcal{H}: \|f\|_{\mathcal{H}} \leq 1\}$ given by the unit ball of an RKHS  $\mathcal{H}$ defined by kernel\index{kernel} $c$.
The IPM\index{metric!IPM} defined by MMD\index{metric!MMD} is a metric\index{metric} if kernel $c$ is 
characteristic\index{kernel!characteristic}.
\end{remark}

\begin{example} \label{ex:aatc}
Gaussian and Laplace kernels\index{kernel!Gaussian}\index{kernel!Laplace} 
are examples of characteristic kernels\index{kernel!characteristic}. 
\end{example}

We defined MMD for positive definite kernels. The energy distance,\index{distance!energy}
which we define next, is obtained by formally setting
$c(u,u')=-|u-u'|$ in the expression defining MMD. Although $c$ is
not positive definite, the resulting expression is non-negative
because the Euclidean distance kernel $|u-u'|$ is negative definite.

\begin{definition} \label{def:energy}
The {\em energy distance}\index{distance!energy} $\den$ between  $\postr$ and $\postv \in \cP(\R^{\du})$ is defined by
\begin{equation}
\den^2(\postr,\postv) \coloneqq 2\bbE^{(u,v) \sim \postr \otimes \postv} |u-v|-\bbE^{(u,u') \sim \postr \otimes \postr} |u-u'|
-\bbE^{(v,v') \sim \postv \otimes \postv} |v-v'|. 
\end{equation}
\end{definition}

In Subsection \ref{ssec:ES} we
discuss energy scores.\index{score!energy}
There we prove the following lemma, based on
an equivalent formulation of energy distance in terms of
characteristic functions.\index{characteristic function}
See, in particular, the proof of Lemma \ref{lemma:den}.

\begin{lemma} \label{lemma:den}
The energy distance $\den$ is a metric on $\cP(\Ru).$
\end{lemma}

\begin{remark}
\label{rem:mmd&e}
A key property of the MMD and the energy distance is that  they are amenable to 
ensemble\index{ensemble} approximation; they can be implemented using only samples
from $\postr$ or $\postv$. Given independent samples $\{u^{(i)}\}_{i=1}^N \sim \postr$ and, independently, given independent samples $\{v^{(i)}\}_{i=1}^M \sim \postv,$ we define the empirical measures
\begin{align}
\label{eq:empz}
\postr^N(u)=\frac{1}{N} \sum_{n=1}^N \delta_{\un}(u), \quad \postv^M(u)=\frac{1}{M} \sum_{m=1}^M \delta_{v^{(m)}}(u).
\end{align}
We can estimate the MMD distance between $\postr$ and $\postv$ by
\begin{subequations}\label{eq:MMD_ens}
\begin{align}
&{\dmmd^2}(\postr,\postv) \approx {\dmmd^2}(\postr^N,\postv^M),\\
   & {\dmmd^2}(\postr^N,\postv^M)  =  \frac{1}{N^2} \sum_{i,j} c(u^{(i)},u^{(j)}) + \frac{1}{M^2} \sum_{i,j} c(v^{(i)},v^{(j)}) - \frac{2}{NM} \sum_{i,j} c(u^{(i)},v^{(j)}).
\end{align}
\end{subequations}
Choosing the kernel $c(u,u')=-|u-u'|$ yields the following
estimate for the energy distance between $\postr$ and $\postv$:
\begin{equation}\label{eq:energy_dist_ens}
   {\den^2}(\postr^N,\postv^M) = \frac{2}{NM} \sum_{i,j} |u^{(i)} - v^{(j)}| - \frac{1}{N^2} \sum_{i,j} |u^{(i)} - u^{(j)}| - \frac{1}{M^2} \sum_{i,j} |v^{(i)} - v^{(j)}|.
\end{equation}
\end{remark}

\begin{remark} \label{rem:jic}
It is standard in practice to use a slightly different empirical approximation of MMD. This
arises from dropping the diagonal term in the first two sums appearing in (\ref{eq:MMD_ens}b),
and renormalizing, resulting in the approximation
$${\dmmd^2}(\postr,\postv) \approx   \frac{1}{N(N-1)} \sum_{i \ne j} c(u^{(i)},u^{(j)}) + \frac{1}{M(M-1)} \sum_{i \ne j} c(v^{(i)},v^{(j)}) - \frac{2}{NM} \sum_{i,j} c(u^{(i)},v^{(j)}).$$
We present the approximation \eqref{eq:MMD_ens} including the diagonal term because it is
simple conceptually. Dropping the diagonal term, however, leads to an unbiased estimator.
Formula \eqref{eq:energy_dist_ens}, in particular, can also be modified similarly to obtain an unbiased 
estimator of the squared energy distance. Note, however, that these unbiased estimators can be negative 
at finite sample size.
\end{remark}

\begin{remark}
An important property of both the MMD with isotropic kernel\index{kernel!isotropic} and the energy distance is that they are rotation invariant. That is, the 
metrics\index{metric!rotation invariant} do not change under the transformations 
$u \leftarrow Ru$ and $v \leftarrow R v$, for any unitary matrix $R$. 
\end{remark}

\subsection{Metrics on the Space of Random Probability Measures}
\label{ssec:rpm}
In these notes, when we consider random variables, they primarily take values in finite-dimensional
Euclidean space. However it is sometimes useful to consider random variables which take values in the space of probability measures on finite-dimensional Euclidean space. This leads to
consideration of the space of random probability measures\index{probability!measure!random}. This
space arises naturally when building methodologies from empirical\index{empirical} sampling or from Monte Carlo\index{Monte Carlo} methods.

We consider functions $\pi: \Omega \to \cP(\Ru)$,
for some abstract probability space $(\Omega,\mathcal{B},\mathbb{P});$ expectation under $\mathbb{P}$ 
is denoted by $\mathbb{E}.$ 
We say that $\pi: \Omega \to \cP(\Ru)$ is an element of the space of random probability measures.\index{probability!measure!random}
For any fixed $\omega \in \Omega$ let $\pi=\pi(\omega)$ and 
define $\bbE^\pi \bigl[f\bigr] \coloneqq \int_{\R^d} f(u)\pi(du)$.
Given $f:\Ru\longrightarrow\mathbb{R}$ recall that $|f|_\infty := \sup_{u\in \Ru} |f(u)|.$

\begin{definition}\label{def:random}
Let $\pi, \pi': \Omega \to \cP(\Ru)$. Then   
$$\dr(\pi,\pi') \coloneqq \sup_{|f|_{\infty} \leq 1} \left|\mathbb{E}\left[\left(\bbE^\pi \bigl[f\bigr]  - \bbE^{\pi'} \bigl[f\bigr] \right)^2\right]\right|^{1/2}.$$
\end{definition}

\begin{remark}
    \label{rem:rtv}
Note that $\dr$ reduces to twice the total variation distance\index{total variation distance} \eqref{eq:TV_VariationalCharacterization}
when $\pi,\pi'$ are not random.
\end{remark}

Given a fixed deterministic probability density function $\post,$ we now define a random Monte Carlo approximation, from independent samples, using Diracs:\index{Dirac measure}
\begin{equation}
\label{eq:pmc0}
\pMC = \frac{1}{\Sam}\sum_{\sam=1}^{\Sam}\delta_{u^{(\sam)}},\,\, u^{(\sam)} \sim \post \quad \index{i.i.d.}{\rm  i.i.d.}
\end{equation}
Note that $\pMC$ is a random probability measure, over space $\Omega$ underlying the random samples
$\{u^{(\sam)}\},$ approximating $\pi.$ From this random probability measure we can compute random approximations of
expectation under $\pi:$
\begin{equation}
\label{eq:pmc}
\bbE^{\pMC} \bigl[f\bigr] = \frac{1}{\Sam}\sum_{\sam=1}^{\Sam}f(u^{(\sam)}),\,\, u^{(\sam)} \sim \post \quad \index{i.i.d.}{\rm  i.i.d.}
\end{equation}
The metric $\dr$ introduced in Definition~\ref{def:random} is well-adapted to quantify the error in approximating the population limit\index{population limit} $\post$ by $\pMC.$ Let $\bbE$ denote expectation with respect to the $N$ i.i.d. samples from $\pi$ defining the Monte Carlo approximation of $\post$, a product measure; likewise $\text{Var}$ denotes variance under the same product measure. The following theorem captures the properties of this random approximation:

\newcommand*\diff{\mathop{}\!\mathrm{d}}
      \begin{theorem}
\label{t:MC}
The  Monte Carlo approximation $\pMC$ of $\pi$ satisfies the following two properties:
       \begin{equation*}
        \begin{split}
        \sup_{|f|_\infty\leq1} \left|\Expect{\left[\bbE^{\pMC} \bigl[f\bigr] -\bbE^\pi \bigl[f\bigr] \right]}\right|  &= 0,\\
        \dr(\pMC, \post)^2 &\leq \frac{1}{\Sam}. 
        \end{split}
        \end{equation*}
      \end{theorem}
      
      \begin{proof}
To prove that the estimator is unbiased,  use linearity of the expected value and that $u^{(\sam)} \sim \post$ to obtain the following identity, from which the desired result follows:
\begin{align*}
\Expect \left[\bbE^{\pMC} \bigl[f\bigr] \right]  &= \Expect \Biggl[   \frac{1}{\Sam} \sum_{\sam=1}^{\Sam} f\bigl(  u^{(n)} \bigr) \Biggr] \\
&= \frac{1}{\Sam} N  \bbE^{\post} \bigl[f\bigr] = \bbE^{\post} \bigl[f\bigr] = \Expect \bigl[  \bbE^{\post} \bigl[f\bigr] \bigr]. 
\end{align*}

For the error estimate expressed in terms of $\dr(\cdot,\cdot)$ we note that, since $\bbE^{\pMC} \bigl[f\bigr]$ is unbiased, its variance coincides
with its mean squared error. Using the fact that the $u^{(\sam)} \sim \post$ are independent we deduce that
\begin{align*}
\text{Var} \left[\bbE^{\pMC} \bigl[f\bigr] \right]  &= \text{Var} \Biggl[   \frac{1}{\Sam} \sum_{\sam=1}^{\Sam} f \bigl( u^{(\sam)}\bigr) \Biggr] \\
&= \frac{1}{\Sam^2} \Sam \text{Var}_\post[f] =  \frac{1}{\Sam} \text{Var}_{\post}[f] . 
\end{align*}
Assuming $|f|_\infty \le 1,$ we have
        \begin{equation*}
        \text{Var}_{\post}[f] = \bbE^{\post} \bigl[f^2\bigr] -\bbE^{\post} \bigl[f\bigr]^2 \leq  \bbE^{\post} \bigl[f^2\bigr] \le 1, 
        \end{equation*}
and therefore
        \begin{equation*}
        \sup_{|f|_\infty\leq1} \left|\Expect{\left[\left(\bbE^{\pMC} \bigl[f\bigr]-\bbE^{\post} \bigl[f\bigr]\right)^2\right]}\right| = \sup_{|f|_\infty\leq1} \left| \frac{1}{\Sam}\text{Var}_{\post}[f]\right| \leq \frac{1}{\Sam}.
        \end{equation*}
      \end{proof}

\section{Divergences}\index{divergence}
\label{sec:div}

Recall the conditions for a metric given in Section~\ref{sec:metric}. If we remove the conditions of
symmetry and the triangle inequality then we obtain a 
{\em statistical divergence}:\index{divergence!statistical} 
a function $\D\colon \cP(\Ru) \times \cP(\Ru) \rightarrow \mathbb{R}$ that satisfies the following two properties for all $\postr,\postv \in \cP(\Ru)$: 
\begin{enumerate} \itemsep0pt
    \item Non-negative: $\D(\postr,\postv) \geq 0.$
    \item Positive: $\D(\postr,\postv) = 0$ if and only if $\postr = \postv.$
\end{enumerate}
To emphasize the asymmetry in their arguments, divergences are often written with double bars $\D(\postr\|\postv)$ or a colon $\D(\postr : \postv)$. We adopt
the former convention. The following is a straightforward consequence of the fact that divergence is obtained by removing two of the four defining characteristics of a metric:

\begin{lemma}
    \label{lem:this2} Every metric is a statistical divergence.
\end{lemma}

\subsection{$\mathsf{f}$-Divergences\index{divergence!$\mathsf{f}$}}
\label{ssec:fdiv}

We now define the special class of $\mathsf{f}$-divergences\index{divergence!$\mathsf{f}$}.
An important feature of this subclass of divergences is that they are defined through expectation of 
a ratio of two probability density functions. To simplify the presentation, throughout this 
section we restrict our attention to positive densities.

\begin{definition}\label{def:fdivergence}
Let $\mathsf{f}\colon [0,\infty) \rightarrow (-\infty,\infty]$ be a convex function that satisfies $\mathsf{f}(0) = \lim_{t \rightarrow 0^+} \mathsf{f}(t)$ (this value can be $\infty$) and $\mathsf{f}(1) = 0$. 
The {\em $\mathsf{f}$-divergence}\index{divergence!$\mathsf{f}$} between distributions with positive densities $\postr, \postv \in \cP(\R^{\du})$ is defined by
$$\D_\mathsf{f}(\postr\|\postv) \coloneqq \bbE^\postv\Bigl[\mathsf{f}\Bigl(\frac{\postr}{\postv}\Bigr)\Bigr] = \int_{\R^d} \mathsf{f}\left(\frac{\postr(u)}{\postv(u)}\right)\postv(u)\,du.$$
\end{definition}

Throughout, we restrict attention to choices of $\mathsf{f}$ for which
$\D_{\mathsf{f}}(\postr\|\postv)=0$ if and only if
$\postr=\postv,$ so that $\D_{\mathsf{f}}$ is a statistical
divergence in the sense defined above.

\begin{remark}
If we changed `convex\index{convex!function} function' to `strictly\index{convex!function!strictly} convex function' in Definition \ref{def:fdivergence}, 
then the preceding restriction would be automatic. However total variation would not, then, be an $\mathsf{f}$-divergence.
Hence the reason for why we have formulated things as we have.
\end{remark}

\begin{remark}
\label{rem:1127}
Some $\mathsf{f}$-divergences\index{divergence!$\mathsf{f}$} define metrics. For example, the choice $\mathsf{f}(t) = \frac{1}{2} | t -1|$ delivers $\D_\mathsf{f}(\postr\|\postv) = \dtv(\postr,\postv)$ and the choice $\mathsf{f}(t) = (1 - \sqrt{t})^2$
delivers $\D_\mathsf{f}(\postr\|\postv)=2\dhell(\postr,\postv)^2.$ It can be shown that the only $\mathsf{f}$-divergence\index{divergence!$\mathsf{f}$} that is also an \index{metric!IPM}IPM is the total variation distance. 
\end{remark}

The next definition introduces two important $\mathsf{f}$-divergences\index{divergence!$\mathsf{f}$} that do not define metrics: the 
Kullback--Leibler divergence\index{divergence!Kullback--Leibler} and the $\chi^2$ 
divergence\index{divergence!$\chi^2$}. They correspond to choosing $\mathsf{f}(t) = t\log t$ and $\mathsf{f}(t) = (t-1)^2,$ respectively.
Reversing the order in the Kullback--Leibler divergence is achieved by setting $\mathsf{f}(t) = -\log t.$

\begin{definition} \label{def:KL} The {\em Kullback--Leibler (KL)  divergence}\index{divergence!Kullback--Leibler} between two positive probability density functions $\postr$ and $\postv$ on $\R^d$ is defined by
\begin{equation}
    \dkl(\postr\|\postv) \coloneqq \int_{\R^d} \log \biggl(\frac{\postr(u)}{\postv(u)} \biggr) \postr(u) \, du.
\end{equation}
The \index{divergence!$\chi^2$} \emph{$\chi^2$ divergence} between two positive probability density functions $\postr$ and $\postv$ on $\R^d$ is defined by \label{def:chi2}
\begin{equation}\label{eq:chi2}
\dchi(\postr \| \postv) \coloneqq \int_{\R^d} \left( \frac{\postr(u)}{\postv(u)} -1  \right)^2 \, \postv(u) \, du.
\end{equation}
\end{definition}

\begin{example}
\label{ex:klg}
The KL and $\chi^2$ 
divergences\index{divergence!Kullback--Leibler}\index{divergence!$\chi^2$} 
can be explicitly computed for $d$-dimensional multivariate Gaussians\index{Gaussian}. Indeed, for
$\postr = \mathcal{N}(m_\postr,\Sigma_\postr)$ and $\postv = \mathcal{N}(m_\postv,\Sigma_\postv)$ we have:
\begin{subequations}
\label{eq:kl_gaussians}
\begin{align}
\dkl(\postr\|\postv) &= \frac{1}{2}\left(\text{Tr}(\Sigma_\postv^{-1}\Sigma_\postr) - d + (m_\postr - m_\postv)^\top \Sigma_\postv^{-1}(m_\postr - m_\postv) + \log\left(\frac{\det\Sigma_\postv}{\det\Sigma_\postr}\right) \right), \\
    \dchi(\postr\|\postv)
&=
\frac{\det(\Sigma_\postv)}
{\sqrt{\det(\Sigma_\postr)\det(2\Sigma_\postv-\Sigma_\postr)}}
\exp \left(
(m_\postr-m_\postv)^\top
(2\Sigma_\postv-\Sigma_\postr)^{-1}
(m_\postr-m_\postv)
\right)
-1.
\end{align}
\end{subequations}
We give references for these formulae in the
bibliography Section \ref{sec:12bib}. Note that neither the KL  
divergence\index{divergence!Kullback--Leibler} nor the $\chi^2$ divergence\index{divergence!$\chi^2$} are symmetric in their arguments, illustrating that the converse of Lemma \ref{lem:this2} does not hold in general. Note further that the formula for the $\chi^2$ divergence requires
$2\Sigma_\postv-\Sigma_\postr$ to be positive definite. If this condition does not hold, then $\dchi(\postr\|\postv) = \infty$.
\end{example}

\begin{remark}
\label{rem:yifan}
The KL divergence\index{divergence!Kullback--Leibler} can be minimized over dependence in its first argument without knowledge of the normalizing constant of the second; this fact is useful in variational inference (see Chapter \ref{ch:VI}).
To show this we must extend the notion of $\mathsf{f}$-divergence\index{divergence!$\mathsf{f}$} in Definition \ref{def:fdivergence} to allow the density in the second argument to be unnormalized. 
The KL divergence\index{divergence!Kullback--Leibler} is the only $\mathsf{f}$-divergence with the property that, for any $c>0$,
$\D_\mathsf{f}(q \|c\pi)-\D_\mathsf{f}(q \|\pi)$ is independent of $q.$ (See Bibliography Section \ref{sec:12bib} for citation.)
For any probability density function $\post$ and statistical $\mathsf{f}$-divergence\index{divergence!$\mathsf{f}$} it holds that
$$\pi \in \argmin_{q \in \cP} \D_\mathsf{f}(q \|\pi)$$
and the minimizer is unique. Consequently, for any $c>0$,
$$\pi \in \argmin_{q \in \cP} \dkl(q \|c\pi).$$
That is, the minimizer is independent of scaling factors on the target density, such as a normalization constant. 
\end{remark}

The following result, which we use repeatedly in Chapter \ref{ch:data-dependence}, is known as the chain rule of KL divergence. \index{divergence!Kullback--Leibler!chain rule} Note the convention that the KL divergence between two conditionals
is defined by averaging over the marginal distribution of the variable used to condition.

\begin{lemma}\label{lemma:chainrule}
    Let $\postr(u,v) = \postr(u)\postr(v|u)$ and $\postv(u,v) = \postv(u)\postv(v|u)$ be two joint probability densities. It holds that 
    \begin{equation*}
        \dkl \bigl( \postr(u,v) \| \postv(u,v) \bigr) = \dkl \bigl( \postr(u) \| \postv(u) \bigr) + \dkl \bigl( \postr(v|u) \| \postv(v|u) \bigr),
    \end{equation*}
where 
\begin{equation*}
    \dkl \bigl( \postr(v|u) \| \postv(v|u) \bigr) := \int_{\R^d} \biggl[ \int_{\R^d} \log \biggl( \frac{\postr(v|u)}{\postv(v|u)} \biggr) \postr(v|u) \,  dv \biggr] \postr(u)\,  du. 
\end{equation*}
\end{lemma}
\begin{proof}
By direct calculation,
    \begin{align*}
        \dkl \bigl( \postr(u,v) \| \postv(u,v) \bigr) &= \int_{\R^d} \int_{\R^d} \log \biggl( \frac{\postr(u,v)}{\postv(u,v)} \biggr) \postr(u,v) \, du dv  \\
        &= \int_{\R^d} \int_{\R^d} \biggl( \log \biggl( \frac{\postr(u)}{\postv(u)} \biggr) + \log \biggl( \frac{\postr(v|u)}{\postv(v|u)} \biggr) \biggr) \postr(u) \postr(v|u) \, du dv\\  
        &= \int_{\R^d}  \log \biggl( \frac{\postr(u)}{\postv(u)} \biggr) \postr(u) \, du
        + \int_{\R^d} \biggl[ \int_{\R^d} \log \biggl( \frac{\postr(v|u)}{\postv(v|u)} \biggr) \postr(v|u) \,  dv \biggr] \postr(u) \, du \\
        & = \dkl \bigl( \postr(u) \| \postv(u) \bigr) + \dkl \bigl( \postr(v|u) \| \postv(v|u) \bigr),
    \end{align*}
as desired.     
\end{proof}

\subsection{Relationships Between $\mathsf{f}$-Divergences\index{divergence!$\mathsf{f}$} and Metrics\index{metric}}
\label{ssec:metdiv}
Here we establish bounds between some metrics\index{metric} and 
$\mathsf{f}$-divergences\index{divergence!$\mathsf{f}$}. 
We continue to work under the simplifying assumption that 
$\postr,\postv$ are positive densities. 

\begin{lemma} \label{l:kllemma}
The \index{distance!Hellinger}Hellinger and \index{distance!total variation}total variation distances
are upper bounded by the \index{divergence!Kullback--Leibler}KL divergence as follows:
$$ \dhell(\postr, \postv)^2 \leq \frac{1}{2}\dkl(\postr \| \postv),
\quad \dtv(\postr, \postv)^2 \leq \dkl(\postr \| \postv).$$
\end{lemma}

\begin{proof}
It suffices to prove only the first inequality since it implies the second by Lemma \ref{l:dh1}.
Define function $\phi:\R^+ \mapsto \R$ by
$$\phi(x): = x - 1 - \log{x}.$$
Then
\begin{equation*}
\phi'(x) = 1 - \frac{1}{x}, \quad
\phi''(x) = \frac{1}{x^2}, \quad
\lim_{x\to \infty} \phi(x)  = \lim_{x\downarrow 0} \phi(x) =\infty,
\end{equation*}
and so the function is convex on its domain.
The minimum of $\phi$ is attained at $x = 1$, and $\phi(1) = 0$;
hence $\phi(x) \geq 0$  for all $x \in (0, \infty).$ It follows that, for all $x > 0,$ 
\begin{equation*}
x-1 \geq \log{x}, \quad
\sqrt{x} - 1 \geq \frac{1}{2}\log{x}.
\end{equation*}
Using this last inequality we can bound the \index{distance!Hellinger}Hellinger distance
as follows:
\begin{equation*}
\begin{split}
\dhell(\postr,\postv)^2 &= \frac{1}{2}\int_{\R^d} \left( 1 - \sqrt{\frac{\postv(u)}{\postr(u)}} \right)^2 \postr(u) \, du \\
&=  \frac{1}{2}\int_{\R^d} \left( 1 + \frac{\postv(u)}{\postr(u)} -  2\sqrt{\frac{\postv(u)}{\postr(u)}} \right) \postr(u) \, du \\
&= \int_{\R^d} \left( 1 - \sqrt{\frac{\postv(u)}{\postr(u)}} \right) \postr(u)\,  du \leq -\frac{1}{2}\int_{\R^d} \log\biggl({\frac{\postv(u)}{\postr(u)}}\biggr)\postr(u) \, du = \frac{1}{2}\dkl(\postr \| \postv ).
\end{split}
\end{equation*}
\end{proof}

The following lemma shows that the \index{divergence!$\chi^2$}$\chi^2$ divergence upper bounds the \index{divergence!Kullback--Leibler}KL divergence. By Lemma \ref{l:kllemma} this implies that
it also bounds the \index{distance!total variation}total variation 
and \index{distance!Hellinger}Hellinger distances. 

\begin{lemma}\label{l:chisq}
The \index{divergence!$\chi^2$}$\chi^2$ divergence upper bounds 
the \index{divergence!Kullback--Leibler}KL divergence
as follows:
$$\dkl(\postr \| \postv) \le \log\Bigl(\dchi(\postr\| \postv) + 1 \Bigr), \quad \quad \dkl( \postr \| \postv) \le \dchi(\postr \| \postv).$$
\end{lemma}
\begin{proof}
The second inequality is a direct consequence of the first one, noting that, for $x \ge 0,$ $\log(x + 1 ) \le x.$
To prove the first inequality note that by \index{Jensen inequality}Jensen inequality
\begin{align*}
\dkl(\postr \| \postv)  &= \int_{\R^d} \log \biggl( \frac{\postr(u)}{\postv(u)} \biggr) \postr(u) \, du \\
& \le \log \biggl(  \int_{\R^d} \frac{\postr(u)}{\postv(u)}   \postr(u) \, du \biggr)  \\
& = \log \biggl(  \int_{\R^d} \frac{\postr(u)}{\postv(u)}   \frac{\postr(u)}{\postv(u)} \postv(u) \, du \biggr)  \\
& = \log\Bigl(\dchi(\postr\| \postv) + 1 \Bigr),
\end{align*}
where for the last equality we used that 
\begin{align*}
\dchi(\postr\| \postv) &= \int_{\R^d} \left( \frac{\postr(u)}{\postv(u)} -1  \right)^2 \, \postv(u) \, du \\
& = \int_{\R^d} \left( \frac{\postr(u)}{\postv(u)} \right)^2 \, \postv(u) \, du  - 2  \int_{\R^d} \left( \frac{\postr(u)}{\postv(u)} \right) \, \postv(u) \, du  + \int_{\R^d}  \postv(u) \, du \\
&= \int_{\R^d} \left( \frac{\postr(u)}{\postv(u)} \right)^2 \, \postv(u) \, du  - 1.
\end{align*}
\end{proof}

\subsection{Invariance of $\mathsf{f}$-Divergences\index{divergence!$\mathsf{f}$} Under Invertible Transformations}
\label{ssec:idiv}
This subsection discusses an important property satisfied by all $\mathsf{f}$-divergences\index{divergence!$\mathsf{f}$}: they are
invariant under invertible transformations of the underlying variables. Recall the notation for
pushforward\index{pushforward} from the preface. We start with the following lemma relating to
pushforwards. The proof is a straightforward consequence of change of variables.

\begin{lemma}
\label{lem:cov}
Let $\varrho \in \cP(\R^{\du})$ and
let $\fgf \in C^1(\R^d,\R^d)$ be invertible everywhere on $\R^d$. 
Assume further that the determinant of the Jacobian of the inverse, 
\begin{equation}
\label{eq:invd}
{\rm det} D(\fgf^{-1})(u)=\Bigl({\rm det} D\fgf\bigl(\fgf^{-1}(u)\bigr)\Bigr)^{-1},
\end{equation}
is positive everywhere on $\R^d$. Then, for $u=g(z)$ and function $\phi: \R^d \to \R$,
$$\int_{\R^d} \phi(z)\varrho(z) \, dz=\int_{\R^d} \bigl(\phi \circ \fgf^{-1}\bigr)(u)
\bigl(\varrho\circ \fgf^{-1}\bigr)(u) {\rm det} D(\fgf^{-1})(u) \, du .$$
In particular, taking $\phi(z)=\mathbb{1}_{g^{-1}(A)}(z)$, we obtain
$$\bbP\bigl(z \in g^{-1}(A)\bigr)=\bbP\bigl(u \in A\bigr)=\int_{A}
\bigl(\varrho\circ \fgf^{-1}\bigr)(u) {\rm det} D(\fgf^{-1})(u) \, du .$$
Thus 
\begin{subequations}
\label{eq:cov}
\begin{align}
    \fgf_\sharp \varrho(u)&= \bigl(\varrho \circ \fgf^{-1}\bigr)(u) {\rm det} D(\fgf^{-1})(u),\\
\log \fgf_\sharp \varrho(u)&= \log \bigl(\varrho \circ \fgf^{-1}\bigr)(u) +
 \log {\rm det} D\bigl(\fgf^{-1}\bigr)(u). 
\end{align}
\end{subequations}
\end{lemma}

The following theorem, concerning the invariance of the $\mathsf{f}$-divergence $\D_\mathsf{f}$ under invertible transformations, is a consequence of the preceding lemma.

\begin{theorem}\label{thm:invariance}
Let $T:\R^d\to\R^d$ be a $C^1$-diffeomorphism such that $\det DT(u)>0$
for every $u\in\R^d.$ Then, for
$\postr,\postv\in\cP(\R^{\du}),$
\[
\D_{\mathsf f}(\postr\|\postv)
=
\D_{\mathsf f}(T_\sharp\postr\|T_\sharp\postv).
\]
\end{theorem}

\begin{proof} Let $v = T(u)$ denote the transformed variable. For an invertible transformation, $u = T^{-1}(v)$ and $du = \det \bigl(\nabla T^{-1}(v)\bigr)\, dv$. Performing a change of variables in the divergence we have 
\begin{align*}
\D_\mathsf{f}(\postr\|\postv) &= \int_{\R^d} \mathsf{f}\left(\frac{\postr(u)}{\postv(u)} \right) \postv(u) \, du \\
&= \int_{\R^d} \mathsf{f} \left(\frac{\postr \bigl(T^{-1}(v)\bigr)}{\postv \bigl(T^{-1}(v)\bigr)} \right) \postv \bigl(T^{-1}(v)\bigr) \det \bigl(\nabla T^{-1}(v)\bigr) \, dv \\
&= \int_{\R^d} \mathsf{f} \left(\frac{\postr \bigl(T^{-1}(v)\bigr)\det \bigl(\nabla T^{-1}(v)\bigr)}{\postv \bigl(T^{-1}(v)\bigr)\det \bigl(\nabla T^{-1}(v)\bigr)} \right) \postv \bigl(T^{-1}(v)\bigr) \det \bigl(\nabla T^{-1}(v)\bigr) \, dv.
\end{align*}
Using Lemma \ref{lem:cov} we see that $\postr \bigl(T^{-1}(v)\bigr)\det \bigl(\nabla T^{-1}(v)\bigr) = T_\sharp \postr(v)$ denotes the density of the pushforward random variable $v$; thus we have
\begin{align*}
\D_\mathsf{f}(\postr\|\postv) = \int_{\R^d} \mathsf{f} \left(\frac{T_\sharp \postr(v)}{T_\sharp \postv(v)} \right) T_\sharp \postv(v) \, dv = \D_\mathsf{f}(T_\sharp\postr\|T_\sharp \postv).
\end{align*}
\end{proof}

\section{Scoring Rules\index{scoring rule}}
\label{sec:psr}

Probabilistic scoring rules\index{scoring rule!probabilistic|textbf} quantify the accuracy of a given distribution, which we will refer to as the {\em forecast},\index{distribution!forecast}
with respect to a {\em true distribution} known only through
samples; the latter is sometimes termed {\em the verification.}\index{distribution!verification} Let $\postr$ and $\postv$ denote the densities of the forecast and verification distributions, respectively. 

\begin{definition}
A {\em scoring rule} is a function $\mS \colon \cP(\Ru) \times \Ru \rightarrow \R$ that assigns a {\em score} $\mS(\postr,v)$ to a sample $v \sim \postv$ with respect
to $\postr$.
\end{definition}

We follow the convention here that scoring rules are negatively oriented, meaning that a lower score indicates a better forecast. To compare the forecast and verification distributions, we define:
\begin{definition}
Given a scoring rule, the \emph{expected score}
$\overline{\mS} \colon \cP(\Ru) \times \cP(\Ru) \rightarrow \R$ 
is defined by
\begin{equation*}
    \overline{\mS}(\postr,\postv) \coloneqq \mathbb{E}^{v \sim \postv}[\mS(\postr, v)] = \int_{\R^d} \mS(\postr, v) \postv(v) \, dv.
\end{equation*}
\end{definition}
It is natural to ask whether an expected score can be used to define a divergence between $\postr, \postv \in \cP(\R^d).$ With this goal in mind we make
the following definition:

\begin{definition}\label{def:proper_score}
    A scoring rule\index{scoring rule} $\mS$ is called \emph{proper}\index{scoring rule!proper} if, for all $\postr, \postv \in \mathcal{P}(\Ru)$, $\overline{\mS}(\postv,\postv) \leq \overline{\mS}(\postr,\postv)$. A scoring rule is called \emph{strictly proper}\index{scoring rule!strictly proper} when equality holds if and only if $\postr = \postv$.
\end{definition}
Thus, for a proper scoring rule, forecasting with the true distribution results in the lowest expected score. For a strictly proper scoring rule, the lowest expected score can only be attained when forecasting with the true distribution. Using this property, scoring rules can
be used to define  a divergence\index{divergence} if strict propriety is imposed:

\begin{theorem} \label{thm:wdwlt}
Let $\mS$ be a strictly proper scoring rule and define
\begin{equation} \label{eq:distance_score}
    \D_{\mS}(\postr \| \postv) \coloneqq \overline{\mS}(\postr,\postv) - \overline{\mS}(\postv,\postv).
\end{equation}
Then $\D_{\mS}$ defines a divergence\index{divergence!scoring rule}.\index{scoring rule!divergence} Conversely, if $\D_{\mS}$ is a divergence, then $\mS$ is strictly proper.
\end{theorem}
\begin{proof}
For a proper scoring rule, the distance is non-negative: $\D_{\mS} (\postr,\postv) \geq 0$. If $\mS$ is strictly proper then $\D_{ \mS}(\postr,\postv) = 0$ if and only if $\postr = \postv$. The converse is immediate.
\end{proof}

In the following subsections we present five (probabilistic) scoring 
rules\index{scoring rules!probabilistic}: the energy score, the continuous ranked probability score,\index{score!continuous ranked probability} the quantile score\index{score!quantile}, the logarithmic score\index{score!logarithmic}, and the Dawid--Sebastiani score.\index{score!Dawid--Sebastiani} We then discuss the class of kernel scores,\index{score!kernel} the topic of noise in the verification\index{distribution!verification}
and we introduce deterministic scoring rules.\index{scoring rules!deterministic}

\subsection{Energy Score}\index{score!energy}
\label{ssec:ES}
We first consider the energy score, which is closely related to the \index{distance!energy}energy distance studied in Subsection \ref{ssec:mmded}.

\begin{definition} \label{d:energy} For $\beta\in(0, 2]$,
the {\em energy score}\index{score!energy} $\ES_\beta \colon \cP(\Ru) \times \Ru \rightarrow \R$ is defined by
\begin{equation}\label{eq:energy_score}
    \ES_\beta(\postr, v) \coloneqq \mathbb{E}^{u \sim \postr}|u - v|^\beta - \frac{1}{2}\mathbb{E}^{(u,u') \sim \postr \otimes\postr}|u - u'|^\beta.
\end{equation}
The \emph{expected energy score}\index{score!expected energy score}
$\overline{\ES}_\beta \colon \cP(\Ru) \times \cP(\Ru) \rightarrow \R$ is then defined by
\begin{equation}\label{eq:exp_energy_score}
    \overline{\ES}_\beta(\postr,\postv) \coloneqq \mathbb{E}^{v \sim \postv}[\ES_\beta(\postr, v)] = \int_{\R^d} \ES_\beta(\postr, v) \postv(v) \, dv.
\end{equation}
\end{definition}

\begin{lemma} \label{lem:ESFT}
The \index{score!expected energy score}expected energy score \eqref{eq:exp_energy_score} can be written, for $\beta\in(0,2)$, as
\begin{equation}
\label{eq:CTM}
    \overline{\ES}_\beta(\postr,\postv) = \frac{\beta 2^{\beta - 2}\Gamma(\frac{d}{2}+\frac{\beta}{2})}{\pi^{d/2}\Gamma(1 - \frac{\beta}{2})}\int_{\R^d} \frac{|\varphi_\postr(u) - \varphi_\postv(u)|^2}{|u|^{d+\beta}} \, du + \overline{\ES}_\beta(\postv,\postv),
\end{equation}
where $\varphi_\postr$ and $\varphi_\postv$ are the characteristic 
functions\index{characteristic function} of $\postr$ and $\postv$, respectively. Hence, for $\beta\in(0,2)$,
the energy score is strictly proper.
\end{lemma}

\begin{proof}{We give a reference for \eqref{eq:CTM} in the
bibliography; its proof is omitted here for reasons of brevity. The strict propriety follows because
two characteristic functions $\varphi_\postr(u),\varphi_\postv(u)$ are identical if and only if
$\postr=\postv.$}
\end{proof}

Recall the squared energy distance\index{distance!energy} in Definition~\ref{def:energy}, given by
\begin{equation}
\den^2(\postr,\postv) := 2\bbE^{(u,v) \sim \postr \otimes \postv} |u-v|-\bbE^{(u,u') \sim \postr \otimes \postr} |u-u'|
-\bbE^{(v,v') \sim \postv \otimes \postv} |v-v'|.
\end{equation}
The following lemma makes an explicit connection between the energy score, with $\beta=1$, and the energy distance.
\begin{lemma}
\label{lem:CTM}
The energy score\index{score!energy} $\ES_1(\postr, v)$ is related to the energy distance by the identity
$$\frac{1}{2}\den^2(\postr,\postv) = \overline{\ES}_1(\postr,\postv) - \overline{\ES}_1(\postv,\postv).$$
\end{lemma}
\begin{proof}
First note that
\begin{align*}
    \overline{\ES}_\beta(\postr,\postv) &= \mathbb{E}^{(u, v)\sim \postr\otimes\postv}|u - v|^\beta - \frac{1}{2}\mathbb{E}^{(u, u')\sim \postr\otimes\postr}|u - u'|^\beta,\\
    \overline{\ES}_\beta(\postv,\postv) &= \frac{1}{2}\mathbb{E}^{(u, u')\sim \postv\otimes\postv}|u - u'|^\beta.
\end{align*}
Then, for $\beta = 1$,
\begin{align*}
    \overline{\ES}_1(\postr,\postv) - \overline{\ES}_1(\postv,\postv) = \mathbb{E}^{(u, v)\sim \postr\otimes\postv}|u - v| - \frac{1}{2}\mathbb{E}^{(u, u')\sim \postr\otimes\postr}|u - u'| - \frac{1}{2}\mathbb{E}^{(u, u')\sim \postv\otimes\postv}|u - u'|,
\end{align*}
which is indeed half of the squared energy distance.
\end{proof}

We are now in a position to prove Lemma \ref{lemma:den}:

\begin{proof}[Proof of Lemma \ref{lemma:den}]
From Lemma \ref{lem:CTM} it follows that $\frac12\den^2(\postr,\postv)$ is a divergence. As
a consequence $\den(\postr,\postv)$ is also a divergence. It remains to show that
$\den(\postr,\postv)$ is symmetric and satisfies the triangle inequality.
The proof follows from noting that, from Lemma \ref{lem:CTM} and equation \eqref{eq:CTM}
\begin{align*}
\frac12\den^2(\postr,\postv) &= \overline{\ES}_1(\postr,\postv) - \overline{\ES}_1(\postv,\postv)\\
&=\frac{\Gamma(\frac{d}{2}+\frac{1}{2})}{2\pi^{d/2}\Gamma(\frac{1}{2})}\int_{\R^d} \frac{|\varphi_\postr(u) - \varphi_\postv(u)|^2}{|u|^{d+1}} \, du .
\end{align*}
Using the fact that $\den(\postr,\postv)$ is defined as a weighted $L^2$-norm
of the difference between the characteristic functions of the pair $(\postr,\postv)$
leads to the desired metric structure. In particular the symmetry and triangle inequality
follow from the properties of norms.
\end{proof}

Lemma \ref{lem:ESFT} holds for $\beta \in (0,2).$ For $\beta=2$ strict propriety
is lost. 

\begin{lemma} \label{rem:nsp}
The energy score\index{score!energy} $\ES_2(\postr, v)$ is proper\index{score!proper}, but not strictly proper\index{score!strictly proper}.  
\end{lemma}

\begin{proof}
Notice that
\begin{align*}
    \ES_2(\postr, v) &= \mathbb{E}^{u \sim \postr}|u|^2 - 2\mathbb{E}^{u \sim \postr}[u]^\top v + |v|^2 - \frac{1}{2}(\mathbb{E}^{u \sim \postr}|u|^2 - 2\mathbb{E}^{u \sim \postr}[u]^\top \mathbb{E}^{u \sim \postr}[u] + \mathbb{E}^{u \sim \postr} |u|^2) \\
    &= \mathbb{E}^{u \sim \postr}[u]^\top \mathbb{E}^{u \sim \postr}[u] - 2\mathbb{E}^{u \sim \postr}[u]^\top v + |v|^2 \\
    &= \bigl|\mathbb{E}^{u \sim \postr}[u] - v\bigr|^2.
\end{align*}
The minimizer of $\mathsf{J}(b):= \mathbb{E}^{v \sim \postv} |b - v|^2$ is $b^\star = \mathbb{E}^{u \sim \postv}[u],$ which shows that $\ES_2$ is proper. It is not strictly proper, since for any other distribution $\widetilde{\postv}$ with $ \mathbb{E}^{u \sim \postv}[u] = \mathbb{E}^{u \sim \widetilde{\postv}}[u] $ it holds that  $\overline{\ES}_2(\postv,\postv) = \overline{\ES}_2(\widetilde{\postv},\postv). $
\end{proof}

\begin{remark} \label{rem:rmse}
When $\beta=2$ the expected energy score $\bbE^{v \sim \postv}[\ES_2(\postr, v)]$ is the \emph{mean-square error} between $\mathbb{E}^{u \sim \postr}[u]$ and $v$. Its square root, the \emph{root-mean-square error (RMSE)}\index{root-mean-square error (RMSE)} is often used for evaluating probabilistic forecasts, despite lacking strict propriety. Furthermore, in various places in Chapters \ref{ch:LG2} 
and \ref{ch:LG10}, we use variants on the mean-square error to define loss functions from which to learn data assimilation algorithms. 
\end{remark}

\subsection{Continuous Ranked Probability Score\index{score!continuous ranked probability|textbf}}

This score\index{score} is only defined for probability measures on $\R$ and real-valued random variables. Similar to the Wasserstein-$1$ distance\index{distance!Wasserstein} 
in the one-dimensional setting studied in Subsection \ref{ssec:metricstransport}, the continuous ranked probability score (CRPS) \index{score!continuous ranked probability} \index{CRPS|see{score, continuous ranked probability}} is  based on comparing cumulative distribution functions.\index{cumulative distribution function} Recall the notation
$F_\rho$ for the cumulative distribution function associated to probability density function $\rho.$
To introduce the definition, we let
 $\mathbb{1}_{u \ge v}$ denote the indicator function of the set $\{u \in \R: u \ge v \}$; this is also known as a Heaviside step function.\index{Heaviside step function}
We will also employ natural generalizations of this notation in which $\ge$ is replaced by
$>$, $\le$ and $<$.

\begin{definition} \label{def:crps} The {\em continuous ranked probability score}  $\CRPS: \cP(\R) \times \R \to \R$ 
 is defined by
    \begin{equation} \label{eq:CRPS}
    \CRPS(\postr, v) \coloneqq \int_{\R} \bigl(\Fr(u) - \mathbb{1}_{u \ge v}(u)\bigr)^2\,du.
\end{equation}
The \emph{expected continuous ranked probability score}
$\overline{\CRPS} \colon \cP(\R) \times \cP(\R) \rightarrow \R$ is then defined by 
\begin{equation*}
    \overline{\CRPS}(\postr,\postv) \coloneqq \mathbb{E}^{v \sim \postv}[\CRPS(\postr, v)] = \int_{\R}  \CRPS(\postr, v) \postv(v) \, dv.
\end{equation*}
\end{definition}

\begin{remark}
The indicator function $\mathbb{1}_{u \ge v}$ is the cumulative distribution function\index{cumulative distribution function} of a Dirac\index{Dirac measure} mass at $v$.
Thus, the CRPS compares two cumulative distribution functions, one for the 
forecast\index{distribution!forecast} and one for the verification\index{distribution!verification}; 
in particular it is the square of the $L^2$ distance between the two cumulative distribution functions.
\end{remark}

The following lemma shows that, in dimension $\du = 1$, the CRPS\index{score!continuous ranked probability} 
is equivalent to the energy\index{score!energy} score in Definition~\ref{d:energy} with $\beta=1$. Thus, the energy
score with $\beta=1$ provides a natural generalization of CRPS to more than one dimension.
\begin{lemma}\label{lemma:CRPSandE1} 
For $\rho \in \cP(\R)$ and $v \in \R,$ it holds that
\begin{equation}
\label{eq:CRPS_energy}
    \CRPS(\postr, v) = \ES_1(\postr,v) = \mathbb{E}^{u \sim \postr}|u - v| - \frac{1}{2} \mathbb{E}^{(u, u') \sim \postr\otimes \postr}|u - u'|.
\end{equation}
\end{lemma}

\begin{proof} First, let us recall that the absolute value of a difference can be written as
$$|u-v| = \int_v^\infty \mathbb{1}_{z \leq u}(z) \, dz + \int_{-\infty}^v \mathbb{1}_{z > u}(z) \, dz,$$
by considering the two cases $v \leq u$ and $v > u$.
Taking an expectation with respect to $u \sim \postr$ and applying Fubini's theorem we have 
\begin{align}
\mathbb{E}^{u \sim \postr}|u-v| &= \int_v^{\infty} \int_{\R} \mathbb{1}_{z \leq u}(z) \postr(u) \, du dz + \int_{-\infty}^v \int_{\R} \mathbb{1}_{z > u}(z) \postr(u) \, du dz \nonumber \\
&= \int_v^{\infty} \bigl(1 - \Fr(z) \bigr) \, dz + \int_{-\infty}^v \Fr(z) \, dz \nonumber \\
&= \int_{\R}  \bigl(1 - \Fr(z)\bigr) \mathbb{1}_{z \geq v}(z) \, dz + \int_{\R} \Fr(z) \mathbb{1}_{z < v}(z) \, dz. \label{eq:CRPS_calibration}
\end{align}
Following similar steps for the second term in~\eqref{eq:CRPS_energy} and taking an expectation over $v \sim \postr$, we have 
\begin{align}
\frac{1}{2}\mathbb{E}^{(u,v) \sim \postr \otimes \postr}\,|u-v| =\frac12 \bbE^{v \sim \postr}
\Bigl( \mathbb{E}^{u \sim \postr}|u-v|\Bigr)
= \int_{\R} \Fr(z) \bigl(1 - \Fr(z)\bigr) \, dz \label{eq:CRPS_spread}.
\end{align}
Lastly, using $\mathbb{1}_{z < v}(z)=1-\mathbb{1}_{z \geq v}(z)$, and subtracting~\eqref{eq:CRPS_spread} 
from~\eqref{eq:CRPS_calibration}, gives
\begin{align*}
\mathbb{E}^{u \sim \postr}|u-v| - \frac12 \mathbb{E}^{(u,u') \sim \postr \otimes \postr}\,|u-u'| 
& = \int_\R  \Bigl( \mathbb{1}_{z \geq v}(z)  - 2 \Fr(z)\mathbb{1}_{z \geq v}(z)  +  \Fr(z)^2 \Bigr) \, dz \\
& = \int_\R  \Bigl( \mathbb{1}_{z \geq v}(z)^2  - 2 \Fr(z)\mathbb{1}_{z \geq v}(z)  +  \Fr(z)^2 \Bigr) \, dz \\
& = \int_{\R} \bigl(\Fr(z) - \mathbb{1}_{z \ge v}(z)\bigr)^2\,dz = \CRPS(\postr, v).
\end{align*}
\end{proof}

The characterization of the CRPS\index{score!expected continuous ranked probability}  in \eqref{eq:CRPS_energy} shows that it is composed of two parts:
a calibration term that quantifies closeness of the forecast to $v$ and a sharpness term that is related to the spread of the distribution.

\begin{example} Consider a forecast distribution which is deterministic: $\postr=\delta_w$ and
$\Fr(u) = \mathbb{1}_{u \ge w}(u)$. Then either \eqref{eq:CRPS} or \eqref{eq:CRPS_energy}
show that the CRPS reduces to the absolute error between the point $w$ and sample $v$.
\end{example}

\begin{lemma}\label{lemma:CRPS}
The expected CRPS for forecast and verification distributions with probability density functions $\postr$ and $\postv$, respectively, is given by 
\begin{equation}
    \overline{\CRPS}(\postr, \postv) = \int_{\R} \bigl(\Fr(u) - \Fv(u)\bigr)^2 \, du + \int_{\R} \Fv(u) \bigl(1-\Fv(u)\bigr) \, du.
\end{equation}
\end{lemma}

\begin{proof}
    Start with form~\eqref{eq:CRPS} of the CRPS. Then,
    \begin{align*}
        \mathbb{E}^{v \sim \postv} \bigl[ \CRPS(\postr, v)\bigr] &= \int_{\R} \mathbb{E}^{v \sim \postv} \Bigl[ \bigl(\Fr(u) - \mathbb{1}_{u \ge v}(u)\bigr)^2 \Bigr]\,du\\
        &= \int_{\R} \mathbb{E}^{v \sim \postv}\Bigl[\Fr(u)^2 - 2\Fr(u)\mathbb{1}_{u \ge v}(u) + \mathbb{1}_{u \geq v}(u)\Bigr]\,du,
    \end{align*}
    noting that $\mathbb{1}_{u \geq v}(u)^2 = \mathbb{1}_{u \geq v}(u)$. Since 
    $\mathbb{E}^{v \sim \postv}[\mathbb{1}_{u \ge v}(u)] = \Fv(u)$, we obtain the result after rearranging.
\end{proof}

\begin{remark}
A direct consequence of Lemma \ref{lemma:CRPS} is that the divergence \eqref{eq:distance_score} defined by the CRPS\index{score!expected continuous ranked probability}  takes the form
    \begin{align*}
       \mathsf{D_{CRPS}} (\postr \| \postv) = \overline{\CRPS}(\postr, \postv) - \overline{\CRPS}(\postv, \postv) &= \int_{\R} \bigl(\Fr(u) - \Fv(u)\bigr)^2 \, du,
    \end{align*}
 which is known as the squared Cram\'er's distance.\index{distance!Cram\'er} Thus, we have shown that in one dimension CRPS \index{score!continuous ranked probability} agrees with the energy score $\mathsf{ES}_1$ (Lemma \ref{lemma:CRPSandE1}), the divergence induced by CRPS\index{score!expected continuous ranked probability} agrees with the squared Cram\'er's distance (Lemma \ref{lemma:CRPS}), and the squared energy distance is exactly twice the squared Cram\'er's distance (Lemma \ref{lem:CTM}).   
\end{remark}

\begin{example} \label{ex:GCRPS}
For a univariate Gaussian\index{Gaussian} density $\postr = \mathcal{N}(m,\sigma^2)$, the 
CRPS\index{score!expected continuous ranked probability}  with respect to a single observation $v$ has the form 
$$\CRPS(\postr,v) = \sigma \left(2\phi\left(\frac{v - m}{\sigma}\right) + \frac{v - m}{\sigma} \left(2\Phi\left(\frac{v - m}{\sigma} \right) - 1 \right) - \frac{1}{\sqrt{\pi}} \right),$$
where $\phi$ and $\Phi$ denote the probability density function and cumulative distribution function of a standard Gaussian\index{Gaussian} distribution, respectively. We give a reference for this formula in the
bibliography Section \ref{sec:12bib}. The formula illustrates that, for a single observation $v,$ the CRPS
favors forecasts whose mean is close to $v$ and whose variance is
small. Indeed, if degenerate Gaussians are allowed, the minimum value
zero is attained when $m=v$ and $\sigma=0,$ corresponding to the
Dirac measure at $v.$
\end{example}

\subsection{Quantile Score}\index{score!quantile}
Like the CRPS\index{score!continuous ranked probability} and
 Wasserstein-$1$ distance\index{distance!Wasserstein} in the one-dimensional setting, the
 quantile score is based on comparing cumulative distribution functions; and like CRPS it is only defined for probability measures on $\R$ and for real-valued random variables. The quantile score introduces additional flexibility, in comparison to the CRPS, through a parameter $\alpha \in (0,1)$ that allows one to penalize deviations at a specific quantile level. To introduce  the  quantile score\index{score!quantile} we first define the
 quantile function\index{quantile function} and then demonstrate how that
 function can be derived through a minimization problem. Finally, we link the 
 quantile score\index{score!quantile} to the CRPS\index{score!continuous ranked probability}.

\begin{definition}
Given a probability density function $\postv$ corresponding
to a random variable on $\R$ with cumulative distribution function $\Fv$, the corresponding \emph{quantile function}\index{quantile function}  $\qv: (0,1) \to \R$
is the inverse of the cumulative distribution function: 
\footnote{For ease of presentation, we assume throughout invertibility of the cumulative distribution function; the ideas generalize to the non-invertible case by considering the generalized inverse distribution function.} 
$\qv(\alpha):=\Fv^{-1}(\alpha)$.
\end{definition}

\begin{example} \label{ex:quant}
Recall that the cumulative distribution function of the Dirac\index{Dirac measure} mass at $v$ is the Heaviside
step function $\mathbb{1}_{u \ge v}(u).$
Informally, by turning this function on its side, we see that its inverse is the constant function taking value $v$ in $(0,1)$. Thus this constant function $v$ is the quantile function associated with the Dirac\index{Dirac measure} mass at $v$. To see this more formally, note that for any $\alpha \in (0,1),$ we have that $F^{-1}_{\delta_v}(\alpha) = \inf\{ u: F_{\delta_v}(u) \ge \alpha \} = v,$ where here we have used the generalized inverse distribution function.
\end{example}

Note that we have already used the quantile function, in Lemma
\ref{lemma:wasserstein1d}, to define the Wasserstein distance\index{distance!Wasserstein} in one
dimension. The quantile function is the solution to the $\alpha$-parameterized family of equations 
$$\Fv\bigl(\qv(\alpha)\bigr)=\alpha.$$
We now prove a lemma which characterizes the quantile function\index{quantile function} in a different way:
showing that the quantile function (of $\alpha$) is also the minimizer of a scalar-valued function $\La(\cdot)$;
the minimizer depends parametrically on both $\alpha$ and $\varrho$ and, for fixed $\varrho$ delivers the
quantile function (of $\alpha$). To this end let $\alpha \in (0,1).$  Start by defining the $\alpha$-parameterized family of functions $\ha: \R \to \R^+$ by 
\begin{equation} \label{eq:hingeloss}
\ha(u)=\begin{cases}
(1-\alpha)|u|, & \quad u<0,\\
\alpha|u|, & \quad u \ge 0.
\end{cases}
\end{equation}
Function $\ha$ is the {\em quantile loss}\index{loss!quantile} and may also be written
$\ha(u)=-(\mathbb{1}_{u\leq 0}-\alpha)u$ or as
\begin{equation*}
\ha(u)=\begin{cases}
-(1-\alpha)u, & \quad u<0,\\
\alpha u, & \quad u \ge 0.
   \end{cases}
\end{equation*}

Using this function, together with a probability density function\index{probability!density function} $\postv$, we define another $\alpha$-parameterized family of functions $\La: \R \to \R^+$ by
\begin{equation}
\La(\theta)= \bbE^{v \sim \postv} \bigl[ \ha(v-\theta) \bigr].
\end{equation} 
The following lemma uses $\La(\cdot)$ to provide an alternative characterization
of the quantile function\index{quantile function}:
\begin{lemma}
\label{l:q}
For $\alpha \in (0,1),$ it holds that 
$$\qv(\alpha) \in\argmin_{\theta \in \R}\,\La(\theta).$$
\end{lemma}

\begin{proof}
Note that, using \eqref{eq:hingeloss},
$$\La(\theta)=(\alpha-1)\int_{-\infty}^\theta (v-\theta)\postv(v) \, dv +
\alpha\int_\theta^\infty (v-\theta)\postv(v) \, dv.$$
Differentiating with respect to $\theta$ yields
\begin{align*}
    D_\theta\La(\theta)&=(1-\alpha)\int_{-\infty}^\theta \postv(v) \, dv -
\alpha\int_\theta^\infty \postv(v)\, dv\\
&=\int_{-\infty}^\theta \postv(v) \, dv-\alpha\\
&=\Fv(\theta)-\alpha.
\end{align*}
Setting the derivative to zero yields the desired result.
\end{proof}

We may now define, and analyze, the \index{score!quantile}quantile score. 

\begin{definition}
For $\alpha \in (0,1)$,
the {\em quantile score}\index{score!quantile} at level $\alpha$, 
${\QS}_{\alpha} \colon \cP(\R) \times \R \rightarrow \R$,
is defined by
$$\QS_\alpha(\postr, v) = h_\alpha \bigl(v - q_{\rho}(\alpha) \bigr).$$ 
The \index{score!expected quantile}\emph{expected quantile score}  
$\overline{\QS}_{\alpha} \colon \cP(\R) \times \cP(\R) \rightarrow \R$ is then defined by
\begin{equation*}
    \overline{\QS}_{\alpha}(\postr,\postv) \coloneqq \mathbb{E}^{v \sim \postv}[\QS_{\alpha}(\postr, v)] = 
  \mathbb{E}^{v \sim \postv}[h_\alpha \bigl(v - q_{\rho}(\alpha) \bigr)].
\end{equation*}
\end{definition}

\begin{remark} In view of Example \ref{ex:quant} we see that the quantile score\index{score!quantile}
is found by applying the hinge function $h_\alpha$ to the difference between the quantile functions
for a Dirac mass at $v$ and for $\rho,$ evaluated at $\alpha.$ We note also the following two 
alternative expressions for the \index{score!quantile}quantile score at level $\alpha:$
\begin{equation}
\label{eq:q}
    \QS_\alpha(\postr, v) = \left(\mathbb{1}_{v \leq \qr(\alpha)} - \alpha\right)(\qr(\alpha) - v), \quad \QS_\alpha(\postr, v) = h_\alpha \bigl(v - F_\rho^{-1}(\alpha) \bigr).
\end{equation} 

\end{remark}

The following is a direct consequence of Lemma~\ref{l:q}:

\begin{proposition}
For every $\alpha \in (0,1)$, the quantile score $\QS_\alpha$ is a proper, but not strictly proper, scoring rule.\index{scoring rule!proper}
\end{proposition}

\begin{proof}
Let
$\postr^\star \in \argmin_{\postr}\, \overline{\QS}_\alpha(\postr, \postv).$
We show that, for all $\alpha \in (0,1),$ $\postr^\star=\postv$
is a solution of this minimization problem. It follows that $\QS_\alpha$ is
a proper scoring rule.\index{scoring rule!proper}
To see this note that 
\begin{align*}
    \bbE^{v\sim \postv}[\QS_\alpha(\postr, v)]&= \bbE^{v\sim \postv}\Bigl[h_\alpha \bigl(v - \qr(\alpha)\bigr) \Bigr] \\
    &=\La\bigl(\qr(\alpha)\bigr).
\end{align*} 
By Lemma \ref{l:q} the optimal $\rho$ will be one for
    which $q_\rho(\alpha)=\qv(\alpha).$ 
    This can be achieved by setting $\postr=\postv.$ Finally, note that $\QS_\alpha$ is not strictly proper since a single quantile does not identify the full forecast distribution. 
\end{proof}

Our final result on \index{score!quantile}quantile scores shows that the CRPS\index{score!continuous ranked probability} can also be computed as twice the integral of the quantile 
score\index{score!quantile} over all quantiles.
\begin{lemma}\label{lemma:CRPSquantile}
Assume that $\postr$ has finite first moment. Then
\begin{equation*}
    \CRPS(\postr, v) = 2\int_0^1 \QS_\alpha(\postr, v)\,d\alpha,
\end{equation*}
and hence
\begin{equation*}
    \overline{\CRPS}(\postr, \postv) = 2\int_0^1 \overline{\QS}_\alpha(\postr, \postv)\,d\alpha.
\end{equation*}

\end{lemma}

\begin{proof}
For $\alpha \in (0,1)$, recall that $\qr(\alpha)=\Fr^{-1}(\alpha)$  so that the first identity in \eqref{eq:q} gives
$$\QS_\alpha(\postr, v) = \left(\mathbb{1}_{v \leq \Fr^{-1}(\alpha)} - \alpha\right)\left(\Fr^{-1}(\alpha) - v\right).$$
We proceed by applying the change of variables $u = \Fr^{-1}(\alpha)$ to the integrated quantile score. Note that this change of variables implies that $d\alpha = \rho(u) \, du$. Thus
\begin{align*} 
2\int_0^1 \QS_\alpha(\postr, v)\,d\alpha &= 
\int_0^1 2\left(\mathbb{1}_{v\leq \Fr^{-1}(\alpha)} - \alpha\right) \left(\Fr^{-1}(\alpha) - v \right) \, d\alpha  \\
&= 2\int_{\R} \left(\mathbb{1}_{v\leq u} - \Fr(u)\right)(u - v)\rho(u) \, du.
\end{align*}
We recognize that, for $u \ne v$, $D_u\bigl(\mathbb{1}_{v\leq u} - \Fr(u)\bigr)^2 = -2\bigl(\mathbb{1}_{v\leq u} - \Fr(u)\bigr)\rho(u)$. Applying integration by parts gives us 
\begin{align*}
    2\int_0^1 \QS_\alpha(\postr, v)\,d\alpha &= 
    -\int_{\R} D_u\left(\mathbb{1}_{v\leq u} - \Fr(u)\right)^2(u - v) \, du  \\
   & = \bigl(\mathbb{1}_{v\leq u} - \Fr(u)\bigr)^2 (u-v) \Bigl|_{u = -\infty}^{u = \infty} + \int_{\R} \bigl(\mathbb{1}_{v\leq u} - \Fr(u)\bigr)^2 \, du. 
\end{align*}
We assume that the boundary terms in the integration by parts vanish, and establish this fact below.
Under this assumption we have shown that
\begin{align*}
    2\int_0^1 \QS_\alpha(\postr, v)\,d\alpha &= \int_{\R} \bigl(\mathbb{1}_{v\leq u} - \Fr(u)\bigr)^2 \, du =  \CRPS(\postr, v),
\end{align*}
as desired.

Now consider the boundary terms. By Markov's inequality we have that, for $u$ large and positive,
$$1 - F_\postr(u) = \bbP(w \ge u) \le \bbP(|w| \ge |u|) \le \frac{\Expect^{w \sim \postr}|w|}{|u|}.$$ 
Similarly, for $u$ large and negative,
$$F_\postr(u) = \bbP(w \le u) \le \bbP(|w| \ge |u|) \le \frac{\Expect^{w \sim \postr}|w|}{|u|}.$$ 
Hence $\bigl(1 - F_\postr(u) \bigr)^2 (u-v)  \to 0$ as $u \to \infty.$ The limit as $u \to -\infty$ is similar.
\end{proof}

\subsection{Logarithmic Score}\index{score!logarithmic}\index{ignorance|see{score,logarithmic}}
The logarithmic score is based on the intuitive idea that  $\postr \in \cP(\R^d)$ provides an accurate forecast for verification sample $v \sim \postv$ if $\rho(v)$ is large. Noting that the logarithm function is monotonically increasing and that scoring rules are negatively oriented by convention,
the logarithmic score evaluates the negative logarithm of the forecast probability density function $\postr \in \cP(\R^d)$ at a sample $v \sim \postv.$ In the following, we tacitly assume we work with positive densities.

\begin{definition} \label{d:logs}
The \emph{logarithmic score}\index{score!logarithmic} 
$\LS : \cP(\R^d) \times \R^d \to \R$ is defined by
\begin{equation}\label{eq:log_score}
    \LS(\postr,v) \coloneqq -\log\postr(v).
\end{equation}
The \index{score!expected logarithmic}\emph{expected logarithmic score}
$\overline{\LS} \colon \cP(\Ru) \times \cP(\Ru) \rightarrow \R$ is then defined by
\begin{equation*}
    \overline{\LS}(\postr,\postv) \coloneqq \mathbb{E}^{v \sim \postv}[\LS(\postr, v)] =  \int_{\Ru} \LS(\postr, v) \postv(v) \, dv.
\end{equation*}
\end{definition}

The logarithmic score is also sometimes referred to as the \emph{ignorance score} when the logarithm is in base 2. Notice that the logarithmic score penalizes heavily forecasts $\postr$ that place low probability in outcomes that materialize: for small $\postr(v),$ $ \mathsf{LS}(\postr,v) \coloneqq -\log\postr(v)$ is very large. Recall the KL divergence\index{divergence!Kullback--Leibler} given in Definition~\ref{def:KL}.

\begin{lemma}
The \index{score!logarithmic}logarithmic score is a strictly proper scoring rule.
The divergence resulting from the \index{score!expected logarithmic}expected logarithmic score is the KL divergence\index{divergence!Kullback--Leibler}:
\begin{equation}
    \dkl(\postv\|\postr) = \overline{\LS}(\postr, \postv) - \overline{\LS}(\postv, \postv).\label{eq:logscore_kl}
\end{equation}
\end{lemma}

\begin{proof}
We have that
\begin{align*}
\overline{\mathsf{LS}}(\postr, \postv)&=\mathbb{E}^{v\sim \postv} [\mathsf{LS}(\postr, v)]=-\int_{\R^d} \log \postr(v) \postv(v)\, dv,\\
\overline{\mathsf{LS}}(\postv, \postv)&=\mathbb{E}^{v\sim \postv}[\mathsf{LS}(\postv, v)]=-\int_{\R^d} \log \postv(v) \postv(v) \, dv.
\end{align*}
Recalling that
$$\dkl(\postv \|\postr)=\int_{\R^d}\log\biggl(\frac{\postv(v)}{\postr(v)}\biggr)\postv(v) \, dv$$
delivers the desired result.
\end{proof}

\subsection{Dawid--Sebastiani Score}\index{score!Dawid--Sebastiani|textbf}
The Dawid--Sebastiani score controls the first and second moments of the forecast. 
In the following definition, we tacitly assume the covariance $C$ to be positive definite.

\begin{definition} \label{d:dawid}
The \emph{Dawid--Sebastiani score} $\mathsf{DS}: \cP(\R^d) \times \R^d \to \R$ is defined by 
\begin{equation}\label{eq:ds_score}
    \mathsf{DS}(\postr,v) := |v - m|_C^2 + \log \bigl(\det(C)\bigr),
\end{equation}
where $m$ and $C$ are the mean and covariance of $\postr$. The
\emph{expected Dawid--Sebastiani score} $\overline{\mathsf{DS}}: \cP(\R^d) \times \cP(\R^d) \to \R$ is then defined by 
\begin{equation}\label{eq:ds_score_exp}
    \overline{\mathsf{DS}}(\postr,\postv) := \int_{\R^d}  \mathsf{DS}(\postr,v) \postv(v) \, dv.
\end{equation}
\end{definition}

\begin{remark} \label{remark:DSequivalence} The Dawid--Sebastiani score is equivalent, up to a linear transformation, to the logarithmic score in the Gaussian\index{Gaussian} case. It can be used, however, for general distributions with finite mean and covariance.
\end{remark}

The Dawid--Sebastiani score is proper, but not strictly proper.
This is similar to the $\ES_2$ score, which reduces to controlling the first moment and is not strictly proper -- see Lemma \ref{rem:nsp}; the Dawid--Sebastiani score controls first and second moments and hence it is not strictly proper.

\subsection{Kernel Scores}\index{score!kernel|textbf}

Recall Definition~\ref{def:kernel} of a kernel\index{kernel}.
A kernel score is a scoring rule associated to a kernel.

\begin{definition}\label{def:kernel_score}
    For kernel\index{kernel} $c: \Ru \times \Ru \to \R$, the \emph{kernel score}\index{score!kernel} $\KS: \cP(\R^d) \times \R^d \to \R$ is defined by
    \begin{equation}
        \KS(\postr, v) = \Expect^{u\sim \postr}[c(u, v)] - \frac{1}{2}\Expect^{(u, u')\sim \postr\otimes \postr}[c(u, u')].
    \end{equation}
    The \emph{expected kernel score} $\overline{\KS}: \cP(\R^d) \times \cP(\R^d) \to \R$ is then defined by
    \begin{equation}
        \overline{\KS}(\postr, \postv) = \Expect^{(u, v)\sim \postr\otimes \postv}[c(u, v)] - \frac{1}{2}\Expect^{(u, u')\sim \postr\otimes \postr}[c(u, u')].
    \end{equation}
\end{definition}

\begin{example}\label{ex:kernel_energy}
    The kernel score with kernel $c(u, v) = |u - v|^\beta$, for $\beta\in (0, 2)$, is equivalent to the energy score $\ES_\beta$ (Definition~\ref{d:energy}).\index{score!energy}
It follows, by Lemma~\ref{lemma:CRPSandE1}, that the continuous ranked probability score (CRPS; Definition~\ref{def:crps})\index{score!continuous ranked probability} is also a kernel score of this type, with $\beta = 1$ and $d = 1$.
\end{example}

Kernel scores have different propriety\index{scoring rule!proper} properties depending on the choice of kernel.
In this context it is natural to define
\begin{equation}\label{eq:natural}
        \D_{\KS}(\postr, \postv) := \overline{\KS}(\postr,\postv) - \overline{\KS}(\postv,\postv) 
    \end{equation}
and to ask when this is indeed a divergence\index{divergence}. As a first step we state
the following theorem, and cite a reference to its proof in the bibliography.
\begin{theorem}\label{th:ks_proper}
    The kernel score with a negative definite\index{kernel!negative definite} kernel is proper.\index{scoring rule!proper}
\end{theorem}
The following lemma establishes the relationship between kernel scores and the maximum mean discrepancy (MMD; Definition~\ref{def:dmmd})\index{maximum mean discrepancy}, and gives a condition under which a kernel score 
is \emph{strictly proper}.

\begin{lemma}
Let $c:\Ru\times\Ru\to\R$ be a positive definite kernel, and let
$\KS$ denote the kernel score associated with the kernel $-c.$ Then
\begin{equation}
\label{eq:mmd_ks}
\frac{1}{2}\dmmd^2(\postr,\postv)
=
\D_{\KS}(\postr,\postv).
\end{equation}
The kernel score $\KS$ is proper. Furthermore, if $c$ is
characteristic, then $\KS$ is strictly proper.
\end{lemma}

\begin{proof}
    Computing the right-hand side of \eqref{eq:mmd_ks} with kernel $-c$, we obtain
    \begin{align*}
        \D_{\KS}(\postr,\postv) = -\Expect^{(u, v)\sim \postr\otimes \postv}[c(u, v)] + \frac{1}{2}\Expect^{(u, u')\sim \postr\otimes \postr}[c(u, u')] + \frac{1}{2}\Expect^{(v, v')\sim \postv\otimes \postv}[c(v, v')];
    \end{align*}
    this is exactly $\frac{1}{2}\dmmd^2(\postr, \postv)$ with kernel $c$.

    From the definition of positive and negative definiteness for kernels (Definition~\ref{def:kernel}), we can see that if $c$ is positive definite, $-c$ is negative definite (the converse is not necessarily true); thus, by Theorem~\ref{th:ks_proper}, $\KS$ with kernel $-c$ is proper.
If $c$ is characteristic\index{kernel!characteristic}, then $\dmmd$ is a metric (Remark~\ref{rem:aatc}), and hence a divergence (Lemma~\ref{lem:this2}). Using Theorem~\ref{thm:wdwlt}, this implies that $\KS$ is strictly proper.
\end{proof}

As with the energy score and CRPS (both types of kernel scores, see Example~\ref{ex:kernel_energy}), a practical advantage of kernel scores is that they can be readily empiricalized.
Specifically, given independent samples
$\{u^{(i)}\}_{i=1}^N\sim\postr,$ the kernel score may be estimated by
$$
\KS(\postr,v)
\approx
\frac1N\sum_{i=1}^N c(u^{(i)},v)
-
\frac{1}{2N(N-1)}
\sum_{i\ne j}c(u^{(i)},u^{(j)}),
$$
where we have dropped the diagonal terms in the second term and renormalized, similarly to the discussion
in Remark \ref{rem:mmd&e}.

\begin{remark}\label{rem:POF}
In inverse problems\index{inverse problems} and data assimilation,\index{data assimilation} it is often useful to consider probabilistic objective functions
that are amenable to empirical approximation; thus they can be used given finite data samples as in the data assumptions arising throughout this book. Many metrics\index{metric} and divergences\index{divergence} will not be amenable to quantifying distance between two probability measures when both are given only through samples, and are hence a sum of 
Dirac (empirical) measures.\index{Dirac measure}\index{probability!measure!empirical} 
Examples include the total variation and Hellinger distances from Definition \ref{def:hellinger} and the KL divergence\index{divergence!Kullback--Leibler} from Definition \ref{def:KL}. On the other hand, the transportation metrics from Subsection \ref{ssec:metricstransport} naturally quantify the closeness between Dirac masses, and, more generally, between empirical measures; see Example \ref{ex:enosiht}. 
Metrics and divergences defined via a scoring rule are in principle naturally
approximated by empirical measures in at least one of their arguments, since in the construction of \emph{expected
score},\index{scoring rule!expected} which leads to a divergence, empiricalization can be used to replace the expectation. 
The \index{scoring rule} energy score\index{score!energy} itself is \emph{also} defined via an expectation and as a consequence the resulting metric, the energy distance\index{distance!energy}, can be evaluated with both inputs given as samples; the same is true of MMD\index{metric!MMD}. However, not all divergences and metrics defined by scoring rules can be evaluated with both inputs given as samples; for example the logarithmic 
score\index{score!logarithmic}, which leads to the KL divergence, cannot. This is because the underlying score requires pointwise evaluation of the underlying density, rather than only an expectation with respect to it. 
For this reason KL divergence\index{divergence!Kullback--Leibler}
needs to be used carefully when designing loss functions that are to be empiricalized in practice
as the resulting objective functions are often only 
empiricalizable with respect to one of their two inputs. 
\end{remark}

\subsection{Noise in the Verification}\label{sec:scores_noise}\index{scoring rule!noisy verification}

In the presence of noise in the verification,\index{distribution!verification} 
the scoring rule will generally favor more dispersed forecasts than if the noise were not present. To account for this we consider scores calculated using noisy data $\widetilde{v}$ drawn from
$\widetilde{v} | v\sim r(\cdot\,|v)$, given the noise-free verification $v \sim \postv$. We can then modify the Definition \ref{def:proper_score}  of propriety as follows:\index{scoring rule!proper with respect to noise}
\begin{definition}
Given noise distribution $r(\cdot\,|\cdot): \Ru \times \Ru \to \R^+$ and probability distribution $\postv$ define
$$\widetilde{\postv}(\widetilde{v}) = \int_{\R^d} r(\widetilde{v} | v) \postv(v) \, dv.$$ 
A scoring rule $\mS$ is called \emph{proper with respect to the noise distribution $r$} if, for all $\postr$, $\mathbb{E}^{\widetilde{v}\sim \widetilde{\postv}} \bigl[{\mS}(\postv, \widetilde{v}) \bigr] \leq \mathbb{E}^{\widetilde{v}\sim \widetilde{\postv}} \bigl[  \mS(\postr,\widetilde{v}) \bigr]$. It is called \emph{strictly proper with respect to $r$} when equality holds if and only if $\postr = \postv$.
\end{definition}

\begin{proposition}
Given noise distribution $r(\cdot\,|\cdot)$ and probability distribution $\postv$ define
$$\widetilde{\postv}(\widetilde{v}) = \int_{\R^d} r(\widetilde{v} | v) \postv(v) \, dv.$$ 
    If $\mS$ is proper, then $\widetilde{\mS}(\postr, \cdot) := \mS(\widetilde{\postr}, \cdot)$  
    is proper with respect to $r$. 
\end{proposition}

\begin{proof}
    We have
    \begin{align*}
        \mathbb{E}^{\widetilde{v}\sim \widetilde{\postv}} \bigl[ \widetilde{\mS}(\postr,\widetilde{v}) \bigr] &= \mathbb{E}^{\widetilde{v}\sim \widetilde{\postv}} \bigl[\mS(\widetilde{\postr}, \widetilde{v}) \bigr]\\
        &\geq \mathbb{E}^{\widetilde{v}\sim \widetilde{\postv}} \bigl[\mS(\widetilde{\postv},\widetilde{v}) \bigr]\\
        &= \mathbb{E}^{\widetilde{v}\sim \widetilde{\postv}} \bigl[ \widetilde{\mS}(\postv,\widetilde{v}) \bigr],
    \end{align*}
    where the second line follows from the propriety of $\mS$.
\end{proof}

Note that propriety does not guarantee an accurate ranking of forecast distributions if the forecast distributions are evaluated on verifications with different levels of noise. To ensure this property, the scoring rule must be modified for each level of noise such that the expected score is the same as it would be if there were no noise. A scoring rule possessing this property is 
known as \emph{unbiased}.\index{scoring rule!unbiased}

\subsection{Distance-Like Deterministic Scoring Rule}
\label{ssec:dldsc}

The scoring rules introduced so far are probabilistic\index{scoring rule!probabilistic}
in that they compare a probability\index{probability!distribution} distribution\index{distribution!probability} with a point in the space over which
that probability distribution is defined. It is useful to have a deterministic scoring rule\index{scoring rule!deterministic} comparing two points; norms form a natural class of examples.

\begin{definition}\label{def:det_score}
    We call a function $\dd(\cdot, \cdot): \Ru\times\Ru\to\mathbb{R}$ a \emph{distance-like deterministic scoring rule}
    \index{scoring rule!distance-like deterministic} 
    if it satisfies the following three properties for all $u, v \in \Ru:$
    \begin{enumerate}
        \item Non-negative: $\dd(u, v) \ge 0.$ 
        \item Positive: $\dd(u, v) = 0$ if and only if $u = v.$
        \item Symmetric: $\dd(u, v) = \dd(v, u).$ 
    \end{enumerate}
\end{definition}

\begin{example}
Let   $\dd(u,v)=\|u-v\|^p$ for any norm $\|\cdot\|$ on $\Ru$ and any $p \in (0,\infty).$ Then $\dd(\cdot,\cdot)$ is a distance-like scoring rule.\index{scoring rule!distance-like} 
\end{example}

\section{Bibliography}
\label{sec:12bib}

The paper \cite{gibbs2002choosing} constitutes a readable introduction to the study of 
metrics, and other distance-like functions, including divergences,
on the set of probability measures. 
The explicit formula for the \index{distance!Wasserstein}Wasserstein distance between Gaussians in Example \ref{ex:enosiht} can be found in \cite{peyre2017computational}; for the \index{divergence!Kullback--Leibler}KL and $\chi^2$ divergence\index{divergence!$\chi^2$} formulae in Example \ref{ex:klg}, see \cite{sanz-alonso_inverse_2023} and \cite{sanz2020bayesian} respectively.

The subject of optimal transport\index{transport!optimal}, and resulting metrics, is covered
in several texts, including \cite{friesecke2024optimal,peyre2017computational,villani2009optimal}. 
For the Euclidean norm metric $\dm(z,u) = |z-u|$ on $\R^d$, and for $\postr$ being absolutely continuous with respect to the Lebesgue measure (it has a probability density function\index{probability!density function}), there exists an optimal transport coupling in~\eqref{eq:WP} solving the Kantorovich\index{transport!optimal, Kantorovich} formulation of the Wasserstein-$p$ \index{distance!Wasserstein} optimal transport problem. Moreover, this coupling has the form in~\eqref{eq:CouplingMap}, which is induced by an optimal transport map $g^\star$ solving the corresponding 
Monge\index{transport!optimal, Monge} problem \eqref{eq:MOT}; see~\cite[Chapter 2]{figalli2023invitation} for this connection in the static setting under conditions on the measures and the cost function, as well as~\cite{benamou2000computational} for this connection in the dynamic setting, which seeks a flow to define the transport map. Furthermore, Theorem \ref{t:KM} provides an example of a setting where this connection is made and the
desired result is implied by \cite[Theorem 6.2.4]{ambrosio2005gradient}; 
and in \cite[Theorem 6.2.7]{ambrosio2005gradient} the
differentiability of the transport map is discussed. 
Note that, under the conditions above, the optimal transport map is
unique for $p>1,$ but uniqueness does not in general extend to $p=1;$
see Section 2.2 of~\cite{santambrogio2015optimal}.
See also \cite[Section 2.2]{santambrogio2015optimal} for discussion of the optimality of the transport map used in the proof of Lemma \ref{lemma:wasserstein1d}.

The $\chi^2$ divergence\index{divergence!$\chi^2$} arises naturally in the analysis of sampling algorithms\index{algorithm!sampling} that involve weighting samples according to the ratio between two densities. In that context, the $\chi^2$ divergence has a natural interpretation as quantifying the variance of the weights. The KL divergence has a distinguished place among $\mathsf{f}$-divergences due to its appealing analytical and computational properties. The paper \cite{chen2023gradient} contains proof of the assertion made in Remark \ref{rem:yifan}.

The energy distance is overviewed in \cite{szekely_energy_2013}. 
Relationships between energy distance and MMD may be found in \cite{sejdinovic2013equivalence}. An efficient algorithm\index{algorithm!energy distance gradient} to compute gradients of the energy distance is provided in \cite{hertrich_generative_2024}. The behavior of the energy distance in high dimensions is studied in \cite{chakraborty_new_2021}.
Relating the characteristic property (and universal property of positive definite kernels) to mean embedding of measures is the subject of \cite{sriperumbudur2011universality}.
An influential paper in the use of probabilistic scoring rules is 
\cite{gneiting2007strictly}, a paper in which the reader will also find a
proof for Theorem~\ref{th:ks_proper}. The recent review 
article \cite{waghmare2025proper} overviews the current state of the art
and reflects the growing use of these ideas.

An important consideration for sample-based metrics is their statistical consistency with empirical measures and the associated rates of convergence to their population-level limits. The paper~\cite{weed2019sharp} shows that the sample-averaged Wasserstein-$p$ metric, for
$p \geq 1$, scales poorly with respect to dimension: $\mathbb{E}[W_p(\nu,\nu^n)]=\mathcal{O}(n^{-1/d})$. On the other hand, the MMD with appropriate choices of the kernel only depends on the mean embedding of the measures in the RKHS and thus converges at the rate $\mathcal{O}(n^{-1/2})$, independent of the dimension of the data; see~\cite{gretton2012kernel, sriperumbudur2010hilbert} for more details on MMD convergence.

The \index{score!continuous ranked probability}CRPS was first introduced in \cite{brown_admissible_1974} and \cite{matheson_scoring_1976}. The relationship between the quantile score and the CRPS stated in Lemma \ref{lemma:CRPSquantile} can be found in \cite{laio2007verification}; see also \cite{fakoor2023flexible,gneiting_comparing_2011}. The discussion of bias in the ensemble CRPS is given in \cite{fricker_three_2013}. 
Closed-form expressions for the CRPS for various distributions are given in \cite{jordan_evaluating_2019}, and references to some additional known expressions are given in Table 9.7 of \cite{wilks_statistical_2019}. The explicit formula for a univariate Gaussian in Example \ref{ex:GCRPS} was established in \cite{gneiting2005calibrated}.
Although not employed here, we mention the \index{distance!Fréchet inception}Fréchet inception distance from \cite{heusel2017gans} 
which is widely used to evaluate image generation.

RKHSs\index{reproducing kernel Hilbert space}
have become a useful framework to understand and develop theoretical results for machine learning methods. The background and properties of 
RKHSs\index{reproducing kernel Hilbert space!RKHS} 
can be found in~\cite{berlinet2011reproducing,smola1998learning};
see also~\cite{muandet2017kernel} for a broad survey on kernel mean embeddings.
The Moore--Aronszajn theorem first appeared in Aronszajn's article \cite{aronszajn1950theory}, where he attributes it to Moore.
Some applications of RKHSs in machine learning include comparing distributions based on two-sample tests from estimators for the maximum-mean discrepancy~\cite{gretton2012kernel} and measuring (conditional) dependence between random variables~\cite{gretton2005measuring}. The statistical properties of MMD have also made it a popular metric for generative modeling~\cite{genevay2018learning}. The proof of Lemma \ref{lem:sep}, containing a precise characterization of when IPMs define a metric on the space of probability measures, can be found in~\cite{muller1997integral, sriperumbudur2010hilbert}.
The paper \cite{sriperumbudur2012empirical} shows that the \index{distance!total variation}total variation distance is the only \index{divergence!$\mathsf{f}$}$\mathsf{f}$-divergence that is also an \index{metric!IPM}IPM.  

A framework for scoring rules\index{scoring rule} in the presence of 
observation error\index{observation error} in the 
verification\index{verification} is given in \cite{ferro_measuring_2017}.
An extensive discussion of scoring rules, with focus on meteorological applications, is given in  Chapter 9 of \cite{wilks_statistical_2019}. A decomposition of proper scoring rules into interpretable terms is introduced in \cite{brocker_reliability_2009}. The relationship between the CRPS and the spread--error relationship is discussed in \cite{leutbecher_understanding_2021}. 
The identity \eqref{eq:CTM} is proved by 
using Proposition 2 in \cite{szekely_energy_2013}. Another quantity that is widely used in the verification of spatial fields, particularly in weather forecasting, is the anomaly correlation coefficient~\cite{murphy1989skill,wilks_statistical_2019}.

\chapter{\Large{\sffamily{Supervised Learning}}}
\label{ch:SL}
This chapter is concerned with the 
\index{supervised learning}{\em supervised learning} task of approximating, 
from data, a function $\psid: D \to R,$ $D$ the \emph{domain}\index{domain} and $R$ the \emph{range}\index{range}. Supervised  learning\index{supervised learning} refers 
specifically to learning $\psid$ from data given
in the form of input--output pairs of $\psid.$
In this chapter we focus on the \emph{regression}\index{regression}
problem in which the domain $D \subseteq \R^d$ and range $R \subseteq \R$ are subsets 
of $\Ru$ and $\R,$ respectively;  the methods we introduce are readily generalized 
to the case where $R$ is a subset of $\R^m$ for any integer $m$. 

When the range of $\psid$ has finite cardinality the function
approximation problem is referred to as \emph{classification}. Classification 
played a central role in the historical development of supervised learning, and
is still one of its major uses in current applications of machine learning? 
Here, however, we focus  on regression: this is a task that arises naturally in the 
context of learning forward maps for inverse problems and learning dynamical systems 
for data assimilation; in particular learning emulators\index{emulator}.
Our data assumption is as follows:

\begin{dataassumption}\index{Data Assumption}
\label{data:sl}
Let $D \subseteq \R^d$, $R \subseteq \R$ be open sets, let $\psid: D \to R,$
let $\{\xi^{(n)}\}_{n=1}^N$ be \index{i.i.d.}i.i.d. centered random variables taking values in $\R$, and let $\lambda \ge 0.$ 
Data is available in the form 
\begin{align}
\label{eq:data}
\Bigl\{(\un, y^{(n)}) \Bigr\}_{n=1}^N,\quad y^{(n)}=\psid(\un) + \sqrt{\lambda} \xi^{(n)},
\end{align}
where the $\{\un\}_{n=1}^N$ are generated \index{i.i.d.}i.i.d. from (unknown)
probability density function\index{probability!density function} $\Upsilon(u)$, supported on $D$, and independent of $\{\xi^{(n)}\}_{n=1}^N$.
\end{dataassumption}

The parameter $\lambda \ge 0$ allows us to consider noiseless data ($\lambda = 0)$ and noisy data $(\lambda >0).$  Since the goal is to determine $\psid$ we are concerned with function 
 \index{approximation}approximation, and to that end we will study three approaches to parameterize and learn functions: \index{neural network}neural networks (Section \ref{sec:NN}), \index{random features}random features (Section  \ref{sec:RF}), and Gaussian processes (Section \ref{sec:GP}). 
We will concentrate on the noiseless case $\lambda = 0$, but will briefly mention the noisy case $\lambda >0$ in the context of \index{random features}random features and \index{Gaussian!process}Gaussian processes. 
Before considering these methods of function approximation we introduce
some notational conventions that will be useful in this and the following chapter.

\section{Notational Conventions}
\label{ssec:not}

Throughout this section $D \subseteq \R^d$ is an open set and $\Upsilon$ is a
probability density function\index{probability!density function} supported on $D$.
Given $\Upsilon$ we may define a Hilbert space structure:

\begin{definition} \label{def:hm}
We let $\hm$ denote the Hilbert space 
of real-valued functions on $D$ 
with inner-product and induced norm
$$\la \psi, \phi \ra_{\hm}=\int_{D} \psi(u)\phi(u)\Upsilon(u) \,du, \quad |\psi|_{\hm}^2=\la \psi,\psi \ra_{\hm}.$$
\end{definition}

Thus the induced norm on $\hm$ is given more explicitly by
\begin{equation}\label{eq:norm}
|\psi|_{\hm}=\biggl( \int_D \psi(u)^2  \Upsilon(u) \, du\biggr)^{1/2}.
\end{equation}

\begin{definition} \label{def:emp}
Given the data from Data Assumption \ref{data:sl} 
we define the \emph{empirical\index{empirical} measure}\index{probability!measure!empirical} 
\begin{align*}
\mn(u)=\frac{1}{N} \sum_{n=1}^N \delta_{\un}(u).
\end{align*}
\end{definition}

The probability density function\index{probability!density function}
$\Upsilon$ is thus approximated by the empirical\index{empirical} measure $\Upsilon^N$
-- see Theorem \ref{t:MC}.
We may then define an empirical\index{empirical} approximation of $|\psi|_{\hm}$
by replacing $\Upsilon$ by $\mn$ in \eqref{eq:norm} to obtain
\begin{equation}
\label{eq:ERM}    
|\psi|_{\hmn}=\Biggl( \frac{1}{N} \sum_{n=1}^N \psi(\un)^2 \Biggr)^{1/2}.
\end{equation}
This does not define a norm on the vector space of functions from $D$ into $\R;$
however it approximates  the  norm $|\psi|_{\hm}$ on that space.

\section{\index{neural network} Neural Networks}\label{sec:NN}
We define neural networks\index{neural network|textbf} acting as real-valued
maps on finite-dimensional Euclidean space. We use neural networks
to perform regression from data.
For simplicity we work, throughout this section,
under Data Assumption \ref{data:sl}, considering only
the noise-free setting $\lambda=0.$
To define a \index{neural network}neural network we make the following
definition:

\begin{definition}
An {\em activation function}\index{activation function} is a map $\sigma: \R \to \R,$ extended to $\sigma: \R^s \to \R^s$ pointwise: $\sigma(u)_m=\sigma(u_m)$ for $u=(u_1,\ldots, u_m,\ldots, u_s)$ for all $m \in \{1,\ldots, s\}.$
\end{definition}

The activation function in the neural network construction that follows below is the only source of nonlinearity in the neural
network; it is thus essential to the ability of neural networks to effectively approximate nonlinear $\psid.$ 
Several commonly made choices of activation function are now defined:

\begin{example} The \emph{Rectified Linear Unit (ReLU)}\index{ReLU} activation function is $\sigma(u)={\rm max}(u,0).$ The  
\emph{Gaussian Error Linear Unit (GELU)}\index{GELU}
activation function is $\sigma(u)=uF(u)$, where $F$ is the univariate standard Gaussian cumulative distribution function.
The \emph{Scaled Exponential Linear Unit (SELU)}\index{SELU} activation function is
\begin{align}
    \sigma(u)=
    \begin{cases}
        \kappa u, &\quad u>0,\\
        \kappa \alpha 
   \bigl(\exp(u)-1\bigr), &\quad u \le 0
   \end{cases}
\end{align}
for fixed positive constants $\kappa$ and $\alpha$.
The \emph{Exponential Linear Unit (ELU)}\index{ELU} activation function is obtained from
the SELU by setting $\kappa = 1.$ 
\end{example}

A \index{neural network}\emph{neural network} is a 
parametric family of functions
found by composing activation functions with affine functions. 
We introduce parameter space $\Theta \subseteq \R^{d_\theta}$ and decompose element $\theta \in \Theta$
into $\theta=(\ttheta,\beta).$ Then we may define 
$\psi: \R^d \times \Theta \to \R$ via the iteration
\begin{align}
\label{eq:dnn}
\psi^{(0)}(u;\ttheta)&=u,\\
\psi^{(\ell+1)}(u;\ttheta)&=\sigma\bigl(W_{\ell}\psi^{(\ell)}(u; \ttheta)+b_\ell\bigr), \quad
\ell=0,\dots, L-1,\\
\psi(u;\theta)&=\beta^\top \psi^{(L)}(u; \ttheta).
\end{align}
Here, for $\ell \in \{0,\ldots, L\}$, $d_\ell \in \bbN$, where $d_0=d$ and the other integers are design choices. Then,
for $\ell \in \{0,\ldots, L-1\}$, $W_{\ell} \in \R^{d_{\ell+1}\times d_{\ell}}$ (the \emph{weights}\index{weights}) 
and $b_{\ell} \in \R^{d_{\ell+1}},$ (the \emph{biases}\index{biases}), define
$\vartheta;$ and $\vartheta$ and $\beta \in \R^{d_L}$ (also weights\index{weights}) together
collectively define $\theta$ (and implicitly $d_\theta \in \bbN$).
Note that $\psi^{(\ell)}: \R^{d} \times \Theta \to \R^{d_{\ell}}.$ 
The resulting function $\psi: \R^d \times \Theta \to \R$ is known as a 
\emph{(deep\footnote{Deep if $L$ is large enough; often three or more is considered large in this context. Lower values are sometimes referred to as \emph{shallow neural networks}\index{neural network!shallow}.}) neural network}\index{neural network!deep}. 

\begin{remark}
\label{rem:NNG}    
The reader will note that the concept of neural networks is readily
generalized to $\psi: \R^d \times \Theta \to \R^q$ for any integer $q;$ 
we concentrate on $q=1$ for simplicity
of exposition only.
\end{remark}

\subsection{Empirical Risk Minimization}

By use of the data, a value of the parameter $\theta$ defining the neural
network may be chosen so as to determine an approximation of $\psid.$
We first study how this might be undertaken by considering this problem where the data is the entire measure $\Upsilon$ -- 
the population loss\index{loss!population} perspective.

\begin{definition} \label{def:bnna} 
Assume function $\psid:\R^d \to \R$ is given. For fixed dimensions $\{d_\ell\}_{\ell=1}^L$,
we define the \emph{best neural network approximation} to $\psid$ by $\psi^*=\psi(\cdot\,;\thetas)$ where
\begin{align}
\thetas \in \argmin_{\theta \in \Theta} \J(\theta),
\end{align}
and $\J(\cdot)$ is the \emph{risk}
\begin{equation} \label{eq:trisk}
   \J(\theta):=\frac12|\psid-\psi(\cdot;\theta)|_{\hm}^2
\end{equation}
\end{definition}

We now discuss what is done in practice to determine a neural network approximation of $\psid$, given Data
Assumption \ref{data:sl}, rather than knowledge of $\Upsilon$ itself. Definition \ref{def:bnna} of best approximation requires
the probability density function $\Upsilon$ and the function $\psid,$ neither of which are known. 
What we do have access to is the data set in Data Assumption \ref{data:sl} which contains
implicit information about both $\Upsilon$ and $\psid$ and, in particular, defines the empirical
measure $\mn(u)$ in Definition \ref{def:emp}. It is then natural to choose the optimal parameter $\thetas$ 
to minimize the  {\em empirical\index{empirical} risk}\index{empirical!risk}
$$\J^N(\theta):= \frac12|\psid-\psi(\cdot;\theta)|_{\hmn}^2.$$
Notice that, since we have assumed that $\lambda=0$,
\begin{equation}
\label{eq:erm}   
\J^N(\theta)=\frac{1}{2N} \sum_{n=1}^N |y^{(n)}-\psi(u^{(n)};\theta)|^2.
\end{equation}
Thus in practice we specify the dimensions $\{d_\ell\}_{\ell=1}^L$ and then numerically estimate $\thetas$ through
empirical risk minimization. 

Minimizing \eqref{eq:erm} leads to an implementable strategy for function approximation.
The empirical\index{empirical!risk} risk is typically non-convex as a function of $\theta$, 
with multiple saddle points and local minima. Robust software to perform the minimization via the use of 
stochastic gradient descent (Chapter \ref{ch:optimization}) is readily
available. In practice $\thetas$ is obtained using this software and, whilst the value obtained is
not expected to be the global minimizer of \eqref{eq:erm}, empirical evidence suggests that the
values obtained perform well at making the loss function $\J^N$ in \eqref{eq:erm} close to optimal.
We write $\psis(u):=\psi(u;\thetas).$

\begin{remark}
\label{rem:slp}
We have derived the loss function $\J^N(\cdot)$ in \eqref{eq:erm} under Data Assumption \ref{data:sl}
with $\lambda=0$. However it may also be derived under the more general assumption
that data set $(u^{(n)},y^{(n)})$ is drawn i.i.d. from (unknown) probability measure $\pi$ on $\R^\du \times \R^\dy.$ 
Examples of such measures may be found from Data Assumption \ref{data:sl} for any $\lambda \ge 0,$ but the setting is much more
general. The risk may then be written as 
$$\J(\theta):=\frac12\Expect^{(u,y) \sim \pi}|y-\psi(u;\theta)|^2$$
and approximated empirically\index{empirical} to again obtain \eqref{eq:erm}.
\end{remark}

\begin{remark} \label{rem:refb}
We derived \eqref{eq:erm} by approximating \eqref{eq:trisk} empirically. If 
the objective functional $\J(\cdot)$ in \eqref{eq:trisk} is viewed as a function of $\psi$,
rather than $\theta$, then we say that the optimization problem is nonparametric\index{nonparametric}. 
It is interesting to note that this problem is convex in $\psi.$ This fact is often used to 
motivate the empirical observation that the parametric optimization problem \eqref{eq:erm} 
becomes simpler when over-parameterized.
\end{remark}

\subsection{Universal Approximation} \label{ssec:NNUA}

We now ask whether, by choosing sufficiently large
dimensions $\{d_\ell\}_{\ell=1}^L$ and choosing optimal parameters, 
we can determine neural networks that approximate a given function to 
arbitrary accuracy. There is a deep  \emph{universal approximation}\index{approximation!universal} 
theory for neural networks that, in broad terms, answers this question 
in the affirmative. We state a typical theorem of this type for a two-layer 
neural network. Citations to proof of the stated theorem, and to other relevant literature, may be found in
the bibliography Section \ref{sec:slbib}.

In this and following sections it is instructive to write the two-layer neural
network in another form. We set $L=1$, and note that integer $d_1$ must be chosen
to specify the neural network. For $i=1, \ldots, d_1$, we let $w^{(i)}$ 
denote the $d_1$ rows of $W_0$, $b^{(i)}$ the $d_1$ 
components of $b_0$ and $\beta_i$ the $d_1$  components of $\beta.$ 
We may also define $\ttheta^{(i)}=(w^{(i)},b^{(i)})$, noting that $\ttheta$
is defined by the concatenation of the $\ttheta^{(i)}$ over $i=1, \ldots, d_1$.
Then we may write the neural network \eqref{eq:dnn} as
\begin{subequations}
\label{eq:twolayer}
\begin{align}
\psi(u;\theta)&=\sum_{i=1}^{d_1}\beta_{i}\varphi(u;\ttheta^{(i)}),\\
\varphi(u;\ttheta^{(i)})&=\sigma\bigl(\langle w^{(i)}, u \rangle + b^{(i)}\bigr).
\end{align}
\end{subequations}

\begin{theorem} \label{thm:nnua}
  Let $\sigma \in C(\R,\R)$ and assume that $\sigma$ is not a polynomial. Consider the neural network \eqref{eq:dnn} with $L=1,$
  which may then be written as \eqref{eq:twolayer}.
  Assume we are given any function $\psid \in C(\R^d,\R)$, any compact set $\mathsf{B} \subset \R^d$ and any $\varepsilon>0.$ Then there is
  integer $d_1$, and parameter $\theta=(W_0,b_0,\beta) \in \R^{d_1 \times d} \times \R^{d_1} \times \R^{d_1}$ such that
  $$\sup_{u \in \mathsf{B}}|\psid(u)-\psi(u;\theta)| < \varepsilon.$$
\end{theorem}

\section{\index{random features}Random Features}\label{sec:RF}

To motivate the idea of random features we return to the two-layer neural network
as defined in \eqref{eq:twolayer}.  Recall that $\theta=(\ttheta,\beta)$ where
$\ttheta=(W_0,b_0)$ and where matrix $W_0$ has rows $\{w^{(i)}\}$ and vector $b_0$
has components $\{b^{(i)}\}$, for $i=1, \ldots, d_1.$ Now imagine that we fix $\ttheta$ and
view the function approximation as being parameterized only by $\beta:$
\begin{subequations}
\label{eq:twolayer2}
\begin{align}
\psi(u;\ttheta,\beta)&:=\sum_{i=1}^{d_1}\beta_{i}\varphi(u;\ttheta^{(i)}),\\
\varphi(u;\ttheta^{(i)})&=\sigma\bigl(\langle w^{(i)}, u \rangle + b^{(i)}\bigr).
\end{align}
\end{subequations}
The functions $\varphi_i(\cdot):=\varphi(\cdot,\ttheta^{(i)})$ are often termed \emph{features}.\index{features}
\begin{align*}
\betas&\in \argmin_{\beta \in \R^{d_1}} \frac12|\psid-\psi(\cdot;\ttheta,\beta)|_{\hmn}^2\\
&= \argmin_{\beta \in \R^{d_1}} \frac{1}{2N}\sum_{n=1}^N |y^{(n)}-\psi(u^{(n)};\ttheta,\beta)|^2
\end{align*}
and $\psis(u):=\psi(u;\ttheta,\betas).$ In contrast to full optimization over $\theta$
discussed in the previous section,
this leads to a convex, indeed quadratic, optimization problem that can be readily solved. 
We are seeking the best solution in the linear span of the features.\index{features}

So far we have simply fixed $\ttheta;$ we have not discussed how to choose it.
One natural approach is to simply pick $\ttheta$ at random from some 
probability measure. Indeed in \eqref{eq:twolayer2} we may pick the pair
$(w^{(i)},b^{(i)})$ i.i.d. at random. It is intuitive that working
with an i.i.d. set of random functions will increase expressivity.
We now take this idea and abstract it, not limiting ourselves
to the specific features\index{features} defined by the two-layer neural network.
To realize this construction we consider $q \in \cP(\R^{d_\vartheta})$
and define $\varphi: \R^d \times \R^{d_\vartheta} \to \R$. We then
choose $m \in \N$ and set
\begin{equation}
\label{eq:RFE}
\psi(u;\beta):=\sum_{i=1}^{m}\beta_{i}\varphi(u;\ttheta_i), \quad \ttheta_i \sim q, \, \index{i.i.d.}\text{i.i.d.}
\end{equation}
We have now defined i.i.d. \emph{random features}\index{random features} $\varphi_i(\cdot):=\varphi(\cdot;\ttheta_i)$ for $i=1,\ldots, m.$

\begin{remark}
\label{rem:RFG}    
The reader will note that the concept of random features is readily
generalized to $\psi: \R^d \times \Theta \to \R^q$ for any integer $q,$
as are neural networks -- see Remark \ref{rem:NNG}.
\end{remark}

Analogous to the situation for the neural network in \eqref{eq:trisk}, the idealized approach to 
determining $\beta$ is to minimize the {\em risk}\index{risk} 
\begin{equation} \label{eq:trisk2}
   \J(\beta):=\frac12|\psid-\psi(\cdot;\beta)|_{\hm}^2.
\end{equation}
Again analogously to our discussion of neural networks, we may employ empirical risk\index{empirical!risk} minimization to determine vector $\beta$ in \eqref{eq:RFE}. In addition we now also introduce a regularization term
and consider the regularized\index{regularizer} quadratic empirical loss function 
\begin{equation}
\label{eq:JNR}    
\J^{N,\mathsf{reg}}(\beta):=\frac12 |\psid-\psi(\cdot;\beta)|_{\hmn}^2
+\frac{\lambdar}{2N}|\beta|^2.
\end{equation}
Informally this recovers $\J(\beta)$ from \eqref{eq:trisk2} in the limit $N \to \infty.$
Here $\lambdar>0$ is the regularization parameter. Although we have considered noiseless data $\lambda=0,$
the specific regularization we use here arises naturally in the context of noisy data $\xi^{(n)} \sim \mathcal{N}(0,1)$ in \eqref{eq:data} 
-- see Subsection \ref{ssec:MCGP}. 

The optimal choice $\beta$ is defined by minimizing  $\J^{N,\mathsf{reg}}(\beta).$ Using \eqref{eq:ERM} this
minimization problem may be written as
\begin{subequations}
\label{eq:erm3}
\begin{align}
\betas&\in \argmin_{\beta \in \R^{m}} \J^{N,\mathsf{reg}}(\beta),\\
\J^{N,\mathsf{reg}}(\beta)&:=\frac{1}{2N} \sum_{n=1}^N |y^{(n)}-\psi(u^{(n)};\beta)|^2
+\frac{\lambdar}{2N}|\beta|^2.
\end{align}
\end{subequations}
Similarly to the previous section, in which we studied neural networks,
we write $\psis(u):=\psi(u;\betas).$ 
The optimization problem \eqref{eq:erm3} is quadratic. Furthermore, 
since $\lambdar>0$, the system of linear equations is
guaranteed to have a unique solution. This we now prove.

\begin{theorem} \label{t:invert} Assume that $\lambdar>0$.
Then the linear system for $\betas$ defined by the optimization problem
\eqref{eq:erm3} takes the form $B\betas=r$ where $B$ is strictly 
positive-definite.
\end{theorem}

\begin{proof} 
The optimization problem in $\beta$ is quadratic, and hence the minimizer
$\betas$ solves the linear system
$B\betas=r$,
where
$$
B_{ij}
:=
\frac1N\sum_{n=1}^N
\varphi(u^{(n)};\ttheta^{(i)})
\varphi(u^{(n)};\ttheta^{(j)})
+\frac{\lambdar}{N}\delta_{ij},
\qquad
r_i
:=
\frac1N\sum_{n=1}^N
y^{(n)}\varphi(u^{(n)};\ttheta^{(i)}).
$$
For each $u \in \R^d$ define $\varphi(u) \in {\R^{m}}$ to be 
the vector with $i$-th entry $\varphi(u;\ttheta_i)$. 
From this it follows that
$$\langle z,Bz \rangle_{\R^{m}}=\bbE^{u \sim \mn}
|\langle \varphi(u), z \rangle_{\R^{m}}|^2+\frac{\lambdar}{N}|z|^2_{\R^{m}} \ge 
\frac{\lambdar}{N}|z|^2_{\R^{m}}.$$
Thus symmetric matrix $B$ is strictly positive definite since we have shown that all eigenvalues are bounded below by $\frac{\lambdar}{N}>0.$
\end{proof}

\begin{example}
\label{ex:RFF}
Random Fourier features\index{random features!Fourier} take the form
$$\varphi(u;\ttheta)=\cos\bigl(\langle \omega, u \rangle + b\bigr)$$
where $\ttheta=(\omega,b)$ is chosen at random.
Note that this is precisely the form of feature arising in the
two-layer neural network \eqref{eq:twolayer2}, with activation function chosen as
the cosine function. It is natural to take $b \sim U[0,2\pi]$,
and we assume $\omega \sim W,$  for some $W \in \cP(\R^d).$ 
We assume also that $\omega \ind b.$ 
\end{example}

\begin{example}
Each $\varphi(\cdot;\ttheta_i)$ may be chosen as a  neural network, for example, with  the parameters of that
neural network chosen i.i.d. at random from a  given distribution $q$.
\end{example}

The spatial correlation in the random features\index{random features}
is a natural object to study. To this end, we introduce kernel (recall Definition \ref{def:kernel})
\begin{align}
\label{eq:kernel}
c(u,u'):=\bbE^q \bigl[\varphi(u;\ttheta)\varphi(u';\ttheta) \bigr], 
\end{align}
where expectation is over $\ttheta \sim q.$   

\begin{example}
Consider Example \ref{ex:RFF}. We note that 
\begin{align*}
    c(u,u')&=\int_{\R^d} I(\omega) W(\omega) \, d\omega,\\
    I(\omega) & =\frac{1}{2\pi} \int_0^{2\pi} \cos\bigl(\langle \omega,u \rangle +b\bigr) \cos\bigl(\langle \omega,u' \rangle +b\bigr) \, db\\
    &=\frac{1}{2} \cos\bigl(\langle \omega, u-u' \rangle \bigr).
    \end{align*}
Thus the resulting kernel\index{kernel} depends only on 
$u-u': c(u,u')=\mathfrak{c}(u-u')$ for some $\mathfrak{c}: \R^\du \to \R.$
We say that the kernel is 
\em{stationary}.\index{stationary!random feature}
\end{example}

The attraction of the random features\index{random features}
approach to supervised learning is that it leads to  a quadratic
finite-dimensional optimization problem over the $m$ random features.
In the next section we discuss Gaussian process regression\index{Gaussian
process regression} which also 
leads to quadratic finite-dimensional optimization problem, but with dimension
$N$, the number of data points. Furthermore, rather than working in
a set of random features, Gaussian process regression works in the
span of the covariance functions, with one argument evaluated at a data point.
The random features approach may be viewed as a Monte Carlo approximation
of Gaussian process regression.

\section{Gaussian Processes \index{Gaussian!process}}\label{sec:GP}

In this section we describe Gaussian process regression.\index{regression!Gaussian process} 
In Subsection \ref{ssec:kernel} we introduce the elements needed to define Gaussian process regression,
namely the notions of kernel\index{kernel} (Definition \ref{def:kernel}) and RKHS (Definition \ref{def:RKHS}). 
We study  the infinite-dimensional regression\index{regression} problem in Subsection \ref{ssec:reg},
showing how it is, in fact, a finite-dimensional problem in Subsection \ref{ssec:RT}. We conclude by
connecting to random features methodology in Subsection \ref{ssec:MCGP}.

 \subsection{Kernels and RKHSs \index{kernel}} \label{ssec:kernel}
 In the study of random features based regression, the kernel given by \eqref{eq:kernel} is a byproduct of the random features. 
In Gaussian process regression it is the starting point for defining the regression methodology.
In the theory of Gaussian process regression, kernels are used to specify the covariance function of a Gaussian process model; the covariance function quantifies spatial correlations of the process and determines a choice of RKHS in which to formulate the regression problem. In this subsection we show, under mild assumptions, an explicit construction of the RKHS associated with the covariance function of a Gaussian process on a compact domain. In the following Subsection  \ref{ssec:reg} we formulate and study Gaussian process regression in this RKHS. 

Let $D \subset \Ru$ be compact and let $c$ be a symmetric, positive definite and 
continuous kernel\index{kernel}, interpreted as the covariance function of a Gaussian process. The integral operator $\cC :L^2(D) \to L^2(D)$ given by 
\begin{align}
\label{eq:IE}
(\cC\phi)(u)=\int_{D} c(u,u')\phi(u') \,du', \qquad \phi \in L^2(D),
\end{align}
is called the \emph{covariance operator} with kernel $c.$ As discussed in the bibliography Section \ref{sec:slbib},
 our assumptions on $c$ imply that the covariance operator
 admits eigenpairs $\{\sigma_i,\phi_i\}_{i=1}^\infty$ satisfying $\cC \phi_i = \sigma_i \phi_i,$  where the eigenvalues $\{\sigma_i\}_{i=1}^\infty$ are non-negative, and, without loss of generality, decreasingly ordered; and where the eigenfunctions $\{\phi_i\}_{i=1}^\infty$ form an orthonormal basis of $L^2(D)$. Furthermore, Mercer's theorem \index{Mercer's theorem} (see the bibliography Section \ref{sec:slbib}) ensures that the eigenfunctions corresponding to non-zero eigenvalues are continuous on $D$ (which implies that pointwise evaluations are well defined) and that the covariance function can be represented using the eigenpairs of $\cC$ as follows:
\begin{equation}\label{eq:kernelMercer}
    c(u,u') = \sum_{i=1}^\infty \sigma_i \phi_i(u) \phi_i(u'), \qquad u,u' \in D,
\end{equation}
where the convergence is absolute and uniform. This representation will be repeatedly used in what follows. For ease of presentation, we will restrict attention to kernels for which $\cC$ is
injective, so that $\sigma_i>0$ for every $i\geq 1$.

Next, we seek to explicitly characterize the RKHS (Definition \ref{def:RKHS}) corresponding to kernel $c$ using again the eigenpairs of $\cC.$ To that end, let $\langle \cdot, \cdot \rangle_{L^2}$ denote the inner product in $L^2(D),$ and define 
\begin{equation*}
   \Kk := \biggl\{ f \in L^2(D): \sum_{i=1}^\infty \frac{\langle f, \phi_i \rangle_{L^2}^2 }{ \sigma_i} < \infty \biggr\}. 
\end{equation*}
The set $\Kk$ equipped with inner product $\langle \cdot, \cdot \rangle_{\Kk}$ as now defined is a Hilbert space: 
\begin{equation*}
    \langle f,g \rangle_{\Kk} = \sum_{i=1}^\infty \frac{\langle f, \phi_i \rangle_{L^2}  \langle g, \phi_i \rangle_{L^2} }{\sigma_i }, \qquad f,g \in \Kk. 
\end{equation*}
The following proposition shows that $\Kk$ is an RKHS with reproducing kernel $c.$

\begin{proposition}
    For $f \in \Kk$ and $u \in D,$ it holds that 
    \begin{equation}\label{eq:reproducingexample}
        \langle f, c(u, \cdot) \rangle_{\Kk} = f(u).
    \end{equation} 
    Consequently, $\Kk$ is an RKHS with reproducing kernel $c.$
\end{proposition}
\begin{proof}
Note that, using \eqref{eq:kernelMercer}, we have
\begin{equation}\label{eq:auxRKHS}
\langle c(u,\cdot),\phi_i\rangle_{L^2}
=
\sum_{j=1}^\infty
\sigma_j\phi_j(u)\langle\phi_j,\phi_i\rangle_{L^2}
=
\sigma_i\phi_i(u),
\end{equation}
where the last equality follows since $(\phi_i)_{i=1}^\infty$ are
orthonormal in $L^2(D)$. It follows from \eqref{eq:auxRKHS} and
\eqref{eq:kernelMercer} that
$$
\sum_{i=1}^\infty
\frac{
\langle c(u,\cdot),\phi_i\rangle_{L^2}^2
}{\sigma_i}
=
\sum_{i=1}^\infty\sigma_i\phi_i(u)^2
=
c(u,u)<\infty.
$$
Hence $c(u,\cdot)\in\Kk$ and
$\|c(u,\cdot)\|_{\Kk}^2=c(u,u)$. Furthermore, by the
Cauchy--Schwarz inequality,
$$
\sum_{i=1}^\infty
\left|
\langle f,\phi_i\rangle_{L^2}\phi_i(u)
\right|
\leq
\|f\|_{\Kk}
\left(
\sum_{i=1}^\infty\sigma_i\phi_i(u)^2
\right)^{1/2}
=
\|f\|_{\Kk}c(u,u)^{1/2}.
$$
Consequently, the $L^2$-expansion of $f$ converges absolutely at every
$u\in D$ and defines a pointwise representative of $f$, also denoted
by $f$. Therefore,
$$
\begin{aligned}
\langle f,c(u,\cdot)\rangle_{\Kk}
&=
\sum_{i=1}^\infty
\frac{
\langle f,\phi_i\rangle_{L^2}
\langle c(u,\cdot),\phi_i\rangle_{L^2}
}{\sigma_i}\\
&=
\sum_{i=1}^\infty
\langle f,\phi_i\rangle_{L^2}\phi_i(u)
=
f(u).
\end{aligned}
$$
\end{proof}

An important example of a kernel widely used in Gaussian process regression is the \emph{Gaussian kernel} given by   
\begin{equation}\label{eq:Gaussiankernel}
    c(u,u') := \exp\Bigl( - \frac{1}{2} |u - u' |^2   \Bigr), \qquad u,u' \in D. 
\end{equation} 
The following theorem provides a universal approximation result for Gaussian kernels, analogous to Theorem \ref{thm:nnua} for neural networks. 
\begin{theorem} \label{thm:GPua}
Let $\Kk$ be the RKHS of a Gaussian kernel on a compact domain $D\subset \R^d.$
  Assume we are given any function $\psid \in C(D,\R)$ and $\varepsilon>0.$ Then there exists $\psi \in \Kk$ such that
  $$\sup_{u \in D}|\psid(u)-\psi(u)| < \varepsilon.$$
\end{theorem}
A reference containing proof of this result, together with additional examples of kernels satisfying a similar universal 
approximation property, can be found in the bibliography Section \ref{sec:slbib}.

\begin{remark}
\label{rem:GPG}    
Recall that the concepts of neural networks and random features are readily
generalized to $\psi: \R^d \times \Theta \to \R^q$ for any integer $q,$
 -- see Remarks \ref{rem:NNG} and \ref{rem:RFG}. Generalizing Gaussian processes
 requires a significant conceptual leap, in that the kernel $c$  from Definition
 \ref{def:kernel} becomes matrix valued: $c: D \times D \to \R^{q \times q}.$
\end{remark}

\subsection{Regression}
\label{ssec:reg}

We work under Data Assumption \ref{data:sl} in the setting $\lambda>0.$
We  assume that $\Kk$ is an RKHS\index{reproducing kernel Hilbert space!RKHS} with reproducing kernel $c$, as constructed in the previous subsection,
and seek to find approximation to $\psid$, function $\psis$, through the following minimization problem:
\begin{subequations}
\label{eq:erm4}    
\begin{align}
\psis &\in \argmin_{\psi \in \Kk} \J^N(\psi),\\
\J^N(\psi)&:=\frac{1}{2N} \sum_{n=1}^{N}|y^{(n)}-\psi(u^{(n)})|^2+\frac{\lambda}{2N}\|\psi\|_{\Kk}^2.
\end{align}
\end{subequations}
This is known as Gaussian process regression.\index{Gaussian process regression} 
In the limit $N \to \infty$ this recovers, informally, the problem 
\begin{subequations}
\label{eq:erm45}    
\begin{align}
\psis &\in \argmin_{\psi \in \Kk} \J(\psi),\\
\J(\psi)&:= \frac12|\psid-\psi|_{\hm}^2.
\end{align}
\end{subequations}
In contrast to the previous sections on neural network and random features based
regression, in which the loss functions \index{loss!function} were defined over finite-dimensional parameter spaces, Gaussian process regression leads to a loss defined over the infinite-dimensional space $\Kk.$  
However, remarkably, the resulting optimization problem reduces to a finite-dimensional problem of
dimension defined by the size of the data set. This we now show.

\subsection{The Representer Theorem}
\label{ssec:RT}

The infinite-dimensional optimization problem \eqref{eq:erm4}
over $\Kk$ has the following remarkable property, which
is often referred to as the 
{\em representer theorem}\index{representer theorem}, and demonstrates
that the optimization problem is intrinsically finite dimensional: 

\begin{theorem}
\label{t:this1}
Function $\psis$ solving \eqref{eq:erm4} has the form
\begin{align}
\label{eq:eform}
\psis(u)=\sum_{n=1}^N \alphas_n c(u,\un).
\end{align}
Furthermore, the coefficients $\alphas$ solve the
following quadratic minimization problem: 
\begin{align*}
\alphas& \in  \argmin_{\alpha \in \R^N} \mathsf{A}^N (\alpha),\\
\mathsf{A}^N(\alpha)&:=\frac{1}{2N} \sum_{r=1}^N\Big|y^{(r)}-  \sum_{ n=1}^N  \alpha_n c(\xr, \un  )\Big|^2+\frac{\lambda}{2N}\sum_{n,r=1}^N \alpha_n \alpha_r c(\xr,\un).
\end{align*}
\end{theorem}

\begin{remark}
In the context of supervised learning, notice that the preceding
theorem shows that the optimization problem \eqref{eq:erm4} is in fact
finite dimensional in nature, despite its infinite-dimensional
formulation. Note, furthermore, the key fact that solution to the 
optimization problem is defined entirely by 
knowledge of the kernel\index{kernel} 
$c$ and by the data $\Bigl\{(\un, y^{(n)}) \Bigr\}_{n=1}^N.$
\end{remark}

\begin{proof}[Proof of Theorem \ref{t:this1}]
Using the reproducing property \eqref{eq:reproducingexample}, the \index{objective}objective function $\J^N$ in \eqref{eq:erm4} can be expressed as 
\begin{align*}
N\J^N(\psi)&=\frac12 \sum_{n=1}^{N}|y^{(n)}-\psi(u^{(n)})|^2+\frac{\lambda}{2}\|\psi\|_{\Kk}^2 \\ 
 & = \frac12 \sum_{n=1}^{N}|y^{(n)}-\langle c(\cdot, u^{(n)}), \psi  \rangle_{\Kk}|^2+\frac{\lambda}{2}\|\psi\|_{\Kk}^2 \, .
\end{align*}
Writing $\J^N(\psi+h)=\J^N(\psi)+\langle D\J^N(\psi),h \rangle_{\Kk}+\mathcal{O}(\|h\|_{\Kk}^2)$
identifies the variational derivative $D\J^N(\psi)$ of $\J^N$, evaluated at any point $\psi.$ Setting this derivative to zero at minimizer $\psi=\psis$ gives, for $u \in D,$
\begin{align*}
\lambda \psis(u)&=\sum_{n=1}^N \bigl(y^{(n)}-\langle c(\cdot, u^{(n)}), \psis  \rangle_{\Kk}\bigr)c(u, u^{(n)})\\ 
&=\sum_{n=1}^N \bigl(y^{(n)}-\psis(\un)\bigr)c(u, u^{(n)}).
\end{align*}
This shows that $\psis$ is in the linear
span of $\{c(\cdot,\un)\}_{n=1}^N,$ establishing the first part of the result.

For the second part, let 
$$\Kk^N= \Bigl\{\psi \in \Kk: \psi=\sum_{n=1}^N \alpha_n c(\cdot,\un), \quad \alpha_n \in \R\,\, \forall n \in \{1,\ldots, N\} \Bigr\}$$
and define the optimization problem 
\begin{subequations}
\label{eq:erm44}    
\begin{align}
\psis &\in  \argmin_{\psi \in \Kk^N} \J^N(\psi),\\
\J^N(\psi)&:=\frac{1}{2N} \sum_{n=1}^{N}|y^{(n)}-\psi(u^{(n)})|^2+\frac{\lambda}{2N}\|\psi\|_{\Kk}^2.
\end{align}
\end{subequations}
Because we have established that \eqref{eq:eform} holds for the minimizer 
$\psis$ of \eqref{eq:erm4}, it follows that  $\psis$ will also be the minimizer
of \eqref{eq:erm44}. 
Finally, using the fact that $\psi \in \Kk^N$, and using the identities \eqref{eq:kernelMercer} and \eqref{eq:auxRKHS}, it follows that 
\begin{align*}
    \|\psi\|_{\Kk}^2 & = \sum_{i=1}^\infty \frac{1}{\sigma_i} \Bigl\langle  \sum_{n=1}^N \alpha_n c(\cdot, u^{(n)}), \phi_i\Bigr\rangle_{L^2} \Bigl\langle  \sum_{r=1}^N \alpha_r c(\cdot, u^{(r)}), \phi_i \Bigr\rangle_{L^2} \\
    & = \sum_{i=1}^\infty  \frac{1}{\sigma_i} \sum_{n=1}^N \alpha_n  \sigma_i \phi_i(u^{(n)}) \sum_{r=1}^N \alpha_r  \sigma_i \phi_i(u^{(r)}) \\
    &= \sum_{n,r=1}^N \alpha_n \alpha_r \sum_{i=1}^\infty \sigma_i \phi_i(u^{(n)}) \phi_i(u^{(r)} ) \\ 
    & = \sum_{n,r =1}^N \alpha_n \alpha_r c(u^{(n)}, u^{(r)}),
\end{align*}
concluding the proof. 
\end{proof}

We define $C\in\R^{N\times N}$ to be the matrix with entries
$C_{nr}=c(\un,\xr)$. We assume that $C$ is invertible and denote its
inverse by $S$, with entries $S_{nr}$. This assumption holds, for
example, if $c$ is strictly positive definite and the input points
$\un$, $n=1,\ldots,N$, are pairwise distinct.
Then we have:

\begin{corollary}
If $\ps \in \R^N$ is vector with entries $\ps_n=\psis(\un),$ then
$\ps=C\alphas$ and
\begin{align*}
\ps&\in  \argmin_{p \in \R^N} \J^N(p),\\
\J^N(p)&:=\frac{1}{2N}\sum_{n=1}^N |y^{(n)}-p_n|^2+\frac{\lambda}{2N}\sum_{n,r=1}^N p_n p_r S_{n,r}.
\end{align*}
\end{corollary}


\subsection{Connection to Random Features}
\label{ssec:MCGP}

The minimization problems \eqref{eq:erm3}, from random feature
regression, and \eqref{eq:erm4}, from Gaussian process regression, are
closely connected. Given $\ttheta_i\sim q$ i.i.d., define the empirical
kernel
\[
c_m(u,u')
:=
\frac1m\sum_{i=1}^m
\varphi(u;\ttheta_i)\varphi(u';\ttheta_i).
\]
Then
\[
\bbE[c_m(u,u')]=c(u,u'),
\]
so that $c_m$ is a Monte Carlo\index{Monte Carlo} approximation of the
kernel $c$ in \eqref{eq:kernel}.

Let $\Kk_m$ denote the RKHS with reproducing kernel $c_m$. Its elements
have representations of the form
\[
\psi(u;\beta)
=
\frac1{\sqrt m}
\sum_{i=1}^m\beta_i\varphi(u;\ttheta_i),
\]
with
\[
\|\psi\|_{\Kk_m}^2
=
\inf\left\{
|\beta|^2:
\psi(\cdot)
=
\frac1{\sqrt m}
\sum_{i=1}^m\beta_i\varphi(\cdot;\ttheta_i)
\right\}.
\]
Replacing $\Kk$ by $\Kk_m$ in \eqref{eq:erm4} therefore reduces the
optimization problem to
\[
\min_{\beta\in\R^m}
\left\{
\frac1{2N}\sum_{n=1}^N
\left|
y^{(n)}
-
\frac1{\sqrt m}\sum_{i=1}^m
\beta_i\varphi(u^{(n)};\ttheta_i)
\right|^2
+
\frac{\lambda}{2N}|\beta|^2
\right\}.
\]
Upon setting $\widetilde\beta=\beta/\sqrt m$, this agrees with
\eqref{eq:erm3} with $\lambdar=m\lambda$. Thus random feature
regression may be viewed as a finite-rank Monte Carlo approximation of
Gaussian process regression.


\section{Approximation Properties}
\label{sec:AP}

Theorem \ref{thm:nnua} demonstrates that two-layer neural networks
are, in principle, able to approximate a given continuous function arbitrarily well;
related results exist for neural networks with more than two layers and are cited
in the bibliography Section \ref{sec:slbib}.
Theorem \ref{thm:GPua} provides a similar result for approximation
using Gaussian kernels; and in view of the Representer Theorem \ref{t:this1}
this theorem suggests that Gaussian process regression\index{Gaussian process regression} also leads to
a methodology that can approximate a given continuous function arbitrarily well;
indeed such results exist, with rates, for many different kernels and are cited
in the bibliography Section \ref{sec:slbib}. In addition, there are also density results
for random features approximation, also cited in the bibliography Section \ref{sec:slbib}.

To be maximally useful, these results on approximation theory need to be tied to
the outcome of the optimizations defining parameter choices found in practice.
For Gaussian process regression the situation is straightforward since the parameter optimization
problem is quadratic with known solution; and indeed there is substantial development
of approximation theory concerning this known solution.
The same is true for approximation using random features, for which explicit solution of the
quadratic optimization problem is also known, and which may be viewed as a form of
Gaussian process regression in a space defined by the span of the random features.
The situation for neural networks, when parameters are trained
on an empirical risk function, is considerably less well-developed.
Statistical learning theory\index{statistical learning theory} 
addresses this question by first studying approximation for minimizers of the risk, and then studying the
effect of empiricalization. However, it still leaves open the question of how typical optimizers
used to learn neural networks, such as those based on  
stochastic gradient descent (see Chapter \ref{ch:optimization}),\index{stochastic gradient descent} perform in practice.

\section{Bibliography}
\label{sec:slbib}

The subject of \index{neural network}neural network function \index{approximation}approximation, the
core task of \index{supervised learning}supervised learning,
is overviewed in \cite{devore2021neural}.
Four key universal \index{approximation!universal}approximation theorems for neural networks
were published in 1989:
\cite{carroll1989construction,cybenko1989approximation,funahashi1989approximate,stinchcombe1989universal}; 
in particular, Theorem \ref{thm:nnua} is stated
and proved, by synthesizing this work, in \cite[Theorem 3.1]{pinkus1999approximation}.
For introductions to statistical learning theory\index{statistical learning theory} 
see the texts \cite{hastie2005elements,vapnik2013nature}.

The use of \index{random features}random features was popularized as a methodology,
and analyzed rigorously, in the collection of papers
\cite{rahimi2007random,rahimi2008uniform,rahimi2008weighted}.
We have articulated a specific link between \index{neural network}neural networks
and \index{random features}random features; a related concept underpins the subject of the
{\em neural tangent kernel}\index{kernel!neural tangent} \cite{jacot2018neural}.

\index{Gaussian!process}Gaussian processes are described in \cite{williams2006gaussian}; in that text the mean function and covariance function characterize the Gaussian probability measure on function
space. In \cite{bogachev1998gaussian} Gaussian probability measure on separable Banach
space $X$ is defined by asking that all linear functionals are Gaussian. Subsequently the mean, viewed as a point in $X$, and the covariance
operator, mapping $X^*$ into $X$, may be defined. In this Gaussian measure perspective, the RKHS is referred to as the Cameron--Martin space\index{Cameron--Martin}. Section 4 of the lecture notes
\cite{hairer2009introduction} provides a concise readable introduction to the subject of Gaussian
measure on separable Banach space. Kernel-based\index{kernel} methods more generally are described in
\cite{owhadi2019operator}. We refer to \cite{micchelli2006universal} for a general theory and numerous examples of kernels with the universal approximation property, including the result for squared exponential kernels in Theorem \ref{thm:GPua}. 
Error estimates for interpolation
using \index{Gaussian!process}Gaussian processes are developed in \cite{wendland2004scattered}.
The link between regression and \index{Bayesian!inversion}Bayesian inversion with
\index{Gaussian!process}Gaussian process \index{prior}priors is developed in depth in
\cite{craven1978smoothing,wahba1990spline}.
Mercer's theorem \index{Mercer's theorem}is established in \cite{mercer1909functions} and it can be now found in numerous textbooks; see for example \cite{dunford1963linear}, which also contains additional background on the spectral decomposition of the covariance operator used in Subsection \ref{ssec:kernel}, and \cite{sullivan2015introduction} in the context of uncertainty quantification\index{uncertainty quantification}. In RKHS theory, the spectral decomposition characterizes the RKHS of symmetric, positive definite continuous kernels, as discussed in this chapter. Mercer's theorem is also useful in the theory of integral equations and in the study of stochastic processes, where it underpins the derivation of Karhunen--Loève expansions \cite{pavliotis2014stochastic}.

The minimization problem \eqref{eq:erm4} may be derived by
application of a generalization of Bayes Theorem\index{Bayes Theorem}
\ref{t:bayes},\footnote{Note that in this interpretation the unknown on which we seek
a posterior distribution is now $\psi \in L^2(D)$ 
(whereas in Bayes Theorem\index{Bayes Theorem} \ref{t:bayes}
it was $u \in \Ru$).}
from $\Ru$ to $L^2(D)$, and then a generalization of the notion
of MAP estimator\index{MAP estimator} -- Definition \ref{def:map} -- 
also from $\Ru$ to $L^2(D).$ The Bayesian inverse problem is defined 
as follows.  We place as \index{prior}prior on $\psi$ a 
\index{Gaussian!random field}Gaussian random field with mean zero and non-negative definite continuous 
covariance\index{covariance!function} function $c(u,u');$ 
this covariance function can be equivalently seen as specifying a Gaussian measure $\cN(0,\cC)$ in $L^2(D),$ where $\cC$ is the covariance
operator\index{covariance!operator} given by \eqref{eq:IE}. 
The \index{likelihood}likelihood 
is defined by assuming noisy data ($\lambda>0$) in \eqref{eq:data}, 
where $\xi^{(n)} \sim \mathcal{N}(0,1)$ \index{i.i.d.}i.i.d.\,.
Then the negative log-likelihood is given by 
$$\frac{1}{2\lambda}\sum_{n=1}^N |y^{(n)}-\psi(\un)|^2.$$  
Because the prior on $\psi$ is Gaussian and the log-likelihood is
quadratic in $\psi$, the generalization of 
Bayes\index{Bayes Theorem} Theorem \ref{t:bayes} shows that the
posterior is conjugate\index{posterior!conjugate} to the prior
and is Gaussian.\index{Gaussian} The \index{posterior}posterior mean, 
which is also a form of \index{MAP estimator}MAP estimator, 
then satisfies \eqref{eq:erm4}. We remark that minimization 
of \eqref{eq:erm3} also
has a similar interpretation as a \index{MAP estimator}MAP estimator with respect to
a zero-mean \index{Gaussian!process}Gaussian process with covariance
function
\[
\frac{m\lambda}{\lambdar}c_m(u,u').
\]
In particular, when $\lambdar=m\lambda$, this covariance function is
$c_m$, the Monte Carlo approximation of $c$ introduced in
Subsection~\ref{ssec:MCGP}. A definition of \index{MAP estimator}MAP estimator in infinite dimensions 
may be found in \cite{dashti2013map}; see also \cite{lambley2023strong,klebanov2023maximum,kretschmann2022minimizers,ayanbayev2021convergence} for related discussion.

The use of various machine-learning-inspired \index{approximation}approximation
methods in the solution of PDEs may be found in
\cite{raissi2019physics,nelsen2021random,chen2021solving}.
In particular, these papers address, respectively, the use of neural networks, random features and
Gaussian processes to solve PDEs.

\chapter{\Large{\sffamily{Unsupervised Learning and Generative Modeling\index{generative model}}}}
\label{ch:UL}

This chapter is concerned with the subject of 
\index{unsupervised learning}{\em unsupervised learning}. 
The following data assumption is made:
\begin{dataassumption}\index{Data Assumption}\label{assumption:unsupervised}
We have available data in the form
\begin{align}
\label{eq:datau}
U:=\{\un\}_{n=1}^N,
\end{align}
assumed to be drawn \index{i.i.d.}i.i.d. from (unknown) 
probability density function $\Upsilon$ on $\R^d$.
\end{dataassumption}
Given data of this form there are numerous questions one might ask.
One widely studied problem is that of detecting
{\em clustering}\index{clustering} within the data; we will not
pursue this important topic because it is not of direct relevance
to the solution of \index{inverse problem}inverse problems and
\index{data assimilation}data assimilation, our focus in these notes.
Here we pursue the problem of \index{generative modeling}{\em generative modeling}: 
we model $\Upsilon$ and/or generate new approximate samples from it, 
given only the data $U.$
Machine learning ideas based on generative modeling are useful
for both \index{inverse problem}inverse problems and
\index{data assimilation}data assimilation.

Data Assumption \ref{assumption:unsupervised} should be compared with the 
\index{supervised learning}{\em supervised learning}
Data Assumption \ref{data:sl} (or the generalization discussed
in Remark \ref{rem:slp}), where we are given input-output pairs,
linked through an unknown function; our focus in that setting is on learning
about that \emph{function}. Under the unsupervised learning Data Assumption \ref{assumption:unsupervised} 
our focus is on learning about a \emph{probability measure} underlying the data.

Given Data Assumption \ref{assumption:unsupervised} we have access to the 
empirical\index{empirical} measure (see Definition \ref{def:emp}) 
\begin{align}
\label{eq:emp}
\mn(u)=\frac{1}{N} \sum_{n=1}^N \delta_{\un}(u).
\end{align}
For any probability measure $\intp \in \cP(\R^d)$, recall the notation
$\intp(f):= \mathbb{E}^{u \sim \intp}f(u)$. 
Recall also the metric
\begin{equation*}
\dr(\pr,\pr') = \sup_{|f|_{\infty} \leq 1} \left|\mathbb{E}\left[\left(\pr(f) - \pr'(f)\right)^2\right]\right|^{1/2},
\end{equation*}
on the space of random probability measures, from Subsection \ref{ssec:rpm}.
Theorem \ref{t:MC} shows that $\dr(\Upsilon,\mn)^2 = \mathcal{O}(\frac{1}{N})$.
This highlights that, for large $N$ and on average with respect to the i.i.d. sampling underlying the data,
the empirical measure is close to the true measure we wish to learn about.

We conclude this introductory section
by illustrating the idea of generative modeling with two examples.

\begin{example}
\label{ex:g1}
A natural starting point for generative modeling\index{generative model} is to discuss 
\emph{density estimation}\index{density estimation}. This methodology
seeks to approximate $\Upsilon(u)$, from within some class of densities, on the
basis of the data summarized in $\mn(u).$ 
Some approximations to $\Upsilon$ within such classes can be used
to generate new (approximate) samples. An example is when the approximation is a 
Gaussian mixture\index{Gaussian!mixture}.
However, the goal of density estimation is not always generative modeling:
it may be to have a smooth density which can be differentiated or otherwise manipulated
in a way that $\mn(u)$ cannot. 
\end{example}

\begin{example}
\label{ex:g2}
We introduce the {\em measure transport}\index{measure transport} task of determining
function $g: \R^{\dz} \to \R^d$ and probability density function $\zeta$ on $\R^{\dz}$
such that if $z \sim \zeta$, then (approximately) $g(z) \sim \Upsilon.$ Thus the perfect pair
$g,\zeta$ would satisfy $g_\sharp \zeta = \Upsilon;$ recall that $g_\sharp$ denotes 
the pushforward\index{pushforward} operation under function $g$. 
Measure $\zeta$ is chosen to be easy to sample from; a common setting is to assume that 
$\zeta$ is \index{Gaussian}Gaussian. Given $\zeta$, the task of identifying $g$ must be undertaken empirically\index{empirical}, since only $\mn$ is available, not $\Upsilon.$
It is also possible to study relaxations of this problem in which a specified measure $\zeta$ is convolved with a specified \index{Gaussian}Gaussian whose mean is function $g$; $g$ is then chosen so that
this convolution approximates $\Upsilon.$
\end{example}

In Section \ref{sec:den} we describe the subject of 
density estimation\index{density estimation}, introduced in
Example \ref{ex:g1}, and taking a particular perspective on it which links to several
themes arising in other places in the notes. In the remainder of the chapter
we study variants on the idea of measure transport\index{measure transport} 
illustrated in Example \ref{ex:g2}.
All the methods connect the target measure $\Upsilon$, which we 
wish to sample, and defined on $\Ru$, to a measure on a latent
space $\R^{\dz}.$
In Section \ref{sec:transport2} we study a general approach to measure transport akin to the density estimation approach studied in Section \ref{sec:den}; typically these methods employ $\dz=d.$  Normalizing flows\index{flow!normalizing} are studied in Section \ref{sec:NORM}.
The choice $\dz \ne d$ is possible here too. But we study a continuum limit, using 
neural ODEs\index{neural ODEs}, where the methodology has at its core an invertible mapping with $\dz=d.$ In Section \ref{sec:score} we study score-based\index{score} approaches that build on Langevin\index{Langevin} sampling. Sections \ref{sec:auto} and \ref{sec:vauto}
introduce autoencoders and variational autoencoders respectively,
where an approximate inverse for $g$ is sought, although typically
$\dz<d$ so caution is required to define this carefully. 
Generative adversarial networks\index{generative adversarial network}
(GANs)\index{GAN|see{generative adversarial network}} are introduced in Section \ref{sec:gan}, applicable with $\dz \ne d.$

\section{\index{density estimation} Density Estimation}\label{sec:den}

Recall notation $\cP(\R^d)$ for the set of all probability density 
functions \index{probability!density function} on $\R^d.$
Consider the problem of estimating the probability density function\index{probability!density function} 
underlying data $\UN:=\{\un\}_{n=1}^N$ given in Data Assumption \ref{assumption:unsupervised}.
We seek to solve the problem by finding a probability density 
function\index{probability!density function} from 
a tractable class of probability density functions $\mathcal{Q} \subset \cP(\R^d)$,  
for the purpose of computations and for the
purpose of revealing an explicit, interpretable form. To be concrete we
assume that, for some parameter set $\Theta \subseteq \R^p$, 
$$\cQ= \bigl\{q(\cdot\,;\theta) \in \cP(\R^d): \theta \in \Theta \bigr\}.$$

We start by formulating the problem of density estimation as one of finding
$\bbP(\theta|U)$ by applying Bayesian inference\index{Bayesian!inference}, the methodological
approach to inversion introduced in Chapter \ref{ch1}. If we place prior\index{prior} $\bbP(\theta)$ on unknown parameter $\theta,$ then, noticing that $\bbP(\UN|\theta)=\prod_{n=1}^N\, q(\un;\theta)$, we obtain from Theorem \ref{t:bayes} that the posterior\index{posterior} on $\theta|\UN$ is given by
\begin{equation}
    \label{eq:denb}
  \bbP(\theta|\UN) \propto   \prod_{n=1}^N\, q(\un;\theta) \bbP(\theta).
\end{equation}
If we seek a maximum \emph{a posteriori} (\index{MAP estimator}MAP) estimator for $\theta,$
then, following the developments in Subsection \ref{ssec:MAP},
\begin{align*}
\thetas &\in \argmin_{\theta\in \Theta} \J_{\mbox{\tiny{\rm{MAP}}}}^N(\theta),\\
\J_{\mbox{\tiny{\rm{MAP}}}}^N(\theta) &= - \sum_{n=1}^N\, \log q(\un;\theta) -\log \bbP(\theta).
\end{align*} 
The first term in the definition of $\J_{\mbox{\tiny{\rm{MAP}}}}^N(\cdot)$ is a specific 
instance of the loss function\index{loss!function} \eqref{eq:loss},
and the second is the regularizer\index{regularizer} \eqref{eq:reg}. 
If we drop the regularizer (see Remark \ref{rem:flat}; we are then using a flat prior\index{prior!flat})
then we obtain the maximum likelihood estimation (MLE)\index{maximum likelihood estimation}\index{maximum likelihood estimation!MLE}
problem
\begin{align*}
\thetas &\in \argmin_{\theta\in\Theta} \J_{\mbox{\tiny{\rm{MLE}}}}^N(\theta),\\
\J_{\mbox{\tiny{\rm{MLE}}}}^N(\theta) &= - \sum_{n=1}^N\, \log q(\un;\theta).
\end{align*} 

This MLE\index{maximum likelihood estimation!MLE} may also be derived directly, simply by asking for the
value of $\theta$ which maximizes $\bbP(\UN|\theta).$ It may also be derived by minimizing an empirical approximation of a KL divergence,\index{divergence!Kullback--Leibler} an idea that we now explain and that we will use repeatedly in other places
in the book. Consider the minimization problem
\begin{align*}
\thetas &\in \argmin_{\theta\in\Theta} \F(\theta),\\
\F(\theta) &:= \dkl\bigl(\Upsilon \|q(\cdot\,;\theta)\bigr).
\end{align*}
Since $\Upsilon$ is unknown under Data Assumption \ref{assumption:unsupervised}, we approximate the solution to this minimization problem by replacing the objective $\F(\cdot)$ with an objective defined using only samples from $\Upsilon$. 
Notice that the KL divergence\index{divergence!Kullback--Leibler} defining the objective $\F$ 
has the form
$$\dkl \bigl(\Upsilon \| q(\cdot\,;\theta)\bigr)=\bbE^{u \sim \Upsilon}\Bigl[\log \Upsilon(u)-\log q(u;\theta)\Bigr].$$
The first of the two terms does not depend on $\theta$, and hence minimizing $\F(\theta)$ is the same as minimizing 
the second term
$$-\bbE^{u \sim \Upsilon} \Bigl[ \log q(u;\theta) \Bigr]$$
over $\theta.$  Replacing expectation with respect to $\Upsilon$ with expectation with respect to the empirical\index{probability!measure!empirical} measure $\mn$ defined in \eqref{eq:emp}, we obtain 
$$-\bbE^{u \sim \mn} \Bigl[ \log q(u;\theta) \Bigr] = - \frac{1}{N} \sum_{n=1}^N\, \log q(\un;\theta) =  \frac{1}{N}\J_{\mbox{\tiny{\rm{MLE}}}}^N(\theta).$$
Thus minimization of an empirical approximation of $\F(\theta)$ is the same as computing the
minimizer of $\J_{\mbox{\tiny{\rm{MLE}}}}^N(\cdot),$ the  MLE.\index{maximum likelihood estimation!MLE}

\begin{remark}
\label{rem:discuss0}
We have shown that finding the MLE\index{maximum likelihood estimator!MLE} is closely connected to finding the closest point in $\cQ$ to $\Upsilon$ in the KL divergence.\index{divergence!Kullback--Leibler} 
Note, however, that the
constant term $\bbE^{u \sim \Upsilon}\log \Upsilon(u)$, which we removed from the
optimization since it does not depend on $\theta$, is undefined if $\Upsilon$ is replaced by $\mn.$ 
This issue arises elsewhere in this book, and in particular in the next subsection. We emphasize 
the issue here, in the context of density estimation, since it constitutes the canonical setting in which it arises.
\end{remark}

\section{Transport Methods}\label{sec:transport2}

\index{transport}

In this section we employ transport to characterize $\Upsilon \in \mathcal{P}(\R^d)$, from the
samples defined by Data Assumption \ref{assumption:unsupervised}. The methodology may be
viewed as a specific subclass of the MLE approach to density estimation described in the preceding
section. The subclass is one in which density $q$ is defined via pushforward of a simple measure.
To be concrete, we specify a probability measure $\zeta \in \mathcal{P}(\R^d)$ and determine
a map $g: \R^d \to \R^d$ with the property that (at least approximately) $\Upsilon=g_\sharp \zeta$.
The map $g$, then, 
will be an approximate transport map (Definition \ref{def:transportmap}).
The desiderata are that $\zeta$ be easy to sample, so that (approximate) samples from $\Upsilon$ may be
generated by applying $g$ to samples from $\zeta$; 
that $g$ may be determined
given only $\Upsilon^N$ as defined in \eqref{eq:emp}, and not $\Upsilon$ itself; and that $g$ be cheap to evaluate.

Imagine that a suitable choice for $\zeta$ has been made; a Gaussian is a common
choice. Let $\Theta \subseteq \R^{p}$. We introduce a parametric family of 
functions $g(\cdot\,; \cdot): \R^d \times \Theta \to \R^d$. For our methodology it is
important that $g(\cdot\,;\theta) \in C^1(\R^d,\R^d)$ and is invertible, for all $\theta \in \Theta.$
To this end we work with a function class $\cTt$ satisfying the condition that
\begin{equation}
\label{eq:thisclass0}
\cTt \subset \Bigl\{g(\cdot\,;\theta) \in C^1(\R^d;\R^d),\, \forall \theta \in \Theta \, \Bigl| \det D_u g(u;\theta) > 0\, \forall (u,\theta) \in \R^d \times \Theta \Bigr\}.
\end{equation}
Such classes of smooth invertible functions 
$g(\cdot\,;\theta): \R^d \to \R^d$ will be discussed in Section \ref{sec:NORM}.

We then seek parameter $\theta \in \Theta \subseteq \R^{p}$ such that $g(\cdot\,;\theta): \R^d \to \R^d$ realizes the desired approximation. 
To achieve this goal, we consider the optimization problem, over all $g$ from $\cTt$, 
\begin{subequations}
\label{eq:FNJ}
\begin{align}
\F(\theta)&=\dkl \bigl(\Upsilon \| g(\cdot\,; \theta)_\sharp \zeta \bigr),\\
\thetas &\in \argmin_{\theta \in \Theta} \F(\theta).
\end{align}
\end{subequations}
We define the resulting approximate pushforward\index{pushforward} 
map by $\gs=g(\cdot\,;\thetas).$
As mentioned, this approach is a special case of the density estimation approach in
the preceding Section \ref{sec:den}: we parameterize the approximating densities $q(\cdot\,; \theta) = g(\cdot\,;\theta)_\sharp \zeta$ as pushforward of a fixed reference density $\zeta$ by a parameterized transport map $g(\cdot\,; \theta).$

As in Section \ref{sec:den}, to optimize $\F(\cdot)$ 
we will identify $\J(\cdot)$, 
a function which differs from the KL divergence 
only by a constant independent of $\theta,$
and optimize $\J^N(\cdot),$ an empirical\index{empirical}
approximation of $\J(\cdot)$ found by replacing $\Upsilon$ with $\mn.$ 
Lemma \ref{lem:cov} is useful in determining 
an explicit form of $\J$. We now deploy this lemma to 
find a pair of expressions for $\J(\cdot)$. 
From the definition of the KL divergence we have that
$$\F(\theta)=\dkl \bigl(\Upsilon \| g(\cdot\,; \theta)_\sharp \zeta \bigr)  = \bbE^{u \sim \Upsilon} \Bigl[\log \Upsilon(u) - \log g_\sharp \zeta(u)\Bigr].$$
Thus, noting that the desired minimization is not affected by removing the first $\theta$-independent term, and using (\ref{eq:cov}b) to simplify the second, we define
\begin{equation}
\label{eq:OT}
\J(\theta) = -\bbE^{u \sim \Upsilon} \Bigl[\log \zeta \circ g^{-1}(u;\theta)+ \log {\rm det}  D_u(g^{-1})(u;\theta) \Bigr].
\end{equation}
Our desired minimization problem for $\F(\cdot)$ 
is thus equivalent to minimizing
\eqref{eq:OT}. Furthermore, using \eqref{eq:invd} in \eqref{eq:OT}, we may write
\begin{equation}
\label{eq:OT99}
\J(\theta)= -\bbE^{u \sim \Upsilon} \Bigl[\log \zeta \circ g^{-1}(u;\theta)- \log {\rm det}  D_u g\bigl(g^{-1}(u;\theta);\theta\bigr) \Bigr].
\end{equation}
In practice, $\Upsilon$ is only available empirically\index{empirical} and hence we work
under Data Assumption \ref{assumption:unsupervised}. Hence, the objective 
\eqref{eq:OT} or \eqref{eq:OT99} can be approximated as follows: 
\begin{align}
\label{eq:OT991}
\begin{split}
\J^N(\theta)&= 
-\frac{1}{N} \sum_{n=1}^N  \Bigl[\log \zeta \circ g^{-1}(u^{(n)};\theta)+ \log {\rm det}  D_u(g^{-1})(u^{(n)};\theta) \Bigr]\\
&=-\frac{1}{N} \sum_{n=1}^N \Bigl[\log \zeta \circ g^{-1}(u^{(n)};\theta)- \log {\rm det}  D_u g\bigl(g^{-1}(u^{(n)};\theta);\theta\bigr) \Bigr].
\end{split}
\end{align}
Then the \index{forward!map}forward map $\gs$ is obtained at $\theta^\star$ which minimizes $\J^N(\cdot)$.

\begin{remark}
\label{rem:discuss}
By minimizing \eqref{eq:OT991} we thus find an approximate 
expression for $\Upsilon$ 
as the pushforward under $\gs(\cdot) = g(\cdot,\theta^\star)$ of $\zeta$ using~\eqref{eq:cov}. That is,
\begin{equation}
    \Upsilon(u) \approx \gs_\sharp \zeta(u)= (\zeta \circ g^{-1})(u;\theta^\star) {\rm det} D_u g^{-1}(u;\theta^\star).
\end{equation}
An important point to note here is that, to determine 
$\gs$, minimization of either form of the loss function  $\J^N$ in \eqref{eq:OT991} 
requires evaluation of $g^{-1}.$ Thus $g$ must be readily invertible. 
Section \ref{sec:NORM} addresses the issue of constructing invertible transports\index{transport!invertible} in the context 
of \index{flow!normalizing}normalizing flows, providing examples of classes of function $\cTt.$
\end{remark}

\begin{remark}\label{rem:invparam} In this section $\zeta$ is a simple measure, from which we can easily draw unlimited
samples -- a unit Gaussian measure is a natural example; measure $\Upsilon$ is a more complex measure only available to
us through $N$ samples, and from which we wish to draw more samples. To this end we seek $g$ such that
$\Upsilon=g_\sharp \zeta$. Once the approximate transport $\gs$ is identified we can sample $\zeta$, apply $\gs$ to the sample, 
and obtain new
(approximate) samples from $\Upsilon.$ 

We may, however, use the same ideas in a different context. Assume now that
$\Upsilon$ is a known simple measure, from which samples are easily drawn, and $\zeta$ is a more complicated measure 
from which we wish to sample. Now the objective is to determine $T$ so that $\zeta=T_\sharp \Upsilon.$ We thus
consider the following variant on the minimization problem \eqref{eq:FNJ}:
\begin{subequations}
\label{eq:FNJ00}
\begin{align}
\F(\theta)&=\dkl \bigl(T(\cdot\,; \theta)_\sharp \Upsilon \| \zeta \bigr),\\
\thetas &\in \argmin_{\theta \in \Theta} \F(\theta).
\end{align}
\end{subequations}
By Theorem \ref{thm:invariance} this is equivalent, for invertible and sufficiently regular $T$, to
\begin{subequations}
\label{eq:FNJ99}
\begin{align}
\F(\theta)&=\dkl \bigl(\Upsilon \| T^{-1}(\cdot\,; \theta)_\sharp \zeta \bigr),\\
\thetas &\in \argmin_{\theta \in \Theta} \F(\theta).
\end{align}
\end{subequations}
and may thus be tackled by the same techniques described in this section, replacing $g^{-1}$ by $T.$
\end{remark}

\begin{remark}
Theorem \ref{t:KM} demonstrates that, through the Monge\index{transport!optimal, Monge} formulation of
optimal transport\index{transport!optimal}, there \emph{exists} at least one perfect transport map $g_{\rm perfect}$
for which $(g_{\rm perfect})_\sharp \zeta=\Upsilon$ and hence $\dkl(\Upsilon \| (g_{\rm perfect})_\sharp \zeta)=0$. 
We aim to get as close as possible to \emph{a} perfect transport solution, within our
parametric class, by minimizing $\F$ over $\theta.$ But we do not impose that we have
an optimal transport. The resulting variational methodologies resonate with the Monge\index{transport!optimal, Monge} 
formulation of optimal transport, in that they are defined by a transport map, but we do not invoke the same optimality criterion.
\end{remark}

\section{\index{flow!normalizing}Normalizing Flows}\label{sec:NORM}

The \index{transport}transport approach from Section \ref{sec:transport2} requires
invertibility of the map $g$, and efficient computation of the determinant
of the Jacobian. Invertibility of the map $g$ is also useful in other
contexts. \index{flow!normalizing}Normalizing flows address this issue; \emph{normalizing}
reflects the fact that distribution $ \zeta$ is often a \index{Gaussian}Gaussian, whilst 
\emph{flow} connotes the breaking up of the map $g$  into a sequence of
simpler maps which are themselves parameterized, rather than
parameterizing $g$ itself.\footnote{And, although we will derive a continuous-time analog of normalizing\index{flow!normalizing} flows, the neural ODE\index{neural ODEs}, the use
of \emph{flow}\index{flow!dynamical systems} 
here should not be confused with its use in the theory
of continuous-time dynamical systems.} These simpler maps can be constructed to be readily
invertible, so that $g$ is itself invertible; thus we address the concept of invertible transport
introduced in the previous section.

To this end we fix $J \in \mathbb{N}$ and introduce the iteration 
\begin{subequations}
\begin{align}
v_{j+1}&=H(v_j\,;\theta), \quad j=0,\dots,J-1, \\
v_0&=z,
\end{align}
\end{subequations}
where we define $u:=v_J$; here $H(\cdot\,; \theta): \R^d \to \R^d.$ Then 
$u=g(z\,;\theta)$ where 
\begin{equation}
    \label{eq:compose}
g(\cdot\,;\theta)=\underbrace{H(\cdot\,;\theta) \circ H(\cdot\,;\theta) \circ \dots \circ H(\cdot\,;\theta)}_{J\text{ times}},
\end{equation}
the $J$-fold
composition of $H(\cdot\,;\theta)$. 
The proposed method is thus simply a transport of the type
discussed in Section \ref{sec:transport2}, with a specific
construction of $g$ in compositional form. The remainder of
the section comprises two components: firstly showing
specifics of the formulation of the minimization problems
\eqref{eq:OT} or \eqref{eq:OT99} in this compositional
setting; and secondly showing a continuum limit of the
compositional setting, which recovers neural ODEs\index{neural ODEs}. 

\begin{remark}
We have concentrated on the simple setting where the same map $H$ is used
at each step of the composition; this simplifies the exposition, especially in
the continuum limit. However it is commonplace to compose different
maps at each step, as discussed in Remark~\ref{rem:nfna}; furthermore Remark~\ref{rem:nfna2}
highlights the implications of doing so in the neural ODEs\index{neural ODEs} setting.
\end{remark}

\subsection{Structure in the Optimization Problem for $\theta$}

We assume that, for all $\theta \in \Theta \subseteq \R^p$, $H(\cdot\,;\theta): \R^d \to \R^d$ 
is invertible, with inverse
$H^{-1}(\cdot\,;\theta): \R^d \to \R^d $, differentiable, with 
derivative $DH(\cdot\,;\theta): \R^d \to \R^{d \times d}$, and that $\det DH(\cdot,\theta) >0$ everywhere
in $\R^d$. 
Thus $g$ given by \eqref{eq:compose} is an invertible transport\index{transport!invertible}.
Assume that $z \sim \zeta$ and let $p_j(v_j)$ 
denote the density of $v_j$. Then the goal of a normalizing
flow is to choose parameter $\theta$ in $H$ so that
$p_J \approx \Upsilon$ and hence (approximately)
$u \sim \Upsilon;$ this is potentially a more 
straightforward task than working with a directly parameterized $g,$ 
the approach described in Section \ref{sec:transport2}.

Let $v_j=H^{(j)}(z;\theta)$, where $H^{(j)}(\cdot\,;\theta)$ denotes the 
$j$-fold composition of $H(\cdot\,;\theta)$ and recall that $u=g(z;\theta)=H^{(J)}(z;\theta)$.
Note, also, that $z$ is found as the $J$-fold
composition of $H^{-1}(\cdot\,;\theta)$ evaluated at $u$.
Formulae \eqref{eq:invd}, \eqref{eq:cov} show that, since $v_j=H^{-1}(v_{j+1};\theta)$, 
\begin{equation*}
\log p_{j+1}(v_{j+1}) = \log p_j(v_j) - \log {\rm det} DH(v_j;\theta).
\end{equation*} 
By induction, we obtain that
\begin{equation*}
\log p_{J}(v_{J}) = \log p_0(v_0) - \sum_{j=0}^{J-1} \log {\rm det} DH(v_j;\theta),
\end{equation*}
and hence
\begin{equation}
\label{eq:cov2}
\log g_\sharp \zeta(u)   = \log \zeta(z) - \sum_{j=0}^{J-1} \log {\rm det} DH(v_j;\theta).
\end{equation}

In what follows we emphasize the dependence of $v_j$ on $\theta,$ writing $v_j(\theta).$
Note that $\F(\theta)$ given by \eqref{eq:FNJ} differs from
$\J(\theta)=-\bbE^{u \sim \Upsilon} \log  g(\cdot\,; \theta)_\sharp \zeta$ 
by a constant that is independent of $\theta$ and that we minimize
an empirical approximation of $\J(\theta).$ To this end, with the specified 
relationships defined between $(z,\{v_j\})$ and $u$, 
we may consider the following specific case of \eqref{eq:OT99}, 
using the normalization flow structure: 
\begin{equation}
\label{eq:OT999}
\J(\theta)= -\bbE^{u \sim \Upsilon} \Bigl[\log \zeta(z)- \sum_{j=0}^{J-1}
\log {\rm det}  DH\bigl(v_j(\theta);\theta\bigr) \Bigr].
\end{equation}
In practice the expectation 
over $\Upsilon$ in \eqref{eq:OT999} can be approximated using the empirical\index{empirical} measure $\mn$ leading to
loss function $\J^N(\cdot).$
As before we set 
$$\thetas\in \argmin_{\theta \in \Theta} \J^N(\theta)$$
and define the resulting approximate pushforward\index{pushforward} map by 
$\gs=g(\cdot\,;\thetas).$
Note that $g(\cdot\,;\thetas)$ is found from the $J$-fold
composition of $H(\cdot\,;\thetas)$.

\begin{remark}
\label{rem:nfna}
    The preceding setting can be generalized to one in which
    each map in the composition is different and carries its
    own set of parameters to be optimized, so that \eqref{eq:compose} is replaced by
    \begin{equation}
    \label{eq:compose2}
g(\cdot\,;\theta)=H_{J-1}\bigl(\cdot\,;\theta^{(J-1)}\bigr) \circ H_{J-2}\bigl(\cdot\,;\theta^{(J-2)}\bigr) \circ \dots \circ H_0\bigl(\cdot\,;\theta^{(0)}\bigr),
\end{equation}
and $\theta= \bigl\{\theta^{(j)} \bigr\}_{j=0}^{J-1}.$ 
Then, for example, each $H_j$ may have a triangular structure with respect to
some ordering of the elements of $\R^d$. Recall that the term triangular refers to the Jacobian of $H_j$ being a (lower-) triangular matrix. One parameterization of a triangular map is given by
$$H_j(v;\theta) = \begin{bmatrix*}[c] v_1 \\ f_2(v_1;\theta) + v_2 \\ \vdots \\ f_{d}(v_1,\dots,v_{d-1};\theta) + v_d \end{bmatrix*},$$
where $f_i(v_1,\dots,v_{i-1};\theta)$ are parameterized functions. 
Since triangular maps are readily
designed to be invertible, this approach can be useful; and choosing different
orderings at each step and hence different $H_j$ at each step, mitigates
against bias from any particular ordering. While the illustrative parameteterization of a triangular map above is volume preserving, and thus potentially restrictive, more expressive parameterizations can be considered.
\end{remark}

\subsection{Neural ODEs} \label{sec:nODEs}
Neural ODEs\index{neural ODEs} are a special case of transport maps 
where map $g$ is the solution map of an ordinary differential equation (ODE) with flow defined by a 
parameterized vector field; finding $g$ is replaced by the task of finding 
the vector field. A continuum limit of \index{flow!normalizing}normalizing 
flows can be used to construct neural ODEs. To this end we define  
\begin{align}
\label{eq:nfode}
\begin{split}
\frac{dv}{dt}&=h(v;\theta),\\
v(0)&=z.
\end{split}
\end{align}
Noting that $v(t)$ depends on $z$ and on $\theta$ 
we set $g_t(z;\theta)=v(t)$ and $g(z;\theta)=g_1(z;\theta)=v(1).$
An advantage of the continuum perspective is that the inverse 
of map $g$ is readily computed as follows. Define the equation
\begin{align*}
\frac{dw}{dt}&=-h(w;\theta),\\
w(0)&=u.
\end{align*}
Then $g^{-1}_t(u;\theta)=w(t)$ and, in particular, $g^{-1}(u;\theta)=w(1).$
Thus, provided the regularity and structural properties of $h$ are such as to ensure existence and uniqueness of solutions on the unit interval, both forward and backward, the map $g$ defines an invertible transport.\index{transport!invertible}
We make this assumption in the remainder of this subsection.

In the following lemma we emphsize dependence of $v$ on $\theta.$

\begin{lemma}
Assume that in \eqref{eq:nfode} the initial condition satisfies
$z \sim \zeta$. Then 
\begin{equation}
\label{eq:cov3}
\log g_\sharp \zeta(u)   = \log \zeta(z) - \int_0^1 
\div h\bigl(v(s;\theta);\theta\bigr) \, ds.
\end{equation}
\end{lemma}

\begin{proof}
To establish this result, we work with the Euler approximation of the
continuum formulation, and pass to the limit of infinitesimally small
time-step. This may be achieved by considering the discrete 
\index{flow!normalizing}normalizing flow with 
$H(\cdot\,;\theta)=\Id+\Delta t h(\cdot\,;\theta)$; setting $J\Delta t=1$,
where $\Id$ is the identity map; then taking $\Delta t \to 0$ will deliver the desired limit.
Note that the discrete and continuum pictures are connected by
$v_j(\theta) \approx v(j\Delta t;\theta).$ We do not provide a detailed $\epsilon-\delta$
characterization of the limit $\Delta t \to 0$, instead proceeding purely formally.

We first note that
$${\rm det} DH={\rm det}(I_d+\Delta t Dh)=1+\Delta t {\rm Tr} Dh
+{\mathcal O}(\Delta t^2).$$
Now note that ${\rm Tr}\, Dh=\div h.$ Thus,
substituting this expression for ${\rm det} DH$
into \eqref{eq:cov2}, expanding the logarithm in powers
of $\Delta t$, summing over $j$ and letting $\Delta t \to 0$ yields
the desired result.
\end{proof}

The continuum analog of \eqref{eq:OT999} is then 
\begin{equation}
    \label{eq:noopt}
\J(\theta)=
 -\bbE^{u \sim \Upsilon} \Bigl[\log \zeta(z)- \int_0^1 
\div h\bigl(v(s;\theta);\theta\bigr) \, ds \Bigr],
\end{equation}
where $z=g^{-1}(u;\theta)$ and $v(t;\theta)=g^{-1}_{1-t}(u;\theta).$ 
As before, up to an additive constant,
$\J(\theta)=\dkl(\Upsilon\|g_\sharp \zeta).$
Minimizing $\J$ over $\theta$, to obtain $\thetas$,
thus leads to $\gs(\cdot)=g(\cdot\,;\thetas)$ and
$(\gs)_\sharp \zeta \approx \Upsilon.$

\begin{remark}
\label{rem:nfna2}
Using the setting of Remark \ref{rem:nfna} leads
to a generalized form of neural ODE taking the form

\begin{align}
    \label{eq:nfodetimedep}
\begin{split}
\frac{dv}{dt}&=h\bigl(v,t;\theta(t)\bigr),\\
v(0)&=z.
\end{split}
\end{align}
Optimization is now over a function of time, $\theta(t)$, and
some form of regularization is needed to enforce continuity
of $\theta(\cdot)$. For example we may employ Tikhonov-Phillips 
regularization,\index{regularizer!Tikhonov-Phillips}
penalizing the $H^1([0,1];\R^p)$
norm of $\theta$, replacing 
\eqref{eq:noopt} by
\begin{equation}
    \label{eq:noopt2}
\J(\theta)=
 -\bbE^{u \sim \Upsilon} \Biggl[\log \zeta(z)- \int_0^1 
\div h\bigl(v(s;\theta),s;\theta(s)\big) \, ds \Biggr]+\lambda
\int_0^1 \Bigl( |\theta(s)|^2 +  \bigl|\frac{d\theta}{dt}(s)\bigr|^2 \Bigr) \, ds.
 \end{equation}
Notice that $z=v_0$ and $v(s)$ are functions of $u=v(1)$, 
defined by the backward evolution of the ODE \eqref{eq:nfodetimedep}. 
\end{remark}

\section{Score-Based\index{score} Approaches}\label{sec:score}
So far in this chapter we have overviewed a variety of techniques for generative modeling, 
all based around density estimation, with our focus being variants
on transport-based methods. An important shift of perspective, leading 
to alternative methodologies, follows from understanding the role
of the Langevin equation\index{Langevin equation} 
in generating samples from a measure with probability density $\Upsilon.$ 
This is the subject of Subsection \ref{ssec:langevin}. In the context of
Data Assumption \ref{assumption:unsupervised} use of the Langevin equation
to generate approximate samples from $\Upsilon$
requires learning the probability density function to be sampled,
or learning the \emph{score function:}\index{score function} the gradient of the logarithm of the 
probability density function. Subsections \ref{ssec:sm} and
\ref{ssec:dnsm} study two approaches to learning the score.\index{score} Throughout this section, we assume that $\Upsilon$ is positive and sufficiently smooth for the expressions and integration-by-parts arguments below to be well defined.

\subsection{Langevin Equation} \label{ssec:langevin}

We start by defining the Langevin equation\index{Langevin equation}
\begin{equation}
\label{eq:lan}
    du(t)=\nabla \log \Upsilon\bigl(u(t)\bigr) dt+\sqrt{2} dW(t), \quad u(0) \sim \zeta. 
\end{equation}
Under certain tail and smoothness assumptions on $\Upsilon$ this equation 
is ergodic\index{ergodic}: for suitable classes of test function $\varphi:\R^d \to \R,$ 
we have that, almost surely with respect to $u(0)$ and $W(\cdot)$ (assumed
independent of one another),
$$\lim_{T \to \infty} \frac{1}{T}\int_0^T \varphi\bigl(u(t)\bigr) \, dt = \bbE^{u \sim \Upsilon} \bigl[\varphi(u)\bigr].$$
Similar convergence results hold in law.
This statement of ergodicity may be viewed as defining a stochastic transport, 
over an infinite time horizon, of $\zeta$ into $\Upsilon;$ stochastic because it depends on
Brownian motion $W(\cdot)$ and infinite time horizon because $u(0) \sim \zeta$
whilst $u(T) \sim \Upsilon$ in the limit $T \rightarrow \infty$. Note that, because of ergodicity,
wide classes of $\zeta$ all get transported to the desired measure $\Upsilon.$

We wish to use simulation of the Langevin equation as the basis for drawing approximate samples
from $\Upsilon.$ However, we do not assume that we are given $\Upsilon$; we simply have
the data $U$ from Data Assumption \ref{assumption:unsupervised}. Thus we must learn
the \emph{score}\index{score function} $\nabla \log \Upsilon$
from data $U.$ One approach to deal with this issue is to employ density estimation from Section \ref{sec:den} to approximate $\Upsilon$ and then, provided the approximation is sufficiently smooth, use the Langevin approach to sample.
But since only the gradient of the logarithm of $\Upsilon$ enters the Langevin equation, we can instead try to directly learn this function from data. Doing so is the topic of the next two subsections.

\begin{remark}
Once the score\index{score function} is known, then discretization of the Langevin equation in
time, for example by the Euler--Maruyama\index{Euler--Maruyama} method, and integration to a large time,
provides a way of approximating samples from $\Upsilon.$ However, discretization
may cause ergodicity to be lost. One approach to overcome this is to view the
discretization as a proposal\index{Markov kernel!proposal} distribution for the
Metropolis--Hastings\index{Metropolis--Hastings} MCMC\index{Monte Carlo!Markov chain} 
methods introduced in Section \ref{sec:MCMC}.
\end{remark}

\subsection{Score-Matching} \label{ssec:sm}

The starting point is to identify a function $s_\theta \colon \R^d \rightarrow \R^d$  so that $s_{\thetas} \approx \nabla \log \Upsilon.$  We assume that $\theta$ is chosen from set $\Theta \subseteq \R^p$.  We work under
Data Assumption \ref{assumption:unsupervised} and hence are given only samples from $\Upsilon$: we do not assume access to the target density $\Upsilon$ itself. 
A natural objective results in the following regression problem:
\begin{subequations} \label{eq:score_matching_regression}
\begin{align}
  \F(\theta) &= \int_{\R^d} |s_\theta(u) - \nabla \log \Upsilon(u)|^2 \Upsilon(u) \, du,\\
  \thetas &\in \argmin_{\theta \in \Theta} \F(\theta).
\end{align}
\end{subequations}
Then $\ss:=s_{\thetas}$ is the approximate score\index{score} at the optimal $\theta.$

This approach looks attractive, but approximate evaluation 
of the objective $\F(\cdot)$, using only knowledge of the empirical\index{empirical} 
measure \eqref{eq:emp}, and directly employing 
\eqref{eq:score_matching_regression}, is not feasible in practice. 
The root cause is the presence of the true score  $\grad \log \Upsilon(u)$ 
in the objective. The reader will note that $\log \Upsilon(u)$ also entered the
objective $\F$ arising in density estimation (Section \ref{sec:den}).
However in that context the terms involving $\log \Upsilon(u)$ additively decouple
from those involving the unknown parameter $\theta.$ The same happens here, but an integration
by parts is necessary to make this apparent.

\begin{proposition} 
Assume that $K:=\int_{\R^d} |\nabla \log \Upsilon(u)|^2 \Upsilon(u) \, du < \infty$ and that, for every $\theta\in\Theta$, the integration-by-parts identity
\begin{equation}
\label{eq:score_matching_ibp_identity}
\int_{\R^d}
\langle s_\theta(u),\nabla\Upsilon(u)\rangle\,du
=
-\int_{\R^d}
\div s_\theta(u)\Upsilon(u)\,du
\end{equation}
holds. Then
\begin{subequations}
\begin{align*}
\F(\theta) & = \J(\theta) + K, \\
\J(\theta) &=  \int_{\R^d} \Bigl(|s_\theta(u)|^2 + 2\,\div  s_\theta(u)\Bigr) \Upsilon(u) \, du.
\end{align*}
\end{subequations}
Thus $\thetas$ defined by \eqref{eq:score_matching_regression} also satisfies
\begin{equation} \label{eq:score_matching_IBP}
\thetas \in \argmin_{\theta \in \Theta} \J(\theta).
\end{equation}
\end{proposition}

\begin{proof}
Expanding the Euclidean norm, the least-squares objective \eqref{eq:score_matching_regression} yields 
\begin{align*}
\F(\theta) 
&= \int_{\R^d} \Bigl(|s_\theta(u)|^2 - 2\bigl\langle s_\theta(u), \nabla \log \Upsilon(u)\bigr\rangle + |\nabla \log \Upsilon(u)|^2\Bigr)\Upsilon(u) \, du.
\end{align*}
Using
$\nabla\log\Upsilon(u)\Upsilon(u)=\nabla\Upsilon(u)$
and the integration-by-parts identity
\eqref{eq:score_matching_ibp_identity}, we obtain
$$\int_{\R^d} \bigl\langle s_\theta(u), \nabla \log \Upsilon(u)\bigr\rangle \Upsilon(u) \, du = 
\int_{\R^d}  \bigl\langle s_\theta(u), \nabla \Upsilon(u) \bigr\rangle \, du = -\int_{\R^d} 
\,\div s_\theta(u)\Upsilon(u) \, du.$$
The desired result follows.  
\end{proof}

Objective function $\J(\cdot)$ is readily approximated under Data Assumption \ref{assumption:unsupervised}, leading to 
$$\J^N(\theta) =  \int_{\R^d} \Bigl(|s_\theta(u)|^2 + 2\,\div  s_\theta(u)\Bigr) \Upsilon^N(u) \, du.$$
Actionable algorithms to approximate the score may then proceed by optimizing $\J^N(\cdot).$

\begin{remark} \label{rem:1313} Let $\Upsilon_\theta$ be a distribution whose score\index{score} is given by $s_\theta = \nabla \log \Upsilon_\theta$. Then, the objective $\F(\cdot)$ corresponds to the squared 
\emph{Fisher divergence}\index{divergence!Fisher} 
$$d_F(\Upsilon \|\Upsilon_\theta)^2 \coloneqq \int_{\R^d} | \nabla \log \Upsilon_\theta(u) - \nabla \log \Upsilon(u)|^2 \Upsilon(u) \, du.$$
\end{remark}

\begin{remark}
We note that, in contrast to \eqref{eq:score_matching_regression}, the objective 
\eqref{eq:score_matching_IBP} requires evaluation of 
the gradient of the approximate score\index{score!function} function $\nabla s_\theta$, in order to evaluate the divergence term. If $s_\theta$ is parameterized as the gradient of a scalar function, then this
gradient is a Hessian matrix of size $d \times d$ and can be challenging to compute and store in high dimensions. However, one may leverage techniques from randomized linear algebra to estimate the trace of this matrix.
The need to evaluate the Hessian can also be avoided by learning the score
not of $\Upsilon$ but of its convolution with a Gaussian.
\end{remark}

\subsection{Denoising Score-Matching} \label{ssec:dnsm}

To avoid the need to evaluate the Hessian we introduce the idea of
\emph{denoising score-matching}\index{score!-matching!denoising}. 
This approach estimates the score\index{score!function} function 
for an approximate distribution defined by the convolution of $\Upsilon$ 
with a kernel $p_\sigma \colon \R^d \times \R^d \rightarrow \R_{+}$ 
with bandwidth $\sigma > 0$. That is,
\begin{equation} \label{eq:smoothed_distribution}
\Upsilon_\sigma(u) \coloneqq \int_{\R^d} p_\sigma(u,w) \Upsilon(w) \, dw.    
\end{equation}
For example, $p_\sigma$ can be a Gaussian kernel\index{kernel!Gaussian} 
(recall the Definition \ref{d:mk} of kernel\index{kernel})
\begin{equation}
\label{eq:stave}
p_\sigma(u,w) \propto \exp\Bigl(-\frac{1}{2\sigma^2}|u-w|^2\Bigr).
\end{equation}
This choice is motivated by the facts that: (i) for bandwidth 
$\sigma \rightarrow 0$ we obtain that $p_\sigma(\cdot\,; w)$ converges 
to the Dirac\index{Dirac measure} measure $\delta_w$, if appropriately normalized, 
and so $\Upsilon_\sigma(u)$ converges to $\Upsilon(u)$, the original 
distribution; (ii) on the other hand, larger bandwidths smooth 
larger-scale features of $\Upsilon$. The following proposition shows 
that score-matching\index{score!-matching} for $\Upsilon_\sigma$ 
can be performed without knowing the score\index{score!function} 
of $\Upsilon$ directly. To this end we define
\begin{subequations} \label{eq:Fsig2}
\begin{align}
\F_\sigma(\theta) &= \int_{\R^d} |s_\theta(u) - \nabla \log \Upsilon_\sigma(u)|^2 \Upsilon_\sigma(u) \, du,\\
\thetas &\in \argmin_{\theta \in \Theta} \F_\sigma(\theta).
\end{align}
\end{subequations}

\begin{proposition} \label{prop:denoising_score_matching} Assume that 
\begin{equation*}\int_{\R^d} |\nabla \log \Upsilon_\sigma(u)|^2 \Upsilon_\sigma(u) \, du + \int_{\R^d} \int_{\R^d} |\nabla_u \log p_\sigma(u,w)|^2 p_\sigma(u,w) \Upsilon({w}) \, dw\,du < \infty.
\end{equation*}
Then $\theta$ solving \eqref{eq:Fsig2} also solves the minimization problem
\begin{equation}
    \label{eq:definedin}
    \thetas \in \argmin_{\theta \in \Theta} \int_{\R^d} \int_{\R^d} |s_\theta(u) - \nabla_u \log p_\sigma(u,w)|^2 p_\sigma(u,w)\Upsilon(w) \, du\,dw.
\end{equation}
\end{proposition}

\begin{proof}
The score\index{score!function} of the smoothed distribution $\Upsilon_\sigma$ is given by
\begin{align}
\nabla \log \Upsilon_\sigma(u) = \frac{\nabla \Upsilon_\sigma(u)}{\Upsilon_\sigma(u)} &= \frac{1}{\Upsilon_\sigma(u)}\int_{\R^d} \nabla_u p_\sigma(u,w) \Upsilon(w) \, dw \notag\\
&= \frac{1}{\Upsilon_\sigma(u)}\int_{\R^d} \nabla_u \bigl(\log p_\sigma(u,w)\bigr) p_\sigma(u,w) \Upsilon(w) \, dw. \label{eq:stave2} 
\end{align}
Thus, the inner product of $s_\theta$ and the score\index{score!function} of the smoothed distribution is given by
$$\int_{\R^d} \bigl \langle s_\theta(u), \nabla \log \Upsilon_\sigma(u) \bigr \rangle \Upsilon_\sigma(u) \, du = \int_{\R^d} \int_{\R^d} \bigl \langle s_\theta(u), \nabla_u \log p_\sigma(u,w) \bigr \rangle p_\sigma(u,w) \Upsilon(w) \, dw\,du.$$
Furthermore
$$ \int_{\R^d} |s_\theta(u)|^2 \Upsilon_\sigma(u) \, du= \int_{\R^d} \int_{\R^d} |s_\theta(u)|^2  p_\sigma(u,w) \Upsilon(w) \, dw\,du$$
by definition of $\Upsilon_\sigma.$
Substituting the two preceding expressions into the objective $\F_\sigma(\theta)$
gives
\begin{align*}
    \F_\sigma(\theta) &= \int_{\R^d} \int_{\R^d} |s_\theta(u) - \nabla_u \log p_\sigma(u,w)|^2 p_\sigma(u,w)\Upsilon(w) \, du\,dw+K,\\
    K &= \int_{\R^d} |\nabla \log\Upsilon_\sigma(u)|^2 \Upsilon_\sigma(u) \, du  - \int_{\R^d} \int_{\R^d}|\nabla_u \log p_\sigma(u,w)|^2 p_\sigma(u,w)\Upsilon(w) \, du\,dw.
\end{align*}
Noting that $K$ is independent of $\theta$ and finite by assumption, we deduce that the minimizers of $\F_\sigma(\cdot)$ coincide with those defined in \eqref{eq:definedin}.
\end{proof}

We now restrict our attention to the specific case where $p_\sigma(u,w)$
is the conditional density of $u$ given $w$
defined by \eqref{eq:stave}.
The score\index{score!function} of smoothed distribution $\Upsilon_\sigma$ that is computed by Proposition~\ref{prop:denoising_score_matching} can then 
be expressed in terms of a conditional expectation of the underlying random variables. The next lemma presents this result, which is known as \emph{Tweedie's formula}\index{Tweedie's formula}. 

\begin{lemma} \label{lem:tweedie} Let $u = w + \sigma \eta$ where $w \sim \Upsilon$ and $\eta \sim \mathcal{N}(0,I_d)$
and let $p_\sigma(\cdot\,;w)$ denote the Gaussian probability density function\index{probability density function!Gaussian} 
defined by kernel\index{kernel}\eqref{eq:stave}. 
Define $\pi^w$ to be the posterior density\footnote{Note that this is the correct normalization 
by~\eqref{eq:smoothed_distribution}.}
$$\pi^w(w|u) = \frac{1}{\Upsilon_\sigma(u)} p_\sigma(u,w)\Upsilon(w).$$
Then the score\index{score!function} $\nabla \log \Upsilon_\sigma(u)$ satisfies the identity 
\begin{equation}
\label{eq:othertweedie}
\mathbb{E}^{w \sim \pi^w}[w] = u + \sigma^2 \nabla \log \Upsilon_\sigma(u).
\end{equation}
\end{lemma}

\begin{proof}
First note that the score\index{score!function} of the Gaussian probability density function\index{probability density function!Gaussian}  is
$$\nabla_u \log p_\sigma(u,w) = -\frac{(u - w)}{\sigma^2}.$$
Using this identity in \eqref{eq:stave2}
we can write the score\index{score!function} of the smoothed distribution as 
\begin{align*}
\nabla \log \Upsilon_\sigma(u) &= \frac{1}{\Upsilon_\sigma(u)}\int_{\R^d} \nabla_u \bigl( \log p_\sigma(u,w)\bigr) p_\sigma(u,w) \Upsilon(w) \, dw \\
&= \int_{\R^d} \frac{-(u - w)}{\sigma^2} \frac{p_\sigma(u,w) \Upsilon(w)}{\Upsilon_\sigma(u)} \, dw \\
&= \frac{1}{\sigma^2}\mathbb{E}^{w \sim \pi^w}[w - u],
\end{align*}
and the desired result follows.
\end{proof}

Tweedie's formula\index{Tweedie's formula} shows that computing a score\index{score!function} function is related to an inverse problem. It may be rearranged
to give:
$$\nabla \log \Upsilon_\sigma(u)=\frac{1}{\sigma^2}\Bigl(\mathbb{E}^{w \sim \pi^w}[w] - u\Bigr).$$
Given a noisy realization $u$ of the random variable $w,$ the score\index{score!function} is given in terms of the conditional expectation $\mathbb{E}[w|u]$, which is the posterior mean for the unknown $w$. Notice that $\mathbb{E}[w|u]$ is the function of $u$ that minimizes $\mathbb{E} | w - \phi(u) |^2$ over measurable maps $\phi: \R^d \to \R^d.$ Thus, to learn $D\log \Upsilon_\sigma(u)$ we can seek parameterized map $\phi_\theta: \R^d \times \Theta \to \R^d$ that minimizes $\mathbb{E} | w - \phi_\theta(u) |^2.$ This problem can be empirically\index{empirical} approximated without having access to the density $\Upsilon$. Specifically, given only data $\{w^{(n)}\}_{n=1}^N $ independently sampled from $\Upsilon,$ we set $u^{(n)}:= w^{(n)} + \sigma \eta^{(n)},$ where $\eta^{(n)} \sim \mathcal{N}(0,I_d)$ i.i.d. We then set $\theta^\star \in \argmin_\theta \sum_{n=1}^N | w^{(n)} - \phi_\theta (u^{(n)}) |^2,$ define $\phi^\star = \phi_{\thetas}$ and approximate 
$D\log \Upsilon_\sigma(u) \approx \frac{1}{\sigma^2} \bigl(\phi^\star(u) - u\bigr)$.

\section{Autoencoders\index{autoencoder}}\label{sec:auto}
Autoencoders are primarily used as a technique for uncovering
latent low-dimensional structure in high-dimensional data. We introduce
them here and then demonstrate in the next section that a natural
probabilistic relaxation of the idea leads to generative models 
termed \index{autoencoder!variational}variational autoencoders.

\index{transport}Transport ideas may be used to (approximately) represent a complicated 
probability measure, perhaps only known through samples, as the 
pushforward\index{pushforward} of a simpler measure. In this section 
we go a step further, by seeking the simpler measure in a 
\index{latent space}latent space of lower dimension than 
that of the ambient space where the data lives. 
\index{autoencoder}Autoencoders are one natural approach to such problems.

The basic idea of \index{autoencoder}autoencoders is to find an approximate factorization
of the identity which is accurate on the support of density $\Upsilon$; furthermore we
need to do this using only the empirical\index{empirical} approximation of $\Upsilon$ by $\mn$. 
Let $f: \R^d \times \Theta_f \to \R^{\dz}$
and  $g: \R^{\dz} \times \Theta_g \to \R^d$ where $\Theta_f \subseteq
\R^{p_f}$ and $\Theta_g \subseteq \R^{p_g}.$ Function $f$ is referred to as the \emph{encoder} and $g$ as the \emph{decoder}. Recalling that $\Id$ denotes the 
identity mapping on $\R^d$,  define\footnote{The notation
$\hm$ and $\hm^N$ was introduced in the previous chapter in the
context of real-valued functions; see equation \eqref{eq:norm} and the material following it. 
This notation readily generalizes to $\R^d$-valued functions}. 
$$\J(\theta_f,\theta_g)= 
\Bigl\|\Id-g\bigl(f(\cdot\,;\theta_f);\theta_g\bigr)\Bigr\|_{\hm}^2:=
\mathbb{E}^{u \sim \Upsilon}\bigl|u-g\bigl(f(u;\theta_f);\theta_g\bigr)\bigr|^2.$$ 
Now consider the following optimization problem: 
\begin{align} 
\label{eq:erm99} 
(\thetas_f,\thetas_g)\in \argmin_{(\theta_f,\theta_g) \in \Theta_f \times
\Theta_g} \J(\theta_f,\theta_g). 
\end{align}
We then define $\fs(u)=f(u;\thetas_f)$ and $\gs(z)=g(z;\thetas_g).$
Roughly speaking, and dropping the dependence on parameters for
expository purposes, we are seeking functions $f$ and $g$ 
such that $g\bigl(f(u)\bigr) \approx u.$ In practice the
optimization is implemented empirically\index{empirical}, and we minimize
$$\J^N(\theta_f,\theta_g)= \Bigl\|\Id-g\bigl(f(\cdot\,;\theta_f);\theta_g\bigr)\Bigr\|_{\hm^N}^2
=\frac{1}{N}\sum_{n=1}^N\bigl|u^{(n)}-g\bigl(f(u^{(n)};\theta_f);\theta_g\bigr)\bigr|^2.$$ 

\begin{remark}
This approach reduces the \index{autoencoder}autoencoder to a particular form of supervised learning\index{supervised learning} in which the inputs and outputs are equal, so we seek to
approximate the identity as composition of $g$ with $f$: $g \circ f \approx \Id.$ 
We impose a specific structure on the class of approximating functions; for example each of $f$ and $g$ can be a \index{neural network}neural network, 
as defined in Section \ref{sec:NN}, generalized to vector-valued output.
We refer to $\R^{\dz}$ as the {\em latent space}\index{latent space}
and note that a primary application of the methodology 
is to identify \index{latent space}latent spaces of dimension
$\dz$ which is much less than the dimension $d$ of the data space. The
approximate factorization of the identity found by composing $g$ with $f$
provides a way of moving between the data space and the \index{latent space}latent space.
\end{remark}

\begin{example}
We demonstrate that \emph{principal component analysis}\index{principal component analysis},
often referred to simply as PCA\index{PCA}, may be viewed
as a form of autoencoder.
After centering
the data, we may assume that
$\mathbb{E}^{u\sim\Upsilon}u=0.$
Define the covariance matrix
$$C=\mathbb{E}^{u\sim\Upsilon}[u\otimes u].$$
Then consider the eigenpairs $(\varphi^{(i)},\lambda^{(i)}) \in
\R^d \times \R^+$ solving
\begin{align*}
    C\varphi^{(i)}&=\lambda^{(i)} \varphi^{(i)},\\
    |\varphi^{(i)}|&=1.
\end{align*}
The eigenvalues are non-negative and we assume them to be decreasingly ordered  
with respect to $i.$ Moreover, the eigenvectors can be chosen to be orthonormal.

For given $\dz \le d,$ we now define the maps
\begin{align*}
   f(u) &=\Bigl(\langle u, \varphi^{(1)}\rangle, \ldots, \langle u, \varphi^{(d_z)}\rangle\Bigr),\\
   g(z) &= \sum_{j=1}^{d_z} z_j \varphi^{(j)}.
\end{align*}
Then, their composition gives us
\begin{align*}
   g\bigl(f(u)\bigr) &= \sum_{j=1}^{d_z} \langle u, \varphi^{(j)}\rangle \varphi^{(j)}\\
   &=\Bigl(\sum_{j=1}^{d_z} \varphi^{(j)}\otimes \varphi^{(j)}\Bigr)u.
\end{align*}
Thus, the PCA method may be viewed as truncating the representation
of the identity defined by the 
spectral theorem\index{spectral!theorem} for positive,
symmetric matrices. The specific basis used is defined by
$\Upsilon$ but it is approximated in practice knowing only $\mn$, after centering the data by their
empirical\index{empirical} mean and computing their empirical covariance.
\end{example}

\section{Variational Autoencoders \index{autoencoder!variational}}\label{sec:vauto}
In the preceding section, note that $\zeta=(\fs)_\sharp \Upsilon$ 
gives the (approximate) distribution 
in the \index{latent space}latent space, but that we do not specify $\zeta$.
Furthermore the  mapping
$g$ (approximately) solves a \index{transport!measure}measure transport problem from $\zeta$ to $\Upsilon$, 
but the choice of measure $\zeta$ is not within our control. 
It is natural to ask whether the methodology can be 
generalized to settings in which it is desirable
to impose a specified distribution $\zeta$. This leads to the topic
of {\em variational autoencoders},\index{autoencoder!variational}
which we now outline.

Let $\pi$ be a \index{coupling}coupling of $\Upsilon$ and $\zeta$ 
and note that the following two identities hold:
\begin{align*}
\pi(u,z)&=\bbP(u|z)\zeta(z),\\
\pi(u,z)&=\bbP(z|u)\Upsilon(u).
\end{align*}
The idea of the \index{autoencoder!variational}variational autoencoder 
is to approximate the two conditional densities $\bbP(u|z), \bbP(z|u)$, 
appearing in these identities, by parameterized families; and then
to align the two different expressions for the coupling. To be explicit 
we assume that
\begin{align*}
u|z & \sim \cN\bigl(g(z;\theta_g),\sigma_g^2 I\bigr),\\
z|u & \sim \cN\bigl(f(u;\theta_f),\sigma_f^2 I\bigr). 
\end{align*}
Here the domains and ranges of $f,g$ are as specified in Section \ref{sec:auto}. 
The assumptions on the conditionals $u|z$ and $z|u$ thus constitute a relaxation of the setting 
for \index{autoencoder}autoencoders.
These two identities yield $\bbP(u|z;\theta_g) \approx \bbP(u|z)$ and
$\bbP(z|u;\theta_f) \approx \bbP(z|u).$
Then parameter choices to define $f$ and $g$ are made to ensure that the two   
resulting approximate expressions for $\pi(u,z)$ are close.
Invoking these \index{Gaussian!approximation}Gaussian approximations for the conditionals we obtain
the following two approximations for the \index{coupling}coupling density:
\begin{align*}
\pi_g(u,z)&=\frac{1}{Z_g}\exp\Bigl(-\frac{1}{2\sigma_g^2}|u-g(z;\theta_g)|^2\Bigr) \zeta(z),\\
\pi_f(u,z)&=\frac{1}{Z_f}\exp\Bigl(-\frac{1}{2\sigma_f^2}|z-f(u;\theta_f)|^2\Bigr) \Upsilon(u).
\end{align*}
A common choice is to assume that $\zeta$ is the density of
a standard unit \index{Gaussian}Gaussian; we make this assumption throughout what follows. We will only discuss learning of the parameters defining $(f,g)$,
but $(\sigma_g,\sigma_f)$ could also be learned.

We ask that $\pi_f$ and $\pi_g$ are close in 
\index{divergence!Kullback--Leibler} KL divergence, 
with $\pi_f$ in the first argument.
Then, using the \index{Gaussian}Gaussian assumptions 
on the conditionals and on $\zeta$,
leads to insightful explicit calculations. 
To see this, we first note that 
\begin{align} \label{eq:KLdivergence_VAE_joint}
\dkl(\pi_f\|\pi_g) = \bbE^{u \sim \Upsilon} \biggl[\bbE^{z \sim \bbP(z|u;\theta_f)}
\Bigl[ \frac{1}{2\sigma_g^2}|u-g(z;\theta_g)|^2 \Bigr] +
\dkl\bigl(\bbP(\cdot|u;\theta_f)\| \zeta \bigr)\biggr]+{\mathsf{const}},
\end{align}
where the constant is independent of the parameters we wish to learn
and where $\bbP(z|u)$ and $\zeta(z)$ are given by their assumed \index{Gaussian}Gaussian structure.
Using the expression in Example \ref{ex:klg} for 
the \index{divergence!Kullback--Leibler}KL
divergence between two \index{Gaussian}Gaussians, and dropping the terms that
depend only on the (assumed known) variances, we obtain
\begin{align*}
\dkl(\pi_f\|\pi_g) = \bbE^{u \sim \Upsilon} \biggl[\bbE^{z \sim \cN\bigl(f(u;\theta_f),\sigma_f^2 I\bigr)} \Bigl[\frac{1}{2\sigma_g^2}\big|u-g(z;\theta_g)\big|^2 \Bigr]+
\frac{1}{2}|f(u;\theta_f)|^2\biggr]+{\mathsf{const}}.
\end{align*}

We may now use the \emph{reparameterization trick}\index{reparameterization trick} 
(see Subsection~\ref{ssec:reparameterization})
to rewrite the desired objective function.
Noting that $z \sim \cN\bigl(f(u;\theta_f),\sigma_f^2 I\bigr)$ is
the same in law as $f(u;\theta_f)+\sigma_f \xi$, where $\xi \sim
\cN(0,I),$ we may write the desired \index{divergence}divergence as
\begin{align*}
\dkl(\pi_f\|\pi_g) = 
\J(\theta_f,\theta_g)+\mathsf{const},
\end{align*} 
where
\begin{align*}
\J(\theta_f,\theta_g) = \bbE^{u \sim \Upsilon} \biggl[\bbE^{\xi \sim \cN(0,I)} \Bigl[\frac{1}{2\sigma_g^2}\Big|u-g\bigl(f(u;\theta_f)+\sigma_f \xi;\theta_g\big)\Big|^2 \Bigr]+
\frac{1}{2}|f(u;\theta_f)|^2\biggr]
\end{align*}
and the constant term is independent of $(\theta_f,\theta_g).$

This constitutes a regularized version of the standard \index{autoencoder}autoencoder;
in particular the preceding expression for the \index{divergence}divergence clearly
regularizes the basic concept that $g\bigl(f(u)\bigr) \approx u$
under density $\Upsilon.$

\begin{remark} \label{rem:319}
In practice this is implemented with expectation 
over $\Upsilon$ approximated empirically\index{empirical} by $\mn.$
This leaves an optimization problem in which the \index{objective}objective
is defined via expectation over $\xi.$ Stochastic
gradient descent\index{gradient descent!stochastic} from
Chapter \ref{ch:optimization} 
may be used to tackle this problem. After training, approximate samples 
from $\Upsilon$ may be obtained by sampling $z\sim \zeta$ and then 
sampling $u|z \sim \Nc \bigl(g(z; \thetas_g), \sigma_g^2 I\big)$. 

Notice that in \eqref{eq:KLdivergence_VAE_joint} we employ the expectation
$$\bbE^{u \sim \Upsilon} \bbE^{z \sim \bbP(z|u;\theta_f)}.$$
Using empiricalization this is approximated by
$$\bbE^{u \sim \mn} \bbE^{z \sim \bbP(z|u;\theta_f)}=\frac{1}{N}
\sum_{n=1}^N \bbE^{z \sim \bbP(z|u^{(n)};\theta_f)},$$
where the $\{u^{(n)}\}_{n=1}^N$ are the i.i.d. samples 
from $\Upsilon$ that comprise $\mn.$
The resulting expression involves the $N$ probability measures $\bbP(z|u^{(n)};\theta_f)$. These are
referred to as $N$ posterior \index{posterior} distributions in the original
literature on this subject as they are (approximate) posteriors on the latent $z$, one for each data point $u^{(n)}.$
\end{remark}

\section{Generative Adversarial Networks\index{generative adversarial network}}\label{sec:gan}

The \index{autoencoder!variational}variational autoencoder trains 
a probabilistic model for $u \in \R^d$ by learning $\bbP(u|z)$ 
and specifying $\bbP(z).$
The \index{generative adversarial network}generative adversarial network is
another method for training such a probabilistic model, which
also allows for low-dimensional \index{latent space}latent space. 
To be concrete we assume that it has the same structural 
form, namely
\begin{subequations}
\label{eq:construct}
\begin{align}
u|z & \sim \cN\bigl(g(z;\theta_g),\sigma_g^2 I\bigr),\\
z & \sim \cN\bigl(0,I\bigr).
\end{align}
\end{subequations}
Once again the domain and range of
$g$ are as specified in Section \ref{sec:auto}.

With the probabilistic construction \eqref{eq:construct},
samples from $u$ are created from samples from $z$. We
write the resulting density for $u$, under this generative model,
as $\Upsilon_g$; here $g$ connotes the fact that the model is generative. 
It is desired that samples from $\Upsilon_g$ are similar to those
specified by the data, which is drawn from density $\Upsilon$.
The \index{generative adversarial network}generative adversarial network starts by defining a
\emph{discriminator} $\mathsf{d}: \R^d \times \Theta_{\mathsf{d}} \to [0,1]$,
to be thought of as taking values which are probabilities.
Here $\Theta_{\mathsf{d}} \subset \bbR^{p_{\mathsf{d}}}.$ 
It is instructive to think of $\mathsf{d}(u;\theta_{\mathsf{d}})$ as being the
probability that $u$ is drawn from the density $\Upsilon$, and
$1- \mathsf{d} (u;\theta_{ \mathsf{d} })$ as the probability of $u$ being drawn
from the \index{generative model}generative model $\Upsilon_g.$
We then define $v: \Theta_g \times \Theta_{ \mathsf{d} } \to \bbR$ by
$$v(\theta_g,\theta_{ \mathsf{d} })=\bbE^{u \sim \Upsilon} \Bigl[\log  \mathsf{d}  (u; \theta_{ \mathsf{d} }) \Bigr] +
\bbE^{u \sim \Upsilon_g} \Bigl[\log \bigl(1-  \mathsf{d}  (u; \theta_{ \mathsf{d} })\bigr)\Bigr].$$
The optimal parameter values are defined by 
\begin{align*}
\widetilde{\theta}_{ \mathsf{d} }(\theta_g) & \in \argmax_{\theta_{ \mathsf{d} }\in\Theta_\mathsf{d}} v(\theta_g,\theta_{ \mathsf{d} }),\\
\thetas_g & \in \argmin_{\theta_g\in\Theta_g} v\bigl(\theta_g, \widetilde{\theta}_{ \mathsf{d} }(\theta_g)\bigr),\\
\thetas_{ \mathsf{d}} & = \widetilde{\theta}_{ \mathsf{d} }(\thetas_g).
\end{align*}
In the maximization step, the parameters of the discriminator are chosen
to maximize $v$ -- the discriminator acts adversarially to try and find data
under $\Upsilon_g$ which does not look like data under $\Upsilon$. In the minimization
step the generator acts to reduce $v$ to try and make the two data sources
look similar. Ideally, through an iterative process, a solution is found in
which $ \mathsf{d}  (\cdot\,;\thetas_{ \mathsf{d} }) \equiv \frac12$, so that the data and generated
data are indistinguishable. The value of $\thetas_g$ defines the generative
model once this indistinguishable state has been reached.

\section{Bibliography}\label{sec:IPMLBib}

Density estimation\index{density estimation} is covered in numerous texts; we refer the reader to
the book \cite{silverman2018density} and the literature cited there. For a foundational reference concerning the use of normalizing 
flows\index{flow!normalizing}
in machine learning, see \cite{rezende2015variational}; more recent
applications may be found in \cite{winkler2019learning}, \cite{gabrie2021efficient}.
The neural ODE perspective was popularized in the paper
\cite{chen2018neural}; earlier papers \cite{haber2017stable, weinan2017proposal} anticipated the idea but did not develop it into a practical tool in the same way as \cite{chen2018neural}. 
The subject of \index{transport!optimal}optimal transport is overviewed
and systematically
developed in \cite{villani2009optimal}; computational
methodology is overviewed and developed in
\cite{peyre2017computational}.
Triangular transport maps are overviewed as a method for \index{sampling}sampling in 
\cite{marzouk2016sampling}.

Score-based\index{score} methods have become popular in the context of generative diffusion models\index{diffusion models}~\cite{song2020score}. 
The integration by parts method and denoising score-matching\index{score!-matching!denoising} approaches were first proposed in~\cite{hyvarinen2005estimation} and~\cite{vincent2011connection}, respectively. These approximate scores\index{score} have been used for sampling using Langevin dynamics; see~\cite{lee2023convergence, block2020generative} for sampling guarantees. They have also been used in other unsupervised learning tasks, like learning probabilistic graphical models. \index{autoencoder}Autoencoders
have a long history; see 
\cite{goodfellow2016deep,hinton1993autoencoders,schmidhuber2015deep} 
and the citations therein. 
Variational autoencoders were introduced concurrently in  
\cite{kingma2013auto} and \cite{rezende2014stochastic}.
For overviews of variational autoencoders see
\index{autoencoder!variational}
\cite{zhang2018advances,kingma2019introduction}.
The idea of conducting \index{unsupervised learning}unsupervised learning with 
\index{generative adversarial network} GANs was introduced
in \cite{goodfellow2014generative}. 

Although not covered here, because our focus is on generative modeling,
\index{clustering}clustering is
a widely used methodology in unsupervised learning; it is overviewed in \cite{von2007tutorial}. The paper \cite{ng2001spectral} describes an 
underlying mathematical framework, based on eigenvalue perturbation theory. Understanding large data limits of spectral clustering methods is a subject
developed in \cite{von2008consistency,hoffmann2022spectral}. The use of graph Laplacians, which underpin spectral \index{clustering}clustering,
in the solution of \index{inverse problem}inverse problems, is undertaken in
\cite{garcia2018continuum,dunlop2020large,trillos2017consistency,harlim2019kernel,sanz2020spde,harlim2022graph}.

\chapter{\Large{\sffamily{Time-Series Forecasting}}}\label{ch:ts_forecasting}

This chapter is devoted to the study of time-series forecasting,
with an emphasis on data-driven approaches.
In most other chapters we have used $n$ to index data and $j$ to
index (discrete) time in (possibly stochastic) dynamical systems.
\index{dynamical system}\index{dynamical system!stochastic}\index{dynamical system!discrete time} In this chapter the two are conflated as the data comes from
a time series as follows:

\begin{dataassumption}\index{Data Assumption} \label{da:forc}
Let $\mathsf{N}=\{0,1,\ldots,N\}.$
    We are given a time series $\Vd=\{\vd_n\}_{n \in \mathsf{N}}$, 
    of elements $\vd_n$ in $\R^{\du}.$
\end{dataassumption}

For this reason the (possibly stochastic) dynamical 
systems\index{dynamical system!stochastic} in this
chapter will be indexed by $n$, not $j.$ 
Our focus in this chapter is to learn models that are consistent
with the data in Data Assumption \ref{da:forc} and which may be
used to predict future outcomes from points not in the data set.

Section \ref{sec:ERG} contains background on ergodic\index{ergodic} 
and stationary\index{stationary} processes.
In Sections \ref{sec:AF} and \ref{sec:RS}
we study deterministic models to fit the time-series data,
and in Section \ref{sec:autoregressive} we study random models.
Section \ref{sec:AF} is devoted to the study of analog forecasting methods;
these methods seek to predict
from a given point by looking at what happened, in the data set,
starting at nearby point(s).
In Section \ref{sec:RS} we look at fitting Markovian deterministic
models to the data; we also study non-Markovian deterministic models, 
using a recurrent structure to capture memory dependence. 
In Section \ref{sec:autoregressive} we introduce
autoregressive models\index{autoregressive!model}; we study linear autoregressive models\index{autoregressive!linear model},
modelling the data as a Gaussian process,
and non-Gaussian autoregressive models, using transport\index{transport}\index{autoregressive!transport} ideas.
We conclude in Section \ref{sec:TSB} with bibliographic remarks.

\section{Ergodic and Stationary Processes}\index{ergodic}
\label{sec:ERG}

Suppose that $\{\vd_n\}_{n \in \bbZ^+}$ is generated by the map $\psi: \R^d \to \R^d$ and the iteration, assumed to hold for
all $n \in \bbZ^+$,
\begin{equation} 
\label{eq:modfit0}
\vd_{n+1} = \psi(\vd_n);
\end{equation}
complete specification of $\{\vd_n\}_{n \in \bbZ^+}$ then follows once $\vd_0$ is given. 
We extend $\psi$ to act on Borel subsets of $\R^d$ and denote by $\psi^{(n)}$ its $n$-fold
composition. We assume that there exists a set $\cA \subseteq \R^d$ which
supports an invariant measure $\mu$; that is, a measure such that if $v \sim \mu$ then $\psi(v) \sim \mu.$
In particular, if $\vd_0 \sim \mu$ then $\vd_n \sim \mu$ for all $n \in \bbZ^+.$
In terms of pushforward\index{pushforward} we have $\psi_\sharp \mu=\mu.$

\begin{example}
Assume that $\psi$ is continuous and has the property that there is a compact \emph{absorbing
set} $\cB \subset \R^d$: for any compact subset $B \subset \R^d$ there is
$M=M(B)$ such that $\psi^{(m)}(B) \subseteq \cB$ for all $m \ge M$. 
Then $\psi$ has a \emph{global attractor}\index{attractor!global}: a unique compact invariant set $\cA \subset \Ru$ that attracts all bounded sets $B$: the Hausdorff distance\index{distance!Hausdorff}  $h(\psi^{(m)}(B),\cA) \to 0$ as $m \to \infty.$ 
The global attractor $\cA$ is defined by
\begin{equation*}
\cA
:=
\bigcap_{M\geq 0}
\overline{\bigcup_{m\geq M}\psi^{(m)}(\cB)}.
\end{equation*}
If $\psi(\cB) \subseteq \cB$ then
the global attractor may be characterized more directly by the formula
$$\cA=\bigcap_{m \in \mathbb{N}} \psi^{(m)}(\cB).$$
The global attractor $\cA$ supports an invariant measure $\mu.$
\end{example}

We employ the following informal definition of an ergodic\index{ergodic} dynamical system.
The process $\{\vd_n\}$, in $\R^{\du},$ is 
\emph{ergodic}\index{ergodic} with respect to invariant measure $\mu$ if, 
for a sufficiently wide class of test functions $\varphi$,
\begin{equation}
\label{eq:erg}    
\lim_{M \to \infty} \frac{1}{M} \sum_{m=0}^{M-1} \varphi(\vd_m)=
\int_{\R^{\du}} \varphi(v) \, \mu(dv),
\end{equation}
with the convergence being almost sure with respect to $\vd_0$ drawn
from $\mu.$
Since the process is deterministic, it is possible that measure 
$\mu$ will not have density with respect to Lebesgue measure.
This is why, in contrast to much of the remainder of the notes,
we have not used a probability density function in this
definition.

\begin{remark}\label{rem:erg-stoc}
    Ergodicity for a \emph{stochastic} rather than deterministic Markov process $\{\vd_n\}$ can be defined similarly by requiring that there exists measure $\mu$ on $\R^d$ such that, for a sufficiently wide class of test functions $\varphi,$ equation \eqref{eq:erg} holds; the almost sure property is then with respect to both the initial condition and
    the noise process driving the stochastic evolution. For both deterministic and stochastic settings, \eqref{eq:erg} encapsulates the idea that averaging over time is equivalent to averaging over space with respect to $\mu$: the expected value of $\varphi$ with respect to measure $\mu$ on $\R^d$ can be computed by averaging the values of $\varphi$ along a trajectory of the process.  This idea also underpins 
    Markov chain Monte Carlo\index{Markov chain Monte Carlo} algorithms; this methodology is
    described in Section \ref{sec:MCMC}.
\end{remark}

The following definition applies to both deterministic dynamical systems, randomly initialized, and to
random dynamical systems:

\begin{definition} \label{def:stat}
A discrete-time dynamical system\index{dynamical system!discrete time}
giving rise to data $\{\vd_n\}_{n \in \mathbb{Z}^+}$ is said to be 
\emph{strictly stationary}\index{stationary!strictly} if,
for all $p \in \mathbb{Z}^+$, the joint distribution of
$(\vd_n, \ldots, \vd_{n+p})$ is the same for all  $n \in \mathbb{Z}^+$.
It is said to be \emph{wide-sense stationary}\index{stationary!wide-sense} if, for all $p \in \mathbb{Z}^+$, the first and second moments of
$(\vd_n, \ldots, \vd_{n+p})$ are the same for all  $n \in \mathbb{Z}^+$.
\end{definition}

For stationary processes it is natural to
make the following definition.

\begin{definition} \label{def:cf}
For given $m \in \mathbb{Z}^+$, the \emph{correlation measure}\index{correlation measure}
$\pi_m \in \mathcal{P}(\R^d \times \R^d)$ of a stationary process $\{\vd_n\}_{n \in \mathbb{Z}^+}$ is the distribution of the pair $(\vd_n, \vd_{n+m}).$ The \emph{two-point correlation function}\index{correlation function} 
of a wide-sense stationary process $\{\vd_n\}_{n \in \mathbb{Z}^+}$ is  function $c: \mathbb{Z}^+ \to \R^{d \times d}$ given by $c(m):= \bbE[\vd_n (\vd_{n+m})^\top].$ 
\end{definition}

\begin{remark}
Note that the two-point correlation function\index{correlation function} is sometimes defined as the scalar-valued
function found as the trace of $c(\cdot).$ The matrix-valued definition we adopt here is, however, more widely useful
and implies the scalar-valued definition by applying the particular linear functional defined by the trace operation.
\end{remark}

\begin{example}
In the deterministic setting of
\eqref{eq:modfit0} randomness can only enter through the initial condition.
If $v_0 \sim \mu$ then the correlation measure $\pi_m(\cdot,\cdot)$ is simply the
pushforward of $\mu$ under the map $u \in \R^d \mapsto 
\bigl(u,\psi^{(m)}(u)\bigr) \in \R^d \times \R^d.$
\end{example}

\subsection{Lyapunov Exponents}\index{Lyapunov exponent}

Under conditions described by Oseledet's Theorem\index{Oseledet's Theorem} (see Bibliography
Section \ref{sec:TSB}), closely related to the preceding informal definition
of ergodicity\index{ergodic}, we can define the maximal 
\emph{Lyapunov exponent}\index{Lyapunov exponent} 
of $\psi$ as
\begin{equation}
\label{eq:lyapunov}
    \lambda_{\text{max}} = \lim_{n\to\infty}\lim_{\epsilon\to 0^+}\frac{1}{n}\log\frac{|\psi^{(n)}(\vd + \epsilon e) - \psi^{(n)}(\vd)|}{\epsilon},
\end{equation}
for $\mu$-almost every $\vd$ and almost every vector $e$  picked uniformly at random from the unit sphere. For \emph{chaotic}\index{chaos} systems
the maximal exponent will be larger than zero.

For expository purposes consider the case $\psi(v)=Av$ where $A$ is symmetric and positive definite, and suppose that its largest eigenvalue $\gamma$ is simple, with corresponding unit eigenvector $\varphi$.
This does not satisfy the standard assumptions of the theory, but is simple 
enough to provide intuition. Provided that
the projection of $e$ onto $\varphi$ is non-zero (which will happen almost surely) 
it is clear that $\lambda_{\text{max}}=\log(\gamma).$ Thus the maximal 
Lyapunov exponent\index{Lyapunov exponent}  delivers, through exponentiation, 
the typical growth rate of perturbations to the system.

\section{Analog Forecasting}\index{forecasting!analog}
\label{sec:AF}

Assume that we are given time-series data set $\Vd$ 
as in Data Assumption \ref{da:forc}.
The goal of \emph{forecasting}\index{forecasting} 
is to predict $v_{q} = \psi^{(q)}(v_0)$ given $v_0$, for $q \in \mathbb{N},$ assuming that the map $\psi$ is unknown.
\emph{Analog forecasting}\index{forecasting!analog}, specifically, produces forecasts by identifying nearby points to $v_0$ in the data set $\Vd$ and using the evolution, in the data set, of these points to make a forecast.

\subsection{Basic Analog Forecasting}\index{forecasting!analog}
\label{ssec:baf}

Let $\dd$ be a distance-like deterministic scoring rule
    \index{scoring rule!distance-like deterministic} 
    on $\R^{\du}$ from Definition \ref{def:det_score}, 
    recalling that any norm on $\R^{\du}$ provides
    an example. A basic instantiation of \emph{analog forecasting} works as follows. Let 
\begin{equation} \label{eq:argmin_analog}
n^\star \in \argmin_{n \in \mathsf{N}} \dd(v_0,\vd_{n}).
\end{equation}
Then the prediction is
\begin{equation} \label{eq:argmin_analog2}
v_{q}=\vd_{n^\star+q}\,\quad q=0, \ldots, N-n^\star.
\end{equation}
For $q > N-n^\star$ no prediction is made because
the data does not support doing so.
In principle the $\argmin$ operation may not deliver a unique $n^\star$. 
The preceding discussion implies that choosing
the minimum of all $n^\star$ will deliver the ability to make predictions over
the longest horizon.

The intuition behind the forecast is simple to articulate:
find the closest point in the data set to $v_0$ and ask what happened $q$ time units later in the data set. This is then used
as the prediction.

\begin{theorem}
\label{t:disc}
Assume that we are given a time-series data set as in Data
Assumption~\ref{da:forc} and that the time series is not constant in
$n$. Then the nearest-analog map
\begin{equation*}
F_0(v):=\vd_{n^\star(v)},
\qquad
n^\star(v)\in\argmin_{n\in\mathsf{N}}\dd(v,\vd_n),
\end{equation*}
is not continuous.
\end{theorem}

\begin{proof}
The range of $F_0$ is contained in the finite set
$\{\vd_0,\ldots,\vd_N\}$. Moreover, $F_0(\vd_n)=\vd_n$ for every $n$,
and hence $F_0$ is not constant. If $F_0$ were continuous, its range
would be connected because $\R^{\du}$ is connected. However, a finite
connected subset of $\R^{\du}$ contains only one point, giving a
contradiction.
\end{proof}
For $q>0$, the forecast map may also be discontinuous: as $v_0$ varies,
the selected nearest analog may change from $\vd_n$ to $\vd_{n'}$,
causing the prediction to jump from $\vd_{n+q}$ to $\vd_{n'+q}$ when
these two outcomes differ.

\begin{remark}
Rather than forecasting from $v_0$ one can consider the setting where one forecasts from an initial trajectory $\{v_0, \ldots, v_l\}$ rather than from a single point $v_0.$ In that setting, analog forecasting proceeds by finding the closest so-called \emph{motif} of length $l+1$ in the data set, and then forecasting by considering what happened from that point onward.
Such an approach may be motivated by Takens' embedding theorem\index{Takens' embedding theorem}, discussed
below in Remark \ref{rem:takens}.
\end{remark}

In the next subsection we describe a generalization which resolves the undesirable 
discontinuous behavior highlighted in Theorem \ref{t:disc}.

\subsection{Kernel Analog Forecasting}\index{forecasting!kernel analog}
\label{ssec:KAF}

The idea we deploy here to overcome the discontinuous behavior of
the simple analog forecasting method in the previous subsection is
an intuitive one: rather than predict using a single closest
point in the data set, we predict using multiple points in the
data set, but weight them \emph{smoothly} according to their relevance
to the initial point $v_0$ from which we wish to make a forecast.
To effect this smooth weighting we use kernels\index{kernel} as
introduced in Definition \ref{def:kernel}.

Let $c:\R^{\du} \times \R^{\du} \to \R^+$ be a positive definite and strictly positive
\emph{kernel} -- see Definition \ref{d:mk}. Then make prediction 
\begin{equation}
\label{eq:houman-style}
v_q=\sum_{n=0}^{N-q} p(v_0,\vd_{n})\vd_{n+q}, \quad q=0, \ldots, N,
\end{equation}
where the weight function $p:\R^{\du} \times \R^{\du} \to \R^+$ normalizes the kernel evaluations 
so that the weights add to one:
\begin{equation*}
    p(v_0, \vd_n) = \frac{c(v_0, \vd_n)}{\sum_{m=0}^{N - q} c(v_0, \vd_m)}.
\end{equation*}
In practice this method will not work well for $q$ close to $N$ as the data available is then sparse.
Note that the basic analog forecast \eqref{eq:argmin_analog2} only worked for $q$ up to $N-n^\star$.

This method, too, has a simple intuitive explanation, generalizing
analog forecasting.  Rather than forecasting using $q$ steps ahead 
from a single closest point in the data set to the initial condition, 
$q$-step forecasts are made from the first $(N+1-q)$ points in the data set. 
These are then weighted according to how close the starting point in
the data set is to $v_0$; the kernel performs this weighting.

\begin{example}
A typical choice of kernel is the radial basis function kernel\index{radial basis function} 
$c(v,w)=\exp(-|v-w|^{2}/\ell)$. The choice of $\ell$ is crucial 
to the success of the methodology and should be viewed as a hyperparameter,
tuned to the specific data set at hand.
\end{example}

\begin{remark}
Assuming that the kernel is continuous in its arguments, it
follows that this forecast will be continuous with respect to
$v_0.$ Note that we have only used points in the data set
from which it is possible to forecast forward $q$ steps. 
In fact we can only do this for a subset of points in the data 
given by Data Assumption \ref{da:forc}. 
Thus the effective size of the data set is $N+1-q$ in \eqref{eq:houman-style}. 
\end{remark}

\subsection{Role of Lyapunov Exponents}
\label{ssec:erg}

The success of analog forecasting depends on the 
Lyapunov exponents\index{Lyapunov exponent} of the system that generated 
$\Vd$. Recall that the analog forecast from $v_0$ is found by identifying $n$ such that $\vd_n$
is the closest point in the data set to $v_0$ and setting $v_q=\vd_{n+q}=\psi^{(q)}(\vd_n).$
Thus, approximately for $|\vd_n-v_0|$ small, we have that on average
\begin{equation}
    |v_{q} - \psi^{(q)}(v_0)| = |\psi^{(q)}(\vd_n) - \psi^{(q)}(v_0)| \approx |\vd_n-v_0|e^{\lambda_{\text{max}} q}.
\end{equation}
When $\psi$ is chaotic\index{chaos} and $\lambda_{\text{max}}>0$, this leads to a potentially large error whenever $q$ is large.

\section{Recurrent Structure}
\label{sec:RS}

As in the previous section we seek to forecast in a purely data-driven way, using 
Data Assumption \ref{da:forc}. Here we attempt to forecast by learning a map consistent
with the data given in Data Assumption \ref{da:forc}, and then use that map to forecast.
In its simplest form the map is from $\R^{\du}$ into itself, and this
map can be composed with itself $q$ times, and applied to any initial
point $v_0$, to provide a forecast $v_q.$ This idea is explained
in Subsection \ref{ssec:TimeSeries_Markovian}. The underlying 
assumption in trying to predict using a map from $\R^{\du}$ into itself
is that the data is generated by a Markovian model on $\R^{\du}$. If the
data is not compatible with this Markovian assumption then memory needs
to be accounted for in prediction. In Subsection \ref{ssec:TimeSeries_memory}
we explain how this memory may be accounted for by learning a Markovian model
on $\R^{\du+r}$, where $r$ itself is a hyperparameter to be learned.
Subsection \ref{ssec:reservoir} makes a link between
the methods introduced in Subsection \ref{ssec:TimeSeries_memory} and
the idea of random features\index{random features} introduced
in Section \ref{sec:RF}.

\subsection{Markovian Prediction} \label{ssec:TimeSeries_Markovian}

Assume that the data $\Vd = \{\vd_n\}_{n=0}^{N}$ from Data Assumption \ref{da:forc} 
is the outcome of deterministic Markovian model \eqref{eq:modfit0}, but that
$\psi$ is not known to us. In order to learn $\psi$ we introduce parameter space
$\Theta \subseteq \R^{d_\theta}$ and family of functions $\psi(\cdot\,;\theta):\R^d \to \R^d$,
for all $\theta \in \Theta.$ We then aim to choose $\theta \in \Theta$ so that dynamical system
\begin{align}\label{eq:modfit}
v_{n+1}=\psi(v_n;\theta)
\end{align}
fits the given data $\Vd.$
Function $\psi(\cdot\,;\theta):\R^d \to \R^d$ may be chosen, for example,
as a neural network, a random features model,
or the mean of a Gaussian process, as in Chapter \ref{ch:SL}. 

We wish to choose $\theta \in \Theta \subseteq \R^{d_\theta}$ so
that the data provided by Data Assumption \ref{da:forc} is plausibly  
generated, approximately, by \eqref{eq:modfit}. 
We can view this as a generalization of the supervised learning problem, 
defined by Data Assumption \ref{data:sl}, where the inputs are 
no longer i.i.d. With this connection in mind, we propose to learn $\theta$ 
by solving the following optimization problem:
\begin{subequations}
\label{eq:noroll}
\begin{align}
\J^N(\theta) & = \frac{1}{N}\sum_{n=0}^{N-1}\bigl|\vd_{n+1}-\psi(\vd_n;\theta)\big|^2,\\
\thetas & \in \argmin_{\theta \in \Theta} \J^N(\theta).
\end{align}
\end{subequations}
 
This may be viewed as a supervised learning\index{supervised learning} objective with \emph{correlated} data. 
Furthermore, the $\vd_n$ are not, in general, identically distributed unless 
in the setting where $\vd_0$ is sampled from an invariant measure.
At first glance these two facts would appear to make the data less informative
than in the standard i.i.d. case arising in supervised learning.\index{supervised learning} 
However, there are additional features that result from assuming the
data comes (at least approximately) from a deterministic map on $\R^{d}.$
In particular, as well as expecting that data points indexed by integers
separated by one are linked via $\psi(\cdot\,;\theta)$, we 
also expect that data points separated by integer $m$ are linked by
$\psi^{(m)}(\cdot\,;\theta)$, the $m$-fold composition of $\psi(\cdot\,;\theta)$
with itself. Using this idea with $m=2$ suggests, for example, consideration
of the loss function, for some $\lambda>0$,
\begin{subequations} 
\label{eq:roll}
\begin{align}
\J^N(\theta) & = \frac{1}{N}\sum_{n=0}^{N-1}\bigl|\vd_{n+1}-\psi(\vd_n;\theta)\big|^2 + \lambda \frac{1}{(N-1)}\sum_{n=0}^{N-2}\bigl|\vd_{n+2}-\psi^{(2)}(\vd_n;\theta)\big|^2 ,\\
\thetas & \in \argmin_{\theta \in \Theta} \J^N(\theta).
\end{align}
\end{subequations}

\begin{remark} \label{rem:ro}
There are many variants on the idea expressed in the optimization
problem \eqref{eq:roll}, often referred to as \emph{roll-out}\index{roll-out}.
For example here we have rolled out over $m=2$ steps, but other choices of
roll-out duration $m$ may be employed. Furthermore the roll-out can be employed
in a two-step optimization process, first solving \eqref{eq:noroll} and
only then solving \eqref{eq:roll}, initialized at the outcome of the first
optimization problem; this is sometimes referred to as fine-tuning\index{fine-tuning}. 
Since neither optimization will be exact in general,
this will typically deliver a different parameter choice than that arising from
directly attempting to solve \eqref{eq:roll}. 

Lyapunov exponents\index{Lyapunov exponent} (recall \eqref{eq:lyapunov}) 
play a role
in interpreting the incorporation of roll-out. In general, the larger $m$ is,
the harder it is to fit the data to $m$-fold compositions of 
$\psi(\cdot\,;\theta)$. In particular, for chaotic\index{chaos} systems,
the maximal Lyapunov exponent will be larger than zero, leading to sensitivity
in maps defined by $m$-fold composition for larger $m.$ The two-stage process, where
a parameter ball-park is identified, without roll-out, and then it is adjusted
to reflect a roll-out, over $m=2$ or more compositions, can help in training
on chaotic systems.
\end{remark}

The population-level optimization problem that \eqref{eq:noroll} approximates
may be defined as follows. Assume that $\vd_0 \sim \mu$ 
is sampled from invariant measure $\mu$ and let $\pi_1 \in \mathcal{P}(\R^d \times \R^d)$ 
be the correlation measure of the process $\{\vd_n \}.$ Then
\begin{subequations} \label{eq:thisadd}
\begin{align}
\J(\theta) & = \bbE^{(v,w) \sim \pi_1(\cdot,\cdot)}\bigl|w-\psi(v;\theta)\big|^2,\\
\thetas & \in \argmin_{\theta \in \Theta} \J(\theta).
\end{align}
\end{subequations}
Furthermore, if the process $\{\vd_n\}_{n=0}^{N-1}$ is ergodic\index{ergodic}, then $\J(\theta)$ is the pointwise limit of $\J^N(\theta).$ 
The following theorem shows that the minimizer of the population-level loss is the posterior mean\index{posterior mean} $\Expect[w|v]$. In particular, if the function that computes the posterior mean is within the class of functions parameterized by $\theta\in\Theta$, it will be recovered by minimizing $\J(\theta)$.

\begin{theorem}
Assume that $w$ and $\psi(v)$ are square-integrable. Then,
\begin{equation}
\Expect^{(v, w)\sim \pi_1(\cdot, \cdot)} \Bigl[|w - \psi(v)|^2\Bigr]\geq \Expect^{(v, w)\sim \pi_1(\cdot, \cdot)} \Bigl[|w - \Expect[w|v]|^2\Bigr].
\end{equation}
Furthermore, the inequality is strict if $\psi(v)\neq \Expect[w|v]$.
\end{theorem}
\begin{proof}
This follows directly from Theorem~\ref{th:postmean_opt}.
\end{proof}

If we include roll-out over two steps then the natural 
population-level data assumption is that we are given, not only $\pi_1(v,w)$
from a stationary\index{stationary} process,
but also the correlation measure $\pi_2(v,w)$ in 
$\mathcal{P}(\R^d \times \R^d)$ of the same  process.
The limiting optimization problem is
\begin{subequations}
\begin{align}
\J(\theta) & = \bbE^{(u,v) \sim \pi_1(\cdot,\cdot)}\bigl|v-\psi(u;\theta)\big|^2+\lambda \bbE^{(u,w) \sim \pi_2(\cdot,\cdot)}\bigl|w-\psi^{(2)}(u;\theta)\big|^2,\\
\thetas & \in \argmin_{\theta \in \Theta} \J(\theta).
\end{align}
\end{subequations}
These ideas can be readily generalized to roll-out over $m$ steps.

\subsection{Memory\index{memory} and Prediction} \label{ssec:TimeSeries_memory}

The premise of the preceding subsection is that 
data in Data Assumption \ref{da:forc} derives from a Markov\index{Markov!model} model,
such as \eqref{eq:modfit0}, but there are numerous settings 
where this is not the case, even approximately. Attempting to
fit a model of the form \eqref{eq:modfit}, using optimization
problems such as \eqref{eq:noroll} or \eqref{eq:roll}, will typically
lead to large values of the objective function at the optimum, and to
poor predictions, when the data is not consistent with a Markovian
model. In such a setting it is of interest to introduce 
an unobserved latent variable $\{r_n\}_{n=0}^{N-1}$, with 
$r_n \in \R^{r}$, and then to hypothesize a Markov\index{Markov!model} model 
for the pair $(v_n,r_n)$ of the form
\begin{subequations}
\label{eq:rnn}
\begin{align}
v_{n+1}&=\psi(v_n,r_n;\theta_v),\\
r_{n+1}&=\phi(v_n,r_n;\theta_r), \quad r_0=\theta_0.
\end{align}
\end{subequations}
If $\psi, \phi$ are neural networks then the overall structure is known as a \emph{recurrent neural network},\index{recurrent neural network} or RNN for short.
We define $\theta:=(\theta_v,\theta_r,\theta_0) \in \Theta := \Theta_v \times \Theta_r \times \Theta_0$ and 
learn $\theta$ by solving the minimization problem
\begin{subequations}
\label{eq:rnn2}
\begin{align}
\J^N(\theta) & = \frac{1}{N}\sum_{n=0}^{N-1}\bigl|\vd_{n+1}-\psi(\vd_n, r_n;\theta_v)\big|^2,\\
r_{n+1}&=\phi(\vd_n,r_n;\theta_r), \, r_0=\theta_0,\\
\thetas & \in \argmin_{\theta \in \Theta} \J^N(\theta).
\end{align}
\end{subequations}

The key fact about (\ref{eq:rnn}b) is that, once $(\theta_r,\theta_0)$ are chosen
it defines a unique function, for each
$n$, taking the history $\{v_{\ell}\}_{\ell=0}^{n-1}$ into $r_n:$
\begin{equation}
\label{eq:thatmap}
r_n=\mathcal{F}_n\bigl(\{v_m\}_{m=0}^{n-1};\theta_r,\theta_0).
\end{equation}
Thus (\ref{eq:rnn}a) becomes memory dependent as a closed model for the evolution of $v_n$. Nonetheless
we retain the benefits of being Markovian with respect to the pair $(v_n,r_n).$

\begin{remark} \label{rem:twda}
Sometimes $r_0$ is found using data assimilation\index{data assimilation}.
The observations comprise elements of the sequence $\{v_n\}$ and the unobserved sequences to
be learned are components of $\{r_n\}$ and, in particular, the initial condition. Indeed there is a connection here
to the expectation maximization algorithm\index{expectation maximization} from Section~\ref{sec:142}, and its application in Chapter \ref{ch:DAML}.
\end{remark}

\begin{remark} \label{rem:takens}
    The need for memory in time-series prediction can be motivated not only for non-Markovian systems, but also for Markovian but partially observed systems. One way to see this is using \emph{Takens' embedding theorem}\index{Takens' embedding theorem}, which says (roughly) that under quite general conditions, given only partial observations of a Markovian system, compensated for by employing enough memory, one can reconstruct the original dynamics.

We state Takens' Theorem informally; see the bibliography for references.
Suppose that $\psi:\mathcal{M}\to\mathcal{M}$ is a smooth, invertible map on a compact
$d$-dimensional state space $\mathcal{M}$, defining the deterministic dynamical
system $\vd_{j+1}=\psi(\vd_j)$. Let $h: \mathcal{M}\to\mathbb{R}$ 
be a smooth scalar-valued observation function and define vector  $\Yd_j \in \R^{m}$ by 
    \begin{equation*}
        \Yd_j := \Bigl(h(\vd_{j-(m-1)}), h(\vd_{j-(m-2)}),\ldots, h(\vd_{j})\Bigr).
    \end{equation*}
    Provided $m$ is large enough, Takens' Theorem states that for typical
smooth observation functions $h$ the delay-coordinate map
$\vd_j\mapsto\Yd_j$ is a smooth embedding; informally, this excludes
exceptional observation functions whose delayed values fail to
distinguish different states. It follows that
$\Yd_{j+1}=\Psi(\Yd_j)$ for some map $\Psi$ defined on the set of delay
vectors, and that this new dynamical system is equivalent to the
original one in the sense that they are related by a smooth, invertible
change of coordinates.
    \end{remark}

\subsection{Memory\index{memory} and Reservoir Computing\index{reservoir computing}}
\label{ssec:reservoir}
Including memory is important in many time-series prediction tasks,
as the data may not come from a Markov process. However, learning
RNN structure can be challenging for two reasons: (i) because $r_n$ is an unobserved
latent variable; and (ii) because, within the context of gradient-based methods (Chapter \ref{ch:optimization}),
propagation of gradients of $(\theta_r,\theta_0)$ through the model is difficult. 
\emph{Reservoir computing} is one approach to overcome this. The model deployed is
again \eqref{eq:rnn}; however, rather than learning
$(\theta_r,\theta_0)$, that parameter is simply picked at random from some probability
measure on space of appropriate dimension and the following
optimization problem is solved:
\begin{subequations}
\label{eq:rfc}
\begin{align}
\J^N(\theta_v) & = \frac{1}{N}\sum_{n=0}^{N-1}\bigl|\vd_{n+1}-\psi(\vd_n, r_n;\theta_v)\big|^2,\\
r_{n+1}&=\phi(\vd_n,r_n;\theta_r), \, r_0=\theta_0,\\
\thetas_v & \in \argmin_{\theta_v \in \Theta_v} \J^N(\theta_v).
\end{align}
\end{subequations}
Crucially the optimization is only over $\theta_v$, as
$(\theta_r,\theta_0)$ is fixed at a randomly chosen value.

\begin{remark}
Typically the method requires $r$ of higher dimension than
for the RNN in order to be successful. This is related to
the issue of how expressive randomly chosen functions are.
Indeed, there is a connection to the random features\index{random
features} methodology of Section \ref{sec:RF}. To see
this, recall that equation (\ref{eq:rfc}b) reveals that
$r_n$ is defined by \eqref{eq:thatmap}.
Thus, once $(\theta_r,\theta_0)$ is chosen at random, $r_n$ 
is a fixed function of the history of the observed data up to time $n-1.$ 
\end{remark}

\section{Autoregressive Models}\index{autoregressive!model} \label{sec:autoregressive}

The models we have fit to data in Sections \ref{sec:AF} and \ref{sec:RS} are all deterministic. Here we
consider learning stochastic models to explain the data.
We continue to work under Data Assumption \ref{da:forc}.
To this end, consider the stochastic dynamical system, for sequences $\{v_n\}_{n \in \mathbb{Z}^+}$ in $\R^d$,
\begin{equation}
\label{eq:LAM}
    v_{n+1} = \Psi\bigl(\{v_{n-\ell}\}_{\ell=0}^{p-1}\bigr)+\varepsilon_n, \qquad n\geq p-1,
\end{equation}
where $\varepsilon_n \sim \nu$ i.i.d. for centered $\nu \in \mathcal{P}(\R^d)$ and $\Psi:\R^{pd}\to\R^d$ is a deterministic mapping. Generating a sequence requires specification of $\{v_{\ell}\}_{\ell=0}^{p-1}$. 

\begin{remark}
When
equipped with initialization of  $\{v_{\ell}\}_{\ell=0}^{p-1}$, equation~\eqref{eq:LAM} defines a 
Markov process on $\R^{pd}$ for variable
\begin{equation}
\label{eq:refback0}
 V_{n} := \{v_{n + \ell}\}_{\ell=0}^{p-1}, \qquad n \in \mathbb{Z}^+.
\end{equation}
However we will not use this formulation on $\R^{pd}$ explicitly in what follows.
\end{remark}

To link with Data Assumption \ref{da:forc} we consider a sequence defined for $n \in \mathbb{N}$ and note the following
autoregressive factorization\index{autoregressive!factorization} of the probability of such a sequence under
the dynamics model \eqref{eq:LAM}:
\begin{subequations}
\label{eq:LAMO}
\begin{align}
\bbP\bigl(\{v_{n}\}_{n=0}^{N}\bigr)&=\Bigl[ \prod_{n=p}^N\, \bbP\bigl(v_n| \{v_{\ell}\}_{\ell=0}^{n-1}\bigr)   \Bigr]  
\bbP\bigl(\{v_{m}\}_{m=0}^{p-1})\\
&=\Bigl[ \prod_{n=p}^N\, \bbP\bigl(v_n| \{v_{\ell}\}_{\ell=n-p}^{n-1}\bigr)   \Bigr] 
\bbP\bigl(\{v_{m}\}_{m=0}^{p-1})\\
&=\Bigl[ \prod_{n=p}^N\, \nu\Bigl(v_n-\Psi\bigl(\{v_{n-1-\ell}\}_{\ell=0}^{p-1}\bigr)\Bigr)\Bigr]  
\bbP\bigl(\{v_{m}\}_{m=0}^{p-1}\bigr).
\end{align}  
\end{subequations}
The model \eqref{eq:LAM} is readily generalized to noise that is not additive, and models of the form
\begin{equation}
\label{eq:LAM2}
    v_{n+1} = \Psi\bigl(\{v_{n-\ell}\}_{\ell=0}^{p-1},\varepsilon_n\bigr).
\end{equation}
In this case  (\ref{eq:LAMO}b) still holds; identity (\ref{eq:LAMO}c) however, is specific to additive noise.

In Subsection~\ref{ssec:linear_autoregressive} we introduce linear autoregressive models as a special case of the autoregressive model \eqref{eq:LAM}. Then, in Subsection~\ref{ssec:NGAR}, we introduce a transport-based generalization of autoregressive models.

\subsection{Linear Autoregressive Models}\index{autoregressive!linear model} \label{ssec:linear_autoregressive}

In this subsection we consider  the  vector autoregressive\index{autoregressive!vector} 
(VAR)\index{autoregressive!VAR} process:
\begin{definition}
The $p$th-order VAR model, written VAR($p$), is
\begin{equation}
\label{eq:VARP}
    v_{n+1} = c + \sum_{\ell=0}^{p-1} A_\ell\,v_{n - \ell} + \varepsilon_n,
\end{equation}
where the $A_\ell \in \R^{d \times d}$ are fixed matrices, $c \in \R^d$ is a fixed vector, and 
$\varepsilon_n\sim\mathcal{N}(0, C)$ is an i.i.d.\thinspace mean-zero Gaussian sequence
in $\R^d$ with fixed covariance $C \in \bbR^{d \times d}_{\textrm{sym},>}.$
\end{definition}

Given initial conditions $\{v_n\}_{n=0}^{p-1}$ this will generate $\{v_n\}_{n \in \mathbb{Z}^+}$, an $\R^d$-valued sequence.
As well as generalizing the foregoing sections by allowing for
stochasticity in the dynamics, this model also generalizes what we
have seen in previous sections in its treatment of memory.
In Subsection \ref{ssec:TimeSeries_Markovian} we employed a deterministic Markovian model 
in $\R^{d}$ with no memory. And in Subsection \ref{ssec:TimeSeries_memory} 
we employed 
a model which, when viewed as a dynamics model in $\R^{d}$, had memory
that, for each $n$, reaches all the way back to time $0$ -- this is demonstrated
in \eqref{eq:thatmap}. The VAR process has memory, if $p>1$, but the
memory is over a fixed finite window, the same for every $n$ and
determined by $p$ -- the memory does not stretch back all the way 
to time $0.$

\begin{remark}
When $v_n$ is a scalar, the process is known simply as an 
autoregressive\index{autoregressive} (AR)\index{autoregressive!AR} model.
It is sometimes useful to consider $\varepsilon_n$ to be
a non-Gaussian strictly stationary and ergodic\index{ergodic} process, but we 
restrict to the i.i.d. Gaussian\index{Gaussian} model here for simplicity.
\end{remark}

In the following we take $c=0$ for simplicity. But the methodological
developments are readily modified to accommodate the setting $c\neq 0$; the resulting optimization problem is 
still quadratic.
We also assume that $C$ is known and positive definite, so that 
the unknown parameters of the model are $\theta:=(A_0,\ldots,A_{p-1}).$
Assume that, for $m=0, \ldots, p-1$, the $v_m$ are i.i.d. random variables with
distribution $v_m \sim \mathcal{N}(0, \gamma^2 I),$ for some $\gamma>0.$
Then, using the data $\Vd = \{ \vd_n \}_{n=0}^N$ given in Data Assumption \ref{da:forc}, 
define the objective function
\begin{equation}
\label{eq:varobj0}
\loss^N(\theta) = \frac{1}{2}\sum_{n=p}^{N}\bigl|C^{-\frac12}\bigl(\vd_{n}-\sum_{\ell=0}^{p-1}
A_{\ell}\vd_{n-\ell-1}\bigr) \big|^2+\frac{1}{2\gamma^2}\sum_{m=0}^{p-1}|\vd_m|^2 
\end{equation}
noting that then, by \eqref{eq:LAMO}, the likelihood\index{likelihood} $\bbP\bigl(\{\vd_{n}\}_{n=0}^{N}|\theta\bigr) \propto \exp \bigl(-\loss^N(\theta)\bigr).$
Thus maximum likelihood estimation\index{maximum likelihood estimation} is defined by the quadratic optimization problem
\begin{equation}
\label{eq:varobj}
\thetas  \in \argmin_{\theta \in \Theta} \loss^N(\theta).
\end{equation}

\begin{remark}
Learning matrix $C$, in addition to the matrices $\{A_\ell\}_{\ell=0}^{p-1}$, is also possible, 
but leads to a non-quadratic objective. Dependence on the normalization constant of the likelihood 
on $\det C$ must be included in the loss function.
\end{remark}

\begin{remark}
If we assume that the data is generated by a strictly stationary process
then there is a population-level limit of the minimization problem defined by $N^{-1}\loss^N.$ 
It may be expressed in terms of the distribution of the underlying stationary 
process on the data. We omit details, but conclude with an example of
a strictly stationary process of VAR type.
\end{remark}

\begin{example}\label{example:var1_dist}
Consider the following VAR(1) process with $c=0$ and defined by matrices $A \in \R^{d \times d}$ and $C \in \bbR^{d \times d}_{\textrm{sym},>}$:
$$v_{n+1}=Av_{n}+\varepsilon_n, \quad n \in \bbZ^+,$$
where $\varepsilon_n\sim\mathcal{N}(0, C)$ i.i.d.\,.
Assume that $A$ is stable, meaning that all its eigenvalues have modulus
strictly less than one. Then the Lyapunov equation
\begin{equation*}
C_\infty=AC_\infty A^\top+C
\end{equation*}
has a unique positive definite solution $C_\infty$. If
$v_0\sim\mathcal{N}(0,C_\infty)$ independently of the i.i.d. noise sequence
$\{\varepsilon_n\}_{n \in \bbZ^+}$, then the process is strictly stationary.
Indeed, then, $v_n \sim \mathcal{N}(0, C_\infty)$ for all $n \in \mathbb{Z}^+.$

To establish these facts we first note that the linear additive Gaussian structure implies by
induction that if $v_0$ is Gaussian, then $v_n$
is Gaussian for all $n \in \mathbb{Z}^+.$ Assume that 
$v_0 \sim \mathcal{N}(m_0, C_0).$ Then $v_n \sim \mathcal{N}(m_n, C_n)$,
and the mean and covariance update according to
\begin{subequations}
\begin{align}
m_{n+1} & = Am_n,\\
C_{n+1} & = AC_{n}A^\top+C.
\end{align}
\end{subequations}
The pair $(0,C_\infty)$ is a fixed point of this iterated map on mean and covariance.
Thus we have established that, if $v_0  \sim \mathcal{N}(0, C_\infty)$
then $v_n  \sim \mathcal{N}(0, C_\infty)$ for all $n \in \mathbb{Z}^+.$
The vector $(v_n, \ldots, v_{n+p})$ will then also be jointly Gaussian
with mean zero. To establish strict stationarity it suffices to show 
that the covariance of $(v_n, \ldots, v_{n+p})$ is independent of $n$, for
any $p \in \mathbb{Z}^+.$ To do this it is sufficient to show that 
the covariance of $v_n$ with $v_{n+m}$ is $n$-independent, for any
$m \in \mathbb{Z}^+.$ We note that, by induction, 
$$v_{n+m}=A^m v_n+\sum_{j=1}^{m} A^{m-j}\varepsilon_{n+j-1}.$$
Because the sequence $\{\varepsilon_n\}_{n \in \mathbb{N}}$ is i.i.d.\thinspace
it follows that, for $m \in \mathbb{N}$,
\begin{align*}
\mathbb{E} [v_{n+m}v_n^\top] &= A^m \mathbb{E} [v_{n}v_n^\top]+\sum_{j=1}^{m} 
A^{m-j}\mathbb{E}[\varepsilon_{n+j-1}v_n^\top],\\
&= A^m C_\infty.
\end{align*}
As desired this depends only on $m$ and not on $n$ and the desired result is
established.
\end{example}

\begin{remark} \label{rem:r2}
    The dynamics and observation processes underlying the steady-state Kalman filter\index{Kalman filter!steady state} (Subsection~\ref{ssec:3dvar}) are VAR(1) processes. Under observability and controllability conditions, moreover, the steady-state Kalman filter itself \eqref{eq:ss3} is a strictly stationary VAR(1) process as in Example~\ref{example:var1_dist}. See the bibliography Section \ref{sec:TSB}.
\end{remark}

\subsection{Transport-Based Autoregressive Models\index{transport}\index{autoregressive!transport}}
\label{ssec:NGAR}

In this section we extend the linear autoregressive models with additive 
Gaussian noise, from Subsection~\ref{ssec:linear_autoregressive}, to 
a more general nonlinear and non-Gaussian probabilistic form. 
We do this by modeling non-Gaussian distributions 
as the pushforward of a Gaussian variable. This is an idea
we introduced through transport\index{transport},
in Section~\ref{sec:transport2}, in the context of
generative modelling.

To motivate the approach we adopt, we first recall the VAR($p$) process \eqref{eq:VARP} from
Subsection~\ref{ssec:linear_autoregressive}. If $(c,C)$ are assumed known, and we define
$\theta=(A_0,\ldots,A_{p-1}),$ then the model can be formulated through use of a conditional 
probability kernel $\pi$ of the form
\begin{equation}
\label{eq:VARP2}
    v_{n+1} \sim \pi(\cdot|v_{n}, \ldots, v_{n-p+1};\theta)=\mathcal{N}\Bigl(c + \sum_{\ell=0}^{p-1} A_\ell\,v_{n - \ell}, C\Bigr).
\end{equation}
Furthermore, the independent noise structure assumed in
Subsection~\ref{ssec:linear_autoregressive} implies the order-$p$
Markov property
\begin{equation*}
\bbP\bigl(v_{n+1} | v_0,\ldots,v_n\bigr)
=
\bbP\bigl(v_{n+1} | v_{n-p+1},\ldots,v_n\bigr).
\end{equation*}

We generalize the linear Gaussian structure of VAR($p$) in what follows.
We let $\refT(v)$ denote the standard Gaussian density on $\R^d.$ We introduce a transport\index{transport} $T(\cdot\,; V,\theta): \R^d \to \R^d$, depending on $V \in \R^{pd}$ and parameters $\theta \in \Theta \subseteq \R^q$, so that
\begin{equation} \label{eq:autoregressive_transport}
    v_{n+1} \sim \pi(\cdot|v_{n}, \ldots, v_{n-p+1};\theta)= T(\cdot\,; v_{n}, \ldots, v_{n-p+1}, \theta)_\sharp \refT.
\end{equation}
We assume that the randomness used at different time indices is
independent; equivalently, the model may be written as
\begin{equation*}
v_{n+1}
=
T\bigl(z_n;v_n,\ldots,v_{n-p+1},\theta\bigr),
\qquad
z_n\stackrel{\mathrm{i.i.d.}}{\sim}\refT.
\end{equation*}

\begin{example}
The VAR \eqref{eq:VARP2} falls into the general class \eqref{eq:autoregressive_transport} with $T$ given by
$$T(z;v_{n}, \ldots, v_{n-p+1}, \theta)=c + \sum_{\ell=0}^{p-1} A_\ell\,v_{n - \ell}+C^{1/2}z,$$
where, if $(c,C)$ are assumed known, $\theta=(A_0,\ldots,A_{p-1}).$
\end{example}

\begin{remark}
Note that \eqref{eq:autoregressive_transport} postulates an autoregressive structure that subsumes \eqref{eq:LAM} 
as a special case. Furthermore, in the case where the $\varepsilon_n$ are Gaussian, \eqref{eq:LAM2} 
provides a state-space formulation of the Markov chain on 
the augmented state-space $\R^{pd}$ (see \eqref{eq:refback})
implied by \eqref{eq:autoregressive_transport}. We also note that the identity
\eqref{eq:autoregressive_transport} can be used in expression (\ref{eq:LAMO}b) 
to define the conditional probabilities arising there.
\end{remark}

In Subsection \ref{ssec:linear_autoregressive} we showed how to learn parameters $\theta$ for the VAR($p$) linear
and Gaussian model, by invoking a maximum
likelihood principle, leading to an optimization problem for $\theta.$
We now aim to choose parameter $\theta$ so that the more general model \eqref{eq:autoregressive_transport} 
explains the data given in Data Assumption \ref{da:forc}, again using a maximum likelihood approach. We assume that
$T(\cdot\,; V,\theta): \R^d \to \R^d$ is invertible for all  $(V,\theta) \in \R^{pd} \times \Theta,$
with inverse $S(\cdot\,; V,\theta): \R^d \to \R^d.$ Map $T(\cdot\,; V,\theta)$ takes as input $z$ in the latent space (which is
$\R^d$ in this case); inverse $S(\cdot\,; V,\theta)$ takes as input $v$ in the original space $\R^d.$
Then we can use the change-of-variables formula from Lemma \ref{lem:cov} to express the density of the transformed variable as 
\begin{align}
     \label{eq:incase}
&\pi(v_{n+1}|v_{n}, \ldots, v_{n-p+1};\theta) \\
&\quad\quad\quad= \refT\bigl(S(v_{n+1};v_{n}, \ldots, v_{n-p+1}, \theta)\bigr)\det D_{v} S\bigl(\cdot\,;v_{n}, \ldots, 
v_{n-p+1},\theta\bigr)\vert_{v_{n+1}}.\nonumber
\end{align}
This expression is useful for formulating an optimization problem to determine the choice of $\theta.$

To simplify the exposition we continue our discussion of transport-based learning
of probabilistic models in the restricted setting where $p=1.$ 
In this case, our goal is to find invertible transport\index{transport map} $T(\cdot\,; V,\theta): \R^d \to \R^d$, with inverse $S(\cdot\,; V,\theta): \R^d \to \R^d$,   so that $\pi(v_{n+1}|v_n,\theta) \coloneqq T(\cdot\,;v_n, \theta)_\sharp \refT(v_{n+1})$ models the conditional distribution 
implied by the data from Data Assumption \ref{da:forc}. 
To this end we work with function classes $\cTt$ that satisfy the inclusion
\begin{align}
\label{eq:thisclass555}
\cTt\subset
\Bigl\{T:\;&T(\cdot\,;w,\theta)\in C^1(\R^d;\R^d)
\text{ is invertible for every }(w,\theta)\in\R^d\times\Theta,
\nonumber\\
&\det D_zT(z;w,\theta)>0
\quad\text{for every }(z,w,\theta)\in\R^d\times\R^d\times\Theta
\Bigr\}.
\end{align}

The following result, which provides a canonical working example in the $p=1$ setting,
follows directly from \eqref{eq:incase}, using the fact that $\refT$ is a standard Gaussian:

\begin{proposition} \label{prop:NonlinearAutoregressive}
Consider the probabilistic model \eqref{eq:autoregressive_transport} in the case where $p=1.$ 
Define $T(z;w,\theta): = \psi(w;\theta) + C^{1/2}z,$ for some class of maps $\psi(\cdot\,;\theta) \colon \R^\du \rightarrow \R^{\du}$ for
all $\theta \in \Theta$ and for matrix $C \in \R^{d \times d}_{\mathrm{sym},>}$. Then we have 
\begin{align*}
S(v; w,\theta) &= C^{-1/2}\bigl(v - \psi(w;\theta)\bigr),\\
    \pi(v|w;\theta) &= \refT\Bigl(C^{-1/2}\bigl(v - \psi(w;\theta)\bigr)\Bigr) \det C^{-1/2}\\
    &= \frac{1}{\sqrt{(2\pi)^\du 
    \,\det{C}}} \exp \left(-\frac{1}{2}|v - \psi(w;\theta)|_{C}^2 \right).
\end{align*}
That is, the conditional distribution for the next state $v_{n+1}$, $\pi(\cdot |v_n;\theta)$, is a multivariate Gaussian of the form $\mathcal{N}\bigl(\psi(v_n),C\bigr)$.
\end{proposition}

\begin{example}
In setting of the preceding proposition we may, for example, 
choose $\psi(\cdot\,;\theta)$ to be from a family of neural networks
of specified depth and widths; see Remark \ref{rem:NNG}.
\end{example}

Given a sequence of data, we can learn the transport parameters by maximizing the log-likelihood of consecutive data 
pairs $\{(\vd_{n},\vd_{n+1})\}_{n=0}^{N-1}$ under the model. From \eqref{eq:incase} we obtain, in case $p=1$,
 \begin{align}
     \label{eq:incase2}
\pi(v_{n+1}|v_{n};\theta) 
&= \refT\bigl(S(v_{n+1}; v_{n}, \theta)\bigr)
\det D_{v}S(v_{n+1}; v_{n}, \theta).
\end{align}
Because of the assumed conditional independence structure, $\pi(v_{n+1}|v_n,\theta)$ defines
a Markov chain\index{Markov!chain} on $\R^d.$ If we assume that $\vd_0 \sim p_0$ then the probability of sequence 
$\{\vd_n\}_{n \in \mathsf{N}}$ from Data Assumption \ref{da:forc} is, using the Markov structure,
given by
\begin{equation*}
   \bbP\bigl(\{\vd_{n}\}_{n=0}^{N}|\theta\bigr)=\Bigl( \prod_{n=0}^{N-1} \pi(v_{n+1}^\dagger|v_n^\dagger;\theta)\Bigr) p_0(\vd_0).
\end{equation*}
This is a particular case of \eqref{eq:LAMO}. We again assume a standard Gaussian reference $\refT$.
We also assume that $p_0$ does not depend on $\theta.$ Then, up to a $\theta$-independent constant, the 
negative log-likelihood $\loss^N(\theta)=\J^N(\theta)+\mathsf{const}$, where
\begin{subequations}
\begin{align}
    \J^N(\theta) &= - \sum_{n=0}^{N-1} \log \pi(v_{n+1}^\dagger|v_n^\dagger;\theta) \label{eq:nll_autoreg}\\
    &= - \sum_{n=0}^{N-1} \left[\log \refT \bigl(S(v_{n+1}^\dagger;v_{n}^\dagger,\theta)\bigr) + \log \det D_{v} S(v_{n+1}^\dagger;v_{n}^\dagger,\theta) \right] \nonumber \\
    &=  \sum_{n=0}^{N-1} \left[\frac{1}{2} |S(v_{n+1}^\dagger;v_{n}^\dagger,\theta)|^2 - \log \det D_{v} S(v_{n+1}^\dagger;v_{n}^\dagger,\theta) \right] + \mathsf{const},     \label{eq:NonlinearAutoRegLoss}
\end{align}
\end{subequations}
where $\mathsf{const}$ is a constant that is again independent of $\theta$. 
The optimal transport parameters are then given by 
$$\theta^\star \in \argmin_{\theta \in\Theta} \J^N(\theta).$$

\begin{remark}
Here we have assumed that $p_0$ is $\theta$-independent. Another natural situation is
where $p_0$ is the invariant measure associated with the process giving rise to
data in Data Assumption \ref{da:forc}. In this case $p_0$ depends on $\theta$ and 
the estimation problem is more complicated. Relatedly, it is also natural to consider 
the population loss limit of $N^{-1}\J^N(\cdot)$, using an ergodic\index{ergodic} invariant measure.
\end{remark}

\begin{example} A commonly used non-Gaussian model is based on a nonlinear diagonal transformation of the form 
$$S(v_{n+1};v_n, \theta) = s(v_{n+1} - \psi(v_n;\theta);\theta),$$
where $s(\cdot\,;\theta): \R^{\du} \rightarrow \R^\du$ is an invertible function, with inverse
$s^{-1}(\cdot\,;\theta): \R^{\du} \rightarrow \R^\du$, for all $\theta \in \Theta$, and $\psi : \R^{\du} \times \Theta \rightarrow \R^\du$. This can be re-written in terms of the direct transport as 
$$T(z; v_n,\theta) = \psi(v_n;\theta) + s^{-1}(z;\theta).$$
Recall that $z\sim\refT$, the standard Gaussian; thus $v_{n+1}$ is
obtained by applying $\psi$ to $v_n$ and adding the generally
non-Gaussian innovation $s^{-1}(z;\theta)$. When $s$ acts
component-wise and monotonically, the distribution of this innovation
is an example of a \emph{nonparanormal}\index{nonparanormal}, or
\emph{Gaussian copula}, distribution\index{copula model}.
\end{example}

\begin{example} Assume that $S(v_{n+1}; v_n, \theta) = v_{n+1} - \psi(v_{n};\theta),$ as in Proposition~\ref{prop:NonlinearAutoregressive}, with $C = I$. Then we have 
$D_{v_{n+1}} S(v_{n+1}; v_n, \theta) = I$. Thus, the learning problem based on the loss function in~\eqref{eq:NonlinearAutoRegLoss} reduces to 
\begin{align*}
    \theta^\star \in \argmin_{\theta \in\Theta} \frac{1}{2N} \sum_{n=0}^{N-1} \left[ |v_{n+1}^\dagger - \psi(v_{n}^\dagger;\theta)|^2 \right].
\end{align*}
That is, we can identify the model parameters by solving a supervised learning problem for the nonlinear map $\psi$. This is the setting of Subsection \ref{ssec:TimeSeries_Markovian}. The scaling factor $N^{-1}$ does not change the optimization problem, but makes clear that under ergodicity assumptions there is a population loss optimization problem of
the form \eqref{eq:thisadd}.
\end{example}

\begin{remark}
In Chapter~\ref{ch:transport} 
we show how to model non-Gaussian distributions, also using 
transport\index{transport}, in the context of Bayesian inverse problems.
\end{remark}

\subsection{Learning the Conditional Distribution Using Strictly Proper Scoring Rules}\label{sec:cond_scoring}\index{scoring rule!proper}

In the preceding Subsection~\ref{ssec:NGAR} we studied the use of transport maps to approximate
\begin{equation}\label{eq:e2e}
    \pi(v_{n+1}|\Vd_n) := \Prob(v_{n+1} | \Vd_n)
\end{equation}
where, recall, $\Vd_n \in \R^{pd}$ is defined by\footnote{Note that the indexing differs from that used to define $V_n$
in \eqref{eq:refback0}.} 
\begin{equation}
\label{eq:refback}
    \Vd_{n} := \{\vd_{n - \ell}\}_{\ell=0}^{p-1}, \qquad n=p-1,\ldots,N.
\end{equation}
We learned parameters of the maps using maximum likelihood estimation. 
In this subsection we learn the same conditional distribution
using scoring rules to define the objective function.
The following theorem provides the basis for learning the conditional $\pi$ using strictly proper scoring rules\index{scoring rule!proper}; recall their definition in Section~\ref{sec:psr}.

\begin{theorem}
Consider a strictly proper scoring rule\index{scoring rule!proper} $\mS: \cP(\R^{\du}) \times \R^{\du} \to \R$,  and let $\pi(v_{n+1}|\Vd_{n}) \in \cP(\R^d)$ be the distribution
 defined by \eqref{eq:e2e}. For any distribution $q(\cdot;\Vd_n)$ on $\R^{\du}$, parameterized by $\Vd_n$, we have
\begin{align}
\label{eq:e2e_inequality}
&\mathbb{E}^{\Vd_n}
\mathbb{E}^{v_{n+1}\sim\pi(\cdot\mid\Vd_n)}
\Bigl[
\mS\bigl(q(\cdot;\Vd_n),v_{n+1}\bigr)
\Bigr]\geq
\mathbb{E}^{\Vd_n}
\mathbb{E}^{v_{n+1}\sim\pi(\cdot\mid\Vd_n)}
\Bigl[
\mS\bigl(\pi(\cdot\mid\Vd_n),v_{n+1}\bigr)
\Bigr].
\end{align}
Furthermore, the inequality is strict if
\begin{equation*}
\mathbb{P}^{\Vd_n}
\Bigl(q(\cdot;\Vd_n)\neq\pi(\cdot\mid\Vd_n)\Bigr)>0.
\end{equation*}
\end{theorem}

\begin{proof}
    This follows directly from Theorem~\ref{th:scoring_rule_joint}.
\end{proof}

Now we use the objective function implied by \eqref{eq:e2e_inequality} as the basis of an implementable algorithm,
under Data Assumption \ref{da:forc}. Approximating $\pi(\cdot| \Vd_n)$ by $\pi(\cdot | \Vd_n; \theta)$ and making
an empirical approximation in time, justified by assuming ergodicity, we obtain
\begin{equation}\label{eq:scoring_rule_autoreg_emp}
    \J^N(\theta) = \frac{1}{N-p+1}\sum_{n=p-1}^{N-1} \mS(\pi(\cdot | \Vd_n; \theta), \vd_{n+1})
\end{equation}
for some choice of strictly proper scoring rule $\mS$.

\begin{remark}
Note that the transport-based\index{transport} conditional distributions \eqref{eq:autoregressive_transport} are a possible way of parameterizing $\pi(v_{n+1} | \Vd_n; \theta)$; however, other classes of distributions can also be considered.
\end{remark}

\begin{remark} Consider the case $p=1.$
    Then the log-likelihood cost function \eqref{eq:nll_autoreg} considered in Subsection~\ref{ssec:NGAR} is a special case of \eqref{eq:scoring_rule_autoreg_emp}, when $\mS$ is chosen to be the logarithmic score\index{score!logarithmic} from Definition~\ref{d:logs}.
\end{remark}

\begin{remark}
    Suppose the process underlying the time-series $\Vd$ is a hidden Markov model\index{hidden Markov model} with a hidden state $z$, i.e.,
    \begin{align*}
        z_{n+1} &= \Psi(z_n) + \xi_n,\quad\xi_n\sim\mathcal{N}(0, \Sigma),\\
        v_{n+1} &= h(z_{n+1}) + \eta_{n+1}, \quad \eta_{n+1}\sim \mathcal{N}(0, \Gamma),\\
        z_0 &\sim \mathcal{N}(m_0, C_0),
    \end{align*}
    as in \eqref{eq:sdm}, \eqref{eq:dm} (note shift of notation to align with that arising in this subsection). Then, a structure similar to \eqref{eq:e2e} arises; however $\Vd_n$ has a different definition, with dimension depending on $n$.
   Now $\Vd_n$ is given by \eqref{eq:VdYd} as $\Vd_n := \{\vd_0, \ldots, \vd_n\} \in \R^{(n+1)d}$, to be contrasted with the fixed lag
   definition of $\Vd_n \in \R^{pd}$ given by \eqref{eq:refback}, and adopted in the discussion preceding this remark.

    We now discuss how, with this modified $\Vd_n$, the conditional probability distribution given by \eqref{eq:e2e}
    can be approximated using filtering algorithms.
    In particular, suppose one has obtained the predicted distribution $\widehat{\pi}_{n+1}(z_{n+1}) := \Prob(z_{n+1} | \Vd_n)$; see the definitions in Subsection~\ref{ssec:filtering}. Then, \eqref{eq:e2e} can be obtained from $\widehat{\pi}_{n+1}$ by application of the stochastic observation map. That is, using convolution\index{convolution} and pushforward,\index{pushforward}
    \begin{equation} \label{eq:1433}
        \Prob(v_{n+1} | \Vd_n) = \mathcal{N}(0, \Gamma) * h_\sharp \widehat{\pi}_{n+1}.
    \end{equation}
    In the context where the time-series has such a hidden Markov structure it is sometimes of interest to predict $v_{n+1}$ from $\Vd_n$ directly, bypassing the prediction--analysis structure of filtering (see Subsection~\ref{ssec:filtering}); this may be the case, for example, when a good approximation for $\Psi$ is not available. This is sometimes called \emph{end-to-end prediction}\index{end-to-end prediction} or \emph{direct observation prediction}.
\end{remark}


\section{Bibliography}
\label{sec:TSB}
An extensive introduction to linear autoregressive processes, including the conditions for stationarity and ergodicity\index{ergodic}, is provided in \cite{hansen_econometrics_2022}.
The Lorenz '63 model is proved to be ergodic\index{ergodic} in
\cite{tucker1999lorenz,tucker2002rigorous} providing a useful example. Indeed, in this context central limit theorems have been proved \cite{holland2007central}, 
characterizing the fluctuations about the law of large numbers limit in \eqref{eq:erg}.

Analog forecasting was first introduced by Lorenz in \cite{lorenz_atmospheric_1969}. A historical overview of analog forecasting and extensions, as well as estimates of error scaling, is given in \cite{farmer_exploiting_1988}. The kernel analog
regularization of Lorenz's idea was introduced and
then studied in \cite{giannakis2019data,alexander2020operator}; it has recently
been extended to an efficient streaming framework \cite{giannakis2023learning}.
 
 The key message contained in Subsection \ref{ssec:erg} is that purely data-driven
methods of this type will be limited by sensitivity to initial conditions.
In contrast, the data assimilation\index{data assimilation} methods,
which combine data with knowledge of the model, can overcome
sensitivity to initialization; the mechanism by which this occurs
is explained in \cite{sanz2015long} and in \cite{moodey2013nonlinear}
and also underlies Theorem \ref{t:ques} concerning the effect of
model error (rather than the effect of initialization error).

Memory in the context of learning model error is discussed in \cite{levine_framework_2022}. Reservoir computers were considered for predicting chaotic systems in \cite{pathak_model-free_2018}. Takens' Theorem and other embedding theorems are reviewed in \cite{sauer_embedology_1991}.
The Remark~\ref{rem:twda} highlights the use of data assimilation in training data-driven
models. Incorporating data assimilation into the training process can be very useful, especially when learning from noisy observations and with unobserved latent variables; see Chapter~\ref{ch:DAML} and the bibliography therein. Furthermore, if available, it is often beneficial to train on analyses produced by an imperfect model rather than on observations directly; see Subsection~\ref{ssec:rean} and \cite{bouallegue_rise_2024}.

The stationarity of the linear autoregressive model referred to in Remark \ref{rem:r2} is established, under 
controllability and observability conditions, by study of matrix Riccati equations -- see \cite{lancaster1995algebraic}.
General non-linear autoregressive models have been considered for time-series modeling using various probabilistic models, such as conditional normalizing flows,  in~\cite{oliva2018transformation, rasul2020multivariate}. Some of these models rely on recurrent neural networks architectures that can summarize the past states and extract relevant features for making predictions. Learning probabilistic time-series forecasts using scoring rules was considered in \cite{pacchiardi_probabilistic_2024}.
\chapter{\Large{\sffamily{Optimization and Sampling}}}\label{ch:optimization}
This chapter provides an introduction to various topics in optimization and sampling that are particularly pertinent to machine learning, inverse problems and data assimilation. 
We start in Section~\ref{sec:preliminaries} with definitions and background results that will be useful throughout this chapter.
Section~\ref{sec:grad} is concerned with gradient-based
 optimization, including deterministic and stochastic
 gradient descent. In Section~\ref{sec:NGN} we discuss Newton and Gauss--Newton optimization methods for nonlinear least squares; these algorithms use full or partial Hessian information to define a preconditioner and thereby accelerate the convergence. 
 Section~\ref{sec:EKI} describes derivative-free ensemble Kalman methods for nonlinear least squares; these methods, which are intimately related to ensemble Kalman filters for data assimilation studied in Chapter \ref{lecture7}, rely on the first two empirical moments of an ensemble of particles to determine the search direction and to define a preconditioner.
 Section~\ref{sec:142} is devoted to expectation maximization, an important optimization algorithm for computing maximum likelihood estimators when the likelihood is intractable but can be expressed as the marginal of a tractable extended likelihood. Section~\ref{sec:MCMC} describes the Metropolis--Hastings algorithm and Gibbs samplers, two Markov chain Monte Carlo  algorithms to produce approximate samples from a target distribution. 
 Finally, since many optimization and sampling algorithms involve derivatives of the objective, and since computing these derivatives can be a limiting computational bottleneck, in Section~\ref{sec:auto-differentiation}
 we study auto-differentiation. These techniques have greatly contributed to the scalability and efficiency of modern optimization and sampling algorithms in machine learning and related tasks. Other topics in optimization and sampling are referenced in the bibliography Section~\ref{sec:OB}.


\section{Background Definitions and Results}\label{sec:preliminaries}
In this section we introduce some definitions
that will be useful in later sections and in other chapters.

\begin{definition} \label{def:convex}
   A function $\J: \R^d\to\R$ is called \emph{convex}\index{convex!function} if, for all $u, v\in \R^d$ and for all $\theta\in [0, 1]$, we have that
    \begin{equation*}
        \J(\theta u + (1-\theta) v) \leq \theta \J(u) + (1-\theta) \J(v).
    \end{equation*}
    It is called \emph{strictly convex}\index{convex!function!strictly} if the inequality is strict whenever $u \ne v$ and $\theta \in (0,1)$. For $\mu \ge 0$, function $\J \in C^1(\R^d,\R)$ is called \emph{$\mu$-strongly convex}\index{convex!function!$\mu$-strongly} if, for all $u, v\in \R^d,$ 
    $$\J(v) \ge \J(u) + \langle D\J(u), v- u \rangle +  \frac{\mu}{2} | v - u |^2.$$
\end{definition}
Notice that for $\mu=0$, $\mu$-strong convexity is equivalent to convexity, while
$\mu$-strong convexity with $\mu>0$ implies strict convexity.
An important consequence of strict convexity is uniqueness of minimizers:

\begin{theorem}
    Let $\J: \R^d\to\R$ be strictly convex.\index{convex!function!strictly} 
    If $u_1$ and $u_2$ are minimizers of $\J,$ then $u_1 = u_2.$
\end{theorem}
\begin{proof}
    Suppose for contradiction that $u_1\neq u_2$ are distinct minimizers, and let $\J_{\rm min}:= \J(u_1) = \J(u_2).$ Let $ \theta \in (0,1)$ and consider the intermediate point $u_\theta:= \theta u_1 + (1- \theta) u_2.$ By strict convexity, since $u_1 \neq u_2$ we have
    \begin{align*}
        \J(u_\theta) &= \J(\theta u_1 + (1- \theta) u_2) 
        < \theta \J(u_1) + (1-\theta) \J(u_2) = \J_{\rm min}.
    \end{align*}
    This contradicts the fact that $u_1$ and $u_2$ are minimizers of $\J.$
\end{proof}

The following result provides a useful equivalent characterization of $\mu$-strong convexity.

\begin{theorem}\label{thm:convexcharacterization}
    Let $\J \in C^1(\R^d,\R).$ The following are equivalent:
    \begin{enumerate}
        \item $\J$ is $\mu$-strongly convex.\index{convex!function!$\mu$-strongly}
        \item For all $u,v \in \R^d$ it holds that
        \begin{equation}\label{eq:equivalentconvexity}
            \langle D \J(u) - D \J(v), u - v \rangle \ge \mu |u - v|^2.
        \end{equation}
    \end{enumerate}
\end{theorem}
\begin{proof}
    Suppose first that $\J$ is $\mu$-strongly convex. Then
    \begin{align*}
        \J(v) &\ge \J(u) + \langle D \J(u), v - u \rangle + \frac{\mu}{2} |v-u|^2, \\
         \J(u) &\ge \J(v) + \langle D\J(v), u - v\rangle + \frac{\mu}{2}| u - v|^2.
    \end{align*}
    Adding these inequalities, we obtain that 
\begin{align*}
    \J(v) + \J(u) \ge \J(u) + \J(v) + \langle D\J(u), v-u\rangle + \langle D\J(v), u - v\rangle + \mu | u - v|^2. 
\end{align*}
Consequently,
\begin{align*}
    0 \ge \langle D\J(u), v-u\rangle + \langle D\J(v), u - v\rangle + \mu | u - v|^2,  
\end{align*}
which implies \eqref{eq:equivalentconvexity}. 

Conversely, suppose that \eqref{eq:equivalentconvexity} holds for all $u,v \in \R^d.$ Fix $u,v \in \R^d.$ Then by the fundamental theorem of calculus, we have
\begin{align}\label{eq:auxiliaryeq}
\begin{split}
    \J(v) - \J(u) &= \int_0^1 \bigl\langle D\J\bigl(u + \theta(v - u)\bigr), v-u \bigr\rangle d\theta  \\ 
    & = \langle D\J(u), v - u \rangle + 
    \int_0^1 \bigl\langle D\J \bigl(u + \theta(v - u)\bigr) - D\J(u), v-u \bigr\rangle d\theta. 
    \end{split}
\end{align}
Let $x := u + \theta(v-u)$ and $y := u.$ Applying \eqref{eq:equivalentconvexity} with $(u,v) \mapsto (x,y)$ gives
\begin{equation*}
    \langle D \J(u + \theta(v-u)) - D \J(u), \theta(v-u) \rangle \ge \mu \theta^2 | v - u |^2.
\end{equation*}
Thus, for $\theta \in (0,1),$
\begin{equation*}
     \langle D \J(u + \theta(v-u)) - D \J(u), v-u \rangle \ge \mu \theta | v - u |^2. 
\end{equation*}
This implies that
\begin{equation*}
    \int_0^1 \langle D \J\bigl(u + \theta(v-u)\bigr)-D \J\bigl(u\bigr), v-u\rangle d\theta \ge \mu | v- u |^2 \int_0^1 \theta d\theta = \frac{\mu}{2} |v - u |^2. 
\end{equation*}
Substituting this bound into \eqref{eq:auxiliaryeq}, we conclude that
\begin{equation*}
    \J(v) \ge \J(u) + \langle D \J(u), v - u \rangle + \frac{\mu}{2} | v - u |^2,
\end{equation*}
establishing $\mu$-strong convexity.
\end{proof}

We will also use the following definition of $L$-smoothness.

\begin{definition}
    Function $\J \in C^1(\R^d,\R)$ is called $L$-smooth\index{$L$-smooth} if for all $u,v \in \R^d$ it holds that 
    \begin{equation*}
        | D \J(u) - D\J(v) | \le L | u - v|.
    \end{equation*}
\end{definition}

The following result provides a useful equivalent characterization of $L$-smoothness.

\begin{theorem}\label{th:smoothcharacterization}
  Let $\J \in C^1(\R^d,\R)$ be convex and let $L>0$. The following are equivalent:
    \begin{enumerate}
        \item $\J$ is $L$-smooth.
        \item For all $u,v \in \R^d$ it holds that
        \begin{equation}\label{eq:equivalentsmooth}
            \J(v) \le \J(u) + \langle D \J(u), v - u \rangle + \frac{L}{2} | v - u |^2.
        \end{equation}
    \end{enumerate}
\end{theorem}
\begin{proof}
Suppose that $\J$ is $L$-smooth.
As in the proof of Theorem \ref{thm:convexcharacterization} we have that for $u,v \in \R^d$
\begin{equation}
    \J(v) - \J(u) =  \langle D\J(u), v - u\rangle + \int_0^1 \langle D \J( u + \theta(v-u)) - D\J(u), v - u \rangle d\theta. 
\end{equation}
Applying the Cauchy--Schwarz inequality and using $L$-smoothness gives 
\begin{align*}
    \J(v) - \J(u)  &\le   \langle D\J(u), v - u\rangle + \int_0^1 L \theta | v - u |^2  d\theta \\
     &=    \langle D\J(u), v - u\rangle + \frac{L}{2} | v - u |^2 .
\end{align*}
Conversely, suppose that \eqref{eq:equivalentsmooth} holds. Fix
$u,v\in\R^d$ and define
\begin{equation*}
\phi(x):=\J(x)-\langle D\J(v),x\rangle.
\end{equation*}
Since $\J$ is convex and $D\phi(v)=0$, point $v$ is a minimizer of
$\phi$. Set
\begin{equation*}
w:=u-\frac{1}{L}D\phi(u).
\end{equation*}
Subtracting a linear function leaves the first-order Taylor remainder
unchanged, so \eqref{eq:equivalentsmooth} also holds for $\phi$ with
the same constant $L$.
Applying \eqref{eq:equivalentsmooth} to $\phi$ gives
\begin{equation*}
\phi(w)
\leq
\phi(u)-\frac{1}{2L}|D\phi(u)|^2.
\end{equation*}
Since $\phi(v)\leq\phi(w)$, it follows that
\begin{equation*}
\J(u)-\J(v)-\langle D\J(v),u-v\rangle
\geq
\frac{1}{2L}|D\J(u)-D\J(v)|^2.
\end{equation*}
Swapping $u$ and $v$ and adding the resulting inequalities yields
\begin{equation*}
\langle D\J(u)-D\J(v),u-v\rangle
\geq
\frac{1}{L}|D\J(u)-D\J(v)|^2.
\end{equation*}
The Cauchy--Schwarz inequality now gives
\begin{equation*}
|D\J(u)-D\J(v)|\leq L|u-v|,
\end{equation*}
establishing $L$-smoothness.
\end{proof}

\begin{remark}\label{rem:relationship}
    Let $\J \in C^1(\R^d,\R)$ be  $\mu$-strongly convex and $L$-smooth. We then claim that $\mu \le L.$
    Indeed, let $u, v \in \R^d$ with $u \neq v.$ Then, using the characterization of $\mu$-strong convexity in 
    Theorem \ref{thm:convexcharacterization}, along with Cauchy--Schwarz and the definition of $L$-smoothness, 
    we deduce that
    \begin{align*}
     \mu | u - v|^2 & \le    \langle D \J(u) - D \J(v) , u - v \rangle \\
     & \le | D\J(u) - D \J(v) | |u -v| \\
     & \le L | u - v|^2.
    \end{align*}
    This shows that $\mu \le L.$ If $\J \in C^2(\R^d, \R)$ then $\mu$ and $L$ provide lower and upper bounds on the eigenvalues of the Hessian: it can be shown that $\mu I \preceq D^2 \J(u) \preceq L I $ for all $u \in \R^d.$
\end{remark}

We conclude this section with the definition of a convex set.

\begin{definition} \label{def:convex_set}
    A set $S$ in a vector space is \emph{convex}\index{convex!set} if, for all $u, v\in S$ and for all $\theta\in [0, 1]$, we have that
    \begin{equation*}
         \theta u+(1-\theta)v \in S.
    \end{equation*}
\end{definition}

\section{Gradient Descent}\index{gradient descent}
\label{sec:grad}

\subsection{Deterministic Gradient Descent}\index{gradient descent!deterministic}

Our starting point is the idea of a gradient flow:\index{gradient!flow}

\begin{definition} Given function $\J \in C^1(\R^\du,\R^+)$,
\emph{gradient flow}\index{gradient!flow} refers to the family of solutions 
generated by the differential equation as the initial condition $u_0$ is varied:
\begin{equation}
\label{eq:gf}
    \frac{du}{dt}=-\nabla \J(u), \quad u(0)=u_0.
\end{equation}
\end{definition}

\begin{theorem}
Assume that $\J \in C^2(\R^\du,\R^+)$ has compact level sets.
Then a solution to \eqref{eq:gf} exists for all initial conditions $u_0 \in \R^\du$ and for all positive times $t.$ Moreover, 
\begin{equation}
\label{eq:gf_obj}
\frac{d}{dt} \J\bigl(u(t))=-|\nabla \J\bigl(u(t)\bigr)|^2.
\end{equation}
\end{theorem}

\begin{proof} Since $\J$ is $C^2$, vector field $\nabla \J$ is $C^1$ and hence a local solution exists for every
initial condition $u_0.$ Straightforward calculation reveals that,
for all $u_0 \in \R^d$, and for all $t \ge 0$ such that a solution exists, \eqref{eq:gf_obj} holds.
Since $\J$ is bounded from below, is $C^2$ and has compact level sets, the fact that $\J(u(t)) \le \J(u_0)$ 
may be used to establish that $|u(t)|<\infty$ for all $t \ge 0$ and hence that, for all initial  conditions $u_0$, a solution does indeed exist for all positive times $t$. 
\end{proof}

Equation \eqref{eq:gf_obj} implies that
$\J(\cdot)$ decreases along trajectories, unless the trajectory is at a critical point
of $\J(\cdot).$ As a consequence, time-discretization
of equation \eqref{eq:gf} provides the basis of algorithms\index{algorithm!gradient descent} to minimize $\J.$ This is motivation for the following:

\begin{definition} Given function $\J \in C^1(\R^\du,\R)$, 
    \emph{gradient descent} refers to
the iterative generation of sequence $\{u_j\}_{j \in \Z^+}$
from a sequence of step-sizes $\{\alpha_j\}_{j \in \Z^+}$ by picking an $u_0
\in \R^\du$ and then iterating as follows:
    \begin{equation}
    \label{eq:gfd}
        u_{j+1} = u_j - \alpha_j \nabla \J(u_j), \qquad j \in \Z^+.
    \end{equation}
\end{definition}

\begin{remark}
Note that equation \eqref{eq:gfd} corresponds to Euler discretization of \eqref{eq:gf}.
In the simplest case, $\alpha_j = \alpha$ is some fixed constant time-step.
Often, however, the time-step sequence $\{\alpha_j\}_{j \in \Z^+}$ is determined
online as the algorithm progresses, with $\alpha_j$ depending on $\{u_i\}_{i=0}^{j}.$ 
In either case, knowing how to choose the time-step to ensure that $\J$ decreases 
at each step, unless it is at a critical point, requires knowledge of the properties of the vector field $D\J.$ 
\end{remark}

The following theorem establishes linear convergence\index{convergence!linear} of gradient descent with a specific fixed step-size $\alpha$. We will assume that the objective $\J$ is $L$-smooth and $\mu$-strongly convex. Recall that, as discussed in Remark \ref{rem:relationship}, $\mu \le L.$

\begin{theorem}\label{thm:GD}
Suppose that $\J \in C^1(\R^d,\R)$ is $\mu$-strongly convex and $L$-smooth with constants $0<\mu \le L$, and has minimizer $u^\star \in \R^d.$ Consider the gradient descent algorithm with step-size $\alpha_j \equiv \alpha:= 1/L.$ Then, at iteration $j \ge 1,$
\begin{equation*}
    \J(u_j) - \J(u^\star) \le \Bigl( 1 - \frac{\mu}{L} \Bigr)^j \bigl( \J(u_0) - \J(u^\star) \bigr).
\end{equation*}
\end{theorem}
\begin{proof}
First we show that $\mu$-strong convexity implies that, for all $u \in \R^d,$ 
    \begin{equation}\label{eq:consequence2}
      | D\J(u) |^2 \ge 2 \mu\bigl( \J(u) - \J(u^\star)\bigr).
      \end{equation}
    Indeed, using $\mu$-strong convexity, we have
    $$\J(u^\star) \ge \J(u) + \langle D\J(u), u^\star- u \rangle +  \frac{\mu}{2} | u^\star - u |^2.$$
    Thus, by the Cauchy--Schwarz inequality, we deduce that 
    \begin{align*}
        \J(u) - \J(u^\star) &\le \langle D \J(u), u - u^\star \rangle - \frac{\mu}{2} | u - u^\star |^2 \\
        & \le | D\J(u) |  | u - u^\star | - \frac{\mu}{2} |u- u^\star |^2.
    \end{align*}
    Let $\phi(t):= |D\J(u)| t - \frac{\mu}{2} t^2.$ The maximum of $\phi$ occurs at $t^\star = | D \J(u)|/\mu.$ Moreover, $\phi(t^\star) = \frac{1}{2\mu} | D \J(u)|^2.$ This implies \eqref{eq:consequence2}.

    Next, using the characterization of $L$-smoothness in Theorem \ref{th:smoothcharacterization} and that $u_{j+1} = u_j - \alpha D \J(u_j)$ with $\alpha = 1/L$ we have 
    \begin{align*}
        \J(u_{j+1}) &\le \J(u_j) - \alpha | D \J(u_j) |^2 + \frac{L \alpha^2}{2} | D \J(u_j) |^2 \\
        & = \J(u_j) - \alpha \Bigl( 1 - \frac{L\alpha}{2} \Bigr) | D \J(u_j) |^2  \\
        & = \J(u_j) - \frac{1}{2L} | D \J(u_j) |^2.
    \end{align*}
    By \eqref{eq:consequence2} we have that
    $$| D\J(u_j) |^2 \ge 2 \mu \bigl( \J(u_j) - \J(u^\star) \bigr).$$
    Therefore we conclude that
      \begin{align*}
        \J(u_{j+1}) - \J(u^\star) &\le \J(u_j) - \J(u^\star) - \frac{\mu}{L} \bigl( \J(u_j) - \J(u^\star) \bigr) \\
        &= \Bigl(1 - \frac{\mu}{L}\Bigr) \bigl( \J(u_j) - \J(u^\star) \bigr),
    \end{align*}
    and the result follows by induction.
\end{proof}

\begin{remark}
    Under the conditions of the previous theorem, it is also possible to show linear convergence of the optimization iterates. More precisely, it holds that $| u_j - u^\star |^2 \le (1 - \mu/L)^j | u_0 - u^\star |^2. $
\end{remark}

 \begin{remark} For convex functions with Lipschitz gradients, there are modifications of gradient descent that have better convergence rates than the one stated above. For instance, Nesterov's accelerated gradient descent achieves an objective error of order $j^{-2}$, which is optimal for this class of functions within an appropriate class of first-order methods.\index{gradient descent!Nesterov's accelerated} We will not discuss these other methods here but provide references in the bibliography.
 \end{remark}

\subsection{Stochastic Gradient Descent}\index{gradient descent!stochastic}\label{ssec:SGD}

Here we consider the minimization of an objective function $\J:\R^\du \to \R$ that is defined by an expectation of a function $F\colon \R^d \times \R^{d_z} \rightarrow \R$. That is, 
\begin{equation}
\label{eq:fF}
\J(u) = \int F(u,z) \, d\zeta(z),   
\end{equation}
where $\zeta$ denotes the probability density function of the random variable $z \in \R^{d_z}$. While explicit evaluations of $\J$ or its gradients are challenging as a result of the expectation, we assume that it is possible to evaluate $D_u F(u,z)$ for any realization of the random variable $z$.

Assuming that $\zeta$ is replaced by an empirical\index{empirical} approximation 
\begin{equation*}
    \zeta^N = \frac{1}{N}\sum_{n=1}^N \delta_{z^{(n)}},
\end{equation*}
with the $z^{(n)}$ sampled i.i.d. from $\zeta,$ then $\J$ is replaced by
\begin{equation}
\label{eq:fF2}
\J^N(u) = \frac{1}{N}\sum_{n=1}^N F(u,z^{(n)}).
\end{equation}
The gradient of $\J^N$ is then given by 
\begin{equation}\label{eq:grad_full}
    D \J^N(u) = \frac{1}{N}\sum_{n =1}^N D_u F(u,z^{(n)}).
\end{equation}
This expression for $D \J^N(u)$ has the interpretation as a Monte Carlo\index{Monte Carlo} estimator of
the gradient  $D\J(u)$ defined by \eqref{eq:fF}.

For large $N$, the evaluation of the sum \eqref{eq:grad_full} may be too expensive. Instead, one can pick $M < N$ i.i.d. samples from $\zeta^N$; this corresponds to picking $M$ random indices $\{\omega^{(m)}\}_{m=1}^M$, where the elements of this
set $\omega^{(m)}$ are sampled uniformly with replacement from $1,\ldots,N$. Then the gradient can be approximated as
\begin{equation*}
    D \J^N(u) \approx \frac{1}{M}\sum_{m=1}^M D_u F(u,z^{(\omega^{(m)})}).
\end{equation*}
This is known as a \emph{mini-batch}.\index{mini-batch} Carrying out gradient descent with (for example) an independent mini-batch is a widely used methodology for minimizing
$\J^N$ given by \eqref{eq:fF2}:

\begin{definition} Given function  $F \in C^{1,0}(\R^\du \times \R^{\dz}, \R),$ 
    \emph{stochastic gradient descent} for minimizing $\J^N$ defined
    by \eqref{eq:fF2}, refers to
the iterative generation of sequence $\{u_j\}_{j \in \Z^+}$
from a sequence of step-sizes $\{\alpha_j\}_{j \in \Z^+}$
and $j$-indexed sets $\{\omega_j^{(m)}\}_{m=1}^M$ with points  sampled uniformly with replacement from $1,\ldots,N$, independently for each $j$,
by picking any $u_0 \in \R^\du$ and then iterating as follows:
$$u_{j+1} = u_{j} - \frac{\alpha_j}{M}\sum_{m=1}^M D_u F(u_j,z^{(\omega^{(m)}_j)}).$$
\end{definition}

\begin{remark}
    Even if the full gradient with respect to the empirical\index{empirical} measure \eqref{eq:grad_full} can be feasibly computed, the stochasticity in stochastic gradient descent can be helpful in escaping from local minima for non-convex objective functions; in the context of optimizing neural networks
    it is also sometimes advocated as a method to avoid overfitting.
\end{remark}

Although stochastic gradient descent is now widely used in the context of training neural networks, its history can be traced back to the 1950s where it was developed as a method to minimize intractable objective functions $\J(\cdot)$ formulated through an integral, as in \eqref{eq:fF}. In this context we introduce the algorithm
\begin{equation}
\label{eq:RM}
u_{j+1} = u_j - \alpha_j g_j,
\end{equation}
where $g_j$ is a random estimator of the gradient of $\J(\cdot).$ For example we might take
$$g_j= D_u F(u_j,z_j), \quad z_j \sim \zeta\,\, \index{i.i.d.}\text{i.i.d.}.$$
The following theorem shows that the stochastic gradient descent algorithm \eqref{eq:RM}, with a constant step-size $\alpha$, can recover the desired minimum up to an error determined by the variance of the gradient estimates. This is followed by a second theorem which removes the error by using a decreasing step-size sequence $\alpha_j$.

\begin{theorem}\label{thm:SGD}
    Suppose that $\J \in C^1(\R^d,\R)$ is defined by \eqref{eq:fF}, is $\mu$-strongly convex and $L$-smooth for constants $0<\mu \le L$, with minimizer $u^\star$. Consider the stochastic gradient descent algorithm \eqref{eq:RM}
    with constant step-size $\alpha_j \equiv \alpha := 1/L.$
    Assume that $g_j$ provides an unbiased estimate of the gradient with bounded variance:
     $\Expect [ g_j | u_j] = D \J(u_j)$
    and $\Expect [ | g_j - D \J(u_j) |^2 | u_j] \le \sigma^2. $ Then
 \begin{equation*}
     \Expect [ \J(u_j) - \J(u^\star) ] \le \Bigl( 1 - \frac{\mu}{L} \Bigr)^j \bigl( \J(u_0) - \J(u^\star) \bigr) + \frac{\sigma^2}{2\mu}.
 \end{equation*}
\end{theorem}
\begin{proof}
    As in the proof of Theorem \ref{thm:GD}, it follows from $L$-smoothness that 
    \begin{equation*}
        \J(u_{j+1}) \le \J(u_j) - \frac{1}{L} \langle D \J(u_j), g_j \rangle + \frac{1}{2L} | g_j |^2.
    \end{equation*}
    Denote by $\Expect_j$ expectation conditioned on $u_j.$ Then,
    \begin{equation*}
        \Expect_j \J(u_{j+1}) \le \J(u_j) - \frac{1}{L} | D \J(u_j) |^2 + \frac{1}{2L} \Expect_j | g_j |^2.
    \end{equation*}
    Now notice that 
    \begin{align*}
        \Expect_j | g_j |^2 &= | D \J(u_j) |^2 + \Expect_j| g_j - D \J(u_j) |^2 \\
        & \le | D \J(u_j) |^2 + \sigma^2.
    \end{align*} 
    Hence,  
    \begin{align*}
        \Expect_j \J(u_{j+1}) \le \J(u_j) - \frac{1}{2L} | D \J(u_j) |^2 + \frac{\sigma^2}{2L}.
    \end{align*}
    Using inequality \eqref{eq:consequence2}, we have 
    $$ | D \J(u_j) |^2 \ge 2 \mu \bigl( \J(u_j) - \J(u^\star) \bigr). $$
    Therefore, 
    $$ \Expect_j[ \J(u_{j+1}) - \J(u^\star) ] \le \Bigl( 1 - \frac{\mu}{L} \Bigr) \bigl( \J(u_j) - \J(u^\star) \Bigr) + \frac{\sigma^2}{2L}. $$
    Taking expectation over $u_j$ and using induction yields the result.
\end{proof}

The next result shows convergence of stochastic gradient descent with variable step size. 

\begin{theorem}
Suppose that $\J\in C^1(\R^d,\R)$ is $\mu$-strongly convex and
$L$-smooth for constants $0<\mu\leq L$, with minimizer $u^\star$.
Assume that $g_j$ is an unbiased estimator of the gradient with bounded
variance:
$\Expect[g_j | u_j]=D\J(u_j),$ and
$\Expect\bigl[|g_j-D\J(u_j)|^2 | u_j\bigr]\leq\sigma^2.$
Consider the stochastic gradient descent algorithm \eqref{eq:RM} with
step-size
\begin{equation*}
\alpha_j=\frac{2}{2L+\mu j}.
\end{equation*}
Then, for $\kappa:=\frac{2L}{\mu},$
\begin{equation*}
\Expect\bigl[\J(u_j)-\J(u^\star)\bigr]
\leq
\frac{c}{j+\kappa}
\end{equation*}
for some constant $c$ independent of $j$, but depending on $L, \mu, \sigma$ and $u_0.$
\end{theorem}

\begin{proof}
Let $\Expect_j$ denote expectation conditioned on $u_j$. By
$L$-smoothness and Theorem \ref{th:smoothcharacterization} we have
$$\J(u_{j+1}) \le \J(u_j) + \langle D \J(u_j), u_{j+1} - u_j \rangle + \frac{L}{2} | u_{j+1} - u_j |^2.$$
Using the update $u_{j+1}=u_j-\alpha_jg_j$, the fact that $g_j$ is an unbiased estimator of the gradient,
and subtracting $\J(u^\star)$ from both sides of the inequality, we obtain 
\begin{equation*}
\Expect_j\J(u_{j+1}) - \J(u^\star)
\leq
\J(u_j) - \J(u^\star)
-\alpha_j|D\J(u_j)|^2
+\frac{L\alpha_j^2}{2}\Expect_j|g_j|^2.
\end{equation*}
The assumptions on $g_j$ imply that
\begin{equation*}
\Expect_j|g_j|^2
\leq
|D\J(u_j)|^2+\sigma^2.
\end{equation*}
Since $\alpha_j\leq 1/L$, it follows that
\begin{equation*}
\Expect_j\J(u_{j+1}) - \J(u^\star)
\leq
\J(u_j) - \J(u^\star)
-\frac{\alpha_j}{2}|D\J(u_j)|^2
+\frac{L\alpha_j^2\sigma^2}{2}.
\end{equation*}
Let $e_j:=\Expect\bigl[\J(u_j)-\J(u^\star)\bigr]$.
Using \eqref{eq:consequence2} and taking expectations gives
the recurrence
\begin{equation*}
e_{j+1}
\leq
(1-\mu\alpha_j)e_j+\frac{L\alpha_j^2\sigma^2}{2}.
\end{equation*}
Define $b:=\frac{2L\sigma^2}{\mu^2}.$ Then
\begin{equation*}
e_{j+1}
\leq
\left(1-\frac{2}{j+\kappa}\right)e_j
+\frac{b}{(j+\kappa)^2}.
\end{equation*}
Note that, since $\mu \le L$, the coefficient of $e_j$ is non-negative. Choose $c\geq\max\{\kappa e_0,b\}$. Then $e_0 \le c/\kappa$ and the postulated bound holds for $j=0.$
Using $b \le c$ straightforward induction shows that, for all $j \in \mathbb{Z}^+$, 
\begin{equation*}
e_j\leq\frac{c}{j+\kappa} \Rightarrow e_{j+1}\leq\frac{c}{j+1+\kappa}
\end{equation*}
The desired result follows.
\end{proof}


\section{Newton and Gauss--Newton}
\label{sec:NGN}

\subsection{Newton's Method}\index{Newton's method}

Newton's method is a common root-finding algorithm which, under suitable conditions and sufficiently close to a solution, benefits from quadratic convergence\index{convergence!quadratic}. This quadratic convergence improves on the
linear convergence\index{convergence!linear} of fixed point methods based on the contraction mapping principle.\index{contraction mapping principle} Newton's method can also be applied to optimization, as we now show, leading to a methodology that improves on the linear convergence\index{convergence!linear} of gradient descent\index{gradient descent}.

\begin{definition} Given function $\J \in C^2(\R^\du,\R)$ 
with invertible Hessian,
    \emph{Newton's method} refers to
the iterative generation of sequence $\{u_j\}_{j \in \Z^+}$
by picking any $u_0 \in \R^\du$ and then iterating as follows:
    \begin{equation}\label{eq:newton}
        u_{j+1} = u_j - \bigl(D^2 \J(u_j)\bigr)^{-1}D\J(u_j).
    \end{equation}
\end{definition}

Note that critical points of $\J$ are fixed points of the Newton iteration.
Newton's method proceeds by optimizing successive quadratic approximations to $\J$, as the following theorem shows.
\begin{theorem}
 Consider the Newton iteration \eqref{eq:newton}. If $\J$ is convex, then $u_{j+1}$ 
 is the unique minimizer of the quadratic approximation of $\J$ around $u_j$.
\end{theorem}
\begin{proof}
    We first approximate $\J$ around $u_j$ by its second-order Taylor expansion:
    \begin{equation}\label{eq:taylor_quad}
        \J(u_j + v) \approx \J(u_j) + \langle \nabla \J(u_j), v\rangle + \frac{1}{2}
        \langle v,D^2 \J(u_j) v \rangle,
    \end{equation}
    where $v$ is any vector and the approximation is good when $v$ is small. Since $\J$ is convex, the Hessian is positive semidefinite, and the right-hand side of \eqref{eq:taylor_quad} is convex as a function of $v$, meaning that it can be minimized by setting the derivative to $0$. Differentiating the
    approximation of $\J(u_j + v)$ given in \eqref{eq:taylor_quad} 
    with respect to $v$, and then setting this derivative to $0$, gives
    \begin{equation*}
        D\J(u_j) + D^2 \J(u_j) v = 0,
    \end{equation*}
    and thus
    \begin{equation*}
        v = -\bigl(D^2 \J(u_j)\bigr)^{-1}D\J(u_j).
    \end{equation*}
\end{proof}

\subsection{Gauss--Newton}\index{Gauss--Newton}
Let $\mathsf{G} \in C^1(\R^d,\R^m)$ and let $\mathsf{G}=(\mathsf{G}_1, \ldots, \mathsf{G}_m)$ where 
$\mathsf{G}_i\in C^1(\R^d,\R),$ $1 \le i \le m$. Consider 
the nonlinear least squares optimization problem 
\begin{subequations}
\label{eq:nnlsq}
\begin{align}
    \J(u) &= \frac12  |\mathsf{G}(u)|^2 = \frac12 \sum_{i=1}^m \mathsf{G}_i(u)^2,\\
    u^\star &\in  \argmin_{u \in \R^d} \J(u).
\end{align}
\end{subequations}

\begin{definition}
Given function $\mathsf{G} \in C^1(\R^\du,\R^m)$ 
such that $\bigl(D\mathsf{G}(u)\bigr)^\top \bigl(D\mathsf{G}(u)\bigr)$
is invertible for all $u \in \R^d$,
    the \emph{Gauss--Newton method} for minimizing nonlinear least squares problems of the form \eqref{eq:nnlsq}
     refers to the iterative generation of sequence 
     $\{u_j\}_{j \in \Z^+}$
by picking any $u_0 \in \R^\du$ and then iterating as follows:
    \begin{align*}
        u_{j+1} &= u_j -\Bigl(\bigl(D\mathsf{G}(u_j)\bigr)^\top \bigl(D \mathsf{G}(u_j)\bigr)\Bigr)^{-1}\bigl(D\mathsf{G}(u_j)\bigr)^\top \mathsf{G}(u_j). 
    \end{align*}
\end{definition}

Gauss--Newton results from applying Newton's method \eqref{eq:newton} to (\ref{eq:nnlsq}a), and neglecting the second-order terms of $\mathsf{G}$ in the Hessian of $\J$. To make this connection precise, suppose additionally that
$\mathsf{G}\in C^2(\R^d,\R^m)$. Notice that
\begin{equation*}
    D\J(u) =  \bigl(D\mathsf{G}(u)\bigr)^\top \mathsf{G}(u),
\end{equation*}
and that, entry-wise,
\begin{equation}\label{eq:hessian_J}
    \bigl(D^2\J(u)\bigr)_{ab} = \bigl(D \mathsf{G}(u)^\top D \mathsf{G}(u)\bigr)_{ab} + \sum_{i=1}^m \mathsf{G}_i(u) \bigl(D^2 \mathsf{G}_i(u)\bigr)_{ab}.
\end{equation}
Substituting these expressions into Newton's method \eqref{eq:newton} and setting the second term in \eqref{eq:hessian_J} to $0$, recovers the Gauss--Newton method. Making this approximation is computationally expedient as it avoids computation of the Hessians of the $\mathsf{G}_k.$ 
It is also a good approximation near a minimizer $u^\star$ of $\J$
with small residual $\mathsf{G}(u^\star)$, since the neglected term in
\eqref{eq:hessian_J} is then small as the iterates approach $u^\star$.

\section{Ensemble Kalman Inversion}\index{Kalman inversion!ensemble}
\label{sec:EKI}
This section discusses \emph{Ensemble Kalman inversion}\index{Kalman inversion!ensemble} (EKI) as a general-purpose ensemble Kalman method for nonlinear least squares. Compared to the variants on gradient descent and Newton methods studied in the previous two sections, EKI has the advantage of being \emph{derivative-free}. For this reason, EKI is widely used in inverse problems with complex forward models; indeed, as its name suggests, EKI was first developed as a method for numerical solution of inverse problems, and only more recently formulated as a general-purpose and broadly useful optimization algorithm. 

Consider the cost function
\begin{equation*}
    \J(u) = \frac{1}{2} | y- G(u)|_{\Gamma}^2,
\end{equation*}
for some (possibly) nonlinear function $G:\R^d \to \R^k$, point $y \in \R^k$ and  matrix $\Gamma \succ 0$. Notice that this nonlinear least squares objective is of the form in equation (\ref{eq:nnlsq}a); this can be seen by setting $m:=k$ and $\mathsf{G}(u):= y - G(u).$
Notice further that in the language of Chapter \ref{ch1}
the objective $\J(u)$ is, up to an additive constant, the negative log-likelihood function determined by inverse problem \eqref{eq:jc0} with Gaussian noise $\eta \sim \Nc(0,\Gamma);$ that is, up to additive constant it holds that   $\J(u) = - \log \Prob(y|u) = - \log \like(y|u)$. Therefore, minimizing $\J(u)$ corresponds to maximizing the likelihood, and this interpretation underpins the original derivation of the method. 

EKI\index{Kalman inversion!ensemble} involves first drawing an initial set of $\Sam$ particles from a pre-specified probability measure; to be concrete
we will employ a Gaussian with a given mean $m_0$ and covariance $C_0$. Thus
\begin{equation*}
    u_0^{(\sam)} \sim \mathcal{N}(m_0, C_0)\quad\text{i.i.d.}, \quad \sam=1,\ldots,\Sam.
\end{equation*}

EKI\index{Kalman inversion!ensemble} then evolves these particles for
$j=0,\ldots,J-1$. At iteration $j$, we first compute
\begin{align*}
\bar{u}_j
&=
\frac{1}{\Sam}\sum_{\sam=1}^{\Sam}u_j^{(\sam)},\\
\bar{G}_j
&=
\frac{1}{\Sam}\sum_{\sam=1}^{\Sam}G(u_j^{(\sam)}),\\
C_j^{ug}
&=
\frac{1}{\Sam}\sum_{\sam=1}^{\Sam}
\bigl(u_j^{(\sam)}-\bar{u}_j\bigr)
\otimes
\bigl(G(u_j^{(\sam)})-\bar{G}_j\bigr),\\
C_j^{gg}
&=
\frac{1}{\Sam}\sum_{\sam=1}^{\Sam}
\bigl(G(u_j^{(\sam)})-\bar{G}_j\bigr)
\otimes
\bigl(G(u_j^{(\sam)})-\bar{G}_j\bigr),\\
K_j
&=
C_j^{ug}\bigl(C_j^{gg}+\Gamma\bigr)^{-1}.
\end{align*}
The particles are then updated according to
\begin{align*}
u_{j+1}^{(\sam)}
&=
u_j^{(\sam)}
+
K_j\bigl(y+\eta_{j+1}^{(\sam)}-G(u_j^{(\sam)})\bigr),
\qquad \sam=1,\ldots,\Sam,\\
\eta_{j+1}^{(\sam)}
&\sim\Nc(0,\Gamma)
\qquad\text{i.i.d.}
\end{align*}



\begin{remark}
The algorithm produces an ensemble $\{u_j^{(\sam)}\}_{\sam=1}^N$ that, for each $j$, remains in the
linear span of the initial ensemble $\{u_0^{(\sam)}\}_{\sam=1}^N.$ Thus the algorithm may be viewed
as seeking to minimize the objective function over a finite-dimensional subspace. In the case
of linear $G(\cdot)$ the methodology is amenable to a complete theoretical analysis,
demonstrating sublinear convergence\index{convergence!sublinear}. However
variants on it, including Tikhonov regularization\index{regularizer!Tikhonov}, restore linear convergence.
\index{convergence!linear} Continuous-time limits may also be derived, 
and then analyzed analogously in the linear setting. It is
empirically observed that EKI\index{Kalman inversion!ensemble} often provides good approximations of the minimizer of the loss function in the nonlinear setting, and does so for relatively small $J$ and $N$. See citations in the bibliography Section~\ref{sec:OB}.
\end{remark}

\begin{remark} \label{rem:ensg}
As the name suggests, there is a link between EKI\index{Kalman inversion!ensemble} and the 
ensemble Kalman filter\index{Kalman filter!ensemble} (EnKF) for data assimilation\index{data assimilation}, 
defined in Subsection~\ref{ssec:enkf}. One way of seeing this connection is via the formulation of certain maximum likelihood optimization problems as filtering\index{filtering} problems. Another way of viewing the connection is
to note that, at every iteration, EKI\index{Kalman inversion!ensemble} approximates the solution to a nonlinear inverse problem using an ensemble approximation to the solution of a linear Gaussian inverse problem (Example \ref{ex:loss-l2}) and formally replacing the linear forward model with the nonlinear one. We also note that in the setting 
of linear $G$ the EKI methodology is connected to a preconditioned gradient descent\index{gradient descent!preconditioned}. Finally we comment that ensemble approximations of other optimization methods, such as Gauss--Newton, are also in use, popular because they avoid the need to compute derivatives.
These connections are detailed in the bibliography Section~\ref{sec:OB}.
\end{remark}

\section{Expectation Maximization}\label{sec:142}\index{expectation maximization}
The \index{expectation maximization}expectation maximization (EM) 
 algorithm\index{algorithm!EM} is a general-purpose approach to \index{maximum likelihood estimation}maximum likelihood estimation with \index{latent variable}latent variables. 
 Let $u \in \Theta \subset \R^p$ be the parameter of interest, $y \in \R^k$ the data, and $z \in \R^{d_z}$ a latent variable. We seek to compute the maximum likelihood estimator of parameter $u$ given data $y;$ that is, we seek to maximize 
 \begin{equation}\label{eq:objectiveEM}
     \Prob(y|u) = \int_{\R^{d_z}} \Prob(y,z |u) \, dz.  
 \end{equation}
The EM algorithm is useful when we cannot readily evaluate the likelihood $\J(u) = \Prob(y|u)$ or compute its gradient, but evaluating the  \emph{full likelihood} $\Prob(y,z|u)$ is possible. Variable $z$ is sometimes referred to as \emph{missing data}.\index{missing data}

\begin{example} \label{ex:1515}
Recall the \index{stochastic dynamics model}stochastic dynamics and \index{data model}observation models 
summarized in equations \eqref{eq:SSM1}:
\begin{align*}
        \vd_{j+1} &= \Psi_\vartheta(\vd_j) + \xid_j, && \xid_j \sim \Nc(0, \Sigma) \index{i.i.d.} \text{ i.i.d.},  \\
        \yd_{j+1} &= h(\vd_{j+1}) + \etad_{j+1}, && \etad_{j+1} \sim \Nc(0, \Gamma) \index{i.i.d.}\text{ i.i.d.} 
\end{align*}
Recall that we refer to $\{\vd_j\}$ as the signal\index{signal} and $\{\yd_j\}$ 
as the observation\index{observation} sequence. Here $\vd_0 \sim \Nc(m_0,C_0)$ 
and we assume the independence structure summarized in Assumption \ref{a:noise}. Take $u=\vartheta$,
$z = \{\vd_0, \ldots, \vd_J\}$ and $y = \{\yd_1, \ldots, \yd_J\}.$ Consider the problem of
computing the maximum likelihood estimator for $u$, given observations $y$; we assume that $z$ is not known to us. 
This has precisely the structure
\eqref{eq:objectiveEM}; in particular $\Prob(y,z |u)$ can be written down in closed form, but in general
$\Prob(y|u)$ cannot be. To solve the maximum likelihood estimation problem it is necessary to obtain information about the missing data\index{missing data}
$z$ -- here, the signal.\index{signal} See Section \ref{sec:DAlearningmodels_setting} for application of the ideas in this section 
to this specific parameter estimation problem for $\vartheta.$
\end{example}

The objective function in \eqref{eq:objectiveEM} for the EM algorithm resembles that in \eqref{eq:fF} for stochastic gradient descent\index{gradient descent!stochastic}, where intractable objective $\J(u)$ is defined by averaging tractable objective $F(u,z)$ over auxiliary variables $z \in \R^{d_z}.$
  Notice, however, that in this section we depart from the convention in previous sections and formulate the optimization problem in terms of \emph{maximizing} rather than \emph{minimizing} objective $\J.$

A key component in the derivation of the EM algorithm is encapsulated in the following definition:

\begin{definition}\label{d:elbo} The \emph{evidence lower bound}\index{evidence! lower bound} ($\ELBO$) of a probability density $q$ in $\R^{d_z}$ with respect to an unnormalized density $\widetilde{\post}$ in $\R^{d_z}$ is given by 
$$\ELBO(\widetilde{\post},q) = \mathbb{E}^q[\log \widetilde{\post}] - \mathbb{E}^q[\log q].$$
\end{definition}

\begin{remark}
Recall the KL divergence introduced in Definition \ref{def:KL}. \index{divergence!Kullback--Leibler}
Note that if $\post$ is a normalized probability density then
$$\ELBO(\post,q)=-\dkl(q \| \post).$$
\end{remark}

It is useful to define
\begin{subequations} \label{eq:elqu}
\begin{align}
\J(u) :=& \Prob(y|u),\\   
\mathcal{L}(q, u) :=& \ELBO\bigl(\Prob(y, \cdot\,  | u), q\bigr).
\end{align}
\end{subequations}
Note that $\Prob (y, z| u)$ is an unnormalized density for variable $z$ alone, with $(y,u)$ fixed; however, it is normalized with respect to the joint distribution on the pair $(y,z)$.
 The following \index{likelihood} result connects the joint distribution $\Prob(y,z|u)$ to $\Prob(y|u)$
 and is therefore useful for maximum likelihood estimation, in the context of the identity \eqref{eq:objectiveEM}.

\begin{theorem}\label{th:EMlikecharacterization}
Let $q$ be a probability density function over latent variable $z \in \R^{d_z}.$
It holds that 
\begin{equation}
\label{eq:identity}
\log \J(u) = \mathcal{L}(q, u) + \dkl \bigl( q \| \Prob(\cdot\, |y,u)\bigr).
\end{equation}
\end{theorem}
\begin{proof}
Using the definition of $\mathcal{L}(q, u),$ product rule, and the Definition \ref{def:KL} of the \index{divergence!Kullback--Leibler}KL divergence, we have
\begin{align*}
\mathcal{L}(q, u) &= \ \int_{\R^{d_z}} \log\biggl( \frac{\Prob (y, z| u)}{q(z)} \biggr)  q(z)  \, dz \\
& = \ \int_{\R^{d_z}}  \log \biggl( \frac{\Prob(z|y,u)}{q(z)} \biggr) q(z) \, dz + \ \int_{\R^{d_z}}  \log \Prob(y|u) q(z) \, dz \\
& = - \dkl \bigl(q \| \Prob(\cdot\, |y, u) \bigr) + \log \J(u),
\end{align*}
as desired.
\end{proof}

Note that the left-hand side of identity \eqref{eq:identity} is independent of density $q$. Noting that $q$
may be chosen freely, and since all divergences\index{divergence} are non-negative, the following is immediate:

\begin{corollary} \label{c:oo} For all $q \in \cP(\R^{d_z})$ and all $u \in \Theta$ it holds that $\log \J(u) \ge \mathcal{L}(q, u).$
\end{corollary}

Recall that we are interested in maximizing $\J(u)$, and hence $\log \J(u)$, over $u$. Recall also that we have no direct
access to the missing data\index{missing data} $z$. Corollary \eqref{c:oo} suggests that we may instead try to maximize $\mathcal{L}(q, u).$
The EM algorithm is an iterative method which generates sequence $\{u_j,q_j\}_{j \in \mathbb{Z}^+}$ by alternating the maximization of $\mathcal{L}(q, u)$ with respect to $q$ and $u.$ The density
$q$ is chosen in a manner which builds into the algorithm learned information about
dependence on the missing data $z$. The high-level structure of the algorithm is:
\begin{enumerate}
    \item Set $q_j \in \argmax_{q \in \cP(\R^{\dz})} \mathcal{L}(q, u_j).$ 
    \item Set $u_{j+1} \in \argmax_{u \in \Theta}\mathcal{L}(q_j,u)$.
\end{enumerate}
Step 1 yields $q_j(z) = \Prob(z|y,u_j)$; this is because
\eqref{eq:identity} shows that the maximization of $\mathcal{L}(q, u_j)$ over $q$ 
is the same as minimization of $\dkl \bigl(q \| \Prob(\cdot\, |y,u_j)\bigr)$ over $q$. Density 
$q_j(z) = \Prob(z|y,u_j)$ achieves the global minimum of this non-negative function and
we have that
\begin{equation} \label{eq:addthis}
  \dkl \bigl( q_j \| \Prob(\cdot\, |y,u_j)\bigr)=0.  
\end{equation}
We thus obtain the identity
\begin{equation}
\label{eq:equality}    
\log \J( u_j)=\mathcal{L}(q_j, u_j).
\end{equation}
Step 2 proceeds by noting that 
\begin{subequations}
\label{eq:equality2}
\begin{align}
\mathcal{L}(q_j, u) &= \ELBO\bigl(\Prob(y, \cdot\,  | u), q_j\bigr),\\
&= \mathbb{E}^{q_j}[\log \Prob(y, \cdot\,  | u)] - \mathbb{E}^{q_j}[\log q_j],\\
&= \int_{\R^{d_z}} \log\Prob(y, z | u) \Prob(z|y, u_j) \, dz + \mathsf{const}.
\end{align}
\end{subequations}
Here $\mathsf{const}$ is independent of $u$, but does depend on $u_j$; since, in this step, maximization is over $u$, it is irrelevant.
Hence, the quantity to maximize is the expected value of the joint log-density $\log  \Prob (y, z | u)$ with respect to $q_j(z) = \Prob(z|y, u_j);$ we see explicitly that $q_j$ thus
represents the distribution
of the missing data $z$, given the data $y$ and current parameter estimate $u_j.$

In summary, the E-step corresponds to computing $\mathcal{L}(q_j, u)$ from \eqref{eq:equality2}; the M-step
corresponds to maximizing the resulting expression over $u$ to obtain $u_{j+1}.$
Combining these two steps gives the \index{expectation maximization}EM algorithm:

\begin{algorithm}
\caption{\label{alg:EM} Expectation Maximization\index{algorithm!EM}}
\begin{algorithmic}[1]
\STATE {\bf Input}:  Initialization $u_0.$
\STATE For $j = 0, 1, \ldots$ do the following expectation and maximization steps:
\STATE {\bf E-Step}: Compute 
\begin{align*}
\Expect^{z \sim  \Prob(z|y, u_j)} \Bigl[ \log \Prob(y, z | u) \Bigr]= \int \log \Prob(y, z | u) \Prob(z|y, u_j) \, dz. 
\end{align*}
\vspace{-0.5cm}
\STATE{{{\bf M-Step}}}: Compute 
\begin{equation*}
u_{j+1} \in \argmax_{u\in\Theta} \Expect^{z \sim  \Prob(z|y, u_j)} \Bigl[ \log \Prob(y, z | u) \Bigr].
\end{equation*}
\STATE{\bf Output}: Parameter estimates $\{u_j\}_{j \in \bbZ^+}.$
\end{algorithmic}
\end{algorithm}
\FloatBarrier

\begin{example} We return to hidden Markov model\index{hidden Markov model} Example \ref{ex:1515}. 
The E-step requires the smoothing\index{smoothing!distribution} distribution $\Prob(z|y, u_j)$,
at current parameter estimation step. The M-step requires formula for the joint distribution of 
signal and observation, at arbitrary parameter $u$; and maximization of this function when averaged
with respect to the smoothing distribution. For practical reasons, the smoothing distribution is sometimes
replaced by the filtering distribution\index{filtering!distribution} as this may be easier to approximate.
\end{example}

The following result shows that the \index{expectation maximization}EM framework has the desirable property that the \index{likelihood}likelihood function increases monotonically along iterates $u_j.$ 

\begin{theorem}\label{th:EMmonotone}
Let $\{u_j \}_{j \ge 0}$ be the iterates of the EM Algorithm\index{algorithm!EM} \ref{alg:EM}. Then, 
\begin{equation*}
\J (u_j) \le \J(u_{j + 1}).
\end{equation*}
\end{theorem}
\begin{proof}
Since the logarithm is an increasing function in its domain, it suffices to show that $\log  \J (u_j) \le \log \J(u_{j + 1}).$
Let $q_j(z) = \Prob(z|y, u_j).$  Using the \index{likelihood}log-likelihood characterization in Theorem \ref{th:EMlikecharacterization}, it holds that 
\begin{align*}
\log \J(u_{j+1}) 
&=\mathcal{L}(q_j, u_{j + 1}) + \dkl \bigl( q_j \| \Prob(\cdot\,|y,u_{j + 1})\bigr) \\ 
&\ge \mathcal{L}(q_j, u_{j}) +  \dkl \bigl( q_j \| \Prob(\cdot\,|y,u_{j + 1})\bigr) \\
&\ge \mathcal{L}(q_j, u_{j})  + \dkl \bigl( q_j \| \Prob(\cdot\,|y,u_{j })\bigr) \\
& = \log \J(u_{j}).
\end{align*}
The first inequality follows because $u_{j + 1} \in \argmax_u \mathcal{L}(q_j, u);$ the second follows from \eqref{eq:addthis} and the fact that all divergences are non-negative; the final equality follows from \eqref{eq:addthis} and \eqref{eq:equality}. 
\end{proof}

\begin{remark}
\label{rem:em}
The monotonic increase of the likelihood does not guarantee, in general, that the EM parameter estimates $\{u_j\}_{j \in \bbZ^+}$ converge in the large $j$ limit.
Convergence of parameter estimates to a local maximizer of the \index{likelihood}likelihood function can be established
under additional assumptions.
It is important to note, however, that the expectation in the E-step and the \index{optimization}optimization in the M-step are often intractable. \index{Monte Carlo}Monte Carlo, \index{filtering}filtering, or \index{smoothing}smoothing algorithms may be employed to approximate the E-step, and \index{optimization}optimization algorithms to approximate the M-step. Such approximations can cause loss of monotonicity and loss of convergence guarantees.
\end{remark}

We conclude with an independent and direct proof of Corollary \ref{c:oo}.

 \begin{lemma} \label{lem:EMKI}
Let $q$ be a probability density function in $\R^{d_z}$ and let $\Prob(y,z |u),$ viewed as a function of $z\in \R^{d_z}$ with $y \in \R^k$ and $u\in \R^d$ fixed, be the unnormalized density of latent variable $z \in \R^{d_z}.$ It holds that
\begin{subequations}
\begin{align*}
\ELBO\bigl(\Prob(y, \cdot\,  | u), q\bigr) &\le \log \Prob(y|u),\\
\mathcal{L}(q, u) &\le \log \J(u).
 \end{align*} 
 \end{subequations}
\end{lemma}
\begin{proof}
This lower bound follows by using the \index{Jensen inequality}Jensen inequality: 
\begin{align*}
 \log   \Prob(y|u) &= \log \ \int_{\R^{d_z}}  \Prob (y, z| u) \, dz \\ 
&  = \log \ \int_{\R^{d_z}}  \frac{\Prob (y, z| u)}{q(z)} q(z) \, dz \\ 
& \ge \ \int_{\R^{d_z}} \log\biggl( \frac{\Prob (y, z| u)}{q(z)} \biggr)  q(z)  \, dz \\
&  = \ELBO(\Prob(y, \cdot\, | u), q). 
\end{align*}
The second inequality in the statement of the lemma follows directly from the definitions.
\end{proof}

\section{Markov Chain Monte Carlo\index{Markov chain Monte Carlo|textbf}}\label{sec:MCMC}
In this section we turn our attention to the problem of drawing samples from a given target distribution $\post \in \cP(\R^d).$ We will focus on Markov chain Monte Carlo (MCMC) \index{MCMC}algorithms, a wide class of methods that share the idea of generating samples from a Markov chain\index{Markov!chain} for which the target is an invariant distribution.  
To describe MCMC we first recall that a \index{Markov kernel}\emph{Markov kernel} is a function $p: \Ru \times \Ru \to \R$ satisfying:
\begin{itemize}
    \item $p(u,v) \ge 0$ for all $(u, v) \in \Ru \times \Ru;$ and
    \item $\int_{\Ru} p(u,v) \, dv = 1 $ for all $u \in \Ru.$
\end{itemize}
For fixed $u\in \Ru,$ $p(u, \cdot)$ defines a probability density function on $\Ru.$
From this kernel we can define a Markov chain through the iteration
\begin{equation}\label{eq:MCMCsampling}
    u^{(n+1)} \sim p(u^{(n)}, \cdot), \qquad 0 \le n \le N-1 .
\end{equation}
Note that if the chain is currently at $u^{(n)} \in \Ru$ then $p(u^{(n)},\cdot)$ defines the probability with which the chain will be in any given Borel subset of $\Ru$ after one time step.
The idea of MCMC is to define a Markov kernel
with the property that the target $\post$ is an invariant distribution for the kernel. 
This means that if we sample an initial draw $u^{(0)} \sim \post$ 
then it holds that $u^{(n)} \sim \post $ for all $1 \le n \le N.$

 It is possible to show, under mild assumptions, that even when the initial draw $u^{(0)}$ is sampled from initial distribution $\pi_0$ different from $\post$, the distribution of the $n$-th draw $u^{(n)}$ defined by \eqref{eq:MCMCsampling} converges to $\post$ in the large $n$ asymptotic. Ergodicity\index{ergodic!Markov chain}
 (see Section \ref{sec:ERG}) then enables approximate computation of expectations 
 under $\pi$ from a single long trajectory.
 In the next two subsections we introduce two classes of MCMC methods that are widely used in inverse problems, data assimilation and other applications: the Metropolis--Hastings algorithm and the Gibbs sampler. 

\subsection{The Metropolis--Hastings Algorithm}
We consider first the \index{Metropolis--Hastings}Metropolis--Hastings algorithm, a versatile \index{MCMC}MCMC approach to define a \index{Markov kernel}Markov kernel $\pmh$ that can be readily sampled and which satisfies \index{detailed balance}\emph{detailed balance} with respect to $\post;$ that is, for all $u, v \in \Ru$ it holds that
\begin{equation}\label{eq:detailedbalance}
    \post(u) \pmh(u,v) = \post(v) \pmh(v,u);
\end{equation}
this means that the probability of being in state $u$ and transitioning from $u$ to $v$ is equal to the probability of being in state $v$ and transitioning from $v$ to $u$.
Integrating over $u$ delivers
$$\int_{\R^d}  \post(u) \pmh(u,v) du = \post(v) \int_{\R^d} \pmh(v,u) du =  \post(v),$$
which is the desired invariance property.
The key idea behind the \index{Metropolis--Hastings}Metropolis--Hastings algorithm is that kernel $\pmh$ satisfying \eqref{eq:detailedbalance} can be defined by sampling from a proposal \index{Markov kernel!proposal}Markov kernel $q,$ chosen to be easy to sample from, and an accept/reject step.  Given 
the $n$-th sample $u^{(n)},$ we obtain the $(n+1)$-th sample through a two-step process. First, we propose $v^* \sim q(u^{(n)}, \cdot)$ by sampling from the \index{Markov kernel!proposal}\emph{proposal} Markov kernel. Second, we accept this proposal and set $u^{(n+1)} = v^*$ with probability $a(u^{(n)}, v^*);$ otherwise, we reject it and set
$u^{(n+1)} = u^{(n)}.$ The Metropolis--Hastings acceptance probability\index{probability!acceptance} is defined by 
\begin{equation}\label{eq:mhaccept}
    a(u,v):= \min \biggl\{1, \frac{\post(v)}{\post(u)} \frac{q(v,u)}{q(u,v)}  \biggr\}, \qquad u, v \in \Ru.  
\end{equation}
This definition ensures that the \index{Markov kernel}Markov kernel $\pmh$ that determines the probabilistic transition from $u^{(n)}$ to $u^{(n+1)}$ satisfies the detailed balance condition \eqref{eq:detailedbalance}. Algorithm \ref{algMH} outlines the procedure.

\FloatBarrier
\begin{algorithm}
\caption{\label{algMH} \index{Metropolis--Hastings}Metropolis--Hastings Algorithm}
\begin{algorithmic}[1]
\vspace{0.1in}
\STATE {\bf Input}: Target distribution $\post$, initial distribution $\post_0,$ proposal Markov kernel $q$,\index{Markov kernel!proposal} test function $f,$ number of samples $\Sam.$  \\
\STATE {\bf Initial Draw}: Draw initial sample $u^{(0)} \sim \post_0.$ 
 \\
\STATE{{{\bf Subsequent Samples}}}: For $\sam = 0,1,\dots,\Sam -1$ do:
\begin{enumerate}
\item Sample $v^{\star} \sim q(u^{(\sam)},\cdot).$
\item Update 
\vspace{-.04in}
\begin{align*}
u^{(\sam+1)} =
\begin{cases}
 v^{\star}, &  \text{with probability} \,\, a(u^{(n)}, v^*) \, \, \text{using} \, \, \eqref{eq:mhaccept}. \\ 
 u^{(\sam)}, & \text{with probability} \,\,  1 - a(u^{(n)}, v^*).
\end{cases}
 \end{align*}
\end{enumerate}
\STATE{\bf Output}: Samples $\{u^{(n)}\}_{n=1}^N$ and approximation $\Expect^{u \sim \post}[ f(u)] \approx  \frac{1}{N} \sum_{n=1}^N f(u^{(n)}).$
\end{algorithmic}
\end{algorithm}
\FloatBarrier

\begin{remark}
\label{rem:nonormagain}
Notice that the target distribution only enters the \index{Metropolis--Hastings}Metropolis--Hastings algorithm in the acceptance probability  \eqref{eq:mhaccept}; since this formula only involves the \emph{ratio} of the target density at the proposed and current values of the chain, the acceptance probability can be computed even when the target distribution is only known up to normalization constant. 
\end{remark}

\setlength{\tabcolsep}{18pt}
\renewcommand{\arraystretch}{1.6}
\begin{table}
\centering
\begin{tabular}{ll}
Algorithm & Proposal      \\\hline \hline 
Random Walk Metropolis--Hastings\index{random walk Metropolis--Hastings (RWMH)} & $q(u, \cdot ) = \cN( u, \beta I_d)$     \\
Metropolis-Adjusted Langevin \index{Metropolis-adjusted Langevin algorithm (MALA)}&   $q(u, \cdot) = \cN( u + \frac{\beta}{2}  D \log \post(u), \beta I_d)$   \\
Preconditioned Crank-Nicolson \index{preconditioned Crank--Nicolson (pCN)} & $q(u,\cdot) = \Nc\bigl( (1 - \beta)^{1/2} u, \beta I_d \bigr)  $ \\
\end{tabular}
\caption{\label{table:proposals}  Three representative instantiations of the Metropolis--Hastings \index{Metropolis--Hastings} algorithm, obtained by choosing specific \index{Markov kernel!proposal}proposal Markov kernels. The parameter $\beta$ in the proposal kernels allows control of the acceptance rate.}
\end{table}

The choice of \index{Markov kernel!proposal}proposal Markov kernel has a major impact on the performance of the Metropolis--Hastings algorithm. Table \ref{table:proposals} contains three illustrative choices of \index{Markov kernel!proposal}proposal Markov kernel that lead to important instantiations of the Metropolis--Hastings algorithm, known as the \index{random walk Metropolis--Hastings (RWMH)}\emph{random walk Metropolis--Hastings} (RWMH) algorithm, the \index{Metropolis-adjusted Langevin algorithm (MALA)}\emph{Metropolis-adjusted Langevin algorithm} (MALA) and the \index{preconditioned Crank--Nicolson (pCN)}\emph{preconditioned Crank--Nicolson} (pCN) method. These three proposals depend on a user-chosen parameter $\beta,$ with $\beta>0$ for RWMH and MALA, and $\beta \in (0,1)$ for pCN. Proposal $v^*$  can be obtained by drawing Gaussian sample $\xi \sim \Nc(0, I_d)$ and setting, respectively, 
\begin{alignat*}{3}
    v^* &:= u + \beta^{1/2} \xi,  &&\qquad \qquad \text{(RWMH proposal)}   \\
    v^*  &:= u + \frac{\beta}{2} D \log  \pi(u) + \beta^{1/2} \xi,&& \qquad \qquad  \text{(MALA proposal)}  \\
    v^* &:= (1 - \beta)^{1/2} u  + \beta^{1/2} \xi. && \qquad \qquad  \text{(pCN proposal)} 
\end{alignat*}
In each case, the parameter $\beta$ controls how far proposed sample $v^* \sim q(u, \cdot)$ is, on average, from the current state $u \in \Ru$ of the chain. Smaller $\beta$ leads to $v^*$ being closer to $u,$ and thus to Metropolis--Hastings algorithms with slow exploration but high acceptance rate. Many theoretical and heuristic guidelines are available to choose the step-size $\beta.$ This choice can be made offline (i.e. before implementing the algorithm) but also online (i.e. adapting the choice as the chain runs, based on its output). Online adaptation must be designed carefully to preserve the
ergodicity of the resulting algorithm.

\begin{remark}
\index{random walk Metropolis--Hastings (RWMH)}RWMH was the first MCMC algorithm to be developed, and it is still widely used due to its simplicity. \index{Metropolis-adjusted Langevin algorithm (MALA)}MALA adds gradient information from the target distribution so that proposed samples tend to move to regions of higher target density; this algorithm is closely related to the score-based approach for generative modeling considered in Section~\ref{sec:score}; indeed the MALA proposal arises from Euler--Maruyama discretization of the Langevin\index{Langevin equation} equation. The
\index{preconditioned Crank--Nicolson (pCN)}pCN proposal modifies RWMH to obtain a \index{Markov kernel!proposal} proposal Markov kernel that satisfies detailed balance\index{detailed balance} with respect to Gaussian density 
$\pr:=\Nc(0, I_d).$  In particular if $\post(u) \propto \exp\bigl(-\Phi(u)\bigr)\pr(u)$ then for
the pCN proposal the acceptance probability \eqref{eq:mhaccept} becomes
\begin{equation}\label{eq:mhaccept2}
    a(u,v):= \min \biggl\{1, \exp\bigl(\Phi(u)-\Phi(v)\bigr) \biggr\}, \qquad u, v \in \Ru.  
\end{equation}
The method is thus useful for sampling target distributions arising as posterior distributions in Bayesian inverse problems with Gaussian prior $\pr$. 
\end{remark}

\subsection{The Gibbs Sampler}\label{ssec:Gibbs}
Another important MCMC algorithm is the \index{Gibbs sampler} \emph{Gibbs sampler}, where each one-dimensional coordinate $u_i$ of $u \in \Ru$ is updated in turn by sampling from the target's \emph{full conditional distributions} \index{full conditional distributions} given by
$$\post_i(u_i | u_{-i}) := \frac{\post(u)}{\post(u_{-i})},$$
where $$\post(u_{-i}):= \int_\R \post(u_1, \ldots, u_{i-1}, v, u_{i+1}, \ldots, u_d) \, dv.$$
Importantly, the full conditionals are probability densities on $\R$ and the cost of sampling from these distributions need not, in general, scale with the dimension $d$ of the parameter $u \in \R^d.$ 
The Gibbs sampler yields a Markov chain whose kernel leaves the target
$\post$ invariant. A random-scan Gibbs kernel ---that is, one in which the coordinate to update is chosen at each iteration according to fixed probabilities--- also satisfies detailed
balance with respect to $\post$.

\FloatBarrier
\begin{algorithm}
\caption{\label{alg:Gibbs} \index{Gibbs sampler}Gibbs Sampler}
\begin{algorithmic}[1]
\vspace{0.1in}
\STATE {\bf Input}: \index{target distribution}Target distribution $\post$, initial distribution $\post_0,$ test function $f,$ number of samples $\Sam.$  \\
\STATE {\bf Initial Draw}: Draw initial sample $u^{(0)} \sim \post_0.$  
 \\
\STATE{{{\bf Subsequent Samples}}}: \\
 For $\sam = 0,1,\dots,\Sam -1$ 
\STATE \hspace{1cm} For $i = 1, \ldots, d$  
\begin{enumerate}[leftmargin=3cm]
\item  Sample $u_i^{(n+1)} \sim \post_i \Bigl(\,\cdot \, | u_1^{(n+1)}, \ldots,  u_{i-1}^{(n+1)}, u_{i+1}^{(n)}, \ldots, u_{d}^{(n)}  \Bigr).$
\item Set 
$u^{(\sam+1)} = \Bigl(u_1^{(n+1)}, \ldots, u_{i-1}^{(n+1)}, u_i^{(n+1)}, u_{i+1}^{(n)}, \ldots, u_d^{(n)} \Bigr)
.$ 
\end{enumerate}
\STATE \hspace{1cm} End for
\STATE Record sample $u^{(n+1)}.$
\STATE End for
\STATE{\bf Output}: Samples $\{u^{(n)}\}_{n=1}^N$ and approximation $\Expect^{u\sim\post}[ f(u)] \approx  \frac{1}{N} \sum_{n=1}^N f(u^{(n)}).$
\end{algorithmic}
\end{algorithm}
\FloatBarrier

\begin{remark}
  The Metropolis--Hastings \index{Metropolis--Hastings} algorithm and the Gibbs sampler \index{Gibbs sampler} were developed independently and have distinct flavors: in Metropolis--Hastings we propose an update $v^*$ of all coordinates of $u$ at once, and accept or reject this update probabilistically; in contrast, in the Gibbs sampler we update each coordinate in turn by sampling from the target's full conditional, and there is no accept/reject step. Despite these differences, it is possible to view the Gibbs sampler as a Metropolis--Hastings method in which at each iteration a proposal Markov kernel is defined via a full conditional. With such a choice of proposal Markov kernel, \index{Markov kernel!proposal} the acceptance probability in the accept/reject step of the Metropolis--Hasting algorithm automatically simplifies  to be $1,$ explaining the lack of an accept/reject step in the Gibbs sampler.   
\end{remark}

Rather than updating a single coordinate at a time, it is often useful to perform Gibbs updates block-wise. To explain this idea, let $u = (u_a \;\; u_b) \in \Ru$ with $u_a \in \R^{d_a},$ $u_b \in \R^{d_b}$ and $d_a + d_b = d.$ We can then update $u_a$ and $u_b$ in turn by sampling from the target's full conditionals
\begin{align*}
    \post_a( u_a|u_{-a}):= \frac{\post(u)}{\post(u_{-a})}, \qquad \post_b( u_b|u_{-b}):= \frac{\post(u)}{\post(u_{-b})},
\end{align*}
where 
\begin{align*}
    \post(u_{-a}): = \int_{\R^{d_a}} \post(v, u_b) \, dv, \qquad \post(u_{-b}): = \int_{\R^{d_b}} \post(u_a, v) \, dv.
\end{align*}
The procedure is outlined in Algorithm \ref{alg:Gibbs_block}:

\FloatBarrier
\begin{algorithm}
\caption{\label{alg:Gibbs_block} \index{Gibbs sampler}Block-Wise Gibbs Sampler}
\begin{algorithmic}[1]
\vspace{0.1in}
\STATE {\bf Input}: \index{target distribution}Target distribution $\post$, initial distribution $\post_0,$ test function $f,$ number of samples $\Sam.$  \\
\STATE {\bf Initial Draw}: Draw initial sample $u^{(0)} \sim \post_0.$  
 \\
\STATE{{{\bf Subsequent Samples}}}: \\
 For $\sam = 0,1,\dots,\Sam -1$ 
\begin{enumerate}[leftmargin=3cm]
\item  Sample $u_a^{(n+1)} \sim \post_a \Bigl(\,\cdot \, | u_{-a}^{(n)}  \Bigr).$
\item Sample 
$u_b^{(\sam+1)} \sim \post_b \Bigl(\,\cdot \, | u_{-b}^{(n+1)}  \Bigr).$
\item Set $u^{(n+1)} = \bigl(u_a^{(n+1)}, u_b^{(n+1)} \bigr).$
\end{enumerate}
\STATE End for
\STATE{\bf Output}: Samples $\{u^{(n)}\}_{n=1}^N$ and approximation $\Expect^{u \sim \post}[ f(u)]  \approx  \frac{1}{N} \sum_{n=1}^N f(u^{(n)}).$
\end{algorithmic}
\end{algorithm}
\FloatBarrier

\begin{remark}
The Gibbs sampler tends to converge more quickly when highly correlated coordinates are  blocked together. On the other hand conditionals can become hard to sample from if many coordinates are blocked together. Optimizing this trade-off is central to the successful implementation of the Gibbs
sampler. In determining how to optimize this trade-off it is useful to note that, when sampling from the full conditionals is challenging, a Metropolis--Hastings algorithm can be employed to sample from them. The resulting algorithm, which goes under the name of \emph{Metropolis-within-Gibbs}, preserves the desired stationary distribution if any number of  Metropolis--Hastings
steps is used, instead of an exact sample from a conditional. However replacing exact conditional
sampling with approximate Metropolis--Hastings steps will typically slow down the rate of convergence of the overall algorithm.
\end{remark}

\section{Automatic Differentiation}\index{differentiation!automatic}
\label{sec:auto-differentiation}

Gradients (and higher-order derivatives) are often difficult to obtain in closed form for complex cost functions. Finite difference approximations, on the other hand, are often inaccurate and expensive for high-dimensional problems.

\emph{Automatic differentiation}\index{differentiation!automatic} involves repeated application of chain rule on elementary operations that enables computing derivatives accurately to working precision. That is, suppose an output variable $y\in\mathbb{R}^{d_n}$ is related to an input variable $x\in\mathbb{R}^{d_0}$ by some function $f$, i.e., $y = f(x),$ and suppose that the Jacobian of $f$ is not readily available in closed form. We assume the implementation of $f$ in computer code is made up of $n$ elementary operations $f_i: \mathbb{R}^{d_{i-1}}\to\mathbb{R}^{d_i}$ (for instance, addition, multiplication, logarithms, etc.), such that
\begin{equation}
    f(x) = f_n \circ f_{n-1} \circ \cdots \circ f_1(x).
\end{equation}
We assume further that the Jacobians for these elementary operations, $\nabla f_i: \mathbb{R}^{d_{i-1}}\to\mathbb{R}^{d_{i}\times d_{i-1}}$, are available in closed form. The goal of automatic differentiation\index{differentiation!automatic} is to evaluate $Df(x)$ from the Jacobians of the elementary operations.

We define the identity function $g_0(x)=x$
and then recursively define function 
$g_i(x) = f_i\bigl(g_{i-1}(x)\bigr))$; note that then $f=g_n.$
By representing the output $g_i$ of each step in the recursion by a node, we can use a graph\index{graph} to represent the function evaluation where the directed edges encode the input dependencies $g_{i-1}$ (the parents\index{parents}) that are required in the evaluation of $f_i$. This is known as the \emph{computational graph}\index{graph!computational}. The nodes without parents correspond to the original inputs $x$ to the function $f$. After constructing the graph, we can define the \emph{forward mode}\index{differentiation!forward mode} as the process of computing the output $y$, and possibly the gradient of $f$, by traversing the graph starting from the inputs $x$. The \emph{reverse mode}\index{differentiation!reverse mode}  starts with the final output node and traverses the graph back to the inputs to compute the gradient of $f$. The following two subsections demonstrate how the full Jacobian of $f$ is assembled in forward and reverse mode automatic differentiation.

\begin{remark}
    In realistic implementations, instead of multiplying the full Jacobians of each operation, only the partial derivatives of variables that interact with each other in the computational graph are used. For simplicity of exposition, we present the algorithms with the full Jacobians here. We state the 
    complexity, that is the number of required arithmetic operations, using standard matrix multiplication; we do not report complexity for optimized algorithms that exploit any specific computational graph.
\end{remark}

\subsection{Forward Mode}
Forward mode automatic differentiation\index{differentiation!forward mode}  
proceeds to accumulate the incremental function evaluations along with their derivatives. With $g_i$ recursively defined as above
we may write, for $i \in \{1, \ldots, n\}$,
\begin{equation*}
    g_i(x) = (f_i\circ\cdots\circ f_1)(x).
\end{equation*}
By the chain rule we have that
\begin{equation}\label{eq:jacob_prod}
    D_x f(x) =  \Bigl(D_{g_{n-1}}f_n\bigl(g_{n-1}(x)\bigr)\Bigr)\cdots \Bigl(D_{g_1}f_2\bigl(g_1(x)\bigr)\Bigr)\Bigl(D_x f_1(x)\Bigr).
\end{equation}
This suggests the following algorithm\index{algorithm!auto-differentiation}
\begin{algorithm}
\caption{\label{alg:forward_autodiff}Forward Mode Automatic Differentiation}
\begin{algorithmic}[1]
\STATE {\bf Input}:  Functions $\{f_i(\cdot)\}_{i=1}^n$, corresponding Jacobians $\{Df_i(\cdot)\}_{i=1}^n$, and input $x$.
\STATE Set $g_1 = f_1(x)$ and $J_1 = D f_1(x)$.
\STATE For $i = 2, \ldots, n$: set $g_i = f_i(g_{i-1})$ and $J_i = \bigl(Df_i(g_{i-1})\bigr) J_{i-1}$.
\STATE{\bf Output}: Function output $y = f(x) = g_n$ and derivative $Df(x) = J_n$.
\end{algorithmic}
\end{algorithm}
\FloatBarrier

The number of floating-point operations required to multiply the Jacobians $J_i$ in forward mode automatic differentiation to compute $Df(\cdot)$ is of order
\begin{equation*}
    d_0 \sum_{i=1}^{n-1} d_{i+1} d_i.
\end{equation*}
Assuming for simplicity that all the $d_i$ except for $d_0$ and $d_n$ are fixed at $d$, we have a cost 
$\mathfrak{c}_f$ given by
\begin{equation}
\label{eq:cf}
    \mathfrak{c}_f=d^2(n-2)d_0 + d_n d d_0.
\end{equation}
We now introduce a different way of computing $Df(\cdot)$ which reverses the order of
arithmetic operations, resulting in a different computational cost $\mathfrak{c}_r.$
The difference in computational cost compared to forward mode differentiation stems from the fact that, while matrix multiplication is associative, the complexity is not.

\subsection{Reverse Mode}

\emph{Reverse mode} automatic differentiation\index{differentiation!reverse mode}  also computes the product of Jacobians \eqref{eq:jacob_prod}, but instead of computing the product from right to left, it computes it from left to right. The algorithmic implementation requires a forward pass to evaluate the function and a backward pass to compute the derivative:
\begin{algorithm}
\caption{\label{alg:reverse_autodiff}Reverse Mode Automatic Differentiation}\index{algorithm!auto-differentiation}
\begin{algorithmic}[1]
\STATE {\bf Input}:  Functions $\{f_i(\cdot)\}_{i=1}^n$, corresponding Jacobians $\{Df_i(\cdot)\}_{i=1}^n$, and input $x$. 
\STATE Set $g_1 = f_1(x)$.
\STATE For $i = 2, \ldots, n$: set and store $g_i = f_i(g_{i-1})$.
\STATE Set $J_n = D f_n(g_{n-1})$.
\STATE For $i = n-1, \ldots, 1$ set $J_i = J_{i+1}Df_i(g_{i-1})$.
\STATE{\bf Output}: Function output $y = f(x) = g_n$ and derivative $Df(x) = J_1$.
\end{algorithmic}
\end{algorithm}
\FloatBarrier

This leads to the number of floating-point operations required to multiply the Jacobians $J_i$ to compute $Df(\cdot)$  being of order
\begin{equation*}
    d_n \sum_{i=1}^{n-1} d_{i-1} d_i.
\end{equation*}
Again assuming that all the $d_i$ are fixed at $d$ except for $d_0$ and $d_n$, the total cost 
$\mathfrak{c}_r$ is given by
\begin{equation}
\label{eq:cr}
    \mathfrak{c}_r=d^2(n-2)d_n + d_n d d_0.
\end{equation}
We may now compare $\mathfrak{c}_f$ and $\mathfrak{c}_r$ given by \eqref{eq:cf} and \eqref{eq:cr}.
This comparison shows that forward mode is advantageous when $d_n$ is large whereas reverse mode is advantageous when $d_0$ is large. As an example, if $d_0$ is fixed and $d_n=d,$ then, in the large $d$ asymptotic, forward mode has quadratic cost in $d$ whereas reverse mode has cubic cost in $d.$ In contrast, if $d_n$ is fixed and $d_0 = d,$ then, in the large $d$ asymptotic, forward mode has cubic cost in $d$ whereas reverse mode has quadratic cost in $d.$

\begin{remark}
In many machine learning tasks, inputs are high dimensional but outputs are real-valued; in that setting, reverse mode differentiation is preferred.
Backpropagation\index{backpropagation}, a common method for computing the gradient of the objective function with respect to the weights of a neural network, is a special case of reverse mode differentiation.
\end{remark}

\subsection{Reparameterization Trick}\label{ssec:reparameterization}\index{reparameterization trick|textbf}

In many machine learning tasks, we are interested in computing the gradient of an empirical expectation with respect to parameters of the distribution.

That is, suppose that we have an expectation of a function $f$ with respect to a probability distribution $\rho(\cdot; \theta)$, where $\theta$ are some parameters:
\begin{equation}\label{eq:expect_reparameterization}
    \J(\theta) = \Expect^{u\sim \rho(\cdot; \theta)} [f(u)].
\end{equation}
We would like to compute the gradient of the expectation with respect to $\theta$. Given i.i.d. samples $\{u^{(n)}\}_{n=1}^N$ from $\rho(\cdot; \theta)$, $\J$ can be approximated by
\begin{equation*}
\J^N(\theta) = \frac{1}{N}\sum_{n=1}^N f(u^{(n)}).
\end{equation*}
We cannot, however, compute the gradient $\D_\theta \J^N(\theta)$ using automatic differentiation, since this would involve differentiating the random samples $u^{(n)}$.

The \emph{reparameterization trick}\index{reparameterization trick} involves rewriting $u$ as a function $g(z; \theta)$ of a random variable $z$ drawn from a distribution $\zeta$ that does not depend on $\theta$ and that can be sampled from. That is,
\begin{equation}
    u = g(z; \theta),\quad z\sim \zeta.
\end{equation}
Equivalently, $\rho = g(\cdot; \theta)_\sharp \zeta$. Then, \eqref{eq:expect_reparameterization} can be rewritten as
\begin{equation}
    \J(\theta) = \Expect^{z\sim \zeta} [f(g(z; \theta))].
\end{equation}
Instead of drawing samples from $\rho$, we can draw $\{z^{(n)}\}_{n=1}^N$ i.i.d. from $\zeta$, and approximate the expectation as
\begin{equation}
\J^N(\theta) = \frac{1}{N}\sum_{n=1}^N f(g(z^{(n)};\theta)).
\end{equation}
Then,
\begin{equation}
\D_\theta\J^N(\theta) = \frac{1}{N}\sum_{n=1}^N \D_\theta f(g(z^{(n)};\theta)),
\end{equation}
where the $z^{(n)}$ are constant with respect to $\theta$, so the gradients of the terms of the sum can be computed using automatic differentiation.

\begin{example}
    The most common example of the reparameterization trick is reparameterizing $u\sim\rho(\cdot; \theta) = \mathcal{N}(m, \Sigma)$, where $\theta$ may consist of $m$, $\Sigma$, or both, as $u = g(z; \theta) =  m + \Sigma^{\frac{1}{2}}z$, where $z\sim \zeta = \mathcal{N}(0, I)$ is the standard normal.
\end{example}

\section{Bibliography}
\label{sec:OB}

Optimization is reviewed in a number of textbooks, including \cite{boyd_convex_2004,nocedal_numerical_2006}.
We refer to \cite{bottou2018optimization} for an accessible survey article on optimization for machine learning.
Convergence
proofs for gradient descent and stochastic gradient descent
may be found in \cite{sanz-alonso_inverse_2023} in simple
settings. Nesterov's accelerated gradient method and its rate of convergence are discussed in detail in \cite{nesterov_introductory_2004}.
Gradient descent can be formulated in continuous time as a gradient flow;\index{gradient!flow} see
\cite{hale,ambrosio2005gradient}.
Newton and Gauss--Newton methods are surveyed in many textbooks; see for instance \cite{nocedal_numerical_2006}. The papers \cite{bell1993iterated,bell1994iterated} established a direct connection between Gauss--Newton methods and the extended Kalman filter (ExKF)\index{Kalman filter!extended} discussed in Subsection~\ref{ssec:exkf}.

The origins of ensemble Kalman methods for inverse problems stem from the observation that the analysis step in data assimilation involves solution of an inverse problem; this observation is formalized, for instance, in \cite{reich2011dynamical}.
Early developments of ensemble Kalman methods for inverse problems include \cite{chen2012ensemble,emerick2013ensemble}. For a history of the use of ensemble Kalman methods to solve inverse problems, see the bibliography sections in \cite{chada2021iterative,calvello22}. Ensemble Kalman inversion\index{Kalman inversion!ensemble} (EKI) as defined here was introduced in \cite{iglesias2013ensemble}, and
various adaptations of it may be found in \cite{iglesias2015iterative,iglesias2016regularizing,chada2019tikhonov,iglesias2021adaptive,chada2021iterative,vernon_nesterov_2025}; it is 
also discussed in the final chapter of the textbook \cite{sanz-alonso_inverse_2023}, which follows the presentation in  \cite{chada2021iterative}. 
For analysis in continuous time see \cite{schillings2017analysis,schillings2018convergence}.
A mean field perspective on EKI\index{Kalman inversion!ensemble} and an interpretation of the method as a covariance-preconditioned gradient flow\index{gradient!flow} is provided in \cite{calvello22}. Variants of EKI\index{Kalman inversion!ensemble} are discussed in \cite{huang2021unscented2,huang2022iterated}. The use of EKI\index{Kalman inversion!ensemble} to train machine learning models, or machine learning models embedded within physical models, is discussed in \cite{kovachki2019ensemble,lopez2022training,wu2023learning,chen_neural_2024}. There is another growing class 
of methodologies, which also solve optimization problems in a derivative-free fashion, based on the idea
of interacting particle systems which achieve consensus \cite{pinnau2017consensus}. In contrast to ensemble-based
methods, these methods do not invoke a Gaussian approximation and hence can achieve provably good approximations of global optima on a wide range of problems \cite{carrillo2018analytical}; however they have yet to be shown to be effective on the large-scale applications where ensemble Kalman methods excel.

The EM framework for \index{maximum likelihood estimation}maximum likelihood estimation was introduced in \cite{dempster1977maximum}; see \cite{meng1997algorithm}
for a review and see \cite{sahu1999convergence} for an insightful connection to Gibbs\index{Gibbs sampler} sampling. The generalization of the EM framework for optimization problems where the objective does not necessarily involve conditional expectations is known as the Minorize--Maximization (MM), or equivalently Majorize--Minimization, algorithm. An overview of MM algorithms and their properties can be found in~\cite{lange2016mm}. 

Reinforcement learning\index{reinforcement learning}, a form of online optimization, is seeing growing use in numerous
applications of machine learning. For historical underpinnings of the field see
\cite{kaelbling1996reinforcement,sutton1999reinforcement}; for recent developments see
\cite{wiering2012reinforcement,arulkumaran2017deep}. Bayesian optimization\index{Bayesian optimization} is another popular online optimization method \cite{frazier2018tutorial} based on the idea of building a surrogate model of the objective by evaluating it at carefully chosen query points. Bayesian optimization is a particularly useful global optimization method in applications where the objective function is expensive to evaluate and the dimension is moderate. 

 As discussed for instance in \cite{trillos2023optimization,sanz2024first}, optimization and sampling are closely related. 
 A workhorse sampling algorithm for Bayesian computation is Markov Chain Monte Carlo \index{Monte Carlo!Markov chain} (MCMC)\index{MCMC}\index{algorithm!MCMC} that builds a sequence of correlated samples from a target distribution; see~\cite{gamerman2006markov, meyn2012markov} for a general introduction and see \cite{roberts_general_2004} for a flavor of convergence results. MCMC methods can also be useful for non-convex optimization by constructing a target distribution from the objective function. We refer to Section~\ref{sec:14} for more references and discussion on Monte Carlo methods, including quasi-Monte Carlo (QMC)\index{QMC} and
 sequential Monte Carlo (SMC)\index{SMC}.

 A detailed overview of automatic differentiation is given in \cite{griewank_evaluating_2008}. For a survey of automatic differentiation in machine learning, we refer to \cite{baydin2018automatic}. Beyond forward and reverse mode automatic differentiation, the problem of computing the Jacobian for a given computational graph with the least number of arithmetic computations is NP-complete \cite{naumann_optimal_2008}. The reparameterization trick was introduced into the machine learning literature by \cite{kingma2013auto}. It can be applied to general multivariate distributions using the method described in \cite{figurnov_implicit_2018}.
 
 Automatic differentiation is a central tool in machine learning, allowing users to implement highly complex neural network architectures without worrying about the calculation of gradients to optimize the parameters. Automatic differentiation has also contributed to popularizing gradient-based sampling algorithms based on Langevin or Hamiltonian dynamics in applications where calculation of derivatives was previously unfeasible. Automatic differentiation is now available in most programming languages as a package or extension.

\newpage
\bibliographystyle{abbrvnat} 

\bibliography{references,references_new}

@book{nickl,
  title={Bayesian Non-linear Statistical Inverse Problems},
  author={Nickl, R.},
  year={2022},
  url={http://www.statslab.cam.ac.uk/~nickl/Site/__files/lecturenotes.pdf},
  series={Zurich Lectures in Advanced Mathematics},
  publisher={EMS Press}
}

@article{jordan1999introduction,
  title={An introduction to variational methods for graphical models},
  author={Jordan, Michael I and Ghahramani, Zoubin and Jaakkola, Tommi S and Saul, Lawrence K},
  journal={Machine Learning},
  volume={37},
  pages={183--233},
  year={1999},
  publisher={Springer}
}

@book{evensen2022data,
  title={Data Assimilation Fundamentals: A Unified Formulation of the State and Parameter Estimation Problem},
  author={Evensen, G. and Vossepoel, F. C. and van Leeuwen, P. J.},
  year={2022},
  publisher={Springer}
}

@article{sacks1989design,
  title={Design and analysis of computer experiments},
  author={Sacks, Jerome and Welch, William J and Mitchell, Toby J and Wynn, Henry P},
  journal={Statistical Science},
  volume={4},
  number={4},
  pages={409--423},
  year={1989},
  publisher={Institute of Mathematical Statistics}
}

@article{kennedy2001bayesian,
  title={Bayesian calibration of computer models},
  author={Kennedy, Marc C and O'Hagan, Anthony},
  journal={Journal of the Royal Statistical Society: Series B (Statistical Methodology)},
  volume={63},
  number={3},
  pages={425--464},
  year={2001},
  publisher={Wiley Online Library}
}

@article{bigoni2020data,
  title={{Data-driven forward discretizations for Bayesian inversion}},
  author={Bigoni, D. and Chen, Y. and Garcia Trillos, N. and Marzouk, Y. and Sanz-Alonso, D.},
  journal={Inverse Problems},
  doi={10.1088/1361-6420/abb2fa},
  volume={36},
  number={10},
  pages={105008},
  year={2020},
  publisher={IOP Publishing}
}

@article{stuart2018posterior,
  title={{Posterior consistency for Gaussian process approximations of Bayesian posterior distributions}},
  author={Stuart, Andrew and Teckentrup, Aretha},
  journal={Mathematics of Computation},
  volume={87},
  number={310},
  pages={721--753},
  year={2018}
}

@article{dunbar2021calibration,
  title={{Calibration and uncertainty quantification of convective parameters in an idealized GCM}},
  author={Dunbar, Oliver RA and Garbuno-Inigo, Alfredo and Schneider, Tapio and Stuart, Andrew M},
  journal={Journal of Advances in Modeling Earth Systems},
  volume={13},
  number={9},
  pages={e2020MS002454},
  year={2021},
  publisher={Wiley Online Library}
}

@article{patel2021gan,
  title={{GAN-based priors for quantifying uncertainty in supervised learning}},
  author={Patel, Dhruv V and Oberai, Assad A},
  journal={SIAM/ASA Journal on Uncertainty Quantification},
  volume={9},
  number={3},
  pages={1314--1343},
  year={2021},
  publisher={SIAM}
}

@inproceedings{rezende2015variational,
  title={Variational inference with normalizing flows},
  author={Rezende, Danilo and Mohamed, Shakir},
  booktitle={International Conference on Machine Learning},
  pages={1530--1538},
  year={2015},
  organization={PMLR}
}

@inproceedings{rezende2014stochastic,
  title={Stochastic backpropagation and approximate inference in deep generative models},
  author={Rezende, Danilo Jimenez and Mohamed, Shakir and Wierstra, Daan},
  booktitle={International Conference on Machine Learning},
  pages={1278--1286},
  year={2014},
  organization={PMLR}
}

@article{margossian2023amortized,
  title={Amortized variational inference: When and why?},
  author={Margossian, Charles C and Blei, David M},
  journal={arXiv preprint arXiv:2307.11018},
  year={2023}
}

@inproceedings{sun2021deep,
  title={Deep probabilistic imaging: Uncertainty quantification and multi-modal solution characterization for computational imaging},
  author={Sun, He and Bouman, Katherine L},
  booktitle={Proceedings of the AAAI Conference on Artificial Intelligence},
  volume={35},
  pages={2628--2637},
  year={2021}
}

@inproceedings{song2020score,
  title={Score-Based Generative Modeling through Stochastic Differential Equations},
  author={Song, Yang and Sohl-Dickstein, Jascha and Kingma, Diederik P and Kumar, Abhishek and Ermon, Stefano and Poole, Ben},
  booktitle={International Conference on Learning Representations},
  year = {2021}
}

@inproceedings{gabrie2021efficient,
  title={{Efficient Bayesian sampling using normalizing flows to assist Markov chain Monte Carlo methods}},
  author={Gabri{\'e}, Marylou and Rotskoff, Grant M and Vanden-Eijnden, Eric},
  booktitle={ICML Workshop on Invertible Neural Networks, Normalizing Flows, and Explicit Likelihood Models},
  year={2021}
}

@article{tabak2010density,
  title={Density estimation by dual ascent of the log-likelihood},
  author={Tabak, Esteban G and Vanden-Eijnden, Eric},
  journal={Communications in Mathematical Sciences},
  volume={8},
  number={1},
  pages={217--233},
  year={2010},
  publisher={International Press of Boston}
}

@article{arridge2019solving,
  title={Solving inverse problems using data-driven models},
  author={Arridge, Simon and Maass, Peter and {\"O}ktem, Ozan and Sch{\"o}nlieb, Carola-Bibiane},
  journal={Acta Numerica},
  doi={10.1017/S0962492919000059},
  volume={28},
  pages={1--174},
  year={2019},
  publisher={Cambridge University Press}
}

@article{soh2019learning,
  title={Learning semidefinite regularizers},
  author={Soh, Yong Sheng and Chandrasekaran, Venkat},
  journal={Foundations of Computational Mathematics},
  volume={19},
  number={2},
  pages={375--434},
  year={2019},
  publisher={Springer}
}

@article{cybenko1989approximation,
  title={Approximation by superpositions of a sigmoidal function},
  author={Cybenko, George},
  journal={Mathematics of Control, Signals and Systems},
  volume={2},
  number={4},
  pages={303--314},
  year={1989},
  publisher={Springer}
}

@article{pinkus1999approximation,
  title={{Approximation theory of the MLP model in neural networks}},
  author={Pinkus, Allan},
  journal={Acta Numerica},
  volume={8},
  pages={143--195},
  year={1999},
  publisher={Cambridge University Press}
}

@article{funahashi1989approximate,
  title={On the approximate realization of continuous mappings by neural networks},
  author={Funahashi, Ken-Ichi},
  journal={Neural networks},
  volume={2},
  number={3},
  pages={183--192},
  year={1989},
  publisher={Elsevier}
}

@inproceedings{carroll1989construction,
  title={Construction of neural nets using the {Radon} transform},
  author={Carroll and Dickinson},
  booktitle={International 1989 joint conference on neural networks},
  pages={607--611},
  year={1989},
  organization={IEEE}
}

@inproceedings{stinchcombe1989universal,
  title={Universal approximation using feedforward networks with non-sigmoid hidden layer activation functions},
  author={Stinchcombe},
  booktitle={International 1989 Joint Conference on Neural Networks},
  pages={613--617},
  year={1989},
  organization={IEEE}
}

@article{devore2021neural,
  title={Neural network approximation},
  author={DeVore, Ronald and Hanin, Boris and Petrova, Guergana},
  journal={Acta Numerica},
  volume={30},
  pages={327--444},
  year={2021},
  publisher={Cambridge University Press}
}

@inproceedings{rahimi2008uniform,
  title={Uniform approximation of functions with random bases},
  author={Rahimi, Ali and Recht, Benjamin},
  booktitle={2008 46th Annual Allerton Conference on Communication, Control, and Computing},
  pages={555--561},
  year={2008},
  organization={IEEE}
}

@article{rahimi2008weighted,
  title={{Weighted sums of random kitchen sinks: Replacing minimization with randomization in learning}},
  author={Rahimi, Ali and Recht, Benjamin},
  journal={Advances in Neural Information Processing Systems},
  volume={21},
  year={2008}
}

@article{rahimi2007random,
  title={Random features for large-scale kernel machines},
  author={Rahimi, Ali and Recht, Benjamin},
  journal={Advances in Neural Information Processing Systems},
  volume={20},
  year={2007}
}

@book{williams2006gaussian,
  title={Gaussian Processes for Machine Learning},
  author={Williams, Christopher K and Rasmussen, Carl Edward},
  year={2006},
  publisher={MIT Press},
  address={Cambridge, MA}
}

@book{owhadi2019operator,
  title={{Operator-Adapted Wavelets, Fast Solvers, and Numerical Homogenization: From a Game Theoretic Approach to Numerical Approximation and Algorithm Design}},
  author={Owhadi, Houman and Scovel, Clint},
  volume={35},
  year={2019},
  publisher={Cambridge University Press}
}

@article{chen2021solving,
  title={{Solving and learning nonlinear PDEs with Gaussian processes}},
  author={Chen, Yifan and Hosseini, Bamdad and Owhadi, Houman and Stuart, Andrew M},
  journal={Journal of Computational Physics},
  volume={447},
  pages={110668},
  year={2021},
  publisher={Elsevier}
}

@article{raissi2019physics,
  title={{Physics-informed neural networks: A deep learning framework for solving forward and inverse problems involving nonlinear partial differential equations}},
  author={Raissi, Maziar and Perdikaris, Paris and Karniadakis, George E},
  journal={Journal of Computational physics},
  volume={378},
  pages={686--707},
  year={2019},
  publisher={Elsevier}
}

@article{nelsen2021random,
  title={{The random feature model for input-output maps between Banach spaces}},
  author={Nelsen, Nicholas H and Stuart, Andrew M},
  journal={SIAM Journal on Scientific Computing},
  volume={43},
  number={5},
  pages={A3212--A3243},
  year={2021},
  publisher={SIAM}
}

@article{harlim2022graph,
  title={Graph-based prior and forward models for inverse problems on manifolds with boundaries},
  author={Harlim, J. and Jiang, S. W. and Kim, H. and Sanz-Alonso, D.},
  journal={Inverse Problems},
  volume={38},
  number={3},
  pages={035006},
  year={2022},
  publisher={IOP Publishing}
}

@article{sanz2020spde,
  title = {The {{SPDE}} Approach to {{Mat{\'e}rn}} Fields: Graph Representations},
  author = {{Sanz-Alonso}, Daniel and Yang, Ruiyi},
  year = {2022},

  journal = {Statistical Science},
  volume = {37},
  number = {4},
  pages = {519--540},
  publisher = {Institute of Mathematical Statistics},
  doi = {10.1214/21-STS838},
  urldate = {2024-10-05}
}

@article{zellner1988optimal,
  title={Optimal information processing and {Bayes's} theorem},
  author={Zellner, Arnold},
  journal={The American Statistician},
  doi={10.2307/2685143},
  volume={42},
  number={4},
  pages={278--280},
  year={1988},
  publisher={Taylor \& Francis}
}

@article{zeng2025ensemble,
  title={Ensemble transport filter via optimized maximum mean discrepancy},
  author={Zeng, Dengfei and Jiang, Lijian},
  journal={Journal of Computational Physics},
  pages={114582},
  year={2025},
  publisher={Elsevier}
}

@inproceedings{al2024nonlinear,
  title={Nonlinear filtering with brenier optimal transport maps},
  author={Al-Jarrah, Mohammad and Jin, Niyizhen and Hosseini, Bamdad and Taghvaei, Amirhossein},
  year={2024},
  booktitle={Proceedings of the 41st International Conference on Machine Learning},
  
}

@article{nguyen2019like,
  title={{EM}-like learning chaotic dynamics from noisy and partial observations},
  author={Nguyen, Duong and Ouala, Said and Drumetz, Lucas and Fablet, Ronan},
  journal={arXiv preprint arXiv:1903.10335},
  year={2019}
}

@inproceedings{naesseth2018variational,
  title={{Variational sequential Monte Carlo}},
  author={Naesseth, Christian and Linderman, Scott and Ranganath, Rajesh and Blei, David},
  booktitle={International Conference on Artificial Intelligence and Statistics},
  pages={968--977},
  year={2018},
  organization={PMLR}
}

@article{maddison2017filtering,
  title={Filtering variational objectives},
  author={Maddison, Chris J and Lawson, John and Tucker, George and Heess, Nicolas and Norouzi, Mohammad and Mnih, Andriy and Doucet, Arnaud and Teh, Yee},
  journal={Advances in Neural Information Processing Systems},
  volume={30},
  year={2017}
}

@inproceedings{le2017auto,
  title={{Auto-encoding sequential Monte Carlo}},
  author={Le, T and Igl, M and Rainforth, T and Jin, T and Wood, F},
  booktitle={International Conference on Learning Representations (ICLR)},
  year={2018}
}

@inproceedings{ranganath2014black,
  title={Black box variational inference},
  author={Ranganath, Rajesh and Gerrish, Sean and Blei, David},
  booktitle={Artificial Intelligence and Statistics},
  pages={814--822},
  year={2014},
  organization={PMLR}
}

@article{karl2016deep,
  title={{Deep variational Bayes filters: Unsupervised learning of state space models from raw data}},
  author={Karl, Maximilian and Soelch, Maximilian and Bayer, Justin and Van der Smagt, Patrick},
  journal={arXiv preprint arXiv:1605.06432},
  year={2016}
}

@inproceedings{rangapuram2018deep,
  title={Deep state space models for time series forecasting},
  author={Rangapuram, Syama Sundar and Seeger, Matthias and Gasthaus, Jan and Stella, Lorenzo and Wang, Yuyang and Januschowski, Tim},
  booktitle={Proceedings of the 32nd International Conference on Neural Information Processing Systems},
  pages={7796--7805},
  year={2018}
}

@article{bocquet2020bayesian,
  title={Bayesian inference of chaotic dynamics by merging data assimilation, machine learning and expectation-maximization},
  author={Bocquet, Marc and Brajard, Julien and Carrassi, Alberto and Bertino, Laurent},
  journal={Foundations of Data Science},
  doi={10.3934/fods.2020004},
  volume={2},
  number={1},
  pages={55--80},
  year={2020},
  publisher={American Institute of Mathematical Sciences}
}

@article{drovandi2021ensemble,
  title={{Ensemble MCMC: accelerating pseudo-marginal MCMC for state space models using the ensemble Kalman filter}},
  author={Drovandi, Christopher and Everitt, Richard G and Golightly, Andrew and Prangle, Dennis and others},
  journal={Bayesian Analysis},
  year={2021},
  publisher={International Society for Bayesian Analysis}
}

@article{stroud2010ensemble,
  title={{An ensemble Kalman filter and smoother for satellite data assimilation}},
  author={Stroud, Jonathan R and Stein, Michael L and Lesht, Barry M and Schwab, David J and Beletsky, Dmitry},
  journal={Journal of the American Statistical Association},
  volume={105},
  number={491},
  pages={978--990},
  year={2010},
  publisher={Taylor \& Francis}
}

@article{stroud2018bayesian,
  title={{A Bayesian adaptive ensemble Kalman filter for sequential state and parameter estimation}},
  author={Stroud, Jonathan R and Katzfuss, Matthias and Wikle, Christopher K},
  journal={Monthly Weather Review},
  volume={146},
  number={1},
  pages={373--386},
  year={2018}
}

@article{stroud2007sequential,
  title={{Sequential state and variance estimation within the ensemble Kalman filter}},
  author={Stroud, Jonathan R and Bengtsson, Thomas},
  journal={Monthly Weather Review},
  volume={135},
  number={9},
  pages={3194--3208},
  year={2007}
}

@article{pulido2018stochastic,
  title={{Stochastic parameterization identification using ensemble Kalman filtering combined with maximum likelihood methods}},
  author={Pulido, Manuel and Tandeo, Pierre and Bocquet, Marc and Carrassi, Alberto and Lucini, Magdalena},
  journal={Tellus A: Dynamic Meteorology and Oceanography},
  volume={70},
  number={1},
  pages={1--17},
  year={2018},
  publisher={Taylor \& Francis}
}

@article{carrassi2017estimating,
  title={Estimating model evidence using data assimilation},
  author={Carrassi, Alberto and Bocquet, Marc and Hannart, Alexis and Ghil, Michael},
  journal={Quarterly Journal of the Royal Meteorological Society},
  doi={10.1002/qj.2972},
  volume={143},
  number={703},
  pages={866--880},
  year={2017},
  publisher={Wiley Online Library}
}

@article{dempster1977maximum,
  title={{Maximum likelihood from incomplete data via the EM algorithm}},
  author={Dempster, Arthur P and Laird, Nan M and Rubin, Donald B},
  journal={Journal of the Royal Statistical Society: Series B (Methodological)},
  volume={39},
  number={1},
  pages={1--22},
  year={1977},
  publisher={Wiley Online Library}
}

@article{wei1990monte,
  title={{A Monte Carlo implementation of the EM algorithm and the poor man's data augmentation algorithms}},
  author={Wei, Greg CG and Tanner, Martin A},
  journal={Journal of the American Statistical Association},
  doi={10.2307/2290005},
  volume={85},
  number={411},
  pages={699--704},
  year={1990},
  publisher={Taylor \& Francis Group}
}

@article{metref2019estimating,
  title={{Estimating model evidence using ensemble-based data assimilation with localization--The model selection problem}},
  author={Metref, Sammy and Hannart, Alexis and Ruiz, Juan and Bocquet, Marc and Carrassi, Alberto and Ghil, Michael},
  journal={Quarterly Journal of the Royal Meteorological Society},
  volume={145},
  number={721},
  pages={1571--1588},
  year={2019},
  publisher={Wiley Online Library}
}

@article{tandeo2015offline,
  title={{Offline parameter estimation using EnKF and maximum likelihood error covariance estimates: Application to a subgrid-scale orography parametrization}},
  author={Tandeo, Pierre and Pulido, Manuel and Lott, Fran{\c{c}}ois},
  journal={Quarterly Journal of the Royal Meteorological Society},
  volume={141},
  number={687},
  pages={383--395},
  year={2015},
  publisher={Wiley Online Library}
}

@article{sanz2015long,
  title={Long-time asymptotics of the filtering distribution for partially observed chaotic dynamical systems},
  author={Sanz-Alonso, Daniel and Stuart, Andrew M},
  journal={SIAM/ASA Journal on Uncertainty Quantification},
  volume={3},
  number={1},
  pages={1200--1220},
  year={2015},
  publisher={SIAM}
}

@article{ueno2014iterative,
  title={Iterative algorithm for maximum-likelihood estimation of the observation-error covariance matrix for ensemble-based filters},
  author={Ueno, Genta and Nakamura, Nagatomo},
  journal={Quarterly Journal of the Royal Meteorological Society},
  volume={140},
  number={678},
  pages={295--315},
  year={2014},
  publisher={Wiley Online Library}
}

@article{dreano2017estimating,
  title={{Estimating model-error covariances in nonlinear state-space models using Kalman smoothing and the expectation--maximization algorithm}},
  author={Dreano, Denis and Tandeo, Pierre and Pulido, Manuel and Ait-El-Fquih, Boujemaa and Chonavel, Thierry and Hoteit, Ibrahim},
  journal={Quarterly Journal of the Royal Meteorological Society},
  volume={143},
  number={705},
  pages={1877--1885},
  year={2017},
  publisher={Wiley Online Library}
}

@article{ueno2016bayesian,
  title={Bayesian estimation of the observation-error covariance matrix in ensemble-based filters},
  author={Ueno, Genta and Nakamura, Nagatomo},
  journal={Quarterly Journal of the Royal Meteorological Society},
  volume={142},
  number={698},
  pages={2055--2080},
  year={2016},
  publisher={Wiley Online Library}
}

@article{cocucci2021model,
  title={{Model error covariance estimation in particle and ensemble Kalman filters using an online expectation--maximization algorithm}},
  author={Cocucci, Tadeo J and Pulido, Manuel and Lucini, Magdalena and Tandeo, Pierre},
  journal={Quarterly Journal of the Royal Meteorological Society},
  volume={147},
  number={734},
  pages={526--543},
  year={2021},
  publisher={Wiley Online Library}
}

@article{raissi2018multistep,
  title={Multistep neural networks for data-driven discovery of nonlinear dynamical systems},
  author={Raissi, Maziar and Perdikaris, Paris and Karniadakis, George Em},
  journal={arXiv preprint arXiv:1801.01236},
  year={2018}
}

@book{brunton2019data,
  title={Data-driven Science and Engineering: Machine Learning, Dynamical Systems, and Control},
  author={Brunton, Steven L and Kutz, J Nathan},
  year={2019},
  publisher={Cambridge University Press}
}

@article{harlim2021machine,
  title={Machine learning for prediction with missing dynamics},
  author={Harlim, J. and Jiang, S. W. and Liang, S. and Yang, H.},
  journal={Journal of Computational Physics},
  volume={428},
  pages={109922},
  year={2021},
  publisher={Elsevier}
}

@article{levine_framework_2022,
  title = {A Framework for Machine Learning of Model Error in Dynamical Systems},
  author = {Levine, Matthew and Stuart, Andrew},
  year = {2022},
  journal = {Communications of the American Mathematical Society},
  volume = {2},
  number = {07},
  pages = {283--344},
  doi = {10.1090/cams/10},
  urldate = {2023-03-17},
  langid = {english}
}

@article{fraccaro2017disentangled,
  title={A disentangled recognition and nonlinear dynamics model for unsupervised learning},
  author={Fraccaro, Marco and Kamronn, Simon and Paquet, Ulrich and Winther, Ole},
  journal={Advances in neural information processing systems},
  volume={30},
  year={2017}
}

@article{andrieu2010particle,
  title={{Particle Markov chain Monte Carlo methods}},
  author={Andrieu, Christophe and Doucet, Arnaud and Holenstein, Roman},
  journal={Journal of the Royal Statistical Society: Series B (Statistical Methodology)},
  doi={10.1111/j.1467-9868.2009.00736.x},
  volume={72},
  number={3},
  pages={269--342},
  year={2010},
  publisher={Wiley Online Library}
}

@article{hannart2016dada,
  title={{DADA}: data assimilation for the detection and attribution of weather and climate-related events},
  author={Hannart, Alexis and Carrassi, Alberto and Bocquet, Marc and Ghil, Michael and Naveau, Philippe and Pulido, Manuel and Ruiz, Juan and Tandeo, Pierre},
  journal={Climatic Change},
  doi={10.1007/s10584-016-1595-3},
  volume={136},
  number={2},
  pages={155--174},
  year={2016},
  publisher={Springer}
}

@article{andrieu2009pseudo,
	title = {The pseudo-marginal approach for efficient {Monte} {Carlo} computations},
	volume = {37},
	doi = {10.1214/07-AOS574},
	number = {2},
	urldate = {2024-10-02},
	journal = {The Annals of Statistics},
	author = {Andrieu, Christophe and Roberts, Gareth O.},

	year = {2009},
	pages = {697--725},
}

@inproceedings{kingma2013auto,
  author    = {Diederik P. Kingma and
               Max Welling},
  title     = {{Auto-encoding variational Bayes}},
  booktitle = {2nd International Conference on Learning Representations},
  year      = {2014},
}

@article{ishizone2020ensemble,
  title={{Ensemble Kalman variational objectives: nonlinear latent trajectory inference with a hybrid of variational inference and ensemble Kalman filter}},
  author={Ishizone, Tsuyoshi and Higuchi, Tomoyuki and Nakamura, Kazuyuki},
  journal={arXiv preprint arXiv:2010.08729},
  year={2020}
}

@article{delsole2010state,
  title={State and parameter estimation in stochastic dynamical models},
  author={DelSole, Timothy and Yang, Xiaosong},
  journal={Physica D: Nonlinear Phenomena},
  volume={239},
  number={18},
  pages={1781--1788},
  year={2010},
  publisher={Elsevier}
}

@inproceedings{krishnan2017structured,
  title={Structured inference networks for nonlinear state space models},
  author={Krishnan, Rahul and Shalit, Uri and Sontag, David},
  booktitle={Proceedings of the AAAI Conference on Artificial Intelligence},
  volume={31},
  year={2017}
}

@article{brajard2020combining,
  title={Combining data assimilation and machine learning to emulate a dynamical model from sparse and noisy observations: A case study with the {Lorenz} 96 model},
  author={Brajard, Julien and Carrassi, Alberto and Bocquet, Marc and Bertino, Laurent},
  journal={Journal of Computational Science},
  doi={10.1016/j.jocs.2020.101171},
  volume={44},
  pages={101171},
  year={2020},
  publisher={Elsevier}
}

@article{gottwald2021supervised,
  title={Supervised learning from noisy observations: Combining machine-learning techniques with data assimilation},
  author={Gottwald, G. A. and Reich, S.},
  journal={Physica D: Nonlinear Phenomena},
  doi={10.1016/j.physd.2021.132911},
  volume={423},
  pages={132911},
  year={2021},
  publisher={Elsevier}
}

@article{caflisch1998monte,
  title={{Monte Carlo and quasi-Monte Carlo methods}},
  author={Caflisch, Russel E},
  journal={Acta Numerica},
  volume={7},
  pages={1--49},
  year={1998},
  publisher={Cambridge University Press}
}

@misc{owen2013monte,
  title={{Monte Carlo theory, methods and examples}},
  author={Owen, Art B},
  year={2013},
  url={https://artowen.su.domains/mc/}
}

@book{trefethen2022numerical,
  title={Numerical Linear Algebra},
  author={Trefethen, Lloyd N and Bau, David},
  year={2022},
  publisher={SIAM}
}

@book{grimmett2020probability,
  title={Probability and Random Processes},
  author={Grimmett, Geoffrey and Stirzaker, David},
  year={2020},
  publisher={Oxford university press}
}

@book{wasserman2013all,
  title={All of Statistics: A Concise Course in Statistical Inference},
  author={Wasserman, Larry},
  year={2013},
  publisher={Springer Science \& Business Media}
}

@article{muller2019neural,
  title={Neural importance sampling},
  author={M{\"u}ller, Thomas and McWilliams, Brian and Rousselle, Fabrice and Gross, Markus and Nov{\'a}k, Jan},
  journal={ACM Transactions on Graphics (ToG)},
  volume={38},
  number={5},
  pages={1--19},
  year={2019},
  publisher={ACM New York, NY, USA}
}

@inproceedings{hoffman2019neutra,
  title={{Neutra-lizing bad geometry in Hamiltonian Monte Carlo using neural transport}},
  author={Hoffman, Matthew and Sountsov, Pavel and Dillon, Joshua V and Langmore, Ian and Tran, Dustin and Vasudevan, Srinivas},
  booktitle={1st Symposium on Advances in Approximate Bayesian Inference},
  year={2018}
}

@article{parno2018transport,
  title={{Transport map accelerated Markov chain Monte Carlo}},
  author={Parno, Matthew D and Marzouk, Youssef M},
  journal={SIAM/ASA Journal on Uncertainty Quantification},
  volume={6},
  number={2},
  pages={645--682},
  year={2018},
  publisher={SIAM}
}

@article{iglesias2021adaptive,
  title={{Adaptive regularisation for ensemble Kalman inversion}},
  author={Iglesias, M. A. and Yang, Y.},
  journal={Inverse Problems},
  volume={37},
  number={2},
  pages={025008},
  year={2021},
  publisher={IOP Publishing}
}

@inproceedings{albergostochastic,
  title={Stochastic Interpolants with Data-Dependent Couplings},
  author={Albergo, Michael Samuel and Goldstein, Mark and Boffi, Nicholas Matthew and Ranganath, Rajesh and Vanden-Eijnden, Eric},
  booktitle={Forty-first International Conference on Machine Learning},
  year={2024}
}

@article{chemseddine2024conditional,
	title = {Conditional {Wasserstein} {Distances} with {Applications} in {Bayesian} {OT} {Flow} {Matching}},
	volume = {26},
	url = {http://jmlr.org/papers/v26/24-0586.html},
	number = {141},
	urldate = {2026-04-19},
	journal = {Journal of Machine Learning Research},
	author = {Chemseddine, Jannis and Hagemann, Paul and Steidl, Gabriele and Wald, Christian},
	year = {2025},
	pages = {1--47},
}

@article{kerrigan2024dynamic,
  title={Dynamic Conditional Optimal Transport through Simulation-Free Flows},
  author={Kerrigan, Gavin and Migliorini, Giosue and Smyth, Padhraic},
  journal={arXiv preprint arXiv:2404.04240},
  year={2024}
}

@article{wu2023learning,
  title = {Learning About Structural Errors in Models of Complex Dynamical Systems},
  author = {Wu, Jin-Long and Levine, Matthew E. and Schneider, Tapio and Stuart, Andrew},
  year = {2024},

  journal = {Journal of Computational Physics},
  pages = {113157},
  doi = {10.1016/j.jcp.2024.113157},
  urldate = {2024-06-03}
}

@article{pulido2019sequential,
  title={{Sequential Monte Carlo with kernel embedded mappings: The mapping particle filter}},
  author={Pulido, Manuel and van Leeuwen, Peter Jan},
  journal={Journal of Computational Physics},
  doi={10.1016/j.jcp.2019.06.060},
  volume={396},
  pages={400--415},
  year={2019},
  publisher={Elsevier}
}

@article{murphy1989skill,
  title={Skill scores and correlation coefficients in model verification},
  author={Murphy, Allan H and Epstein, Edward S},
  journal={Monthly Weather Review},
  doi={10.1175/1520-0493(1989)117<0572:SSACCI>2.0.CO;2},
  volume={117},
  number={3},
  pages={572--582},
  year={1989},
  publisher={American Meteorological Society}
}

@article{iglesias2016regularizing,
  title={{A regularizing iterative ensemble Kalman method for PDE-constrained inverse problems}},
  author={Iglesias, M. A.},
  journal={Inverse Problems},
  volume={32},
  number={2},
  pages={025002},
  year={2016},
  publisher={IOP Publishing}
}

@article{huang2021unscented2,
  title={{Unscented Kalman inversion: Efficient Gaussian approximation to the posterior distribution}},
  author={Huang, Daniel Z and Huang, Jiaoyang},
  journal={arXiv preprint arXiv:2103.00277},
  year={2021}
}

@article{bell1993iterated,
  title={{The iterated Kalman filter update as a Gauss-Newton method}},
  author={Bell, B. M. and Cathey, F. W.},
  journal={IEEE Transactions on Automatic Control},
  volume={38},
  number={2},
  pages={294--297},
  year={1993},
  publisher={IEEE}
}

@article{bell1994iterated,
  title={{The iterated Kalman smoother as a Gauss--Newton method}},
  author={Bell, B. M.},
  journal={SIAM Journal on Optimization},
  volume={4},
  number={3},
  pages={626--636},
  year={1994},
  publisher={SIAM}
}

@article{evensen2000ensemble,
  title={{An ensemble Kalman smoother for nonlinear dynamics}},
  author={Evensen, Geir and van Leeuwen, Peter Jan},
  journal={Monthly Weather Review},
  volume={128},
  number={6},
  pages={1852--1867},
  year={2000}
}

@article{anderson2001ensemble,
  title={{An ensemble adjustment Kalman filter for data assimilation}},
  author={Anderson, J. L.},
  journal={Monthly Weather Review},
  doi={10.1175/1520-0493(2001)129<2884:AEAKFF>2.0.CO;2},
  volume={129},
  number={12},
  pages={2884--2903},
  year={2001},
  publisher={American Meteorological Society}
}

@article{chada2021iterative,
  title={{Iterative ensemble Kalman methods: A unified perspective with some new variants}},
  author={Chada, Neil K and Chen, Yuming and Sanz-Alonso, Daniel},
  journal={Foundations of Data Science},
  volume={3},
  number={3},
  pages={331--369},
  year={2021},
  publisher={Foundations of Data Science}
}

@article{tippett2003ensemble,
  title={Ensemble square root filters},
  author={Tippett, M. K. and Anderson, J. L. and Bishop, C. H. and Hamill, T. M. and Whitaker, J. S.},
  journal={Monthly Weather Review},
  doi={10.1175/1520-0493(2003)131<1485:ESRF>2.0.CO;2},
  volume={131},
  number={7},
  pages={1485--1490},
  year={2003}
}

@article{kwiatkowski2015convergence,
  title={{Convergence of the square root ensemble Kalman filter in the large ensemble limit}},
  author={Kwiatkowski, E. and Mandel, J.},
  journal={SIAM/ASA Journal on Uncertainty Quantification},
  volume={3},
  number={1},
  pages={1--17},
  year={2015},
  publisher={SIAM}
}

@article{blei2017variational,
  title={Variational inference: A review for statisticians},
  author={Blei, D. M. and Kucukelbir, A. and McAuliffe, J. D.},
  journal={Journal of the American Statistical Association},
  volume={112},
  number={518},
  pages={859--877},
  year={2017},
  publisher={Taylor \& Francis}
}

@article{wainwright2008graphical,
  title={Graphical models, exponential families, and variational inference},
  author={Wainwright, Martin J and Jordan, Michael I and others},
  journal={Foundations and Trends{\textregistered} in Machine Learning},
  doi={10.1561/2200000001},
  volume={1},
  number={1--2},
  pages={1--305},
  year={2008},
  publisher={Now Publishers, Inc.}
}

@article{mandel2011convergence,
  title={{On the convergence of the ensemble Kalman filter}},
  author={Mandel, J. and Cobb, L. and Beezley, J. D.},
  journal={Applications of Mathematics},
  volume={56},
  number={6},
  pages={533--541},
  year={2011},
  publisher={Springer}
}

@book{robert2013monte,
  title={{Monte Carlo Statistical Methods}},
  author={Robert, Christian and Casella, George},
  year={2013},
  publisher={Springer Science \& Business Media}
}

@book{liu2008monte,
	Author = {Liu, J. S.},
	Publisher = {Springer Science \& Business Media},
	Title = {{Monte Carlo Strategies in Scientific Computing}},
	Year = {2008}}

@article{sanz2024first,
  title={{A First Course in Monte Carlo Methods}},
  author={Sanz-Alonso, Daniel and Al-Ghattas, Omar},
  journal={arXiv preprint arXiv:2405.16359},
  year={2024}
}

@article{lorenc2000met,
  title={{The Met. Office global three-dimensional variational data assimilation scheme}},
  author={Lorenc, A. C. and Ballard, S. P. and Bell, R. S. and Ingleby, N. B. and Andrews, P. L. F.  and Barker, D. M. and Bray, J. R. and Clayton, A. M. and Dalby, T. and Li, D. and others},
  journal={Quarterly Journal of the Royal Meteorological Society},
  volume={126},
  number={570},
  pages={2991--3012},
  year={2000},
  publisher={Wiley Online Library}
}

@book{crisan2011oxford,
  title={{The Oxford Handbook of Nonlinear Filtering}},
  author={Crisan, D. and Rozovskii, B.},
  year={2011},
  publisher={{Oxford University Press}}
}

@book{bain2008fundamentals,
  title={{Fundamentals of Stochastic Filtering}},
  author={Bain, Alan and Crisan, Dan},
  volume={60},
  year={2008},
  publisher={Springer Science \& Business Media}
}

@article{reich2019data,
  title={{Data assimilation: the Schr{\"o}dinger perspective}},
  author={Reich, S.},
  journal={Acta Numerica},
  doi={10.1017/S0962492919000011},
  volume={28},
  pages={635--711},
  year={2019},
  publisher={{Cambridge University Press}}
}

@book{sarkka2013bayesian,
  title={{Bayesian Filtering and Smoothing}},
  author={S{\"a}rkk{\"a}, S.},
  volume={3},
  year={2013},
  publisher={{Cambridge University Press}}
}

@book{asch2016data,
  title={Data Assimilation: Methods, Algorithms, and Applications},
  author={Asch, M. and Bocquet, M. and Nodet, M.},
  volume={11},
  year={2016},
  publisher={SIAM}
}

@book{chopin2020introduction,
  title={{An Introduction to Sequential Monte Carlo}},
  author={Chopin, N. and Papaspiliopoulos, O.},
  year={2020},
  publisher={Springer}
}

@book{vogel2002computational,
  title={{Computational Methods for Inverse Problems}},
  author={Vogel, C. R},
  year={2002},
  publisher={SIAM}
}

@book{tikhonov1977solutions,
  title={{Solutions of Ill-posed Problems}},
  author={Tikhonov, A. N. and Arsenin, V. Y.},
  publisher={Washington,  Winston \& Sons},
  year={1977}
}

@book{bal2012introduction,
  title={{Introduction to Inverse Problems}},
  author={Bal, G.},
  publisher={Unpublished, Department of Applied Physics and Applied Mathematics, Columbia University, New York},
  year={2012}
}

@book{miller2003fundamentals,
  title={{Fundamentals of Inverse Problems}},
  author={Miller, E. L. and Karl, W. C.},
  publisher={Unpublished},
  year={2003},
  url={https://ece.northeastern.edu/fac-ece/elmiller/eceg398f03/notes.pdf}
}

@article{sanz2020bayesian,
  title={Bayesian Update with Importance Sampling: Required Sample Size},
  author={Sanz-Alonso, Daniel and Wang, Zijian},
  journal={Entropy},
  volume={23},
  number={1},
  pages={22},
  year={2021},
  publisher={Multidisciplinary Digital Publishing Institute}
}

@article{bottou2018optimization,
  title={Optimization methods for large-scale machine learning},
  author={Bottou, L. and Curtis, F. E. and Nocedal, J.},
  journal={SIAM Review},
  volume={60},
  number={2},
  pages={223--311},
  year={2018},
  publisher={SIAM}
}

@article{latz2020well,
  title={{On the well-posedness of Bayesian inverse problems}},
  author={Latz, J.},
  journal={SIAM/ASA Journal on Uncertainty Quantification},
  volume={8},
  number={1},
  pages={451--482},
  year={2020},
  publisher={SIAM}
}

@article{kamilov2023plug,
  title={Plug-and-play methods for integrating physical and learned models in computational imaging: Theory, algorithms, and applications},
  author={Kamilov, Ulugbek S and Bouman, Charles A and Buzzard, Gregery T and Wohlberg, Brendt Egon},
  journal={IEEE Signal Processing Magazine},
  volume={40},
  number={LA-UR-22-23403},
  year={2023},
  publisher={Los Alamos National Laboratory (LANL), Los Alamos, NM (United States)}
}

@article{romano2017little,
  title={The Little Engine That Could:: Regularization by Denoising (RED)},
  author={Romano, Yaniv and Elad, Michael and Milanfar, Peyman},
  journal={SIAM Journal on Imaging Sciences},
  volume={10},
  number={4},
  pages={1804--1844},
  year={2017},
  publisher={Society for Industrial and Applied Mathematics}
}

@article{chada2019tikhonov,
  title={{Tikhonov regularization within ensemble Kalman inversion}},
  author={Chada, Neil K and Stuart, Andrew M and Tong, Xin T},
  journal={SIAM Journal on Numerical Analysis},
  volume={58},
  number={2},
  pages={1263--1294},
  year={2020},
  publisher={SIAM}
}

@article{kovachki2019ensemble,
  title={{Ensemble Kalman inversion: A derivative-free technique for machine learning tasks}},
  author={Kovachki, N. B. and Stuart, A. M.},
journal={Inverse Problems},
  volume={35},
  number={9},
  pages={095005},
  year={2019},
  publisher={IOP Publishing}
}

@article{trillos2017consistency,
  title={{On the consistency of graph-based Bayesian semi-supervised learning and the scalability of sampling algorithms}},
  author={Garcia Trillos, N. and Kaplan, Z. and Samakhoana, T. and Sanz-Alonso, D.},
  journal={Journal of Machine Learning Research},
  volume={21},
  number={28},
  pages={1--47},
  year={2020}
}

@article{trillos2018bayesian,
  title={{The Bayesian update: variational formulations and gradient flows}},
  author={Garcia Trillos, Nicolas and Sanz-Alonso, Daniel},
  journal={Bayesian Analysis},
  volume={15},
  number={1},
  pages={29--56},
  year={2020},
  publisher={International Society for Bayesian Analysis}
}

@article{sanz2018importance,
  title={Importance sampling and necessary sample size: An information theory approach},
  author={Sanz-Alonso, Daniel},
  journal={SIAM/ASA Journal on Uncertainty Quantification},
  volume={6},
  number={2},
  pages={867--879},
  year={2018},
  publisher={SIAM}
}

@article{agapiou2017sparsity,
  title={{Sparsity-promoting and edge-preserving maximum \textit{a posteriori} estimators in non-parametric Bayesian inverse problems}},
  author={ S. Agapiou and  M. Burger and  M. Dashti and T. Helin.},
  journal={Inverse Problems},
  doi={10.1088/1361-6420/aaacac},
  volume={34},
  number={4},
  pages={045002},
  year = {2017}}

@article{agapiou2017importance,
  title={Importance sampling: Intrinsic dimension and computational cost},
  author={S. Agapiou and  O. Papaspiliopoulos and D. Sanz-Alonso and A. M. Stuart},
  journal={Statistical Science},
  doi={10.1214/17-STS611},
  volume={32},
  number={3},
  pages = {405--431},
  year = {2017}
}

@book{bishop,
  title={{Pattern Recognition and Machine Learning}},
  author={C. M. Bishop},
  publisher={Springer},
  year = {2006}  
}

@article{chen2012ensemble,
  title={Ensemble randomized maximum likelihood method as an iterative ensemble smoother},
  author={Y. Chen and D. Oliver},
  journal={Mathematical Geosciences},
  volume={44},
   number={1},
  pages = {1--26},
  year = {2002}  
}

@article{cotter2010approximation,
  title={{Approximation of Bayesian inverse problems for PDE's}},
  author={S. Cotter and M. Dashti and A. M. Stuart},
  journal={SIAM Journal on Numerical Analysis},
  volume={48},
   number={1},
  pages = {322--345},
  year = {2010}  
}

@article{dashti2013bayesian,
  title={ Bayesian approach to inverse problems},
  author={M. Dashti and A. M. Stuart},
  journal={Handbook of Uncertainty Quantification},
  pages={311--428},
  year = {2017}  
}

@article{dashti2013map,
  title={{MAP estimators  and their consistency in Bayesian nonparametric  inverse problems}},
  author={M. Dashti and K. J. H. Law and A. M. Stuart and J. Voss},
  journal={Inverse Problems},
  volume={29},
   number={9},
   pages = {095017},
  year = {2013}  
}

@incollection{doucet2001introduction,
  title={{An introduction to sequential Monte Carlo methods}},
  author={Doucet, Arnaud and Freitas, Nando de and Gordon, Neil},
  booktitle={Sequential Monte Carlo Methods in Practice},
  pages={3--14},
  year={2001},
  publisher={Springer}
}

@article{doucet2000sequential,
  title={{On sequential Monte Carlo sampling methods for Bayesian filtering}},
  author={Doucet, Arnaud and Godsill, Simon and Andrieu, Christophe},
  journal={Statistics and Computing},
  volume={10},
  number={3},
  pages={197--208},
  year={2000},
  publisher={Springer}
}

@article{carrillo2018analytical,
  title={An analytical framework for consensus-based global optimization method},
  author={Carrillo, Jos{\'e} A and Choi, Young-Pil and Totzeck, Claudia and Tse, Oliver},
  journal={Mathematical Models and Methods in Applied Sciences},
  volume={28},
  number={06},
  pages={1037--1066},
  year={2018},
  publisher={World Scientific}
}

@article{pinnau2017consensus,
  title={A consensus-based model for global optimization and its mean-field limit},
  author={Pinnau, Ren{\'e} and Totzeck, Claudia and Tse, Oliver and Martin, Stephan},
  journal={Mathematical Models and Methods in Applied Sciences},
  volume={27},
  number={01},
  pages={183--204},
  year={2017},
  publisher={World Scientific}
}

@article{emerick2013ensemble,
  title={Ensemble smoother with multiple data assimilation},
  author={Emerick, Alexandre A and Reynolds, Albert C},
  journal={Computers \& Geosciences},
  volume={55},
  pages={3--15},
  year={2013},
  publisher={Elsevier}
}

@book{engl1996regularization,
  title={{Regularization of Inverse Problems}},
  author={H. Engl and M. Hanke and A. Neubauer},
   publisher={{Springer Science and Business Media}},
  year = {1996}  
}

@article{gamerman2006markov,
 title={{Markov chain Monte Carlo: stochastic simulation for Bayesian inference}},
  author={D. Gamerman and H. Lopes},
  journal={{CRC Press}}, 
   year = {2006}
}

@article{gibbs2002choosing,
 title={On choosing and bounding probability metrics},
  author={A. Gibbs and F. Su},
  journal={{International Statistical Review}}, 
   pages = {419--435},
    volume={70},
   number={3},
   year = {2002}
}

@incollection{ghil_applications_1981,
  title={Applications of estimation theory to numerical weather prediction},
  author={Ghil, Michael and Cohn, S and Tavantzis, John and Bube, K and Isaacson, Eugene},
  booktitle={Dynamic Meteorology: Data Assimilation Methods},
  pages={139--224},
  year={1981},
  publisher={Springer}
}

@article{helin2015maximum,
 title={{Maximum a posteriori probability estimates in infinite-dimensional Bayesian inverse problems}},
  author={T. Helin and M. Burger},
  journal={{Inverse Problems}}, 
 volume={31},
   number={8},
   pages = {085009},
     year = {2015}
}

@article{iglesias2013ensemble,
 title={{Ensemble Kalman methods for inverse problems}},
  author={M. A. Iglesias and K. J. H. Law and A. M. Stuart},
  journal={{Inverse Problems}}, 
 volume={29},
   number={4},
     pages = {045001},
     year = {2014}
}

@article{iglesias2015iterative,
 title={Iterative regularization for ensemble data assimilation in reservoir models},
  author={M. A. Iglesias},
  journal={{Computational Geosciences}}, 
 volume={19},
   number={1},
     pages = {177--212},
     year = {2015}
}

@article{kalman1960new,
 title={A new approach to linear filtering and prediction problems},
  author={R. Kalman},
  journal={{Journal of Basic Engineering}}, 
 volume={82},
   number={1},
        pages = {35--45},
     year = {1960}
}

@book{kalnay2003atmospheric,
  title = {Atmospheric {{Modeling}}, {{Data Assimilation}} and {{Predictability}}},
  author = {Kalnay, Eugenia},
  year = {2002},

  publisher = {{Cambridge University Press}},
  address = {{New York}},
  isbn = {978-0-521-79629-3},
  langid = {english}
}

@article{kantas2014sequential,
 title={{Sequential Monte Carlo methods for high-dimensional inverse problems: a case study for the Navier Stokes equations}},
  author={N. Kantas and A. Beskos and A. Jasra},
  journal={{SIAM Journal on Uncertainty Quantification}}, 
 volume={2},
   number={1},
        pages = {464--489},
     year = {2014}
}

@article{kaipio2006statistical,
 title={{Statistical and Computational Inverse Problems}},
  author={J. Kaipio and E. Somersalo},
  journal={{Springer Science \& Business Media}}, 
 volume={160},
     year = {2006}
}

@article{knapik2011bayesian,
 title={{Bayesian inverse problems with Gaussian priors}},
  author={B. Knapik and A. van~der Vaart and J. van Zanten},
  journal={{Annals of Statistics}}, 
 volume={39},
   number={5},
           pages = {2626--2657},
     year = {2011}
}

@article{le2009large,
 title={{ Large sample asymptotics for the ensemble Kalman filter}},
  author={Le Gland, F. and Monbet, V. and Tran, V.},
  journal={{PhD Thesis}}, 
     year = {2009}
}

@article{law2012analysis,
 title={{Analysis of the 3DVAR filter for the partially observed Lorenz'63 model}},
  author={K. J. H. Law and A. Shukla and A. M. Stuart},
  journal={{Discrete and Continuous Dynamical Systems}}, 
  volume= {34},
  number = {3},
  pages = {1061--1078},
     year = {2014}
}

@book{law2015data,
  title={{Data Assimilation}},
  author={Law, K. J. H. and Stuart, A. M. and Zygalakis, K.},
  year={2015},
  publisher={Springer}
}

@article{lorenc1986analysis,
 title={Analysis methods for numerical weather prediction},
  author={A. Lorenc},
  journal={{Quarterly Journal of the Royal Meteorological Society}}, 
 volume={112},
    number={474},
           pages = {1177--1194},
     year = {1986}
}

@book{majda2012filtering,
 title={{Filtering Complex Turbulent Systems}},
  author={A. J. Majda and J. Harlim},
  publisher={{Cambridge University Press}}, 
  year = {2012}
}

@article{marzouk2009stochastic,
 title={{A stochastic collocation approach to Bayesian inference in inverse problems}},
  author={Y. Marzouk and D. Xiu},
  journal={{Communications in Computational Physics}}, 
    volume={6},
    number={4},
           pages = {826--847},
     year = {2009}

}

@book{meyn2012markov,
 title={{Markov Chains and Stochastic Stability}},
  author={S. Meyn and R. Tweedie},
  publisher={{ Springer Science and Business Media}}, 
  year = {2012}
}

@book{del2004feynman,
  title={{Feynman-Kac Formulae: Genealogical and Interacting Particle Systems with Applications}},
  author={Del Moral, Pierre},
  year={2004},
  publisher={Springer Science \& Business Media}
}

@article{del2006sequential,
  title={{Sequential Monte Carlo samplers}},
  author={Del Moral, Pierre and Doucet, Arnaud and Jasra, Ajay},
  journal={Journal of the Royal Statistical Society: Series B (Statistical Methodology)},
  volume={68},
  number={3},
  pages={411--436},
  year={2006},
  publisher={Wiley Online Library}
}

@article{rauch1965maximum,
 title={Maximum likelihood estimates of linear dynamic systems},
  author={H. Rauch and F. Tung and C. Striebel},
  journal={{AIAA Journal}}, 
     volume={3},
    number={8},
             pages = {1445--1450},
     year = {1965}
}

@book{reich2015probabilistic,
 title={Probabilistic Forecasting and Bayesian Data Assimilation},
  author={Reich, Sebastian and Cotter, Colin},
  publisher={{Cambridge University Press}}, 
     year = {2015}
}

@article{sanz2016gaussian,
  title={{Gaussian approximations of small noise diffusions in Kullback-Leibler divergence}},
  author={Sanz-Alonso, D. and Stuart, A. M.},
  journal={Communications in Mathematical Sciences},
  volume={15},
  number={7},
  pages={2087-2097},
  year={2017}
}

@article{schillings2017analysis,
 title={{Analysis of the ensemble Kalman filter for inverse problems}},
  author={C. Schillings and A. M. Stuart},
  journal={{SIAM Journal on Numerical Analysis}}, 
     volume={55},
    number={3},
    pages={1264--1290},
     year = {2017}
}

@article{schillings2018convergence,
  title={{Convergence analysis of ensemble Kalman inversion: the linear, noisy case}},
  author={Schillings, Claudia and Stuart, Andrew M},
  journal={Applicable Analysis},
  volume={97},
  number={1},
  pages={107--123},
  year={2018},
  publisher={Taylor \& Francis}
}

@article{stuart2010inverse,
 title={{Inverse problems: a Bayesian perspective}},
  author={A. M. Stuart},
  journal={{Acta Numerica}}, 
       pages = {451--559},
volume={19},
     year = {2010}
}

@book{tarantola2005inverse,
 title={Inverse Problem Theory and Methods for Model Parameter Estimation},
  author={A. Tarantola},
  publisher={{SIAM}}, 
     year = {2015}
}

@book{abarbanel2013predicting,
 title={{Predicting The Future: Completing Models Of Observed Complex Systems}},
  author={H. Abarbanel},
publisher={{Springer}}, 
     year = {2013}
}

@article{carrassi2018data,
 title={Data assimilation in the geosciences: An overview of methods, issues, and perspectives},
  author={A. Carrassi and M. Bocquet and L. Bertino and G. Evensen},
  journal={{Wiley Interdisciplinary Reviews: Climate Change}},
  doi={10.1002/wcc.535},
  volume = {9},
  number = {5},
   year = {2018}
}

@article{evensen1994sequential,
 title={{Sequential data assimilation with a nonlinear quasi-geostrophic model using Monte Carlo methods to forecast error statistics}},
  author={G. Evensen},
  journal={{Journal of Geophysical Research: Oceans}}, 
  volume = {99},
  number = {c5},
  pages = {10143--10162},
   year = {1995}
}

@article{snyder2008obstacles,
 title={Obstacles to high-dimensional particle filtering},
  author={C. Snyder and T. Bengtsson and P. Bickel and J. L. Anderson},
  journal={{Monthly Weather Review}}, 
  volume = {136},
  number = {12},
  pages = {4629--4640},
   year = {2016}
}

@article{beskos2015sequentialb,
  title={{Sequential Monte Carlo methods for Bayesian elliptic inverse problems}},
  author={Beskos, Alexandros and Jasra, Ajay and Muzaffer, Ege A and Stuart, Andrew M},
  journal={Statistics and Computing},
  volume={25},
  number={4},
  pages={727--737},
  year={2015},
  publisher={Springer}
}

@article{beskos2015sequential,
 title={{Multilevel Sequential Monte Carlo Samplers}},
  author={A. Beskos and A. Jasra and K. J. H. Law and R. Tempone and Y. Zhou},
  journal={{Stochastic Processes and their Applications}}, 
  volume = {127},
  number = {5},
  pages = {1417--1440},
   year = {1994}
}

@article{reich2011dynamical,
 title={A dynamical systems framework for intermittent data assimilation},
  author={S. Reich},
  journal={{BIT Numerical Mathematics}}, 
  volume = {51},
  number = {1},
  pages = {235--249},
   year = {2017}
}

@incollection{bickel2008sharp,
  title={Sharp failure rates for the bootstrap particle filter in high dimensions},
  author={Bickel, Peter and Li, Bo and Bengtsson, Thomas},
  booktitle={Pushing the Limits of Contemporary Statistics: Contributions in Honor of Jayanta K. Ghosh},
  pages={318--329},
  year={2008},
  publisher={Institute of Mathematical Statistics},
  doi={10.1214/074921708000000228}
}

@article{hosseini2017well,
  title={{Well-posed Bayesian inverse problems: Priors with exponential tails}},
  author={Hosseini, B. and Nigam, N.},
  journal={SIAM/ASA Journal on Uncertainty Quantification},
  volume={5},
  number={1},
  pages={436--465},
  year={2017},
  publisher={SIAM}
}

@article{hosseini2017wellb,
  title={{Well-posed Bayesian inverse problems with infinitely divisible and heavy-tailed prior measures}},
  author={Hosseini, B.},
  journal={SIAM/ASA Journal on Uncertainty Quantification},
  volume={5},
  number={1},
  pages={1024--1060},
  year={2017},
  publisher={SIAM}
}

@book{goodfellow2016deep,
  title={{Deep Learning}},
  author={Goodfellow, I. and Bengio, Y. and Courville, A.},
  year={2016},
  publisher={MIT Press}
}

@book{pavliotis2014stochastic,
  title={{Stochastic Processes and Applications: Diffusion Processes, the Fokker-Planck and Langevin Equations}},
  author={Pavliotis, G. A.},
  volume={60},
  year={2014},
  publisher={Springer}
  }

@book{sullivan2015introduction,
  title={{Introduction to Uncertainty Quantification}},
  author={Sullivan, T. J.},
  volume={63},
  year={2015},
  publisher={Springer}
}

@article{harlim2019kernel,
  title={{Kernel methods for Bayesian elliptic inverse problems on manifolds}},
  author={Harlim, John and Sanz-Alonso, Daniel and Yang, Ruiyi},
  journal={SIAM/ASA Journal on Uncertainty Quantification},
  volume={8},
  number={4},
  pages={1414--1445},
  year={2020},
  publisher={SIAM}
}

@article{lasanen2012non,
  title={{Non-Gaussian statistical inverse problems. Part I: Posterior distributions}},
  author={Lasanen, Sari},
  journal={Inverse Problems \& Imaging},
  volume={6},
  number={2},
  pages={215--266},
  year={2012}
}

@article{lasanen2012nonb,
  title={{Non-Gaussian statistical inverse problems. Part II: Posterior convergence for approximated unknowns}},
  author={Lasanen, Sari},
  journal={Inverse Problems \& Imaging},
  volume={6},
  number={2},
  pages = {267},
  year={2012}
}

@article{calvello22,
  title = {Ensemble {{Kalman}} Methods: {{A}} Mean-Field Perspective},
  shorttitle = {Ensemble {{Kalman}} Methods},
  author = {Calvello, Edoardo and Reich, Sebastian and Stuart, Andrew M.},
  year = {2025},
  journal = {Acta Numerica},
  volume = {34},
  pages = {123--291},
  doi = {10.1017/S0962492924000060},
  urldate = {2025-07-23},
  langid = {english},
}

@book{wendland2004scattered,
  title={Scattered Data Approximation},
  author={Wendland, Holger},
  volume={17},
  year={2004},
  publisher={Cambridge University Press}
}

@inproceedings{asim2020invertible,
  title={Invertible generative models for inverse problems: mitigating representation error and dataset bias},
  author={Asim, Muhammad and Daniels, Max and Leong, Oscar and Ahmed, Ali and Hand, Paul},
  booktitle={International Conference on Machine Learning},
  pages={399--409},
  year={2020},
  organization={PMLR}
}

@inproceedings{gao2023image,
  title={Image reconstruction without explicit priors},
  author={Gao, Angela F and Leong, Oscar and Sun, He and Bouman, Katherine L},
  booktitle={ICASSP 2023-2023 IEEE International Conference on Acoustics, Speech and Signal Processing (ICASSP)},
  pages={1--5},
  year={2023},
  organization={IEEE}
}

@article{kingma2018glow,
  title={Glow: Generative flow with invertible 1x1 convolutions},
  author={Kingma, Durk P and Dhariwal, Prafulla},
  journal={Advances in Neural Information Processing Systems},
  volume={31},
  year={2018}
}

@article{jacot2018neural,
  title={Neural tangent kernel: Convergence and generalization in neural networks},
  author={Jacot, Arthur and Gabriel, Franck and Hongler, Cl{\'e}ment},
  journal={Advances in Neural Information Processing Systems},
  volume={31},
  year={2018}
}

@article{ng2001spectral,
  title={On spectral clustering: Analysis and an algorithm},
  author={Ng, Andrew and Jordan, Michael and Weiss, Yair},
  journal={Advances in Neural Information Processing Systems},
  volume={14},
  year={2001}
}

@article{von2007tutorial,
  title={A tutorial on spectral clustering},
  author={Von Luxburg, Ulrike},
  journal={Statistics and Computing},
  volume={17},
  number={4},
  pages={395--416},
  year={2007},
  publisher={Springer}
}

@article{von2008consistency,
  title={Consistency of spectral clustering},
  author={Von Luxburg, Ulrike and Belkin, Mikhail and Bousquet, Olivier},
  journal={The Annals of Statistics},
  pages={555--586},
  year={2008},
  publisher={JSTOR}
}

@article{hoffmann2022spectral,
  title={{Spectral analysis of weighted Laplacians arising in data clustering}},
  author={Hoffmann, Franca and Hosseini, Bamdad and Oberai, Assad A and Stuart, Andrew M},
  journal={Applied and Computational Harmonic Analysis},
  volume={56},
  pages={189--249},
  year={2022},
  publisher={Elsevier}
}

@article{dunlop2020large,
  title={Large data and zero noise limits of graph-based semi-supervised learning algorithms},
  author={Dunlop, Matthew M and Slep{\v{c}}ev, Dejan and Stuart, Andrew M and Thorpe, Matthew},
  journal={Applied and Computational Harmonic Analysis},
  volume={49},
  number={2},
  pages={655--697},
  year={2020},
  publisher={Elsevier}
}

@article{garcia2018continuum,
  title={{Continuum limits of posteriors in graph Bayesian inverse problems}},
  author={Garcia Trillos, N. and Sanz-Alonso, D.},
  journal={SIAM Journal on Mathematical Analysis},
  volume={50},
  number={4},
  pages={4020--4040},
  year={2018},
  publisher={SIAM}
}

@book{villani2009optimal,
  title={Optimal Transport: Old and New},
  author={Villani, C{\'e}dric},
  volume={338},
  year={2009},
  publisher={Springer}
}

@article{peyre2017computational,
  title={Computational optimal transport: with applications to data science},
  author={Peyr{\'e}, Gabriel and Cuturi, Marco},
  journal={Foundations and Trends{\textregistered} in Machine Learning},
  volume={11},
  number={5-6},
  pages={355--607},
  year={2019},
  publisher={Now Publishers, Inc.},
  doi={10.1561/2200000073}
}

@article{goodfellow2014generative,
  title={Generative adversarial nets},
  author={Goodfellow, Ian and Pouget-Abadie, Jean and Mirza, Mehdi and Xu, Bing and Warde-Farley, David and Ozair, Sherjil and Courville, Aaron and Bengio, Yoshua},
  journal={Advances in Neural Information Processing Systems},
  volume={27},
  year={2014}
}

@article{el2012bayesian,
  title={Bayesian inference with optimal maps},
  author={El Moselhy, Tarek A and Marzouk, Youssef M},
  journal={Journal of Computational Physics},
  volume={231},
  number={23},
  pages={7815--7850},
  year={2012},
  publisher={Elsevier}
}

@article{marzouk2016sampling,
  title={Sampling via measure transport: An introduction},
  author={Marzouk, Youssef and Moselhy, Tarek and Parno, Matthew and Spantini, Alessio},
  journal={Handbook of Uncertainty Quantification},
  pages={1--41},
  year={2016},
  publisher={Springer Cham, Switzerland}
}

@article{schmidhuber2015deep,
  title={Deep learning in neural networks: An overview},
  author={Schmidhuber, J{\"u}rgen},
  journal={Neural Networks},
  volume={61},
  pages={85--117},
  year={2015},
  publisher={Elsevier}
}

@article{hinton1993autoencoders,
  title={{Autoencoders, minimum description length and Helmholtz free energy}},
  author={Hinton, Geoffrey E and Zemel, Richard},
  journal={Advances in Neural Information Processing Systems},
  volume={6},
  year={1993}
}

@article{ongie2020deep,
  title={Deep learning techniques for inverse problems in imaging},
  author={Ongie, Gregory and Jalal, Ajil and Metzler, Christopher A and Baraniuk, Richard G and Dimakis, Alexandros G and Willett, Rebecca},
  journal={IEEE Journal on Selected Areas in Information Theory},
  volume={1},
  number={1},
  pages={39--56},
  year={2020},
  publisher={IEEE}
}

@article{lunz2018adversarial,
  title={Adversarial regularizers in inverse problems},
  author={Lunz, Sebastian and {\"O}ktem, Ozan and Sch{\"o}nlieb, Carola-Bibiane},
  journal={Advances in Neural Information Processing Systems},
  volume={31},
  year={2018}
}

@book{wahba1990spline,
  title={Spline Models for Observational Data},
  author={Wahba, Grace},
  year={1990},
  publisher={SIAM}
}

@article{basu1977nuissance,
 doi = {10.2307/2286800},
 author = {Debabrata Basu},
 journal = {Journal of the American Statistical Association},
 number = {358},
 pages = {355--366},
 publisher = {[American Statistical Association, Taylor & Francis, Ltd.]},
 title = {On the Elimination of Nuisance Parameters},
 urldate = {2023-09-10},
 volume = {72},
 year = {1977}
}

@article{craven1978smoothing,
  title={Smoothing noisy data with spline functions},
  author={Craven, Peter and Wahba, Grace},
  journal={Numerische Mathematik},
  volume={31},
  number={4},
  pages={377--403},
  year={1978},
  publisher={Springer}
}

@article{SMC,
  title={},
  author={Del Moral Doucet Jasra},
  journal={},
  volume={},
  number={},
  pages={},
  year={2006},
}

@article{huang2022iterated,
  title={{Iterated Kalman methodology for inverse problems}},
  author={Huang, Daniel Zhengyu and Schneider, Tapio and Stuart, Andrew M},
  journal={Journal of Computational Physics},
  doi={10.1016/j.jcp.2022.111262},
  volume={463},
  pages={111262},
  year={2022},
  publisher={Elsevier}
}

@article{kim2022hierarchical,
  title={{Hierarchical ensemble Kalman methods with sparsity-promoting generalized gamma hyperpriors}},
  author={Kim, H. and Sanz-Alonso, D. and Strang, A.},
  journal={arXiv preprint arXiv:2205.09322},
  year={2022}
}

@article{sejdinovic2013equivalence,
  title={{Equivalence of distance-based and RKHS-based statistics in hypothesis testing}},
  author={Sejdinovic, Dino and Sriperumbudur, Bharath and Gretton, Arthur and Fukumizu, Kenji},
  journal={The Annals of Statistics},
  pages={2263--2291},
  year={2013},
  publisher={JSTOR}
}

@article{rasul2020multivariate,
  title={Multivariate probabilistic time series forecasting via conditioned normalizing flows},
  author={Rasul, Kashif and Sheikh, Abdul-Saboor and Schuster, Ingmar and Bergmann, Urs and Vollgraf, Roland},
  journal={International Conference on Learning Representations},
  year={2021}
}

@article{hoffman2013stochastic,
  title={Stochastic Variational Inference},
  author={Hoffman, Matthew D and Blei, David M and Wang, Chong and Paisley, John},
  journal={Journal of Machine Learning Research},
  volume={14},
  pages={1303--1347},
  year={2013}
}

@inproceedings{hernandez2016black,
  title={Black-box alpha divergence minimization},
  author={Hernandez-Lobato, Jose and Li, Yingzhen and Rowland, Mark and Bui, Thang and Hern{\'a}ndez-Lobato, Daniel and Turner, Richard},
  booktitle={International Conference on Machine Learning},
  pages={1511--1520},
  year={2016},
  organization={PMLR}
}

@article{li2016renyi,
  title={R{\'e}nyi divergence variational inference},
  author={Li, Yingzhen and Turner, Richard E},
  journal={Advances in Neural Information Processing Systems},
  volume={29},
  year={2016}
}

@article{lambert2022variational,
  title={{Variational inference via Wasserstein gradient flows}},
  author={Lambert, Marc and Chewi, Sinho and Bach, Francis and Bonnabel, Silv{\`e}re and Rigollet, Philippe},
  journal={Advances in Neural Information Processing Systems},
  volume={35},
  pages={14434--14447},
  year={2022}
}

@article{sriperumbudur2011universality,
  title={{Universality, characteristic kernels and RKHS embedding of measures}},
  author={Sriperumbudur, Bharath K and Fukumizu, Kenji and Lanckriet, Gert RG},
  journal={Journal of Machine Learning Research},
  volume={12},
  number={7},
  year={2011}
}

@article{kim2024enhancinggaussianprocesssurrogates,
  title={{Enhancing Gaussian process surrogates for optimization and posterior approximation via random exploration}},
  author={Kim, Hwanwoo and Sanz-Alonso, Daniel},
  journal={SIAM/ASA Journal on Uncertainty Quantification},
  volume={13},
  number={3},
  pages={1054--1084},
  year={2025},
  publisher={SIAM}
}

@article{frazier2018tutorial,
  title={{A tutorial on Bayesian optimization}},
  author={Frazier, Peter I},
  journal={arXiv preprint arXiv:1807.02811},
  year={2018}
}

@article{Bayesoptgraph,
author = {Kim, Hwanwoo and Sanz-Alonso, Daniel and Yang, Ruiyi},
title = {{Optimization on manifolds via graph Gaussian processes}},
journal = {SIAM Journal on Mathematics of Data Science},
volume = {6},
number = {1},
pages = {1-25},
year = {2024},
doi = {10.1137/22M1529907},

URL = { 
    
        https://doi.org/10.1137/22M1529907
    
    

},
eprint = {    
        https://doi.org/10.1137/22M1529907
}
}

@article{sanz2024long,
  title={{Long-time accuracy of ensemble Kalman filters for chaotic dynamical systems and machine-learned dynamical systems}},
  author={Sanz-Alonso, Daniel and Waniorek, Nathan},
  journal={SIAM Journal on Applied Dynamical Systems},
  volume={24},
  number={3},
  pages={2246--2286},
  year={2025},
  publisher={SIAM}
}

@book{bengtsson_dynamic_1981,
    address = {New York},
    series = {Applied {Mathematical} {Sciences}},
    title = {Dynamic {Meteorology}: {Data} {Assimilation} {Methods}},
    isbn = {978-0-387-90632-4},
    shorttitle = {Dynamic {Meteorology}},
    url = {https://www.springer.com/gp/book/9780387906324},
    language = {en},
    number = {36},
    urldate = {2020-06-21},
    publisher = {Springer-Verlag},
    editor = {Bengtsson, L. and Ghil, M. and Källen, E.},
    year = {1981},
    doi = {10.1007/978-1-4612-5970-1},
}

@article{aronszajn1950theory,
  title={Theory of reproducing kernels},
  author={Aronszajn, Nachman},
  journal={Transactions of the American Mathematical Society},
  volume={68},
  number={3},
  pages={337--404},
  year={1950}
}

@article{silva_adaptive_2025,
    title = {An {Adaptive} {Hierarchical} {Ensemble} {Kalman} {Filter} with {Reduced} {Basis} {Models}},
    volume = {13},
    url = {https://epubs.siam.org/doi/10.1137/24M1653690},
    doi = {10.1137/24M1653690},
    number = {1},
    urldate = {2025-08-17},
    journal = {SIAM/ASA Journal on Uncertainty Quantification},
    author = {Silva, Francesco A. B. and Pagliantini, Cecilia and Veroy, Karen},
    month = mar,
    year = {2025},
    note = {Publisher: Society for Industrial and Applied Mathematics},
    pages = {140--170},
}

@phdthesis{filoche_variational_2022,
    type = {{PhD} thesis},
    title = {Variational {Data} {Assimilation} with {Deep} {Prior}. {Application} to {Geophysical} {Motion} {Estimation}},
    url = {https://theses.hal.science/tel-03986457},
    school = {Sorbonne Université},
    author = {Filoche, Arthur},
    year = {2022}
}

@article{bouallegue_rise_2024,
    title = {The {Rise} of {Data}-{Driven} {Weather} {Forecasting}: {A} {First} {Statistical} {Assessment} of {Machine} {Learning}–{Based} {Weather} {Forecasts} in an {Operational}-{Like} {Context}},
    volume = {105},
    shorttitle = {The {Rise} of {Data}-{Driven} {Weather} {Forecasting}},
    doi = {10.1175/BAMS-D-23-0162.1},
    language = {EN},
    number = {6},
    urldate = {2025-08-18},
    journal = {Bulletin of the American Meteorological Society},
    author = {Bouallègue, Zied Ben and Clare, Mariana C. A. and Magnusson, Linus and Gascón, Estibaliz and Maier-Gerber, Michael and Janoušek, Martin and Rodwell, Mark and Pinault, Florian and Dramsch, Jesper S. and Lang, Simon T. K. and Raoult, Baudouin and Rabier, Florence and Chevallier, Matthieu and Sandu, Irina and Dueben, Peter and Chantry, Matthew and Pappenberger, Florian},
    year = {2024},
    pages = {E864--E883},
}

@article{pasmans_ensemble_2025,
	title = {Ensemble {Kalman} filter in latent space using a variational autoencoder pair},
	volume = {152},
	doi = {10.1002/qj.70070},
	language = {en},
	number = {775},
	urldate = {2026-04-18},
	journal = {Quarterly Journal of the Royal Meteorological Society},
	author = {Pasmans, Ivo and Chen, Yumeng and Finn, Tobias Sebastian and Bocquet, Marc and Carrassi, Alberto},
	year = {2026},
	pages = {e70070},
}

@article{roberts_general_2004,
    title = {General state space {Markov} chains and {MCMC} algorithms},
    volume = {1},
    doi = {10.1214/154957804100000024},
    urldate = {2025-11-08},
    journal = {Probability Surveys},
    author = {Roberts, Gareth O. and Rosenthal, Jeffrey S.},
    year = {2004},
    note = {Publisher: Institute of Mathematical Statistics and Bernoulli Society},
    pages = {20--71},
}

@misc{keller_ai-based_2024,
    title = {{AI}-based data assimilation: {Learning} the functional of analysis estimation},
    shorttitle = {{AI}-based data assimilation},
    url = {http://arxiv.org/abs/2406.00390},
    urldate = {2024-06-05},
    publisher = {arXiv},
    author = {Keller, Jan D. and Potthast, Roland},
    month = jun,
    year = {2024},
    note = {arXiv:2406.00390 [physics]},
}

@misc{zinchenko_combined_2024,
    title = {Combined {Optimization} of {Dynamics} and {Assimilation} with {End}-to-{End} {Learning} on {Sparse} {Observations}},
    url = {http://arxiv.org/abs/2409.07137},
    doi = {10.48550/arXiv.2409.07137},
    urldate = {2024-09-15},
    publisher = {arXiv},
    author = {Zinchenko, Vadim and Greenberg, David S.},
    month = sep,
    year = {2024},
    note = {arXiv:2409.07137 [physics]},
}

@article{raanes_ensemble_2016,
    title = {On the ensemble {Rauch}-{Tung}-{Striebel} smoother and its equivalence to the ensemble {Kalman} smoother},
    volume = {142},
    doi = {10.1002/qj.2728},
    language = {en},
    number = {696},
    urldate = {2025-12-17},
    journal = {Quarterly Journal of the Royal Meteorological Society},
    author = {Raanes, Patrick Nima},
    year = {2016},
    pages = {1259--1264},
}

@article{ramgraber_ensemble_2023,
    title = {Ensemble transport smoothing. {Part} {I}: {Unified} framework},
    volume = {17},
    shorttitle = {Ensemble transport smoothing. {Part} {I}},
    doi = {10.1016/j.jcpx.2023.100134},
    urldate = {2025-12-21},
    journal = {Journal of Computational Physics: X},
    author = {Ramgraber, Maximilian and Baptista, Ricardo and McLaughlin, Dennis and Marzouk, Youssef},
    year = {2023},
    pages = {100134},
}

@article{banerjee_optimality_2005,
    title = {On the optimality of conditional expectation as a {Bregman} predictor},
    volume = {51},
    doi = {10.1109/TIT.2005.850145},
    number = {7},
    urldate = {2026-01-03},
    journal = {IEEE Transactions on Information Theory},
    author = {Banerjee, A. and Guo, Xin and Wang, Hui},
    year = {2005},
    pages = {2664--2669},
}

@book{williams_probability_1991,
    title = {Probability with {Martingales}},
    isbn = {978-0-521-40605-5},
    language = {en},
    urldate = {2026-01-03},
    publisher = {Cambridge University Press},
    author = {Williams, David},
    month = feb,
    year = {1991},
    doi = {10.1017/CBO9780511813658},
}

@article{vernon_nesterov_2025,
    title = {Nesterov acceleration for ensemble {Kalman} inversion and variants},
    volume = {535},
    doi = {10.1016/j.jcp.2025.114063},
    urldate = {2025-05-15},
    journal = {Journal of Computational Physics},
    author = {Vernon, Sydney and Bach, Eviatar and Dunbar, Oliver R. A.},
    month = aug,
    year = {2025},
    pages = {114063},
}

@article{hodyss_using_2026,
    title = {Using {Diffusion} {Models} to {Do} {Data} {Assimilation}},
    volume = {154},
    doi = {10.1175/MWR-D-25-0125.1},
    language = {EN},
    number = {2},
    urldate = {2026-01-28},
    journal = {Monthly Weather Review},
    author = {Hodyss, Daniel and Morzfeld, Matthias},
    year = {2026},
    pages = {165--182},
}

@article{tian_evaluating_2026,
    title = {Evaluating {Machine} {Learning} {Weather} {Models} for {Data} {Assimilation}: {Fundamental} {Limitations} in {Tangent} {Linear} and {Adjoint} {Properties}},
    volume = {53},
    shorttitle = {Evaluating {Machine} {Learning} {Weather} {Models} for {Data} {Assimilation}},
    doi = {10.1029/2025GL119402},
    language = {en},
    number = {2},
    urldate = {2026-01-31},
    journal = {Geophysical Research Letters},
    author = {Tian, Xiaoxu and Holdaway, Daniel and Kleist, Daryl},
    year = {2026},
    pages = {e2025GL119402},
}

@article{slivinski_assimilating_2025,
    title = {Assimilating {Observed} {Surface} {Pressure} {Into} {ML} {Weather} {Prediction} {Models}},
    volume = {52},
    doi = {10.1029/2024GL114396},
    language = {en},
    number = {6},
    urldate = {2025-03-14},
    journal = {Geophysical Research Letters},
    author = {Slivinski, L. C. and Whitaker, J. S. and Frolov, S. and Smith, T. A. and Agarwal, N.},
    year = {2025},
    pages = {e2024GL114396},
}

@article{pathak_model-free_2018,
    title = {Model-{Free} {Prediction} of {Large} {Spatiotemporally} {Chaotic} {Systems} from {Data}: {A} {Reservoir} {Computing} {Approach}},
    volume = {120},
    shorttitle = {Model-{Free} {Prediction} of {Large} {Spatiotemporally} {Chaotic} {Systems} from {Data}},
    doi = {10.1103/PhysRevLett.120.024102},
    number = {2},
    urldate = {2019-11-27},
    journal = {Physical Review Letters},
    author = {Pathak, Jaideep and Hunt, Brian and Girvan, Michelle and Lu, Zhixin and Ott, Edward},
    year = {2018},
    pages = {024102},
}

@article{naumann_optimal_2008,
    title = {Optimal {Jacobian} accumulation is {NP}-complete},
    volume = {112},
    doi = {10.1007/s10107-006-0042-z},
    language = {en},
    number = {2},
    urldate = {2024-04-21},
    journal = {Mathematical Programming},
    author = {Naumann, Uwe},
    year = {2008},
    pages = {427--441},
}

@article{brocker_reliability_2009,
    title = {Reliability, sufficiency, and the decomposition of proper scores},
    volume = {135},
    copyright = {Copyright © 2009 Royal Meteorological Society},
    doi = {10.1002/qj.456},
    number = {643},
    urldate = {2026-03-05},
    journal = {Quarterly Journal of the Royal Meteorological Society},
    author = {Br{\"o}cker, Jochen},
    year = {2009},
    pages = {1512--1519},
}

@inproceedings{sonderby_amortised_2017,
    title = {Amortised {MAP} {Inference} for {Image} {Super}-resolution},
    url = {https://openreview.net/forum?id=S1RP6GLle},
    language = {en},
    urldate = {2026-03-28},
    author = {Sønderby, Casper Kaae and Caballero, Jose and Theis, Lucas and Shi, Wenzhe and Huszár, Ferenc},
    month = feb,
    year = {2017},
booktitle={International Conference on Learning Representations},
}

@article{wang_artificial_2026,
    title = {Artificial {Intelligence} techniques in data assimilation: {Emerging} approaches, key challenges, and future prospects},
    shorttitle = {Artificial {Intelligence} techniques in data assimilation},
    doi = {10.1007/s11430-025-1807-8},
    language = {en},
    urldate = {2026-03-28},
    journal = {Science China Earth Sciences},
    author = {Wang, Wuxin and Ni, Weicheng and Yuan, Taikang and Huang, Lilan and Han, Tao and Duan, Boheng and Li, Xiaoyong and Zhao, Yanlai and Fei, Ben and Bai, Lei and Ren, Kaijun},
    month = mar,
    year = {2026},
}

@incollection{bach_interface_2026,
	series = {Developments in {Weather} and {Climate} {Science}},
	title = {The {Interface} of {Data} {Assimilation} and {Machine} {Learning}},
	isbn = {978-0-443-40360-6},
	number = {13},
	booktitle = {Weather and {Climate}: {Applications} of {Machine} {Learning} and {Artificial} {Intelligence}},
	publisher = {Elsevier},
	author = {Bach, Eviatar},
	collaborator = {Driscoll, Simon and Hunt, Kieran M. R. and Mansfield, Laura and Swaminathan, Ranjini and Wei, Hong and Bach, Eviatar and Peard, Alison},
	year = {2026},
}

@inproceedings{figurnov_implicit_2018,
    address = {Red Hook, NY, USA},
    series = {{NIPS}'18},
    title = {Implicit reparameterization gradients},
    url = {https://dl.acm.org/doi/10.5555/3326943.3326984},
    urldate = {2026-03-31},
    booktitle = {Proceedings of the 32nd {International} {Conference} on {Neural} {Information} {Processing} {Systems}},
    publisher = {Curran Associates Inc.},
    author = {Figurnov, Michael and Mohamed, Shakir and Mnih, Andriy},
    month = dec,
    year = {2018},
    pages = {439--450},
}

@article{pacchiardi_probabilistic_2024,
    title = {Probabilistic {Forecasting} with {Generative} {Networks} via {Scoring} {Rule} {Minimization}},
    volume = {25},
    url = {http://jmlr.org/papers/v25/23-0038.html},
    number = {45},
    urldate = {2024-11-19},
    journal = {Journal of Machine Learning Research},
    author = {Pacchiardi, Lorenzo and Adewoyin, Rilwan A. and Dueben, Peter and Dutta, Ritabrata},
    year = {2024},
    pages = {1--64},
}

@article{arcucci_convergence_2026,
    title = {The convergence of machine learning and data assimilation in {Earth} system science},
    volume = {2},
    copyright = {2026 The Author(s)},
    doi = {10.1038/s44387-026-00107-0},
    language = {en},
    number = {1},
    urldate = {2026-04-24},
    journal = {npj Artificial Intelligence},
    author = {Arcucci, Rossella and Healy, Sean and Dance, Sarah and Lei, Lili and Bach, Eviatar and Weaver, Anthony T. and Miyoshi, Takemasa and Dillon, Maria Eugenia and Draper, Clara and Schneider, Rochelle and others},
    year = {2026},
    pages = {48},
}

@book{vannitsem_statistical_2018,
    title = {Statistical {Postprocessing} of {Ensemble} {Forecasts}},
    isbn = {978-0-12-812372-0},
    language = {en},
    urldate = {2020-01-04},
    publisher = {Elsevier},
    editor = {Vannitsem, Stéphane and Wilks, Daniel S. and Messner, Jakob W.},
    year = {2018},
    doi = {10.1016/C2016-0-03244-8},
}

@article{chakraborty_new_2021,
    title = {A new framework for distance and kernel-based metrics in high dimensions},
    volume = {15},
    issn = {1935-7524, 1935-7524},
    url = {https://projecteuclid.org/journals/electronic-journal-of-statistics/volume-15/issue-2/A-new-framework-for-distance-and-kernel-based-metrics-in/10.1214/21-EJS1889.full},
    doi = {10.1214/21-EJS1889},
    number = {2},
    urldate = {2024-10-05},
    journal = {Electronic Journal of Statistics},
    author = {Chakraborty, Shubhadeep and Zhang, Xianyang},
    month = jan,
    year = {2021},
    note = {Publisher: Institute of Mathematical Statistics and Bernoulli Society},
    pages = {5455--5522},
}

@article{knoblauch_optimization-centric_2022,
    title = {An {Optimization}-centric {View} on {Bayes}' {Rule}: {Reviewing} and {Generalizing} {Variational} {Inference}},
    volume = {23},
    issn = {1533-7928},
    shorttitle = {An {Optimization}-centric {View} on {Bayes}' {Rule}},
    url = {http://jmlr.org/papers/v23/19-1047.html},
    number = {132},
    urldate = {2024-11-30},
    journal = {Journal of Machine Learning Research},
    author = {Knoblauch, Jeremias and Jewson, Jack and Damoulas, Theodoros},
    year = {2022},
    pages = {1--109},
}

@misc{bach_learning_2026,
    title = {Learning {Probabilistic} {Filters} with {Strictly} {Proper} {Scoring} {Rules}},
    url = {http://arxiv.org/abs/2606.26497},
    doi = {10.48550/arXiv.2606.26497},
    urldate = {2026-06-26},
    publisher = {arXiv},
    author = {Bach, Eviatar and Baptista, Ricardo and Bröcker, Jochen and Chen, Bohan and Stuart, Andrew},
    month = jun,
    year = {2026},
    note = {arXiv:2606.26497 [cs.LG]},
}

@misc{kaveh_amortized_2026,
    title = {{Energy}-{based} transport for amortized {Bayesian} {Inference}},
    url = {http://arxiv.org/abs/2605.15407},
    doi = {10.48550/arXiv.2605.15407},
    urldate = {2026-05-19},
    publisher = {arXiv},
    author = {Baptista, Ricardo and Kaveh, Hojjat and Stuart, Andrew M.},
    month = may,
    year = {2026},
    note = {arXiv:2605.15407 [math.NA]},
}

@misc{talts_validating_2020,
    title = {Validating {Bayesian} {Inference} {Algorithms} with {Simulation}-{Based} {Calibration}},
    url = {http://arxiv.org/abs/1804.06788},
    doi = {10.48550/arXiv.1804.06788},
    urldate = {2025-11-17},
    publisher = {arXiv},
    author = {Talts, Sean and Betancourt, Michael and Simpson, Daniel and Vehtari, Aki and Gelman, Andrew},
    month = oct,
    year = {2020},
    note = {arXiv:1804.06788 [stat]},
}

@inproceedings{talagrand_evaluation_1997,
    address = {Shinfield Park, Reading},
    title = {Evaluation of probabilistic prediction systems},
    language = {eng},
    publisher = {ECMWF},
    author = {Talagrand, O. and Vautard, R. and Strauss, B.},
    year = {1997},
    booktitle = {Workshop on Predictability},
}

@article{hamill_interpretation_2001,
    title = {Interpretation of {Rank} {Histograms} for {Verifying} {Ensemble} {Forecasts}},
    volume = {129},
    issn = {1520-0493, 0027-0644},
    url = {https://journals.ametsoc.org/view/journals/mwre/129/3/1520-0493_2001_129_0550_iorhfv_2.0.co_2.xml},
    doi = {10.1175/1520-0493(2001)129<0550:IORHFV>2.0.CO;2},
    language = {EN},
    number = {3},
    urldate = {2026-07-17},
    journal = {Monthly Weather Review},
    author = {Hamill, Thomas M.},
    month = mar,
    year = {2001},
    note = {Publisher: American Meteorological Society
Section: Monthly Weather Review},
    pages = {550--560},
}

@misc{calvello_operator_2026,
    title = {Operator {Learning} for {Smoothing} and {Forecasting}},
    url = {http://arxiv.org/abs/2603.20359},
    doi = {10.48550/arXiv.2603.20359},
    urldate = {2026-03-31},
    publisher = {arXiv},
    author = {Calvello, Edoardo and Carlson, Elizabeth and Kovachki, Nikola and Manta, Michael N. and Stuart, Andrew M.},
    month = mar,
    year = {2026},
    note = {arXiv:2603.20359 [stat]},
}

@misc{li_continuous_2026,
    title = {Continuous {Data} {Assimilation} with {Learned} {Surrogate} {Dynamics}},
    url = {http://arxiv.org/abs/2606.00480},
    doi = {10.48550/arXiv.2606.00480},
    urldate = {2026-07-21},
    publisher = {arXiv},
    author = {Li, Wenwen and Sanz-Alonso, Daniel},
    month = may,
    year = {2026},
    note = {arXiv:2606.00480 [math.DS]
version: 1},
}

@article{sun2024provable,
  title={Provable probabilistic imaging using score-based generative priors},
  author={Sun, Yu and Wu, Zihui and Chen, Yifan and Feng, Berthy T and Bouman, Katherine L},
  journal={IEEE Transactions on Computational Imaging},
  volume={10},
  pages={1290--1305},
  year={2024},
  publisher={IEEE}
}

@article{adrian2026auto,
  title={{Auto-differentiable data assimilation: Co-learning of states, dynamics, and filtering algorithms}},
  author={Adrian, Melissa and Sanz-Alonso, Daniel and Willett, Rebecca},
  journal={arXiv preprint arXiv:2603.20891},
  year={2026}
}

@inproceedings{chung2022diffusion,
  title={Diffusion posterior sampling for general noisy inverse problems},
  author={Chung, Hyungjin and Kim, Jeongsol and Mccann, Michael Thompson and Klasky, Marc Louis and Ye, Jong Chul},
  booktitle={The Eleventh International Conference on Learning Representations},
  year={2022}
}

@article{zhang2024flow,
  title={Flow priors for linear inverse problems via iterative corrupted trajectory matching},
  author={Zhang, Yasi and Yu, Peiyu and Zhu, Yaxuan and Chang, Yingshan and Gao, Feng and Wu, Ying Nian and Leong, Oscar},
  journal={Advances in Neural Information Processing Systems},
  volume={37},
  pages={57389--57417},
  year={2024}
}

@inproceedings{zheng2025inversebench,
  title={Inversebench: Benchmarking plug-and-play diffusion priors for inverse problems in physical sciences},
  author={Zheng, Hongkai and Chu, Wenda and Zhang, Bingliang and Wu, Zihui and Wang, Austin and Feng, Berthy and Zou, Caifeng and Sun, Yu and Kovachki, Nikola and Ross, Zachary and others},
  booktitle={International Conference on Learning Representations},
  volume={2025},
  pages={90912--90940},
  year={2025}
}

@inproceedings{zhang2025improving,
  title={Improving diffusion inverse problem solving with decoupled noise annealing},
  author={Zhang, Bingliang and Chu, Wenda and Berner, Julius and Meng, Chenlin and Anandkumar, Anima and Song, Yang},
  booktitle={Proceedings of the Computer Vision and Pattern Recognition Conference},
  pages={20895--20905},
  year={2025}
}

@article{wu2024principled,
  title={Principled probabilistic imaging using diffusion models as plug-and-play priors},
  author={Wu, Zihui and Sun, Yu and Chen, Yifan and Zhang, Bingliang and Yue, Yisong and Bouman, Katherine L},
  journal={Advances in Neural Information Processing Systems},
  volume={37},
  pages={118389--118427},
  year={2024}
}

@article{micchelli2006universal,
  title={Universal Kernels},
  author={Micchelli, Charles A and Xu, Yuesheng and Zhang, Haizhang},
  journal={Journal of Machine Learning Research},
  volume={7},
  number={12},
  year={2006}
}

@article{calvetti2018inverse,
  title={{Inverse problems: From regularization to Bayesian inference}},
  author={Calvetti, Daniela and Somersalo, Erkki},
  journal={Wiley Interdisciplinary Reviews: Computational Statistics},
  volume={10},
  number={3},
  pages={e1427},
  year={2018},
  publisher={Wiley Online Library}
}

@article{dunlop2017hierarchical,
  title={{Hierarchical Bayesian level set inversion}},
  author={Dunlop, Matthew M and Iglesias, Marco A and Stuart, Andrew M},
  journal={Statistics and Computing},
  volume={27},
  number={6},
  pages={1555--1584},
  year={2017},
  publisher={Springer}
}

@article{glaubitz2025efficient,
  title={Efficient sampling for sparse {Bayesian} learning using hierarchical prior normalization},
  author={Glaubitz, Jan and Marzouk, Youssef},
  journal={arXiv preprint arXiv:2505.23753},
  year={2025}
}

@article{sanz2025hierarchical,
  title={{Hierarchical Bayesian inverse problems: A high-dimensional statistics viewpoint}},
  author={Sanz-Alonso, Daniel and Waniorek, Nathan},
  journal={SIAM Review},
  volume={67},
  number={3},
  pages={543--575},
  year={2025},
  publisher={SIAM}
}

@article{waniorek2023bayesian,
  title={Bayesian hierarchical dictionary learning},
  author={Waniorek, N and Calvetti, D and Somersalo, E},
  journal={Inverse Problems},
  volume={39},
  number={2},
  year={2023}
}

@article{waghmare2025proper,
  title={Proper scoring rules for estimation and forecast evaluation},
  author={Waghmare, Kartik and Ziegel, Johanna},
  journal={arXiv preprint arXiv:2504.01781},
  year={2025}
}

@article{laio2007verification,
  title={Verification tools for probabilistic forecasts of continuous hydrological variables},
  author={Laio, Francesco and Tamea, Stefania},
  journal={Hydrology and Earth System Sciences},
  volume={11},
  number={4},
  pages={1267--1277},
  year={2007},
  publisher={Copernicus GmbH}
}

@article{fakoor2023flexible,
  title={Flexible model aggregation for quantile regression},
  author={Fakoor, Rasool and Kim, Taesup and Mueller, Jonas and Smola, Alexander J and Tibshirani, Ryan J},
  journal={Journal of Machine Learning Research},
  volume={24},
  number={162},
  pages={1--45},
  year={2023}
}

@article{bocquet_surrogate_2023,
  title = {Surrogate Modeling for the Climate Sciences Dynamics with Machine Learning and Data Assimilation},
  author = {Bocquet, Marc},
  year = {2023},
  journal = {Frontiers in Applied Mathematics and Statistics},
  volume = {9},

  doi = {10.3389/fams.2023.1133226},
  urldate = {2023-03-06}
}

@article{cheng_machine_2023,
  title = {Machine learning with data assimilation and uncertainty quantification for dynamical systems: a review},
  shorttitle = {Machine learning with data assimilation and uncertainty quantification for synamical systems},
  author = {Cheng, Sibo and {Quilodr{\'a}n-Casas}, C{\'e}sar and Ouala, Said and Farchi, Alban and Liu, Che and Tandeo, Pierre and Fablet, Ronan and Lucor, Didier and Iooss, Bertrand and Brajard, Julien and Xiao, Dunhui and Janjic, Tijana and Ding, Weiping and Guo, Yike and Carrassi, Alberto and Bocquet, Marc and Arcucci, Rossella},
  year = {2023},

  journal = {IEEE/CAA Journal of Automatica Sinica},
  volume = {10},
  number = {6},
  pages = {1361--1387},
  doi = {10.1109/JAS.2023.123537}
}

@article{grooms2022comparison,
  title={{A comparison of nonlinear extensions to the ensemble Kalman filter: Gaussian anamorphosis and two-step ensemble filters}},
  author={Grooms, Ian},
  journal={Computational Geosciences},
  volume={26},
  number={3},
  pages={633--650},
  year={2022},
  publisher={Springer}
}

@book{lange2016mm,
  title={MM Optimization Algorithms},
  author={Lange, Kenneth},
  year={2016},
  publisher={SIAM}
}

@inproceedings{oliva2018transformation,
  title={Transformation autoregressive networks},
  author={Oliva, Junier and Dubey, Avinava and Zaheer, Manzil and Poczos, Barnabas and Salakhutdinov, Ruslan and Xing, Eric and Schneider, Jeff},
  booktitle={International Conference on Machine Learning},
  pages={3898--3907},
  year={2018},
  organization={PMLR}
}

@article{anderson2010non,
  title={{A non-Gaussian ensemble filter update for data assimilation}},
  author={Anderson, Jeffrey L},
  journal={Monthly Weather Review},
  doi={10.1175/2010MWR3253.1},
  volume={138},
  number={11},
  pages={4186--4198},
  year={2010},
  publisher={American Meteorological Society}
}

@article{chen2018neural,
  title={Neural ordinary differential equations},
  author={Chen, Ricky TQ and Rubanova, Yulia and Bettencourt, Jesse and Duvenaud, David K},
  journal={Advances in Neural Information Processing Systems},
  volume={31},
  year={2018}
}

@article{lopez2022training,
  title={{Training physics-based machine-learning parameterizations with gradient-free ensemble Kalman methods}},
  author={Lopez-Gomez, Ignacio and Christopoulos, Costa and Langeland Ervik, Haakon Ludvig and Dunbar, Oliver RA and Cohen, Yair and Schneider, Tapio},
  journal={Journal of Advances in Modeling Earth Systems},
  volume={14},
  number={8},
  pages={e2022MS003105},
  year={2022},
  publisher={Wiley Online Library}
}

@book{hale,
  title={Asymptotic Behavior of Dissipative Systems},
  author={Hale, Jack K},
  series={Mathematical Surveys and Monographs},
  number={25},
  year={2010},
  publisher={American Mathematical Society}
}

@book{ambrosio2005gradient,
  title={Gradient Flows: In Metric Spaces and in the Space of Probability Measures},
  author={Ambrosio, Luigi and Gigli, Nicola and Savar{\'e}, Giuseppe},
  year={2005},
  publisher={Springer Science \& Business Media},
  doi={10.1007/978-3-7643-8722-8}
}

@book{friesecke2024optimal,
  title={Optimal Transport: A Comprehensive Introduction to Modeling, Analysis, Simulation, Applications},
  author={Friesecke, Gero},
  year={2024},
  publisher={SIAM}
}

@article{alexander2020operator,
  title={Operator-theoretic framework for forecasting nonlinear time series with kernel analog techniques},
  author={Alexander, Romeo and Giannakis, Dimitrios},
  journal={Physica D: Nonlinear Phenomena},
  doi={10.1016/j.physd.2020.132520},
  volume={409},
  pages={132520},
  year={2020},
  publisher={Elsevier}
}

@article{giannakis2019data,
  title={Data-driven spectral decomposition and forecasting of ergodic dynamical systems},
  author={Giannakis, Dimitrios},
  journal={Applied and Computational Harmonic Analysis},
  volume={47},
  number={2},
  pages={338--396},
  year={2019},
  publisher={Elsevier}
}

@article{giannakis2023learning,
  title={Learning to forecast dynamical systems from streaming data},
  author={Giannakis, Dimitrios and Henriksen, Amelia and Tropp, Joel A and Ward, Rachel},
  journal={SIAM Journal on Applied Dynamical Systems},
  volume={22},
  number={2},
  pages={527--558},
  year={2023},
  publisher={SIAM}
}

@article{sahu1999convergence,
  title={{On convergence of the EM algorithmand the Gibbs sampler}},
  author={Sahu, Sujit K and Roberts, Gareth O},
  journal={Statistics and Computing},
  volume={9},
  number={1},
  pages={55--64},
  year={1999},
  publisher={Springer}
}

@article{meng1997algorithm,
  title={{The EM algorithm—an old folk-song sung to a fast new tune}},
  author={Meng, Xiao-Li and Van Dyk, David},
  journal={Journal of the Royal Statistical Society Series B: Statistical Methodology},
  volume={59},
  number={3},
  pages={511--567},
  year={1997},
  publisher={Oxford University Press}
}

@article{bonavita2020machine,
  title={Machine learning for model error inference and correction},
  author={Bonavita, Massimo and Laloyaux, Patrick},
  journal={Journal of Advances in Modeling Earth Systems},
  doi={10.1029/2020MS002232},
  volume={12},
  number={12},
  pages={e2020MS002232},
  year={2020},
  publisher={Wiley Online Library}
}

@article{bolton2019applications,
  title={Applications of deep learning to ocean data inference and subgrid parameterization},
  author={Bolton, Thomas and Zanna, Laure},
  journal={Journal of Advances in Modeling Earth Systems},
  doi={10.1029/2018MS001472},
  volume={11},
  number={1},
  pages={376--399},
  year={2019},
  publisher={Wiley Online Library}
}

@book{santambrogio2015optimal,
  title={Optimal Transport for Applied Mathematicians},
  author={Santambrogio, Filippo},
  volume={87},
  year={2015},
  publisher={Springer}
}

@article{bonavita2023limitations,
  title = {On Some Limitations of Current Machine Learning Weather Prediction Models},
  author = {Bonavita, Massimo},
  year = {2024},
  journal = {Geophysical Research Letters},
  volume = {51},
  number = {12},
  pages = {e2023GL107377},
  doi = {10.1029/2023GL107377},
  urldate = {2024-06-15},
  langid = {english}
}

@inproceedings{finn2017model,
  title={Model-agnostic meta-learning for fast adaptation of deep networks},
  author={Finn, Chelsea and Abbeel, Pieter and Levine, Sergey},
  booktitle={International conference on machine learning},
  pages={1126--1135},
  year={2017},
  organization={PMLR}
}

@article{andrychowicz2016learning,
  title={Learning to learn by gradient descent by gradient descent},
  author={Andrychowicz, Marcin and Denil, Misha and Gomez, Sergio and Hoffman, Matthew W and Pfau, David and Schaul, Tom and Shillingford, Brendan and De Freitas, Nando},
  journal={Advances in neural information processing systems},
  volume={29},
  year={2016}
}

@article{dayan1995helmholtz,
  title={The helmholtz machine},
  author={Dayan, Peter and Hinton, Geoffrey E and Neal, Radford M and Zemel, Richard S},
  journal={Neural computation},
  volume={7},
  number={5},
  pages={889--904},
  year={1995},
  publisher={MIT Press One Rogers Street, Cambridge, MA 02142-1209, USA journals-info~…}
}

@article{rasp2018deep,
  title={Deep learning to represent subgrid processes in climate models},
  author={Rasp, Stephan and Pritchard, Michael S and Gentine, Pierre},
  journal={Proceedings of the National Academy of Sciences},
  volume={115},
  number={39},
  pages={9684--9689},
  year={2018},
  publisher={National Acad Sciences}
}

@book{smola1998learning,
  title={Learning with Kernels},
  author={Smola, Alex J and Sch{\"o}lkopf, Bernhard},
  volume={4},
  year={1998},
  publisher={Citeseer}
}

@inproceedings{genevay2018learning,
  title={{Learning generative models with Sinkhorn divergences}},
  author={Genevay, Aude and Peyr{\'e}, Gabriel and Cuturi, Marco},
  booktitle={International Conference on Artificial Intelligence and Statistics},
  pages={1608--1617},
  year={2018},
  organization={PMLR}
}

@article{gretton2012kernel,
  title={A kernel two-sample test},
  author={Gretton, Arthur and Borgwardt, Karsten M and Rasch, Malte J and Sch{\"o}lkopf, Bernhard and Smola, Alexander},
  journal={The Journal of Machine Learning Research},
  volume={13},
  number={1},
  pages={723--773},
  year={2012},
  publisher={JMLR. org}
}

@article{muandet2017kernel,
  title={{Kernel mean embedding of distributions: A review and beyond}},
  author={Muandet, Krikamol and Fukumizu, Kenji and Sriperumbudur, Bharath and Sch{\"o}lkopf, Bernhard and others},
  journal={Foundations and Trends{\textregistered} in Machine Learning},
  volume={10},
  number={1-2},
  pages={1--141},
  year={2017},
  publisher={Now Publishers, Inc.}
}

@inproceedings{gretton2005measuring,
  title={{Measuring statistical dependence with Hilbert-Schmidt norms}},
  author={Gretton, Arthur and Bousquet, Olivier and Smola, Alex and Sch{\"o}lkopf, Bernhard},
  booktitle={International Conference on Algorithmic Learning Theory},
  pages={63--77},
  year={2005},
  organization={Springer}
}

\newpage
\printindex
\end{document}